\documentclass[
11pt, %
english, %
singlespacing, %
headsepline, %
]{MastersDoctoralThesis} %

\usepackage[utf8]{inputenc} %
\usepackage[T1]{fontenc} %

\usepackage{mathpazo} %

\usepackage[backend=bibtex,style=authoryear,natbib=true]{biblatex} %

\usepackage[autostyle=true]{csquotes} %

\newcommand{\printpublication}[1]{%
  \begin{refsection}
    \AtNextCite{\defcounter{maxnames}{99}}%
    \fullcite{#1}%
  \end{refsection}
}

\usepackage{amsmath}
\usepackage{amsbsy}
\usepackage{bm}
\usepackage{tensor}

\DeclareMathAlphabet{\mathsfit}{\encodingdefault}{\sfdefault}{m}{sl}
\SetMathAlphabet{\mathsfit}{bold}{\encodingdefault}{\sfdefault}{bx}{n}
\newcommand{\tens}[1]{\bm{\mathsfit{#1}}}

\usepackage{booktabs}
\usepackage{forest}

\usepackage{url}

\usepackage{outlines}
\usepackage{multirow}
\usepackage{float} 
\usepackage[normalem]{ulem}
\usepackage{subcaption}
\usepackage{multirow}
\usepackage{booktabs}
\usepackage{tabularx}
\usepackage{multirow}
\usepackage{makecell}
\usepackage{framed,enumitem} 
\usepackage{amsmath,amssymb}
\usepackage{lscape} %

\usepackage{makecell} %
\usepackage{pifont}%
\newcommand{\cmark}{\ding{51}}%
\newcommand{\xmark}{\ding{55}}%

\usepackage{algorithm}%
\usepackage{algorithmicx}%
\usepackage{algpseudocode}%
\usepackage{graphicx}
\usepackage{amssymb} %
\usepackage{bbding}
\usepackage{wrapfig}
\usepackage{adjustbox}

\usepackage{array}
\usepackage{rotating}

\newcommand{\RNum}[1]{\uppercase\expandafter{\romannumeral #1\relax}}

\newcolumntype{C}[1]{>{\centering\arraybackslash}m{#1}}
\newcolumntype{R}[1]{>{\raggedleft\arraybackslash}m{#1}}
\newcolumntype{L}[1]{>{\raggedright\arraybackslash}m{#1}}

\newcommand{\bftab}{\fontseries{b}\selectfont}

\usepackage{listings}

\usepackage{verbatim}

\usepackage[final,babel]{microtype}

\thesistitle{Learning Tree-Based Models with Gradient Descent} %
\supervisor{Prof. Dr. Heiner \textsc{Stuckenschmidt}} %
\examiner{Prof. Dr. Stefan \textsc{L\"udtke}} %
\degree{Doctor of Science} %
\author{Sascha \textsc{Marton}} %
\addresses{} %

\subject{Computer Science} %
\university{\href{https://www.uni-mannheim.de/}{University of Mannheim}} %
\department{\href{https://www.uni-mannheim.de/ines}{Institute for Enterprise Systems}} %
\group{\href{https://www.uni-mannheim.de/dws/}{Data and Web Science Group}} %
\faculty{\href{http://faculty.university.com}{Faculty Name}} %

\AtBeginDocument{
\hypersetup{pdftitle=\ttitle} %
\hypersetup{pdfauthor=\authorname} %
\hypersetup{pdfkeywords=\keywordnames} %
}

\begin{document}

\frontmatter %

\pagestyle{plain} %

\begin{titlepage}
\begin{center}

\includegraphics[width=0.5\textwidth]{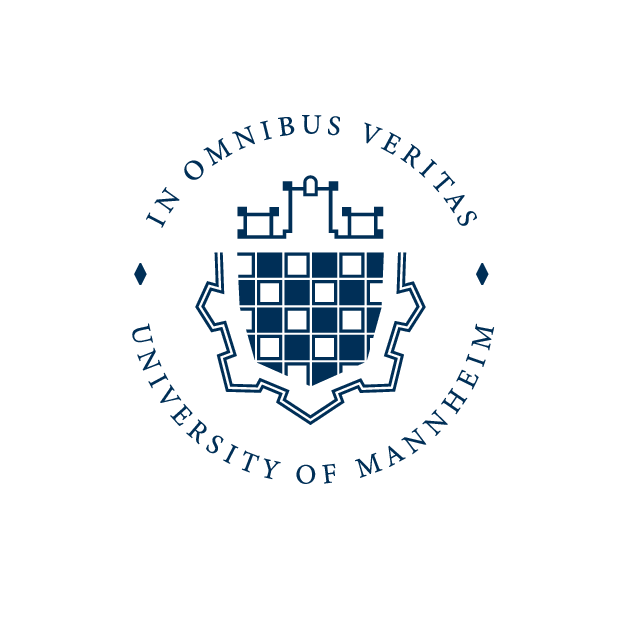} \\
{\scshape\LARGE \univname\par}\vspace{1.5cm} %
\textsc{\Large Doctoral Thesis}\\[0.5cm] %

\HRule \\[0.4cm] %
{\huge \bfseries \ttitle\par}\vspace{0.4cm} %
\HRule \\[1.5cm] %
 
\begin{minipage}[t]{0.4\textwidth}
\begin{flushleft} \large
\emph{Author:} \\ \authorname
\end{flushleft}
\end{minipage}
\begin{minipage}[t]{0.5\textwidth}
\begin{flushright} \large
\emph{Supervisor:} \\ \supname
\end{flushright}
\end{minipage}\\[3cm]
 
\vfill

\large \textit{A thesis submitted in fulfillment of the\\requirements for the degree of \degreename\\at the \univname}\\[2cm] %
 
\vfill

{\large March 3, 2025}\\[4cm] %
 
\vfill
\end{center}

\newpage

\vspace*{\fill} 

\begin{center}
    \begin{tabular}{ll}
        \textit{Examiners:} & Prof. Dr. Heiner Stuckenschmidt, University of Mannheim \\
        & Prof. Dr. Christian Bartelt, Technical University of Clausthal \\
        & Jun.-Prof. Dr. Stefan L\"udtke, University of Rostock
    \end{tabular}
\end{center}

\vspace{2cm}

\end{titlepage}

\begin{abstract}
\addchaptertocentry{\abstractname} %
    \sloppy
    Tree-based models are widely recognized for their interpretability and have proven effective in various application domains, particularly in high-stakes domains. In addition, their hierarchical structure and axis-aligned splits can provide a beneficial inductive bias, especially for heterogeneous tabular data. However, learning decision trees poses a significant challenge due to their combinatorial complexity and discrete, non-differentiable nature. As a result, traditional methods such as CART, which rely on greedy search procedures, remain the most widely used approaches. These methods make locally optimal decisions at each node, constraining the search space and often leading to suboptimal tree structures. Additionally, their demand for custom training methods precludes a seamless integration into modern machine learning approaches, such as those employed in multimodal or reinforcement learning, where optimization is typically achieved via gradient descent and backpropagation. 

    In this thesis, we propose a novel method for learning hard, axis-aligned decision trees through gradient descent. Our approach utilizes backpropagation with a straight-through operator on a dense decision tree representation, enabling the joint optimization of all tree parameters. We further extend this method to tree ensembles including a novel, instance-wise weighting scheme, allowing for a trade-off between performance and interpretability. 
    By introducing gradient-based optimization for hard, axis-aligned decision trees, our method addresses the two primary limitations of traditional decision tree algorithms. First, gradient-based training is not constrained by the sequential selection of locally optimal splits but, instead, jointly optimizes all tree parameters. Second, by leveraging gradient descent for optimization, our approach seamlessly integrates into existing machine learning approaches e.g., for multimodal and reinforcement learning tasks, which inherently rely on gradient descent.
    These advancements allow us to achieve state-of-the-art results across multiple domains, including interpretable decision trees for small tabular datasets, advanced models for complex tabular data, multimodal learning, and interpretable reinforcement learning without information loss. By bridging the gap between decision trees and gradient-based optimization, our method significantly enhances the performance
    and applicability of tree-based models across various machine learning domains.
    
\end{abstract}

\begin{acknowledgements}
\addchaptertocentry{\acknowledgementname} %
I would like to sincerely thank my supervisor, Christian Bartelt, for his unwavering support and dedication. Christian spent countless hours discussing ideas with me, offering insightful feedback, and guiding me through every stage of this dissertation. His advice was instrumental, not only in shaping the direction of my research but also in navigating the broader challenges of the PhD process.

I am also deeply grateful to Stefan L\"udtke, who joined as a co-supervisor during my third year when I was feeling uncertain about the direction of my work. Stefan’s guidance helped me refocus, and his detailed feedback during our many meetings was invaluable. He was always available for questions, providing both practical advice and critical insights that significantly improved the quality of my work.

A special thanks goes to Heiner Stuckenschmidt, who first brought me to the Institute for Enterprise Systems. Heiner was an honest, thoughtful and caring supervisor, always available for discussions and offering balanced, helpful feedback on both my work and important decisions related to my research. His mentorship from the start laid a strong foundation for this dissertation.

I would also like to acknowledge my great colleagues and friends — Patrick Knab, Tim Grams, Andrej Tschalzev, Jannik Brinkmann, Tobi Sesterhenn, Janis Zenkner, Alina Klerings, Lars Hoffmann, Patrick Betz and many others — whose discussions and feedback were vital. Whether offering technical insights or simply sharing their perspectives on my work, their contributions helped me refine my ideas and stay motivated throughout the process.

I extend my thanks to everyone who, in one way or another, supported me throughout this journey. Their encouragement and assistance have made a significant impact on the completion of this dissertation.

Lastly, I want to express my deepest gratitude to Laura for her constant support, love, and patience throughout this journey. Her understanding during long working hours, her encouragement, and optimism in moments of doubt, and her belief in me were a constant source of strength. I could not have made it through this process without her by my side.

\end{acknowledgements}

\tableofcontents %

\listoffigures %

\listoftables %

\dedicatory{To Laura, who has always been there for me and has supported me in all my decisions, especially in my choice to pursue a PhD.}

\mainmatter %

\pagestyle{thesis} %

\def\chapterautorefname{Chapter}%

\chapter{Introduction} %
\label{cha:introduction} %

\section{Motivation}

\begin{figure}[th]
    \centering
    \includegraphics[width=0.6\textwidth]{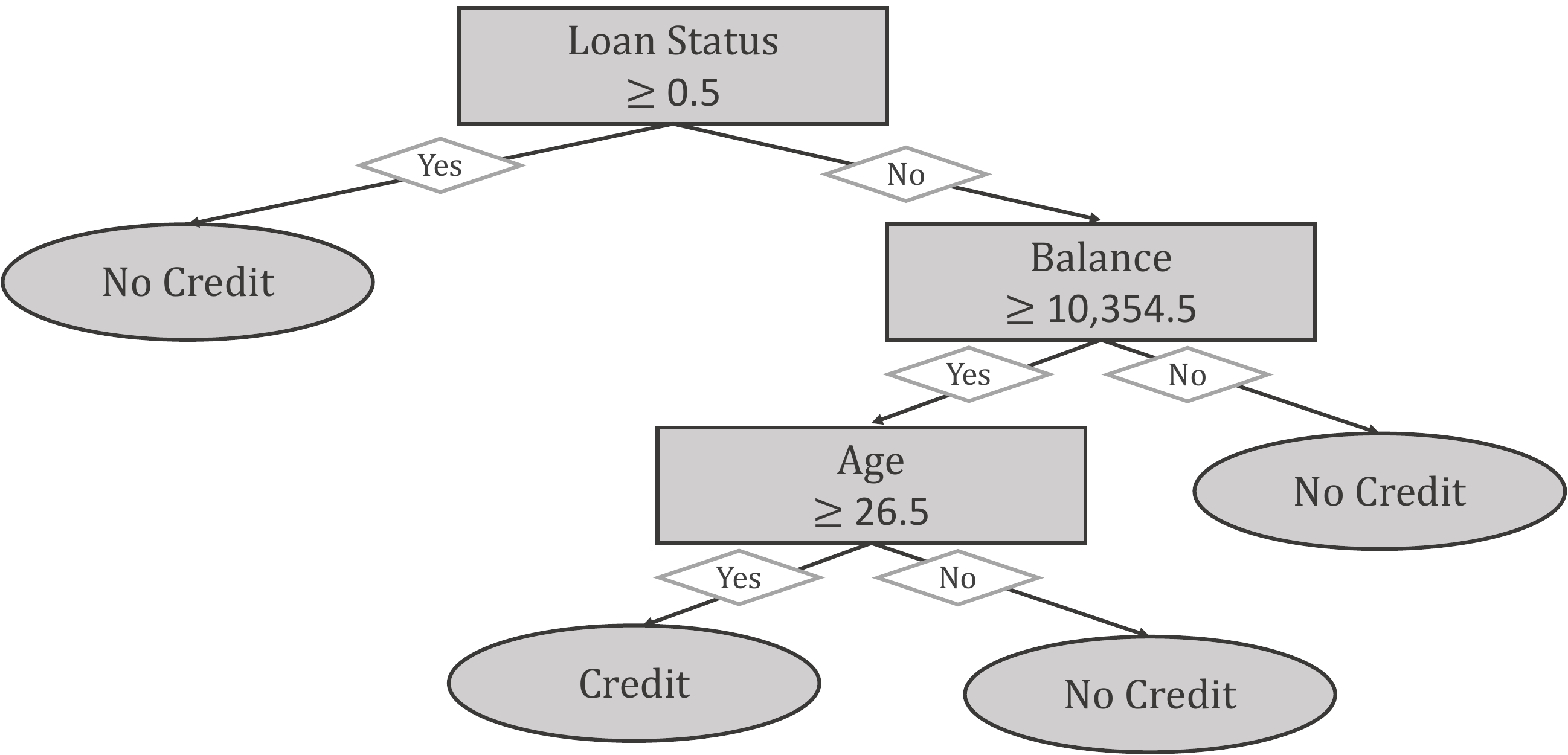}
    \caption[Exemplary DT]{\textbf{Exemplary DT.} This is an example of a simple DT trained on a simulated dataset to predict whether a bank should approve a credit application. The decision process begins by checking if the \emph{Loan Status} is greater than or equal to $0.5$, indicating whether the customer already holds a personal loan with the bank. If the customer already has a loan ($1$), the bank should not grant additional credit.
    If the customer does not hold a loan ($0$), the model proceeds to assess the customer's \emph{Balance}, comparing it to a threshold of $10{,}354.5$. If the balance is below the threshold, the credit application is denied. For higher balances, the model further evaluates the applicant’s \emph{Age}: If the customer is older than 26.5 years, the credit is approved and otherwise, it is denied.
    }
    \label{fig:dt_example_credit}
\end{figure}

\paragraph{Everybody loves decision trees for their interpretability} 
Decision trees (DTs) are a fundamental class of machine learning models that partition data by applying a series of if-then conditions at each node, ultimately arriving at leaf nodes where a final prediction is made~\citep{Tan2018}. This branching structure allows for a clear and intuitive mapping of how inputs lead to outputs, making DTs inherently transparent. 
An exemplary DT is visualized in Figure~\ref{fig:dt_example_credit} illustrating the strong explanatory power of DTs, providing clear and transparent reasoning by breaking down decisions into intuitive, rule-based steps.
Consequently, tree-based models are indispensable across a wide range of areas and have demonstrated their utility in numerous application scenarios, particularly in high-stakes domains~\citep{med1,med2,med3,credit1,fraud1,multimodal2}. A key reason for their widespread adoption is their inherent interpretability. Because of their symbolic nature, tree-based architectures are often considered to be more transparent and easier to verify than other machine learning models~\citep{rudin2019stop}. Unlike complex neural networks (NNs), which function as opaque black boxes, DTs provide a clear pathway to trace how each decision is reached, allowing stakeholders to build trust in the system. This interpretability makes them particularly suitable for applications where the ability to critically evaluate and understand model decisions is crucial. Their clear structure can be easily visualized (Figure~\ref{fig:dt_example_credit}), which aids in both the comprehension and validation processes, thereby making the decision-making process more transparent and explainable~\citep{molnar2020}. 

\begin{figure}[th]
\centering
\begin{subfigure}{0.5\textwidth}
  \centering
  \includegraphics[width=0.95\columnwidth]{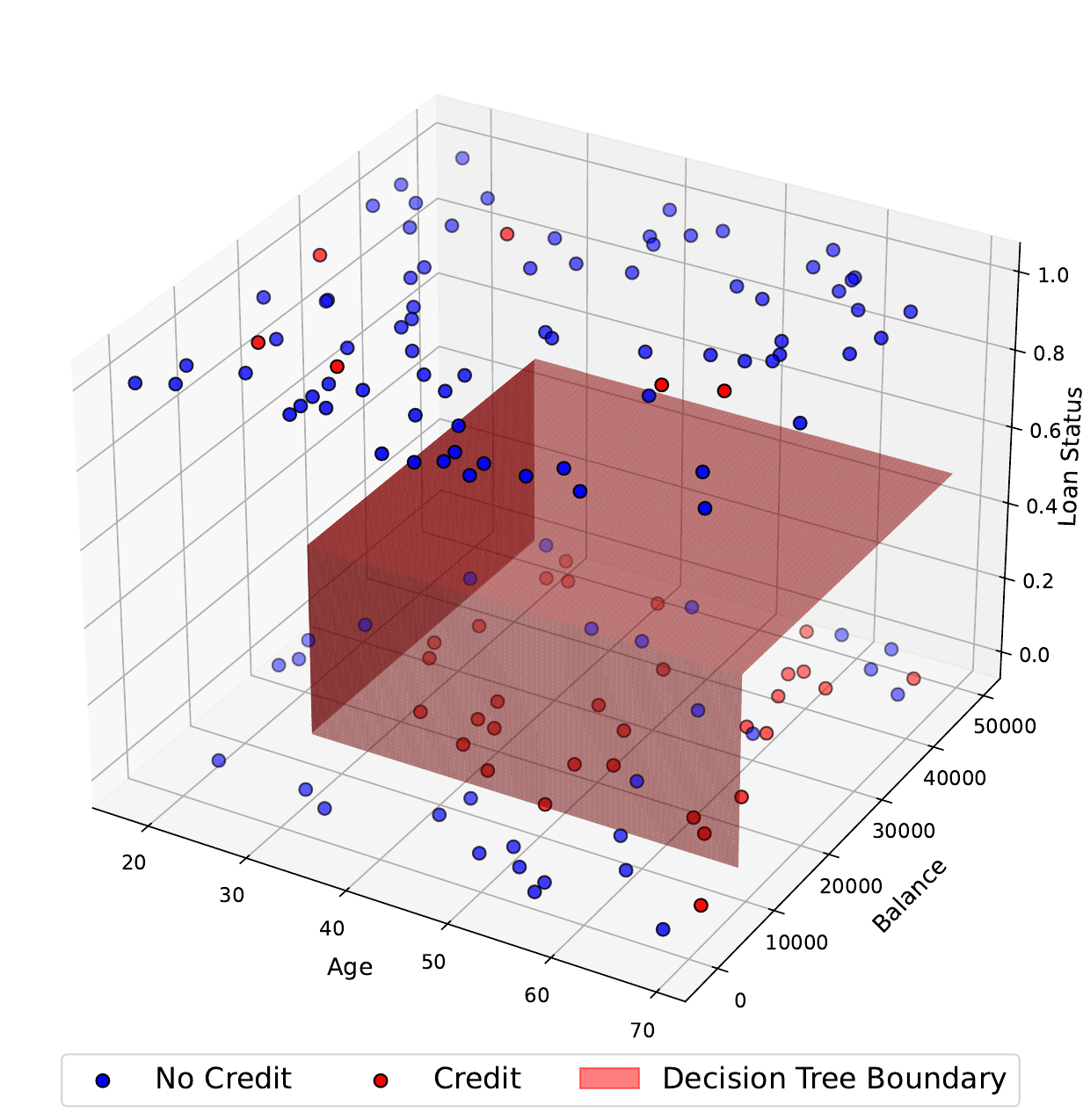}
  \caption{Decision Tree Decision Boundary}
  \label{fig:boundary_example_DT}
  
\end{subfigure}%
\begin{subfigure}{0.5\textwidth}
  \centering
  \includegraphics[width=0.95\columnwidth]{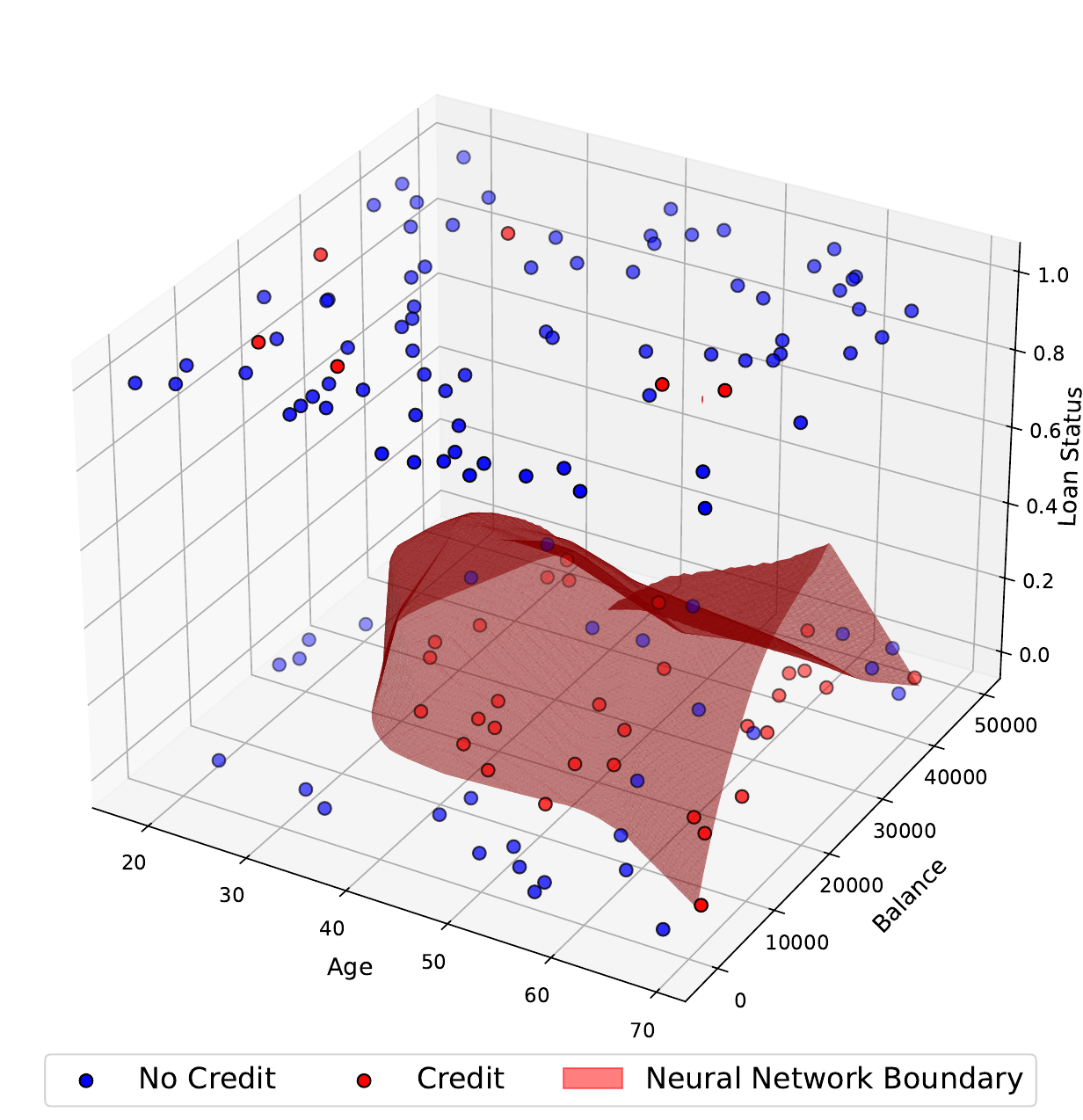}
  \caption{Neural Network Decision Boundary}
  \label{fig:boundary_example_NN}
\end{subfigure}
\caption[Decision Boundary Comparison]{\textbf{Decision Boundary Comparison.} This figure illustrates the decision boundaries, i.e., the hypersurface that separates different classes in the input space based on the learned function, of a DT (A) and an NN (B) on a simulated credit dataset. The decision boundary of the DT in Figure~\ref{fig:dt_example_credit} clearly partitions the data along axis-aligned splits. In contrast, the NN exhibits a more flexible decision boundary, which lacks a similarly interpretable structure. Notably, the DT effectively captures the underlying patterns in the data, such as the distinct separation between Loan Status values of 0 and 1. The NN, however, introduces artificial variance by treating values between 0 and 1 differently, even though such values are not present in the dataset. This highlights the advantageous inductive bias of axis-aligned splits, which efficiently capture the relevant structure of the Loan Status variable.}
\label{fig:boundary_example}
\end{figure}

\paragraph{Decision trees can provide a good inductive bias}
Inductive bias refers to the set of assumptions that machine learning algorithms use to generalize from observed data to previously unencountered instances. These assumptions can prioritize certain solutions over others, independent of the observed data. In the context of DTs, their hierarchical structure and discrete decisions as inductive biases facilitate a natural partitioning of data into meaningful subsets, enabling efficient handling of diverse datasets~\citep{breiman1984classification,breiman2001randomforest}. 
Furthermore, the axis-aligned splits, a characteristic feature of standard DTs, introduce a beneficial inductive bias by creating decision boundaries parallel to the feature axes while considering only a single feature at a time (e.g., dividing data based on whether \emph{Balance} is above or below a threshold). In doing so, this natural structuring of decision boundaries along feature dimensions favors models that align well with the feature space and effectively capture relevant patterns in heterogeneous tabular data~\citep{grinsztajn2022tree}.
This also aligns well with many real-world datasets, where features often have direct, independent relationships to the target variable (like the loan status), allowing the tree to capture relevant patterns effectively, as shown in Figure~\ref{fig:boundary_example}.
The DT in Figure~\ref{fig:boundary_example_DT} clearly demonstrates how restricting each decision to a single feature axis (in this case, first \emph{Loan Status} followed by \emph{Balance} and then \emph{Age}) yields an intuitive, stepwise partition of the data. 
Furthermore, by focusing on one variable at a time and splitting it at a simple threshold, the model avoids overly complex decision boundaries which can favor generalization as well as interpretability. 
The simplicity of these splits contributes to robustness, as DTs are less likely to overfit on noisy or non-informative features, especially when regularized or combined in ensembles~\citep{breiman2001randomforest}.

\paragraph{Tree-based models dominate tabular data but struggle elsewhere}
Tree-based models have demonstrated particular success in the tabular domain, which remains their most prominent application area. Notably, despite the widespread success of deep learning in other domains, tabular data is the last \emph{unconquered castle} for traditional deep learning models~\citep{kadra2021well}. DT ensembles, especially gradient-boosted DT~\citep{friedman2001gradientboostingtree} libraries like XGBoost~\citep{chen2016xgboost} and CatBoost~\citep{prokhorenkova2018catboost} are still considered as state-of-the-art~\citep{grinsztajn2022tree,mcelfresh2023neural_whitepaper,tschalzev2024data}. The structured nature of tabular data, coupled with the advantageous inductive bias of axis-aligned splits, contributes to their ongoing success in this domain~\citep{grinsztajn2022tree}.
However, tree-based models face significant limitations in other domains such as reinforcement learning (RL), time series analysis, and complex feature spaces like images and text. These limitations stem primarily from their fundamental structure and learning process, which leads us to the key challenges that need to be addressed.

\section{Problem Description}

Despite their wide use in machine learning, conventional methods to learning DTs suffer from two fundamental problems that hinder their effectiveness and broader applicability. These limitations not only affect their performance in various tasks but also complicate their integration into modern machine learning frameworks. This thesis specifically addresses the following issues:

\paragraph{(1) Common methods for learning tree-based models are prone to locally optimal solutions}
Learning DTs is a challenging optimization problem due to the combinatorial complexity of their branching structure and the discrete nature of their decisions.
To address these challenges, the most prevalent approach for learning DTs has been through greedy, recursive partitioning algorithms such as CART and C4.5. These methods sequentially split the data at each internal node by minimizing a local impurity measure, typically Gini impurity or entropy for classification tasks, and variance for regression tasks. While computationally efficient, these approaches often lead to suboptimal tree structures, resulting in models that may have poor performance, fail to generalize well to unseen data, or become overly large~\citep{zharmagambetov2021non,vos2023optimal,bertsimas2017optimal}.
The limitations of current approaches have motivated exploration into alternative, non-greedy optimization techniques including evolutionary algorithms and advanced methods for exhaustive search procedures. \\ %

\begin{figure}[tb]
\centering
\begin{subfigure}{0.5\textwidth}
  \centering
  \includegraphics[width=0.875\columnwidth]{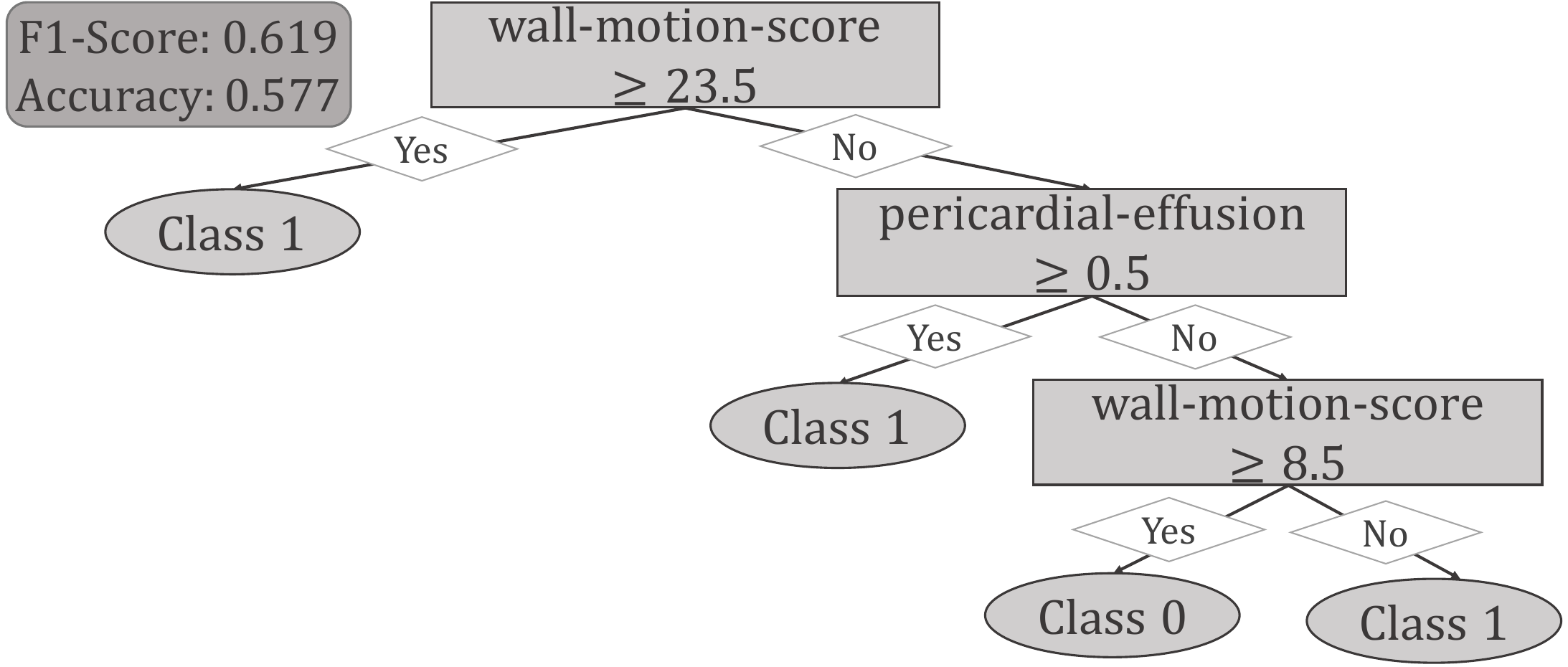}
  \caption{Greedy DT (CART)}
  \label{fig:cart_dt_example}
  
\end{subfigure}%
\begin{subfigure}{0.5\textwidth}
  \centering
  \includegraphics[width=0.875\columnwidth]{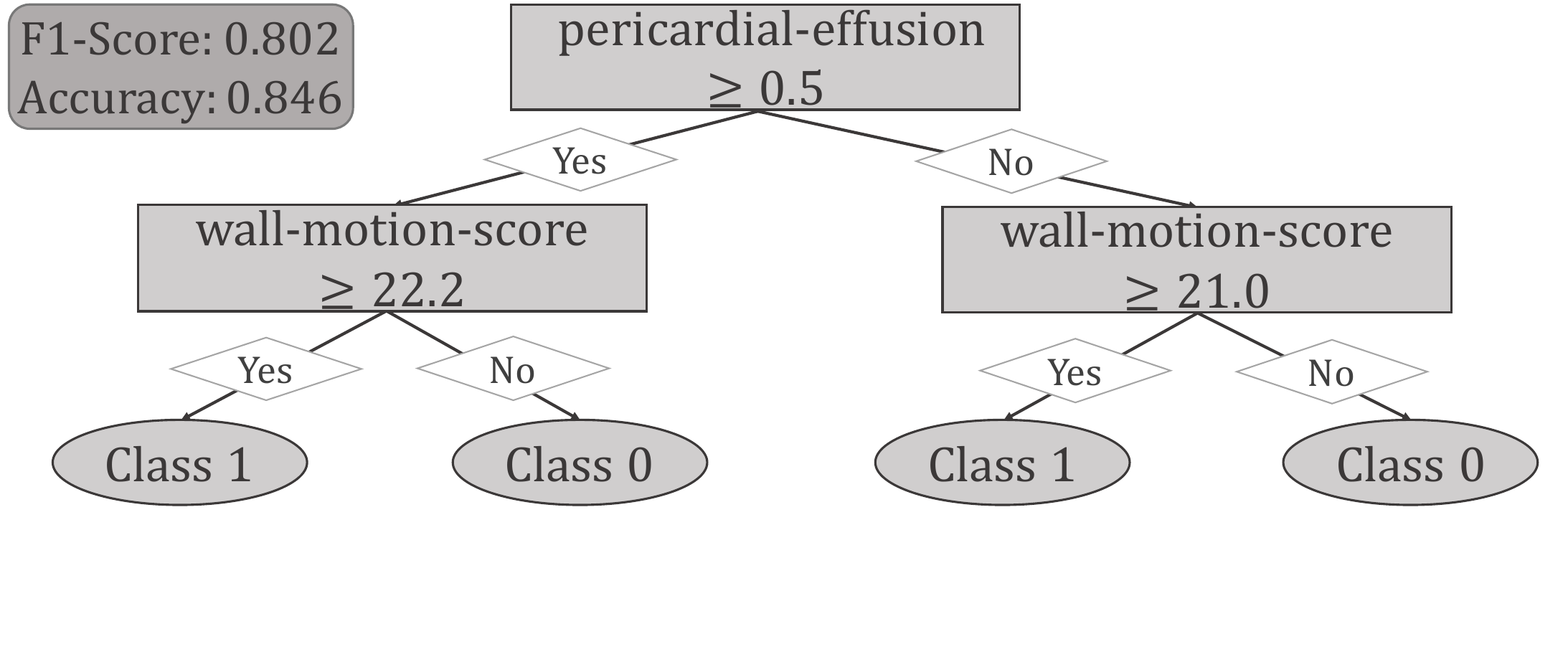}
  \caption{Non-Greedy DT}
  \label{fig:gdt_example}
\end{subfigure}
\caption[Greedy vs. Non-Greedy DT]{\textbf{Greedy vs. Gradient-Based DT.} Two DTs trained on the Echocardiogram dataset. 
The CART DT (A) sequentially performs only locally optimal splits, thereby constraining the search space, whereas the non-greedy algorithm (B) explores the entire search space, enabling the discovery of a solution that achieves significantly better performance.
}
\label{fig:dt_comparison_example}
\end{figure}

\noindent \emph{Example} \hspace{0.25cm}
The Echocardiogram dataset~\citep{ml_repository} deals with predicting one-year survival of patients after a heart attack based on tabular data from an echocardiogram. Figure~\ref{fig:dt_comparison_example} shows two DTs.
The tree on the left is learned by a greedy algorithm (CART) while the one on the right is learned with a non-greedy approach.
We can observe that the greedy procedure leads to a tree with a significantly lower performance. Splitting on the \emph{wall-motion-score} is the locally optimal split (see Figure~\ref{fig:cart_dt_example}), but globally, it is beneficial to split based on the \emph{wall-motion-score} with different values conditioned on the \emph{pericardial-effusion} in the second level (Figure~\ref{fig:gdt_example}). 

\paragraph{(2) Gradient-based optimization is required for many modern ML approaches}
Many modern machine learning approaches, particularly in reinforcement learning and multimodal learning frameworks, rely heavily on gradient-based optimization techniques such as stochastic gradient descent. These methods require that both the model and its objective function be differentiable so that gradients can be computed and used to iteratively minimize loss functions. %
While tree-based models excel on structured data due to their simplicity and axis-aligned splits, they often struggle in other domains because they are not optimized using gradient descent. Their inherent non-differentiability makes it challenging to integrate them with methods that rely on gradient computations, such as backpropagation and end-to-end learning frameworks~\citep{lecun2015deep,goodfellow2016deep}. This static nature also limits their adaptability in dynamic environments like those encountered in RL~\citep{silver2014deterministic}.
In the context of RL, the necessity of gradients becomes particularly evident. Many state-of-the-art RL algorithms, such as policy gradient methods~\citep{reinforce,schulman2017proximal}, rely on the computation of gradients to update the policy parameters in a way that maximizes the expected cumulative reward. They calculate the gradient of the expected return with respect to the policy parameters, using this information to adjust the policy in the direction that yields improved performance. This process of incremental adjustment based on gradient information is essential for enabling an agent to learn and adapt to complex, changing environments. Without the ability to compute and propagate gradients through the model, as is the case with traditional DTs, incorporating such models directly into RL frameworks becomes highly challenging.
Given the success of gradient-based optimization, there is growing interest in developing differentiable or gradient-based variants of DTs and tree-based models~\citep{grinsztajn2022tree}, enabling their integration into modern end-to-end differentiable methods and allowing practitioners to apply gradient-based techniques directly to tree models. 
This would allow us to avoid creating custom solutions and optimization algorithms tailored to specific problem settings, instead leveraging a general-purpose DT model that can be applied across different scenarios, much like how neural networks are currently utilized. \\

\noindent To tackle these two key problems, in this thesis, we propose a gradient-based approach to learning DTs. The ability to use gradient descent to learn a DT from scratch would directly address the previously mentioned challenges: (1) A gradient-based optimization enables a global, joint optimization of all tree parameters, overcoming the constrained search space, which leads to local optimality issues inherent in the traditional greedy, sequential splitting procedure (see Figure~\ref{fig:dt_comparison_example}). Similar to neural network optimization, a gradient-based approach could converge to a local optimum that generalizes well, providing a significant advantage over alternative non-greedy methods like optimal DTs~\citep{murtree,aglin2020learning}, which frequently suffer from severe computational constraints and overfitting~\citep{zharmagambetov2021non}. (2) A DT learned via gradient descent would seamlessly integrate into existing gradient-based frameworks, such as those used for multimodal learning or RL. 
However, maintaining the inductive bias of hard, axis-aligned splits while optimizing the DT through gradient descent remains a highly non-trivial challenge that, to date, has not been effectively solved. This difficulty primarily arises from the following challenges:

\begin{itemize}
    \item \textbf{Learning decision trees is a difficult optimization problem} \hspace{0.25cm}
        Unlike continuous models where gradient-based optimization can be applied, DTs require a combinatorial search to find the optimal split points for nodes, which significantly increases the computational complexity~\citep{bengio2021machine}. This search often involves evaluating numerous potential splits and selecting the one that minimizes a chosen split objective (typically a purity measure), leading to a highly non-convex optimization landscape. 
        While neural networks also face non-convex optimization challenges, they benefit from the continuity of their parameter space, allowing the use of gradient-based methods. In contrast, the discrete nature of DTs typically precludes the direct application of such techniques. 
        The difficulty is increased by the fact that learning an optimal DT is NP-hard~\citep{laurent1976constructing}. The search space grows exponentially with the depth of the tree, making exhaustive methods impractical for anything beyond shallow trees, even with modern algorithms~\citep{bertsimas2017optimal,aglin2020learning,murtree}.
    \item \textbf{Decision trees are non-differentiable} \hspace{0.25cm}
        DTs are widely employed in various machine learning applications due to their inherent interpretability, ease of implementation, and ability to handle both categorical and continuous data. However, the learning process for DTs presents significant challenges, primarily because the associated optimization problem is non-convex and non-differentiable. The structure of a DT involves discrete choices at each node, which makes the gradient undefined. 
        An additional challenge arises in selecting the feature (index) and corresponding threshold at each node. Since the DT structure relies on selecting discrete feature splits, the optimization problem becomes combinatorial. In traditional methods, this selection is done by evaluating all possible splits and choosing the one that minimizes impurity. However, when attempting to apply gradients to this process, the discrete nature of feature and path selection poses a significant challenge. Unlike continuous variables, where gradients can guide parameter updates, the discrete choice of which feature to split on is inherently a non-differentiable decision. 
        This non-differentiability prevents the straightforward application of many conventional gradient-based optimization methods, like gradient-descent through backpropagation.

\end{itemize}

\section{Research Questions} \label{sec:research_questions}
Building on the problems and challenges outlined above, in this section, we formulate the main research questions addressed in this work. Thereby, we focus on the feasibility of learning DTs using gradient descent. By exploring a new approach to DT learning, this research seeks to connect classic interpretable models with modern gradient-based optimization, potentially enabling integration within existing learning frameworks. The following research questions guide this investigation, addressing both the fundamental challenges of learning DTs with gradient descent and their broader applicability across different machine learning research fields.

\begin{enumerate}[label={\textbf{RQ\arabic*}}, left=0pt, itemsep=1ex]
\item \textbf{Is it possible to efficiently learn hard, axis-aligned DTs with gradient descent?} \label{rq1}

The potential to learn DTs through gradient-based optimization methods offers a promising alternative to traditional approaches by allowing for the simultaneous optimization of all tree parameters, potentially overcoming certain inherent limitations. This research question seeks to explore whether it is feasible to learn hard, axis-aligned DTs using gradient descent techniques. The investigation is motivated by the potential advantages of using gradient descent, such as the ability to employ a global search over all model parameters and to integrate tree learning into broader deep learning frameworks. The challenge lies in the non-differentiable nature of the traditional DT splits, which requires careful reformulation to deal with gradient propagation. This research will investigate whether our approach can yield an efficient learning process that maintains the interpretability and performance of traditional DTs.

    \begin{enumerate}[label={\textbf{RQ\arabic{enumi}.\arabic*}}, left=0em, itemsep=1ex]
        \item \textbf{Can we efficiently optimize all tree parameters (threshold, feature to split on, prediction in leaf) jointly?} \label{rq1.1}
        
        Traditional DT algorithms typically use a greedy, stepwise process to select split points, which prevents the joint optimization of tree parameters. Gradient-based methods, however, could potentially allow for the simultaneous optimization of all relevant parameters (thresholds, feature selection, and predictions in the leaf nodes). 
        This sub-question seeks to investigate the feasibility of such joint optimization by finding differentiable approximations to ensure a reasonable gradient flow and examining the associated trade-offs. A key focus will be on balancing effective gradient updates with preserving the tree’s hard, axis-aligned structure. To address the challenge of non-differentiable operations during gradient propagation, we will explore the use of the straight-through (ST) operator as a mechanism for maintaining this structure. Additionally, we will conduct both theoretical and empirical comparisons of the ST operator with alternative approaches proposed in the literature.

        \item \textbf{Does a gradient-based optimization of decision trees solve the issue of local optimality?} \label{rq1.2}
        
        While conventional DT construction relies on a sequential, greedy selection of splits, often resulting in locally optimal, but globally suboptimal, solutions, a gradient-based approach offers a more integrated exploration of the decision space. This sub-question investigates whether applying gradient descent to DT learning can effectively address the limitations imposed by greedy, sequential split selection. Moreover, given the non-convex nature of the problem, it is crucial to examine whether the alternative local optima encountered in a gradient-based framework can be managed as effectively as those in neural network training, where convergence to a sufficiently good local optimum often results in robust performance.

        \item \textbf{Is learning decision trees with gradient descent runtime efficient?}\label{rq1.3}
        
        One of the primary concerns when applying gradient descent to DTs is the computational efficiency compared to traditional tree learning algorithms. Gradient descent typically involves iterative updates over multiple epochs, which can be computationally expensive. This sub-question addresses the runtime efficiency of gradient-based learning for DTs, examining how the computational cost scales with the depth of the tree, the number of features, and the size of the dataset. The question will also explore optimization strategies such as batching and parallelization that can make the learning process more efficient. 
    
        \item \textbf{Are the resulting trees interpretable?} \label{rq1.4}
        
        Interpretability is a key advantage of DTs over more complex models such as deep neural networks. However, the introduction of gradient-based optimization raises questions about whether the resulting trees retain the same level of interpretability. 
        This sub-question investigates the interpretability of trees learned through gradient descent, particularly whether the final trees maintain the straightforward, axis-aligned decision boundaries of traditional DTs. The investigation will include empirical evaluations of tree size and simplicity, with a focus on how these trees compare to their traditionally trained counterparts.
    
    \end{enumerate}

\item \textbf{Can gradient-based decision trees be extended to form an effective ensemble?} \label{rq2}

Extending from single DTs to ensembles is crucial for enhancing model robustness, scalability to high-dimensional datasets, and predictive power. This approach allows for leveraging multiple weak learners to improve performance, reduce variance, and increase generalization capabilities. An effective transition to ensembles involves integrating the gradient-based optimization while preserving essential DT characteristics, such as axis-aligned splits and deterministic routing.

    \begin{enumerate}[label={\textbf{RQ\arabic{enumi}.\arabic*}}, left=0em, itemsep=1ex]

        \item \textbf{Does an extension from individual trees to a tree ensemble yield a considerable performance increase?} \label{rq2.1}
        
        Extending individual trees to a tree ensemble significantly increases complexity and decreases the overall interpretability. However, typically this can lead to a substantial performance increase, similar to the extension from DTs to random forests as a DT ensemble. In this research question, we want to examine whether we can observe a similar performance increase for our gradient based methods compared to using traditional ensembling methods for DTs.
    
        \item \textbf{Can an ensemble of gradient-based decision trees maintain efficient tensor computations?} \label{rq2.2}
        
        When transitioning to ensembles, an essential concern is computational efficiency. In this question, we want to elaborate whether it is possible to formulate a whole ensemble in a way that allows for efficient tensor-based operations. We want to ensure that the transition from individual trees to ensembles does not compromise the scalability and speed offered by gradient-based optimization methods.

        \item \textbf{Can gradient-based optimization enable advanced weighting techniques for tree ensembles?} \label{rq2.3} 
        
        Traditionally, ensemble methods like random forests trees rely on fixed weight combinations (such as simple averaging) for their final predictions, since re-learning weights is challenging and costly. Using gradient-based optimization, however, allows for more sophisticated, potentially instance-specific weighting schemes. This would allow each estimator in the ensemble to focus on specific areas of the feature space, improving the model's local interpretability and performance. In an end-to-end gradient-based setup, we can include learning the weights directly into the optimization without compromising efficiency.
        
    \end{enumerate}  

\item \textbf{Can gradient-based optimization enable the integration of hard, axis-aligned decision trees into existing approaches for alternative domains?} \label{rq3}

The utility of gradient-based DTs extends beyond standalone tasks on tabular data. They have the potential to be integrated into broader machine learning architectures and frameworks, enabling new capabilities and applications. This research question explores whether gradient-based DTs can be effectively incorporated into existing approaches for alternative domains, such as multimodal learning and reinforcement learning, and how they can contribute to improving model performance and interpretability. The focus will be on adapting gradient-based tree learning to fit within established frameworks and on evaluating the impact of such integration on model performance, interpretability, and computational efficiency.

    \begin{enumerate}[label={\textbf{RQ\arabic{enumi}.\arabic*}}, left=0em, itemsep=1ex]
        \item \textbf{Can we incorporate a tree-based structure, learned with gradient descent, as a tabular component into a multimodal learning architecture?} \label{rq3.1}    
        
        Multimodal learning involves integrating information from different data modalities, such as images, and structured tabular data. Gradient-based DTs, due to their ability to be trained using backpropagation, could be seamlessly integrated into such architectures and trained end-to-end. This sub-question investigates whether a tree-based structure, learned using gradient descent, can serve as the tabular component within a multimodal model, potentially leading to improved performance over traditional approaches that use NNs to handle the tabular component or handle tabular data entirely separate. The research will focus on how the tree-based component interacts with other modalities in the architecture and the potential benefits in terms of performance, interpretability, and training efficiency.
    
        \item \textbf{Can we integrate gradient-based decision trees into existing reinforcement learning approaches like Proximal Policy Optimization to learn interpretable policies?} \label{rq3.2}   
        
        Reinforcement learning approaches often rely on (deep) neural networks for policy representation, which can be difficult to interpret. Integrating gradient-based DTs into these approaches, particularly in methods like Proximal Policy Optimization~\citep{schulman2017proximal}, could enhance interpretability and potentially improve performance in certain scenarios. This sub-question explores the feasibility of integrating DTs as RL policies and the impact on learning performance. Thereby, we will consider the impact of the inherent structure of DTs on the stability of policy updates. Furthermore, we will explore how interpretable policies as DTs can provide insights into the decision-making process.
    
    \end{enumerate}    
\end{enumerate}

\section{Contributions \& Published Work}
\label{sec:contributions}

In this thesis, we propose a novel method that enables the learning of hard, axis-aligned DTs using gradient descent. 
The proposed method uses backpropagation with a straight-through operator on a dense DT representation, to jointly optimize all tree parameters. Our method represents a significant advancement in the field, opening the door to a variety of new application scenarios for DTs. 
Specifically, our contributions are as follows:

\paragraph{A novel method for learning axis-aligned tree-based models with gradient descent (Chapter~\ref{cha:methodology}).}
\begin{itemize}
    \item We propose \textbf{Grad}ient-based decision \textbf{Tree}s (GradTree) as novel method for learning hard, axis-aligned decision trees with gradient descent (Section~\ref{sec:method_gradtree}).
    \begin{itemize}
        \item We introduce a dense DT representation that enables a joint, gradient-based optimization of all tree parameters (Section~\ref{ssec:dense_rep}).
        \item We present a procedure to deal with the non-differentiable nature of DTs using backpropagation with a straight-through (ST) operator (Section~\ref{ssec:adjusted_backprop}).
        \item We propose a novel tree routing that allows an efficient, parallel optimization of all tree parameters with gradient descent (Section~\ref{ssec:training}).
    \end{itemize}
    \item We present an extension of GradTree to \textbf{GRA}die\textbf{N}t-based \textbf{D}ecision tree \textbf{E}nsembles (GRANDE) as a performance-interpretability trade-off (Section~\ref{sec:method_grande}).
    \begin{itemize}
        \item We extend GradTree from individual trees to an end-to-end gradient-based tree ensemble, maintaining efficient computation (Section~\ref{ssec:ensembles}).
        \item We propose a novel weighting technique that emphasizes instance-wise estimator importance (Section~\ref{ssec:weighting}).       
    \end{itemize}
\end{itemize}

\paragraph{Gradient-based decision trees as new state-of-the-art method for tabular data (Chapter~\ref{cha:tabular}).}

\begin{itemize}
    \item State-of-the-art results in learning interpretable DTs for small tabular data problems (Section~\ref{ssec:experiment_results}).
    \begin{itemize}
        \item We verify that our gradient-based optimization can mitigate the challenges of local optimality arising from a constrained search space and instead converge to a local optimum with strong generalization performance.
        \item We show that DTs learned by GradTree outperform alternative methods on most benchmark datasets, where the performance difference is substantial for several datasets.
        \item We demonstrate that our method is more robust to overfitting compared to existing methods.
        \item We confirm that the DTs learned by GradTree are interpretable and have a similar effective tree size after pruning compare to alternative methods.
    \end{itemize}
    \item State-of-the-art results for complex tabular datasets with gradient-based DT ensembles (Section~\ref{sec:eval_grande}).
    \begin{itemize}
        \item We find that GRANDE achieves state-of-the-art results for complex tabular datasets, outperforming tree-based and gradient-based methods on most datasets.
        \item We show that GRANDE is computationally efficient for large and high-dimensional datasets, scaling well with the number of estimators by leveraging parallel computation.
        \item We verify that the novel instance-wise weighting increases the model performance and can support local interpretability.
    \end{itemize}
    \item Performance gain using a tree-based tabular backbone in multimodal learning with end-to-end gradient-based optimization (Section~\ref{ssec:evaluation_multimodal}).
    \begin{itemize}
        \item We demonstrate that using GRANDE as a tabular backbone in a multimodal learning setup can lead to a performance gain out-of-the-box.
    \end{itemize}    
\end{itemize}

\paragraph{Mitigating information loss in tree-based reinforcement learning via direct optimization (Chapter~\ref{cha:sympol}).}
\begin{itemize}
    \item We extend GradTree to efficiently learn interpretable, axis-aligned DT policies end-to-end using gradient descent.
    \item We show that our interpretable policy can directly be optimized using policy gradients without information loss.
    \item We demonstrate that our method allows a seamless integration into existing RL approaches by the example of proximal policy optimization (PPO) and advantage actor critic (A2C).
    \item We propose a dynamic rollout buffer to enhance exploration stability and a dynamic batch size through gradient accumulation to improve gradient stability.
    \item We confirm that our policies are compact and interpretable.
\end{itemize}

By bridging the gap between DTs and gradient descent, our approach paves the way for broader adoption of tree-based models in areas where interpretability, efficiency, and integration with modern optimization frameworks are of high importance. \\

\noindent This dissertation is grounded in articles that have already been presented on prestigious international conferences. I am the first author and have made significant contributions to all the following papers:

\begin{itemize}
    \item \printpublication{gradtree_bold} \\
    \textbf{CORE2023 Ranking: A$^{*}$}
    \paragraph{Personal contribution}
    I conceived the idea of learning axis-aligned DTs using gradient descent and designed the initial proof-of-concept implementation. I also developed the dense architecture, integrated the straight-through estimator to address non-differentiable operations, and implemented the necessary modifications to support gradient flow. Additionally, I designed the efficient parallel algorithm and implemented the final version, which included various runtime and performance optimizations. I was the primary author of the manuscript. Stefan Lüdtke, Christian Bartelt, and Heiner Stuckenschmidt contributed to the paper's development and the approach's design through thoughtful discussions and insightful feedback throughout the process. They also assisted in proofreading and refining the final version.
    
    \item \printpublication{grande_bold} \\
    \textbf{CORE2023 Ranking: A$^{*}$}
    \paragraph{Personal contribution}
    I initiated the idea of extending gradient-based DTs to DT ensembles and designed the initial proof-of-concept implementation. I subsequently implemented the final algorithm, incorporating several runtime and performance optimizations. The concept of using an advanced, instance-wise weighting scheme based on the leaf nodes emerged from a collaborative discussion, and I implemented the idea and designed the experiments to demonstrate its effectiveness. I also introduced the use of the softsign function to enhance gradient flow, replacing the commonly used sigmoid function. I was the primary author of the manuscript. Stefan Lüdtke, Christian Bartelt, and Heiner Stuckenschmidt provided valuable insights and contributed to the paper and the approach's design through rigorous discussions. They also helped proofread and refine the final manuscript.

    \item \printpublication{marton2024sympol_bold} \\
    \textbf{CORE2023 Ranking: A$^{*}$}
    \paragraph{Personal contribution}
    I developed the idea of using gradient-based DTs as a policy in a reinforcement learning setup and created the initial implementation. Florian Vogt enhanced this implementation by optimizing its efficiency, particularly through the integration of JAX. I also designed and implemented the dynamic rollout buffer and the modified Adam optimizer with weight decay. Together with Tim Grams, I further refined the approach, designed and conducted the experiments, and co-authored the manuscript. Through in-depth discussions with Christian Bartelt, Stefan Lüdtke, and Heiner Stuckenschmidt, we refined the paper, with all contributing to the structuring and proofreading of the manuscript. Stefan Lüdtke, Christian Bartelt, and Heiner Stuckenschmidt provided valuable insights and contributed to the overall design and development of the approach throughout the process.
    
\end{itemize}

\section{Outline}

This thesis is structured to first give an introduction into the field, including the corresponding background and related work. Next, we will describe the overall methodology of gradient-based DTs, which are the main novelty and contribution of this thesis and also form the foundation of the remaining dissertation. After describing our method in general, we will go through several application scenarios, namely tabular data, multimodal learning and reinforcement learning, showcasing the effectiveness in a wide range of settings and introducing relevant adjustments to achieve state-of-the-art performance. We will then conclude the dissertation with a discussion, conclusion and future work. The content of the subsequent chapters will be outlined more detailed in the following:

\subsubsection*{\autoref{cha:background} - \nameref{cha:background}}
This chapter provides the essential background, covering both DTs and gradient descent optimization as foundational elements of our approach. We introduce the structure and mechanics of DTs, including comparisons between axis-aligned and oblique splits, and present the fundamentals of gradient descent optimization and backpropagation. Advanced techniques such as the Adam optimizer and weight decay are also explored to support the novel gradient-based DT approach. This background equips the reader with the theoretical knowledge needed for the methods and applications introduced later.

\subsubsection*{\autoref{cha:related} - \nameref{cha:related}}
In this chapter, we review related work on DT learning algorithms, examining traditional greedy approaches, evolutionary algorithms, and optimal DTs. We also explore recent advancements in gradient-based DT methods, including soft DTs and ensemble techniques. By highlighting the limitations of these existing approaches, particularly their inability to support non-differentiable, hard, axis-aligned splits, we establish a basis for our methodology and identify a research gap addressed within this thesis.

\subsubsection*{\autoref{cha:methodology} - \nameref{cha:methodology}}
In this chapter, we introduce our primary methodology for learning axis-aligned DTs through gradient descent, which forms the core of our contributions. We detail the GradTree framework, which reformulates DT learning to enable joint optimization of parameters using gradient-based methods. We leverage techniques like the straight-through operator and a dense tree representation to maintain hard, axis-aligned splits while taking full advantage of gradient-based optimization. 
After presenting the core method, we provide a theoretical discussion and compare it with the learning methods introduced in Chapter~\ref{cha:related}.
In the final part of the chapter, we extend GradTree from individual trees to GRANDE, a tree ensemble, as a performance-interpretability trade-off.

\subsubsection*{\autoref{cha:tabular} - \nameref{cha:tabular}}
This chapter focuses on applying gradient-based DTs to tabular data, where DTs have traditionally excelled. We first discuss the specific challenges of tabular data before presenting our evaluation of gradient-based DTs on smaller tabular datasets. These experiments demonstrate GradTree’s ability to maintain interpretability while delivering competitive performance. We then scale the method to more complex real-world datasets, showing how our gradient-based DT ensemble, GRANDE, achieves state-of-the-art performance in the tabular data domain.

\subsubsection*{\autoref{cha:sympol} - \nameref{cha:sympol}}
In this chapter, we explore how gradient-based DTs can be adapted for reinforcement learning through a symbolic, tree-based on-policy approach called SYMPOL. 
We outline relevant background and motivation from the reinforcement learning literature and position SYMPOL as an innovative method for interpretable reinforcement learning, mitigating the information loss typically associated with DT policies.
We demonstrate how gradient-based DTs offer unique benefits in reinforcement learning, a field where traditional tree-based methods are rarely applied due to their incompatibility with gradient-based optimization.

\subsubsection*{\autoref{cha:conclusion} - \nameref{cha:conclusion}}
In this final chapter, we synthesize the key findings and contributions of the thesis, reflecting on the advantages and challenges of the gradient-based DT method introduced. We discuss the broader implications of this work, including potential applications across fields requiring interpretability, efficiency, and integration with modern machine learning frameworks. In this context, we outline future research directions and explore additional applications. Finally, we conclude the thesis by answering the research questions formulated in the introduction, highlighting the impact of the proposed method.

\chapter{Background: Decision Trees and Gradient Descent Optimization} \label{cha:background}

In this chapter, we present the foundational background necessary for this thesis, focusing on the two main components of our approach: \emph{Decision Trees} (Section~\ref{sec:decision_trees_background}) and \emph{Gradient Descent Optimization} (Section~\ref{sec:gradient_descent_background}). We provide a comprehensive overview of the essential concepts required for the remainder of the thesis. For DTs, this includes the structure of the tree (Section~\ref{sec:dt_formalization}) and a comparison between axis-aligned and oblique splits (Section~\ref{ssec:split_types}), as well as hard and soft decisions (Section~\ref{ssec:sdt_background}). For gradient descent optimization, we discuss the general process of gradient descent (Section~\ref{sec:gradient_descent_optimization}) with backpropagation and automatic differentiation (Section~\ref{sec:backpropagation}), along with advancements for more efficient optimization (Section~\ref{sec:advancements_gradient_descent}), specifically the Adam optimizer \citep{kingma2014adam}, which is critically important to our approach. We provide a summary of the notation used throughout the thesis in Appendix~\ref{A:notation}.

\section{Decision Trees} \label{sec:decision_trees_background}

\subsection{Decision Tree Formalization} \label{sec:dt_formalization}

DTs are fundamental structures in machine learning and statistical modeling, used for classification and regression tasks. From a mathematical standpoint, a DT represents a function that maps input variables to an output variable by partitioning the input space into mutually exclusive regions. This section provides a comprehensive mathematical formalization of DTs, focusing on their structure and functioning.

\paragraph{Basic graph structure}

A decision tree \( T \) is a finite, rooted tree defined as a tuple \( T = (V, E) \), where  \( V \) is a finite set of nodes and \( E \subseteq V \times V \) is a set of directed edges connecting the nodes.
The node set \( V \) can be partitioned into three disjoint subsets:

\begin{enumerate}
    \item \textbf{Root Node (\( v_{root} \))}: The unique node with no incoming edges.
    \item \textbf{Internal Nodes (\( V_I \))}: Nodes with exactly one incoming edge and two or more outgoing edges.
    \item \textbf{Leaf Nodes (\( V_L \))}: Nodes with exactly one incoming edge and no outgoing edges.
\end{enumerate}

\paragraph{Function representation}

A DT represents a piecewise constant function \( t: \mathbb{R}^n \rightarrow \mathbb{R} \) (for regression) or \( t: \mathbb{R}^n \rightarrow \mathcal{C} \) (for classification), where \( \mathbb{R}^n \) is the input space with \( n \) features and \( \mathcal{C} \) is the set of classes in classification tasks.
The tree partitions the input space \( \mathbb{R}^n \) into a set of non-overlapping regions \( \{R_0, R_1, \dots, R_{L-1}\} \), each corresponding to a leaf node with $L = |V_L|$. The function \( t \) can be expressed as:

\begin{equation}
t(\boldsymbol{x}) = \sum_{l=0}^{L-1} \lambda_l \cdot \mathbb{L}_{\{\boldsymbol{x} \in R_l\}},
\end{equation}

where \( \boldsymbol{x} \in \mathbb{R}^n \) is an input vector, \( \lambda_l \) is the constant value assigned to region \( R_l \) and \( \mathbb{L}_{\{\cdot\}} \) is an indicator function.

\paragraph{Internal nodes / split nodes}

Each internal node \( v \in V_I \) implements a decision rule that partitions its incoming region into two or more subregions. Formally, a decision rule at node \( v \) is a function \( \mathbb{S}_v: \mathbb{R}^n \rightarrow \{1, 2, \dots, k_v\} \), where \( k_v \geq 2 \) is the number of child nodes of \( v \). For binary trees, \( k_v = 2 \) for all \( v \in V_I \).
Specifically, \( \mathbb{S}_v(\boldsymbol{x}) \) returns an index representing which child node the input \( \boldsymbol{x} \in \mathbb{R}^n \) should be routed to, based on the decision criteria at node \( v \).
In the most general form, a decision rule for a binary tree can be defined as:

\begin{equation}
\mathbb{S}_v(\boldsymbol{x}) = \begin{cases}
\mathrm{index}(v_\text{left}) & \text{if a given condition is true,} \\
\mathrm{index}(v_\text{right}) & \text{otherwise,}

\end{cases}
\end{equation}

where \( \mathbb{S}_v \) routes \( \boldsymbol{x} \) to either the left ($v_\text{left}$) or right ($v_\text{right}$) child depending on a predefined condition. 
The condition at each node can typically take one of several forms. The most common form compares a single feature value \( x_\iota \) of the input vector \( \boldsymbol{x} = (x_0, x_1, \dots, x_{n-1}) \) with a threshold value \( \tau \), such as \( x_\iota \geq \tau \). This results in axis-aligned splits of the input space. More complex conditions can involve multiple features, such as linear combinations \( \boldsymbol{w}^T\boldsymbol{x} + \tau \geq 0 \) for some weight vector \( \boldsymbol{w}_v \) and bias term \( \tau \), leading to oblique splits. We will go into detail on these two split types in Section~\ref{ssec:split_types}.
The decision rule at node \( v \) partitions the input space \( R_v \) associated with node \( v \) into \( k_v \) disjoint subsets:

\begin{equation}
R_v = \bigcup_{i=0}^{k_v} R_{v,i}
\; \text{such that:} \;
R_{v,i} = \{\boldsymbol{x} \in R_v \mid \mathbb{S}_v(\boldsymbol{x}) = i\}.
\end{equation}

\paragraph{Leaf nodes and output values}

Each leaf node \( l \in V_L \) corresponds to a region \( R_l \) and assigns an output value \( \lambda_l \). In classification tasks, \( \lambda_l \in \mathcal{C} \); in regression tasks, \( \lambda_l \in \mathbb{R} \).

\paragraph{Traversal mechanism}

To compute \( t(\boldsymbol{x}) \) for a given input \( \boldsymbol{x} \), the tree is traversed from the root node according to the decision rules at each internal node until it reaches a leaf node (Algorithm~\ref{alg:traversal}).

\begin{algorithm}
\caption{Tree Traversal}
\label{alg:traversal}
\begin{algorithmic}[1]
\Function{traverse}{$\boldsymbol{x}$}
\State \textbf{Initialize} $v \gets v_{root}$ \Comment{Start at root node}
\While{$v$ is not a leaf node}
    \State $i \gets \mathbb{S}_v(\boldsymbol{x})$ \Comment{Apply decision rule $\mathbb{S}_v(\boldsymbol{x})$}
    \State $v \gets v_i$ \Comment{Move to the child node determined by $\mathbb{S}_v(\boldsymbol{x})$} 
\EndWhile

\Return $\lambda_v$ \Comment{Output the value at leaf node $v$}
\EndFunction

\end{algorithmic}

\end{algorithm}

The sequence of decision rules effectively partitions the input space \( \mathbb{R}^n \) into hyperrectangles (in the case of axis-aligned splits) or more complex shapes (e.g., for oblique splits). Each leaf node corresponds to one such partition, and the function \( t \) assigns a constant value within each partition.

\begin{figure}[t]
    \centering
    \begin{subfigure}[t]{0.375\textwidth}
        \centering
        \includegraphics[width=\textwidth]{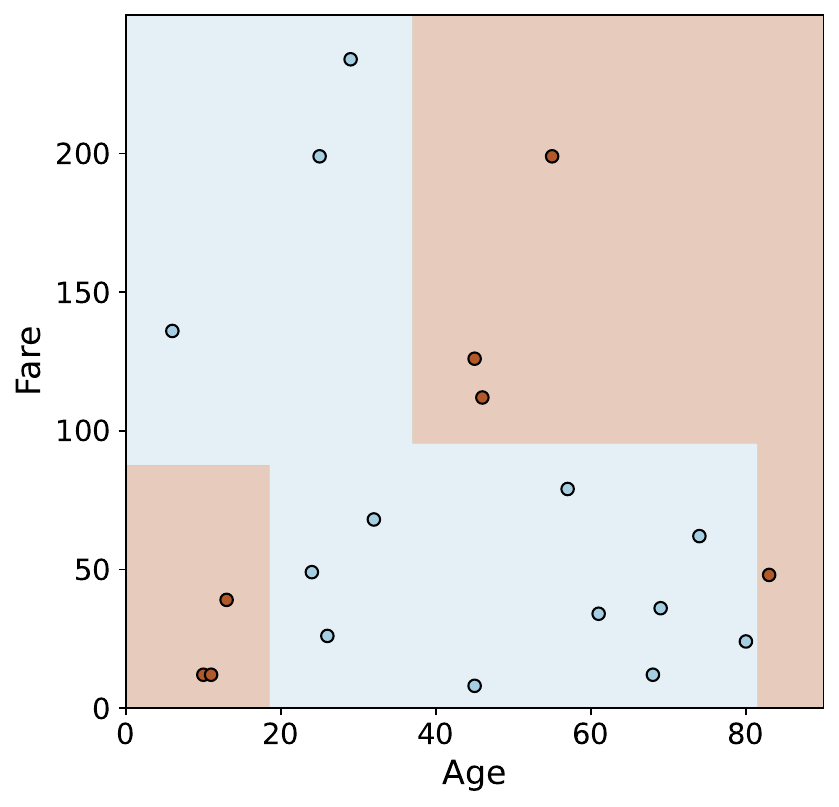}
        \caption{Decision boundary visualization}
        \label{fig:decision_boundary}
    \end{subfigure}%
    \hfill
    \begin{subfigure}[t]{0.60\textwidth}
        \centering
        \includegraphics[width=\textwidth]{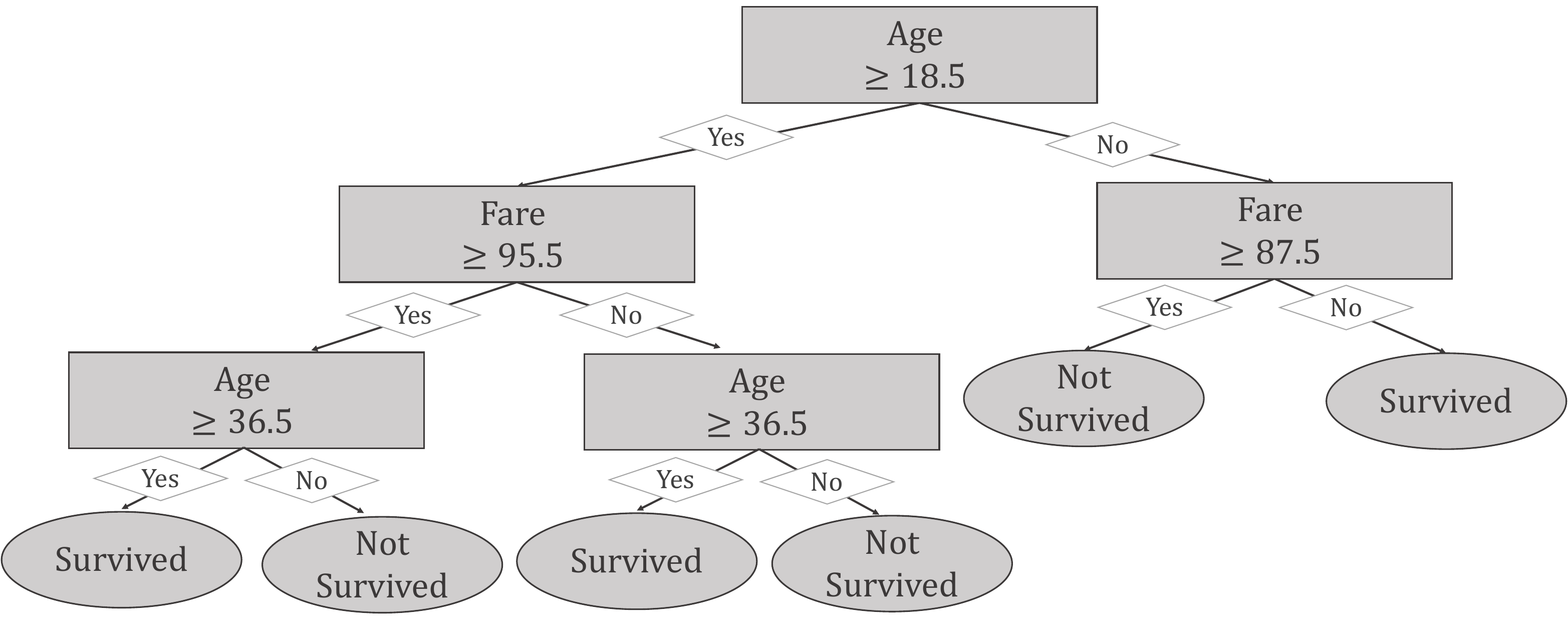}
        \caption{DT structure}
        \label{fig:decision_tree}
    \end{subfigure}
    \caption[Visualization of a DT's Decision-Making Process]{\textbf{Visualization of a DT's Decision-Making Process.} (A) shows how the DT partitions the 2D feature space (\emph{Age} and \emph{Fare}) into regions, with each region corresponding to a prediction of whether a passenger survived. (B) shows the corresponding tree structure, detailing the decision rules at each node.}
    \label{fig:decision_tree_example}
\end{figure}

\paragraph{Exemplary decision tree}
The visualized example in Figure \ref{fig:decision_tree_example} demonstrates how a hard\footnote{In this context, a hard DT refers to a tree that deterministically assigned a sample to only one branch opposed to soft DTs (see Section~\ref{ssec:sdt_background}), where samples are assigned to multiple branches based on a probability.}, axis-aligned, binary DT splits the feature space into distinct regions based on the features \emph{Age} and \emph{Fare} of an exemplary titanic dataset where the task is to predict survival. 
This exemplary dataset will be introduced in more detail in Section~\ref{ssec:2D_titanic} and be used throughout the background and related work section of this thesis. %
The left subfigure (\ref{fig:decision_boundary}) shows the two-dimensional decision boundaries created by the DT classifier, where each region corresponds to a prediction for survival. The right subfigure (\ref{fig:decision_tree}) presents the structure of the DT, with internal nodes representing decision rules and leaf nodes representing final predictions. The tree uses these decision rules to traverse from the root node to the leaf nodes, thereby classifying data points into the respective classes.

\paragraph{Complexity}

The complexity of a DT is determined primarily with respect to the tree's depth \( d \) and the number of nodes \( |V| \):
    
\begin{itemize}
    \item \textbf{Depth (\( d \))}: The length of the longest path from the root to a leaf node. This corresponds to the maximum number of conditions required to define a region.
    \item \textbf{Number of nodes (\( |V| = |V_I| + |V_L| \))}: Total number of nodes in the tree, where $|V_I|$ nodes are internal nodes and $L=|V_L|$ nodes are leaf nodes.
\end{itemize}

The computational complexity of a DT can be analyzed in terms of both time and space.

\begin{itemize}
    \item \textbf{Traversal time complexity}: To evaluate an input \( \boldsymbol{x} \), the tree is traversed from the root to a leaf node. For a balanced tree with depth \( d \), the time complexity of this traversal is \( O(d) \). In the best-case scenario, if the tree is perfectly balanced, \( d \) is proportional to \( \log L \), where \( L \) is the number of leaf nodes, i.e., \( d = O(\log L) \). However, in the worst case, where the tree is unbalanced and resembles a chain, \( d \) can be as large as \( L \), leading to a worst-case time complexity of \( O(L) \).
    
    \item \textbf{Space complexity}: The space complexity of storing a DT is determined by the total number of nodes \( |V| \). Each internal node stores a decision rule along with pointers to its children, while each leaf node contains an output value. Consequently, the space complexity is proportional to the number of nodes in the tree. For a binary tree that is reasonably balanced, the number of nodes can be approximated by \( |V| \approx 2L - 1 \), leading to a space complexity of \( O(L) \). For more general cases, where internal nodes may have \( k_v \) children, the space complexity remains \( O(|V|) \), as the total number of nodes continues to dictate the storage requirements.
    In the case of balanced binary trees, the number of leaves \( L \) is related to the tree's depth \( d \). Since each additional level in a balanced binary tree doubles the number of leaves, \( L \) grows exponentially with depth, such that \( L \approx 2^d \). Consequently, the total number of nodes becomes \( |V| \approx 2(2^d) - 1 \), which simplifies to a space complexity of \( O(2^d) \). Therefore, for balanced binary trees, the space complexity is governed by the depth \( d \), resulting in an overall space complexity of \( O(2^d) \).

\end{itemize}

Furthermore, the \textbf{training complexity} is defined by the time required to learn a decision tree. However, the training time depends on the learning algorithm, which will be discussed in Chapter~\ref{cha:related}.

\paragraph{Universal approximation}

Neural networks are widely recognized as universal function approximators due to their ability to approximate any continuous function under certain conditions~\citep{cybenko1989approximation,hornik1989multilayer}. Nevertheless, they are not the sole class of models capable of this universal approximation property. Interestingly, DTs have also been demonstrated to approximate any function defined over a finite input space, according to the simple function approximation theorem~\citep{royden2010real}. Given sufficient depth and appropriate splits, a DT can achieve an arbitrary level of granularity in partitioning the input space~\citep{watt2020machine}.  However, this comes at the cost of the tree potentially becoming highly complex (large number of nodes and depth) and, especially for smooth functions, the representation is not as efficient as a neural network’s.

\begin{figure}[t]
    \centering
    \begin{subfigure}[t]{0.45\textwidth}
        \centering
        \includegraphics[width=\textwidth]{Figures/Background/decision_boundaries_axis_aligned.pdf}
        \caption{Axis-Aligned DT.}
        \label{fig:decision_tree_axis_aligned_boundary}
    \end{subfigure}
    \hfill
    \begin{subfigure}[t]{0.45\textwidth}
        \centering
        \includegraphics[width=\textwidth]{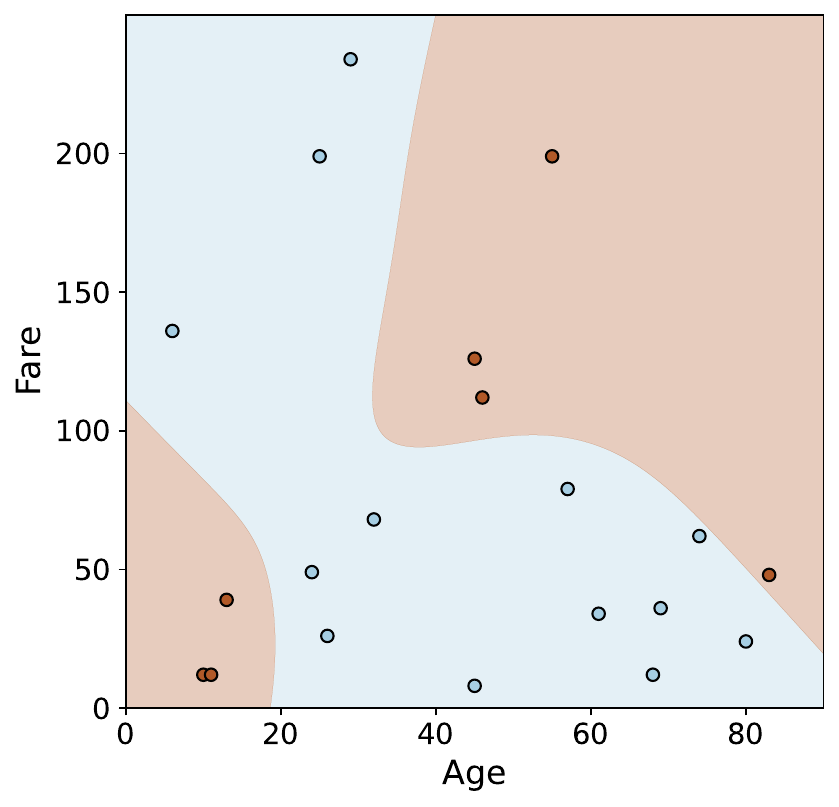}
        \caption{Oblique DT.}
        \label{fig:decision_tree_soft_boundary}
    \end{subfigure}
    \caption[Decision Boundaries Comparison]{\textbf{Decision Boundaries of Axis-Aligned vs. Oblique DT on Reduced Titanic Dataset.} The oblique DT's decision boundaries (B) reflect the ability to form linear combinations of features, resulting in more flexible boundaries of arbitrary shapes. This flexibility is evident in the oblique tree's decision boundaries, which are not limited to being parallel to the feature axes, unlike the axis-aligned tree (A).}
    \label{fig:decision_boundary_comparison}
\end{figure}

\subsection{Split Types}\label{ssec:split_types}
In this section, we discuss the different types of decision rules that can be used at the internal nodes of a DT. The decision rules determine how the feature space \( \mathcal{X} \subseteq \mathbb{R}^n \) is partitioned and thus how the tree is structured. Specifically, we distinguish between \textit{axis-aligned} splits and \textit{oblique} splits. A visualization on an exemplary titanic dataset comparing axis-aligned and oblique splits is given in Figure~\ref{fig:decision_boundary_comparison}.

\subsubsection*{Axis-Aligned Splits}

DTs are well-known for their use of \textit{axis-aligned} splits, a defining characteristic that contributes to their simplicity and interpretability. These splits divide the data space using boundaries that are aligned with the feature axes, making them both efficient to compute and straightforward to understand.
An \textit{axis-aligned} split represents the simplest form of split, where the decision rule at an internal node \( v \) is based on a single feature \( x_\iota \). The decision function \( \mathbb{S}_v \) for an axis-aligned split is defined as:

\begin{equation}
\mathbb{S}_v(\boldsymbol{x}|\iota, \tau) = \begin{cases}
\mathrm{index}(v_\text{left}) & \text{if } x_\iota \geq \tau, \\
\mathrm{index}(v_\text{right}) & \text{if } x_\iota < \tau,
\end{cases}
\end{equation}

where \( \boldsymbol{x} = (x_0, x_1, \dots, x_{n-1}) \in \mathcal{X} \) is an input vector, \( x_\iota \) is the value of the feature at index \( \iota \), and \( \tau \) is the corresponding threshold. Here, \( v_\text{left} \) and \( v_\text{right} \) denote the left and right child nodes of \( v \), respectively.
Axis-aligned splits create decision boundaries that are orthogonal to one of the feature axes. While these splits are easy to interpret and computationally efficient, they may require a deeper tree or ensembling strategies to accurately model complex relationships between features.

\subsubsection*{Oblique Splits}\label{ssec:oblique splits}

An \textit{oblique} split, in contrast, allows the decision rule to be based on a linear combination of multiple features. The decision function \( \mathbb{S}_v \) for an oblique split at node \( v \) is given by:

\begin{equation} \label{eq:oblique_split}
\mathbb{S}_v(\boldsymbol{x}|\boldsymbol{w},\tau) = 
\begin{cases} 
\mathrm{index}(v_\text{left}) & \text{if } \boldsymbol{w}^\top \boldsymbol{x} \geq \tau, \\
\mathrm{index}(v_\text{right}) & \text{if } \boldsymbol{w}^\top \boldsymbol{x} < \tau,
\end{cases}
\end{equation}

where \( \boldsymbol{w} = (w_{0}, w_{1}, \dots, w_{n-1})^\top \in \mathbb{R}^n \) is a vector of weights corresponding to the features \( x_0, x_1, \dots, x_{n-1} \), and \( \tau \) is the corresponding threshold. The linear combination \( \boldsymbol{w}^\top \boldsymbol{x} \) defines a hyperplane in the feature space that serves as the decision boundary.
Oblique splits can capture more complex relationships between features with fewer nodes compared to axis-aligned splits. However, this can lead to less interpretable decision rules~\citep{molnar2020}, especially if the weight vector \( \boldsymbol{w} \) is not sparse (i.e., many features are considered at once).

\subsubsection*{Comparison and Implications}

The choice between axis-aligned and oblique splits has significant implications for both the interpretability and complexity of the resulting DT. Axis-aligned splits are favored for their simplicity and ease of interpretation, making them suitable for many practical applications. Additionally, for domains like tabular data~\citep{grinsztajn2022tree} or categorical reinforcement learning~\citep{fuhrer2024gradient}, axis-aligned splits can provide a beneficial inductive bias. For instance, \citet{grinsztajn2022tree} and \citet{beyazit2024inductive} showed that axis-aligned splits are well suited for irregular, high-frequency target functions typically present in tabular data tasks.
On the other hand, oblique splits can model complex decision boundaries more effectively, but at the cost of increased computational complexity and reduced interpretability.

\begin{figure}[t]
    \centering
    \begin{subfigure}[t]{0.42\textwidth}
        \centering
        \includegraphics[width=\textwidth]{Figures/Background/decision_tree_axis_aligned.pdf}
        \caption{Axis-Aligned DT}
        \label{fig:decision_tree_axis_aligned}
    \end{subfigure}
    \hfill
    \begin{subfigure}[t]{0.55\textwidth}
        \centering
        \includegraphics[width=\textwidth]{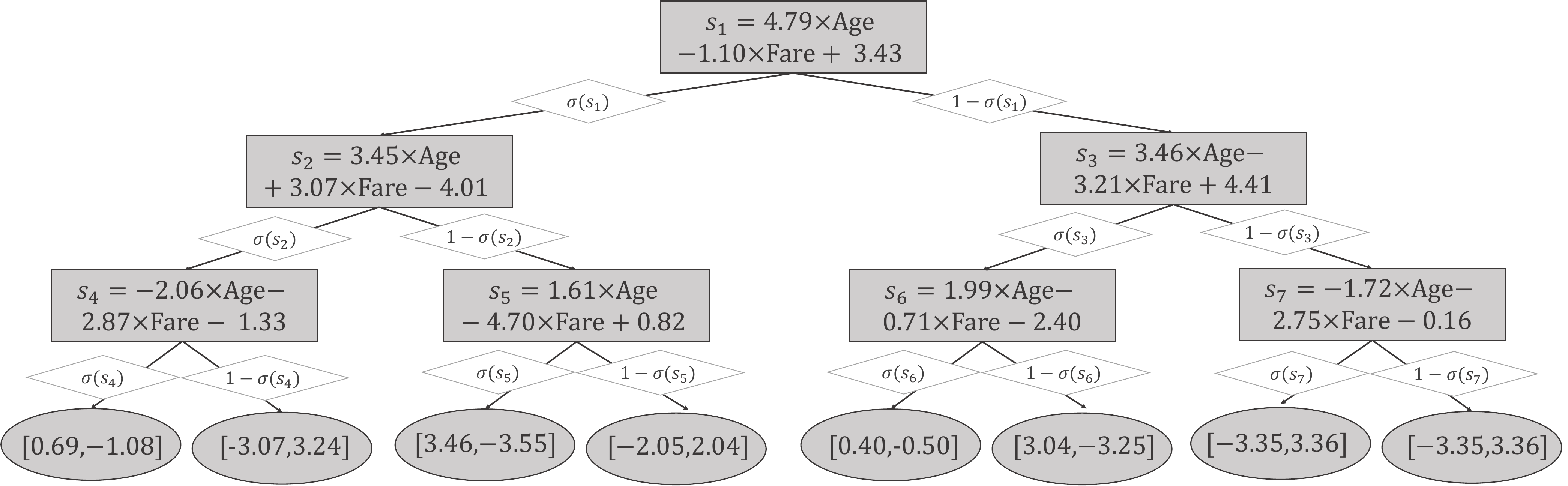}
        \caption{Soft, Oblique DT}
        \label{fig:decision_tree_soft}
    \end{subfigure}
    \caption[Axis-Aligned vs. Soft, Oblique DT on Reduced Titanic Dataset]{\textbf{Axis-Aligned vs. Soft, Oblique DT on Reduced Titanic Dataset.} The comparison between the axis-aligned DT (A) and the soft, oblique DT (B), as shown in the decision boundary plots, highlights significant differences in how each model handles data splits. The axis-aligned DT (Figure A) relies on splits that are parallel to the feature axes, leading to decision boundaries that are horizontal or vertical. This method simplifies interpretation, but may require multiple splits to approximate complex decision surfaces. On the other hand, the oblique DT (Figure B) introduces splits at arbitrary angles, leading to oblique decision boundaries that can better capture complex patterns in the data with individual splits but complicates interpretation. Furthermore, soft DTs do not have explicit assignments to a certain class in the leafs, but probability distributions that are averaged over all paths to obtain a prediction.}
    \label{fig:decision_tree_comparison}
\end{figure}

\subsection{Soft Decision Trees} \label{ssec:sdt_background}
In contrast to vanilla DTs that make a hard decision at each internal node, many hierarchical mixtures of expert models~\citep{soft1} have been proposed. Nowadays, these kinds of models are commonly referred to as Oblique Decision Trees or Soft Decision Trees (SDTs).
SDTs are a variant of DTs that incorporate probabilistic splits, allowing them to form a more flexible model compared to traditional hard DTs~\citep{soft1,soft3}. Unlike conventional DTs, which make discrete decisions at each internal node, SDTs utilize oblique splits and assign probabilities to each path, making them a form of hierarchical mixture of experts. This approach enables SDTs to capture both linear and non-linear patterns in data effectively. A comparison of STDs with standard, axis-aligned DTs is given in Figure~\ref{fig:decision_tree_comparison}. 

\paragraph{Oblique splits}
Traditional DTs, as discussed in Section~\ref{sec:dt_formalization} typically rely on axis-aligned splits, where each node makes a hard decision based on the value of a single input feature. In contrast, SDTs~\citep{soft1,soft2,soft3} make use of oblique splits, which involve a linear combination of multiple input features as shown in Section~\ref{ssec:split_types}. 
In the context of SDTs, the oblique split defined in Equation~\ref{eq:oblique_split} is slightly reformulated to account for the probabilistic splits of SDTs (i.e., multiple paths are taken with a certain probability compared to only taking the path that fulfills a certain condition). In this context, we can interpret the bias as the threshold\footnote{The notation is adjusted to match existing implementation frameworks like TensorFlow and PyTorch, typically working with a weight vector and a bias.}.
Formally, at each internal node \( v \), a decision is made based on a learned weight \( \boldsymbol{w} \) and a bias \( b \):

\begin{equation}
p_v(\boldsymbol{x}|\boldsymbol{w},b) = S(\boldsymbol{w}^\top \boldsymbol{x} + b),
\end{equation}

where \( \boldsymbol{x} \) is the input vector, and \( S \) is the sigmoid logistic function. The probability \( p_v(x) \) determines the likelihood of traversing the left branch of the node, while \( 1 - p_v(x) \) represents the probability of taking the right branch.

\paragraph{Soft splits}
Because SDTs do not make hard decisions at each node, they explore multiple paths in the tree, each path being assigned a probability ($p_v$) that reflects how well it matches the input data. The final prediction is derived by weighting the leaf nodes’ outputs by the product of the path probabilities leading to those leaves. 
Therefore, we typically do not have disjoint subsets in the leafs of a DT anymore.
This contrasts with traditional DTs that select a single path based on hard threshold decisions, making SDTs more flexible but potentially less interpretable. \\

\paragraph{Leaf Nodes}
Specifically, each leaf node \( l \) in the tree comprises a probability distribution \( Q_{l,\cdot} \) in contrast to a class membership, which is defined as:

\begin{equation}
Q_{l,c} = \frac{\exp(\lambda_{l,c})}{\sum_{c'=0}^{|\mathcal{C}|-1} \exp(\lambda_{l,c'})}
\end{equation}

for each class $c \in \mathcal{C}$, where \( \lambda_{l} \) are the learned parameters associated with the leaf. The prediction of the tree for an input \( \boldsymbol{x} \) is determined by combining the output distributions of the leaf nodes, weighted by their respective path probabilities.
The probability of reaching a leaf node \( l \) is given by:

\begin{equation}
P_l(\boldsymbol{x}) = \prod_{v \in \text{path to } l} p_v(\boldsymbol{x}),
\end{equation}

where the product is taken over all internal nodes \( v \) along the path from the root to the leaf \( l \). Consequently, the final prediction of the model can be obtained by summing the leaf distributions weighted by their path probabilities as follows:

\begin{equation}
t(\boldsymbol{x}) = \sum_{l \in \text{leaves}} P_l(\boldsymbol{x}) Q_{l,\cdot},
\end{equation}
where \( P_l(\boldsymbol{x}) \) is the probability of reaching leaf \( l \), and \( Q_{l,\cdot} \) is the learned probability distribution at the leaf.
An alternative approach is selecting the leaf with the maximum path probability for higher interpretability, which can be defined as:

\begin{equation}
t(\boldsymbol{x}) = Q_{l^*,\cdot}, \quad \text{where} \quad l^* = \arg\max_{l \in \text{leaves}} P_l(\boldsymbol{x}),
\end{equation}
where, \( l^* \) is the leaf with the highest path probability, and the prediction is simply the distribution \( Q_{l^*,\cdot} \) corresponding to that leaf.
However, this interpretability increase is only local (= on a level of individual instances) and the global interpretability remains unchanged.

\paragraph{SDTs as hierarchical mixture of experts}

\begin{figure}[t]
    \centering
    \includegraphics[width=\textwidth]{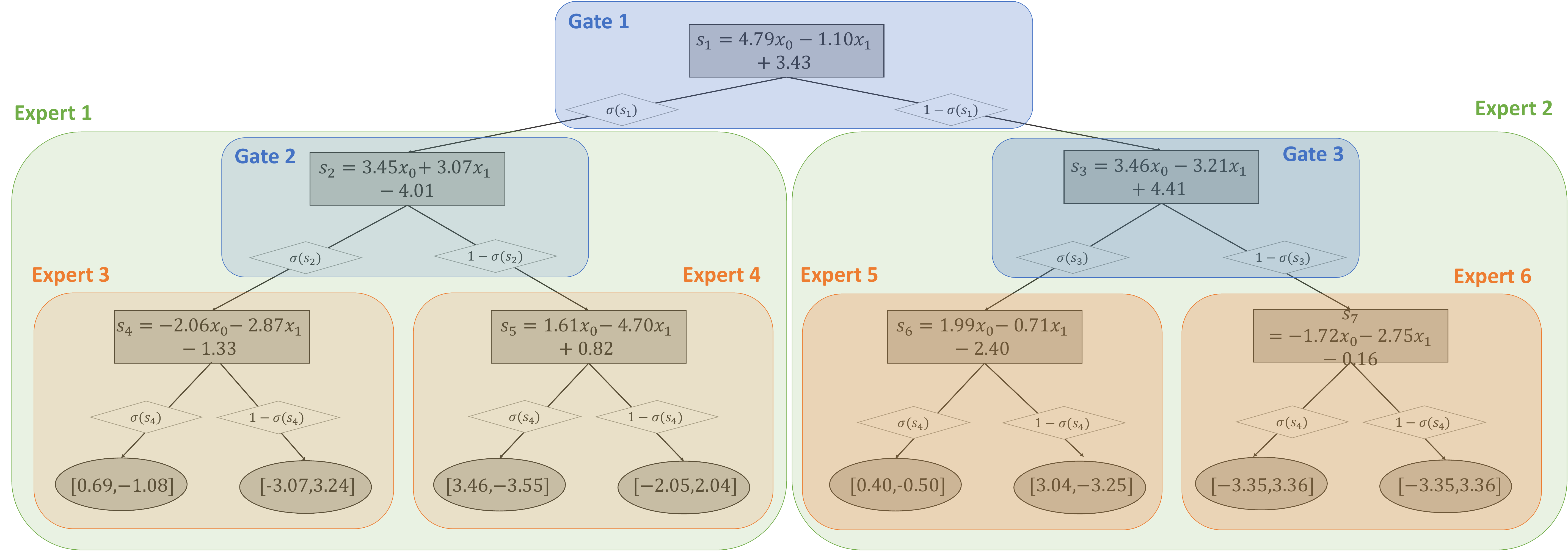}
    \caption[SDT as Mixture of Experts]{\textbf{SDT as Mixture of Experts.} This image visualizes the concept of a soft DT as a hierarchical mixture of expert model.}
    \label{fig:sdt_gating}
\end{figure}

SDTs can be interpreted as a hierarchical, dense mixture of experts model, where each path through the tree represents an expert model, and the soft decisions serve as probabilistic gating mechanisms. This concept is illustrated in Figure~\ref{fig:sdt_gating}. Similarly, one could view standard axis-aligned DTs as a hierarchical top-k mixture of experts with $k=1$.

\paragraph{Summary}
SDTs provide a powerful extension to classical DTs by enabling soft, probabilistic decisions and oblique splits. This makes them suitable for capturing complex patterns in data. However, this only accounts for SDTs as standalone models and is mitigated in ensembling where commonly weak learners are required.
Additionally, their flexibility comes at the cost of interpretability. A comparison of an SDT with a standard, axis-aligned DT is given in Figure~\ref{fig:decision_tree_comparison} showcasing the distinct differences in the interpretability of the two models. Both trees are learned with equal depth on an exemplary titanic dataset, which we will introduce in the following section.
SDTs do not comprise univariate, axis-aligned splits, but are oblique with respect to the axes, as shown in Section~\ref{ssec:oblique splits}.
Therefore, they do not preserve the corresponding inductive bias, which can be beneficial in various domains, especially for tabular data~\citep{grinsztajn2022tree}.
Furthermore, SDTs are considered as less interpretable, especially at the level of individual splits. This is supported by the fact that humans have problems to comprehend explanations involving more than three dimensions at once \citep{molnar2020}, which is often the case when splitting in a SDT.

\section{Gradient Descent Optimization and Backpropagation} \label{sec:gradient_descent_background}

Gradient Descent Optimization and Backpropagation are essential techniques in various optimization problems across different domains. Gradient descent is used to minimize a target function, commonly referred to as a cost or loss function. In addition, backpropagation provides an efficient way to compute gradients necessary for the optimization process. This section provides a detailed explanation of these concepts, focusing on their mathematical foundations and practical applications based on \citet{goodfellow2016deep}. This serves as the basic building block for gradient-based DTs.

\subsection{Gradient Descent Optimization} \label{sec:gradient_descent_optimization}
Gradient descent is an iterative method used to minimize a differentiable function. The function to be minimized, often called the ost function, is a scalar function that quantifies the error or cost associated with a particular set of parameters. The goal is to find the parameter values that minimize this function.

\paragraph{Cost function}
In machine learning, a \emph{cost function} (often referred to as a loss function\footnote{Theoretically, the loss function quantifies the error for individual samples, while the cost function represents the aggregate error over the entire dataset. Nonetheless, these terms are often used interchangeably in practice.}) is a mathematical function that quantifies the difference between the predicted output of a model and the actual target values. The purpose of the cost function is to guide the optimization process by indicating how well the model is performing. The objective is to minimize this cost function, which in turn should improve the model's performance.
For a general model with parameters \( \theta \), input data \( \boldsymbol{X} \), and labels (targets) \( \boldsymbol{y} \), the cost function \( J(\theta) \) can be defined as:

\begin{equation}
J(\theta) = \frac{1}{m} \sum_{j=0}^{m-1} \mathcal{L}(f_\theta(\boldsymbol{x^{(j)}}), y^{(j)}),
\end{equation}

where \( m \) is the number of training examples, \( f_\theta(\boldsymbol{x}^{(j)}) \) is the model's prediction for the \(j\)-th input \( x^{(j)} \) given the parameters \( \theta \), \( y^{(j)} \) is the true label (target) for the \(j\)-th input and \( \mathcal{L} \) is the loss function for the individual example \( j \), which measures the error between the prediction and the true label.
For regression problems, a common cost function is the Mean Squared Error (MSE):

\begin{equation}
J(\theta) = \frac{1}{m} \sum_{j=0}^{m-1} (f_\theta(\boldsymbol{x}^{(j)}) - y^{(j)})^2.
\end{equation}

For binary classification problems, a common cost function is the Binary Cross-Entropy Loss:

\begin{equation}
J(\theta) = - \frac{1}{m} \sum_{j=0}^{m-1} \left( y^{(j)} \log(f_\theta(\boldsymbol{x}^{(j)})) + (1 - y^{(j)}) \log(1 - f_\theta(\boldsymbol{x}^{(j)})) \right).
\end{equation}

In multi-class classification settings with $|\mathcal{C}|$ classes, this loss function generalizes to the Categorical Cross-Entropy Loss, where the sum extends over all classes and predictions are typically obtained using a softmax function.
Minimizing the cost function helps in finding the optimal parameters \( \theta \) that improve the model's performance.

\paragraph{Gradient}
The \emph{gradient} of the cost function \( J(\theta) \) is the vector of partial derivatives of \( J(\theta) \) with respect to each parameter in \( \theta \). Mathematically, the gradient is represented as:

\begin{equation}
\nabla_\theta J(\theta) = \left[ \frac{\partial J(\theta)}{\partial \theta_0}, \frac{\partial J(\theta)}{\partial \theta_1}, \dots, \frac{\partial J(\theta)}{\partial \theta_{p-1}} \right],
\end{equation}

where \( \theta = [\theta_0, \theta_1, \dots, \theta_{p-1}] \) represents the parameters of the model, and each component \( \frac{\partial J(\theta)}{\partial \theta_i} \) is the partial derivative of the cost function \( J(\theta) \) with respect to the corresponding parameter \( \theta_i \), for \( i = 0, 1, \dots, p-1 \).
The gradient indicates the direction of the steepest ascent of the cost function. To minimize \( J(\theta) \), we need to move in the opposite direction of the gradient.

\paragraph{Gradient descent update rule}
The gradient descent update rule adjusts the parameters \( \theta \) in the direction that reduces the cost function. In the following, the superscript \( t \) denotes the iteration index of the algorithm, indicating the discrete step at which the parameters are evaluated. For example, \( \theta^{(t)} \) represents the parameter values at the current iteration, while \( \theta^{(t+1)} \) represents the values after the update. The update rule for each parameter \( \theta_i \) is given by:

\begin{equation}
\theta_{i}^{(t+1)} = \theta_i^{(t)} - \eta \frac{\partial J(\theta)}{\partial \theta_i^{(t)}},
\end{equation}

where  \( \eta \) is the learning rate, which controls the step size of each update and \( \frac{\partial J(\theta)}{\partial \theta_i} \) is the partial derivative (gradient) of the cost function with respect to \( \theta_i \), representing the slope of the cost function in relation to that parameter.
The learning rate controls the step size of the parameter updates during optimization. A properly chosen learning rate ensures that the optimization process converges efficiently. However, using a fixed learning rate can be suboptimal, especially when different parameters require different step sizes.

\subsection{Backpropagation} \label{sec:backpropagation}

Backpropagation is a fundamental algorithm used to compute the gradients of the cost function with respect to the model's parameters in neural networks. These gradients are essential for updating the model parameters during the training process using gradient descent. The core idea behind backpropagation is to apply the \emph{chain rule} of calculus to compute the gradient of the cost function layer by layer, starting from the output layer and working backward to the input layer. This method is computationally efficient and allows for the training of deep neural networks.

\paragraph{The chain rule and layer-wise gradient calculation}

Consider a neural network composed of \( L \) layers, where the input to the network is denoted as \( \boldsymbol{x} \), and the output at layer \( l \) is denoted as 
\begin{align}
\boldsymbol{z}_l &= \boldsymbol{W}_l \boldsymbol{a}_{l-1} + \boldsymbol{b}_l, \\
\boldsymbol{a}_l &= \phi\big(\boldsymbol{z}_l\big),
\end{align}
where \( \boldsymbol{z}_l \) represents the pre-activation values, $\phi$ represents an activation function, \( \boldsymbol{a}_l \) is the activated output of layer \( l \), \( \boldsymbol{W}_l \) and \( \boldsymbol{b}_l \) are the weight matrix and bias vector of the \( l \)-th layer, respectively, and \( \boldsymbol{a}_{l-1} \) is the activation from the previous layer.
The goal of backpropagation is to compute the derivative of the cost function \( J(\theta) \) with respect to each parameter in layer $l$, i.e., \( \theta_l = \{ \boldsymbol{W}_l, \boldsymbol{b}_l \} \). To do this, we propagate the error backward through the network, updating each layer using the chain rule. The error \( \delta_l \) at layer \( l \) is defined as the derivative of the cost function with respect to the pre-activation values \( \boldsymbol{z}_l \):

\begin{equation}
\delta_l = \frac{\partial J}{\partial \boldsymbol{z}_l}.
\end{equation}

Using the chain rule, the gradient of the cost function with respect to the weights \( \boldsymbol{W}_l \) and biases \( \boldsymbol{b}_l \) can be calculated as:

\begin{equation}
\frac{\partial J}{\partial \boldsymbol{W}_l} = \delta_l (\boldsymbol{a}_{l-1})^T,
\end{equation}
\begin{equation}
\frac{\partial J}{\partial \boldsymbol{b}_l} = \delta_l.
\end{equation}

The backpropagation algorithm proceeds as follows:
\begin{enumerate}
    \item Compute the error at the output layer \( \delta_L = \nabla_{\boldsymbol{z}_L} J \), where \( \nabla_{\boldsymbol{z}_L} J \) is the gradient of the cost function with respect to the output layer activations.
    \item Propagate the error backward to the previous layers using the formula:
    \begin{equation}
    \delta_l = (\boldsymbol{W}_{l+1})^T \delta_{l+1} \odot \phi'(\boldsymbol{z}_l),
    \end{equation}
    where \( \odot \) denotes the element-wise product (Hadamard product), and \( \phi'(\boldsymbol{z}_l) \) is the derivative of the activation function at layer \( l \).
    \item Compute the gradients of the weights and biases using the above equations for each layer.
\end{enumerate}

\paragraph{Computational efficiency and complexity}

Backpropagation is computationally efficient because it reuses intermediate computations, such as the activations and gradients, during the backward pass. The time complexity of backpropagation is proportional to the number of parameters in the network, as each parameter needs to be updated during training. For a fully connected network, the computational complexity is \( O(m \times p) \), where \( m \) is the number of training examples and \( p \) is the total number of parameters in the network.

\paragraph{Activation functions}
However, the efficiency of backpropagation depends on factors such as the choice of activation functions and the architecture of the neural network. For instance, certain activation functions, such as the sigmoid function, can cause problems like the vanishing gradient problem, which hinders the learning of deep networks. Modern architectures often employ activation functions like ReLU (Rectified Linear Unit) to mitigate these issues and improve both the speed and stability of training.

\paragraph{Mini-batch gradient descent}
Additionally, mini-batch gradient descent is commonly used in practice. Instead of computing the gradients for the entire dataset at once (which can be computationally expensive), mini-batch gradient descent computes the gradients on smaller subsets (batches) of the data. This reduces the amount of computation and can take advantage of parallelism on modern hardware (e.g., GPUs).

\paragraph{Automatic differentiation}
In practice, backpropagation is implemented with automatic differentiation libraries, which further optimize the computation by maintaining a computational graph and automatically applying the chain rule to compute gradients for arbitrary model architectures. Furthermore, modern computation frameworks vectorize computations by using matrix operations (e.g., matrix multiplications) rather than loops to compute the forward and backward passes, which allows leveraging optimized linear algebra libraries (e.g., BLAS, cuBLAS).

\subsection{Advanced Techniques in Gradient Descent} \label{sec:advancements_gradient_descent}

In the following, we will introduce several adjustments from the literature that are commonly used when learning a model with gradient descent. These adjustments are also relevant for the method introduced in this thesis and are a major factor for the success of the method (see Section~\ref{ssec:gradient_descent_tree}).

\subsubsection*{Adam Optimizer}

The Adam (Adaptive Moment Estimation) optimizer~\citep{kingma2014adam} is an advanced optimization algorithm that combines the benefits of two other popular methods: Momentum and RMSprop~\citep{tieleman2012rmsprop}. Adam is widely used in deep learning due to its ability to adapt the learning rate individually for each parameter, leading to faster convergence and better performance on complex problems.

\paragraph{Momentum}

Momentum is a technique designed to accelerate gradient descent by considering the past gradients in the update process. It helps in smoothing the optimization path and preventing oscillations. The update rule with momentum is:

\begin{equation}
\mu^{(t+1)} = \beta_1 \mu^{(t)} + (1 - \beta_1) \nabla_{\theta^{(t)}} J(\theta^{(t)}),
\end{equation}
\begin{equation}
\theta^{(t+1)} = \theta^{(t)} - \eta \mu^{(t+1)},
\end{equation}

where \( \mu^{(t)} \) is the exponentially decaying average of past gradients (also called velocity) and \( \beta_1 \) is the momentum coefficient.

\paragraph{Adaptive learning rates (RMSprop)}

RMSprop is an adaptive learning rate method that adjusts the learning rate for each parameter based on the magnitude of its gradient. This approach helps to mitigate issues with vanishing or exploding gradients by scaling the learning rate accordingly. The update rule for RMSprop is:

\begin{equation}
\nu^{(t+1)} = \beta_2 \nu^{(t)} + (1 - \beta_2) (\nabla J(\theta^{(t)}))^2,
\end{equation}
\begin{equation}
\theta^{(t+1)} = \theta^{(t)} - \frac{\eta}{\sqrt{\nu^{(t+1)}} + \epsilon} \nabla_{\theta^{(t)}} J(\theta^{(t)}),
\end{equation}

where \( \nu^{(t)} \) is the exponentially decaying average of squared gradients, \( \beta \) is the decay rate and \( \epsilon \) is a small constant (e.g., \( 10^{-8} \)) added for numerical stability.

\paragraph{Adam optimizer}

The Adam optimizer combines the advantages of both Momentum and RMSprop by computing adaptive learning rates for each parameter while maintaining momentum. The update rules for Adam are as follows:

\begin{enumerate}
    \item \textbf{Compute biased first moment estimate (Momentum):}
    \[
    \mu^{(t+1)} = \beta_1 \mu^{(t)} + (1 - \beta_1) \nabla_{\theta^{(t)}} J(\theta^{(t)})
    \]
    \item \textbf{Compute biased second moment estimate (RMSprop-like):}
    \[
    \nu^{(t+1)} = \beta_2 \nu^{(t)} + (1 - \beta_2) (\nabla_{\theta^{(t)}} J(\theta^{(t)}))^2
    \]
    \item \textbf{Bias correction:}
    \[
    \hat{\mu}^{(t+1)} = \frac{\mu^{(t+1)}}{1 - \beta_1^{(t+1)}}, \quad \hat{\nu}^{(t+1)} = \frac{\nu^{(t+1)}}{1 - \beta_2^{(t+1)}}
    \]
    \item \textbf{Update parameters:}
    \[
    \theta^{(t+1)} = \theta^{(t)} - \frac{\eta}{\sqrt{\hat{\nu}^{(t+1)}} + \epsilon} \hat{\mu}^{(t+1)}
    \]
\end{enumerate}

where \( \mu^{(t)} \) and \( \nu^{(t)} \) are the first and second moment estimates, \( \beta_1 \) and \( \beta_2 \) are decay rates for the first and second moment estimates and \( \epsilon \) is a small constant added for numerical stability.

\paragraph{Stochastic Weight Averaging}

Stochastic Weight Averaging (SWA) is an optimization technique that improves the generalization performance of deep learning models~\citep{izmailov2018averaging}. Instead of relying on a single set of weights obtained at the end of training, SWA maintains a running average of weights collected from different points along the training trajectory. Formally, if \( \theta^{(t)} \) represents the weights at iteration \( t \), the SWA weights \( \theta_{\text{SWA}} \) are computed as:

\begin{equation}
\theta_{\text{SWA}} = \frac{1}{k} \sum_{i=0}^{k-1} \theta^{(t_i)},
\end{equation}

where \( t_i \) denotes the iteration at which the weights are sampled, and \( k \) is the number of sampled points. By averaging weights from different stages of training, SWA helps smooth out sharp minima, leading to a flatter and more generalizable solution. 

\paragraph{Weight decay and decoupled weight decay} \label{sec:weight_dacay}
Weight Decay~\citep{hanson1988comparing, loshchilov2017decoupled} is a regularization technique used to prevent overfitting in machine learning models by penalizing large weights. The idea is to add a penalty term to the cost function that encourages the model to prefer smaller weights, thereby reducing the model's complexity.
Decoupled weight decay~\citep{loshchilov2017decoupled} modifies how weight decay is applied during optimization, addressing the limitations of standard weight decay~\citep{hanson1988comparing}, or L$_2$ regularization, when used with adaptive optimizers like Adam. 
In standard weight decay, the regularization term is included in the gradient computation and thus gets scaled by the adaptive learning rates \(\frac{\eta}{\sqrt{\hat{\nu}^{(t+1)}} + \epsilon}\), which vary across parameters and over time. This scaling can lead to inconsistent regularization effects, with parameters that have larger adaptive learning rates receiving less regularization. Decoupled weight decay addresses this issue by applying the weight decay term directly to the parameters using the global learning rate \(\eta\), independent of the adaptive learning rates. This ensures consistent regularization across all parameters, preserving the advantages of adaptive optimizers while maintaining effective regularization. Consequently, decoupled weight decay leads to improved generalization performance and is compatible with adaptive methods.
Decoupled weight decay has become the most commonly used form, and nowadays is often referred to simply as weight decay.

\chapter{Learning Algorithms for Decision Trees} \label{cha:related}

\textbf{The following section was already partially published in \citet{gradtree}. For a detailed summary of the individual contributions, please refer to Section~\ref{sec:contributions}.} \\

\noindent In the previous section, we introduced DTs along with their distinct characteristics and variants (Section~\ref{sec:decision_trees_background}). 
As DT learning is the core component of the method proposed in this thesis, it is highly relevant to understand and differentiate existing methods on this topic, which is the focus of this chapter. Besides the most commonly used, greedy procedure (Section~\ref{sec:greedy_dt}), the methods introduced in the following include lookahead DTs (Section~\ref{sec:lookahead_dt}), evolutionary algorithms for DT learning (Section~\ref{sec:evolutionary_dt}) and optimal DTs (Section~\ref{sec:optimal_dt}). Furthermore, we will introduce existing gradient-based methods for DT learning (Section~\ref{sec:gradient_dt}) providing greater flexibility, but usually focusing on soft DTs with oblique splits. 
In this context, we will critically examine the limitations of existing approaches and identify a research gap (Section~\ref{sec:summary_related}) that is specifically addressed by the method proposed in this thesis (Chapter~\ref{cha:methodology}).
Please note, that we focus explicitly on methods for learning individual trees and do not give a detailed overview on ensembling methods like gradient-boosted DTs, which are commonly used to achieve state-of-the-art performance with DTs at the cost of interpretability. The main reason for this decision is the fact that ensembling methods are usually method-agnostic and can be applied to arbitrary models as weak learners.

\paragraph{Decision tree learning study} In addition to explaining the relevant algorithms, we will give an example for
each algorithm based on a two-dimensional version of a Titanic dataset with 20 samples that will be introduced in the following section (Section~\ref{ssec:2D_titanic}). In this context, we illustrate how greedy approaches can get stuck in local optima, and demonstrate how alternative methods address this constraint by mitigating the impact of local optimality. 
Please note that this dataset was intentionally designed to showcase differences in the results of the different algorithms. While this is necessary for a good comparison, we want to clarify that this is not necessarily the case in all real-world scenarios. Also note that in the following, we focus on how well a DT can fit the (training) data and do not focus on the generalization of the learned model in this example. Nonetheless, we will discuss the generalization of the different approaches and come back to this during the evaluation on more complex real-world datasets. Please note that we limit the depth of each algorithm to $3$ for this experiment to maintain a model that is easy to understand and compare.

\section{Machine Learning from Disaster: A 2D Dataset for Comparing Different Decision Tree Learning Algorithms} \label{ssec:2D_titanic}

\paragraph{Dataset definition}
Let \( \boldsymbol{x} \in \mathbb{R}^n \) be a sample from the feature space \( \mathcal{X} \), and let \( y \in \mathcal{Y} \) be the corresponding label, where \( \mathcal{X} \subseteq \mathbb{R}^n \) and \( \mathcal{Y} \subseteq \mathbb{R}^k \). Further, we define a dataset of input-output pairs \( \mathcal{D} \) as

\[
    \mathcal{D} = \left\{ (\boldsymbol{x}^{(j)}, y^{(j)}) \mid \boldsymbol{x}^{(j)} \in \mathcal{X}, y^{(j)} \in \mathcal{Y}, j = 0, 1, \dots, m-1 \right\}
\]

where \( \boldsymbol{x}^{(j)} \) represents the feature vector of the \( j \)-th sample, and \( y^{(j)} \in \mathcal{Y} \) is the corresponding label or target value. The set \( \mathcal{Y} \subseteq \mathbb{R}^k \) defines the output space, which can either be continuous (in the case of regression) or discrete (in the case of classification).

\paragraph{Synthetic dataset generation}

We generated a dataset containing two continuous features, \emph{Fare} and \emph{Age} to predict the survival of a passenger. 
To create a synthetic dataset for analysis, a function was developed to generate a simple two-dimensional dataset consisting of the features \textit{Passenger ID}, \textit{Fare}, \textit{Age}, and a label \textit{Survived}. The generation process can be summarized as follows:

\begin{enumerate}
    \item \textbf{Passenger Identification (ID):} Each instance is assigned a unique identifier, 
    \[
    \text{Passenger ID} \in \{1, 2, \dots, m\},
    \]
    where \( M \) represents the total number of samples in the dataset.

    \item \textbf{Fare:} Passenger fares are randomly drawn from a uniform distribution between 1 and 250, i.e.,
    \[
    \text{Age} \sim U(1, 250).
    \]

    \item \textbf{Age:} Passenger ages are randomly drawn from a uniform distribution between 1 and 90, i.e.,
    \[
    \text{Age} \sim U(1, 90).
    \]

    \item \textbf{Survival Label (Skewed):} The label \textit{Survived}, indicating whether a passenger survived, is generated based on the following conditions:
    \[
    \text{P(Survived} = \text{`Yes'} \mid \text{Age} > 20 \text{ or Fare} \leq 50) = 0.2,
    \]
    \[
    \text{P(Survived} = \text{`Yes'} \mid \text{Age} \leq 20 \text{ and Fare} > 50) = 0.6.
    \]
    This introduces a probabilistic skew in the survival label, influenced by a passenger’s age and fare.
\end{enumerate}

It is important to emphasize that the primary goal in designing this dataset was not to achieve the most realistic representation, but rather to create a dataset that is straightforward, interpretable, and suitable for illustrating the functioning of DTs. To this end, we intentionally sampled \textit{Age} and \textit{Fare} from a uniform distribution, allowing precise control over the parameter ranges. Although alternative distributions could yield a more realistic dataset, this approach effectively fulfills its intended explanatory purpose.

\paragraph{Duplicate handling} To ensure data variability, the algorithm checks for and removes any duplicates in the combinations of the \textit{Fare} and \textit{Age} features. If duplicates are found, the dataset is regenerated.

\paragraph{Dataset versions}
We binarized the fare attribute for the algorithm comparison to ensure that the dataset includes a combination of continuous and categorical features. Some methods only allow categorical (or binary) features and cannot deal with continuous features of the box. In this case, we also binarized the age attribute. 
The binarization was applied as follows:
\begin{itemize}
    \item \textbf{Fare Category}: Low (0--50), High (50+)
    \item \textbf{Age Category}: Young (0--21), Old (21+)
\end{itemize}
Based on this, we generated three datasets as summarized in Tables~\ref{tab:titanic_2d_num}-\ref{tab:titanic_2d_num_cat}\footnote{In practice, two alternative heterogeneous versions could be generated: One by discretizing the fare and another by discretizing the age. For simplicity, we chose to use only a single heterogeneous version, obtained by discretizing the fare.}. We will use this dataset throughout this section to demonstrate the characteristics of different DT types and algorithms.

\begin{table}[H]
\centering
\caption[Exemplary 2D Titanic Datasets]{\textbf{Exemplary 2D Titanic Datasets.}}

\subcaptionbox{\textbf{Numeric Version.} This is a version of the 2D Titanic Dataset with two numeric features.\label{tab:titanic_2d_num}}[.32\linewidth]{%
\small
\begin{tabular}{|R{0.3cm}|R{0.7cm}|R{0.6cm}|C{1.0cm}|}
\hline
\textbf{ID} & \textbf{Fare (N)} & \textbf{Age (N)} & \textbf{Sur- vived} \\
\hline
1  & 34  & 61  & \text{No}  \\
2  & 126 & 45  & \text{Yes} \\
3  & 234 & 29  & \text{No}  \\
4  & 12  & 10  & \text{Yes} \\
5  & 68  & 32  & \text{No}  \\
6  & 36  & 69  & \text{No}  \\
7  & 12  & 68  & \text{No}  \\
8  & 112 & 46  & \text{Yes} \\
9  & 79  & 57  & \text{No}  \\
10 & 199 & 25  & \text{No}  \\
11 & 39  & 13  & \text{Yes} \\
12 & 49  & 24  & \text{No}  \\
13 & 12  & 11  & \text{Yes} \\
14 & 48  & 83  & \text{Yes} \\
15 & 136 & 6   & \text{No}  \\
16 & 24  & 80  & \text{No}  \\
17 & 199 & 55  & \text{Yes} \\
18 & 8   & 45  & \text{No}  \\
19 & 26  & 26  & \text{No}  \\
20 & 62  & 74  & \text{No}  \\
\hline
\end{tabular}
}
\hfill
\subcaptionbox{\textbf{Categorical Version.} This is a version of the 2D Titanic Dataset with both features discretized to binary features.\label{tab:titanic_2d_cat}}[.32\linewidth]{%
\small
\begin{tabular}{|R{0.3cm}|C{0.7cm}|C{0.9cm}|C{1.0cm}|}
\hline
\textbf{ID} & \textbf{Fare (C)} & \textbf{Age (C)} & \textbf{Sur- vived} \\
\hline
1  & \text{Low}   & \text{Old}  & \text{No}  \\
2  & \text{High}  & \text{Old}  & \text{Yes} \\
3  & \text{High}  & \text{Old}  & \text{No}  \\
4  & \text{Low}   & \text{Young}  & \text{Yes} \\
5  & \text{High}  & \text{Old}  & \text{No}  \\
6  & \text{Low}   & \text{Old}  & \text{No}  \\
7  & \text{Low}   & \text{Old}  & \text{No}  \\
8  & \text{High}  & \text{Old}  & \text{Yes} \\
9  & \text{High}  & \text{Old}  & \text{No}  \\
10 & \text{High}  & \text{Old}  & \text{No}  \\
11 & \text{Low}   & \text{Young}  & \text{Yes} \\
12 & \text{Low}   & \text{Old}  & \text{No}  \\
13 & \text{Low}   & \text{Young}  & \text{Yes} \\
14 & \text{Low}  & \text{Old}  & \text{Yes} \\
15 & \text{High}   & \text{Young}   & \text{No}  \\
16 & \text{Low}  & \text{Old}  & \text{No}  \\
17 & \text{Low}   & \text{Old}  & \text{Yes} \\
18 & \text{High}   & \text{Old}  & \text{No}  \\
19 & \text{Low}  & \text{Old}  & \text{No}  \\
20 & \text{High}   & \text{Old}  & \text{No}  \\
\hline
\end{tabular}
}
\hfill
\subcaptionbox{\textbf{Heterogeneous Version.} This is a version of the 2D Titanic Dataset with one numeric and one discretized feature.\label{tab:titanic_2d_num_cat}}[.32\linewidth]{%
\small
\begin{tabular}{|R{0.3cm}|C{0.7cm}|R{0.6cm}|C{1.0cm}|}
\hline
\textbf{ID} & \textbf{Fare (C)} & \textbf{Age (N)} & \textbf{Sur- vived} \\
\hline
1  & \text{Low}   & 61  & \text{No}  \\
2  & \text{High}  & 45  & \text{Yes} \\
3  & \text{High}  & 29  & \text{No}  \\
4  & \text{Low}   & 10  & \text{Yes} \\
5  & \text{High}  & 32  & \text{No}  \\
6  & \text{Low}   & 69  & \text{No}  \\
7  & \text{Low}   & 68  & \text{No}  \\
8  & \text{High}  & 46  & \text{Yes} \\
9  & \text{High}  & 57  & \text{No}  \\
10 & \text{High}  & 25  & \text{No}  \\
11 & \text{Low}   & 13  & \text{Yes} \\
12 & \text{Low}   & 24  & \text{No}  \\
13 & \text{Low}   & 11  & \text{Yes} \\
14 & \text{Low}  & 83  & \text{Yes} \\
15 & \text{High}   & 6   & \text{No}  \\
16 & \text{Low}  & 80  & \text{No}  \\
17 & \text{High}   & 55  & \text{Yes} \\
18 & \text{Low}   & 45  & \text{No}  \\
19 & \text{Low}  & 26  & \text{No}  \\
20 & \text{High}   & 74  & \text{No}  \\
\hline
\end{tabular}
}

\end{table}

\section{Greedy, Purity-Based Decision Tree Induction} \label{sec:greedy_dt}

Starting with the publication of the first DT learning algorithm in 1963~\citep{morgan1963problems} the task of learning a DT from data remained well-researched until today. 
The most prominent and still frequently used DT learning algorithms, namely C4.5~\citep{quinlan1993c45} (Section~\ref{sec:c45}) as an extension of ID3.0~\citep{quinlan1986_id30} (Section~\ref{sec:id3}) and CART~\citep{cart_breiman1984} (Section~\ref{sec:cart}), date back to the 1980s. %
They all follow a greedy procedure to learn a DT and build onto \emph{Hunt's algorithm} as a basis for DT induction~\citep{hunt1966experiments}. 
While algorithms like Hunt's algorithm, ID3, and C4.5 support non-binary splits, i.e., more than two branches from a split node, we restrict our discussion to binary DTs for simplicity and consistency with subsequent algorithms. 
Although some algorithms are specifically designed to support regression tasks, all approaches can, in principle, be adapted for regression with varying degrees of modification. For clarity, the following discussion will focus on classification.
In general, purity-based DT induction relies on an impurity measure \( I \) to determine the next split. The choice of impurity measure significantly influences the split-selection process and varies across methods.
In this work, we focus on CART (Section~\ref{sec:cart}) as the most commonly employed greedy method, also widely implemented in libraries like scikit-learn~\citep{sklearn_api}. We will walk through the first iteration of the DT induction process to clearly explain the role of the impurity measure and the inherent local optimality issues of the sequential, greedy procedure. A more detailed overview and comparison of existing greedy methods, including extensions like GUIDE and C5.0, is provided in Appendix~\ref{A:greedy_dt}.

\paragraph{Algorithm overview} 
Greedy DT learners recursively partition the training data into increasingly homogeneous subsets with respect to the target class. Algorithm~\ref{alg:cart} summarizes the core steps of such an algorithm, which forms the basis of CART. The primary differences among greedy DT learners (e.g., CART vs. C4.5) lie in their choice of splitting criteria, handling of categorical variables, and various algorithm-specific enhancements (such as pruning methods). For a detailed comparison, refer to Appendix~\ref{sec:greedy_dt_comparison}.

\begin{algorithm}[t]
\caption{Greedy Decision Tree Algorithm}
\label{alg:cart}
\begin{algorithmic}[1]
\Function{GreedyDT}{$\mathcal{D}$}
    \If{StoppingCriteria is met}
        \State Compute output \(\lambda\) as the majority class (classification) or mean (regression) of \(\{y : (\boldsymbol{x},y) \in \mathcal{D}\}\)
        \State \Return Leaf node \(v_l\) with region \(R_l = \mathcal{D}\) and output \(\lambda_l = \lambda\)
    \EndIf
    \State \textbf{Determine} best split parameters \((\iota^*, \tau^*)\) by minimizing an impurity loss
    \State \textbf{Partition} the dataset $\mathcal{D}$:
    \[
        \mathcal{D}_L = \{\boldsymbol{x} \in \mathcal{D} \mid x_{\iota^*} \geq \tau^*\}, \quad \mathcal{D}_R = \{\boldsymbol{x} \in \mathcal{D} \mid x_{\iota^*} < \tau^*\}
    \]
    \State \textbf{Create} an internal node \(v_i\) with the decision rule:
    \[
    S_{v_i}(\boldsymbol{x} \mid \iota^*,\tau^*) =
    \begin{cases}
    \text{index}(v_{\text{left}}) & \text{if } x_{\iota^*} \geq \tau^*, \\[1mm]
    \text{index}(v_{\text{right}}) & \text{if } x_{\iota^*} < \tau^*.
    \end{cases}
    \]
    \State Set \(v_{\text{left}} \gets \) \Call{GreedyDT}{$\mathcal{D}_L$}
    \State Set \(v_{\text{right}} \gets \) \Call{GreedyDT}{$\mathcal{D}_R$}
    \State \Return \(v_i\)
\EndFunction
\end{algorithmic}
\end{algorithm}

\paragraph{Stopping criteria} In the context of DT learning, stopping criteria can be defined as follows:
\begin{itemize} 
\item All leaf nodes are homogeneous.
\item No instances remain.
\item No further splits are possible.
\item Any possible split would increase impurity\footnote{This special case is not explicitly discussed in the literature (except for a corresponding hyperparameter), but selecting a split that increases impurity would contradict the motivation behind using an impurity measure.}.
\item The maximum recursion depth is reached (optional).
\item An optional, user-specified stopping criterion is met (e.g., a minimum number of samples required for a leaf).
\end{itemize}

\paragraph{Complexity} The complexity of learning a binary DT using these greedy algorithms depends primarily on the number of data points ($m$), the number of features ($n$), and the depth ($d$). In the best case of a balanced tree, $d = \log L$ for a tree with $L$ leaf nodes. In the worst case of an unbalanced tree, this can go up to $d = L$.
In general, the time complexity of learning a binary DT is 
\[O(n \cdot m^2 \cdot d)\] 
but can be reduced to 
\[O(n \cdot m \log m \cdot d)\] 
if continuous features are being sorted. Using caching tricks to keep track of sorting results, the time complexity can be reduced further to \[O(n \cdot m \log m)\] but at the cost of a higher space complexity for storing the complete data.

\subsection{Classification and Regression Trees (CART)} \label{sec:cart}

The CART (Classification and Regression Trees) algorithm is a robust DT learning technique introduced by \citet{breiman1984classification}, used for both, classification and regression tasks. Unlike other tree-based methods like ID3 and C4.5, CART produces only binary trees by recursively splitting the data at each node into two child nodes. The splits are chosen to maximize the purity of the resulting nodes, and the tree can be pruned using cross-validation to avoid overfitting.

\paragraph{Impurity measures}

To determine the best split at each node, CART uses different impurity measures based on the type of task:
\begin{itemize}
    
    \item \textbf{Classification:} For classification tasks, CART uses the Gini impurity to quantify the homogeneity of the target variable within the subsets created by the split.
  
       The Gini impurity at node \( v \) is defined as:

      \begin{equation} \label{eq:gini}
      I_{\text{Gini}}(v) = 1 - \sum_{c=0}^{|\mathcal{C}|-1} p(c|v)^2,
      \end{equation}
    
      where \( p(c|v) \) is the proportion of class \( c \) instances at node \( v \), and \( |\mathcal{C}| \) is the total number of classes.
    
    \item \textbf{Regression:} For regression tasks, CART uses the variance (mean squared error) of the target variable to evaluate potential splits.

      The variance impurity at node \( v \) is defined as:
    
      \begin{equation} \label{eq:variance}
      I_{\text{Var}}(v) = \frac{1}{m_{v}} \sum_{i=0}^{m_{v}-1} (y_i - \bar{y}_v)^2,
      \end{equation}
    
      where \( y_i \) is the target value of the \( i \)-th instance at node \( v \), \( \bar{y}_v \) is the mean of the target values at node \( v \), and \( m_v \) is the number of instances at node \( v \).
    
\end{itemize}

\paragraph{Split selection}
The algorithm evaluates all possible splits at each node and selects the split that maximizes the reduction in impurity. For a node \( v \), if a split \( \mathbb{S}_v \) divides it into two child nodes \( v_{\text{left}} \) and \( v_{\text{right}} \), the decrease in Gini impurity due to the split is given by:

\begin{equation} \label{eq:gini_decrease}
\Delta I_{\text{Gini}}(v | \mathbb{S}_v) = I_{\text{Gini}}(v) - \left( \frac{m_{v,\text{left}}}{m_v} \, I_{\text{Gini}}(v_{\text{left}}) + \frac{m_{v,\text{right}}}{m_v} \, I_{\text{Gini}}(v_{\text{right}}) \right),
\end{equation}

where \( I_{\text{Gini}}(v) \) is the impurity of node \( v \), \( I_{\text{Gini}}(v_{\text{left}}) \) and \( I_{\text{Gini}}(v_{\text{right}}) \) are the impurities of the left and right child nodes after the split \( \mathbb{S}_v \) and \( m_{v,\text{left}} \) and \( m_{v,\text{right}} \) are the numbers of instances in the left and right child nodes, respectively.
The goal is to find the split \( \mathbb{S}_v \) that maximizes \( \Delta I_{\text{Gini}}(\mathbb{S}_v, v)  \), thereby achieving the greatest reduction in impurity and resulting in more homogeneous child nodes.

\subsubsection*{Example 1: Constructing a Decision Tree using CART}

Again, we use the simplified version of the Titanic dataset from Table~\ref{tab:titanic_2d_num_cat} including the age as continuous and the fare as categorical attribute.
In the following, we provide a step-by-step example for constructing the root node of the DT using the Gini impurity criterion.

\begin{enumerate}
    \item \textbf{Calculate the Gini impurity for the whole dataset} \hspace{0.25cm}
    The initial Gini impurity \(I_{\text{Gini}}\) is calculated based on the proportion of passengers who survived (Yes) and who did not survive (No):
    
    20 passengers, 7 Survived (Yes), 13 Did Not Survive (No)
    
    \begin{align*}   
    \small
    I_{\text{Gini}}(v_0) = 1 - \left(\frac{7}{20}\right)^2 - \left(\frac{13}{20}\right)^2 = 0.455
    \quad \triangleright \text{Eq.~\ref{eq:gini}} 
    \end{align*}   
    
    \textbf{Calculate decrease in Gini impurity for each potential split}

        \begin{table}[tb]
            \centering
            \small
            \begin{tabular}{c|cccc}
                \toprule
                \textbf{Threshold} & \textbf{Left Gini} & \textbf{Right Gini} & \textbf{Weighted Gini} & \textbf{Gini Gain} \\ 
                \midrule
                Fare Category    & 0.463 & 0.444 & 0.455 & 0.000 \\
                Age $\geq$ \phantom{0}8.0   & 0.000 & 0.465 & 0.442 & 0.013 \\
                Age $\geq$ 10.5  & 0.500 & 0.444 & 0.450 & 0.005 \\
                Age $\geq$ 12.0  & 0.444 & 0.415 & 0.420 & 0.035 \\
                Age $\geq$ 18.5  & 0.375 & 0.375 & 0.375 & \bftab 0.080 \\
                Age $\geq$ 24.5  & 0.480 & 0.391 & 0.413 & 0.042 \\
                Age $\geq$ 25.5  & 0.500 & 0.408 & 0.436 & 0.019 \\
                Age $\geq$ 27.5  & 0.490 & 0.426 & 0.448 & 0.007 \\
                Age $\geq$ 30.5  & 0.469 & 0.444 & 0.454 & 0.001 \\
                Age $\geq$ 38.5  & 0.444 & 0.463 & 0.455 & 0.000 \\
                Age $\geq$ 45.0  & 0.463 & 0.444 & 0.455 & 0.000 \\
                Age $\geq$ 45.5  & 0.463 & 0.444 & 0.455 & 0.000 \\
                Age $\geq$ 50.5  & 0.486 & 0.375 & 0.442 & 0.013 \\
                Age $\geq$ 56.0  & 0.497 & 0.245 & 0.409 & 0.046 \\
                Age $\geq$ 59.0  & 0.490 & 0.278 & 0.426 & 0.029 \\
                Age $\geq$ 64.5  & 0.480 & 0.320 & 0.440 & 0.015 \\
                Age $\geq$ 68.5  & 0.469 & 0.375 & 0.450 & 0.005 \\
                Age $\geq$ 71.5  & 0.457 & 0.444 & 0.455 & 0.000 \\
                Age $\geq$ 77.0  & 0.444 & 0.500 & 0.450 & 0.005 \\
                Age $\geq$ 81.5  & 0.432 & 0.000 & 0.411 & 0.044 \\
                \bottomrule
            \end{tabular}
            \caption[Split Summary CART]{\textbf{Split Summary CART.} Gini Values and Gini Gain for each possible split of the root node.}
            \label{tab:gini_values_table}
        \end{table}
        
        \textbf{Gini Impurity for \emph{Age} Attribute} \hspace{0.25cm}
        We evaluate possible thresholds for the \emph{Age} attribute. Sorted values for \emph{Age} are 6, 10, 11, 13, 24, 25, 26, 29, 32, 45, 45, 46, 55, 57, 61, 68, 69, 74, 80, 83. We consider thresholds at the midpoints:
        
        \textbf{Threshold = 18.5}:
        \begin{itemize}
            \item \textbf{Age $\geq$ 18.5}: 4 passengers, 3 Survived (Yes), 1 Did not survive (No)
        \end{itemize}        
        \begin{align*}   
        \small
        I_{\text{Gini}}(v_{0,\text{left}}) = 1 - \left(\frac{3}{4}\right)^2 - \left(\frac{1}{4}\right)^2 = 0.375
        \quad \triangleright \text{Eq.~\ref{eq:gini}} 
        \end{align*}   
        
        \begin{itemize}
            \item \textbf{Age < 18.5}: 16 passenger, 4 Survived (Yes), 12 Did not survive (No)
        \end{itemize}
        \begin{align*}   
        \small
        I_{\text{Gini}}(v_{0,\text{right}}) = 1 - \left(\frac{4}{16}\right)^2 - \left(\frac{12}{16}\right)^2 = 0.375
        \quad \triangleright \text{Eq.~\ref{eq:gini}} 
        \end{align*}   
        
        Decrease in Gini impurity:
        \begin{align*}   
        \small
        \Delta I_{\text{Gini}}(v_0 | \text{Age} \geq 18.5) = 0.455 -\frac{4}{20} \cdot 0.375 + \frac{16}{20} \cdot 0.375  = 0.080
        \quad \triangleright \text{Eq.~\ref{eq:gini_decrease}} 
        \end{align*}    
        
        This calculation is repeated for all potential thresholds (8.0, 10.5, 12.0, 18.5, 24.5, 25.5, 27.5, 30.5, 38.5, 45.0, 45.5, 50.5, 56.0, 59.0, 64.5, 68.5, 71.5, 77.0, 81.5). We summarized the results in Table~\ref{tab:gini_values_table}.

        \textbf{Gini Impurity for \emph{Fare Category} attribute} \hspace{0.25cm}
        We calculate the Gini impurity for each subset:
        \begin{itemize}
            \item \textbf{Low Fare}: 11 passenger, 4 Survived (Yes), 7 Did not survive (No)
        \end{itemize}        
        \begin{align*}   
        \small
        I_{\text{Gini}}(v_{0,\text{left}}) = 1 - \left(\left(\frac{4}{11}\right)^2 + \left(\frac{7}{11}\right)^2\right) = 0.463
        \quad \triangleright \text{Eq.~\ref{eq:gini}} 
        \end{align*}   
        
        \begin{itemize}
            \item \textbf{High Fare}: 9 passengers, 3 Survived (Yes), 6 Did not survive (No)
        \end{itemize}
        \begin{align*}   
        \small
        I_{\text{Gini}}(v_{0,\text{right}}) = 1 - \left(\left(\frac{3}{9}\right)^2 + \left(\frac{6}{9}\right)^2\right) = 0.444
        \quad \triangleright \text{Eq.~\ref{eq:gini}} 
        \end{align*}   
        
        Decrease in Gini impurity:
        \begin{align*}  
        \small
        \Delta I_{\text{Gini}}(v_0 | \text{Fare Category}) = 0.455 - \frac{11}{20} \cdot 0.463 + \frac{9}{20} \cdot 0.444 = 0.000
        \quad \triangleright \text{Eq.~\ref{eq:gini_decrease}} 
        \end{align*}           
        
    \item \textbf{Select the best split for the current node} \hspace{0.25cm}
        After computing the decrease in Gini impurity for all possible splits of the attributes \emph{Fare Category} and \emph{Age} the attribute-threshold pair with the highest reduction in impurity (Age $\geq$ 18.5 with a decrease of 0.080) is selected as the root node of the DT. The impurity decrease for each potential split is summarized in Table~\ref{tab:gini_values_table}.                
    \begin{figure}[t]
        \centering
        \includegraphics[width=0.7\textwidth]{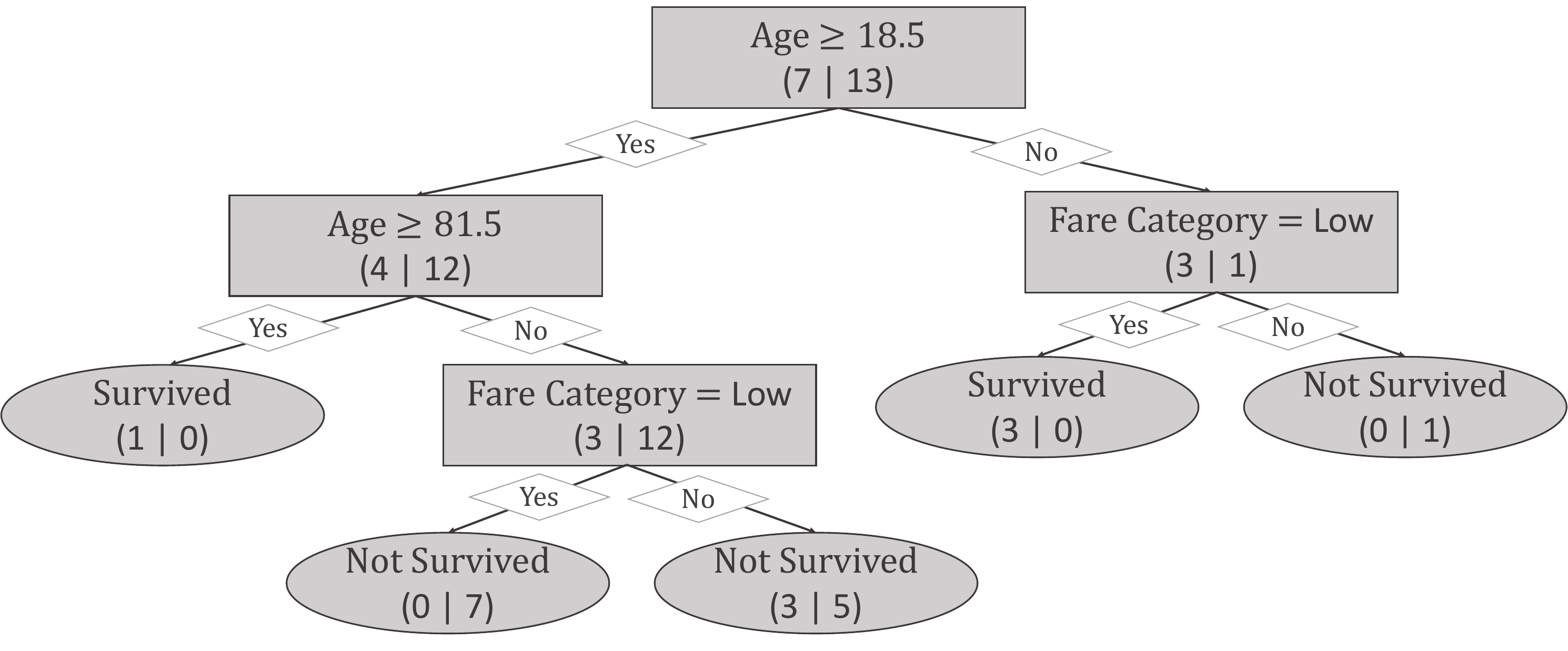}
        \caption[DT learned with CART]{\textbf{DT learned with CART.} The DT achieves an accuracy of 85\% on the training data and covers a total of 4 splits. It encodes the rule that if a passenger is older than 18.5, it is checked if the passenger is also older than 81.5. Those older survived and those younger did not survive, but with different certainties based on the fare category. If passengers are younger than 18.5, the tree splits based on Fare Category, where those with a low fare did survive, while those with a higher fare survived.}
        \label{fig:decision_tree_cart}
    \end{figure}        

    \item \textbf{Recursive Partitioning} \hspace{0.25cm}
        The DT then branches based on the chosen attribute and threshold. Each subset of the data is processed recursively, following the same procedure, until all nodes are pure, or another stopping criterion is met. This results in the DT visualized in Figure~\ref{fig:decision_tree_cart}.
    
\end{enumerate}

\paragraph{Limitations} While CART’s recursive partitioning and impurity-based split selection offer a straightforward and effective mechanism for constructing DTs, this greedy approach can inherently lead to suboptimal solutions from a global perspective. As shown in the exemplary iteration, it is evident that greedy algorithms, by their nature, focus on making locally optimal decisions at each step without accounting for the broader impact on the overall model.
This local optimality often prevents the discovery of globally optimal solutions, as decisions are made based solely on immediate impurity reduction, without considering how early splits influence the tree's future growth and accuracy. 
This can result in suboptimal models, especially when early splits do not align well with the underlying data distribution.
For instance, in Figure~\ref{fig:decision_tree_cart}, the initial split on the \emph{Age} attribute at a threshold of 18.5 maximizes immediate Gini impurity reduction. However, a more flexible approach that considers the global tree structure might have preferred an initial split on the \emph{Fare Category}, followed by splits on \emph{Age} at deeper levels, ultimately yielding a more accurate model. We will show that these more optimal solutions exist for the given example when moving to more advanced methods like lookahead (Section~\ref{sec:lookahead_dt}) DTs, evolutionary algorithms (Section~\ref{sec:evolutionary_dt}) or optimal DTs (Section~\ref{sec:optimal_dt}).
Moreover, greedy methods struggle to integrate seamlessly with existing gradient-based optimization frameworks, which are fundamental in modern machine learning domains such as multimodal learning and reinforcement learning. This limitation reduces their adaptability when combined with complex models that rely heavily on gradient-based updates. Additionally, defining and optimizing custom objectives in greedy algorithms can be challenging, as their inherent structure is tightly coupled with traditional impurity measures. This restricts the flexibility needed to tailor models for specific tasks or domain-specific requirements.

\subsection{Lookahead Decision Trees} \label{sec:lookahead_dt}
To overcome the issues of a greedy DT induction, many researchers focused on finding an efficient alternative. One solution to mitigate the impact of a greedy procedure are methods that look ahead during the induction~\citep{lookahead}. 
Lookahead DTs are an extension of traditional DTs that incorporate a multistep evaluation strategy when deciding how to split nodes. Instead of evaluating splits based only on their immediate effect, the algorithm simulates the outcomes of potential future splits up to a certain depth (the \emph{lookahead} depth). By doing so, it aims to make more globally optimal decisions that could lead to better overall tree performance.

\paragraph{Complexity}
The computational complexity of a lookahead DT algorithm is generally higher than that of a simple greedy approach due to the additional exploration required. When introducing a lookahead depth \( k \), at each node, we do not only evaluate the immediate splits but also consider potential splits for up to \( k \) levels ahead. This involves considering combinations of splits over multiple levels, which adds a combinatorial aspect to the complexity.
For each possible split at the current node, we recursively evaluate potential splits at the next \( k \) levels. In the worst case, if the tree has a branching factor \( b \), the number of combinations grows exponentially with \( k \), leading to \( b^k \) combinations (for a binary tree, \( b = 2 \)). Therefore, the time complexity for learning a DT with lookahead depth \( k \) becomes

\[
O(n \cdot m \log m \cdot 2^k)
\]

when storing the sorting results\footnote{When not storing the sorting results, the runtime increases to \( O(n \cdot m \log m \cdot d \cdot 2^k) \).} compared to \(O(n \cdot m \log m) \) for standard greedy algorithms.

\paragraph{Advantages over greedy algorithms}
The primary advantage of lookahead DTs over standard greedy algorithms is their ability to mitigate the limitations of a constrained search space and avoid suboptimality caused by sequentially selecting locally optimal splits. Greedy algorithms typically select splits that are optimal in the short term but may lead to suboptimal overall solutions due to their reliance on local information.  In contrast, lookahead DTs consider a range of possible future states, thus effectively balancing the trade-off between solution quality and computational complexity.

\begin{figure}[t]
    \centering
    \includegraphics[width=0.8\textwidth]{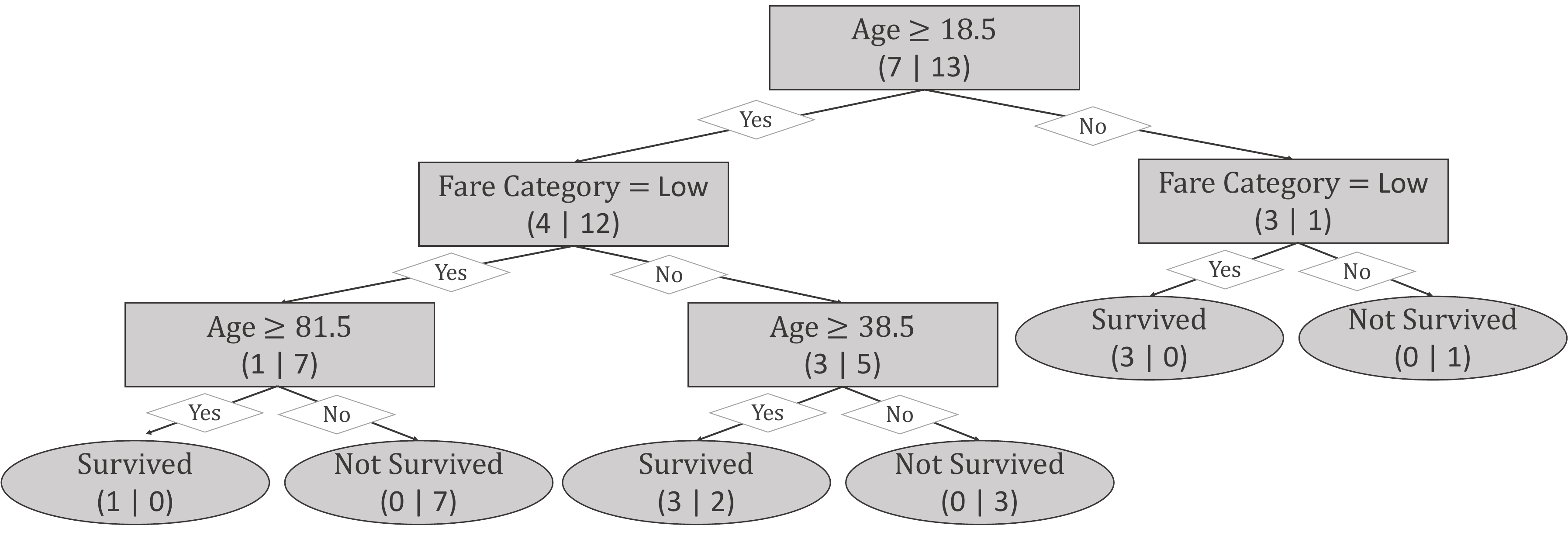}
    \caption[DT learned with Lookahead-CART]{\textbf{DT learned with Lookahead CART.} The DT was learned using CART with a lookahead depth of 2, i.e. looking one additional step forward. It achieves an accuracy of 90\% compared to 85\% for plain CART (Figure~\ref{fig:decision_tree_cart}) and includes 5 splits. The root node splits passengers by whether they are older than 18.5 years or not. In the second level, the tree splits on the fare for both branches and in the left branch additionally on the age.
    Essentially, the tree encodes three rules to decide whether a passenger survived or not. If a passenger is older than 18.5 years, paid a low fare and is older than 81.5 years, it is predicted to survive. Similarly, a passenger is predicted to survive if they are older than 18.5 years, paid a high fare and are older than 38.5 years. If a passenger is younger than 18.5 years and paid a low fare, they are also predicted to survive. In all other cases, the passenger is predicted not to survive.
    }
    \label{fig:decision_tree_lookahead}
\end{figure}

\subsubsection*{Example 2: A Decision Tree Learned with Lookahead}
For this example, we use the simplified version of the Titanic dataset from Table~\ref{tab:titanic_2d_num_cat} including the age as continuous and the fare as categorical attribute. As providing a step-by-step procedure similar to greedy algorithms is not in scope due to the increased complexity, we provide the final tree learned with CART using a lookahead depth of 2, i.e., by looking one additional step forward compare to standard CART. The resulting DT is visualized in Figure~\ref{fig:decision_tree_lookahead}. While the spit at the root node and the right branch is identical to standard CART (Figure~\ref{fig:decision_tree_cart}), we can observe that the left branch changed. Instead of splitting $\text{Age} \geq 81.5$ at the second level, we now split on the fare first and on the age in the third level. This allows for an additional split $\text{Age} \geq 38.5$ which was not possible with standard CART. These changes result in a performance increase as lookahead CART achieves 90\% train accuracy compared to 85\% for standard CART. However, as already mentioned, looking ahead comes at a high cost and potentially becomes infeasible for high lookahead depths.

\paragraph{Limitations}
Despite these theoretical advantages, there are several issues with lookahead DTs that limit their practical utility~\citep{murthy1996growing}. Firstly, the benefits of lookahead search over greedy induction are often marginal. Experimental evidence suggests that limited lookahead search produces DTs with roughly the same classification accuracy and size as those produced by greedy methods, with only slight reductions in the longest paths. Moreover, lookahead search can sometimes lead to poorer DTs that are less accurate, larger, or deeper than those generated by greedy induction. This phenomenon, known as \emph{pathology}, arises because more extensive search can result in poor generalization to unseen data. %
Another significant critique is that post-processing techniques, such as pruning, often provide benefits similar to or greater than those achieved by limited lookahead, without the associated computational expense. Pruning effectively reduces the size and depth of DTs, mitigating the issues that lookahead search aims to address but at a lower computational cost.
Additionally, the computational cost of lookahead DTs is a major drawback. The deeper search requires evaluating potential splits several levels down the tree, significantly increasing computational complexity compared to the greedy approach. This added computational requirement often outweighs any marginal benefits gained from the potentially improved tree structure.
Furthermore, lookahead DTs share the same limitations as standard greedy methods, often resulting in suboptimal trees and missing flexibility in terms of integration into existing frameworks and custom objectives.

\section{Evolutionary Algorithms for Learning Decision Trees} \label{sec:evolutionary_dt}
One common way to learn DTs in a non-greedy fashion and thereby overcome the local optimality limitation of sequential splitting is through evolutionary algorithms. Evolutionary Algorithms (EAs) perform a robust global search in the space of candidate solutions based on the concept of survival of the fittest~\citep{barros2011survey}. This usually results in smaller trees and a better identification of feature interactions compared to a greedy, local search~\citep{freitas2002data}. %
EAs and specifically Genetic Algorithms (GAs) as a subclass, offer a compelling approach to the induction of DTs by leveraging the principles of natural selection and genetic inheritance to evolve tree structures that optimize various performance metrics. 

\paragraph{Evolutionary search for decision tree induction}
EAs for DT induction involve initializing a population of candidate DTs, which are then evolved over successive generations. This process typically includes operations such as selection, crossover, mutation, and elitism, where trees are evaluated based on a fitness function that reflects their classification accuracy or other relevant metrics. The fitness function might also incorporate other objectives, such as minimizing tree size for simplicity. %

\paragraph{Learning complexity in evolutionary decision tree learning}
The learning complexity of evolutionary DT induction is generally higher than traditional methods due to the iterative nature of evolving a population of trees. Each generation requires evaluating $O(P)$ candidate trees, where $P$ is the population size, over $G$ generations. The fitness evaluation for each tree has a complexity of $O(m \cdot d)$, where $m$ is the number of samples and $d$ is the depth of the tree. Thus, the overall complexity is

\[
O(G \cdot P \cdot m \cdot d).
\] 

However, with maximal parallelism, where each tree in the population is evaluated on a separate processing unit, $P$ can effectively become $1$, reducing the runtime complexity to $O(G \cdot m \cdot d)$. Despite this, evolutionary methods still tend to be more computationally expensive than greedy algorithms, which operate in $O(m \cdot n \log n \cdot d)$ without reuse of sorting. %
An exception occurs with very high-dimensional datasets, where greedy algorithms often become runtime-inefficient, whereas the runtime of EAs remains unaffected by dimensionality. Although dimensionality does not influence runtime, it still impacts the optimization procedure, complicating the process and potentially affecting performance. %

\paragraph{Advantages of evolutionary algorithms in decision tree learning}
One of the most significant advantages of using EAs for DT induction is their ability to escape local optima. This capability stems from their global search process, which systematically explores the solution space, rather than being confined to a local region. This approach allows EAs to identify complex patterns and attribute interactions that traditional greedy algorithms might overlook.
Moreover, EAs offer a natural framework for multi-objective optimization. This is particularly relevant in scenarios where DTs must satisfy multiple criteria, such as high accuracy and minimal complexity. For instance, in cost-sensitive classification tasks like medical diagnosis, EAs can optimize DTs to minimize different types of misclassification costs explicitly. Additionally, EAs can be tailored to incorporate parsimony pressure, encouraging the development of simpler, more interpretable trees, which is often a critical requirement in practical applications.

\begin{figure}[t]
    \centering
    \includegraphics[width=0.8\textwidth]{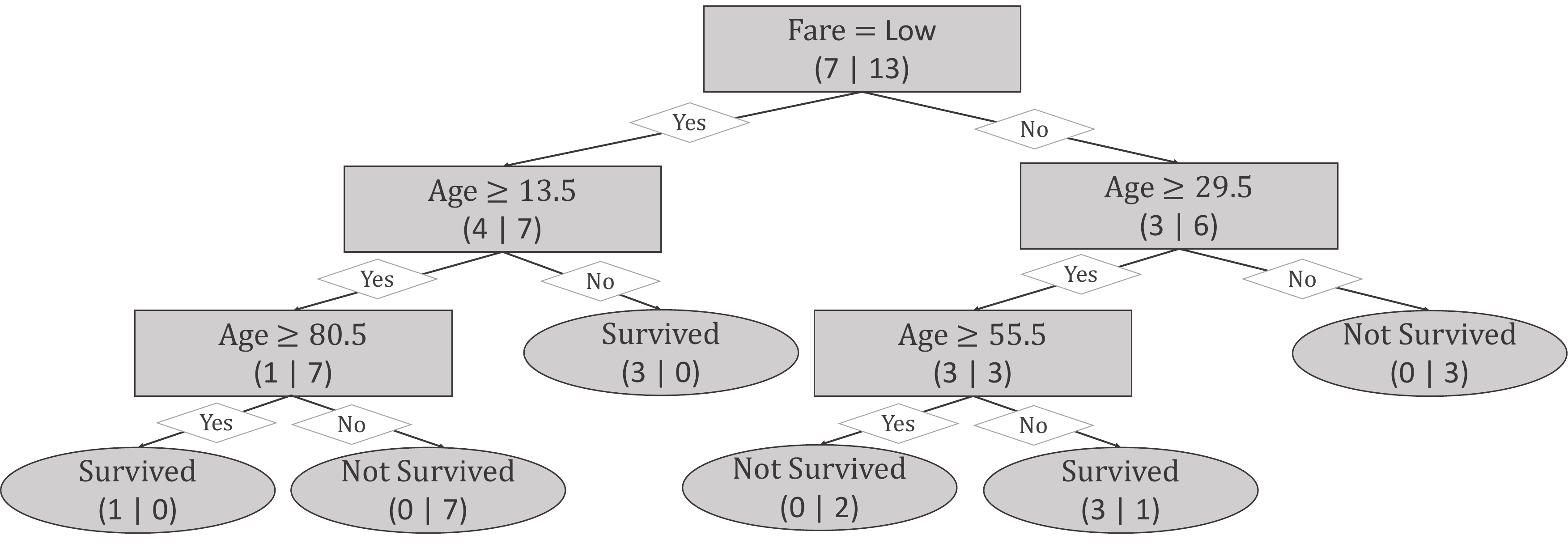}
    \caption[DT learned with a Genetic Algorithm]{\textbf{DT learned with a Genetic Algorithm.} The DT learned with a GA achieves an accuracy of 95\% compared to 85\% for CART (Figure~\ref{fig:decision_tree_cart}) and includes 5 splits. The root node splits on the fare, and subsequent splits are made on the age attribute.
    Essentially, the tree encodes three rules to decide whether a passenger survived or not. If a passenger paid a low fare, is older than 80.5 years, they are predicted to survive. Similarly, a passenger is predicted to survive if they paid a low fare and are younger than 13.5 years. If a passenger paid a high fare, is older than 29.5 years, but younger than 55.5 years, they are also predicted to survive. In all other cases, the passenger is predicted not to survive.}
    \label{fig:decision_tree_genetic}
\end{figure}

\subsubsection*{Example 3: A Decision Tree Learned with a Genetic Algorithm}
For this example, we use the simplified version of the Titanic dataset from Table~\ref{tab:titanic_2d_num_cat} including the age as continuous and the fare as categorical attribute. As providing a step-by-step procedure similar to greedy algorithms is not in scope due to the high complexity of EAs, especially with a reasonably great population, we provide the final tree learned with a genetic algorithm\footnote{We used the GeneticTree package for learning the DT \url{https://github.com/pysiakk/GeneticTree} (Last accessed: March 3, 2025).}. The resulting DT is visualized in Figure~\ref{fig:decision_tree_genetic}.
The DT learned with the genetic algorithm has a very different structure compared to greedy, purity-based methods (Figure~\ref{fig:decision_tree_c45} and Figure~\ref{fig:decision_tree_cart}), starting with the fare at the root node. Not relying on the sequential, greedy procedure allows learning a DT that has the same number of nodes compared to the CART DT with lookahead (Figure~\ref{fig:decision_tree_lookahead}), but achieves a 5\% higher performance. This is mainly caused by the fact that the split on fare is made in the root node (which is not possible with a greedy procedure), avoiding duplicate splits at lower levels.

\paragraph{Limitations}
Despite their advantages, EAs also come with several drawbacks. The most notable is their computational expense. The iterative nature of the evolutionary process, involving multiple generations of selection, crossover, and mutation, is computationally intensive. This issue has been somewhat mitigated by advances in computational power and parallel processing techniques, which allow for more efficient execution of these algorithms. %
Furthermore, EAs are hard to tune, as they have a lot of hyperparameters such as population size, mutation rate, and crossover rate all significantly impact the performance of the algorithm and must be carefully adjusted depending on the specific dataset and problem context. 
In addition, EAs often struggle to effectively handle high-dimensional data. As dimensionality increases, the optimization process becomes more complex. For instance, finding a good split through random mutations becomes increasingly difficult, potentially leading to slower convergence or suboptimal solutions.
While they offer greater flexibility compared to greedy methods, for example in selecting suitable objectives, they remain difficult to integrate into modern machine learning approaches, which typically rely on efficient, gradient-based optimization.

\section{Optimal Decision Trees} \label{sec:optimal_dt}

Another way to overcome the issues of a greedy DT induction, many researchers focused on finding an efficient alternative. Optimal DTs aim to optimize an objective (e.g., the purity) through an approximate brute force search to find a globally optimal tree with a certain specification~\citep{zharmagambetov2021non}. 
Unlike traditional DT methods like CART and C4.5, which build trees using a greedy, top-down approach, optimal DTs aim to find the globally optimal tree by considering all possible tree structures and splits simultaneously.
Most commonly, optimal DT algorithms use Mixed Integer Optimization (MIO) formulations \citep{bertsimas2017optimal} and optionally use advanced algorithms like caching-based branch-and-bound search algorithms to restrict the search space and improve efficiency\citep{aglin2020learning}.

\paragraph{Mixed integer optimization}

Mixed Integer Optimization (MIO) is a type of mathematical optimization that involves finding the best solution to a problem from a finite set of feasible solutions. The \emph{mixed} part refers to the fact that the problem involves both continuous variables (real numbers) and integer variables. An MIO problem can be formulated as:

\begin{equation}
\text{minimize} \; \mathcal{L}(\boldsymbol{x}) \; \text{subject to} \; g_i(\boldsymbol{x}) \leq 0, \; \forall i \in I, \; \boldsymbol{x} \in \mathbb{Z}^p \times \mathbb{R}^q,
\end{equation}

where \( \mathcal{L}(\boldsymbol{x}) \) is the objective function to minimize (e.g., misclassification error). In this context, \( g_i(\boldsymbol{x}) \) represents the set of \emph{constraints} that define the feasible region of the optimization problem. Each \( g_i(\boldsymbol{x}) \leq 0 \) imposes a condition that the solution must satisfy, such as limiting the range of variables, enforcing logical relationships, or encoding problem-specific requirements (e.g., tree depth, feature splits, or class assignments). Furthermore, \( \boldsymbol{x} \in \mathbb{Z}^p \times \mathbb{R}^q \) indicates that some variables are integers (discrete choices) and others are continuous (real values).
MIO is particularly suitable for DT problems because it can naturally model discrete decisions, such as whether to split at a particular node, which feature to use for the split, and which class to assign to a leaf node.

\paragraph{Formulating decision trees as an MIO problem}
The objective of constructing an optimal DT is to minimize the misclassification error \( I_{\text{Mis}}(T) \) of the tree \( T \) on the training data, while also controlling the complexity of the tree through a penalty term \( \alpha |V| \). This leads to the following optimization problem~\citep{bertsimas2017optimal}:

\begin{equation}
\text{minimize} \; I_{\text{Mis}}(T) + \alpha |V| \; \text{subject to} \; m_l \geq m_{\text{min}}, \; \forall l \in V_l,
\end{equation}

where \( I_{\text{Mis}}(T) \) is the misclassification error of the tree \( T \) on the training data, \( \alpha |V| \) is a complexity penalty, where \( |V| \) is the number of nodes in the tree, \( m_l \) is the number of training points in leaf node \( l \) and \( m_{\text{min}} \) is the minimum number of points required in any leaf node\footnote{Please note that we use the number of nodes as a complexity penalty and the minimum number of points required in a leaf node as a constraint. While this is the most common choice, alternative penalty terms and constraints are possible as well.}.
To solve this problem, the MIO formulation introduces binary decision variables to represent whether a node is split or not, which feature to split on, and which direction each data point should take at each split.

\paragraph{Solving MIO problems with branch-and-bound search}
Branch-and-bound is a popular algorithmic technique used to solve combinatorial optimization problems, including MIO problems. It systematically explores the solution space by dividing it (branching) into smaller subspaces and using bounds to eliminate subspaces that cannot contain the optimal solution (bounding). In the context of DTs, branch-and-bound works as follows:

\begin{enumerate}
    \item \textbf{Branching}: The search space of possible DTs is divided recursively into smaller subsets. Each subset corresponds to a partial solution (e.g., a partially constructed tree).
    \item \textbf{Bounding}: A bound on the best possible solution within each subset is computed. If this bound is worse than the current best-known solution, that subset is pruned (eliminated from further consideration).
\end{enumerate}

The algorithm continues until all subsets have been either explored or pruned, ensuring that the best possible solution is found.

\paragraph{Improving branch-and-bound search}
Caching-based branch-and-bound methods, such as DL8.5~\citep{aglin2020learning}, offer several advantages. First, they enhance efficiency by reusing results from previously explored paths, significantly speeding up computations. Second, they demonstrate strong scalability, performing well even with larger datasets and deeper tree structures, often outperforming traditional and other optimal methods. Finally, they provide flexibility by accommodating various data types and optimization criteria, making them suitable for a wide range of applications.
In addition, MurTree~\citep{murtree} further uses dynamic programming, which reduces the runtime significantly. 

\paragraph{Advantages of MIO-based decision trees} The use of MIO for DTs offers several distinct advantages. First, MIO enables global optimization, ensuring the construction of a globally optimal tree rather than settling for a locally optimal solution often produced by greedy, top-down approaches. Second, it provides flexible constraints, allowing the seamless incorporation of requirements such as maximum tree depth or minimum samples per leaf — essential for applications where fairness or interpretability is a priority. Finally, the MIO framework facilitates easy extensions like the use of a custom objective function or multivariate splits, where decision nodes can rely on linear combinations of features rather than single-feature splits.

\begin{figure}[t]
    \centering
    \includegraphics[width=0.8\textwidth]{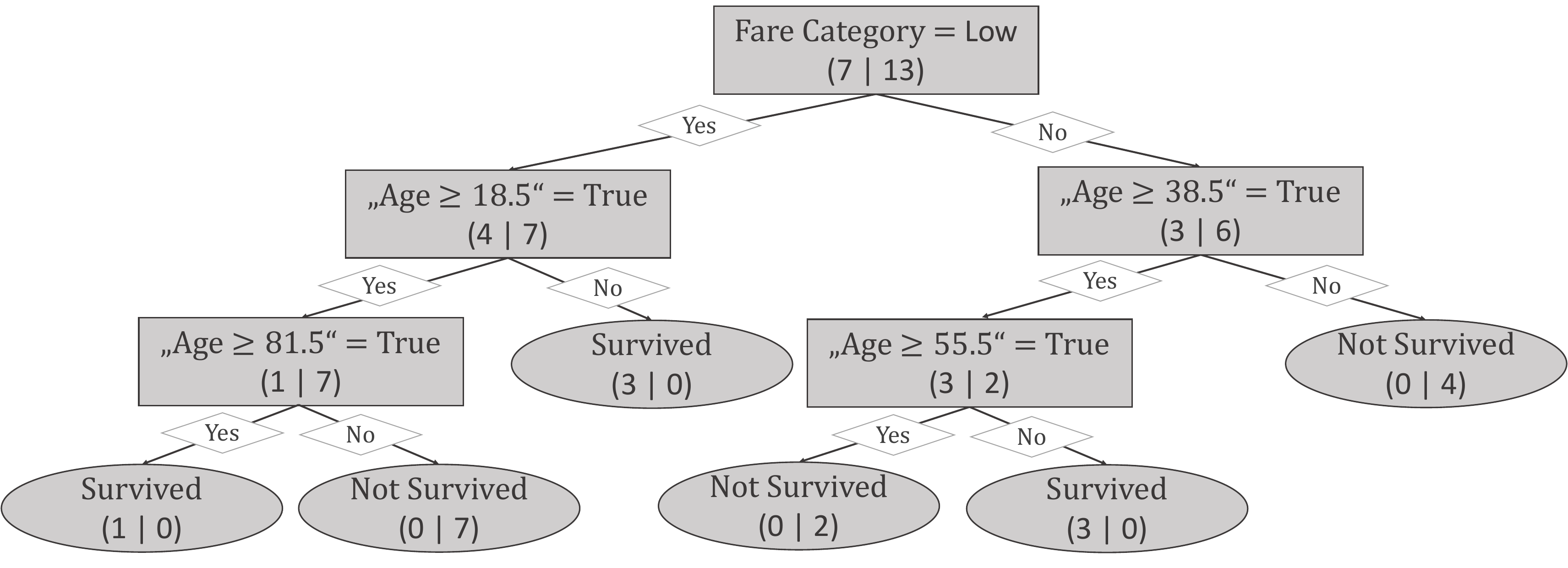}
    \caption[DT learned with an Optimal DT Algorithm]{\textbf{DT learned with an Optimal DT Algorithm.} The optimal DT achieves a perfect accuracy of 100\% compared to 85\% for CART (Figure~\ref{fig:decision_tree_cart}) and includes 5 splits. The root node splits on the fare, and subsequent splits are made on the continuous age attribute.
    Essentially, the tree encodes three rules to decide whether a passenger survived or not. If a passenger paid a low fare, is older than 81.5 years, they are predicted to survive. Similarly, a passenger is predicted to survive if they paid a low fare and are younger than 18.5 years. If a passenger paid a high fare, is older than 38.5 years, but younger than 55.5 years, they are also predicted to survive. In all other cases, the passenger is predicted not to survive.}
    \label{fig:decision_tree_optimal}
\end{figure}

\begin{table}[t]
\caption[Exemplary Titanic 2D Dataset (Binarized)]{\textbf{Exemplary Titanic 2D Dataset (Binarized).}}
\label{tab:titanic_2d_binarized}
\small
\centering
\begin{tabular}{|c|c|c|c|c|c|c|c|c|c|}
\hline
\textbf{ID} & \makecell{\textbf{Fare} \\ \textbf{$=$} \\ \textbf{'Low'}}  & \makecell{\textbf{Age} \\ \textbf{$\geq$ 8.0}} & \makecell{\textbf{Age} \\ \textbf{$\geq$ 10.5}} & \makecell{\textbf{Age} \\ \textbf{$\geq$ 12.0}} & $\cdots$ & \makecell{\textbf{Age} \\ \textbf{$\geq$ 71.5}} & \makecell{\textbf{Age} \\ \textbf{$\geq$ 77.0}} & \makecell{\textbf{Age} \\ \textbf{81.5 $\geq$}} & \textbf{Survived} \\
\hline
1  & 1  & 1 & 1 & 1 & $\cdots$ & 0 & 0 & 0 & No \\
2  & 0  & 1 & 1 & 1 & $\cdots$ & 0 & 0 & 0 & Yes \\
3  & 0  & 1 & 1 & 1 & $\cdots$ & 0 & 0 & 0 & No \\
4  & 1  & 1 & 0 & 0 & $\cdots$ & 0 & 0 & 0 & Yes \\
5  & 0  & 1 & 1 & 1 & $\cdots$ & 0 & 0 & 0 & No \\
6  & 1  & 1 & 1 & 1 & $\cdots$ & 0 & 0 & 0 & No \\
7  & 1  & 1 & 1 & 1 & $\cdots$ & 0 & 0 & 0 & No \\
8  & 0  & 1 & 1 & 1 & $\cdots$ & 0 & 0 & 0 & Yes \\
9  & 0  & 1 & 1 & 1 & $\cdots$ & 0 & 0 & 0 & No \\
10 & 0  & 1 & 1 & 1 & $\cdots$ & 0 & 0 & 0 & No \\
11 & 1  & 1 & 1 & 1 & $\cdots$ & 0 & 0 & 0 & Yes \\
12 & 1  & 1 & 1 & 1 & $\cdots$ & 0 & 0 & 0 & No \\
13 & 1  & 1 & 1 & 0 & $\cdots$ & 0 & 0 & 0 & Yes \\
14 & 1  & 1 & 1 & 1 & $\cdots$ & 1 & 1 & 1 & Yes \\
15 & 0  & 0 & 0 & 0 & $\cdots$ & 0 & 0 & 0 & No \\
16 & 1  & 1 & 1 & 1 & $\cdots$ & 1 & 1 & 0 & No \\
17 & 0  & 1 & 1 & 1 & $\cdots$ & 0 & 0 & 0 & Yes \\
18 & 1  & 1 & 1 & 1 & $\cdots$ & 0 & 0 & 0 & No \\
19 & 1  & 1 & 1 & 1 & $\cdots$ & 0 & 0 & 0 & No \\
20 & 0  & 1 & 1 & 1 & $\cdots$ & 1 & 0 & 0 & No \\
\hline
\end{tabular}
\end{table}

\subsubsection*{Example 4: An Optimal Decision Tree}

To showcase the potential of optimal DTs, we now go back to our exemplary titanic dataset from Section~\ref{ssec:2D_titanic}. While greedy algorithms struggled with this dataset, we were able to show that lookahead DTs as a way to mitigate the greedy nature already resulted in an improvement. Unfortunately, looking ahead too far becomes computationally infeasible fast. Optimal DTs however mitigate this by using advanced optimization techniques. 
As providing a step-by-step procedure similar to greedy algorithms is not in scope due to the high complexity of optimal DT algorithms, we provide the final tree learned with an optimal DT algorithm. The resulting optimal DT is visualized in Figure~\ref{fig:decision_tree_optimal}\footnote{Please note that there can exist multiple optimal DTs for a certain dataset, and optimal DT algorithms find one of these trees. Therefore, different optimal DT algorithms might find different trees.}. In theory, finding an optimal DT is equal to learning a lookahead DT where the lookahead depth $k$ equals the depth of the tree, i.e. $k=d$. The main difference between lookahead and optimal DTs in this case is their efficiency. While lookahead DTs have a complexity scaling exponentially with the lookahead depth, optimal DT algorithms use advanced optimization algorithms (e.g., efficient MIO solvers) and eliminate irrelevant parts of the search space, making them more efficient.
Yet, most advanced optimization techniques require binarized features for their application. As a result, continuous features must be binarized, e.g., through binning. This usually results in an information loss, as, especially for larger datasets with many continuous features, a loss-free binarization is not feasible. For the purpose of this demonstrative example, we chose an alternative method by encoding all potential splits in the DT, resulting in $20$ new binary features, as summarized in Table~\ref{tab:titanic_2d_binarized}. This dataset represents a loss-free binarization where the resulting DT can be interpreted similar to the trees learned by alternative methods.
The DT produced by the optimal DT algorithm differs significantly from those generated by CART (Figure~\ref{fig:decision_tree_cart}) and lookahead CART (Figure~\ref{fig:decision_tree_lookahead}). Notably, with increasing lookahead depth, the split on fare progressively shifts closer to the root node, aligning with its position in the optimal DT. Furthermore, the tree learned by the optimal DT algorithm closely resembles the one generated by a genetic algorithm (Figure~\ref{fig:decision_tree_genetic}), differing only in the thresholds at three nodes. Two thresholds are nearly identical, and only one affects predictions on the training data. At the second level's right node, the genetic algorithm tree splits age at 29.5, while the optimal tree splits at 38.5. This adjustment enables the optimal tree to correctly classify one additional instance as not survived at the rightmost leaf node, which the genetic algorithm misclassifies as survived.
These differences in thresholds arise from the genetic algorithm's mutation-based procedure, where thresholds are not predefined from the data but are randomly generated, potentially leading to suboptimal decisions.

\paragraph{Limitations}
Optimal DTs are only optimal with respect to training data and therefore do not guarantee optimal generalization. They are difficult to regularize and tend to overfit~\citep{zharmagambetov2021non}. 
This raises a critical question: Is it truly beneficial to pursue global optimality on the training set, or should we instead target a sufficiently good local optimum that may generalize better? However, if we decide for the second approach, it remains essential to develop methods that go beyond the limitations of greedy, sequential splitting techniques, which inherently constrain the search space, and thereby allow the model to explore more promising regions of the solution space.
Additionally, optimal DTs typically have poor scalability and can only rarely be learned for depths larger than 4, especially for large or high-dimensional datasets.
Furthermore, most state-of-the-art approaches still require binary data and therefore a discretization of continuous features~\citep{bertsimas2017optimal,aglin2020learning,murtree}, which can lead to information loss. An exception is the approach by \citet{mazumder2022quant}, which can handle continuous features out-of-the-box. However, their method is optimized for very sparse trees and limited to a maximum depth of $3$.
Furthermore, although optimal DTs offer greater flexibility compared to greedy methods, they still do not support seamless integration into modern machine learning approaches, such as those in multimodal or reinforcement learning approaches, which typically rely on gradient-based optimization.

\section{Gradient-Based Decision Tree Learning} \label{sec:gradient_dt}

Gradient-based DT learning has emerged as a powerful approach to optimize DTs by leveraging techniques traditionally used in neural networks. Unlike classical DTs, which rely on greedy algorithms to find optimal splits, gradient-based methods apply continuous optimization techniques such as gradient descent. The method proposed in this thesis can also be categorized as a gradient-based method, but significantly differs from existing methods by the flexible use of hard, axis-aligned splits, which is not possible with the existing methods we will introduce in the following.
In this section, we outline the development of gradient-based approaches and introduce key modifications that have influenced advancements in gradient-based DT learning~\citep{popov2019neural,node_gam}. We will first discuss a general optimization of soft DTs with gradient descent (Section~\ref{sec:sdt_related}) including a discussion of further optimizations and modifications to the tree structure. Next, we will introduce the most prominent works on gradient-based DT learning that are closely related to the method proposed in this thesis, including Deep Neural DTs (Section~\ref{sec:dndt}), Neural Oblivious Decision Ensembles (Section~\ref{ssec:node}) and Representations for Axis-Aligned Decision Forests (Section~\ref{sec:neural_axisaligned_forests}).

\paragraph{Distinction between gradient-based and gradient-boosted decision trees} In this context, we want to emphasize the difference between \emph{gradient-based} DTs and \emph{gradient-boosted} DT~\citep{friedman2001gradientboostingtree} frameworks like XGBoost~\citep{chen2016xgboost} or CatBoost~\citep{prokhorenkova2018catboost}. While both gradient-based DT learning and gradient-boosted DTs incorporate gradients to guide their optimization, they do so in fundamentally different ways. In gradient-based DT learning, the gradient is computed directly with respect to the parameters of a DT. Here, the gradient explicitly refines the tree’s structure and decision boundaries through end-to-end training, much like parameter updates in neural networks. 
In contrast, gradient-boosted DTs form an ensemble of multiple trees, each trained independently to greedily fit the residual errors from the previous stage. They make use of gradients indirectly, deriving them from the loss function’s sensitivity to current predictions. Each new tree in the ensemble is trained to greedily fit the residual errors, which are effectively the negative gradients of the loss with respect to the model’s predictions. This involves constructing each subsequent tree using conventional DT algorithms (like CART), but treating the residuals from previous predictions as the target values.
As a result, gradient boosting leverages gradients to iteratively improve a sequence of discrete, greedily built trees rather than adjusting a single tree’s parameters continuously.  In the following, we focus on gradient-based methods, as we aim to directly learn the parameters of a DT end-to-end with gradient descent.

\subsection{Soft Decision Trees} \label{sec:sdt_related}
In contrast to vanilla DTs that make a hard decision at each internal node, many hierarchical mixtures of expert models~\citep{soft1} have been proposed. The concept of SDTs was already introduced in Section~\ref{ssec:sdt_background} and an exemplary visualization comparing a SDT with a standard DT can be found in Figure~\ref{fig:decision_tree_soft}. In addition to the oblique decision boundary, they usually make soft splits, where each branch is associated with a probability~\citep{soft2,soft3} and the final prediction is a weighted average of the leafs. 
This inherent adjustments to the tree architecture allows for the application of common optimization algorithms not specifically tailored towards DTs. 
The first SDT models~\citep{soft1} were trained using the Expectation Maximization algorithm~\citep{dempster1977maximum}, which was originally used for unsupervised learning, adapted for supervised learning tasks.
With the resurgence of deep learning in the beginning of the 21st century~\citep{hinton2006fast,krizhevsky2012imagenet,bengio2006greedy,salakhutdinov2009deep}, optimization with gradient descent was successfully applied to SDTs~\citep{soft2,soft3} and is still used by state-of-the-art algorithms~\citep{popov2019neural,node_gam}. Building on a standard gradient-based optimization procedure, several modifications and optimizations were proposed and will be summarized in the following.

\paragraph{Limitations}
While relaxing hard, axis-aligned decisions to soft, oblique splits enables end-to-end gradient-based optimization and offers greater flexibility, for instance in the selection of custom objectives, it also eliminates the inductive bias of axis-aligned splits, which can be advantageous in several domains, such as tabular data. Moreover, soft, oblique splits typically lead to significantly reduced interpretability, particularly at the split level. This is supported by \citet{molnar2020}, where the authors argue that humans struggle to comprehend explanations involving more than three dimensions simultaneously.

\subsubsection*{Modifications and Optimizations for Soft Decision Trees}
While SDTs have gained traction due to their differentiability and flexibility, they also introduce limitations, as discussed above. To address these limitations, various extensions and alternative formulations have been proposed, each focusing on different aspects such as sparsity or structure optimization. The following paragraphs discuss related approaches that seek to enhance SDTs with respect to their soft or oblique nature.

\paragraph{Sparsity in optimal randomized classification trees}  
\citet{blanquero2020sparsity} aim to increase the interpretability of SDTs by optimizing for sparsity at both local and global levels. Their approach formulates the problem as a continuous optimization task, where sparsity is enforced using polyhedral norm regularizations. Specifically, they employ $\ell_1$-norm regularization to encourage local sparsity, reducing the number of predictor variables involved in each individual split, and $\ell_{\infty}$-norm regularization to promote global sparsity by limiting the total number of variables used throughout the entire tree. 
Unlike traditional DT algorithms, which typically rely on greedy heuristics, their method constructs optimal classification trees through a continuous, non-greedy optimization process, allowing for improved trade-offs between performance and interpretability. 
Their results demonstrate that significant gains in sparsity can be achieved without substantial reductions in classification performance. %

\paragraph{Adaptive Neural Trees (ANTs)}
\citet{tanno2019ant} combine the benefits of neural networks and DTs by using so-called Adaptive Neural Trees (ANTs), a hybrid approach that integrates representation learning into the edges, routing functions, and leaf nodes of a DT. Unlike conventional DTs, which rely on predefined features and axis-aligned splits, ANTs employ stochastic routing based on a Bernoulli distribution, enabling soft partitioning of the data. Additionally, they utilize non-linear transformer modules at the edges, which results in oblique decision boundaries and deep feature learning within the tree structure. This design removes the static architecture of traditional DTs while retaining their hierarchical structure. ANTs are trained using a backpropagation-based algorithm that incrementally grows the tree by dynamically determining whether to split a node or deepen an existing path based on validation loss. This adaptive architecture learning allows the model to adjust its complexity to the available data, balancing efficiency and performance. However, the resulting trees are still soft and oblique. %

\paragraph{One-Stage Tree}
\citet{one-stage-tree} propose One-Stage Tree as a novel method for learning SDTs that jointly optimizes both the tree structure and its parameters in a single stage. Unlike traditional SDTs that rely on a two-stage process of training and subsequent pruning, One-Stage Tree integrates both steps into a bilevel optimization framework. This approach ensures that the tree structure remains discrete during training, leveraging the reparameterization trick and proximal iterations to maintain interpretability while enabling end-to-end differentiability. By jointly learning both node and architecture parameters, One-Stage Tree reduces the performance gap between training and testing, which is a common issue in existing SDTs.
However, despite its improved interpretability compared to other SDTs, its instance-wise routing rules still present limitations in providing a globally interpretable decision process. Since each instance is routed based on a locally determined discretization of the parameters rather than a fixed global rule, the resulting decision boundaries can vary across instances, making it difficult to extract a single, human-interpretable rule set that holds consistently for all inputs.

\paragraph{Efficient non-greedy optimization of decision trees}
\citet{norouzi2015efficient} proposed an approach to overcome the need for soft decisions to apply gradient-based algorithms by minimizing a convex-concave upper bound on the tree's empirical loss. Their method introduces a structured prediction perspective, associating a latent binary decision variable with each split node and using these variables to define the empirical loss of the DT. By leveraging a gradient-based optimization strategy, they efficiently train DTs while maintaining hard splits, which is in contrast to SDT methods that rely on smooth approximations.
Additionally, their method provides a natural way to regularize tree parameter optimization, discouraging overfitting. While this allows the use of hard splits, the approach is still limited to oblique trees. 

\paragraph{Argmin differentiation for learning decision trees}
\citet{zantedeschi2021learning} use argmin differentiation to learn binary DTs by simultaneously optimizing both discrete and continuous parameters. Their method reformulates the learning process as a mixed-integer program and introduces a sparse relaxation that enables gradient-based optimization. This formulation preserves the ability to enforce hard splits in the learned trees while ensuring that gradients can flow through the optimization process, facilitating end-to-end training. Unlike standard probabilistic relaxation methods that lack a principled pruning mechanism and often require heuristic post-processing, their approach incorporates tree pruning directly into the optimization by introducing a quadratic regularization term. This enables efficient pruning of unnecessary nodes while maintaining interpretability. The authors demonstrate that their method can be integrated as a differentiable module within deep networks and optimized for various downstream objectives. Their framework supports arbitrary differentiable split functions, like linear models, which naturally prevents axis-aligned splits and typically leads to oblique DTs. %

\paragraph{Gradient-based learning of hard decision trees}
\citet{karthikeyanlearning} developed a gradient-based approach to learn hard DTs, addressing the challenge of optimizing discrete, non-differentiable decision boundaries. Their method, called Dense Gradient Trees (DGT), leverages overparameterization and the straight-through operator to enable efficient end-to-end gradient-based training of DTs. A key component of their approach is the use of a structured sum formulation instead of the traditional product-based path function, improving gradient conditioning and stability during training. They also incorporate a quantization-aware gradient descent method to enforce hard decision boundaries without sacrificing differentiability. While they allow for hard splits, the method remains constrained to oblique trees due to its reliance on a linear decision function at each internal node. \\

Despite the diverse range of enhancements proposed for soft DTs, none of these approaches fully resolve the fundamental trade-off between a hard, axis-aligned structure and gradient-based optimization. While many methods improve sparsity, enable joint learning of tree structure and parameters, or facilitate efficient training through novel optimization techniques, they either retain oblique splits or rely on probabilistic routing, thereby failing to achieve both hard splits and axis-aligned decision boundaries simultaneously. This gap remains an open challenge in the development of hard, axis-aligned DTs that can be trained with gradient descent.

\subsubsection*{Gradient-Based Tree Altering Methods}
In addition to methods that aim to learn a DT from scratch, there are approaches focused on optimizing an already trained DT using gradient information, which we will briefly discuss in the following.
\citet{suarez1999globally} relax the hard splits derived from an already existing DT (e.g., learned using CART) and optimizes the soft splits with backpropagation. 
\citet{gouk2019stochastic} propose stochastic gradient trees (SGTs) which use stochastic gradient information as source of supervision to incrementally learn a DT. However, SGTs do not apply gradient descent to learn the entire tree but evaluate the loss function for each possible split to find the one that yields the maximum loss reduction.
\citet{carreira2018alternating} proposed a tree altering optimization algorithm (TAO) that, given an input tree (its structure and the parameter values at its nodes), generates a new tree with the same or a smaller structure, but optimized parameter values. %
TAO can be applied for both, oblique and axis-aligned trees. For axis-aligned trees, TAO uses alternating optimization~\citep{bezdek2002_alernating_optimization}, to deal with the non-differentiable nature of DTs. %
While optimizing an already trained DT is an interesting research area, it fundamentally differs from the task of learning an entire tree from scratch and is therefore not considered further in this thesis.

\subsection{Deep Neural Decision Trees (DNDT)} \label{sec:dndt}
\citet{dndt} introduce DNDT, a hybrid model that leverages the strengths of both neural networks and DTs. 
DNDT seeks to combine the interpretability and axis-aligned nature of DTs with the flexibility of neural networks and a gradient-descent optimization. Therefore, DNDTs are among the most closely related methods and will be discussed in more detail in the following.

\paragraph{Method}
DNDT models the DT structure through a neural network by employing a differentiable approximation of a binning function. The core idea is to replace the traditional hard splits used in DTs with a soft binning mechanism that can be optimized through backpropagation. 
Thereby, the number of bins is a hyperparameter that must be chosen by the user, where setting it to 2 results in a binary decision tree.
Their soft binning function takes as input a continuous variable $x_\iota$ and outputs the likelihood that $x_\iota$ falls into predefined intervals (bins). Formally, the binning function is defined as:

\begin{equation}
f(\boldsymbol{x}|\boldsymbol{w},\boldsymbol{b},\iota) = \text{softmax}\left(\frac{\boldsymbol{w}x_\iota + \boldsymbol{b}}{s}\right),
\end{equation}

where $\boldsymbol{w}$ is a fixed vector, $\boldsymbol{b}$ is a vector of biases corresponding to the bin boundaries, and $s$ is a temperature parameter that controls the sharpness of the output. As $s \to 0$, the softmax function behaves more like a hard assignment, which effectively implements a DT-like structure.
Once the input features are binned, the decision process can be expressed as a \textit{Kronecker product} $\otimes$ of the feature-specific binning functions:

\begin{equation}
z = f_0(x_0) \otimes f_1(x_1) \otimes \cdots \otimes f_{n-1}(x_{n-1}).
\end{equation}

Here, $z$ is an almost one-hot vector representing the path an instance takes through the tree, and the final classification is performed via a linear classifier at each leaf node. The differentiable nature of DNDT allows it to optimize its parameters (e.g., binning cut-points and leaf classifiers) using gradient descent. 

\paragraph{Limitations} Since DNDTs are generated via the Kronecker product of the binning layers, the structure depends on the number of features and classes (and the number of bins). As discussed by the authors, this results in  poor scalability with respect to the number of features, which currently can only be solved by using random forests for high-dimensional datasets ($>12$ features). 
Furthermore, using the Kronecker product to build the tree prevents splitting on the same feature with different thresholds in the same path, which can be crucial to achieve a good performance. 
Moreover, although the trees in DNDTs are axis-aligned, their standard formulation relies on soft split decisions, requiring additional modifications to achieve hard, deterministic splits.

\subsection{Oblique Decision Ensembles} \label{ssec:oblique_ensembles}
The methods introduced in the following were proposed to learn ensembles of DTs with a gradient-based procedure aiming for performance instead of interpretability. However, they can (at least partially) also be applied to learn individual trees and are very closely related to the method proposed in this thesis. Therefore, they will be examined in greater detail in the following.

\subsubsection*{Neural Oblivious Decision Ensembles (NODE)} \label{ssec:node}

\citet{popov2019neural} introduce Neural Oblivious Decision Ensembles (NODE), a novel architecture designed for deep learning on tabular data, addressing the gap between deep neural networks and traditional gradient-boosting DTs. NODE extends oblivious DTs to enable end-to-end gradient-based optimization by relaxing discrete operations. Consequently, the trees used by NODE are oblivious and oblique\footnote{While the terms are very similar, it is important to note that oblivious and oblique DTs are two distinct types of trees. Oblivious DTs use the same feature and threshold across all nodes at a given depth, ensuring uniformity in decision-making at each level. In contrast, oblique DTs are soft DTs and split data using linear combinations of features, allowing for more flexible and complex decision boundaries.}.

\begin{figure}[t]
    \centering
    \begin{subfigure}[t]{0.425\textwidth}
        \centering
        \includegraphics[width=\textwidth]{Figures/Background/decision_tree_axis_aligned.pdf}
        \caption{Standard DT structure.}
        \label{fig:decision_tree_standard}
    \end{subfigure}%
    \hfill
    \begin{subfigure}[t]{0.55\textwidth}
        \centering
        \includegraphics[width=\textwidth]{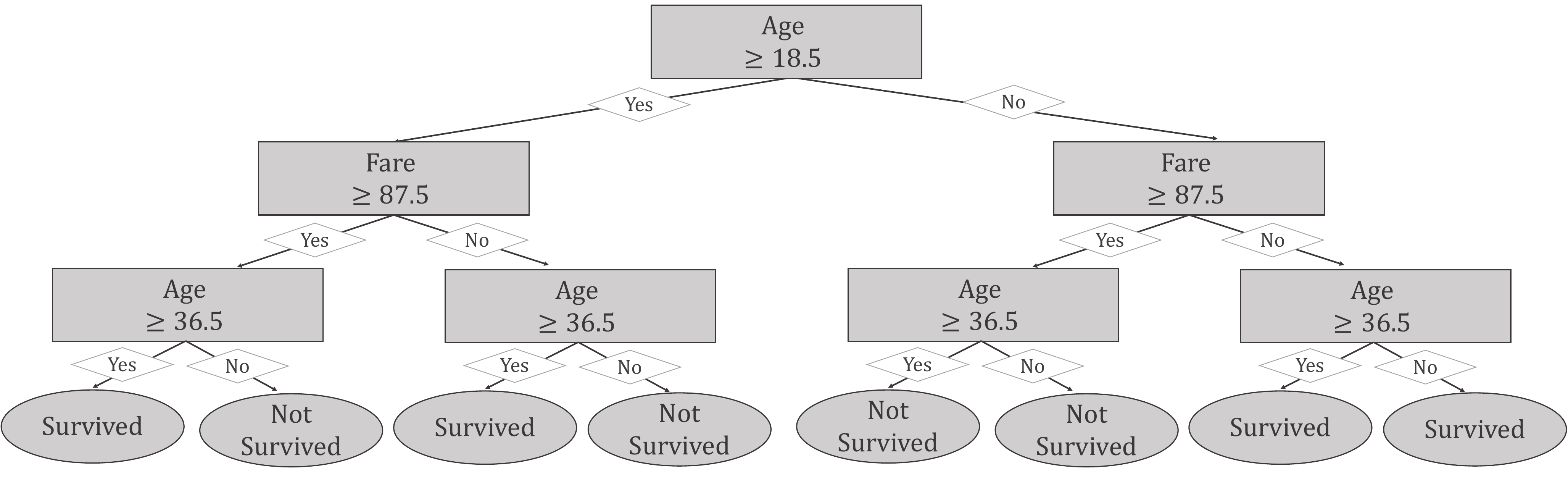}
        \caption{Oblivious DT structure.}
        \label{fig:decision_tree_oblivious}
    \end{subfigure}
    \caption[Comparison of Standard and Oblivious DTs]{\textbf{Comparison of Standard and Oblivious DTs.} (A) shows a standard DT for the exemplary Titanic dataset and (B) shows a similar oblivious DT, where all nodes at a certain level share the same splitting feature and threshold.}
    \label{fig:odt}
\end{figure}

\paragraph{Method}
NODE builds upon the concept of oblivious DTs (ODTs), where all nodes at the same depth share the same splitting feature and threshold (see Figure~\ref{fig:odt}). ODTs are not novel to NODE, but have already previously gained attention as they are the core building block of CatBoost~\citep{prokhorenkova2018catboost}. While ODTs lose explanatory power over standard DTs with the same depth, they are highly efficient in computation and therefore frequently used in ensembles. Traditional ODTs are non-differentiable, but NODE introduces a differentiable approximation using the \textit{entmax} transformation~\citep{entmax} for both feature selection and decision routing. Accordingly, the DTs used in NODE are oblique and oblivious.

\paragraph{Oblivious DTs}
Formally, in an oblivious DT of depth \( d \), each node splits the data based on a splitting feature \( x_\iota \in \mathbb{R} \) and a threshold \( \tau \in \mathbb{R} \). The leaf output of the tree is represented as a vector \( \boldsymbol{\lambda} \in \mathbb{R}^{2^d} \), with the tree output defined as:

\begin{equation}
o(\boldsymbol{x}) = \boldsymbol{\lambda}[\mathbb{S}_{\text{Heaviside}}(\boldsymbol{x}|\iota_0, \tau_0), \dots, \mathbb{S}_{\text{Heaviside}}(\boldsymbol{x}|\iota_{d-1}, \tau_{d-1})],
\end{equation}

where \( \mathbb{S}_{\text{Heaviside}}(\cdot) \) is the Heaviside function, resulting in binary decisions at each node.

\paragraph{Differentiable feature choice}
In NODE, this process is made differentiable by replacing the splitting mechanism with a soft version based on the entmax function. The choice of feature is now determined by a weighted sum of input features through a learnable feature selection matrix \( \boldsymbol{\iota} \in \mathbb{R}^{N}  \) at each depth, where the weights are computed as:

\begin{equation}
c(\boldsymbol{x} | \boldsymbol{\iota}) = \sum_{i=0}^{n-1} x_i \cdot \text{entmax}_\alpha(\iota_i).
\end{equation}

\paragraph{Differentiable splitting}
The decision function \( \mathbb{S}_{\text{Heaviside}}(\boldsymbol{x}|\iota, \tau) \) is relaxed using the entmax function \( S_\alpha(x) = \text{entmax}_\alpha([x, 0]) \), scaled by a learnable temperature parameter. Thus, the decision process at each node becomes a smooth function of the input.
As a result, the choice function \( \mathbb{S}(\boldsymbol{x}| \boldsymbol{\iota}, \boldsymbol{\tau}) = S_\alpha\left(\frac{c(\boldsymbol{x} | \boldsymbol{\iota}) - \tau}{s}\right) \) represents the probability that the feature choice \( c(\boldsymbol{x} | \boldsymbol{\iota}) \) takes a value below the learned threshold \( \tau \), scaled by a learnable parameter \( s \).

\paragraph{Decision routing}
The indicator function \( \mathbb{L}(x) \), which determines the decision routing within the tree, is constructed as a tensor of size \( 2^d \), where each entry corresponds to a possible path through the tree. It is defined as the outer product of the decision probabilities at each depth of the tree:

\begin{equation}
\mathbb{L}(\boldsymbol{x}) = \begin{bmatrix} 
\mathbb{S}(\boldsymbol{x}| \boldsymbol{\iota}_0, \boldsymbol{\tau}_0) \\ 
1 - \mathbb{S}(\boldsymbol{x}| \boldsymbol{\iota}_0, \boldsymbol{\tau}_0)
\end{bmatrix} 
\otimes
\begin{bmatrix} 
\mathbb{S}(\boldsymbol{x}| \boldsymbol{\iota}_1, \boldsymbol{\tau}_1) \\ 
1 - \mathbb{S}(\boldsymbol{x}| \boldsymbol{\iota}_1, \boldsymbol{\tau}_1)
\end{bmatrix} 
\otimes \cdots \otimes
\begin{bmatrix} 
\mathbb{S}(\boldsymbol{x}| \boldsymbol{\iota}_{d-1}, \boldsymbol{\tau}_{d-1}) \\ 
1 - \mathbb{S}(\boldsymbol{x}| \boldsymbol{\iota}_{d-1}, \boldsymbol{\tau}_{d-1})
\end{bmatrix}.
\end{equation}

\paragraph{Output definition}
The final output of the tree is computed as a weighted sum of the leaf output tensor \( \boldsymbol{\lambda} \) with weights from the indicator function \( \mathbb{L} \), which can be calculated as a dot product:

\begin{equation}
\hat{o}(x) = \boldsymbol{\lambda} \cdot \mathbb{L}(\boldsymbol{x}).
\end{equation}

\paragraph{Architecture}
NODE layers can be stacked into a multi-layer architecture, similar to neural networks.
Each NODE layer consists of multiple differentiable ODTs, and the outputs of all trees in the layer are concatenated to form the input for the next layer. This design enables hierarchical representation learning and allows the model to capture complex feature interactions. Specifically, NODE follows a DenseNet architecture~\citep{huang2017densely} where each layer output is additionally concatenated with the input layer and the prediction is a sum from all layers (compared to only the last layer for standard deep networks).

\paragraph{Training and optimization}
NODE models are trained end-to-end with an Adam optimizer including stochastic weight averaging~\citep{gouk2019stochastic}, where all model parameters (feature selectors, thresholds, and response tensors) are optimized simultaneously. Additionally, NODE employs a data-aware initialization to improve training stability and convergence.

\paragraph{Limitations}
While NODE is an efficient ensembling method that enables end-to-end gradient-based training, its individual trees exhibit comparatively limited explanatory power, as the authors opted for oblivious trees over standard DTs to enhance efficiency. Moreover, the trees are oblique and soft, which diminishes both the beneficial inductive bias of axis-aligned splits and the overall interpretability.

\subsubsection*{Learning Representations for Axis-Aligned Decision Forests} \label{sec:neural_axisaligned_forests}

While existing methods for learning tree ensembles often require structural changes, such as relaxing axis-aligned splits to oblivious ones, \citet{bruch2020learning} use input perturbation as a purely analytical procedure to approximate the gradients of the ensemble. 
As a result, it is possible to learn and fine-tune a hard, axis-aligned DT ensemble. Although their method makes this feasible, the authors decided not to learn the split index but only the split threshold, in order to avoid potential changes to the model's structure. Learning both, the split index and threshold can be particularly important when aiming for an interpretable DT model, as opposed to a large ensemble, since the number of possible splits is limited. %

\paragraph{Method}

The core idea is to combine a decision forest with an embedding function which can be for instance an identity function, a linear function or a neural network. This setup allows the decision forest to either learn from tabular data directly (i.e., using an identity function) as well as to learn and fine-tune data representations, effectively bridging the gap between decision forests and deep learning. To achieve differentiability, the authors employ an input perturbation technique instead of smoothing the structure of the tree: The input to the decision forest is perturbed using a zero-mean Gaussian noise, enabling an approximation of the gradient through an expected value calculation.
The input perturbation approach is defined as follows. Given an input vector $\boldsymbol{x} \in \mathbb{R}^n$, where $n$ is the dimensionality of the original feature space, a projection function $\mathcal{E}: \mathbb{R}^n \to \mathbb{R}^k$ is used to map the input to a $k$-dimensional embedding space, where $k$ is usually smaller than $n$. The embedding function $\mathcal{E}$ is implemented as a neural network, which learns to extract relevant features from the raw input.
The perturbed input, denoted as $z$, is sampled from a Gaussian distribution centered around the embedding $\mathcal{E}(x)$. Formally:

\begin{equation}
    \boldsymbol{z} \sim \mathcal{N}(\mathcal{E}(\boldsymbol{x}), \sigma^2 \mathcal{I}),
\end{equation}

where $\sigma$ is the standard deviation that controls the degree of perturbation, and $\mathcal{I} \in \mathbb{R}^{k \times k}$ is the identity matrix to ensure that the perturbation is equally distributed across all directions in the embedding space.
The decision forest takes the perturbed input $\boldsymbol{z}$ and produces an output. Since the perturbation introduces randomness, the output of the decision forest becomes a random variable. To approximate gradients for training, the expected value of the decision forest output over the distribution of $\boldsymbol{z}$ is computed:

\begin{equation}
    \mathcal{F}_{\sigma}(\boldsymbol{x}) = \mathbb{E}_{\boldsymbol{z} \sim \mathcal{N}(\mathcal{E}(\boldsymbol{x}), \sigma^2 \mathcal{I})}[\mathcal{F}(\boldsymbol{z})],
\end{equation}

where  $\mathbb{E}$ is the expectation operator, which represents the expected value of a random variable, $\mathcal{F}(\boldsymbol{z})$ represents the output of the decision forest for a given perturbed input $\boldsymbol{z}$ and $\mathcal{F}_{\sigma}(x)$ denotes the expected output of the decision forest under the Gaussian perturbation of the embedding $\mathcal{E}(\boldsymbol{x})$.
The expectation $\mathcal{F}_{\sigma}(\boldsymbol{x})$ effectively smooths out the non-differentiable nature of the decision forest by averaging over multiple noisy versions of the input. The final objective is to minimize a loss function that depends on the smoothed output $\mathcal{F}_{\sigma}(\boldsymbol{x})$. This smoothing allows for an approximate gradient to be computed, which can then be used to update the parameters during training. This approach makes it possible to train the entire model end-to-end, effectively combining the strengths of both deep learning and DTs. 

\paragraph{Limitations}
In contrast to existing work, the decisions remain hard and axis-aligned.
Although their method makes this feasible in theory, the authors decided not to learn the split index but only the split threshold. This decision was made in order to avoid potential changes to the model's structure, which can lead to instability in the training. However, learning the split index is particularly important when aiming for an interpretable DT model, as opposed to a large ensemble, since the number of possible splits is limited. Moreover, the analytical approximation of gradients through input perturbation can pose challenges when integrating with existing approaches in frameworks such as reinforcement learning or multimodal learning.

\section{Summary and Research Gap} \label{sec:summary_related}

In this chapter, we provided a comprehensive overview of various DT learning algorithms, including greedy, lookahead, evolutionary, optimal, and gradient-based methods. Each approach offers distinct advantages and limitations, which we have critically evaluated using illustrative examples and synthetic datasets:

\begin{itemize}
    \item \textbf{Greedy methods} \hspace{1mm} Algorithms like CART and C4.5 employ a top-down, recursive partitioning strategy, relying on impurity measures to guide splits. While computationally efficient, these methods are prone to local optima and constrain the search space, as decisions are made based on immediate impurity reduction without considering long-term effects.
    \item \textbf{Lookahead decision trees} \hspace{1mm} To mitigate the limitations of greedy approaches, lookahead DTs incorporate foresight by evaluating multiple potential future splits. Although this can enhance performance compared to purely greedy methods, the computational cost increases exponentially with the lookahead depth, limiting scalability.
    \item \textbf{Evolutionary algorithms} \hspace{1mm} These algorithms leverage global optimization principles inspired by natural selection. They are effective in escaping local optima and discovering complex feature interactions. However, their high computational demands and extensive hyperparameter tuning requirements pose significant challenges.
    \item \textbf{Optimal decision trees} \hspace{1mm} Optimal DTs aim to find globally optimal solutions using advanced optimization techniques like MIO and branch-and-bound algorithms. Despite their theoretical advantages, these methods often suffer from scalability issues, poor generalization, and require data discretization, which can lead to information loss.
\end{itemize}

In addition, a critical limitation shared by all non-gradient-based methods is their reduced flexibility, particularly regarding integration within modern machine learning pipelines. Contemporary architectures, for instance those used in multimodal and reinforcement learning, rely on differentiability and joint optimization across model components. The reliance on alternative optimization algorithms prevents seamless integration into these differentiable systems, thereby limiting their applicability in such contexts.

\paragraph{What about neural networks?} Given the limitations of traditional DT methods, one might question: \emph{Why not simply use neural networks for these tasks?} 
Neural networks excel in predictive performance in several domains and can naturally integrate with gradient-based optimization frameworks. However, they lack the axis-aligned inductive bias, can be is essential in domains like tabular data~\citep{grinsztajn2022tree,beyazit2024inductive}. Furthermore, Neural Networks tend to operate as black-box models, which prevents interpretability, a key advantage of DTs, especially in applications demanding transparent decision-making.

\paragraph{Gradient-based decision tree methods} 
Therefore, recent approaches have integrated DT learning with gradient-based optimization, enabling end-to-end training. 
These methods typically leverage continuous relaxation techniques to allow for differentiable decision functions, which facilitates integration with deep learning frameworks. This integration supports not only end-to-end learning but also the potential for joint optimization with other model components in complex architectures. However, they typically rely on soft, oblique splits, which compromise beneficial inductive biases and diminish interpretability. While there are methods that attempt to address these limitations, they often introduce their own constraints in terms of scalability or flexibility. To date, no approach enables a seamless and scalable integration with existing gradient-based methods in frameworks such as multimodal learning or reinforcement learning, while preserving the hard, axis-aligned structure.

\paragraph{Research gap}
Despite the advancements in DT learning algorithms, there remains a significant gap in the ability to simultaneously achieve a global optimization of all tree parameters, strong inductive biases, scalability, and flexibility. Each existing method tends to optimize one or two of these aspects at the expense of others:

\begin{itemize}
    \item \textbf{Local optimality and constrained search space} \hspace{1mm} The most commonly used method for DT learning, a greedy procedure, suffers from a constrained search space and local optimality. Decisions are made sequentially, which limits the model's ability to discover solutions across the full search space potentially leading to suboptimal solutions.
    \item \textbf{Maintaining axis-alignment} \hspace{1mm} While oblique and soft DTs enhance flexibility, they compromise the beneficial inductive biases inherent in hard, axis-aligned splits. Axis-alignment supports learning efficiency through structured bias, which can also contribute to model interpretability as a secondary advantage.
    \item \textbf{Flexibility and integration} \hspace{1mm} Existing methods often lack the flexibility to accommodate custom objective functions and seamless integration into approaches for broader machine learning frameworks, such as those for multimodal learning or reinforcement learning which typically require a gradient-based optimization
\end{itemize}

In this thesis, we address the identified gap by proposing a novel, gradient-based method for learning hard, axis-aligned DT that overcomes the limitations of existing approaches. The proposed method aims to eliminate local optimality originated from sequential splitting by performing a global search over the entire parameter space, ensuring that the learned models are not constrained by sequential or local decisions. It also aims to maintain axis-alignment and hard splits to preserve beneficial inductive biases, with interpretability as an additional advantage when applicable. Furthermore, the method is designed to enhance flexibility, allowing the use of custom objective functions and facilitating integration into existing approaches for machine learning frameworks, including those for multimodal learning and reinforcement learning. By bridging these gaps, our method contributes a versatile, robust, and efficient solution to the ongoing challenges in DT learning and interpretable machine learning.

\chapter[Learning Axis-Aligned Decision Trees with Gradient Descent]{Learning Axis-Aligned Decision Trees with Gradient Descent} \label{cha:methodology}

\textbf{The following section was already partially published in \citet{gradtree} and \citet{grande}. For a detailed summary of the individual contributions, please refer to Section~\ref{sec:contributions}.} \\

In the previous chapter, we have introduced different methods to learn DTs and identified a research gap that, until now, has not been addressed appropriately.
In this chapter, we propose \textbf{Grad}ient-based decision \textbf{Tree}s (GradTree), a novel approach for learning hard, axis-aligned DTs based on a joint optimization of all tree parameters through gradient descent. 
Using a gradient-based optimization, we can overcome the limitations of greedy approaches, which are constrained by sequentially selecting optimal splits, as illustrated by the examples from the previous chapter.
Similar to the optimization in neural networks, our method can converge to a local optimum that offers good generalization, and thus provides an advantage over alternative non-greedy methods like optimal DTs~\citep{murtree,aglin2020learning}, which often suffer from severe overfitting~\citep{zharmagambetov2021non}.
Thereby, GradTree maintains the beneficial inductive bias of hard, axis-aligned splits which is in contrast to existing gradient-based methods typically using soft and oblique splits.
Furthermore, a gradient-based optimization offers significant flexibility, enabling the straightforward optimization of custom objectives and the integration of tailored loss functions. Moreover, the use of backpropagation to adjust model parameters facilitates seamless incorporation into existing frameworks, such as multimodal learning (Section~\ref{sec:multimodal}) or reinforcement learning (Chapter~\ref{cha:sympol}). 
We begin by introducing the general framework and core formulation of GradTree (Section~\ref{sec:method_gradtree}), outlining its key components.
Building on this foundation, we provide a theoretical discussion and comparative analysis of DT learning methods (Section~\ref{sec:discussion_related}). 
Furthermore, we extend the proposed method from individual trees to DT ensembles (Section~\ref{sec:method_grande}) as a performance-interpretability trade-off.

\section{Core Model Formulation for Gradient-Based Decision Trees} \label{sec:method_gradtree}

In the following, we will present the core model formulation for GradTree\footnote{Our implementation is publically available under: \url{https://github.com/s-marton/GradTree}.}, which is the fundamental framework for this thesis. This includes an arithmetic DT formulation (Section~\ref{ssec:arithmetic_dt}), a dense DT representation (Section~\ref{ssec:dense_rep}), and techniques to address non-differentiability during backpropagation (Section~\ref{ssec:adjusted_backprop}). Additionally, we discuss the selection of differentiable split functions (Section~\ref{ssec:splitting_function}), the tree routing function (Section~\ref{ssec:training}), and the optimization procedure based on gradient descent (Section~\ref{ssec:gradient_descent_tree}).

\subsection{Arithmetic Decision Tree Formulation} \label{ssec:arithmetic_dt}
Here, we introduce a notation for DTs with respect to their parameters. We formulate DTs as an arithmetic function based on addition and multiplication, rather than as a nested concatenation of rules (Section~\ref{sec:dt_formalization}), which is necessary for a gradient-based learning. 
Note that our notation and training procedure assume fully-grown (i.e. complete, full) DTs. After training, we apply a basic post-hoc pruning to reduce the tree size for application.
Our formulation aligns with \citet{kontschieder2015neuraldecisionforests}. 
However, they only consider stochastic routing and oblique trees, whereas our formulation emphasizes deterministic routing and axis-aligned trees.
For a DT of depth $d$, the parameters include one split threshold and one feature index for each internal node, represented as vectors \(\boldsymbol{\tau} \in \mathbb{R}^{2^d-1}\) and \(\boldsymbol{\iota} \in \mathbb{N}^{2^d-1}\) respectively, where \(2^d-1\) equals the number of internal nodes in a fully-grown DT.
Additionally, each leaf node comprises a class membership, in the case of a classification task, which we denote as the vector $\boldsymbol{\lambda} \in \mathcal{C}^{2^d}$, where $\mathcal{C}$ is the set of classes and $2^d$ equals the number of leaf nodes in a fully-grown DT.%
Formally, a DT can be expressed as a function $ g(\cdot | \boldsymbol{\tau}, \boldsymbol{\iota}, \boldsymbol{\lambda}): \mathbb{R}^n \rightarrow \mathcal{C}$ with respect to its parameters:
\begin{equation}\label{eq:dt_prediction}
    g(\boldsymbol{x} | \boldsymbol{\tau}, \boldsymbol{\iota}, \boldsymbol{\lambda}) = \sum_{l=0}^{2^d-1} \lambda_l \, \mathbb{L}(\boldsymbol{x} | l, \boldsymbol{\tau}, \boldsymbol{\iota}).
\end{equation}
The function $\mathbb{L}$ indicates whether a sample $\boldsymbol{x} \in \mathbb{R}^n$ belongs to a leaf $l$, and can be defined as a multiplication of the split functions of the preceding internal nodes. We define the split function $\mathbb{S}$ as a Heaviside step function
\begin{equation}\label{eq:split_heaviside}
    \mathbb{S}_{\text{Heaviside}}(\boldsymbol{x} | \iota, \tau) =
    \begin{cases}
        1 \text{if } x_\iota \ge \tau, \\
        0 \text{otherwise},
    \end{cases}
\end{equation}
where $\iota$ is the index of the feature considered at a certain split and $\tau$ is the corresponding threshold.
By enumerating the internal nodes of a fully-grown tree with depth $d$ in a breadth-first order, we can now define the indicator function $\mathbb{L}$ for a leaf $l$ as

\begin{equation}\label{eq:indicator_func}
\begin{split}
    \mathbb{L}(\boldsymbol{x} | l, \boldsymbol{\tau}, \boldsymbol{\iota}) = \prod^{d-1}_{j=0} & \left(1 - \mathfrak{p}(l,j) \right) \, \mathbb{S}(\boldsymbol{x} | \tau_{\mathfrak{i}(l,j)}, \iota_{\mathfrak{i}(l,j)}) \\
    + & \mathfrak{p}(l,j) \, \left( 1 - \mathbb{S}(\boldsymbol{x} | \tau_{\mathfrak{i}(l,j)}, \iota_{\mathfrak{i}(l,j)}) \right).
\end{split}
\end{equation}

Here, $\mathfrak{i}$ is the index of the internal node preceding a leaf node $l$ at a certain depth $j$ and can be calculated as

\begin{equation}\label{eq:index_calc}
    \mathfrak{i}(l,j) = 2^{j} + \left\lfloor \frac{l}{2^{d-j}} \right\rfloor - 1.
\end{equation}

Additionally, $\mathfrak{p}$ indicates whether the left ($\mathfrak{p} = 0$) or the right branch ($\mathfrak{p} = 1$) was taken at the internal node preceding a leaf node $l$ at a certain depth $j$. We can calculate $\mathfrak{p}$ as

\begin{equation}\label{eq:path_calc}
    \mathfrak{p}(l,j) = \left\lfloor \frac{l}{2^{d-(j+1)}} \right\rfloor \bmod 2.
\end{equation}

As becomes evident, DTs involve non-differentiable operations in terms of the split function, including the split feature selection (Equation~\ref{eq:split_heaviside}). This precludes the application of backpropagation for learning the parameters. 
Specifically, to efficiently learn a DT using backpropagation, we must address three challenges:
\begin{itemize}
    \item[] \textbf{C1} The index $\iota$ for the split feature selection is defined as $\iota \in \mathbb{N}$. However, the index $\iota$ is a parameter of the DT and a gradient-based optimization requires $\iota \in \mathbb{R}$.    
    \item[] \textbf{C2} The split function $\mathbb{S}(\boldsymbol{x}| \iota, \tau)$ is a Heaviside step function with an undefined gradient for $x_\iota = \tau$ and $0$ gradient elsewhere, which precludes an efficient optimization. 
    \item[] \textbf{C3} Leafs in a vanilla DT comprise a class membership $\lambda \in \mathcal{C}$. To calculate an informative loss and optimize the leaf parameters with gradient descent, we need $\lambda \in \mathbb{R}$.       
\end{itemize}

Additionally, the computation of the internal node index $\mathfrak{i}$ (Equation~\ref{eq:index_calc}) and path position $\mathfrak{p}$ (Equation~\ref{eq:path_calc}) involves non-differentiable operations. However, given our focus on fully-grown trees, these values remain constant, allowing for their computation prior to the optimization process.

\begin{figure}[tb]
\centering
\begin{subfigure}{0.5\textwidth}
   \centering
  \includegraphics[width=0.9\columnwidth]{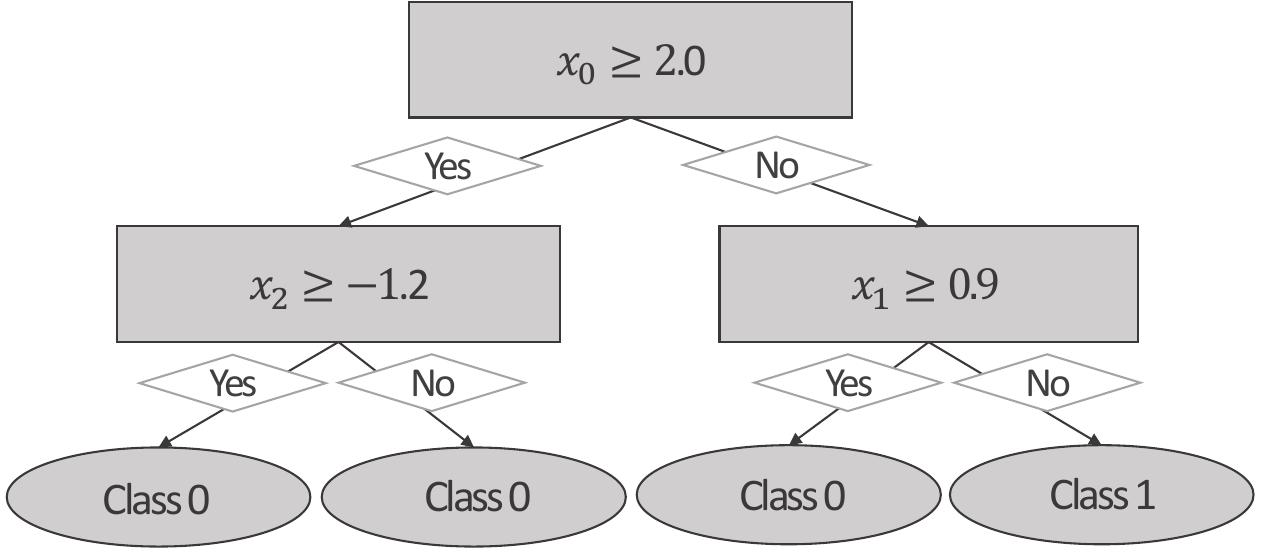}
  \caption{Vanilla DT Representation}
  \label{fig:old_rep_2}
  
\end{subfigure}%
\begin{subfigure}{0.5\textwidth}
  \centering
  \includegraphics[width=0.9\columnwidth]{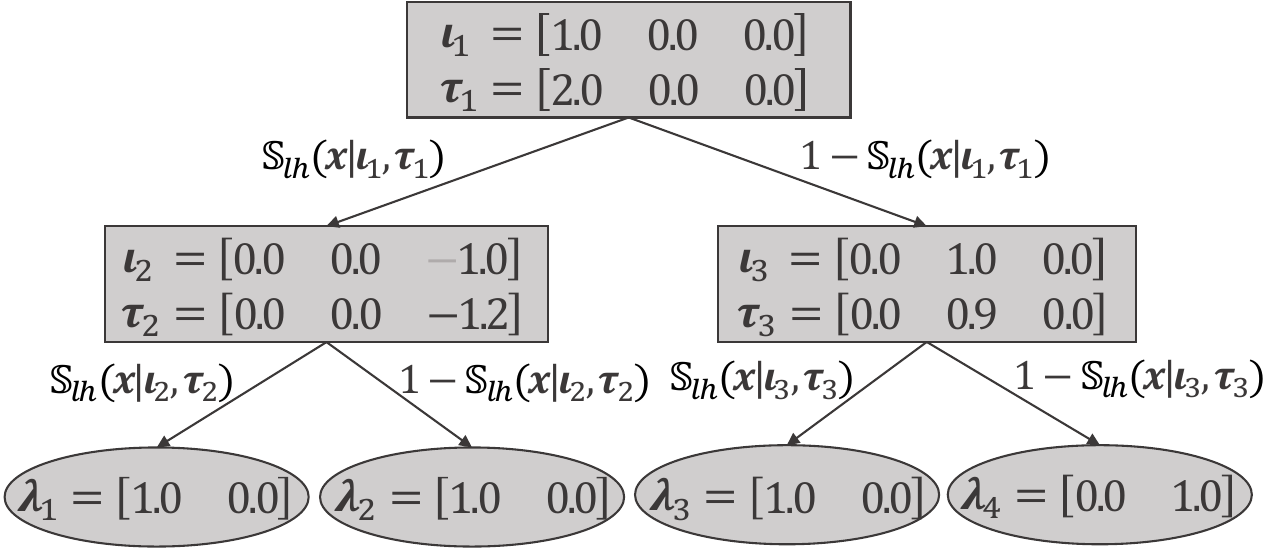}
  \caption{Dense DT Representation}
  \label{fig:new_rep_dense}
\end{subfigure}
\caption[Standard vs. Dense DT Representation]{\textbf{Standard vs. Dense DT Representation.} Comparison of a standard (A) and its equivalent dense representation (B) for an exemplary DT with depth $2$ and a dataset with $3$ variables and $2$ classes. Here, $\mathbb{S}_{\text{lh}}$ stands for $\mathbb{S}_\text{logistic\_hard}$ (Equation~\ref{eq:split_sigmoid_round}). }
\label{fig:dt_reps_dense}
\end{figure}

\subsection{Dense Decision Tree Representation}\label{ssec:dense_rep}
In this section, we present a differentiable representation of the feature indices $\boldsymbol{\iota}$ to facilitate gradient-based optimization, which is illustrated in Figure~\ref{fig:dt_reps_dense}.
To this end, we expand the vector $\boldsymbol{\iota} \in \mathbb{R}^{2^d-1}$ to a matrix $I \in \mathbb{R}^{2^d-1} \times \mathbb{R}^{n}$. This is achieved by one-hot encoding the feature index as $\boldsymbol{\iota} \in \mathbb{R}^{n}$ for each internal node.
This adjustment is necessary for the optimization process to account for the fact that feature indices are categorical instead of ordinal. 
Although our matrix representation for feature selection has parallels with that proposed by \citet{popov2019neural}, we introduce a novel aspect: A matrix representation for split thresholds. We denote this representation as $T \in \mathbb{R}^{2^d-1} \times \mathbb{R}^{n}$. Instead of representing a single value for all features, we store individual values for each feature, denoted as $\boldsymbol{\tau} \in \mathbb{R}^{n}$.
This modification is tailored to support the optimization process, particularly in recognizing that split thresholds are feature-specific and non-interchangeable. In essence, a viable split threshold for one feature may not be suitable for another. This adjustment acts as a memory mechanism, ensuring that a given split threshold is exclusively associated with the corresponding feature. 
Consequently, this refinement enhances the exploration of feature selection at every split during the training.
Besides the previously mentioned advantages, using a dense DT representation allows the use of matrix multiplications for an efficient computation.
Accordingly, we can reformulate the Heaviside split function (Equation~\ref{eq:split_heaviside}) as

\begin{equation} \label{eq:split_sigmoid}
    \mathbb{S}_{\text{logistic}}(\boldsymbol{x}| \boldsymbol{\iota}, \boldsymbol{\tau}) =  S \left( \sum_{i=0}^{n-1} \iota_i  x_i - \sum_{i=0}^{n-1} \iota_i  \tau_i \right),
\end{equation}

\begin{equation} \label{eq:split_sigmoid_round}
    \mathbb{S}_{\text{logistic\_hard}}(\boldsymbol{x}| \boldsymbol{\iota}, \boldsymbol{\tau}) =  \left\lfloor \mathbb{S}_{\text{logistic}}(\boldsymbol{x}| \boldsymbol{\iota}, \boldsymbol{\tau}) \right\rceil,
\end{equation}

where \(S(z)=\frac{1}{1+e^{-z}}\) denotes the logistic function and \(\lfloor \cdot \rceil\) represents rounding to the nearest integer.
In our context, with $\boldsymbol{\iota}$ being one-hot encoded, $\mathbb{S}_{\text{logistic\_hard}}(\boldsymbol{x}| \boldsymbol{\iota}, \boldsymbol{\tau}) = \mathbb{S}_{\text{Heaviside}}(\boldsymbol{x}| \iota, \tau)$ holds. %

\subsection{Backpropagation of Decision Tree Loss}\label{ssec:adjusted_backprop}
While the dense representation introduced previously emphasizes an efficient learning of axis-aligned DTs, it does not solve \textbf{C1}-\textbf{C3}. 
In this section, we will address those challenges by using the ST operator for backpropagation. 

\begin{figure}[tb]
\centering
\begin{subfigure}{0.5\textwidth}
   \centering
  \includegraphics[width=0.95\columnwidth]{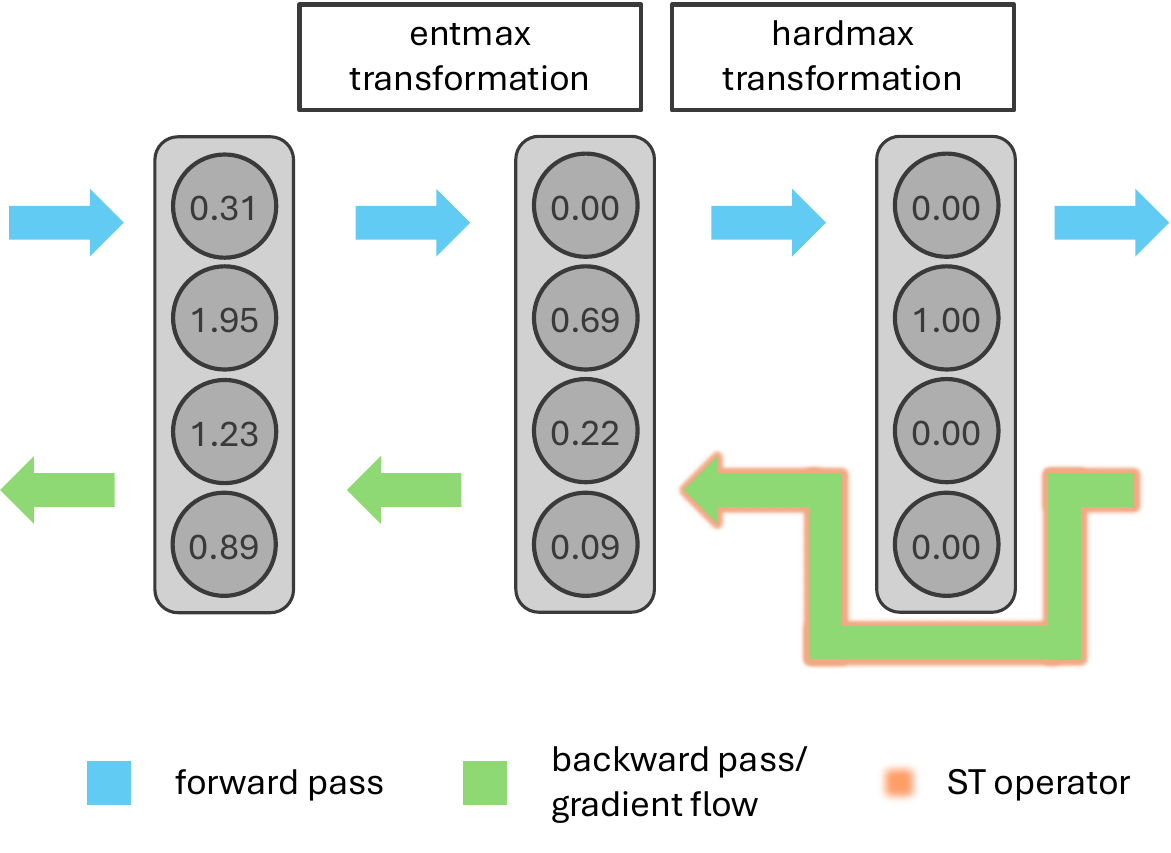}
  \caption{ST Operator for Axis-Aligned Splits}
  \label{fig:st_axis}
  
\end{subfigure}%
\begin{subfigure}{0.5\textwidth}
  \centering
  \includegraphics[width=0.95\columnwidth]{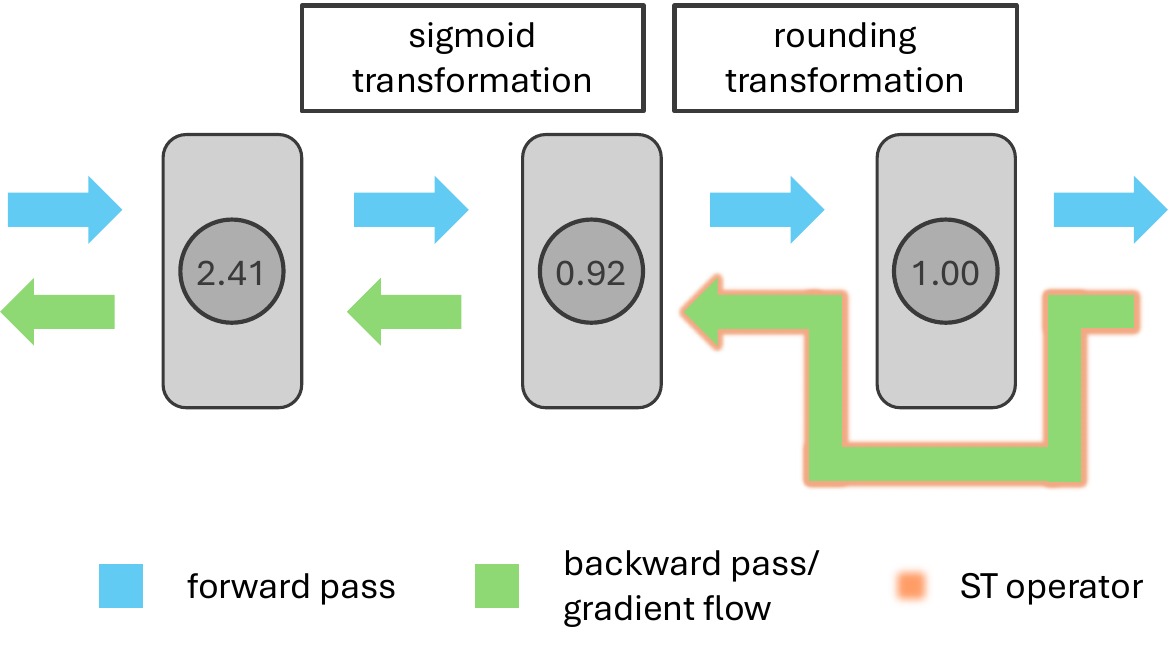}
  \caption{ST Operator for Hard Splits}
  \label{fig:st_hard}
\end{subfigure}
\caption[ST Operator Visualization.]{\textbf{ST Operator Visualization.} This figure illustrates the ST operator for handling discrete decisions in gradient-based optimization. 
For axis-aligned splits (A), an entmax transformation produces sparse probabilities, followed by a hardmax operation to obtain discrete outputs. The forward pass (blue arrows) propagates the hard values, while the backward pass (green arrows) surpasses the non-differentiable hardmax transformation and allows gradients to flow through the entmax transformation. 
For hard splits (B), a sigmoid transformation maps values to a continuous range in $[0, 1]$ before applying a rounding operation to get a discrete decision for whether to take the left ($1$) or right ($0$) path. The forward pass follows the discrete outputs, while the backward pass skips the rounding operation and enables gradients to pass through the sigmoid function only, approximating the non-differentiable rounding step. 
Both approaches leverage the ST operator (red lines) to ensure meaningful gradient flow despite discrete computations.}
\label{fig:st_visualization}
\end{figure}

\paragraph{Using the ST operator to deal with non-differentiable operations}
For the function value calculation in the forward pass, we need to assure that $\boldsymbol{\iota}$ is a one-hot encoded vector. This can be achieved by applying a hardmax function on the feature index vector for each internal node. However, applying a hardmax is a non-differentiable operation, which precludes gradient computation. To overcome this issue, we use the ST operator~\citep{bengio2013propagating}: For the forward pass, we apply the hardmax as is. For the backward pass, we exclude this operation and directly propagate back the gradients of $\boldsymbol{\iota}$. Accordingly, we can optimize the parameters of $\boldsymbol{\iota}$ where $\boldsymbol{\iota} \in \mathbb{R}$ while still using axis-aligned splits during training (\textbf{C1}).
However, this procedure introduces a mismatch between the forward and backward pass. To reduce this mismatch, we additionally perform an entmax transformation~\citep{entmax} to generate a sparse distribution over $\boldsymbol{\iota}$ before applying the hardmax. 
This is visualized conceptually in Figure~\ref{fig:st_axis}. The non-differentiable hardmax operation is excluded from the gradient flow, and therefore only the gradients from the entmax transformation are backpropagated.
Similarly, we employ the ST operator to ensure hard splits (Equation~\ref{eq:split_sigmoid_round}) by excluding rounding for the backward pass (\textbf{C2}). Using the sigmoid logistic function before applying the ST operator (see Equation~\ref{eq:split_sigmoid}) utilizes the distance to the split threshold as additional information for the gradient calculation. If the feature considered at an internal node is close to the split threshold for a specific sample, this will result in larger gradients compared to a sample that is more distant. Figure~\ref{fig:st_hard} provides a conceptual visualization of applying the ST operator for ensuring hard splits while maintaining a reasonable gradient flow.

\paragraph{On alternative approaches for addressing non-differentiability} %
When addressing non-differentiable operations in DT learning, we opted for the ST operator, though several alternatives exist, primarily from the field of discrete optimization. Here, we discuss these alternative approaches and explain why using plain ST operator proved most suitable for our needs.
A prominent alternative is the Gumbel-Softmax estimator~\citep{maddison2016concrete, gumbel}, which can be used either independently or in combination with the ST operator~\citep{paulus2020rao}. While both approaches address non-differentiability, they differ fundamentally in their underlying principles: Gumbel-Softmax (and ST Gumbel-Softmax) relies on a probabilistic interpretation of discrete variables and employs sampling for gradient estimation, whereas the ST operator directly approximates gradients.
The probabilistic approach of Gumbel-Softmax is well-suited for many applications, particularly in scenarios where uncertainty modeling is natural and beneficial. Consider discrete variational autoencoders~\citep{kingma2013auto,rolfe2017discrete}, where the goal is to learn discrete latent representations. For example, when learning interpretable factors from facial images (such as age groups, presence of glasses, or hair color) without explicit labels. Each latent variable represents a discrete attribute, and the probabilistic nature of Gumbel-Softmax aligns well with the inherent uncertainty in inferring these abstract features.
However, DTs present two types of discrete decisions where a probabilistic interpretation becomes problematic: (1) Feature selection at each node involves finding the optimal splitting feature through discrete optimization. This is fundamentally a deterministic selection problem where we seek the objectively best split, rather than modeling uncertainty over feature choices. (2) Binary splitting based on thresholds, where samples are deterministically routed based on feature values compared against fixed thresholds. Adding probabilistic elements to this process would contradict the deterministic nature of threshold-based routing.
Beyond theoretical concerns, stochastic routing and feature selection can severely impact optimization: Random variations in sample routing or feature selection at any node can propagate through the tree, disrupting the entire optimization process. Our empirical evaluation (Section~\ref{ssec:experiment_results}) demonstrates that direct gradient approximation using the ST operator consistently outperforms probabilistic alternatives like Gumbel-Softmax.
Similar limitations apply to other discrete optimization techniques, such as Implicit Maximum Likelihood Estimation~\citep{niepert2021implicit}, which requires probabilistic interpretation and sampling-based approaches that would introduce undesirable stochasticity into routing and feature selection.

\paragraph{Using probability distributions in the leafs for efficient loss calculation}
Furthermore, we need to adjust the leaf nodes of the DT to allow an efficient loss calculation (\textbf{C3}). Vanilla DTs contain the predicted class for each leaf node and are functions $g: \mathbb{R}^n \rightarrow \mathcal{C}$. We use a probability distribution at each leaf node and therefore define DTs as a function $g: \mathbb{R}^n \rightarrow \mathbb{R}^c$ where $c$ is the number of classes. Consequently, the parameters of the leaf nodes are defined as $L \in \mathbb{R}^{2^n} \times \mathbb{R}^c$ for the whole tree and $\boldsymbol{\lambda} \in \mathbb{R}^{c}$ for a specific leaf node. This adjustment allows the application of standard loss functions.%

\subsection{Differentiable Split Functions}\label{ssec:splitting_function}
\begin{figure}[tb]
\centering
\begin{subfigure}{0.33\textwidth}
   \centering
  \includegraphics[width=0.975\textwidth]{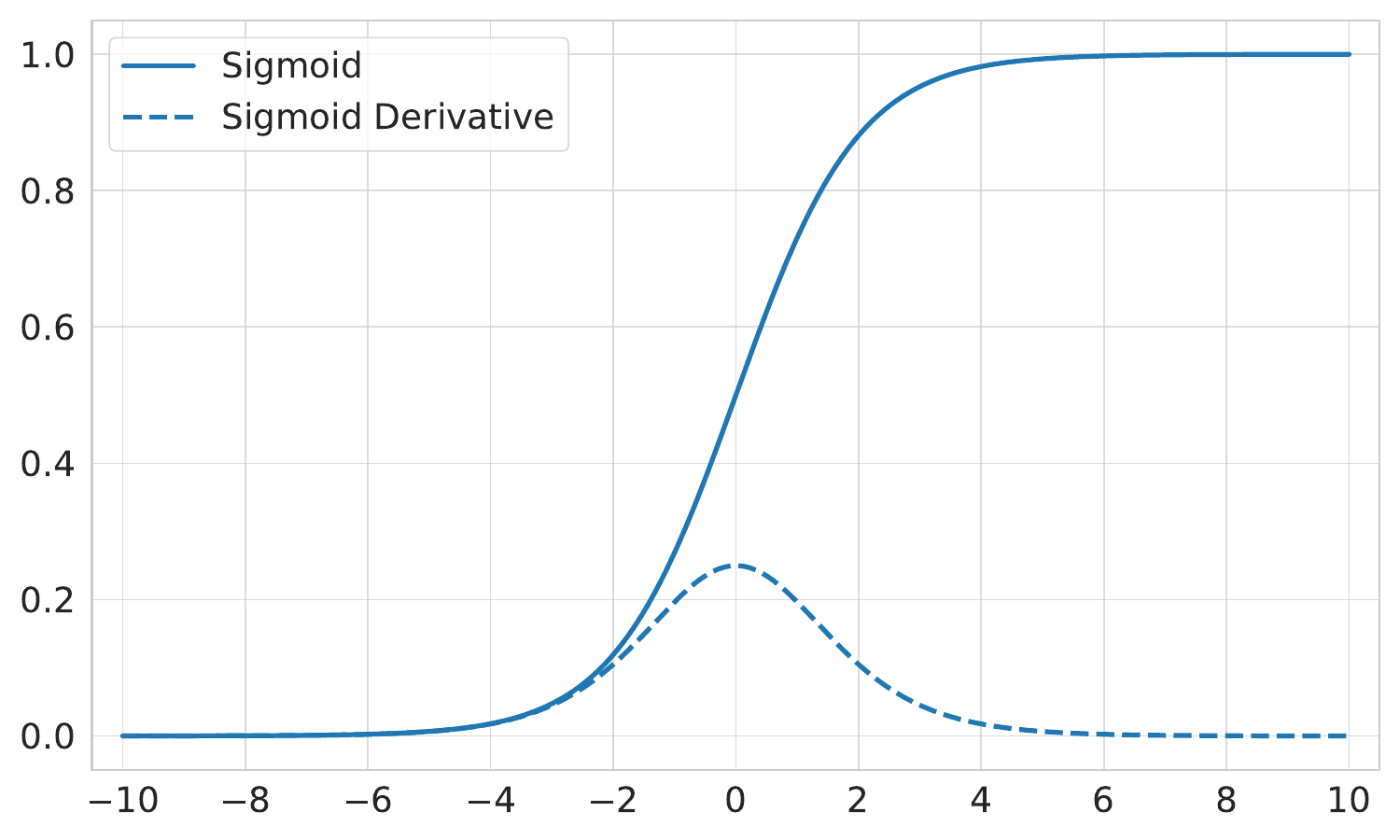}
  \caption{Sigmoid}
  \label{fig:sigmoid}
\end{subfigure}%
\begin{subfigure}{0.33\textwidth}
  \centering
  \includegraphics[width=0.975\textwidth]{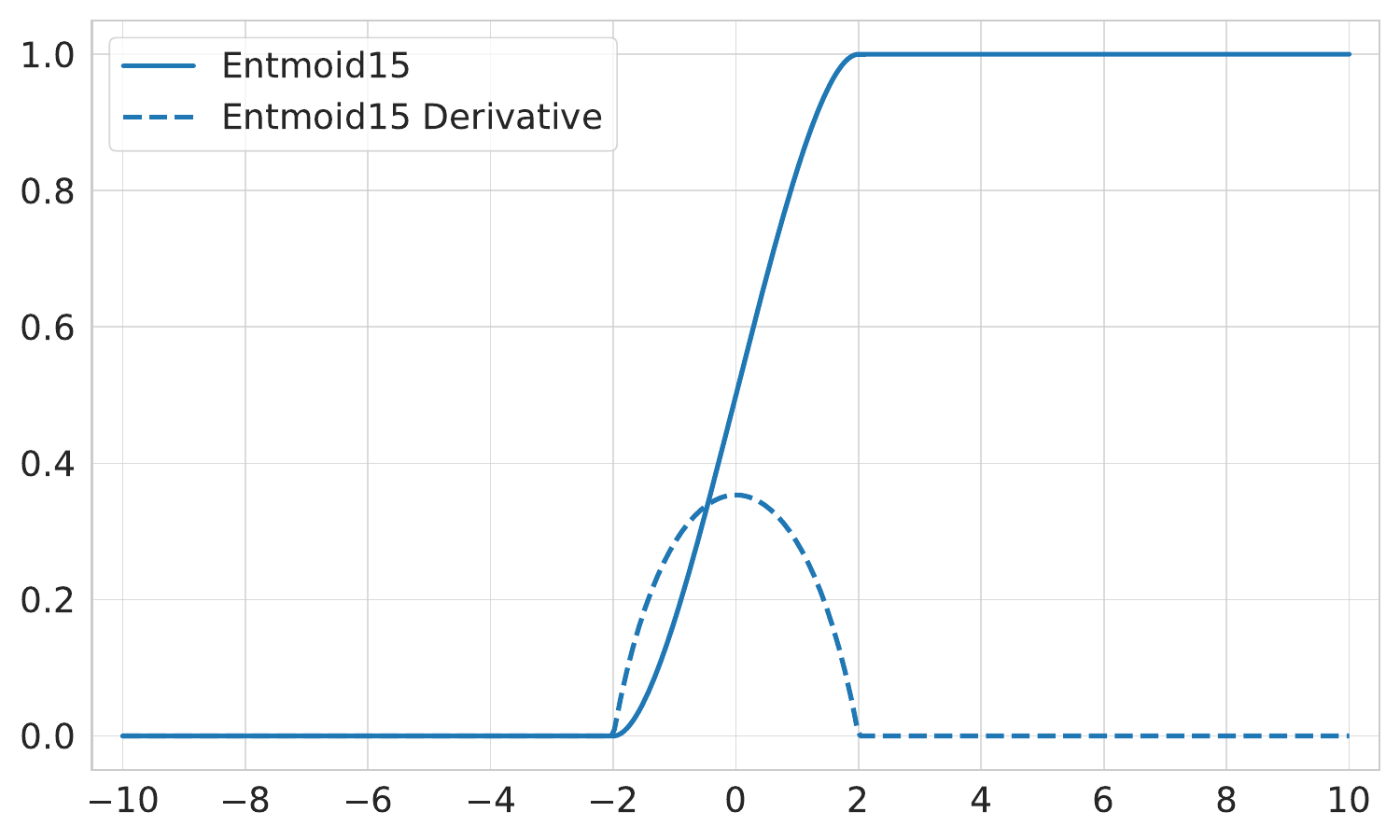}
  \caption{Entmoid}
  \label{fig:endmoid}
\end{subfigure}%
\begin{subfigure}{0.33\textwidth}
  \centering
  \includegraphics[width=0.975\textwidth]{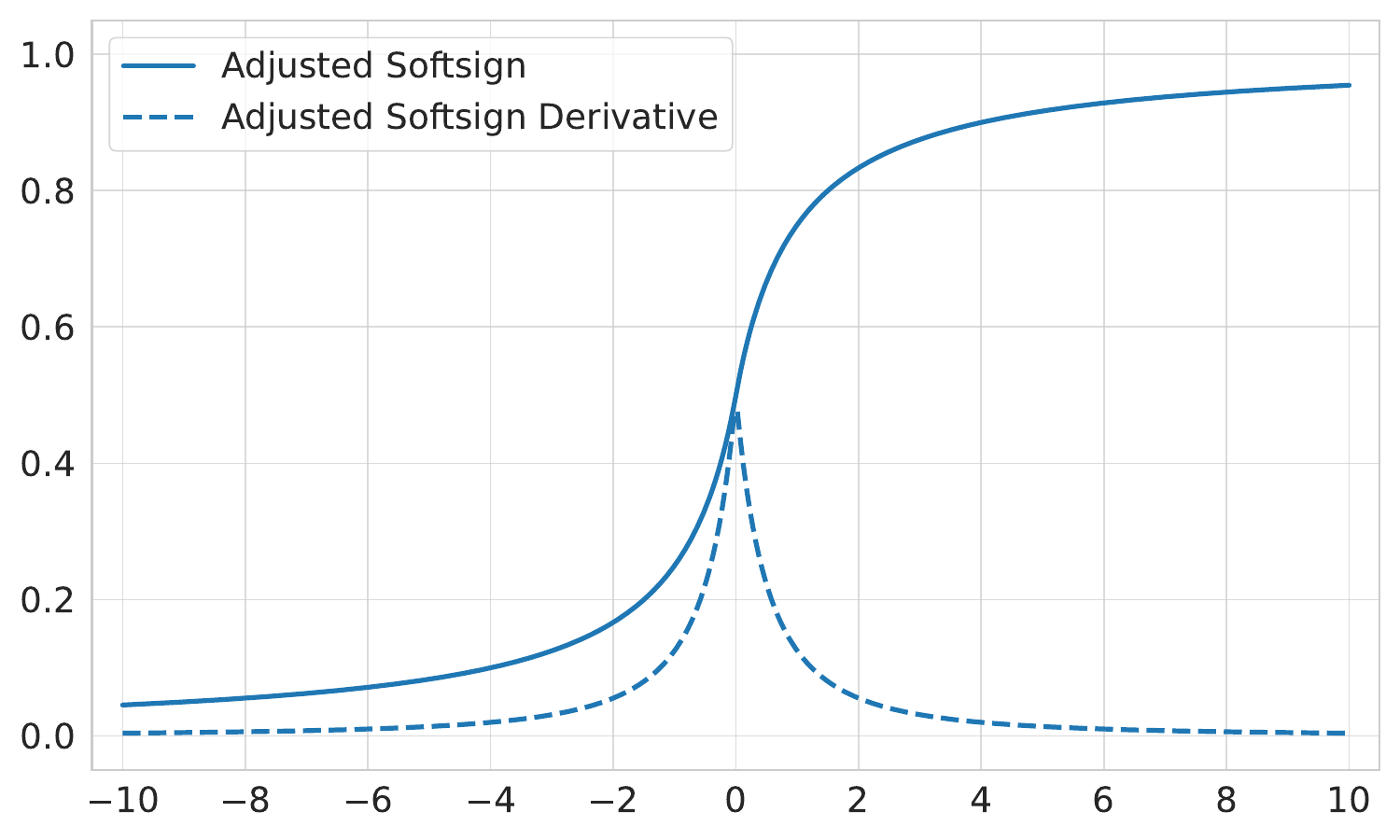}
  \caption{Adjusted Softsign}
  \label{fig:softsign}
\end{subfigure}
\caption[Comparison of Differentiable Split Functions]{\textbf{Comparison of Differentiable Split Functions.} The sigmoid function's gradient (A) declines smoothly, while entmoid's gradient (B) decays more rapidly but becomes zero for large values. The scaled softsign activation (C) has high gradients for small values but maintains a responsive gradient for large values, offering greater sensitivity.}
\label{fig:activation_comparison}            
\end{figure}

The Heaviside step function, which is commonly used as a split function in DTs, is non-differentiable. To address this challenge, various studies have proposed the employment of differentiable split functions. A predominant approach is the adoption of the sigmoid function, which facilitates soft decisions~\citep{soft1,soft2,soft3}. A more recent development in this field originated with the introduction of the entmax transformation~\citep{entmax}. Researchers utilized a two-class entmax (entmoid) function to turn the decisions more sparse~\citep{popov2019neural}. Further, \citet{node_gam} proposed a temperature annealing procedure to gradually turn the decisions hard. 
In Section~\ref{ssec:dense_rep} and Section~\ref{ssec:adjusted_backprop}, we introduced an alternative method for generating hard splits by using an ST operator after a sigmoid split function to generate hard splits (Equation~\ref{eq:split_sigmoid}). While this allows using hard splits for calculating the function values, it also introduces a mismatch between the forward and backward pass. 
However, applying a sigmoid function before the ST operator allows us to leverage additional information from the gradient, capturing the distance between a feature value and the threshold.
Accordingly, the gradient behavior plays a pivotal role in ensuring effective differentiation (see Figure~\ref{fig:st_visualization}), especially in scenarios where input values are close to the decision threshold. The traditional sigmoid function can be suboptimal due to its smooth gradient decline. Entmoid, although addressing certain limitations of sigmoid, still displays an undesirable gradient behavior. Specifically, its gradient drops to zero when the difference in values is too pronounced. This can hinder the model's ability to accommodate samples that exhibit substantial variances from the threshold. 
Therefore, we propose using a softsign function, scaled to $(0,1)$, as a differentiable split function: %
\begin{equation}
    S_{\text{ss}}(z) = \frac{1}{2} \left( \frac{z}{1 + |z|} + 1 \right).
\end{equation}
The distinct gradient characteristics of the softsign, which are pronounced if samples are close to the threshold, reduce sharply but maintain responsive gradients if there is a large difference between the feature value and the threshold. These characteristics make it well-suited for differentiable splitting. This concept is visualized in Figure~\ref{fig:activation_comparison}. 
During preliminary experiments, we observed that for the case of individual trees, there is no significant difference between using a sigmoid or softsign function as differentiable split function. However, for the case of DT ensembles, which will be introduced in Section~\ref{sec:grande}, we can empirically show a substantial performance when using softsign as split function (Table~\ref{tab:ablation_study_summary}). We argue that this effect arises because the drastic changes in split values induced by using softsign rather than sigmoid lead to substantial modifications in the tree architecture, introducing instability during training of individual trees. For ensembles, however, this impact is mitigated, yielding faster convergence and greater diversity within the ensemble when using a softsign function for splitting.

\begin{algorithm}[tb]

\caption{Tree Pass Function}   \label{alg:forward} 

\begin{algorithmic}[1] 
    
    \Function{pass}{$I, T, L, \boldsymbol{x}$}
        \State{$I \leftarrow \text{entmax}(I)$} %
        \State{$I \leftarrow I - c^*_1 \quad \text{where } c^*_1 = I - \text{hardmax}(I)$} \Comment{ST operator}
        
        \State{$\hat{\boldsymbol{y}} \leftarrow {[0]}^c$}
        \For{$l = 0, \dots, 2^d-1$}
            \State{$p \leftarrow 1$}
            \For{$j = 1, \dots, d$}
                \State{$ \mathfrak{i} \leftarrow 2^{j-1} + \left\lfloor \frac{l}{2^{d-(j-1)}} \right\rfloor - 1 $} \Comment{Equation~\ref{eq:index_calc}}            
            
                \State{$\mathfrak{p} \leftarrow \left\lfloor \frac{l}{2^{d-j}} \right\rfloor \bmod 2$} \Comment{Equation~\ref{eq:path_calc}}

                \State{$s \leftarrow S\left(\sum\limits^{n-1}_{i=0} T_{\mathfrak{i}, i} \, I_{\mathfrak{i}, i} - \sum\limits^{n-1}_{i=0} x_{i} \, I_{\mathfrak{i}, i}\right)$} \Comment{Equation~\ref{eq:split_sigmoid}}

                \State{$s \leftarrow s - c^*_2 \quad \text{where } c^*_2 = s - \lfloor s \rceil$} \Comment{ST operator}%

                \State{$p \leftarrow p \, \left( (1-\mathfrak{p}) \, s + \mathfrak{p} \, (1 - s) \right)$} \Comment{Equation~\ref{eq:indicator_func}}
            \EndFor
            \State{$\hat{\boldsymbol{y}} \leftarrow \hat{\boldsymbol{y}} + \mathbb{L}_l \, p$} \Comment{Equation~\ref{eq:dt_prediction}}
        \EndFor
        \State \Return{$\sigma \left( \hat{\boldsymbol{y}} \right)$} \Comment{Softmax  $\sigma$ to get probability distribution}
    \EndFunction
\end{algorithmic}

\end{algorithm}

\subsection{Deterministic Tree Routing and Gradient Calculation}\label{ssec:training}
In the previous sections, we introduced the adjustments that are necessary to apply gradient descent to DTs.
During the optimization, we calculate the gradients with backpropagation based on the computation graph of the tree pass function.
The tree pass function to calculate the function values is summarized in Algorithm~\ref{alg:forward} and utilizes the adjustments introduced in the previous sections. 
Our tree routing facilitates the computation of the tree pass function over a complete batch as a single set of matrix operations, which allows an efficient computation.
We also want to note that our dense representation can be converted into an equivalent vanilla DT representation at each point in time. Similarly, the fully-grown nature of GradTree is only required during the gradient-based optimization and standard pruning techniques to reduce the tree size are applied post-hoc.%

\begin{algorithm}[t]
\caption{Gradient Descent Training for Decision Trees}   \label{alg:train} 

\begin{algorithmic}[1] 
    \Function{trainDT}{$I, T, L, X, \boldsymbol{y}, n, c, d, \xi$}
        \State{$I \sim \mathcal{U}\left(-\sqrt{\frac{6}{2^{2d-1} + n}},\sqrt{\frac{6}{2^{2d-1} + n}}\right)$} 
        \State{$T \sim \mathcal{U}\left(-\sqrt{\frac{6}{2^{2d-1} + n}},\sqrt{\frac{6}{2^{2d-1} + n}}\right)$} 
        \State{$L \sim \mathcal{U}\left(-\sqrt{\frac{6}{2^{2d} + c}},\sqrt{\frac{6}{2^{2d} + c}}\right)$} 
        
        \item[]

        \For{$i = 1,\dots,\xi$}
            \State{$\boldsymbol{\hat{y}} \leftarrow \emptyset$} 
            \For{$j = 1,\dots,|X|$}
                \State{$\hat{y}_j = $ \Call{pass}{$I, T, L, X_j$}}
                \State{$\boldsymbol{\hat{y}} \leftarrow \boldsymbol{\hat{y}} \cup \hat{y}_j$}
            \EndFor
            \State{$I \leftarrow I + \eta \frac{\partial}{\partial I} \mathcal{L}(\boldsymbol{y}, \boldsymbol{\hat{y}})$} \Comment{Calculate gradients with backpropagation} 
            \State{$T \leftarrow T + \eta \frac{\partial}{\partial T} \mathcal{L}(\boldsymbol{y}, \boldsymbol{\hat{y}})$}\Comment{Calculate gradients with backpropagation} 
            \State{$L \leftarrow L + \eta \frac{\partial}{\partial L} \mathcal{L}(\boldsymbol{y}, \boldsymbol{\hat{y}})$}\Comment{Calculate gradients with backpropagation} 

        \EndFor

    \EndFunction
\end{algorithmic}

\end{algorithm}

\subsection{Gradient Descent Optimization} 
\label{ssec:gradient_descent_tree}

We use gradient descent to minimize the loss function of GradTree, which is outlined in Algorithm~\ref{alg:train}. Specifically, we use automatic differentiation to compute the gradients for backpropagation in Lines 11–13.
In this context, we implement mini-batch processing to improve the efficiency of Algorithm~\ref{alg:train} and leverage common SGD techniques, including adaptive learning rates and momentum, by using the Adam~\citep{kingma2014adam} optimizer (see Section~\ref{sec:advancements_gradient_descent}).
Both momentum and adaptive learning rates are crucial for effective training of our method, as gradient updates can be sparse (i.e., only a comparatively small subset of the parameters is active and receives gradients), posing challenges for standard gradient descent. Momentum helps prevent oscillations in gradient updates, leading to smoother convergence. Adaptive learning rates enable rapid adjustments for crucial parameters while maintaining stability for more sensitive ones.
Consequently, plain stochastic gradient descent is insufficient for effectively training our method, as confirmed by preliminary experiments and further demonstrated empirically in our tree ensemble extension (see Section~\ref{sec:eval_grande}).
Moreover, our novel tree pass function allows formulating Line 7-9 as a single set of matrix operations for an entire batch, which results in a very efficient optimization.
We additionally apply weight averaging~\citep{izmailov2018averaging} over 5 consecutive checkpoints, similar to \citet{popov2019neural}.
Furthermore, we implement an early stopping procedure based on the validation loss. To avoid bad initial parametrizations during the initialization, we additionally implement random restarts where the best parameters are selected based on the validation loss.

\begin{figure}[t]
    \centering
    \includegraphics[width=0.8\textwidth]{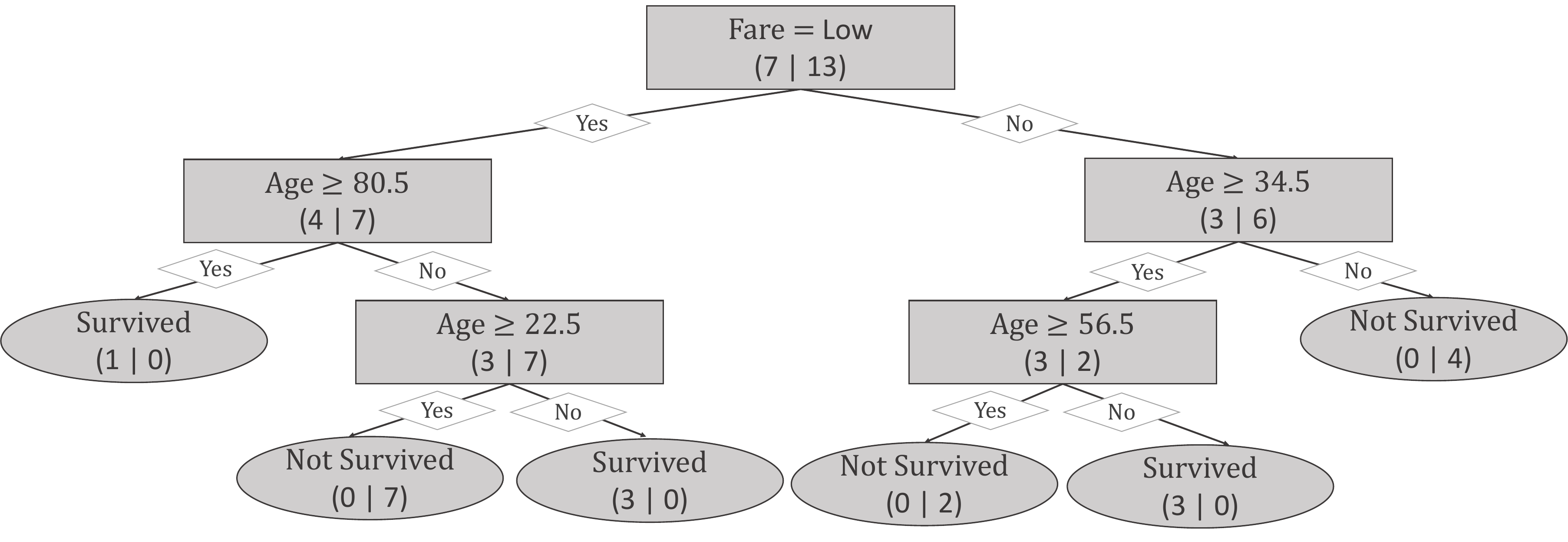}
    \caption[DT learned with GradTree]{\textbf{DT learned with GradTree.} GradTree achieves a perfect accuracy of 100\% compared to 85\% for CART (Figure~\ref{fig:decision_tree_cart}) and includes 5 splits, similar to the optimal DT (Figure~\ref{fig:decision_tree_optimal}). The root node splits on the fare, and subsequent splits are made on the continuous age attribute. Essentially, the tree encodes three rules to decide whether a passenger survived or not. If a passenger paid a low fare and is older than 80.5 years, they are predicted to survive. Similarly, a passenger is predicted to survive if they paid a low fare, are younger than 80.5 yearns and younger than 18.5 years. If a passenger paid a high fare, is older than 34.5 years and younger than 56.5 years, they are also predicted to survive. In all other cases, the passenger is predicted not to survive.}
    \label{fig:decision_tree_gradtree}
\end{figure}

\subsubsection*{Example 5: A Decision Tree Learned with GradTree}
In the following example, we will show the advantages of a joint, gradient-based optimization of all tree parameters with GradTree over alternative methods like greedy approaches, which can result in suboptimal DTs by sequentially selecting the locally optimal split.
For this example, we use the simplified version of the Titanic dataset from Table~\ref{tab:titanic_2d_num_cat} including the age as continuous and the fare as categorical attribute.
The DT learned by GradTree (Figure~\ref{fig:decision_tree_gradtree}) closely resembles the tree generated by an optimal DT algorithm (Figure~\ref{fig:decision_tree_optimal}), correctly classifying all training instances. The primary distinction lies in the swapped split nodes for the second and third level in the left branch of the root node. Additionally, minor differences in split thresholds are observed, which do not affect the classification of the training data. These discrepancies can be attributed to the gradient-based optimization of split thresholds in GradTree, as opposed to the fixed thresholds used in optimal and greedy methods. 
This example clearly demonstrates that GradTree effectively overcomes the issue of local optimality inherent in greedy, purity-based DT induction methods. Specifically, while a DT constructed using a traditional greedy, purity-based approach (Figure~\ref{fig:decision_tree_cart}) achieved an accuracy of only 85\%, the DT learned with GradTree reached a perfect accuracy of 100\%. This performance gain can be attributed to the joint optimization of all parameters, enabling the selection of \emph{Fare Category} as the root split - an attribute that represents the globally optimal choice. In contrast, greedy methods, which rely on locally optimal decisions at each step, fail to identify this globally optimal split.
A more comprehensive discussion of these advantages is provided in Section~\ref{sec:discussion_related}, along with an extensive empirical evaluation in Section~\ref{sec:gradtree}, where we compare GradTree to both greedy and alternative non-greedy methods.

\section{Theoretical Discussion and Comparison} \label{sec:discussion_related}
In the following, we provide a detailed discussion and comparison between the methods introduced in Chapter~\ref{cha:related} and the previously introduced GradTree algorithm. GradTree presents fundamental differences and advantages compared to existing gradient-based methods, as we will show in the next subsection (Section~\ref{sec:comparison_gradient_based}). After this direct comparison of gradient-based methods, we provide a broader evaluation of the various high-level approaches for learning DTs introduced in this chapter (Section~\ref{sec:comparison_general}).

\subsection{Comparison of Gradient-Based Methods} \label{sec:comparison_gradient_based}
The following comparison aims to highlight how GradTree differs from existing approaches that allow learning DTs with gradient descent, and is summarized in Table~\ref{tab:sdt_method_comparison}. Soft DTs generally show two key characteristics: Oblique decision boundaries and soft, probabilistic splitting. These characteristics are the primary factors that enable gradient-based optimization. However, they also influence interpretability and remove the inductive bias of hard, axis-aligned decisions~\citep{grinsztajn2022tree}. Several methods have been developed to address the effects of one or both of these characteristics, though most focus on only one (by using either axis-aligned or hard splits). In contrast, \citet{bruch2020learning} achieves both axis-aligned and hard splits. However, existing methods that aim to achieve hard, axis-aligned splits often introduce certain limitations. For example, as discussed in earlier sections, \citet{bruch2020learning} and \citet{dndt} do not learn the features to split on. Furthermore, \citet{bruch2020learning} have a custom gradient calculation which are hard to integrate into existing gradient-based frameworks where model are trained through backpropagation with automatic differentiation\footnote{In this context, we distinguish between gradient-based methods and approaches that learn DTs using backpropagation via automatic differentiation, as it is the case for GradTree. While gradient-based training offers advantages such as global optimization and increased flexibility, it does not necessarily enable seamless integration into existing frameworks, which typically rely on backpropagation through automatic differentiation.}.
GradTree is the only method that learns a hard, axis-aligned DT without imposing any restrictions on the training procedure, while seamlessly be integrating into existing gradient-based frameworks, as we will show in Section~\ref{sec:multimodal} and Chapter~\ref{cha:sympol}.

\newcommand{\SDTs}{SDTs~\citep{soft1}}

\begin{table}[tb]
    \centering
    \small
    \caption[Gradient-Based Decision Tree Comparison]{\textbf{Gradient-Based Decision Tree Comparison.} Comparison of gradient-based DT learning methods based on characteristic aspects.}    
    \resizebox{0.99\textwidth}{!}{
    \begin{tabular}{|C{2.0cm}|C{1.2cm}|C{1.6cm}|C{2.0cm}|C{1.3cm}|C{1.3cm}|C{2.0cm}|C{1.3cm}|C{1.5cm}|} \hline
      & \textbf{SDTs \citep{soft1}} & \textbf{NODE \citep{popov2019neural}} & \textbf{Axis-Aligned Decision Forests \citep{bruch2020learning}} & \textbf{\citet{norouzi2015efficient}} & \textbf{\citet{karthikeyanlearning}} & \textbf{Argmin Differentiation \citep{zantedeschi2021learning}} & \textbf{DNDT \citep{dndt}} & \textbf{GradTree (ours)} \\ \hline
    \textbf{Decision Boundary} & oblique & oblique & axis-aligned & oblique & oblique & oblique & axis-aligned & axis-aligned \\ \hline
    \textbf{Split Type} & soft & soft & hard & hard & hard & hard & soft & hard \\ \hline
    \textbf{Learnable Split Choice} & - & - & no & - & - & - & no & yes \\ \hline
    \textbf{Back-propagation} & yes & yes & no & yes & yes & no & yes & yes \\ \hline
    \end{tabular}
    }
    \label{tab:sdt_method_comparison}
\end{table}

\subsection{Comparison of Different Decision Tree Learning Methods} \label{sec:comparison_general}

\begin{table}[tb]
    \centering
    \small
    \caption[Comparison of Different Types of Decision Tree Learning]{\textbf{Comparison of Different Types of Decision Tree Learning.} In this table, we compare the different types of methods for DT learning based on their most characteristic aspects. A checkmark (\cmark) indicates that a characteristic is present, while a cross (\xmark) denotes its absence or infeasibility and $\boldsymbol{\sim}$ indicated that an integration is possible but by default not implemented. Additionally, we assess methods using an ordered scale: $[\mathbf{--}, \mathbf{-}, \mathbf{\circ}, \mathbf{+}, \mathbf{++}]$. Although this evaluation is grounded in the theoretical properties of each method as introduced in the previous sections, it is important to note that the ratings inherently involve a degree of subjectivity.}
    \resizebox{0.99\textwidth}{!}{
    \begin{tabular}{|C{2.50cm}|C{1.9cm}|C{2.2cm}|C{1.9cm}|C{2.0cm}|C{2.2cm}|C{1.4cm}|} \hline 
        \textbf{} & \textbf{Greedy (Section~\ref{sec:greedy_dt})} & \textbf{Lookahead (Section~\ref{sec:lookahead_dt})} & \textbf{Optimal (Section~\ref{sec:optimal_dt})} & \textbf{Evolutionary (Section~\ref{sec:evolutionary_dt})} & \textbf{Vanilla SDT (Section~\ref{sec:sdt_related})} & \textbf{GradTree (ours)} \\ \hline 
        
        \textbf{Optimization} & Local & Local & Global & Global & Global & Global \\ \hline
        \textbf{Optimality (based on training data)} & \xmark & \xmark & \cmark & \xmark & \xmark & \xmark \\ \hline
        
        \textbf{Decision Boundary} & axis-aligned & axis-aligned & axis-aligned (oblique) & axis-aligned (oblique) & oblique & axis-aligned \\ \hline
        \textbf{Split Type} & hard & hard & hard & hard & soft & hard \\ \hline
        \textbf{Overfitting Potential} & $\mathbf{\circ}$ & $\mathbf{+}$ & $\mathbf{++}$ & $\mathbf{--}$ & $\mathbf{-}$ & $\mathbf{--}$ \\ \hline
        \textbf{Continuous Features} & \cmark & \cmark & \xmark & \cmark & \cmark & \cmark \\ \hline

        \textbf{Runtime Scalability with Dataset Size} & $\mathbf{\circ}$ & $\mathbf{-}$ & $\mathbf{--}$ & $\mathbf{\circ}$ & $\mathbf{+}$ & $\mathbf{+}$ \\ \hline
        \textbf{Runtime Scalability with Features} & $\mathbf{\circ}$ & $\mathbf{-}$ & $\mathbf{--}$ & $\mathbf{++}$ & $\mathbf{+}$ & $\mathbf{+}$ \\ \hline
        \textbf{Storage Scalability with Features} & $\mathbf{\circ}$ & $\mathbf{-}$ & $\mathbf{\circ}$ & $\mathbf{+}$ & $\mathbf{-}$ & $\mathbf{-}$ \\ \hline        
        \textbf{Tree Size Constraints} & $\infty$ & $k \leq 4$ & $d \leq 4$ & $\infty$ & $d \leq 10$ & $d \leq 10$ \\ \hline
        \textbf{Runtime} & $\mathbf{++}$ & $\mathbf{-}$ & $\mathbf{--}$ & $\mathbf{\circ}$ & $\mathbf{\circ}$ & $\mathbf{\circ}$ \\ \hline

        \textbf{Customizability and Flexibility} & $\mathbf{-}$ & $\mathbf{-}$ & $\mathbf{\circ}$ & $\mathbf{\circ}$ & $\mathbf{++}$ & $\mathbf{++}$ \\ \hline
        \textbf{Custom Objective} & \xmark & \xmark & $\boldsymbol{\sim}$ & \cmark & \cmark & \cmark \\ \hline 
        \textbf{Multi-Way Splitting} & \cmark & \cmark & $\boldsymbol{\sim}$ & $\boldsymbol{\sim}$ & $\boldsymbol{\sim}$ & $\boldsymbol{\sim}$ \\ \hline
        \textbf{Multi-Class} & \cmark & \cmark & $\boldsymbol{\sim}$ & \cmark & \cmark & \cmark \\ \hline
    \end{tabular}
    }
    \label{tab:dt_method_comparison}
\end{table}

In the following, we compare key aspects of DT learning methods and how they are approached by different categories of algorithms. The goal is to highlight strengths and limitations in the context of specific high-level methods to learn DTs. We provide an overview of all key characteristics of the discussed algorithms in Table~\ref{tab:dt_method_comparison} and discuss the most relevant aspects detailed in the following. 

\paragraph{Local vs. global optimization}
Optimization in DT learning can be broadly categorized into local and global approaches, each with distinct trade-offs. Local optimization methods, such as greedy algorithms~\citep{cart_breiman1984,quinlan1993c45}, make decisions incrementally by greedily selecting the best split at each node based on a local criterion, like maximizing purity. While computationally efficient, these methods are prone to suboptimal solutions as they fail to consider the overall tree structure. In contrast, global optimization approaches including optimal DTs~\citep{bertsimas2017optimal}, evolutionary algorithms~\citep{barros2011survey} and gradient-based methods~\citep{dndt} aim to find an entire tree configuration that minimizes a global objective. For global optimization, it can additionally be distinguished between optimal and non-optimal methods. While optimal DTs aim to find the optimal tree based on the training data, evolutionary and gradient-based methods do not necessarily pursue this. Instead, like optimization in neural networks, they aim for a local optimum that offers good generalization to test data~\citep{soft2}.
Further, the local optimum that can be reached by evolutionary and gradient-based methods has a significant advantage over the local optimum of greedy approaches, which is constrained by sequentially selecting the optimal split, as all model parameters are optimized jointly.

\paragraph{Decision boundaries and decisions}
Traditional greedy algorithms like CART and C4.5 create axis-aligned decision boundaries by splitting data based on single features at each node. These univariate splits make DTs easily interpretable, which is a key reason for their popularity~\citep{cart_breiman1984,quinlan1993c45}.
While optimal DTs and genetic algorithms can use more complex, oblique decision boundaries, the most common versions still employ axis-aligned splits~\citep{aglin2020learning}.
For gradient-based DT methods, oblique splits are often essential as they facilitate gradient-based optimization~\citep{soft2}. Although this can enhance performance, particularly in shallow trees, it comes at the cost of interpretability. As \citet{molnar2020} points out, humans struggle to understand explanations involving more than three dimensions, which is frequently the case with oblique splits.
Additionally, gradient-based DTs often employ soft decisions, allowing traversal of all paths in the tree simultaneously, weighted by probabilities, rather than following a single path. 

\paragraph{Overfitting potential}
Greedy algorithms like ID3, C4.5, and CART are prone to overfitting, especially without proper pruning strategies~\citep{cart_breiman1984,quinlan1993c45}. They aim to maximize purity at each split, which can lead to trees that fit the training data too closely and fail to generalize to unseen data. Methods like CART and C4.5 incorporate post-pruning techniques to simplify the tree after construction, mitigating overfitting. Evolutionary algorithms and optimal DTs can incorporate multiple objectives, balancing performance and complexity to reduce overfitting~\citep{barros2011survey,murtree}. 
Nevertheless, overfitting remains a common issue in optimal DTs, as their optimal performance on the training data can limit their capacity to generalize effectively to unseen data~\citep{zharmagambetov2021non}.
In gradient-based approaches, overfitting is generally less problematic, as mini-batch optimization inherently serves as a form of regularization, effectively reducing the risk of overfitting~\citep{goodfellow2016deep,hoffer2017train}.

\paragraph{Scalability with dataset size and features}
Scalability is a significant challenge, particularly for methods that involve global search processes like optimal DTs, evolutionary algorithms and gradient-based methods. Greedy methods like CART scale better with larger datasets due to their lower computational cost. However, even they can struggle with very large datasets or high-dimensional feature spaces, as both impact the computational complexity~\citep{cart_breiman1984}. 
Considerable effort has been dedicated to reducing the runtime of optimal DTs~\citep{murtree}. Indeed, the runtime of current methods is significantly lower than that of those considered state-of-the-art 5 to 10 years ago. Nevertheless, scalability remains a major limitation for optimal DTs, with learning typically being feasible only for trees with a maximum depth of 4 on most datasets~\citep{aglin2020learning,murtree}. In contrast, genetic algorithms scale relatively well, particularly with the number of features, as they do not influence the computational complexity~\citep{barros2011survey}. However, for large datasets, genetic algorithms can require substantial runtime, especially when high numbers of generations and large population sizes are necessary to achieve strong performance.
Frameworks like TensorFlow have significantly enhanced the efficiency of gradient-based methods by simplifying both their design and training, while also providing tools for scalable computation and efficient tensor operations. As a result, gradient-based methods tend to scale well with respect to dataset size and number of features in terms of runtime, although they are not as competitive as greedy methods. However, since the number of features directly influences the number of model parameters, gradient-based methods do not scale as efficiently in terms of space complexity. Furthermore, the optimization process becomes increasingly challenging as the number of features grows, making it more difficult to train models on very high-dimensional datasets.

\paragraph{Tree size}
Greedy algorithms can result in very deep trees, especially if not properly pruned, making them less interpretable and more prone to overfitting~\citep{quinlan1993c45}. Advanced pruning methods in algorithms like CART help control tree size while preserving accuracy, enhancing interpretability without significantly compromising performance. Evolutionary algorithms and optimal DTs tend to produce smaller, more compact trees by including tree size or depth in their optimization objectives~\citep{aglin2020learning,barros2011survey}. However, it must be noted that optimal DTs are generally constrained to a maximum depth of 4. Gradient-based methods also have constraints on the tree depth, as the number of parameters scales exponential with the depth. However, they do not have as strict constraints as, e.g., optimal DTs and can efficiently be learned efficiently until a depth of $10$.

\paragraph{Customizability and flexibility}
The ability to customize and adapt DT algorithms to specific problem requirements is a significant advantage of more advanced methods. Traditional algorithms like CART and C4.5 offer some flexibility through parameter adjustments and pruning strategies but are generally limited by their predefined structures and single-objective optimization criteria, such as maximizing purity. Conversely, evolutionary algorithms and optimal DT methods offer a high degree of customizability and naturally support multi-objective optimization~\citep{aglin2020learning,barros2011survey}. These approaches enable the incorporation of domain-specific constraints and custom optimization objectives directly into the model-building process. However, this added flexibility comes with trade-offs, including increased computational complexity and the need for careful parameter tuning to ensure a balance between competing objectives. Similarly, gradient-based methods offer a high level of flexibility, with custom objectives easily incorporated through modifications to the loss function~\citep{soft3}. Moreover, these methods usually integrate seamlessly into frameworks that rely on gradient descent, such as reinforcement learning or multimodal learning~\citep{silva2020optimization}. Methods that do not rely on gradient descent and an optimization through backpropagation typically need custom solutions which are challenging to develop, maintain and integrate efficiently~\citep{roth2019conservative,gupta2015policy}. Additionally, gradient-based methods benefit significantly from ongoing advances in deep learning research, allowing for the adoption of novel techniques that can further enhance their performance.

\section{Extending Gradient-Based Decision Trees to Ensembles} \label{sec:method_grande}

In the following, we build upon GradTree, transitioning from individual trees to a weighted tree ensemble (Section~\ref{ssec:ensembles}), while maintaining an efficient computation. As a result, we propose GRANDE\footnote{Our implementation is publically available under: \url{https://github.com/s-marton/GRANDE}.}, \textbf{GRA}die\textbf{N}t-Based \textbf{D}ecision Tree \textbf{E}nsembles, a novel approach for learning DT ensembles using end-to-end gradient descent. In this context, we propose an instance-wise weighting mechanism (Section~\ref{ssec:weighting}), and explore the integration of regularization techniques from tree ensembles with gradient-based methods (Section~\ref{ssec:regularization}).
As a result, GRANDE can be learned end-to-end with gradient descent, leveraging the potential and flexibility of a gradient-based optimization.

\subsection{From Decision Trees to Weighted Tree Ensembles}\label{ssec:ensembles} 
One advantage of GRANDE over existing gradient-based methods is the inductive bias of axis-aligned splits for tabular data.
Combining this property with ensembling and an end-to-end gradient-based optimization is at the core of GRANDE. 
This is also a major difference to existing deep learning methods for hierarchical representations like NODE~\citep{popov2019neural}, where soft, oblique splits are used (Section~\ref{ssec:oblique_ensembles}).
Therefore, we can define GRANDE as
\begin{equation}\label{eq:grande}
    G(\boldsymbol{x} | \boldsymbol{\omega},\bm{L},\tens{T},\tens{I}) = \sum_{e=0}^{E-1} \omega_{e} \, g(\boldsymbol{x}| \bm{L}_e,\tens{T}_e,\tens{I}_e),
\end{equation}
where $E$ is the number of estimators in the ensemble and $\boldsymbol{\omega}$ is a weight vector. By extending $\bm{L}$ to a matrix and $\tens{T}$, $\tens{I}$ to tensors for the complete ensemble instead of defining them individually for each tree, we can leverage parallel computation for an efficient training.
As GRANDE can be learned end-to-end with gradient descent, we keep an important advantage over existing, non-gradient-based tree methods like XGBoost and CatBoost. Both, the sequential induction of the individual trees and  the sequential combination of individual trees via boosting are greedy. This results in constraints on the search space and can favor overfitting, as highlighted in Chapter~\ref{cha:related}. 
In contrast, GRANDE learns all parameters of the ensemble jointly and overcomes these limitations.

\begin{figure}[t]
    \centering
    \includegraphics[width=\columnwidth]{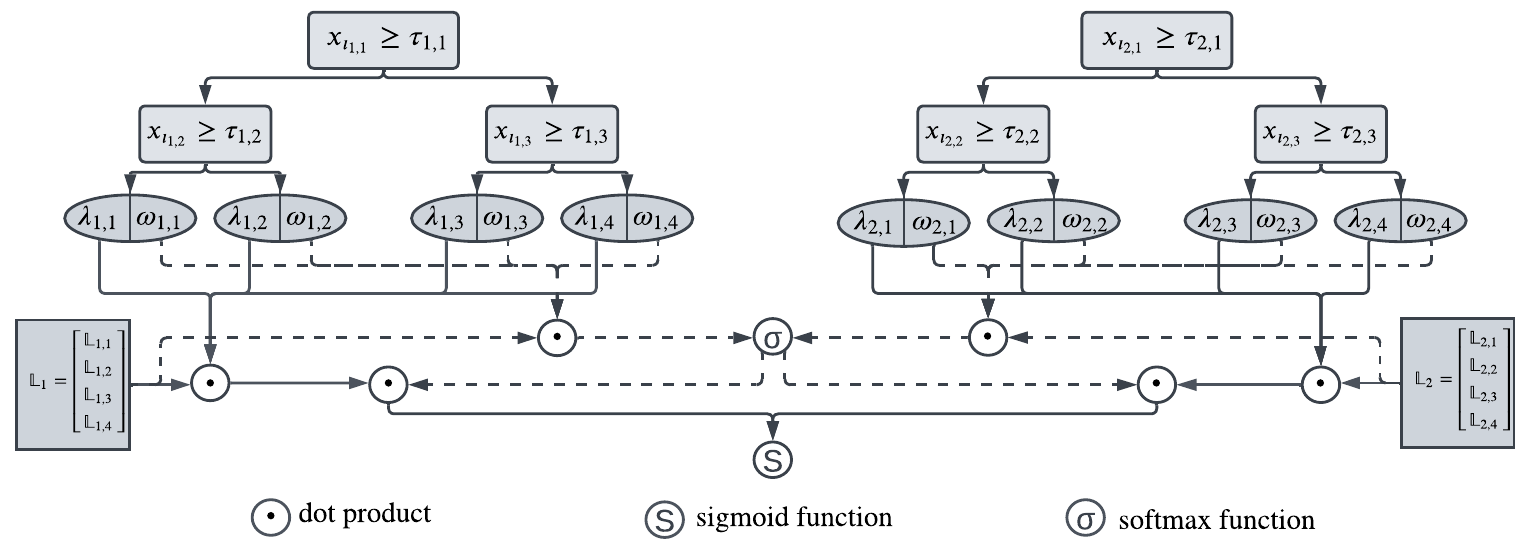}
    \caption[GRANDE Architecture]{\textbf{GRANDE Architecture.} This figure visualizes the structure and weighting of GRANDE for an exemplary ensemble with two trees of depth two. 
    For each tree in the ensemble, and for every sample, we determine the weight of the leaf which the sample is assigned to. 
    Subsequently, a softmax is applied on these chosen weights. Multiplying the post-softmax weights by the predictions equates a weighted average of the individual estimators. 
    }
    \label{fig:weighting}
\end{figure}
\subsection{Instance-Wise Estimator Weights} \label{ssec:weighting}
One challenge of ensemble methods is learning a good weighting scheme of the individual estimators. The flexibility of an end-to-end gradient-based optimization allows including learnable weight parameters to the optimization. A simple solution would be learning one weight for each estimator and using for instance a softmax over all weights, resulting in a weighted average.
However, this forces a very homogeneous ensemble, in which each tree aims to make equally good predictions for all samples. In contrast, it would be beneficial if individual trees can account for different areas of the target function, and are not required to make confident predictions for each sample. 
To address this, we propose an advanced weighting scheme that allows calculating instance-wise weights that can be learned within the gradient-based optimization. Instead of using one weight per \emph{estimator}, we use one weight for each \emph{leaf} of the estimator as visualized in Figure~\ref{fig:weighting} and thus define the weights as $\bm{W} \in \mathbb{R}^{E} \times \mathbb{R}^{2^d}$ instead of $\boldsymbol{\omega} \in \mathbb{R}^{E}$. 
We define $p(\boldsymbol{x} | \bm{L}, \tens{T}, \tens{I}):\mathbb{R}^n \rightarrow \mathbb{R}^E$ as a function to calculate a vector comprising the individual prediction of each tree. Further, we define a function $w(\boldsymbol{x} | \bm{W}, \bm{L}, \tens{T}, \tens{I}):\mathbb{R}^n \rightarrow \mathbb{R}^E$ to calculate a weight vector with one weight for each tree based on the leaf which the current sample is assigned to. %
Subsequently, a softmax is applied to these chosen weights for each sample. The process of multiplying the post-softmax weights by the predicted values from each tree equates to computing a weighted average.
This results in
\begin{equation}
    G(\boldsymbol{x} | \bm{W},\bm{L},\tens{T},\tens{I}) = \sigma\left(w(\boldsymbol{x} |\bm{W}, \bm{L}, \tens{T}, \tens{I})\right) \cdot p(\boldsymbol{x} | \bm{L}, \tens{T}, \tens{I}),
\end{equation}

\begin{align*}
\text{where} \quad w(\boldsymbol{x} | \bm{W}, \bm{L}, \tens{T}, \tens{I}) &= 
\begin{bmatrix}
    \sum_{l=0}^{2^{d-1}-1} \bm{W}_{0,l} \, \mathbb{L}(\boldsymbol{x} | \bm{L}_{0,l}, \tens{T}_0, \tens{I}_0) \\
    \sum_{l=0}^{2^{d-1}-1} \bm{W}_{1,l} \, \mathbb{L}(\boldsymbol{x} | \bm{L}_{1,l}, \tens{T}_1, \tens{I}_1) \\
    \vdots \\
    \sum_{l=0}^{2^{d-1}-1} \bm{W}_{E-1,l} \, \mathbb{L}(\boldsymbol{x} | \bm{L}_{E-1,l}, \tens{T}_{E-1}, \tens{I}_{E-1}) \\
\end{bmatrix} \, , \\ 
p(\boldsymbol{x} | \bm{L}, \tens{T}, \tens{I}) &= 
\begin{bmatrix}
    g(\boldsymbol{x}| \bm{L}_0,\tens{T}_0,\tens{I}_0) \\
    g(\boldsymbol{x}| \bm{L}_1,\tens{T}_1,\tens{I}_1) \\
    \vdots \\
    g(\boldsymbol{x}| \bm{L}_{E-1},\tens{T}_{E-1},\tens{I}_{E-1}) \\
\end{bmatrix},
\end{align*}

and $\sigma(\boldsymbol{z})$ is the softmax function.
It is important to note that when calculating $\mathbb{L}$ (see Equation~\ref{eq:indicator_func}), only the value for the leaf to which the sample is assigned in a given tree is non-zero.
We want to note that our weighting scheme permits calculating instance-wise weights even for unseen samples. Furthermore, our weighting allows GRANDE to learn representations for simple and complex rules within one model. 
In our evaluation, we demonstrate that instance-wise weights significantly enhance the performance of GRANDE and emphasize local interpretability.

\subsection{Regularization: Feature Subset, Data Subset and Dropout}\label{ssec:regularization}
The combination of tree-based methods with gradient-based optimization opens the door to applying a variety of regularization techniques that can improve both model performance and generalization. These techniques aim to prevent overfitting, reduce variance, and improve scalability. In this section, we discuss three key regularization strategies: Feature subset selection, data subset selection, and dropout.

\paragraph{Feature subset selection} For each tree in the ensemble, we select a subset of the available features. This technique has two main advantages: Firstly, it reduces the complexity of each individual tree by limiting the number of features it can split on, thereby regularizing the model. Secondly, it helps to address the poor scalability of GradTree  when dealing with a large number of features. By selecting only a fraction of the features for each tree, we decrease computational requirements and mitigate overfitting, as models trained on different subsets will capture different aspects of the data. This induces diversity within the ensemble, creating a more heterogeneous collection of trees, which can improve overall model performance by reducing correlation among the trees and improving generalization to unseen data. This technique is commonly used in methods such as Random Forests and Gradient Boosted DTs, where it has been shown to enhance both performance and robustness.

\paragraph{Data subset selection} In addition to feature subset selection, we apply data subset sampling. For each estimator in the ensemble, we sample a fraction of the training data, with or without replacement. This process further promotes model diversity by training individual estimators on slightly different distributions of the data. The key advantage of data subset selection is its ability to improve generalization by reducing variance across the ensemble. Moreover, it enhances computational efficiency by allowing the training of smaller, faster models on subsets of the data. The combination of feature and data subsampling creates models that are diverse not only in the features they consider but also in the patterns they learn from different data samples, which leads to a stronger ensemble.

\paragraph{Dropout} To further regularize the ensemble, we introduce a dropout technique, inspired by the method commonly used in neural networks. In this context, dropout involves randomly deactivating a predefined fraction of the estimators in the ensemble during training. By selectively deactivating estimators, we prevent any single tree from dominating the prediction, which can reduce the risk of overfitting to the training data. Additionally, after dropping out a set of estimators, we rescale the weights of the remaining active estimators to ensure that the predictions remain unbiased. This technique has the added benefit of reducing computational costs during training, as only a subset of the trees is active at any given time. Dropout can be particularly effective when combined with data and feature subsampling, as it creates a more diverse ensemble and helps the model generalize better to unseen data. \\

Overall, these regularization techniques (feature subset selection, data subset selection, and dropout) work together to create a more robust and scalable model, improving its ability to generalize while maintaining computational efficiency. The introduction of these techniques into gradient-based DTs provides a powerful mechanism to control model complexity and mitigate overfitting, making it an essential component of the proposed method.

\chapter{Gradient-Based Decision Trees for Tabular Data} \label{cha:tabular}

\textbf{The following section was already partially published in \citet{gradtree} and \citet{grande}. The implementations and experiments are publically available under \url{https://github.com/s-marton/GradTree} and \url{https://github.com/s-marton/GRANDE}, respectively. For a detailed summary of the individual contributions, please refer to Section~\ref{sec:contributions}.} \\

Heterogeneous tabular data is the most frequently used form of data~\citep{tabularData,shwartztabular} and is indispensable in a wide range of applications such as medical diagnosis~\citep{med1,med2,med3}, %
estimation of creditworthiness~\citep{credit1} and fraud detection~\citep{fraud1}.
In the following, we will first give an introduction into the modality of tabular data (Section~\ref{sec:background_tabular}), its characteristics and challenges. Next, we will give a short introduction into state-of-the-art models for tabular data, ranging from tree-based over hybrid to deep learning methods. In the second section (Section~\ref{sec:gradtree}), we will focus on simpler, small- to medium-sized real-world datasets where individual DTs are a realistic modeling choice and show that DTs learned with GradTree are competitive and often even outperform existing methods for learning DTs in terms of performance while still maintaining interpretability.
In the subsequent section (Section~\ref{sec:grande}), we move to more complex real-world datasets where individual DTs typically cannot achieve a high performance and show that extending our gradient-based DTs to GRANDE as a tree ensembles results in a method that is able to achieve state-of-the-art performance while maintaining local interpretability through its unique instance-wise weighting. 
Additionally, we show the potential of a gradient-based optimization by including our method into a multimodal learning setup (Section~\ref{sec:multimodal}), resulting in performance gains against commonly used DL alternatives out-of-the-box.

\begin{figure}
    \centering
    \includegraphics[width=0.75\linewidth]{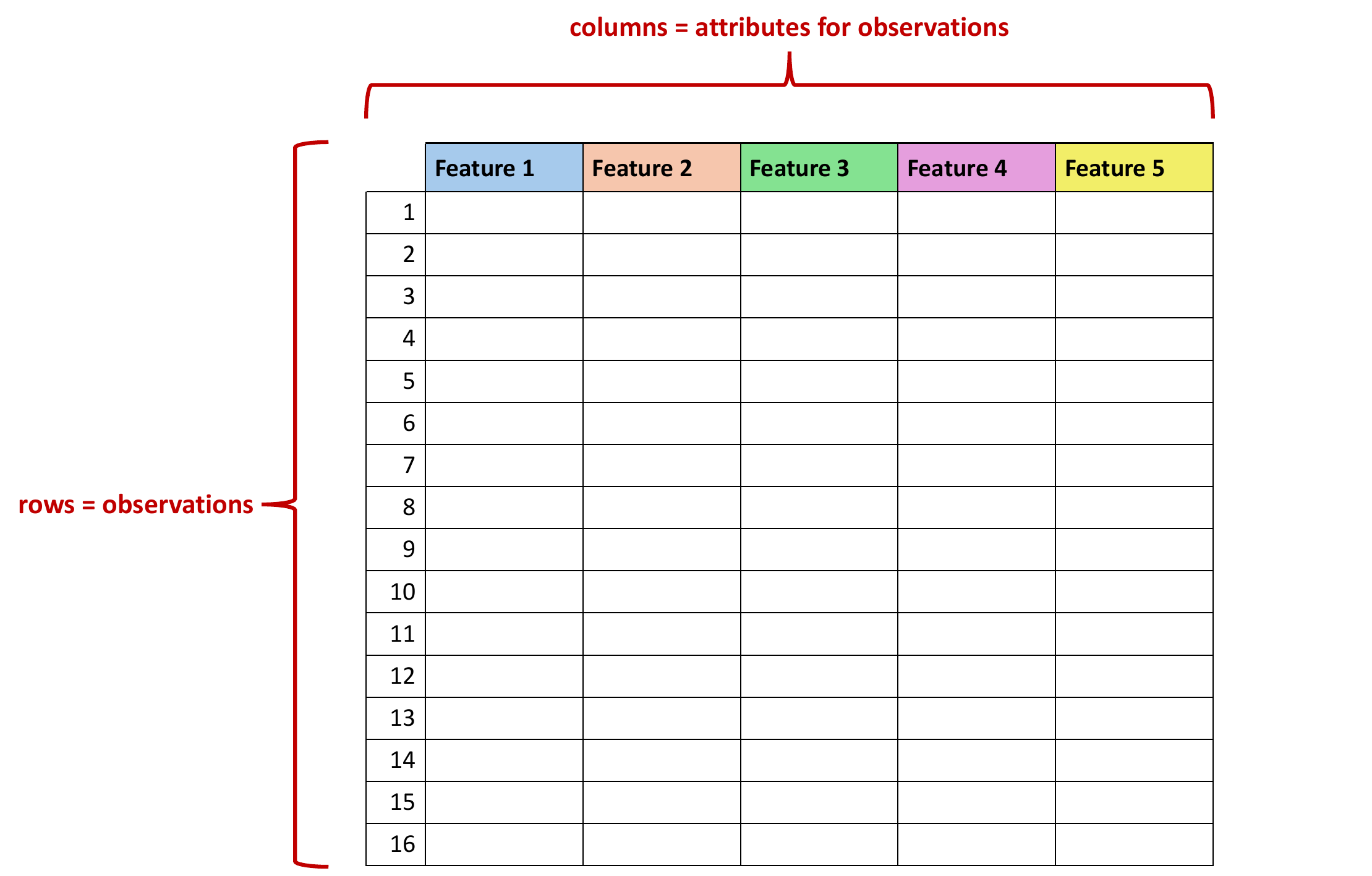}
    \caption[Tabular Data]{\textbf{Tabular Data.} This figure visualizes the straightforward structure of tabular data, with rows as observations and columns as features.}
    \label{fig:tabular_data}
\end{figure}

\section{Background: Tabular Data} \label{sec:background_tabular}
The layout of tabular data is generally simple and intuitive: Rows represent observations or samples, while columns denote features or attributes that describe those observations, as visualized in Figure~\ref{fig:tabular_data}.
Tabular data is often referred to as \emph{structured data} referring to its structured representation in rows and columns compared to unstructured data like images or text. The major difference to these other modalities is the heterogeneity of tabular data. 
For image or text, each signal in the input has the same internal structure, like a pixel in an image encoded as grayscale. In such a case, each value in an input corresponds to the same signal type, i.e., the intensity of light at that point. For tabular data, this is very different in most cases. 
Features in tabular data cannot only have different ranges, but also be completely different feature types. In tabular data, we can have continuous values along with categorical variables for the same data point. 
As a result, despite its straightforward structure, working with tabular data in machine learning can be very challenging, especially for neural networks which usually excel with unstructured and homogeneous data.

\subsection{Structure of Tabular Data}
Tabular data is characterized by its highly organized and rectangular format, which can be directly mapped to tables, databases, or spreadsheets as visualized in Table~\ref{fig:tabular_data}. In this structure:
\begin{itemize}
    \item \textbf{Rows correspond to observations or instances}: Each row represents a single entity or example, such as a customer, a product, a sensor reading at a point in time, or a patient record.
    \item \textbf{Columns correspond to features or variables}: Each column is a different variable that describes some property of the instances, ranging from numerical values (e.g., income, temperature) to categorical labels (e.g., product type, region).
\end{itemize}
The values within each cell in the table could be numerical, categorical, text-based, or even date/time values. The schema of the table is well-defined, meaning the number of columns (features) is fixed, and each feature has a designated type (e.g., integer, float, string).

\paragraph{Types of features in tabular data}
One unique characteristic is the presence of different feature types in tabular datasets. Therefore, tabular data is considered as heterogeneous compared to e.g., text or image datasets which are considered as homogeneous. We will shortly explain the different data type in the following:

\begin{itemize}
    \item \textbf{Numerical features}: These include both continuous variables (e.g., height, temperature) and discrete variables (e.g., count of items). They are represented as real numbers or integers and can be directly used in most machine learning models. As they can strongly vary in their scale, depending on the model, preprocessing such as scaling or normalization can be necessary.
    
    \item \textbf{Categorical features}: These variables represent discrete categories or classes, such as gender, product type, or country. They can be nominal (no inherent order, like color: Red, green, blue) or ordinal (having an order, like ratings: Low, medium, high). Machine learning models typically require converting these categories into numerical representations through techniques like one-hot encoding, label encoding, or embedding representations.

    \item \textbf{Boolean/binary features}: These are special cases of categorical variables, where there are only two possible values (e.g., yes/no, 0/1, true/false). Although straightforward to encode, they can hold great predictive power in models.

\end{itemize}

\subsection{Challenges in Machine Learning for Tabular Data}
Despite its clear structure, tabular data presents several complexities in machine learning. The key characteristics and challenges are discussed in the following.

\paragraph{Heterogeneity of data types and encoding}
As shown previously, tabular datasets can contain multiple types of features, each requiring different preprocessing steps~\citep{borisov2022deep,grinsztajn2022tree,ren2025deep}. Numerical features may need to be normalized or standardized, while categorical features must be encoded into a numerical format. The way data is encoded can influence model performance, and inappropriate encoding can introduce bias or noise into the model~\citep{tschalzev2024data}. For example, encoding ordinal variables using one-hot encoding may remove the inherent order in the data, while using label encoding for nominal variables may introduce artificial ordering. %

\paragraph{Feature scaling and normalization}
Machine learning models such as neural networks, support vector machines, and k-nearest neighbors are sensitive to the scale of numerical features\footnote{A discussion on the importance of feature scaling can be found under: \url{https://scikit-learn.org/1.5/auto_examples/preprocessing/plot_scaling_importance.html} (Last accessed: March 3, 2025).}. For instance, if one feature has values ranging from 0 to 1,000, and another ranges from 0 to 1, the model could give undue importance to the former simply due to its scale. Therefore, normalization (scaling features to a range, like [0, 1] or [-1, 1]) or standardization (adjusting features to have zero mean and unit variance) is essential to ensure balanced model performance~\citep{borisov2022deep,ren2025deep}.

\paragraph{High dimensionality and feature selection}
Many real-world tabular datasets have a large number of features relative to the number of samples, resulting in a high-dimensional space. This phenomenon, known as the \emph{curse of dimensionality}, can lead to overfitting, where the model learns noise rather than meaningful patterns, and computational inefficiency~\citep{yu2003feature}. 
Furthermore, some algorithms, like k-nearest neighbors kernel methods and evolutionary algorithms, generally struggle with high-dimensional data, as the search space grows exponentially, making optimization increasingly difficult.
Techniques such as dimensionality reduction (e.g., PCA~\citep{wold1987pca}, t-SNE~\citep{van2008sne}) or feature selection (e.g., recursive feature elimination, correlation-based filtering, Lasso regression) are often employed to reduce the feature space and improve model generalization.

\paragraph{High cardinality categorical features}
High cardinality categorical features, i.e., features with a large number of unique categories, pose significant challenges in machine learning for tabular data~\citep{sigrist2023comparison}. Standard encoding techniques like one-hot encoding can lead to an explosion in feature dimensionality, making the model computationally expensive and prone to overfitting, especially when the number of samples is small relative to the number of categories~\citep{liang2025efficientrepresentationshighcardinalitycategorical}. Label encoding, while more compact, may unintentionally imply an ordinal relationship between categories, introducing bias. 
Target encoding mitigates these issues by replacing categories with target-dependent statistics, but it can be problematic for high cardinality features. Rare categories may lead to unreliable estimates, increasing variance and overfitting risks. Additionally, improper handling of target encoding can cause data leakage, inflating performance metrics while negatively affecting generalization. Advanced techniques such as embeddings or cross-validation-based encoding help mitigate these challenges by balancing dimensionality reduction with robust information preservation.

\paragraph{Dealing with missing data}
Missing values are common in tabular datasets and can arise due to a variety of reasons (e.g., sensor failure, human error, incomplete data entry). Handling missing data effectively is crucial for model performance~\citep{zhou2024missing,ren2025deep}. Imputation methods (e.g., filling missing values with mean, median, or using model-based techniques like k-nearest neighbors) can mitigate this issue, but they must be chosen carefully to avoid introducing biases or skewing data distributions.

\paragraph{Noise}
Noise in tabular datasets refers to irrelevant, erroneous, or random variations in data that do not reflect underlying patterns or relationships. This can arise from measurement errors, data entry mistakes, or inherent variability in the data-generating process. Noise can obscure meaningful signals, making it harder for models to learn accurate patterns and leading to reduced predictive performance~\citep{hasan2022noise,ren2025deep}. Moreover, models with high capacity, such as neural networks, are particularly susceptible to overfitting noise, mistaking it for important information. Effective noise handling strategies include robust preprocessing techniques, feature engineering, and regularization methods such as dropout or weight decay. These approaches can help mitigate the impact of noise and improve model robustness.

\paragraph{Imbalanced classes and data distribution issues}
In many classification problems with tabular data, the classes may be imbalanced (e.g., fraud detection, rare disease prediction), meaning one class has significantly fewer samples than another~\citep{panagiotou2024synthetic}. This can lead to models that are biased toward the majority class and have poor performance on minority classes~\citep{leevy2018survey}. Techniques like class weights, oversampling, undersampling, and cost-sensitive learning can help mitigate class imbalance and improve model performance on rare events. \\

Tabular data plays a fundamental role in many machine learning applications due to its flexibility and structure. However, successfully applying machine learning to tabular data requires addressing a series of challenges, from preprocessing diverse feature types and handling missing data to capturing complex feature interactions or ensuring model interpretability. 
In this context, data preprocessing plays a particularly critical role, with a wide repertoire of existing methods available, varying across different model types. For instance, neural networks typically do not require specific preprocessing for categorical features, as these are often handled through an embedding layer. In contrast, tree-based models may employ techniques such as one-hot encoding or target encoding, with the choice depending on factors such as feature cardinality and model size. The diversity of preprocessing techniques further complicates the comparison of different model types, as it is challenging to establish a unified evaluation framework applicable to all methods~\citep{tschalzev2024data}.

\subsection{Distinction between Typical Tabular Data and Time Series Data}
While common tabular data and time series data both share a similar grid-like structure, they have distinct characteristics that lead to different modeling challenges. In standard tabular data, rows represent independent observations with no inherent order or sequence. Each sample is treated as being independent and identically distributed (i.i.d.), meaning that models focus on learning patterns from individual feature relationships rather than capturing dependencies between rows\footnote{Although the assumption that tabular data is i.i.d. remains prevalent, recent studies have demonstrated that many tabular datasets inherently contain temporal components that are often overlooked or not explicitly modeled~\citep{tschalzev2024data}. By employing techniques such as test-time adaptation, it is possible to capture distribution shifts in the data over time. In such cases, the i.i.d. assumption is no longer valid. However, this thesis does not explore test-time adaptation and will adhere to the i.i.d. assumption throughout.}. In contrast, time series data involves observations that are ordered in time, where each row not only represents an instance but is also linked to a specific temporal sequence. This ordering introduces dependencies across rows, meaning that the value of a data point at one time step can be influenced by its preceding time steps. Consequently, time series data requires specialized techniques to handle temporal dependencies, such as lag-based features, sliding windows, or models that can inherently capture temporal dependencies (e.g., autoregressive models or recurrent neural networks). Moreover, the temporal aspect often necessitates dealing with issues like seasonality, trends, and stationarity, which are not typically present in standard tabular data. Therefore, while time series data can be represented in a tabular format, its sequential nature requires unique preprocessing and modeling strategies distinct from those used for common tabular datasets.
In addition to i.i.d. tabular data and time series data, some datasets can be viewed as tabular with a temporal component. These datasets lack the strict temporal ordering characteristic of time series but include temporal features (e.g., a datetime field in transaction data) that violate the i.i.d. assumption. Prior studies have acknowledged these issues~\citep{tschalzev2024data,rubachev2024tabred}, but still, such datasets are still predominantly treated as standard tabular data. In this thesis, we use common benchmarks for tabular data and do not specifically exclude any datasets from our evaluation.

\subsection{State-of-the-art Methods for Tabular Data}

Existing methods for tabular data can be divided into tree-based, deep learning and hybrid methods. 
In the following, we categorize the most prominent methods based on these three categories and differentiate our approach from existing work.
For a more comprehensive review, we refer to \citet{borisov2022deep}, \citet{shwartztabular} and \citet{grinsztajn2022tree}.

\paragraph{Tree ensemble methods}
Tree-based methods have been widely used for tabular data due to their interpretability and ability to capture non-linear relationships. While individual trees usually offer a higher interpretability, tree ensembles are commonly used to achieve superior performance~\citep{breiman2001randomforest}. State-of-the-art approaches usually rely on gradient boosting, wherein trees are constructed sequentially, with each tree corrects the errors of its predecessor~\citep{friedman2001gradientboostingtree}.
The most prominent tree-based methods for tabular data further improve upon the gradient boosting algorithm. %
XGBoost introduces a more regularized model formulation to control overfitting, making it one of the dominant methods in machine learning competitions~\citep{chen2016xgboost}. 
CatBoost is a gradient boosting algorithm for oblivious trees that specifically handles categorical variables without the need for extensive preprocessing and excels on datasets with many categorical variables~\citep{prokhorenkova2018catboost}. 
LightGBM employs histogram-based techniques, allowing it to bundle features into discrete bins, thus accelerating training and reducing memory usage. The algorithm uses a leaf-wise growth strategy, as opposed to the traditional depth-first or level-wise approaches, which can lead to increased accuracy given the same number of leaves.~\citep{ke2017lightgbm}.
Regarding the structure, GRANDE is similar to existing tree-based models. The main difference is the end-to-end gradient-based training procedure, which offers additional flexibility, and the instance-wise weighting. %

\paragraph{Deep learning methods for tabular data}
With the success of deep learning in various domains, researchers have started to adjust deep learning architectures, including transformers, to tabular data~\citep{ft_transformer, arik2021tabnet, huang2020tabtransformer, cai2021armnet, kossen2021selfattention}.
For example, \citet{ft_transformer} proposed Tabular ResNet, which uses a deep residual network optimized for tabular data, and MLP-PLR~\citep{gorishniy2022embeddings}, a multi-layer perceptron variant that includes periodic embeddings for numerical features which is still used in newly proposed methods like TabM~\citep{gorishniy2024tabm}. Additionally, their FT-Transformer extends transformer-based methods specifically for tabular tasks by incorporating effective tokenization and embedding strategies~\citep{ft_transformer}. Another method that recently gained attention is Self-Attention and Intersample Attention Transformer (SAINT) using attention over both, rows and columns~\citep{somepalli2021saint} to boost the performance.
Although GRANDE, similar to deep learning methods, uses end-to-end gradient descent for training, it has a shallow, hierarchical structure comprising hard, axis-aligned splits.

\paragraph{Hybrid methods}
Hybrid methods aim to combine the strengths of a gradient-based optimization with other algorithms, most commonly tree-based methods~\citep{abutbul2020dnfnet,hehn2020end,chen2020attentionforests,ke2019deepgbm,ke2018tabnn,katzir2020net}. One prominent way to achieve this is using SDTs to apply gradient descent by replacing hard decisions with soft ones, and axis-aligned with oblique splits~\citep{soft3,kontschieder2015neuraldecisionforests,luo2021sdtr}. %
Neural Oblivious Decision Ensembles (NODE) is one prominent hybrid method, already introduced in Section~\ref{ssec:node} which learns ensembles of oblivious DTs with gradient descent and is therefore closely related to our work~\citep{popov2019neural}. %
NODE-GAM extends NODE by integrating generalized additive models (GAM), while maintaining the benefits of deep learning~\citep{node_gam}. NODE-GAM is considered as more interpretable, as it is a GAM, but this comes with a decrease in performance compared to NODE.
Net-DNF~\citep{katzir2020net} use the inductive bias of Boolean formulas to design a neural network architecture which aims to imitate the characteristics of gradient boosting algorithms. However, they did not propose a solution to handling (high-dimensional) categorical features~\citep{borisov2022deep}.
\citet{cheng2016wide} propose Wide\&Deep which combines a deep neural network with categorical embeddings and simple linear connections. 
Furthermore, there are hybrid methods, that either combine different neural networks architectures~\citep{cheng2016wide,luo2020network,liu2020dnn2lr} or neural networks with non-tree-based methods like Factorization Machines~\citep{guo2017deepfm,lian2018xdeepfm}, which are frequently used for specific tasks like click-through-rate prediction.
GRANDE can also be categorized as a hybrid method. The main difference to existing methods is the use of hard, axis-aligned splits, which prevents overly smooth solutions typically inherent in soft, oblique trees. 
While GRANDE also allows stacking multiple layers to compose a deep model similar to, e.g., NODE, we use GRANDE as a shallow model with only one layer. \\

Recent studies indicate that, despite huge effort in finding high-performing deep learning methods, tree-based models still outperform deep learning for tabular data~\citep{grinsztajn2022tree, borisov2022deep, shwartztabular,tschalzev2024data}. However, they also highlight the need for gradient-based methods due to their flexibility.
One main reason for the superior performance of tree-based methods lies in the use of axis-aligned splits~\citep{grinsztajn2022tree, borisov2022deep}. Typically, DL methods are biased towards smooth solutions~\citep{rahaman2019spectral}. As the target function in tabular datasets is usually not smooth, DL methods struggle to find these irregular functions. In contrast, models that are based on hard, axis aligned DTs learn piece-wise constant functions and therefore do not show such a bias~\citep{grinsztajn2022tree}. Therefore, GRANDE aligns with this argument and utilizes hard, axis-aligned splits, while incorporating the benefits and flexibility of a gradient-based optimization.

\section[Evaluation of Interpretable Decision Trees]{An Evaluation of Interpretable Decision Trees for Tabular Data}\label{sec:gradtree} %

Interpretability has become a key concern in machine learning, especially in domains like healthcare, finance, and law, where understanding how a model arrives at a decision is crucial for gaining trust and ensuring accountability. DTs  have long been favored for their inherent interpretability~\citep{molnar2020}. 
They provide a clear and visual representation of decision-making processes through simple, rule-based structures, enabling users to trace how specific features lead to predictions. This transparency arises from the fact that each path from the root to a leaf node represents a coherent sequence of decisions, making DTs particularly interpretable as they closely mirror human decision-making processes~\citep{hastie2009elements}. Consequently, we can regard hard, axis-aligned DTs as \emph{human-understandable} models, making them especially valuable in applications where model transparency is essential.
However, despite their advantages, conventional DT learning methods are often limited by their reliance on greedy algorithms that locally optimize the model at each node. We already showed this issue based on a small synthetic dataset in Chapter~\ref{cha:related} and will provide a real-world example during the evaluation.
While we already introduced alternative methods that tackle this issue, they often come with their own limitations.
Optimal trees, while globally optimized, often result in overfitting and are only feasible until a depth of 4. Evolutionary algorithms, while effectively overcoming the limitations of sequentially selecting the optimal split, are often challenging to learn and often struggle to provide good results even for moderately complex datasets, as we will show during our evaluation.
Methods that introduce oblique splits, where decisions are made based on linear combinations of features, make it harder for non-experts to understand how the model reaches its decisions. Such methods fail to clearly indicate feature relationships and decision paths that make DTs easy to interpret in the first place.
GradTree was designed to address these issues without compromising on interpretability. By using a gradient-based approach, GradTree can optimize the entire tree structure jointly rather than sequentially, ensuring that the model performs well while still maintaining the intuitive, axis-aligned splits that DTs are known for. Unlike oblique trees, GradTree focuses on preserving simple, univariate splits at each decision point, making the model’s reasoning process easy to follow. In this section, we will focus on hard, axis-aligned DTs and evaluate GradTree against existing methods regarding interpretability and performance.

\subsection{Experimental Setup}

The following experiments are designed to evaluate GradTree's predictive performance and interpretability in comparison to existing methods. 
In this context, we define interpretability as the degree to which a model’s internal decision-making process is transparently encoded in its structure. 
In this context, DTs are regarded as intrinsically interpretable models~\citep{molnar2020, gunning2019xai}, as they are inherently understandable by humans and can be easily visualized. 
Specifically, this means that each split decision is based on a single feature with a clear threshold, creating a transparent path from input to prediction that users can follow and reason about. 
This definition is rooted in the view proposed by \citet{miller2019explanation}, following \citet{biran2017explanation}, characterizing interpretability as “the degree to which an observer can understand the cause of a decision”. Similarly, \citet{gunning2019xai} emphasize that fully interpretable models provide complete and transparent explanations by adhering to domain-specific constraints, such as clearly defined feature relationships. 
This stands in contrast to more complex tree-based models, such as random forests and DT ensembles, where predictions rely on multiple individual trees, or SDTs, which consider multiple features at a single split. Therefore, assessing the interpretability of GradTree involves verifying that its splits are genuinely hard and axis-aligned and ensuring that the resulting trees remain sufficiently compact to support straightforward interpretation.

\paragraph{Dataset selection} In this section, we focus on simpler, small- to medium-sized datasets, mainly from the UCI repository \citep{ml_repository}. The primary reason is that individual DTs, including GradTree, are not well-suited for handling highly complex datasets with many features. Since our goal is interpretability, we prioritize datasets where DTs are a realistic modeling choice. While more complex models, such as tree ensembles, SDTs, or deep learning methods, can achieve higher performance, their interpretability is often limited. By using smaller datasets, we ensure that DTs remain effective and interpretable. In the next section (Section~\ref{sec:grande}), we will extend our evaluation to a more complex benchmark, specifically selecting datasets where simple models struggle, such as those where the performance gap between a DT and a state-of-the-art model like XGBoost is substantial.

\paragraph{Preprocessing}
For all datasets, we performed a standard preprocessing: Similar to \citet{popov2019neural}, we applied leave-one-out encoding to all categorical features and further performed a quantile transform, making each feature follow a normal distribution. %
We used a $80\%/20\%$ train-test split for all datasets. To account for class imbalance, we rebalanced datasets using SMOTE~\citep{chawla2002smote} if the minority class accounts for less than $\frac{25}{c-1} \%$ of the data, where $c$ is the number of classes.
For GradTree and DNDT, we used $20\%$ of the training data as validation data for early stopping.
As DL8.5 requires binary features, we discretized numeric features using quantile binning with $5$ bins and one-hot encoded categorical features.
Details and sources of the datasets are available in Appendix~\ref{A:dataset}

\paragraph{Methods}
We compared GradTree to the most prominent approach from each category (see Section \ref{cha:related}) to ensure a concise, yet holistic evaluation focusing on hard, axis-aligned DTs. Specifically, we selected the following methods:
\begin{itemize}
    \item \textbf{CART}: We use the sklearn~\citep{scikit-learn} implementation, which uses an optimized version of the CART algorithm. 
    While CART usually only uses Gini as impurity measure, we also allow information gain/entropy as an option during our hyperparameter optimization. Since this is also the relevant difference between CART and C4.5 nowadays, we decided to only use optimized CART, which is also the stronger benchmark.

    \item \textbf{Evolutionary DTs}: We use GeneticTree~\citep{genetictree_code} for an efficient learning of DTs with a genetic algorithm (Section~\ref{sec:evolutionary_dt}). 
    \item \textbf{DNDT}: We use the official DNDT implementation~\citep{dndt_code}. For a fair comparison, we enforce binary trees by setting the number of cut points to $1$ and ensure hard splits during inference.
    As suggested by \citet{dndt}, we limited DNDTs to datasets with no more than $12$ features, due to scalability issues.
    \item \textbf{Optimal DTs}: We use the official DL8.5 implementation~\citep{pydl85_code} including improvements from MurTree~\citep{murtree} which reduces the runtime substantially.

\end{itemize}

To ensure a fair comparison, we further applied a simple post-hoc pruning for GradTree to remove all branches with zero samples based on one pass of the training data. Similar to DNDT, we used a cross-entropy loss.

\paragraph{Hyperparameters}
We conducted a random search with cross-validation to determine the optimal hyperparameters. %
The complete list of relevant hyperparameters for each approach along with additional details on the selection are in the Appendix~\ref{A:hyperparams}.

\begin{table}[t]
\centering
\small
\caption[Binary Classification Performance]{\textbf{Binary Classification Performance.} We report macro F1-scores (mean $\pm$ stdev over $10$ trials) on test data with optimized hyperparameters. The rank of each method is presented in brackets. The datasets are sorted by the number of features.}
\resizebox{0.99\textwidth}{!}{
\begin{tabular}{lccccc}
\toprule
{} &                           \multicolumn{2}{l}{Gradient-Based} &            \multicolumn{2}{l}{Non-Greedy} &                       \multicolumn{1}{l}{Greedy} \\ \cmidrule(lr){2-3}  \cmidrule(lr){4-5} \cmidrule(lr){6-6}
  
{} &                           \multicolumn{1}{l}{GradTree (ours)} &                   \multicolumn{1}{l}{DNDT} &            \multicolumn{1}{l}{GeneticTree} &                          \multicolumn{1}{l}{DL8.5 (Optimal)} &                       \multicolumn{1}{l}{CART} \\
\midrule
Blood Transfusion                  &  \bftab 0.628 $\pm$ .036 (1) &         0.543 $\pm$ .051 (5) &         0.575 $\pm$ .094 (4) &  0.590 $\pm$ .034 (3) &         0.613 $\pm$ .044 (2) \\
Banknote Auth.          &  \bftab 0.987 $\pm$ .007 (1) &         0.888 $\pm$ .013 (5) &         0.922 $\pm$ .021 (4) &  0.962 $\pm$ .011 (3) &         0.982 $\pm$ .007 (2) \\ %
Titanic                            &  \bftab 0.776 $\pm$ .025 (1) &         0.726 $\pm$ .049 (5) &         0.730 $\pm$ .074 (4) &  0.754 $\pm$ .031 (2) &         0.738 $\pm$ .057 (3) \\
Raisins                            &         0.840 $\pm$ .022 (4) &         0.821 $\pm$ .033 (5) &  \bftab 0.857 $\pm$ .021 (1) &  0.849 $\pm$ .027 (3) &         0.852 $\pm$ .017 (2) \\
Rice                               &         0.926 $\pm$ .007 (3) &         0.919 $\pm$ .012 (5) &         0.927 $\pm$ .005 (2) &  0.925 $\pm$ .008 (4) &  \bftab 0.927 $\pm$ .006 (1) \\
Echocardiogram                     &  \bftab 0.658 $\pm$ .113 (1) &         0.622 $\pm$ .114 (3) &         0.628 $\pm$ .105 (2) &  0.609 $\pm$ .112 (4) &         0.555 $\pm$ .111 (5) \\
Wisconcin BC &         0.904 $\pm$ .022 (2) &  \bftab 0.913 $\pm$ .032 (1) &         0.892 $\pm$ .028 (4) &  0.896 $\pm$ .021 (3) &         0.886 $\pm$ .025 (5) \\ %
Loan House                         &  \bftab 0.714 $\pm$ .041 (1) &         0.694 $\pm$ .036 (2) &         0.451 $\pm$ .086 (5) &  0.607 $\pm$ .045 (4) &         0.662 $\pm$ .034 (3) \\
Heart Failure                      &         0.750 $\pm$ .070 (3) &         0.754 $\pm$ .062 (2) &         0.748 $\pm$ .068 (4) &  0.692 $\pm$ .062 (5) &  \bftab 0.775 $\pm$ .054 (1) \\
Heart Disease                      &  \bftab 0.779 $\pm$ .047 (1) &             $n>12$ &         0.704 $\pm$ .059 (4) &  0.722 $\pm$ .065 (2) &         0.715 $\pm$ .062 (3) \\
Adult                              &         0.743 $\pm$ .034 (2) &             $n>12$ &         0.464 $\pm$ .055 (4) &  0.723 $\pm$ .011 (3) &  \bftab 0.771 $\pm$ .011 (1) \\
Bank Marketing                     &  \bftab 0.640 $\pm$ .027 (1) &             $n>12$ &         0.473 $\pm$ .002 (4) &  0.502 $\pm$ .011 (3) &         0.608 $\pm$ .018 (2) \\
Congressional Voting               &  \bftab 0.950 $\pm$ .021 (1) &             $n>12$ &         0.942 $\pm$ .021 (2) &  0.924 $\pm$ .043 (4) &         0.933 $\pm$ .032 (3) \\ %
Absenteeism                        &  \bftab 0.626 $\pm$ .047 (1) &             $n>12$ &         0.432 $\pm$ .073 (4) &  0.587 $\pm$ .047 (2) &         0.564 $\pm$ .042 (3) \\
Hepatitis                          &         0.608 $\pm$ .078 (2) &             $n>12$ &         0.446 $\pm$ .024 (4) &  0.586 $\pm$ .083 (3) &  \bftab 0.622 $\pm$ .078 (1) \\
German                             &  \bftab 0.592 $\pm$ .068 (1) &             $n>12$ &         0.412 $\pm$ .006 (4) &  0.556 $\pm$ .035 (3) &         0.589 $\pm$ .065 (2) \\
Mushroom                           &  \bftab 1.000 $\pm$ .001 (1) &             $n>12$ &         0.984 $\pm$ .003 (4) &  0.999 $\pm$ .001 (2) &         0.999 $\pm$ .001 (3) \\
Credit Card                        &         0.674 $\pm$ .014 (4) &             $n>12$ &  \bftab 0.685 $\pm$ .004 (1) &  0.679 $\pm$ .007 (3) &         0.683 $\pm$ .010 (2) \\
Horse Colic                        &  \bftab 0.842 $\pm$ .039 (1) &             $n>12$ &         0.496 $\pm$ .169 (4) &  0.708 $\pm$ .038 (3) &         0.786 $\pm$ .062 (2) \\
Thyroid                            &         0.905 $\pm$ .010 (2) &             $n>12$ &         0.605 $\pm$ .116 (4) &  0.682 $\pm$ .018 (3) &  \bftab 0.922 $\pm$ .011 (1) \\
Cervical Cancer                    &  \bftab 0.521 $\pm$ .043 (1) &             $n>12$ &         0.514 $\pm$ .034 (2) &  0.488 $\pm$ .027 (4) &         0.506 $\pm$ .034 (3) \\
Spambase                           &         0.903 $\pm$ .025 (2) &             $n>12$ &         0.863 $\pm$ .019 (3) &  0.863 $\pm$ .011 (4) &  \bftab 0.917 $\pm$ .011 (1) \\
\midrule
MRR $\uparrow$                    &  \bftab 0.758 $\pm$ .306 (1) &         0.370 $\pm$ .268 (3) &         0.365 $\pm$ .228 (4) &  0.335 $\pm$ .090 (5) &         0.556 $\pm$ .293 (2) \\
MRD $\downarrow$ &        \bftab 0.008 $\pm$ .012 (1) &         0.056 $\pm$ .051 (3) &  0.211 $\pm$ .246 (5) &  0.084 $\pm$ .090 (4) &   0.035 $\pm$ .048 (2) \\
\bottomrule
\end{tabular}%
}

\label{tab:eval-results-binary}
\end{table}

\begin{table}[t]
\centering
\small
\caption[Multi-Class Classification Performance]{\textbf{Multi-Class Classification Performance.} We report macro F1-scores (mean $\pm$ stdev over $10$ trials) on test data with optimized hyperparameters and the rank of each method in brackets. The datasets are sorted by the number of features.}

\resizebox{0.99\textwidth}{!}{
\begin{tabular}{lccccc}
\toprule
{} &                           \multicolumn{2}{l}{Gradient-Based} &            \multicolumn{2}{l}{Non-Greedy} &                       \multicolumn{1}{l}{Greedy} \\ \cmidrule(lr){2-3}  \cmidrule(lr){4-5} \cmidrule(lr){6-6}
  
{} &                           \multicolumn{1}{l}{GradTree (ours)} &                   \multicolumn{1}{l}{DNDT} &            \multicolumn{1}{l}{GeneticTree} &                          \multicolumn{1}{l}{DL8.5 (Optimal)} &                       \multicolumn{1}{l}{CART} \\
\midrule
Iris                 &  \bftab 0.938 $\pm$ .057 (1) &         0.870 $\pm$ .063 (5) &  0.912 $\pm$ .055 (3) &  0.909 $\pm$ .046 (4) &         0.937 $\pm$ .046 (2) \\
Balance Scale        &  \bftab 0.593 $\pm$ .045 (1) &         0.475 $\pm$ .104 (5) &  0.529 $\pm$ .043 (3) &  0.525 $\pm$ .039 (4) &         0.574 $\pm$ .030 (2) \\
Car                  &         0.440 $\pm$ .085 (3) &         0.485 $\pm$ .064 (2) &  0.306 $\pm$ .068 (4) &  0.273 $\pm$ .063 (5) &  \bftab 0.489 $\pm$ .094 (1) \\
Glass                &         0.560 $\pm$ .090 (3) &         0.434 $\pm$ .072 (5) &  0.586 $\pm$ .090 (2) &  0.501 $\pm$ .100 (4) &  \bftab 0.663 $\pm$ .086 (1) \\
Contraceptive        &  \bftab 0.496 $\pm$ .050 (1) &         0.364 $\pm$ .050 (3) &  0.290 $\pm$ .048 (5) &  0.292 $\pm$ .036 (4) &         0.384 $\pm$ .075 (2) \\
Solar Flare          &         0.151 $\pm$ .033 (3) &  \bftab 0.171 $\pm$ .051 (1) &  0.146 $\pm$ .018 (4) &  0.144 $\pm$ .034 (5) &         0.157 $\pm$ .022 (2) \\
Wine                 &  \bftab 0.933 $\pm$ .031 (1) &         0.858 $\pm$ .041 (4) &  0.888 $\pm$ .039 (3) &  0.852 $\pm$ .022 (5) &         0.907 $\pm$ .042 (2) \\
Zoo                  &         0.874 $\pm$ .111 (3) &             $n>12$ &  0.782 $\pm$ .111 (4) &  0.911 $\pm$ .106 (2) &  \bftab 0.943 $\pm$ .076 (1) \\
Lymphography         &  \bftab 0.610 $\pm$ .191 (1) &             $n>12$ &  0.381 $\pm$ .124 (4) &  0.574 $\pm$ .196 (2) &         0.548 $\pm$ .154 (3) \\
Segment              &         0.941 $\pm$ .009 (2) &             $n>12$ &  0.715 $\pm$ .114 (4) &  0.808 $\pm$ .013 (3) &  \bftab 0.963 $\pm$ .010 (1) \\
Dermatology          &         0.930 $\pm$ .030 (2) &             $n>12$ &  0.785 $\pm$ .126 (4) &  0.885 $\pm$ .036 (3) &  \bftab 0.957 $\pm$ .026 (1) \\
Landsat              &         0.807 $\pm$ .011 (2) &             $n>12$ &  0.628 $\pm$ .084 (4) &  0.783 $\pm$ .008 (3) &  \bftab 0.835 $\pm$ .011 (1) \\
Annealing            &         0.638 $\pm$ .126 (3) &             $n>12$ &  0.218 $\pm$ .053 (4) &  0.787 $\pm$ .121 (2) &  \bftab 0.866 $\pm$ .094 (1) \\
Splice               &         0.873 $\pm$ .030 (2) &             $n>12$ &  0.486 $\pm$ .157 (3) &      $>$ 60 min &  \bftab 0.881 $\pm$ .021 (1) \\
\midrule
MRR $\uparrow$  &         0.619 $\pm$ .303 (2) &         0.383 $\pm$ .293 (3) &  0.288 $\pm$ .075 (5) &  0.315 $\pm$ .116 (4) &  \bftab 0.774 $\pm$ .274 (1) \\
MRD $\downarrow$ &         0.069 $\pm$ .102 (2) &         0.188 $\pm$ .200 (3) &  0.521 $\pm$  .749 (5) &  0.215 $\pm$ .249 (4) &  \bftab 0.040 $\pm$ .081 (1) \\

\bottomrule
\end{tabular}%
}

\label{tab:eval-results-multi}
\end{table}

\subsection{Results}\label{ssec:experiment_results}

\begin{figure}[t]
\centering
\begin{subfigure}{0.5\textwidth}
  \centering
  \includegraphics[width=0.875\columnwidth]{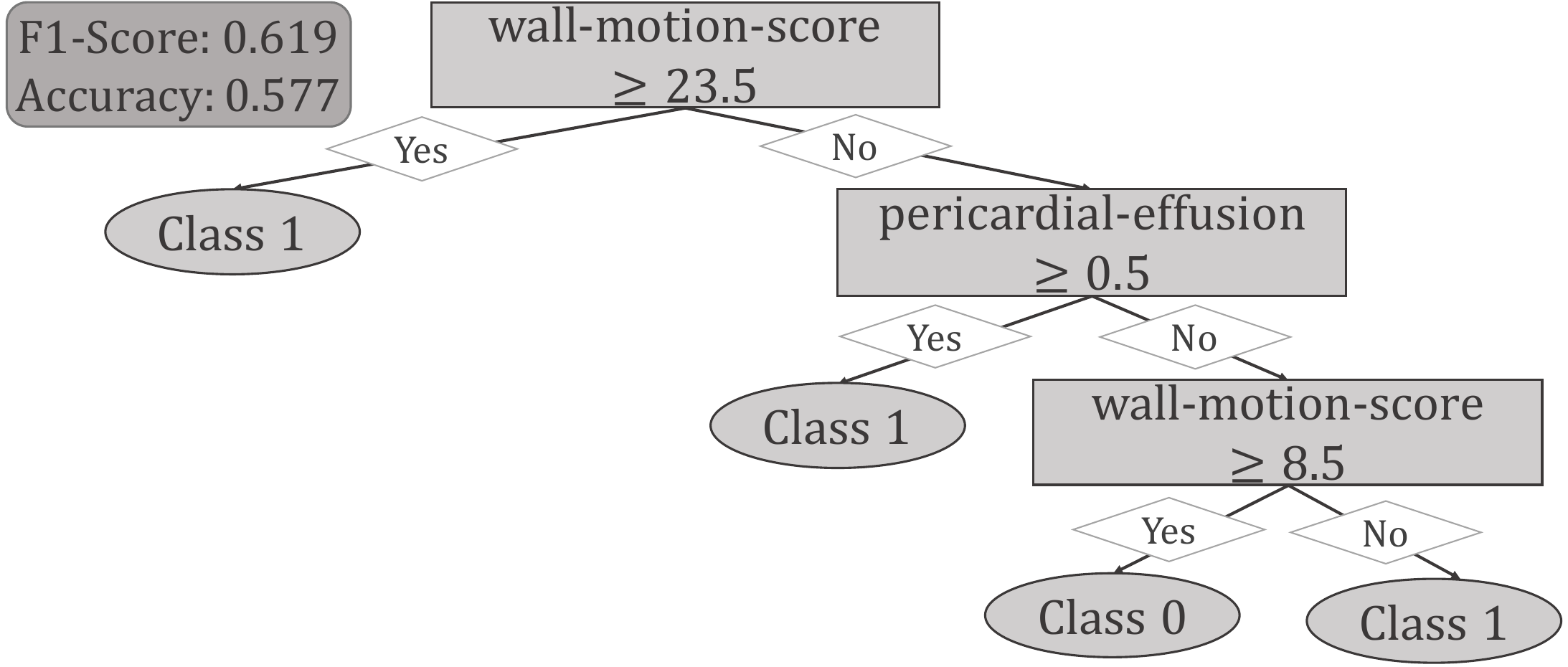}
  \caption{Greedy DT (CART)}
  \label{fig:cart_dt_example_revisited}
  
\end{subfigure}%
\begin{subfigure}{0.5\textwidth}
  \centering
  \includegraphics[width=0.875\columnwidth]{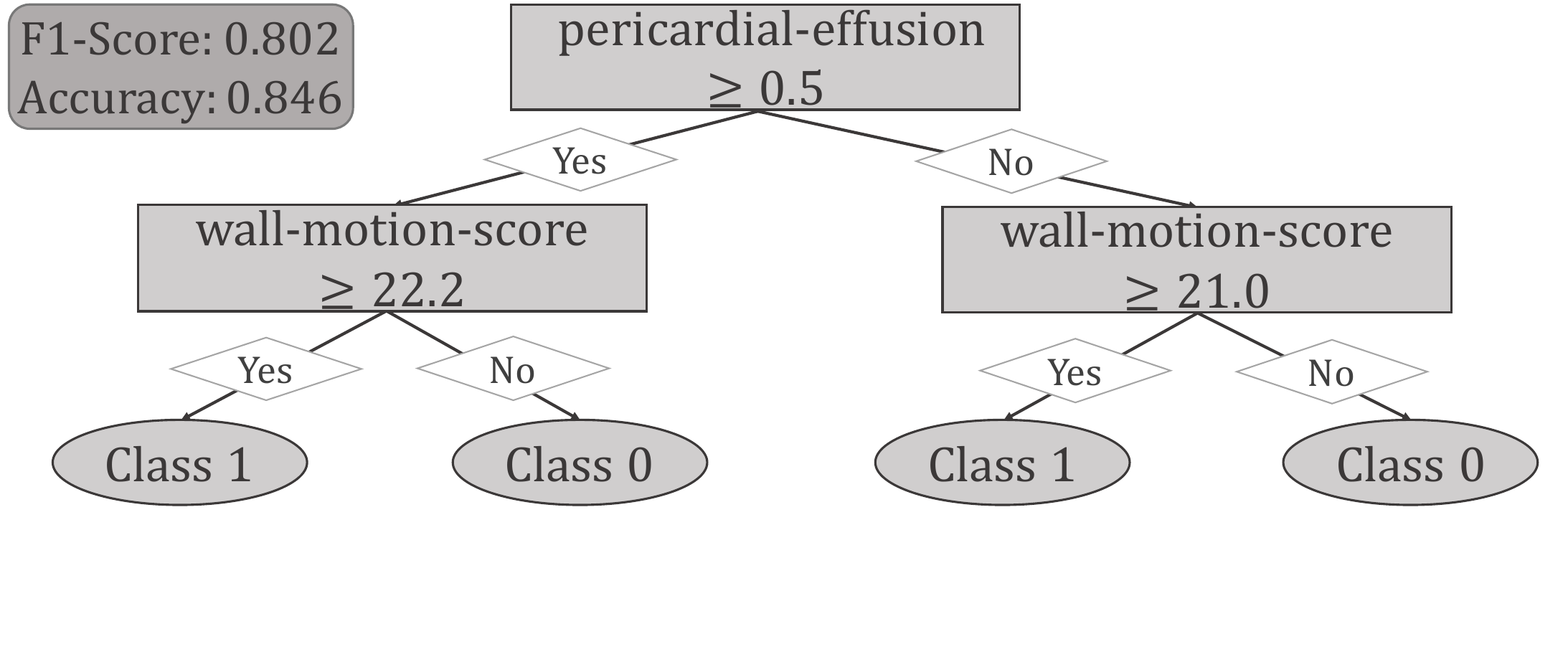}
  \caption{Gradient-Based DT (GradTree)}
  \label{fig:gdt_example_revisited}
\end{subfigure}
\caption[Greedy vs. Gradient-Based DT]{\textbf{Greedy vs. Gradient-Based DT.} Two DTs trained on the Echocardiogram dataset. 
The CART DT (A) makes only locally optimal splits, while GradTree (B) jointly optimizes all parameters, leading to significantly better performance.
}
\label{fig:dt_comparison_example_revisited}
\end{figure}

\paragraph{GradTree can learn interpretable DTs and is not prone to locally optimal solutions}
First, we revisit the Echocardiogram dataset~\citep{ml_repository}, which is used to predict the one-year survival of patients following a heart attack. This dataset, based on tabular data from echocardiograms, was previously introduced in our introductory example. Figure~\ref{fig:dt_comparison_example_revisited} shows two DTs. %
The tree on the left is learned by CART while the one on the right is learned with GradTree approach.
It is evident that the greedy procedure produces a tree with considerably poorer performance. While splitting on the \emph{wall-motion-score} is the best local choice (see Figure~\ref{fig:cart_dt_example_revisited}), on a global scale it proves more advantageous to split on the \emph{wall-motion-score} with different values based on the \emph{pericardial-effusion} at the second level (Figure~\ref{fig:gdt_example_revisited}).
Moreover, it is evident that GradTree can effectively learn DTs for real-world tasks, offering the same level of interpretability as traditional DTs with hard, axis-aligned splits.

\paragraph{GradTree outperforms existing DT learners for binary classification}
Next, we evaluated the performance of GradTree against existing approaches on the benchmark datasets in terms of the macro F1-Score, which inherently considers class imbalance. We report the relative difference to the best model (MRD) and mean reciprocal rank (MRR), following the approach of \citet{dndt}.
Overall, GradTree outperformed existing approaches for binary classification tasks (best MRR of $0.758$ and MRD of $0.008$) and achieved competitive results for multi-class tasks (second-best MRR of $0.619$ and MRD of $0.069$). 
More specifically, GradTree substantially outperformed state-of-the-art non-greedy DT methods, including DNDTs as our gradient-based benchmark.
For binary classification (Table~\ref{tab:eval-results-binary}), GradTree demonstrated superior performance over CART, achieving the best performance on $13$ datasets as compared to only $6$ datasets for CART. 
Notably, the performance difference between GradTree and existing methods was substantial for several datasets, such as \emph{Echocardiogram}, \emph{Heart Disease} and \emph{Absenteeism}.
For multi-class datasets, GradTree achieved the second-best overall performance. 
While GradTree still achieved a superior performance for low-dimensional datasets (top part of Table~\ref{tab:eval-results-multi}), CART achieved the best results for higher-dimensional datasets with a high number of classes.
We can explain this by the dense representation used for the gradient-based optimization. Using our representation, the difficulty of the optimization task increases with the number of features (more parameters at each internal node) and the number of classes (more parameters at each leaf node).

\begingroup
\setlength{\intextsep}{0pt}%
\begin{wraptable}{r}{0.525\textwidth}
\begin{minipage}{0.525\textwidth}
\small
    \centering
    \caption[GradTree Overfitting Comparison]{\textbf{Train-Test Difference.} Mean difference between train and test performance as overfitting indicator. Detailed results are in Table~\ref{tab:eval-results-train}.}%
        \begin{tabular}{lrrrrrr}
        \toprule 
        & \multicolumn{1}{l}{Binary} & \multicolumn{1}{l}{Multi}  \\
        
        \midrule
        GradTree                &  0.051     & 0.174 \\ 
        DNDT                    & 0.039      & 0.239  \\
        GeneticTree             & 0.204      & 0.258  \\
        DL8.5                   & 0.202      & 0.260  \\
        CART                    & 0.183      & 0.247 \\
        \bottomrule
        \end{tabular}

        \label{tab:train_test_gradtree}
\end{minipage}
\end{wraptable}
\paragraph{GradTree is robust to overfitting}
We can observe that gradient-based approaches were more robust and less prone to overfitting compared to a greedy optimization with CART and alternative non-greedy methods. We measure overfitting by the difference between the mean train and test performance (see Table~\ref{tab:train_test_gradtree}). 
For binary tasks, GradTree exhibits a train-test performance difference of $0.051$, considerably smaller than that of CART ($0.183$), GeneticTree ($0.204$), and DL8.5 ($0.202$).
DNDTs, which are also gradient-based, achieved an even smaller difference of $0.039$. For multi-class tasks, the difference was substantially smaller for GradTree compared to any other approach. %

\begin{table}[t]

\centering
\caption[Performance Comparison Default Hyperparameters]{\textbf{Performance Comparison Default Hyperparameters.} We report macro F1-scores (mean $\pm$ stdev over $10$ trials) with default parameters and the rank of each approach in brackets. The datasets are sorted by the number of features. The top part comprises binary classification tasks and the bottom part multi-class datasets.}
\resizebox{1.0\columnwidth}{!}{

\begin{tabular}{lccccc}
\toprule
{} &                           \multicolumn{2}{l}{Gradient-Based} &            \multicolumn{2}{l}{Non-Greedy} &                       \multicolumn{1}{l}{Greedy} \\ \cmidrule(lr){2-3}  \cmidrule(lr){4-5} \cmidrule(lr){6-6}
  
{} &                           \multicolumn{1}{l}{GradTree (ours)} &                   \multicolumn{1}{l}{DNDT} &            \multicolumn{1}{l}{GeneticTree} &                          \multicolumn{1}{l}{DL8.5 (Optimal)} &                       \multicolumn{1}{l}{CART} \\
\midrule
Blood Transfusion                  &  \bftab 0.627 $\pm$ .056 (1) &         0.558 $\pm$ .059 (5) &         0.574 $\pm$ .093 (3) &         0.590 $\pm$ .034 (2) &         0.573 $\pm$ .036 (4) \\
Banknote Auth.            &         0.980 $\pm$ .008 (2) &         0.886 $\pm$ .024 (5) &         0.928 $\pm$ .022 (4) &         0.962 $\pm$ .011 (3) &  \bftab 0.981 $\pm$ .006 (1) \\
Titanic                            &  \bftab 0.782 $\pm$ .035 (1) &         0.740 $\pm$ .026 (4) &         0.763 $\pm$ .041 (2) &         0.754 $\pm$ .031 (3) &         0.735 $\pm$ .041 (5) \\
Raisins                            &         0.850 $\pm$ .025 (2) &         0.832 $\pm$ .026 (4) &  \bftab 0.859 $\pm$ .022 (1) &         0.849 $\pm$ .027 (3) &         0.811 $\pm$ .028 (5) \\
Rice                               &         0.926 $\pm$ .006 (2) &         0.902 $\pm$ .018 (5) &  \bftab 0.927 $\pm$ .005 (1) &         0.925 $\pm$ .008 (3) &         0.905 $\pm$ .010 (4) \\
Echocardiogram                     &  \bftab 0.648 $\pm$ .130 (1) &         0.543 $\pm$ .117 (5) &         0.563 $\pm$ .099 (3) &         0.609 $\pm$ .112 (2) &         0.559 $\pm$ .075 (4) \\
Wisconsin Breast Cancer &         0.902 $\pm$ .029 (3) &  \bftab 0.907 $\pm$ .037 (1) &         0.888 $\pm$ .021 (5) &         0.896 $\pm$ .021 (4) &         0.904 $\pm$ .025 (2) \\
Loan House                         &  \bftab 0.695 $\pm$ .035 (1) &         0.475 $\pm$ .043 (4) &         0.461 $\pm$ .093 (5) &         0.607 $\pm$ .045 (3) &         0.671 $\pm$ .056 (2) \\
Heart Failure                      &         0.745 $\pm$ .063 (2) &         0.580 $\pm$ .077 (5) &         0.740 $\pm$ .055 (3) &         0.692 $\pm$ .062 (4) &  \bftab 0.755 $\pm$ .060 (1) \\
Heart Disease                      &  \bftab 0.736 $\pm$ .069 (1) &                        $n>12$ &         0.704 $\pm$ .059 (3) &         0.722 $\pm$ .065 (2) &         0.670 $\pm$ .090 (4) \\
Adult                              &         0.749 $\pm$ .028 (2) &                        $n>12$ &         0.478 $\pm$ .068 (4) &         0.723 $\pm$ .011 (3) &  \bftab 0.773 $\pm$ .008 (1) \\
Bank Marketing                     &  \bftab 0.626 $\pm$ .017 (1) &                        $n>12$ &         0.473 $\pm$ .002 (4) &         0.502 $\pm$ .011 (3) &         0.616 $\pm$ .007 (2) \\
Congressional Voting               &  \bftab 0.953 $\pm$ .021 (1) &                        $n>12$ &         0.932 $\pm$ .034 (3) &         0.924 $\pm$ .043 (4) &         0.933 $\pm$ .032 (2) \\
Absenteeism                        &  \bftab 0.620 $\pm$ .050 (1) &                        $n>12$ &         0.417 $\pm$ .035 (4) &         0.587 $\pm$ .047 (2) &         0.580 $\pm$ .045 (3) \\
Hepatitis                          &         0.609 $\pm$ .103 (2) &                        $n>12$ &         0.486 $\pm$ .074 (4) &         0.586 $\pm$ .083 (3) &  \bftab 0.610 $\pm$ .123 (1) \\
German                             &         0.584 $\pm$ .057 (2) &                        $n>12$ &         0.412 $\pm$ .005 (4) &         0.556 $\pm$ .035 (3) &  \bftab 0.595 $\pm$ .028 (1) \\
Mushroom                           &         0.997 $\pm$ .003 (3) &                        $n>12$ &         0.984 $\pm$ .003 (4) &  \bftab 0.999 $\pm$ .001 (1) &         0.999 $\pm$ .001 (2) \\
Credit Card                        &         0.676 $\pm$ .009 (4) &                        $n>12$ &  \bftab 0.685 $\pm$ .004 (1) &         0.679 $\pm$ .007 (3) &         0.679 $\pm$ .007 (2) \\
Horse Colic                        &  \bftab 0.842 $\pm$ .033 (1) &                        $n>12$ &         0.794 $\pm$ .042 (2) &         0.708 $\pm$ .038 (4) &         0.758 $\pm$ .053 (3) \\
Thyroid                            &         0.872 $\pm$ .015 (2) &                        $n>12$ &         0.476 $\pm$ .101 (4) &         0.682 $\pm$ .018 (3) &  \bftab 0.912 $\pm$ .013 (1) \\
Cervical Cancer                    &         0.496 $\pm$ .047 (3) &                        $n>12$ &  \bftab 0.514 $\pm$ .034 (1) &         0.488 $\pm$ .027 (4) &         0.505 $\pm$ .033 (2) \\
Spambase                           &         0.893 $\pm$ .015 (2) &                        $n>12$ &         0.864 $\pm$ .014 (3) &         0.863 $\pm$ .011 (4) &  \bftab 0.917 $\pm$ .011 (1) \\
\midrule
Mean $\uparrow$                            &  \bftab 0.764 $\pm$ .144 (1) &                             - &         0.678 $\pm$ .197 (4) &         0.723 $\pm$ .154 (3) &         0.747 $\pm$ .153 (2) \\
MRR $\uparrow$               &  \bftab 0.670 $\pm$ .289 (1) &         0.306 $\pm$ .262 (5) &         0.427 $\pm$ .287 (3) &         0.371 $\pm$ .164 (4) &         0.571 $\pm$ .318 (2) \\
\bottomrule
\end{tabular}%
}
\label{tab:eval-results_noHPO}
\end{table}

\paragraph{GradTree does not rely on extensive hyperparameter optimization}
Besides their interpretability, a distinct advantage of DTs over more sophisticated models is that they typically do not rely on an extensive hyperparameter optimization.
In this experiment, we show that the same is true for GradTree by evaluating the performance with default configurations. When using the default parameters, GradTree still outperformed the other methods on binary tasks (highest MRR and most wins, see Table~\ref{tab:eval-results_noHPO}) and achieved competitive results for multi-class tasks (second-highest MRR, see Table~\ref{tab:eval-results_noHPO_complete}).

\begingroup
\setlength{\intextsep}{0pt}%
\begin{wraptable}{r}{0.525\textwidth}
\begin{minipage}{0.525\textwidth}
\small
    \centering
    \caption[GradTree Tree Size]{\textbf{Tree Size Summary.} Average tree size. Detailed results are in Table~\ref{tab:eval-results-tree-size}.}%
        \begin{tabular}{lrrrrrr}
        \toprule 
        & \multicolumn{1}{l}{Binary} & \multicolumn{1}{l}{Multi}  \\
        
        \midrule
        GradTree                & \phantom{0}54           & \phantom{0}86    \\ 
         DNDT                    & 887                     & 907             \\
        GeneticTree             & \phantom{00}7           & \phantom{0}20    \\
        DL8.5                   & \phantom{0}28           & \phantom{0}29    \\
        CART                    & \phantom{0}67           & \phantom{0}76    \\
        \bottomrule
        \end{tabular}

        \label{tab:tree_size_gradtree}
\end{minipage}
\end{wraptable}
\paragraph{GradTree has a small effective tree size}
The effective tree size (= size after pruning) of GradTree is smaller than CART for binary and marginally higher for multi-class tasks (Table~\ref{tab:tree_size_gradtree}). Only the tree size for GeneticTree is substantially smaller, which is caused by the complexity penalty of the genetic algorithm. DL8.5 also has a smaller average tree size than CART and GradTree. 
We can attribute this to DL8.5 being only feasible up to a depth of $4$ due to the high computational complexity.
The tree size for DNDTs scales with the number of features (and classes) which quickly results in large trees. Furthermore, pruning DNDTs is non-trivial due to the use of the Kronecker product (it is not sufficient to prune subtrees bottom-up).%

\paragraph{GradTree is efficient for large and high-dimensional datasets}
For each dataset, a greedy optimization using CART was substantially faster than other methods, taking less than a second.
Nevertheless, for most datasets, training GradTree took less than $30$ seconds (mean runtime of $35$ seconds). DNDT had comparable runtimes to GradTree. 
For most datasets, DL8.5 had a low runtime of less than 10 seconds.
However, scalability issues become apparent with DL8.5, especially with an increasing number of features and samples. Its runtime surpassed GradTree on various datasets, taking around 300 seconds for \emph{Credit Card} and exceeding 830 seconds for \emph{Landsat}.
For \emph{Splice}, DL8.5 did not find a solution within $60$ minutes and the optimization was terminated. Detailed runtimes are in Table~\ref{tab:eval-results-runtime}.

\paragraph{ST entmax outperforms alternative methods}\label{sssec:ablation}
In an ablation study (Table~\ref{tab:ablation_study}), we evaluated our design choice of utilizing an ST operator directly after an entmax transformation to address the non-differentiability of DTs. We contrasted this against alternative strategies found in the literature. 
As discussed in Section~\ref{ssec:adjusted_backprop}, alternative methods for addressing non-differentiability often adopt a probabilistic perspective on discrete parameters, employing sampling techniques to approximate gradients. This approach, however, introduces stochasticity into routing and split selection, which can significantly compromise the training process. The findings of this ablation study substantiate these theoretical assumptions.
Our approach notably surpassed ST Gumbel Softmax~\citep{gumbel} and outperformed the temperature annealing technique proposed by \citet{node_gam} to gradually turn the entmax one-hot.

\begin{table}[t]
\centering
\small

\caption[Ablation Study]{\textbf{Ablation Study.} We compare our approach to deal with the non-differentiable nature of DTs with alternative methods, reporting the average macro F1-scores over $10$ trials with optimized and default hyperparameters. The complete results are listed in Table~\ref{tab:ablation_HPO}-\ref{tab:ablation_noHPO}.} %
\resizebox{0.85\columnwidth}{!}{
\begin{tabular}{llrrr}
\toprule 
\small
 & & \multicolumn{1}{L{2.75cm}}{ST Entmax (ours)} & \multicolumn{1}{L{2.0cm}}{ST Gumbel} & \multicolumn{1}{L{2.75cm}}{Temp. Annealing} \\

\midrule

\multirow{2}{*}{Default} & Binary         & \bftab 0.764      & 0.560     & 0.757 \\ 
                        & Multi    & \bftab 0.638      & 0.272    & 0.602 \\ %
                        \midrule
                             
\multirow{2}{*}{Optimized}    & Binary         & \bftab 0.771      & 0.569     & 0.759 \\                 
                        & Multi    & \bftab 0.699      & 0.297     & 0.601 \\                   
\bottomrule
\end{tabular}%

}

\label{tab:ablation_study}
\end{table}

\subsection{Summary}
In this section, we evaluated GradTree, a novel gradient-based method for learning axis-aligned DTs. GradTree consistently outperformed alternative methods, demonstrating a substantial advantage over alternative non-greedy approaches, including DNDTs, across binary classification tasks. Notably, GradTree achieved the best MRR ($0.758$) and MRD ($0.008$) for binary tasks, surpassing even CART as non-greedy benchmark, which is still considered as state-of-the art. The substantial performance increase achieved by GradTree across multiple datasets highlights its importance as a noteworthy contribution to the existing repertoire of DT learning methods. 
GradTree's design allows for robust optimization of both split indices and thresholds, which is unique for gradient-based methods. Furthermore, a gradient-based optimization results in less overfitting compared to other methods. 
Moreover, gradient-based optimization provides greater flexibility, allowing for easy integration of custom loss functions tailored to specific application scenarios.
Another advantage is the ability to relearn the threshold value as well as the split index. Therefore, GradTree is suitable for dynamic environments, such as online learning tasks or reinforcement learning, as we will show in Chapter~\ref{cha:sympol}.

\section[Evaluation of Gradient-Based Decision Tree Ensembles]{An Evaluation of Gradient-Based Decision Tree Ensembles for Complex Tabular Datasets.} \label{sec:grande} %

Despite the success of deep learning in various domains, recent studies indicate that tabular data still poses a major challenge and tree-based models like XGBoost and CatBoost outperform them in most cases~\citep{borisov2022deep,grinsztajn2022tree,shwartztabular}.
At the same time, employing end-to-end gradient-based training provides several advantages over traditional machine learning methods~\citep{borisov2022deep}. They offer a high level of flexibility by allowing an easy integration of arbitrary, differentiable loss functions tailored towards specific problems and support iterative training~\citep{online1}. Moreover, gradient-based methods can be incorporated easily into multimodal learning (see Section~\ref{sec:multimodal}), with tabular data being one of several input types~\citep{multimodal1,multimodal2,multimodal3,multimodal4}. %
Therefore, creating tabular-specific, gradient-based methods is a very active field of research and the need for well-performing methods is intense~\citep{grinsztajn2022tree}.
In this section, we will conduct an extensive evaluation on 19 binary classification tasks based on the predefined tabular benchmark proposed by \citet{benchmarksuites}. 
Thereby, we will evaluate GRANDE against existing state-of-the-art approaches, including GBDT and DL methods.
Additionally, we will compare GRANDE with standard DTs and GradTree to show the increase in performance when using a more sophisticated model.

\subsection{Experimental Setup} %

As pointed out by \citet{grinsztajn2022tree}, most papers presenting a new method for tabular data have a highly varying evaluation methodology, with a small number of datasets that might be biased towards the authors’ model. As a result, recent surveys showed that tree boosting methods like XGBoost and CatBoost are still state-of-the-art and outperform new architectures for tabular data on most datasets~\citep{grinsztajn2022tree,shwartztabular,borisov2022deep}.
This highlights the necessity for an extensive and unbiased evaluation, as we will carry out in the following, to accurately assess the performance of a new method and draw valid conclusions. 
We want to emphasize that recent surveys and evaluation on predefined benchmarks indicate that there is no “one-size-fits-all” solution for tabular data. Consequently, we should view new methods as an extension to the existing repertoire and set our expectations accordingly. %

\begin{table}[t]
    \centering
    \small
    \caption[Datasets]{\textbf{Datasets.} A list of tabular datasets along with their characteristics used for the evaluation.}
    \resizebox{\columnwidth}{!}{
    \begin{tabular}{lrrrrrr}
    \toprule
     & \makecell[l]{Samples} & \makecell[l]{Features} & \makecell[l]{Categorical \\ Features} & \makecell[l]{Features \\ (preprocessed)} & \makecell[l]{Minority \\ Class} & \makecell[l]{OpenML ID} \\
    \midrule
        dresses-sales & 500 & 12 & 11 & 37 & 42.00\% & 23381 \\
        climate-simulation-crashes & 540 & 18 & 0 & 18 & 8.52\% & 40994 \\
        cylinder-bands & 540 & 37 & 19 & 82 & 42.22\% & 6332 \\
        wdbc & 569 & 30 & 0 & 30 & 37.26\% & 1510 \\
        ilpd & 583 & 10 & 1 & 10 & 28.64\% & 1480 \\
        tokyo1 & 959 & 44 & 2 & 44 & 36.08\% & 40705 \\
        qsar-biodeg & 1,055 & 41 & 0 & 41 & 33.74\% & 1494 \\
        ozone-level-8hr & 2,534 & 72 & 0 & 72 & 6.31\% & 1487 \\
        madelon & 2,600 & 500 & 0 & 500 & 50.00\% & 1485 \\
        Bioresponse & 3,751 & 1,776 & 0 & 1,776 & 45.77\% & 4134 \\
        wilt & 4,839 & 5 & 0 & 5 & 5.39\% & 40983 \\
        churn & 5,000 & 20 & 4 & 22 & 14.14\% & 40701 \\
        phoneme & 5,404 & 5 & 0 & 5 & 29.35\% & 1489 \\
        SpeedDating & 8,378 & 120 & 61 & 241 & 16.47\% & 40536 \\
        PhishingWebsites & 11,055 & 30 & 30 & 46 & 44.31\% & 4534 \\
        Amazon\_employee\_access & 32,769 & 9 & 9 & 9 & 5.79\% & 4135 \\
        nomao & 34,465 & 118 & 29 & 172 & 28.56\% & 1486 \\
        adult & 48,842 & 14 & 8 & 37 & 23.93\% & 1590 \\
        numerai28.6 & 96,320 & 21 & 0 & 21 & 49.43\% & 23517 \\
    \bottomrule
    \end{tabular}
    }
    \label{tab:datasets}
\end{table}

\paragraph{Benchmark dataset selection}\label{A:datasets}
In this section, we move to a more complex benchmark, including larger and high-dimensional datasets where simple models are not able to achieve competitive results (see Table~\ref{tab:eval-results_binary_test_optimized}).
Therefore, we used a predefined collection of datasets that was selected based on objective criteria from OpenML Benchmark Suites %
and comprises a total of 19 binary classification datasets (see Table~\ref{tab:datasets} for details). The selection process was adopted from \citet{benchmarksuites} and therefore is not biased towards our method.
We decided against using the original CC18 benchmark, as the number of datasets ($72$) is extremely high and as reported by \citet{grinsztajn2022tree}, the selection process was not strict enough, as there are many simple datasets contained. Similarly, we decided not to use the tabular benchmark presented by \citet{grinsztajn2022tree}, as the datasets were adjusted to be extremely homogenous by removing all side-aspects (e.g., class imbalance, high-dimensionality, dataset size). We argue that this also removes some main challenges when dealing with tabular data.
As a result, we decided to use the benchmark proposed by \citet{benchmarksuites}\footnote{The notebook for dataset selection can be accessed under \url{https://github.com/openml/benchmark-suites/blob/master/OpenML\%20Benchmark\%20generator.ipynb} (Last accessed: March 3, 2025).}, which has a more strict selection process than CC18. The benchmark includes both originally, binary and multi-class tasks. For this evaluation, we focused on binary classification tasks for a clear evaluation protocol. Yet, our benchmark has a large overlap with CC18, as $16/19$ datasets are also contained in CC18. The overlap with \citet{grinsztajn2022tree} in contrast is rather small. This is mainly caused by the fact that most datasets in their tabular benchmark are binarized versions of multi-class or regression datasets, which was not allowed during the selection of our benchmark. %
Table~\ref{tab:datasets} lists the used datasets, along with relevant statistics and the source based on the OpenML-ID.\\

\paragraph{Preprocessing} 
We one-hot encoded low-cardinality categorical features and used leave-one-out encoding for high-cardinality categorical features (more than 10 categories). %
To make them suitable for a gradient-based optimization, we gaussianized features using a quantile transformation, as it is common practice~\citep{grinsztajn2022tree}.
In line with \citet{borisov2022deep}, we report the mean and standard deviation of the test performance over a 5-fold cross-validation to ensure reliable results while accounting for the long runtimes of some methods.

\begingroup
\begin{wraptable}{r}{0.6\textwidth}
\vspace{-\intextsep} %
\begin{minipage}{0.6\textwidth}
    \centering
   
        \caption[Categorization of Approaches]{\textbf{Categorization of Approaches.}}
            \small
            \begin{tabular}{|c|c|c|}        
            \hline
             & \textbf{Standard DTs} & \textbf{Oblivious DTs} \\
            \hline
            \textbf{Tree-based} & XGBoost & CatBoost \\
            \hline
            \textbf{Gradient-based} & GRANDE & NODE \\
            \hline
            \end{tabular}

        \label{tab:approaches}
        
\end{minipage}
\end{wraptable}
\paragraph{Methods}
In our main experiments, we compare GRANDE to XGBoost and CatBoost, which achieved superior results according to recent studies, and NODE, which is most related to our approach. With this setup, we have one state-of-the-art tree-based and one gradient-based approach for each tree type (see Table~\ref{tab:approaches}). In addition, we provide an extended evaluation including SAINT, RandomForest and ExtraTree as additional benchmarks in Appendix~\ref{A:additional_results}. These additional results are in line with the results presented in the following. %

\begin{table}[t]
\centering
\small
\caption[Ensembling Performance Comparison]{\textbf{Ensembling Performance Comparison.} This table compares the base learners GradTree and CART along with their basic ensembling counterparts GRANDE and RandomForest respectively. We report the test macro F1-score (mean $\pm$ stdev over 10 trials) with default parameters and the ranking of each approach in parentheses. The datasets are sorted based on the data size.}
\resizebox{\columnwidth}{!}{
\begin{tabular}{lrrrr}
\toprule
  & GradTree & GRANDE & CART & RandomForest \\
\midrule
dresses-sales                    & 0.528 $\pm$ .037 (4) & \bftab 0.596 $\pm$ .014 (1) & 0.550 $\pm$ .039 (3) &        0.576 $\pm$ .037 (2) \\
climate-simulation-crashes       & 0.540 $\pm$ .059 (4) & \bftab 0.758 $\pm$ .065 (1) & 0.705 $\pm$ .050 (2) &        0.558 $\pm$ .040 (3) \\
cylinder-bands                   & 0.697 $\pm$ .051 (4) & \bftab 0.813 $\pm$ .023 (1) & 0.713 $\pm$ .034 (3) &        0.794 $\pm$ .033 (2) \\
wdbc                             & 0.925 $\pm$ .033 (3) & \bftab 0.962 $\pm$ .008 (1) & 0.922 $\pm$ .033 (4) &        0.958 $\pm$ .021 (2) \\
ilpd                             & 0.613 $\pm$ .073 (2) & \bftab 0.646 $\pm$ .021 (1) & 0.568 $\pm$ .042 (4) &        0.584 $\pm$ .018 (3)\\
tokyo1                           & 0.917 $\pm$ .010 (3) & \bftab 0.922 $\pm$ .014 (1) & 0.908 $\pm$ .009 (4) &        0.921 $\pm$ .017 (2) \\
qsar-biodeg                      & 0.784 $\pm$ .027 (4) &        0.851 $\pm$ .032 (2) & 0.788 $\pm$ .012 (3) & \bftab 0.856 $\pm$ .022 (1) \\
ozone-level-8hr                  & 0.595 $\pm$ .014 (3) & \bftab 0.735 $\pm$ .011 (1) & 0.616 $\pm$ .021 (2) &        0.547 $\pm$ .019 (4) \\
madelon                          & 0.654 $\pm$ .046 (4) & \bftab 0.768 $\pm$ .022 (1) & 0.721 $\pm$ .018 (2) &        0.690 $\pm$ .011 (3) \\
Bioresponse                      & 0.710 $\pm$ .035 (4) &        0.789 $\pm$ .014 (2) & 0.721 $\pm$ .011 (3) & \bftab 0.799 $\pm$ .006 (1) \\
wilt                             & 0.876 $\pm$ .029 (4) & \bftab 0.933 $\pm$ .021 (1) & 0.888 $\pm$ .027 (2) &        0.887 $\pm$ .033 (3) \\
churn                            & 0.845 $\pm$ .016 (3) & \bftab 0.896 $\pm$ .007 (1) & 0.836 $\pm$ .006 (4) &        0.873 $\pm$ .019 (2) \\
phoneme                          & 0.792 $\pm$ .016 (4) &        0.860 $\pm$ .008 (2) & 0.834 $\pm$ .009 (3) & \bftab 0.887 $\pm$ .011 (1) \\
SpeedDating                      & 0.612 $\pm$ .038 (4) & \bftab 0.725 $\pm$ .007 (1) & 0.628 $\pm$ .016 (2) &        0.617 $\pm$ .016 (3) \\
PhishingWebsites                 & 0.923 $\pm$ .008 (4) &        0.969 $\pm$ .006 (2) & 0.962 $\pm$ .005 (3) & \bftab 0.971 $\pm$ .006 (1) \\
Amazon\_employee\_access         & 0.591 $\pm$ .026 (4) &        0.602 $\pm$ .006 (3) & 0.665 $\pm$ .010 (2) & \bftab 0.684 $\pm$ .010 (1) \\
nomao                            & 0.905 $\pm$ .005 (4) &        0.955 $\pm$ .004 (2) & 0.938 $\pm$ .004 (3) & \bftab 0.962 $\pm$ .002 (1) \\
adult                            & 0.760 $\pm$ .005 (3) &        0.785 $\pm$ .008 (2) & 0.753 $\pm$ .004 (4) & \bftab 0.793 $\pm$ .003 (1) \\
numerai28.6                      & 0.505 $\pm$ .004 (2) &        0.503 $\pm$ .003 (4) & 0.505 $\pm$ .004 (2) & \bftab 0.513 $\pm$ .002 (1) \\\midrule
Mean                             & 0.7249 & \bftab 0.7931 & 0.7486 & 0.7616 \\
\bottomrule
\end{tabular}
}
\label{tab:gradtree_grande_comparison}
\end{table}

\paragraph{Hyperparameters}
We optimized the hyperparameters using Optuna~\citep{akiba2019optuna} with 250 trials and selected the search space as well as the default parameters for related work in accordance with \citet{borisov2022deep}. The best parameters were selected based on a 5×2 cross-validation as suggested by \citet{raschka2018modelevaluation}, where the test data of each fold was held out of the HPO to get unbiased results. To deal with class imbalance, we further included class weights. Additional information along with the hyperparameters for each approach are in Appendix~\ref{A:hyperparameters}.

\subsection{Results} \label{sec:eval_grande}

\paragraph{Extending GradTree to GRANDE yields a substantial performance increase}
GradTree struggles to perform competitive to CART on complex, often high-dimensional datasets with a mean performance of $0.7249$ for GradTree compared to $0.7486$ for CART, as shown in Table~\ref{tab:gradtree_grande_comparison}. %
However, when ensembling is applied, transitioning from GradTree to GRANDE, the performance improves substantially. On average, GRANDE achieves a $0.0681$ higher performance than GradTree (Table~\ref{tab:gradtree_grande_comparison_difference}). This improvement is much larger than the $0.0129$ gain seen when moving from CART to RandomForest. This highlights the strong impact of ensembling on gradient-based DTs. With ensembling, GRANDE not only surpasses GradTree but also achieves a substantially higher performance compared to RandomForest and even outperforms XGBoost and CatBoost as we will show in the following (Table~\ref{tab:eval-results_binary_test_optimized} and Table~\ref{tab:eval-results_binary_test_default}). These results demonstrate the potential of gradient-based methods to handle complex datasets effectively when combined with ensembling techniques to mitigate the impact of instability and randomness during the training.

\begin{table}[t]
\centering
\small
\caption[Optimized Parameter Performance Comparison]{\textbf{Optimized Parameter Performance Comparison.} We report the test macro F1-score (mean $\pm$ stdev for a 5-fold CV) with optimized parameters. The datasets are sorted based on the data size.}
\resizebox{\columnwidth}{!}{
\begin{tabular}{lcccc}
\toprule
{} &                        \multicolumn{1}{l}{GRANDE} &                           \multicolumn{1}{l}{XGB} &                      \multicolumn{1}{l}{CatBoost} &                          \multicolumn{1}{l}{NODE} \\
\midrule
dresses-sales                    &  \bftab 0.612 $\pm$ .049 (1) &         0.581 $\pm$ .059 (3) &         0.588 $\pm$ .036 (2) &         0.564 $\pm$ .051 (4) \\
climate-simulation-crashes &  \bftab 0.853 $\pm$ .070 (1) &         0.763 $\pm$ .064 (4) &         0.778 $\pm$ .050 (3) &         0.802 $\pm$ .035 (2) \\
cylinder-bands                   &  \bftab 0.819 $\pm$ .032 (1) &         0.773 $\pm$ .042 (3) &         0.801 $\pm$ .043 (2) &         0.754 $\pm$ .040 (4) \\
wdbc                             &  \bftab 0.975 $\pm$ .010 (1) &         0.953 $\pm$ .030 (4) &         0.963 $\pm$ .023 (3) &         0.966 $\pm$ .016 (2) \\
ilpd                             &  \bftab 0.657 $\pm$ .042 (1) &         0.632 $\pm$ .043 (3) &         0.643 $\pm$ .053 (2) &         0.526 $\pm$ .069 (4) \\
tokyo1                           &         0.921 $\pm$ .004 (3) &         0.915 $\pm$ .011 (4) &  \bftab 0.927 $\pm$ .013 (1) &         0.921 $\pm$ .010 (2) \\
qsar-biodeg                      &  \bftab 0.854 $\pm$ .022 (1) &         0.853 $\pm$ .020 (2) &         0.844 $\pm$ .023 (3) &         0.836 $\pm$ .028 (4) \\
ozone-level-8hr                  &  \bftab 0.726 $\pm$ .020 (1) &         0.688 $\pm$ .021 (4) &         0.721 $\pm$ .027 (2) &         0.703 $\pm$ .029 (3) \\
madelon                          &         0.803 $\pm$ .010 (3) &         0.833 $\pm$ .018 (2) &  \bftab 0.861 $\pm$ .012 (1) &         0.571 $\pm$ .022 (4) \\
Bioresponse                      &         0.794 $\pm$ .008 (3) &         0.799 $\pm$ .011 (2) &  \bftab 0.801 $\pm$ .014 (1) &         0.780 $\pm$ .011 (4) \\
wilt                             &         0.936 $\pm$ .015 (2) &         0.911 $\pm$ .010 (4) &         0.919 $\pm$ .007 (3) &  \bftab 0.937 $\pm$ .017 (1) \\
churn                            &         0.914 $\pm$ .017 (2) &         0.900 $\pm$ .017 (3) &         0.869 $\pm$ .021 (4) &  \bftab 0.930 $\pm$ .011 (1) \\
phoneme                          &         0.846 $\pm$ .008 (4) &         0.872 $\pm$ .007 (2) &  \bftab 0.876 $\pm$ .005 (1) &         0.862 $\pm$ .013 (3) \\
SpeedDating                      &  \bftab 0.723 $\pm$ .013 (1) &         0.704 $\pm$ .015 (4) &         0.718 $\pm$ .014 (2) &         0.707 $\pm$ .015 (3) \\
PhishingWebsites                 &  \bftab 0.969 $\pm$ .006 (1) &         0.968 $\pm$ .006 (2) &         0.965 $\pm$ .003 (4) &         0.968 $\pm$ .006 (3) \\
Amazon\_employee\_access         &         0.665 $\pm$ .009 (2) &         0.621 $\pm$ .008 (4) &  \bftab 0.671 $\pm$ .011 (1) &         0.649 $\pm$ .009 (3) \\
nomao                            &         0.958 $\pm$ .002 (3) &  \bftab 0.965 $\pm$ .003 (1) &         0.964 $\pm$ .002 (2) &         0.956 $\pm$ .001 (4) \\
adult                            &         0.790 $\pm$ .006 (4) &  \bftab 0.798 $\pm$ .004 (1) &         0.796 $\pm$ .004 (2) &         0.794 $\pm$ .004 (3) \\
numerai28.6                      &  \bftab 0.519 $\pm$ .003 (1) &         0.518 $\pm$ .001 (3) &         0.519 $\pm$ .002 (2) &         0.503 $\pm$ .010 (4) \\
\midrule
Normalized Mean $\uparrow$ &  \bftab 0.776 (1) &         0.483 (3) &         0.671 (2) &         0.327 (4) \\
MRR $\uparrow$                &  \bftab 0.702  (1) &         0.417 (3) &         0.570  (2) &         0.395  (4) \\
\bottomrule
\end{tabular}
}
\label{tab:eval-results_binary_test_optimized}
\end{table}

\paragraph{GRANDE outperforms existing methods on most datasets} 
We evaluated the performance with optimized hyperparameters based on the macro F1-Score in Table~\ref{tab:eval-results_binary_test_optimized} to account for class imbalance. Additionally, we report the accuracy and ROC-AUC score in the Appendix~\ref{A:additional_results}, which are consistent with the results presented in the following.
GRANDE outperformed existing methods and achieved the highest mean reciprocal rank (MRR) of 0.702 and the highest normalized mean of 0.776. CatBoost yielded the second-best results (MRR of 0.570 and normalized mean of 0.671) followed by XGBoost (MRR of 0.417 and normalized mean of 0.483) and NODE (MRR of 0.395 and normalized mean of 0.327). Yet, our findings are in line with existing work, indicating that there is no universal method for tabular data.
However, on several datasets such as \emph{climate-simulation-crashes} and \emph{cylinder-bands} the performance difference to other methods was substantial, which highlights the importance of GRANDE as an extension to the existing repertoire. Furthermore, as the datasets are sorted by their size, we can observe that the results of GRANDE are especially good for small datasets, which is an interesting research direction for future work.

\paragraph{GRANDE is computationally efficient for large and high-dimensional datasets}
GRANDE averaged 47 seconds across all datasets, with a maximum runtime of 107 seconds. Thereby, the runtime of GRANDE is robust to high-dimensional (37 seconds for \emph{Bioresponse} with 1,776 features) and larger datasets (39 seconds for \emph{numerai28.6} with 96,320 samples). GRANDE achieved a substantially lower runtime compared to our gradient-based benchmark NODE, which has an approximately three times higher average runtime of 130 seconds. However, it is important to note that GBDT frameworks, especially XGBoost, are highly efficient when executed on the GPU and achieve substantially lower runtimes compared to gradient-based methods. The complete runtimes are listed in the appendix (Table~\ref{tab:eval-results_RUNTIME}).

\begin{table}[t]
\centering
\small
\caption[Default Parameter Performance Comparison]{\textbf{Default Parameter Performance Comparison.} We report the test macro F1-score (mean $\pm$ stdev over 10 trials) with default parameters and the ranking of each approach in parentheses. The datasets are sorted based on the data size.}
\resizebox{\columnwidth}{!}{
\begin{tabular}{lcccc}
\toprule
{} &                        \multicolumn{1}{l}{GRANDE} &                           \multicolumn{1}{l}{XGB} &                      \multicolumn{1}{l}{CatBoost} &                          \multicolumn{1}{l}{NODE} \\
\midrule
dresses-sales                    &  \bftab 0.596 $\pm$ .014 (1) &         0.570 $\pm$ .056 (3) &         0.573 $\pm$ .031 (2) &         0.559 $\pm$ .045 (4) \\
climate-simulation-crashes &         0.758 $\pm$ .065 (4) &  \bftab 0.781 $\pm$ .060 (1) &         0.781 $\pm$ .050 (2) &         0.766 $\pm$ .088 (3) \\
cylinder-bands                   &  \bftab 0.813 $\pm$ .023 (1) &         0.770 $\pm$ .010 (3) &         0.795 $\pm$ .051 (2) &         0.696 $\pm$ .028 (4) \\
wdbc                             &         0.962 $\pm$ .008 (3) &  \bftab 0.966 $\pm$ .023 (1) &         0.955 $\pm$ .029 (4) &         0.964 $\pm$ .017 (2) \\
ilpd                             &  \bftab 0.646 $\pm$ .021 (1) &         0.629 $\pm$ .052 (3) &         0.643 $\pm$ .042 (2) &         0.501 $\pm$ .085 (4) \\
tokyo1                           &  \bftab 0.922 $\pm$ .014 (1) &         0.917 $\pm$ .016 (4) &         0.917 $\pm$ .013 (3) &         0.921 $\pm$ .011 (2) \\
qsar-biodeg                      &  \bftab 0.851 $\pm$ .032 (1) &         0.844 $\pm$ .021 (2) &         0.843 $\pm$ .017 (3) &         0.838 $\pm$ .027 (4) \\
ozone-level-8hr                  &  \bftab 0.735 $\pm$ .011 (1) &         0.686 $\pm$ .034 (3) &         0.702 $\pm$ .029 (2) &         0.662 $\pm$ .019 (4) \\
madelon                          &         0.768 $\pm$ .022 (3) &         0.811 $\pm$ .016 (2) &  \bftab 0.851 $\pm$ .015 (1) &         0.650 $\pm$ .017 (4) \\
Bioresponse                      &         0.789 $\pm$ .014 (2) &         0.789 $\pm$ .013 (3) &  \bftab 0.792 $\pm$ .004 (1) &         0.786 $\pm$ .010 (4) \\
wilt                             &  \bftab 0.933 $\pm$ .021 (1) &         0.903 $\pm$ .011 (3) &         0.898 $\pm$ .011 (4) &         0.904 $\pm$ .026 (2) \\
churn                            &         0.896 $\pm$ .007 (3) &         0.897 $\pm$ .022 (2) &         0.862 $\pm$ .015 (4) &  \bftab 0.925 $\pm$ .025 (1) \\
phoneme                          &         0.860 $\pm$ .008 (3) &  \bftab 0.864 $\pm$ .003 (1) &         0.861 $\pm$ .008 (2) &         0.842 $\pm$ .005 (4) \\
SpeedDating                      &  \bftab 0.725 $\pm$ .007 (1) &         0.686 $\pm$ .010 (4) &         0.693 $\pm$ .013 (3) &         0.703 $\pm$ .013 (2) \\
PhishingWebsites                 &  \bftab 0.969 $\pm$ .006 (1) &         0.969 $\pm$ .007 (2) &         0.963 $\pm$ .005 (3) &         0.961 $\pm$ .004 (4) \\
Amazon\_employee\_access         &         0.602 $\pm$ .006 (4) &         0.608 $\pm$ .016 (3) &  \bftab 0.652 $\pm$ .006 (1) &         0.621 $\pm$ .010 (2) \\
nomao                            &         0.955 $\pm$ .004 (3) &  \bftab 0.965 $\pm$ .003 (1) &         0.962 $\pm$ .003 (2) &         0.955 $\pm$ .002 (4) \\
adult                            &         0.785 $\pm$ .008 (4) &         0.796 $\pm$ .003 (2) &         0.796 $\pm$ .005 (3) &  \bftab 0.799 $\pm$ .003 (1) \\
numerai28.6                      &         0.503 $\pm$ .003 (4) &         0.516 $\pm$ .002 (2) &  \bftab 0.519 $\pm$ .001 (1) &         0.506 $\pm$ .009 (3) \\
\midrule
Normalized Mean $\uparrow$ & \bftab 0.637 (1) &         0.587 (3) &         0.579 (1) &  0.270  (4) \\
MRR $\uparrow$                   &  \bftab 0.640  (1) &         0.518 (3) &         0.522 (2) &         0.404 (4) \\
\bottomrule
\end{tabular}
}
\label{tab:eval-results_binary_test_default}
\end{table}

\paragraph{GRANDE achieves impressive results with default parameters} 
Many methods, especially DL methods, are heavily reliant on a proper hyperparameter optimization. Yet, it is a desirable property that a method achieves good results even with their default setting. GRANDE achieves superior results with default hyperparameters, and substantially outperforms existing methods on most datasets. More specifically, GRANDE has the highest normalized mean performance (0.6371) and the highest MRR (0.6404) as summarized in Table~\ref{tab:eval-results_binary_test_default}. %

\begin{table}[tb]
\centering
\caption[Optimizer Comparison for GRANDE]{\textbf{Optimizer Comparison for GRANDE.} In this table, we compare different optimizers for GRANDE. We report the test macro F1-score (mean $\pm$ stdev over 10 trials) with default parameters and the ranking in parentheses.}
\label{tab:optimizer_ablation}
\resizebox{\textwidth}{!}{
\begin{tabular}{lcccc}
\toprule
 & Adam + SWA & Adam & RMSprop & SGD \\
\midrule
dresses-sales                  & \bftab 0.596 $\pm$ .014 (1)   & 0.586 $\pm$ .038 (2)      & 0.507 $\pm$ .012 (3) & 0.457 $\pm$ .042 (4) \\
climate-simulation-crashes     & \bftab 0.758 $\pm$ .065 (1)   & 0.740 $\pm$ .068 (2)       & 0.528 $\pm$ .101 (3) & 0.406 $\pm$ .075 (4) \\
cylinder-bands                 & 0.813 $\pm$ .023 (2)          & \bftab 0.814 $\pm$ .034 (1) & 0.706 $\pm$ .038 (3) & 0.452 $\pm$ .071 (4) \\
wdbc                           & 0.962 $\pm$ .008 (2)          & \bftab 0.964 $\pm$ .016 (1) & 0.950 $\pm$ .027 (3) & 0.321 $\pm$ .096 (4) \\
ilpd                           & \bftab 0.646 $\pm$ .021 (1)   & 0.621 $\pm$ .056 (2)        & 0.518 $\pm$ .034 (3) & 0.438 $\pm$ .051 (4) \\
tokyo1                         & 0.922 $\pm$ .014 (2)          & \bftab 0.926 $\pm$ .014 (1) & 0.918 $\pm$ .009 (3) & 0.516 $\pm$ .122 (4) \\
qsar-biodeg                    & 0.851 $\pm$ .032 (2)          & \bftab 0.853 $\pm$ .028 (1) & 0.807 $\pm$ .027 (3) & 0.477 $\pm$ .044 (4) \\
ozone-level-8hr                & \bftab 0.735 $\pm$ .011 (1)   & 0.6949 $\pm$ .027 (2)       & 0.523 $\pm$ .054 (3) & 0.328 $\pm$ .053 (4) \\
madelon                        & 0.768 $\pm$ .022 (2)          & \bftab 0.776 $\pm$ .006 (1) & 0.563 $\pm$ .033 (3) & 0.483 $\pm$ .024 (4) \\
Bioresponse                    & \bftab 0.789 $\pm$ .014 (1)   & 0.784 $\pm$ .018 (2)        & 0.741 $\pm$ .016 (3) & 0.454 $\pm$ .086 (4) \\
wilt                           & \bftab 0.933 $\pm$ .021 (1)   & 0.929 $\pm$ .016 (2)        & 0.792 $\pm$ .029 (3) & 0.426 $\pm$ .033 (4) \\
churn                          & 0.896 $\pm$ .007 (2)          & \bftab 0.919 $\pm$ .014 (1) & 0.668 $\pm$ .101 (3) & 0.450 $\pm$ .036 (4) \\
phoneme                        & 0.860 $\pm$ .008  (2)         & \bftab 0.863 $\pm$ .006 (1) & 0.805 $\pm$ .007 (3) & 0.471 $\pm$ .074 (4) \\
SpeedDating                    & \bftab 0.725 $\pm$ .007 (1)   & 0.699 $\pm$ .018 (2)        & 0.662 $\pm$ .023 (3) & 0.398 $\pm$ .053 (4) \\
PhishingWebsites               & \bftab 0.969 $\pm$ .006 (1)   & 0.968 $\pm$ .006 (2)        & 0.936 $\pm$ .006 (3) & 0.573 $\pm$ .085 (4) \\
Amazon\_employee\_access       & 0.602 $\pm$ .006 (3)          & \bftab 0.669 $\pm$ .008 (1) & 0.650 $\pm$ .011 (2) & 0.500 $\pm$ .013 (4) \\
nomao                          & \bftab 0.955 $\pm$ .004   (1) & 0.954 $\pm$ .003  (2)       & 0.921 $\pm$ .022 (3) & 0.554 $\pm$ .101 (4) \\
adult                          & 0.785 $\pm$ .008  (2)         & \bftab 0.798 $\pm$ .006  (1) & 0.785 $\pm$ .008 (2) & 0.612 $\pm$ .103 (4) \\
numerai28.6                    & 0.503 $\pm$ .003 (3)          & 0.510 $\pm$ .005 (2) & \bftab 0.515 $\pm$ .002 (1) & 0.488 $\pm$ .014 (4) \\ \midrule
Mean                           & \bftab 0.7931              & 0.7929                    & 0.7103  & 0.4632 \\

\bottomrule
\end{tabular}
}
\end{table}

\paragraph{Advanced optimizers significantly impact performance}
In the background chapter (Section~\ref{sec:advancements_gradient_descent}), we introduced various advancements in gradient descent optimization, which have been successfully applied to both GradTree and GRANDE. To further investigate the impact of these individual advancements, we conducted an additional experiment using GRANDE as a practical example (Table~\ref{tab:optimizer_ablation}). Starting with Adam in combination with Stochastic Weight Averaging (SWA), we systematically ablate pure Adam (i.e., SGD with adaptive learning rates and momentum), RMSprop (i.e., SGD with adaptive learning rates), and plain SGD.
Our results indicate only a minor increase in performance when using SWA compared to plain Adam. However, for specific datasets, such as \emph{ozone-level-8hr} and \emph{SpeedDating}, SWA still shows a substantial performance boost.
Switching from RMSprop to Adam, effectively introducing momentum in addition to dynamic learning rates, yields a substantial performance increase of over $8\%$ on average. Momentum contributes to optimization by smoothing the updates, preventing oscillations, and enabling more consistent convergence. This is particularly important when gradients are sparse, a characteristic inherent to GRANDE due to its selective use of only a subset of parameters.
The impact of adaptive learning rates in RMSprop is even more pronounced, with performance improvements of approximately $25\%$ on average. This enhancement can also be attributed to the dense parameter representation in GRANDE, where only a subset of parameters is actively utilized. Adaptive learning rates enable the model to rapidly adjust critical parameters of higher relevance while permitting subtle changes to more sensitive parameters, thereby enhancing efficiency and performance.
This emphasizes the importance of advanced optimization techniques and highlights the potential for gradient-based DTs to benefit from ongoing research advancements.

\begin{table}[t]
\centering
\small
\caption[Ablation Study Summary]{\textbf{Ablation Study Summary.} Left: Comparison of differentiable split functions. Right: Comparison of our instance-wise weighting with a global weight for each estimator. Complete results in Table~\ref{tab:ablation_study_activation} and Table~\ref{tab:ablation_study_weighting}.}
\resizebox{0.9\columnwidth}{!}{
\begin{tabular}{lrrrrr}
\toprule
& \multicolumn{3}{l}{Differentiable Split Function} & \multicolumn{2}{l}{Weighting Technique} \\ \cmidrule(lr){2-4}  \cmidrule(lr){5-6}
 & \multicolumn{1}{L{1.4cm}}{Softsign} & \multicolumn{1}{L{1.4cm}}{Entmoid} & \multicolumn{1}{L{1.4cm}}{Sigmoid} & \multicolumn{1}{L{1.7cm}}{Leaf Weights} & \multicolumn{1}{L{2.35cm}}{Estimator Weights} \\ 
\midrule
Normalized Mean $\uparrow$ & \bftab 0.7906 (1) & 0.4188 (2) & 0.2207 (3) & \bftab 0.8235 (1) & 0.1765 (2) \\ 
MRR $\uparrow$ & \bftab 0.8246 (1) & 0.5526 (2) & 0.4561 (3) & \bftab 0.9211 (1) & 0.5789 (2)  \\ 
\bottomrule
\end{tabular}
}
\label{tab:ablation_study_summary}
\end{table}

\paragraph{Softsign improves performance}
As discussed in Section~\ref{ssec:splitting_function}, we argue that employing softsign as split index activation propagates informative gradients beneficial for the optimization. In Table~\ref{tab:ablation_study_summary} we support these claims by showing a superior performance of GRANDE with a softsign activation (before discretizing with the ST operator) compared to sigmoid as the default choice as well as an entmoid function which is commonly used in related work~\citep{popov2019neural,node_gam}. Interestingly, we observed during preliminary experiments that employing softsign over sigmoid only yields a consistent performance increase in the context of tree ensembles (i.e., for GRANDE), while there is no significant difference for the case of individual trees (i.e., GradTree).
We suggest that this effect occurs because the significant changes in split values caused by using the softsign function instead of sigmoid can result in notable alterations to the tree structure, potentially leading to instability during the training of individual trees. However, in ensembles, this effect is less pronounced, allowing for faster convergence and greater diversity when the softsign function is used for splitting.

\paragraph{Instance-wise weighting increases model performance}
GRANDE uses instance-wise weighting to assign varying weights to estimators for each sample based on selected leaves. This promotes ensemble diversity and encourages estimators to capture unique local interactions. We argue that the ability to learn and represent simple, local rules with individual estimators in our ensemble can have a positive impact on the overall performance (see Table~\ref{tab:ablation_study_summary}) as it simplifies the task that has to be solved by the remaining estimators. As a result, GRANDE can efficiently learn compact representations for simple rules, where complex models usually tend to learn overly complex representations, as we show in the following case study. 

\subsection{Case Study: PhishingWebsites Dataset}
In the following case study, we demonstrate the ability of GRANDE to learn compact representations for simple rules within a complex ensemble:
The \emph{PhishingWebsites} dataset is concerned with identifying malicious websites based on metadata and additional observable characteristics. Although the task is challenging (i.e., it is not possible to solve it sufficiently well with a simple model, as shown in Table~\ref{tab:case_study_performance}), there exist several clear indicators for phishing websites.
Thus, some instances can be categorized using simple rules, while assigning other instances is more difficult. 
Ideally, if an instance can be easily categorized, the model should follow simple rules to make a prediction.
\begingroup
\begin{wrapfigure}{r}{0.45\textwidth}
\begin{minipage}{0.45\textwidth}
\small
    \centering 
   \centering
  \includegraphics[width=0.85\textwidth]{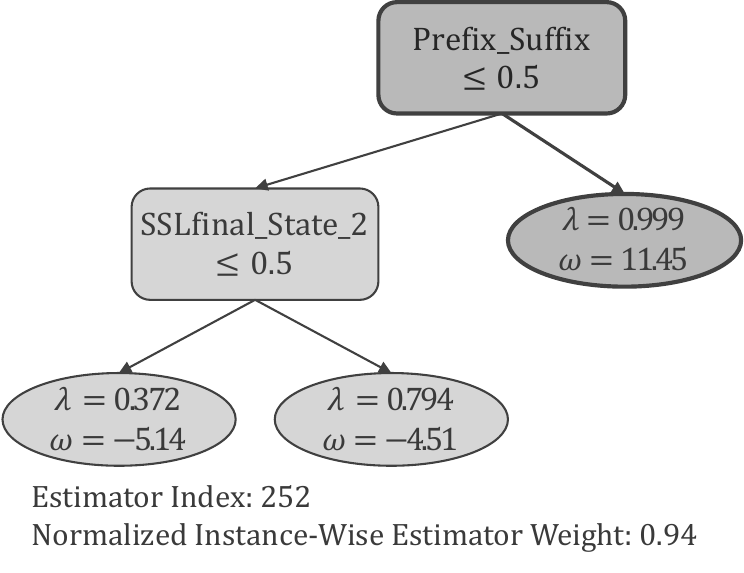}
  \caption[Highest-Weighted Estimator]{\textbf{Highest-Weighted Estimator.} This figure shows the DT from GRANDE (1024 estimators) with the highest weight for an example instance.}
  \label{fig:tree_example}   
\end{minipage}
\end{wrapfigure}
\noindent One example of a rule, which holds universally in the given dataset, is that an instance can be classified as \emph{phishing} if a prefix or suffix was added to the domain name. %
By assessing the weights for an exemplary instance fulfilling this rule, we can observe that the DT visualized in Figure~\ref{fig:tree_example} accounts for 
94\% of the prediction. %
Accordingly, GRANDE has learned a very simple representation and the classification is derived by applying an easily comprehensible rule.
Notably, for the other methods, it is not possible to assess the importance of individual estimators out-of-the-box similarly, as the prediction is either derived by either sequentially summing up the predictions (e.g., XGBoost and CatBoost) or equally weighting all estimators.
Furthermore, this has a strong positive impact on the average performance of GRANDE compared to using one single weight for each estimator (see Table~\ref{tab:ablation_study_summary}). 

\begin{figure}[tb]
    \centering 
        \includegraphics[width=\columnwidth]{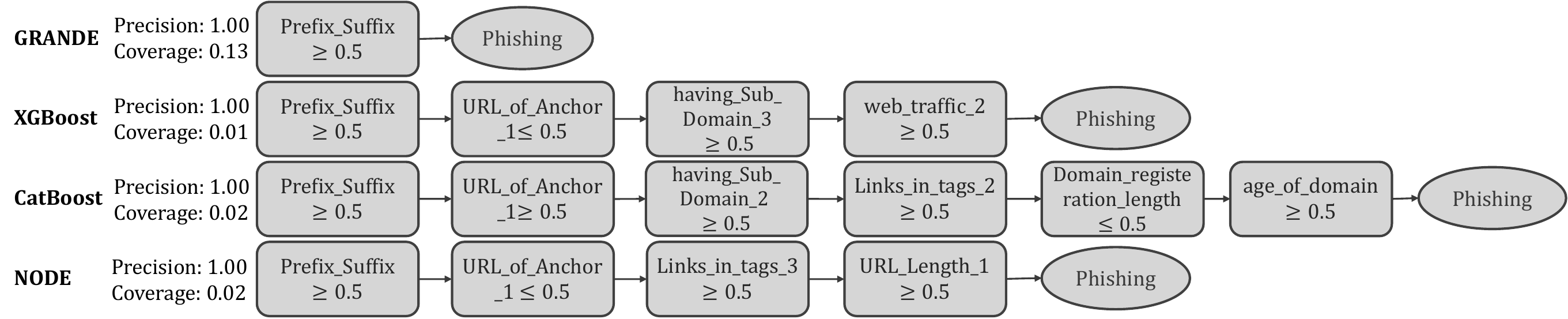}       
        \caption[Anchors Explanations]{\textbf{Anchors Explanations.} This figure shows the local explanations generated by Anchors for the given instance. The explanation for GRANDE only comprises a single rule. %
        In contrast, the corresponding explanations for the other methods have substantially higher complexity, which indicates that these methods are not able to learn simple representations within a complex model.
        }
        \label{fig:anchors}       
\end{figure}
\paragraph{Instance-wise weighting can be beneficial for local interpretability} 
In addition to the performance increase, our instance-wise weighting has a notable impact on the local interpretability of GRANDE. For each instance, we can assess the weights of individual estimators and inspect the estimators with the highest importance to understand which rules have the greatest impact on the prediction. %
For the given example, we only need to observe a single tree of depth two (Figure~\ref{fig:tree_example}) to understand why the given instance was classified as \emph{phishing}, even though the complete model is very complex. 
In contrast, existing ensemble methods require a global interpretation of the model and do not provide simple, local explanations out-of-the-box.
However, similar explanations can be extracted using Anchors~\citep{ribeiro2018anchors}. Anchors, as an extension to LIME~\citep{lime}, provides model-agnostic explanations by identifying conditions (called \emph{anchors}) which, when satisfied, guarantee a certain prediction with a high probability (noted as precision). These anchors are interpretable, rules-based conditions derived from input features that consistently lead to the same model prediction. Figure~\ref{fig:anchors} shows the extracted rules for each approach. We can clearly see that the anchor extracted for GRANDE matches the rule we have identified based on the instance-wise weights in Figure~\ref{fig:tree_example}. Furthermore, it is evident that the prediction derived by GRANDE is much simpler compared to any other approach, as it only comprises a single rule. Notably, this comes without suffering a loss in the precision, which is 1.00 for all methods. Furthermore, the rule learned by GRANDE has a substantially higher coverage, which means that the rule applied by GRANDE is more broadly representative.
Furthermore, we provide a detailed discussion and analysis of the weighting statistics in Appendix~\ref{A:weighting} and show that local experts can be identified on most datasets.

\subsection{Summary}
In this section, we evaluated GRANDE, a novel gradient-based approach for learning axis-aligned DT ensembles which combines the inductive bias of axis-aligned splits with the flexibility of gradient descent optimization. Thereby, GRANDE addresses key challenges in tabular data modeling, including handling heterogeneous data and learning from complex relationships. 
Through a comprehensive evaluation on a rigorously selected benchmark of 19 binary classification datasets, we demonstrated that GRANDE outperforms state-of-the-art methods, including gradient-boosting frameworks like XGBoost and CatBoost, and deep learning models like NODE. Importantly, GRANDE achieves high performance with both optimized and default hyperparameters, indicating its robustness and practical applicability. 
We also introduced a novel instance-wise weighting mechanism, which allows GRANDE to balance learning simple, local rules alongside complex global patterns. This capability not only improves predictive performance but also enhances local interpretability, as illustrated in a detailed case study. By assigning varying weights to individual estimators based on input characteristics, GRANDE facilitates the identification of localized decision-making processes within a complex ensemble.
In the next section, we show GRANDE’s adaptability by integrating it into a multimodal learning pipeline. As a gradient-based framework, GRANDE seamlessly complements other modalities, such as image data, while retaining the inductive bias required for tabular tasks.

\section{Multimodal Learning} \label{sec:multimodal}
As motivated earlier, the gradient-based nature of GRANDE allows a flawless integration into deep learning workflows, for instance in a multimodal learning setup, where tabular data is one of several input types~\citep{multimodal1,multimodal2,multimodal3,multimodal4}. Thereby, GRANDE can serve as a tree-based backbone and boost performance as well as interpretability. In the following, we will show that integrating GRANDE into a basic multimodal learning architecture immediately results in a performance boost compared to commonly used DL methods for tabular data. 

\begin{figure}[tb]
\centering
\begin{subfigure}{\textwidth}
   \centering
   \includegraphics[width=0.95\columnwidth]{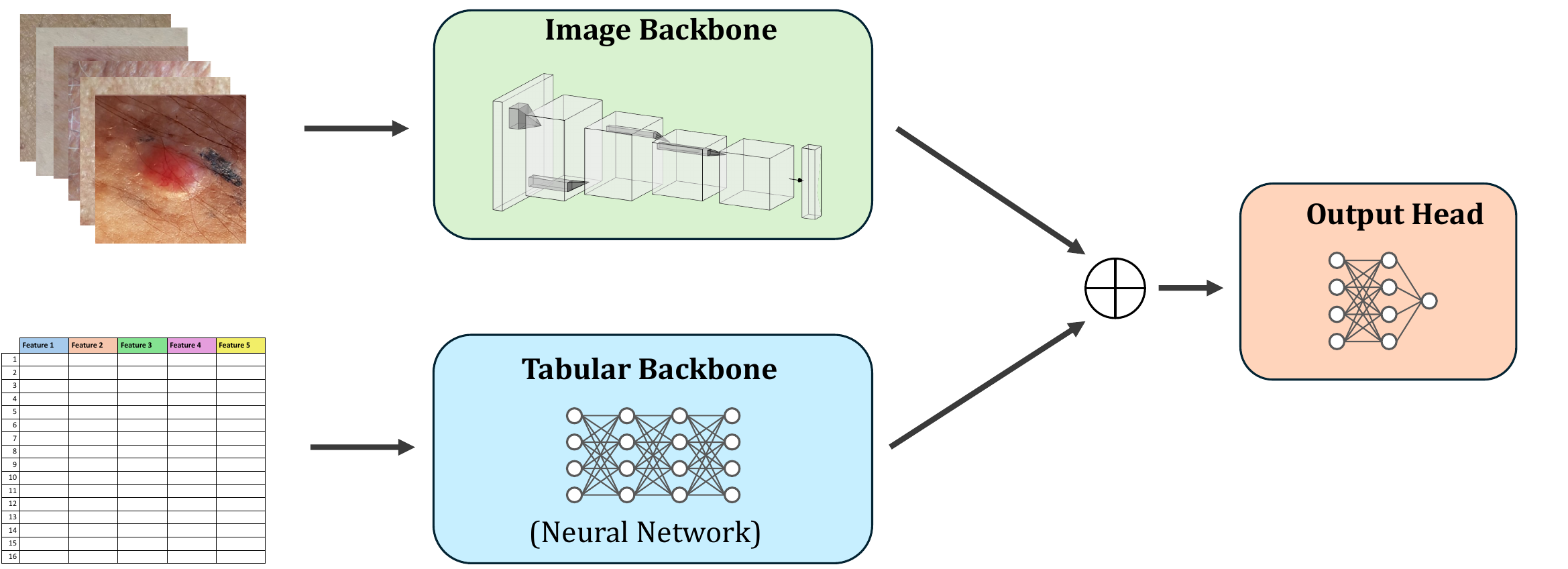}
   \caption{Basic Multimodal Architecture with Intermediate Fusion}
   \label{fig:multimodal_base}
\end{subfigure}

\vspace{0.5cm}

\begin{subfigure}{\textwidth}
   \centering
   \includegraphics[width=0.95\columnwidth]{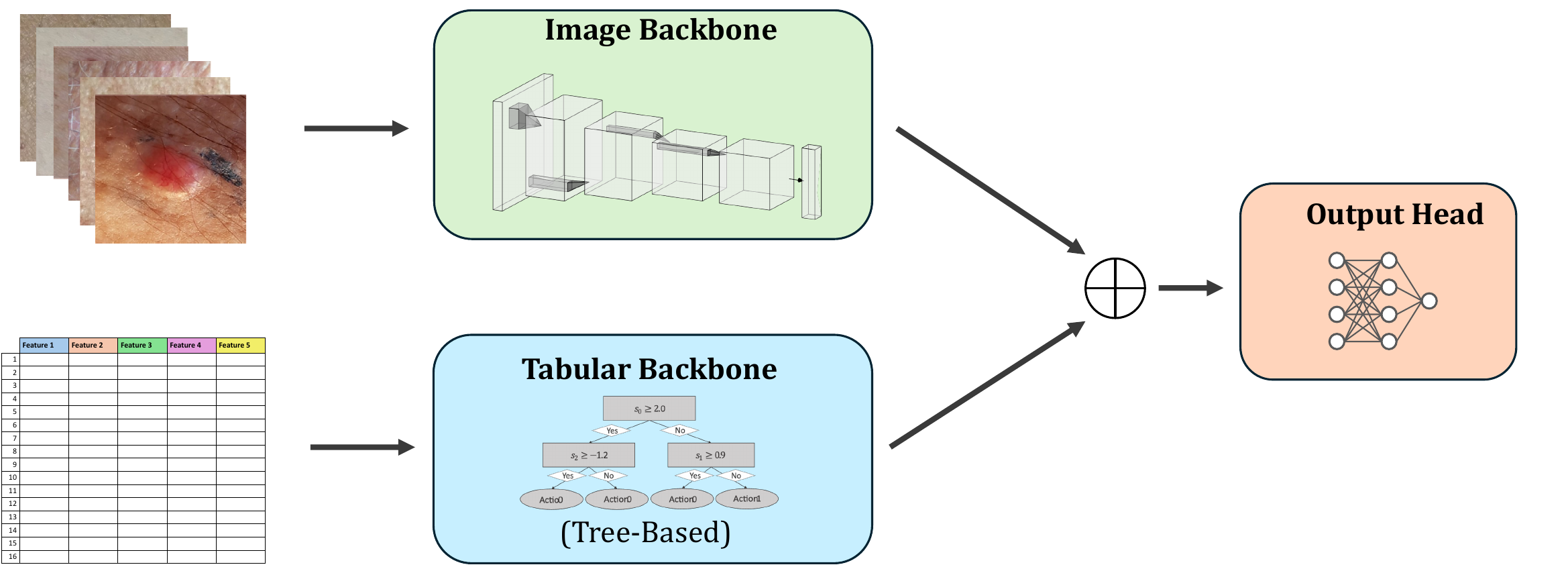}
   \caption{GRANDE Multimodal Architecture with Intermediate Fusion}
   \label{fig:multimodal_grande}
\end{subfigure}

\caption[Standard Multimodal Pipeline vs. GRANDE Multimodal Pipeline]{\textbf{Standard Multimodal Pipeline vs. GRANDE Multimodal Pipeline.} This figure shows how GRANDE can seamlessly be integrated into a multimodal learning setup. Therefore, we simply replace a neural network (A) which is commonly used as tabular backbone with GRANDE (B). In contrast to alternative tree-based methods like XGBoost, we can still learn our model end-to-end with gradient descent.}
\label{fig:multimodal_overview}
\end{figure}

\subsection{Evaluation} \label{ssec:evaluation_multimodal}

Please note that the goal of this experiment is not to achieve state-of-the-art results for multimodal learning, but highlight the potential and flexibility of the proposed method. However, we believe that GRANDE can become a core component for state-of-the art multimodal learning architectures by integrating the inductive bias of axis-aligned splits for the tabular component, which is subject to future work.

\subsubsection{Experimental Setup}

\paragraph{Dataset} We use the PAD-UFES-20 skin lesion dataset introduced by \citet{PAD-UFES}. The PAD-UFES-20 dataset is a publicly available skin lesion dataset comprising 2,298 clinical images and related patient metadata collected from 1,373 patients. Sourced from smartphone devices, these images capture six types of skin lesions, including three skin cancers (Basal Cell Carcinoma, Squamous Cell Carcinoma, and Melanoma) and three non-cancerous skin conditions. Each image is accompanied by up to 21 clinical features, such as patient demographics, lesion location, and history of cancer, providing comprehensive metadata for research. This is also the main reason why we decided for this dataset as the comprehensive tabular component is relevant to achieve a good performance, making it well suited to evaluate our method in a multimodal learning context.

\paragraph{Model setup} Our architecture is summarized in Figure~\ref{fig:multimodal_overview}. For the Image Backbone, we used a ConvNeXt~\citep{liu2022convnet} model pretrained on ImageNet~\citep{deng2009imagenet}, where we only fine-tuned the last layer for prediction, as this yielded the best performance in preliminary experiments. For the tabular component, we used a standard MLP as baseline (Figure~\ref{fig:multimodal_base}) which is the most commonly used option in end-to-end multimodal setups. In addition, we included a tabular ResNet~\citep{ft_transformer} as a stronger benchmark. As visualized in Figure~\ref{fig:multimodal_grande} to employ GRANDE, we can just replace the tabular backbone and still train the model end-to-end. In addition to the multimodal setup, we report the performance of the individual components for a better comparability.

\paragraph{Hyperparameters} To ensure good results for each method, we performed a hyperparameter search for each method, including the individual components. Thereby, we optimized the learning rate and the size of the output head for 60 trials. Additionally, we tuned the individual architectures according to related work~\citep{ft_transformer,grinsztajn2022tree,borisov2022deep}.

\subsubsection{Results} \label{ssec:results_multimodal}
The results in Table~\ref{tab:multimodal} highlight the superior performance of GRANDE in both, the tabular only and multimodal case. For tabular data, GRANDE consistently outperforms both MLP and ResNet across all metrics, achieving the highest ROC (0.943), accuracy (0.792), balanced accuracy (0.675), and F1 score (0.690). Similarly, in the multimodal setting, where both image and tabular data are combined, the ConvNeXt + GRANDE model achieves the highest performance across all metrics (ROC of 0.956, accuracy of 0.823, balanced accuracy of 0.735, and F1 score of 0.747). In contrast, the ConvNeXt model alone shows weaker performance on image data (ROC of 0.870, accuracy of 0.670), similar to only using a tabular model, demonstrating the benefits of multimodal approaches. This again supports the flexibility and easy adoption of GRANDE for further tasks outside of standalone tabular data.

\begin{table}[t]
    \centering
    \caption[PAD-UFES Multi-Class Performance Comparison]{\textbf{PAD-UFES Multi-Class Performance Comparison.} We report the scores for multi-class skin lesion classification on the PAD-UFES dataset with image and tabular component. Pretrained Image Model}
    \label{tab:multimodal}
    \resizebox{0.9\columnwidth}{!}{
    \begin{tabular}{llrrrr}
    \toprule
           &  & ROC $\uparrow$ & ACC $\uparrow$ & Balanced ACC $\uparrow$ & F1 $\uparrow$ \\ \midrule
          \multirow{3}{*}{tabular}&MLP                                                       & 0.921 & 0.761 & 0.622 & 0.636  \\
          &ResNet                                                                            & 0.889 & 0.712 & 0.528 & 0.526  \\
          &GRANDE                                                                            & \bftab 0.943 & \bftab 0.792 & \bftab 0.675 & \bftab 0.690  \\ \midrule
          image &\begin{tabular}[c]{@{}l@{}}ConvNeXt\end{tabular}     & 0.870 & 0.670 & 0.518 & 0.534  \\ \midrule
  \multirow{3}{*}{multimodal} & ConvNeXt + MLP   & 0.940 & 0.791 & 0.689 & 0.703  \\
 & ConvNeXt + ResNet & 0.933 & 0.778 & 0.676 & 0.693  \\
  & ConvNeXt + GRANDE & \bftab 0.956 & \bftab 0.823 & \bftab 0.735 & \bftab 0.747 \\
  \bottomrule
  \end{tabular}
  }
\end{table}

\chapter[Tree-Based Reinforcement Learning via Direct Optimization]{Mitigating Information Loss in Tree-Based Reinforcement Learning via Direct Optimization} \label{cha:sympol}

\textbf{The following section was already partially published in \citet{marton2024sympol}. The implementation and experiments are publically available under \url{https://github.com/s-marton/SYMPOL}. For a detailed summary of the individual contributions, please refer to Section~\ref{sec:contributions}.} \\

\paragraph{Reinforcement learning lacks transparency}%
Reinforcement learning (RL) has achieved remarkable success in solving complex sequential decision-making problems, ranging from robotics and autonomous systems to game playing and recommendation systems. However, the policies learned by traditional RL algorithms, represented by neural networks, often lack interpretability and transparency, making them difficult to understand, trust, and deploy in safety-critical or high-stakes scenarios \citep{landajuela2021discovering}.

\paragraph{Symbolic policies increase trust}%
Symbolic policies, on the other hand, offer a promising alternative by representing decision-making strategies in terms of RL policies as compact and interpretable structures \citep{guo2023efficient}. These symbolic representations do not only facilitate human understanding and analysis but also ensure predictable and explainable behavior, which is crucial for building trust and enabling effective human-AI collaboration.
Moreover, the deployment of symbolic policies in high-stakes domains, such as autonomous vehicles or industrial robots, could significantly improve their reliability and trustworthiness. By providing human operators with a clear understanding of the decision-making process, symbolic policies can facilitate effective monitoring, intervention, and debugging, ultimately enhancing the safety and robustness of these systems. In this context, DTs are particularly effective as symbolic policies for RL, as their hierarchical structure provides natural interpretability.

\begin{figure}[t] %
    \centering
    \begin{subfigure}[b]{0.32\columnwidth}    
        \centering
        \includegraphics[width=\textwidth]{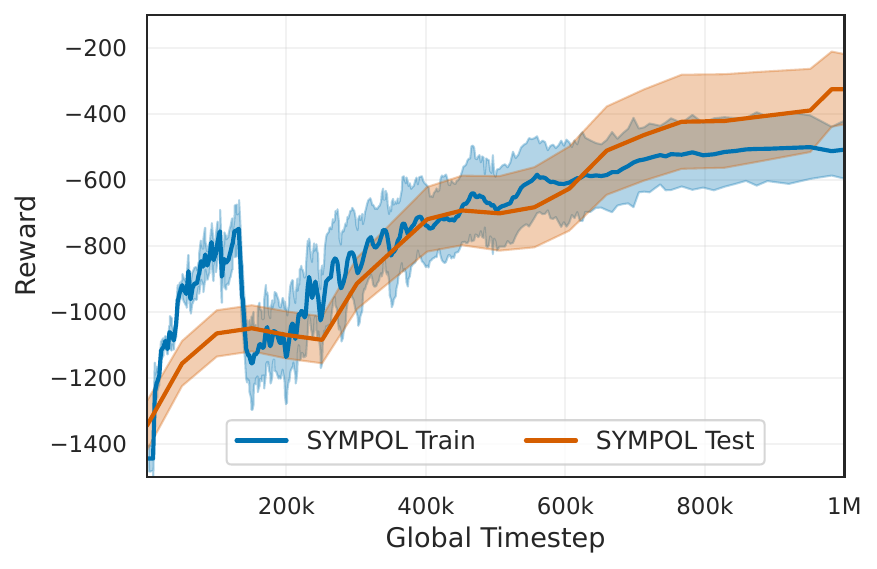}
        \caption{SYMPOL (ours)}
        \label{fig:sympol}
    \end{subfigure}
    \;
    \begin{subfigure}[b]{0.32\columnwidth}    
        \centering
        \includegraphics[width=\textwidth]{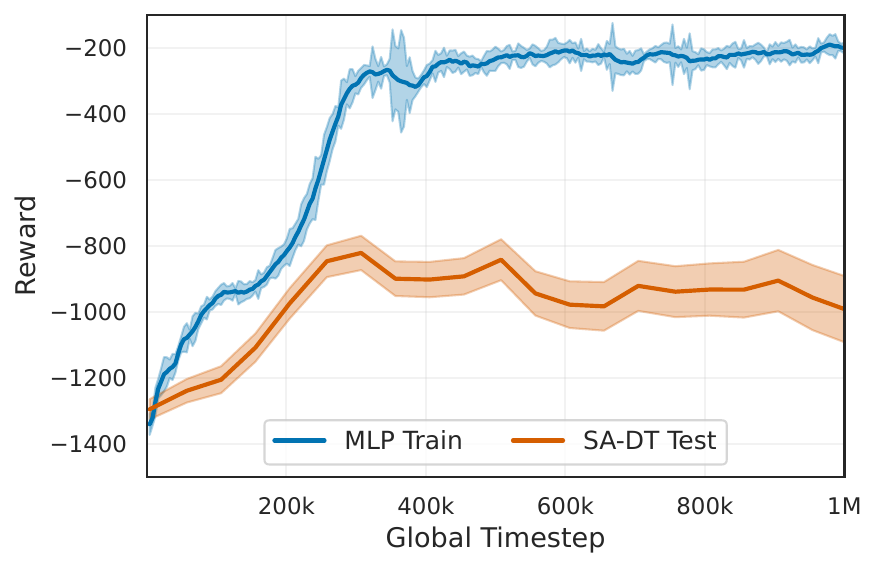}
        \caption{State-Action DT}
        \label{fig:sadt}
    \end{subfigure}
    \;
    \begin{subfigure}[b]{0.32\columnwidth}    
        \centering
        \includegraphics[width=\textwidth]{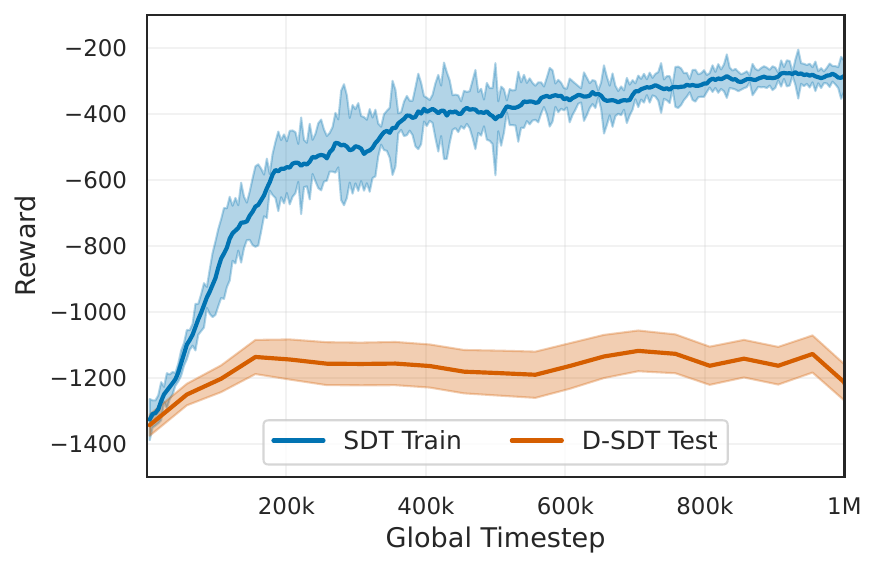}
        \caption{Discretized Soft DT}
        \label{fig:d-sdt}
    \end{subfigure}

    \caption[Information Loss in Tree-Based Reinforcement Learning on Pendulum]{\textbf{Information Loss in Tree-Based Reinforcement Learning on Pendulum.} Existing methods for symbolic, tree-based RL (see (B) and (C)) suffer from severe information loss when converting the differentiable policy used for training (e.g., the MLP for SA-DT) into the symbolic policy used for interpretation (i.e., the DT). Using SYMPOL (A), we can directly optimize the symbolic policy with PPO and therefore have no information loss during the application.}
    \label{fig:methods_comparison}
\end{figure}

\paragraph{Existing challenges}%
Despite these promising prospects, the field of symbolic RL faces several challenges. One main reason is given by the fact that many symbolic models, like DTs, are non-differentiable and cannot be integrated in existing RL approaches. 
Therefore, traditional methods for learning symbolic policies often rely on custom and complex training procedures 
\citep{costa2024evolving, vos2023optimal, kanamori2022counterfactual}, 
limiting their applicability and scalability. 
Alternative methods involve pre-trained neural network policies combined with some post-processing to obtain an interpretable model 
\citep{silva2020optimization, liu2019toward, liu2023effective, bastani2018viper}. 
However, post-processing introduces a mismatch between the optimized policy and the model obtained for interpretation, which can lead to loss of crucial information, as we show in Figure~\ref{fig:methods_comparison}.

\paragraph{A new method for a direct optimization of tree-based policies}%
To mitigate the impact of information loss\footnote{Information loss refers to the discrepancy between the trained and applied policy, which can arise when a model undergoes post-processing such as discretization or distillation. These modifications may introduce deviations from the original learned policy, potentially leading to suboptimal performance during application. This contrasts with information gain in the decision tree literature, which quantifies the reduction in uncertainty when splitting a node. While information gain measures how well a feature contributes to decision-making within a tree, information loss in this context highlights the degradation of policy fidelity e.g., due to some post-processing step.}, a direct optimization of the policy is crucial. 
Existing methods that employ a direct optimization of a tree-based policy, such as those described by \citet{silva2020optimization} learn differentiable, soft DTs, which do not provide a high level of interpretability. To obtain interpretable, axis-aligned DTs, these methods require post-hoc distillation or discretization and therefore suffer from information loss (see Figure~\ref{fig:methods_comparison}). 
In this chapter, we introduce SYMPOL, \textbf{SYM}bolic tree-based on-\textbf{POL}icy RL, a novel method for efficiently learning interpretable, axis-aligned DT policies end-to-end. SYMPOL does not depend on pre-trained NN policies, complex search procedures, or post-processing steps, but can be seamlessly integrated into existing RL algorithms (Section~\ref{sec:sympol}). \\

We start this section with an overview of the RL background (Section~\ref{sec:background_rl}), followed by introducing the relevant related work for our proposed method (Section~\ref{sec:related_rl}). Then, we introduce SYMPOL in Section~\ref{sec:sympol}, before evaluating it in the following section (Section~\ref{sec:sympol_eval}).

\section{Background: Reinforcement Learning} \label{sec:background_rl}

RL is a branch of machine learning concerned with how agents should take actions in an environment to maximize some notion of cumulative reward. This decision-making problem is often modeled using Markov Decision Processes (MDPs) or, when the environment is only partially observable, Partially Observable Markov Decision Processes (POMDPs) as we will introduce in Section~\ref{ssec:mdp}. After establishing the foundations of RL, we will introduce various RL methods (Section~\ref{ssec:on_off_policy}). Specifically, we will differentiate between on-policy and off-policy approaches and present actor-critic methods, along with the fundamentals of proximal policy optimization. These concepts provide the theoretical groundwork for embedding the proposed method within the broader RL framework.

\subsection{Reinforcement Learning as a Decision-Making Problem} \label{ssec:mdp}

\begin{figure}[t]
    \centering
    \includegraphics[width=0.8\textwidth]{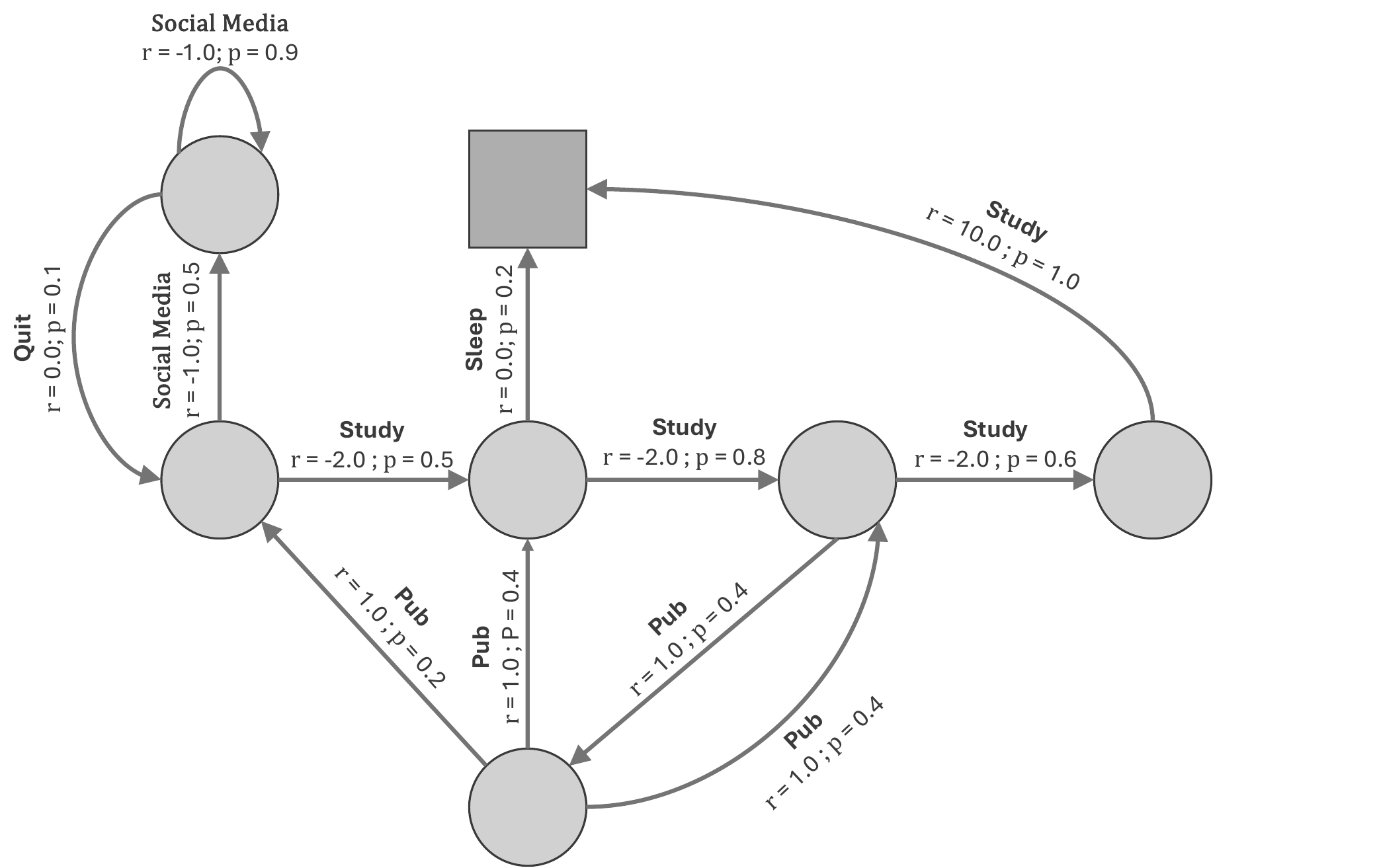}
    \caption[Exemplary MDP]{\textbf{Exemplary MDP.} 
    A schematic representation of the decision-making process in an MDP for daily activity choices from \citet{silver2015mdp}. Each node represents an action (e.g., \emph{Study}, \emph{Social Media}, \emph{Pub}, \emph{Quit}, and \emph{Sleep}), with associated rewards (\(r\)) and transition probabilities (\(p\)) indicated. Arrows illustrate the possible state transitions and their respective probabilities. %
    }
    \label{fig:mdp_structure}
\end{figure}

\paragraph{Markov Decision Process (MDP)}
An MDP is a mathematical framework used to model decision-making problems in which an agent interacts with an environment to achieve certain goals. The agent observes the current state of the environment, selects actions, and receives rewards that guide its future decisions. Formally, an MDP is defined by the tuple $(\mathcal{S}, \mathcal{A}, \mathcal{P}, r, \gamma)$, where~\citep{silver2015mdp}:

\begin{itemize}
    \item $\mathcal{S}$ is the finite state space, representing all possible configurations of the environment.
    \item $\mathcal{A}$ is the finite action space, which contains all actions the agent can take.
    \item $\mathcal{P}: \mathcal{S} \times \mathcal{A} \times \mathcal{S} \rightarrow [0,1]$ denotes the transition dynamics of the environment, specifying the probability of transitioning from state $s_t \in \mathcal{S}$ to state $s_{t+1} \in \mathcal{S}$ when action $a_t \in \mathcal{A}$ is taken.
    \item $r: \mathcal{S} \times \mathcal{A} \rightarrow \mathbb{R}$ is the reward function, which provides feedback to the agent by associating each state-action pair with a scalar reward.
    \item $\gamma \in [0, 1]$ is the discount factor, which determines the present value of future rewards.
\end{itemize}

Figure~\ref{fig:mdp_structure} illustrates an exemplary MDP representing choices between activities like \emph{Study}, \emph{Social Media}, \emph{Pub}, and \emph{Sleep}. Each action has an associated reward (\(r\)) and transition probability (\(p\)). For example, \emph{Study} often has a short-term cost (\(r = -2\)) but can occasionally result in a high positive reward (\(r = 10\)). Transition probabilities capture the stochastic nature of outcomes, such as varying success rates when studying (\(p = 0.5, 0.6, 0.8\)). This example highlights the trade-offs between immediate rewards and long-term benefits, central to MDPs.

\paragraph{The agent-environment cycle}

\begin{figure}[t]
    \centering
    \includegraphics[width=0.8\textwidth]{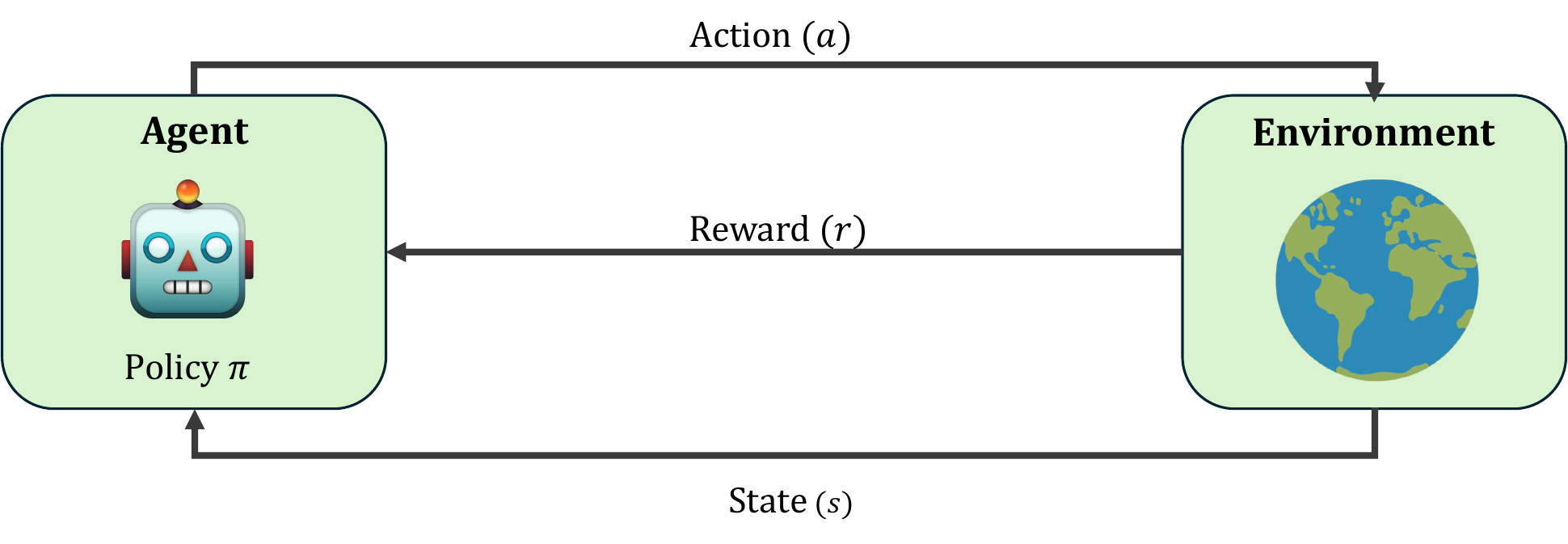}
    \caption[Agent-Environment Interaction Cycle]{\textbf{Agent-Environment Interaction Cycle.} This figure demonstrates how the agent interacts with the environment through observations, actions, transitions, and rewards, forming a feedback loop that drives learning in RL.}
    \label{fig:agent_environment_cycle}
\end{figure}

At each time step $t$, the agent observes the current state $s_t \in \mathcal{S}$, selects an action $a_t \in \mathcal{A}$ based on its policy $\pi: \mathcal{S} \rightarrow \mathcal{A}$, and executes this action in the environment. Consequently, the environment transitions to a new state $s_{t+1} \sim \mathcal{P}(s_t, a_t)$, and the agent receives a reward $r_t = r(s_t, a_t)$.

The interaction between the agent and the environment can be formalized as a cycle:
\begin{itemize}
    \item \textbf{State}: At time step $t$, the environment is in state $s_t$.
    \item \textbf{Observation}: The agent receives an observation $o_t$ (which may be identical to the state $s_t$ in fully observable MDPs).
    \item \textbf{Action}: The agent selects an action $a_t$ based on its policy $\pi$.
    \item \textbf{Transition}: The environment transitions to a new state $s_{t+1}$ according to the transition dynamics $\mathcal{P}(s_t, a_t)$.
    \item \textbf{Reward}: The agent receives a reward $r_t = r(s_t, a_t)$.
\end{itemize}

\paragraph{Action-value function and rewards}
The objective of RL is to learn a policy $\pi$ that maximizes the expected cumulative reward over time~\citep{silver2015mdp}. This cumulative reward is often referred to as the \emph{return} and is given by:
\begin{equation}
G_t = \sum_{k=0}^{\infty} \gamma^k r_{t+k}.
\end{equation}

In addition, the value function, $\mathcal{V}^\pi(s)$, and the action-value function, $\mathcal{Q}^\pi(s, a)$, are key concepts in RL and are defined as follows:
\begin{itemize}
    \item \textbf{State value function}: The value function $\mathcal{V}^\pi(s)$ estimates the expected return when starting in state $s$ and following policy $\pi$:
    \begin{equation}
    \mathcal{V}^\pi(s) = \mathbb{E}_{a_t \sim \pi, s_{t+1} \sim \mathcal{P}}\left[\sum_{t=0}^{\infty} \gamma^t r(s_t, a_t) \mid s_t = s\right].
    \end{equation}
    \item \textbf{Action-value function}: The action-value function $\mathcal{Q}^\pi(s, a)$ estimates the expected return when taking action $a$ in state $s$ and then following policy $\pi$:
    \begin{equation}
    \mathcal{Q}^\pi(s, a) = \mathbb{E}_{a_t \sim \pi, s_{t+1} \sim \mathcal{P}}\left[\sum_{t=0}^{\infty} \gamma^t r(s_t, a_t) \mid s_t = s, a_t = a\right].
    \end{equation}
\end{itemize}

Additionally, the \emph{advantage function}, defined as $\mathcal{A}^\pi(s, a) = \mathcal{Q}^\pi (s, a) - \mathcal{V}^\pi(s)$, provides a measure of how much better taking action $a$ in state $s$ is compared to following the policy $\pi$ directly.

\paragraph{Partially Observable MDPs (POMDPs)}

In many real-world scenarios, the agent does not have full access to the state of the environment, making the problem a POMDP~\citep{silver2015mdp}. Formally, a POMDP is defined by the tuple $(\mathcal{S}, \mathcal{A}, \mathcal{P}, r, \mathcal{O}, \mathcal{Z}, \gamma)$. Thereby, the state space $\mathcal{S}$, the action space $\mathcal{A}$, the transition dynamics $\mathcal{P}$, the reward function $r$ and the discount factor $\gamma$ are defined equal to the standard MDP and further:

\begin{itemize}
    \item $\mathcal{O}$ is the finite observation space, representing all possible observations the agent can receive about the underlying state of the environment.
    \item $\mathcal{Z}: \mathcal{S} \times \mathcal{A} \times \mathcal{O} \rightarrow [0, 1]$ is the observation function, which defines the probability of receiving observation $o_t \in \mathcal{O}$ given that the environment is in state $s_t \in \mathcal{S}$ and action $a_t \in \mathcal{A}$ was taken. This reflects the partial observability of the environment.
\end{itemize}

In the context of a POMDP, the key difference from the MDP framework lies in the agent's interaction with the environment. Instead of observing the true state $s_t$ at each time step, the agent receives an observation $o_t \in \mathcal{O}$, which only provides partial information about the underlying state. Therefore, the agent must infer the hidden state based on its belief or history of previous observations and actions. This impacts the agent's decision-making process, as actions are now selected based on the belief state or history rather than the true state, and the observation function $\mathcal{Z}(o_t | s_t, a_t)$ replaces the direct state observation. Consequently, the learning objective remains to maximize the expected cumulative reward, but the agent must adapt to the uncertainty in the environment's state by relying on indirect, incomplete observations.

\subsection{On-Policy vs. Off-Policy Reinforcement Learning} \label{ssec:on_off_policy}

RL methods can be broadly categorized into on-policy and off-policy approaches, depending on how they handle data collection and learning. Understanding the distinction between these two approaches is crucial for the categorization of our approach and will be elaborated in the following.

\begin{figure}[t]
\centering
\begin{subfigure}{0.475\columnwidth}
   \centering
  \includegraphics[width=\textwidth]{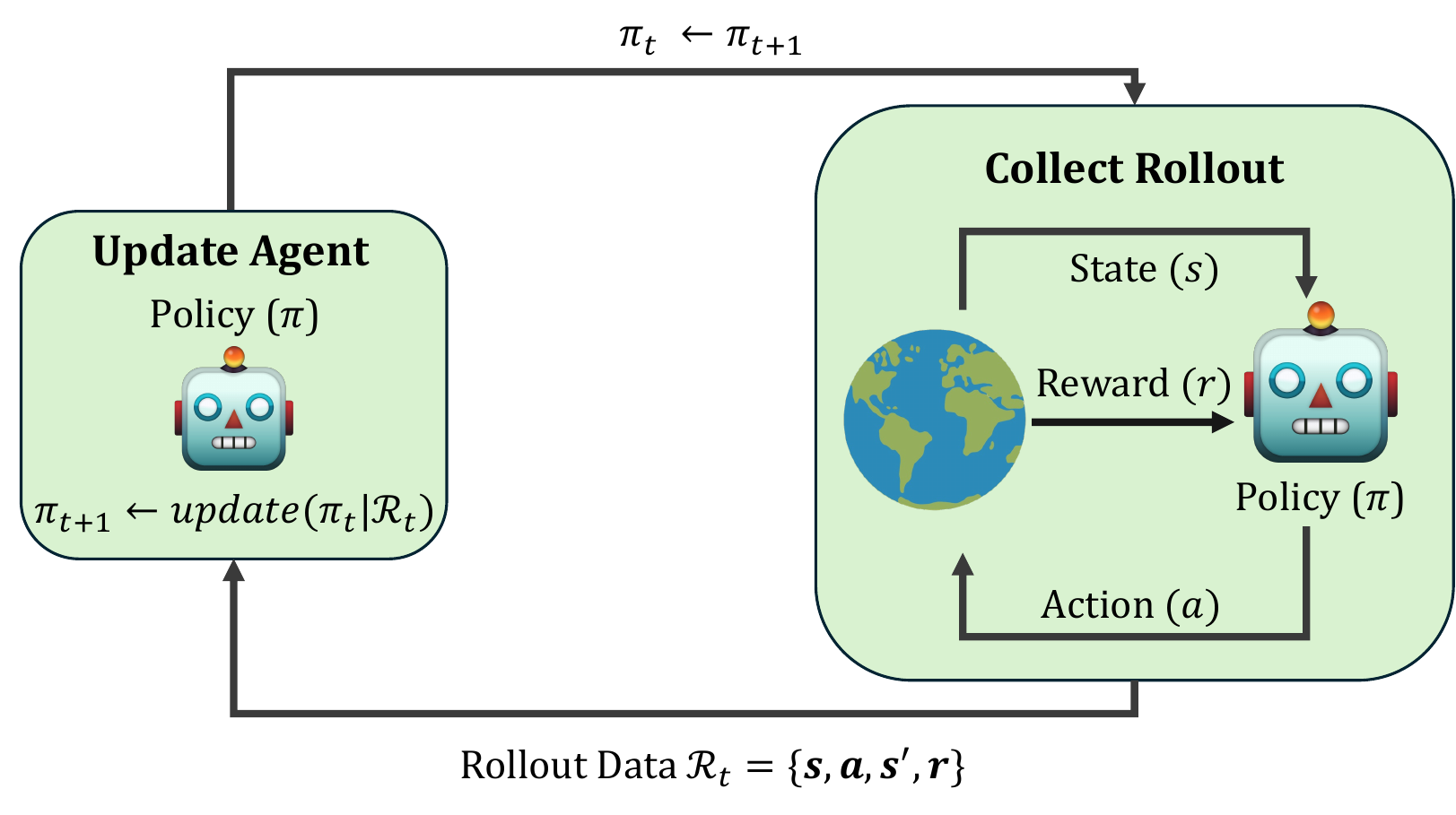}
  \caption[On-Policy RL]{\textbf{On-Policy RL.} This figure demonstrates how the current policy is used to generate data from the environment, and how it is subsequently updated using rollout data, which is derived from the same policy that is being updated.}
  \label{fig:on_policy}
\end{subfigure}%
\quad
\begin{subfigure}{0.475\columnwidth}
  \centering
  \includegraphics[width=\textwidth]{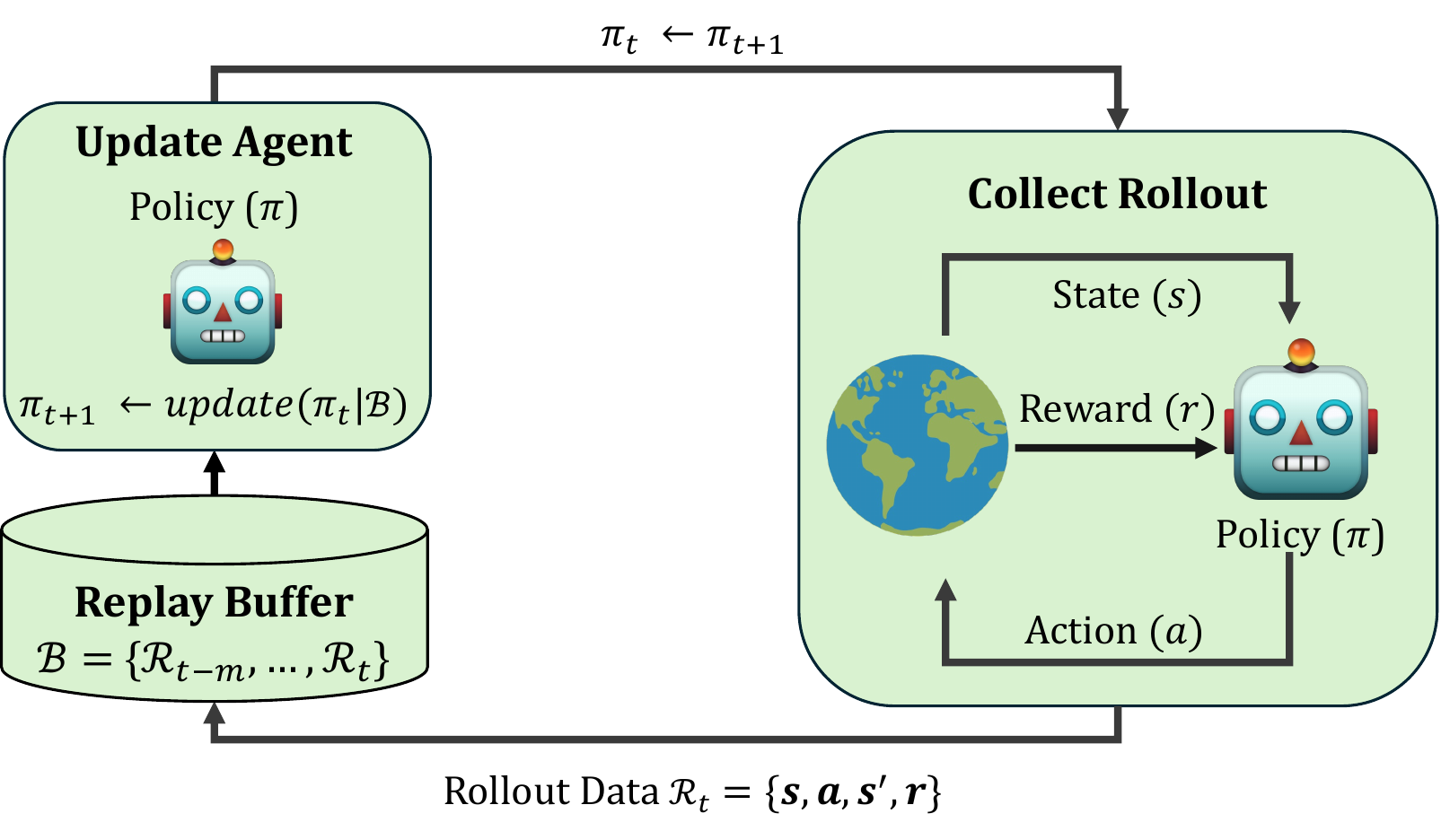}
  \caption[Off-Policy RL]{\textbf{Off-Policy RL.} This figure illustrates how off-policy methods decouple data generation from the current policy. The agent can store past interactions in a replay buffer and sample from this buffer to update the policy.}
  \label{fig:off_policy}
\end{subfigure}
\caption[On-Policy vs. Off-Policy RL]{\textbf{On-Policy vs. Off-Policy RL.}  This figure illustrates the differences in data generation and reuse between on-policy methods (A), which rely on fresh interactions, and off-policy methods (B), which leverage past experiences for greater sample efficiency.}
\label{fig:on_off_policy}
\end{figure}

\paragraph{On-policy learning}

In on-policy RL, the agent uses its current policy to both interact with the environment and update the policy. This tight coupling between data collection and learning means that the agent relies entirely on fresh data generated by the most recent policy for each learning update. A common example of an on-policy algorithm is the Policy Gradient method (e.g., REINFORCE~\citep{reinforce}), where the policy is explicitly optimized based on trajectories collected by the current policy.
Data generated by an on-policy RL algorithm is often stored in a \emph{rollout buffer}. This buffer contains sequences of states, actions, rewards, and next states collected over a fixed number of steps $T$. Each trajectory corresponds to the agent's interactions with the environment under the current policy. Specifically, the rollout buffer can be defined as a collection of trajectories over $T$ steps:

\begin{itemize}
\item \textbf{State buffer}:
   \[
   S = [\mathbf{s}_0, \mathbf{s}_1, \mathbf{s}_2, \ldots, \mathbf{s}_{T-1}] \in \mathbb{R}^{T \times N},
   \]
   where \( \mathbf{s}_t \in \mathbb{R}^{N} \) is the state at time \( t \).

\item \textbf{Action buffer}:
   \[
   A = [\mathbf{a}_0, \mathbf{a}_1, \mathbf{a}_2, \ldots, \mathbf{a}_{T-1}] \in \mathbb{R}^{T \times |\mathcal{A}|},
   \]
   where \( \mathbf{a}_t \in \mathbb{R}^{|\mathcal{A}|} \) is the action at time \( t \).

\item \textbf{Reward buffer}:
   \[
   R = [r_0, r_1, r_2, \ldots, r_{T-1}] \in \mathbb{R}^{T},
   \]
   where \( r_t \in \mathbb{R} \) is the reward received after taking action(s) \( \mathbf{a}_t \) in state \( \mathbf{s}_t \).

\item \textbf{Next state buffer}:
   \[
   \hat{S} = [\mathbf{s}_1, \mathbf{s}_2, \mathbf{s}_3, \ldots, \mathbf{s}_T] \in \mathbb{R}^{T \times N},
   \]
   where \( \mathbf{s}_{t+1} \in \mathbb{R}^{N} \).

\item \textbf{Done buffer}:
   \[
   D = [d_0, d_1, \ldots, d_T] \in \{0, 1\}^{T},
   \]
   where \( d_t \) indicates whether the episode terminated at time step \( t \)\footnote{The done buffer is not included in all definitions of the rollout buffer, as some environments or algorithms assume fixed-length trajectories where episodes do not terminate early. However, it becomes essential for environments with variable-length episodes or when the step size $T$ is larger than the trajectory length (or not a multiple of it).}.
   
\end{itemize}

The complete rollout can be represented as the set:
\[
\mathcal{R} = \{S, A, R, \hat{S}, D\}
\]
In on-policy methods, this data is used for a single update to the policy before being discarded. While this approach prevents learning from outdated data, it typically results in lower sample efficiency, as each data point is used only once and is then discarded.

\paragraph{Off-policy learning}
Off-policy RL, %
on the other hand, decouples data collection from the current policy. This allows the agent to learn from data generated by any policy, including past versions of the policy or entirely different policies. This allows for the reuse of past experiences, improving sample efficiency and making the learning process more data-efficient. 
Off-policy methods store past interactions with the environment in a \emph{replay buffer}. Unlike the rollout buffer, which is temporary and policy-aligned, the replay buffer is a persistent data structure that allows the agent to accumulate a large collection of transitions over time. This enables the agent to revisit older experiences and sample them randomly, breaking the sequential correlation in the data and improving stability during training. 
The replay buffer stores \( T \) transitions and can be represented as
\( \mathcal{B} = \{S, A, R, \hat{S}, D\} \).
However, unlike the rollout buffer, which is cleared after every policy update, the replay buffer maintains its data across updates, with older transitions being removed when the buffer reaches its maximum capacity. 
By randomly sampling from this buffer, off-policy methods like Deep Q-Networks~\citep{dqn} and Soft Actor-Critic~\citep{sac} can learn from past experiences, leading to more sample efficient training. This is in contrast to on-policy methods, which are constrained to use each interaction only once.

\begin{figure}[t]
    \centering
    \includegraphics[width=0.8\textwidth]{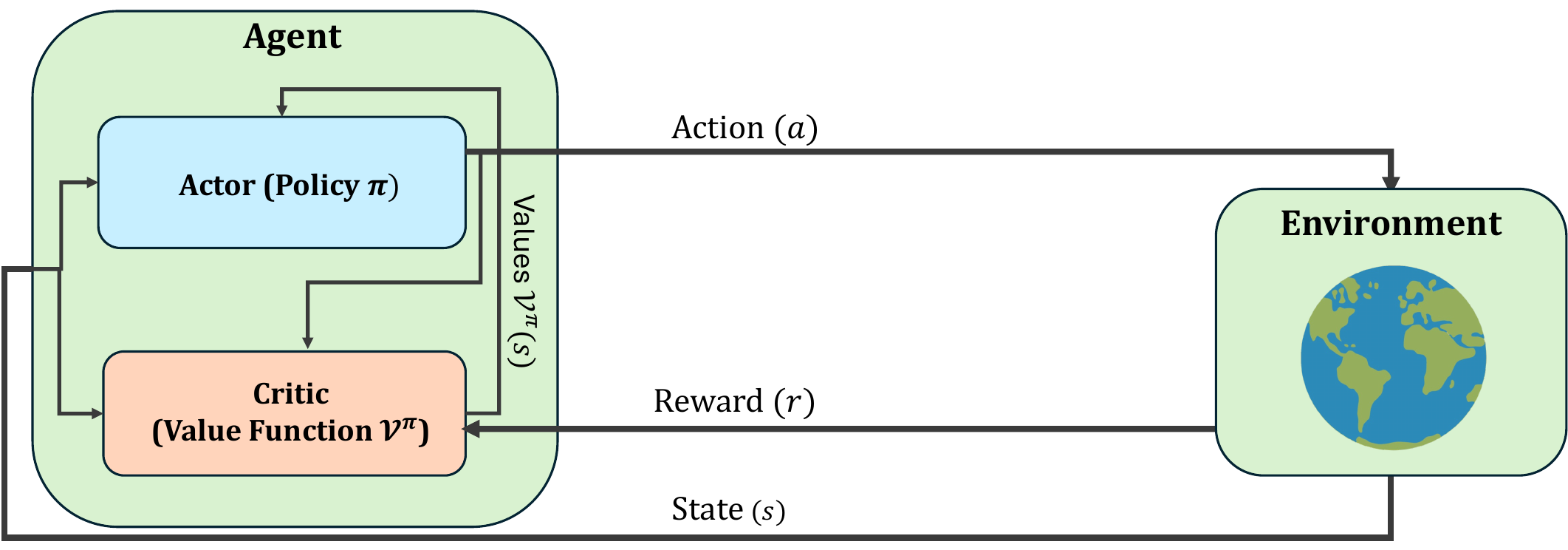}
    \caption[Actor-Critic Architecture]{\textbf{Actor-Critic Architecture.} The actor selects actions based on its policy, while the critic evaluates these actions to provide feedback in the form of value estimates or advantages, which are used to update the policy.}
    \label{fig:actor_critic}
\end{figure}

\paragraph{Actor-critic methods}

One popular class of RL algorithms is \emph{actor-critic methods}. Actor-critic methods can be both, on-policy (e.g., Advantage Actor-Critic~\citep{mnih2016asynchronous} or Proximal Policy Optimization~\citep{schulman2017proximal}) or off-policy (e.g., Deep Deterministic Policy Gradient~\citep{lillicrap2015continuous} or Soft Actor-Critic~\citep{sac}). These methods combine the benefits of value-based and policy-based approaches by introducing two components:
\begin{itemize}
    \item \textbf{Actor}: The actor is responsible for choosing actions based on a policy $\pi_\theta$. It interacts with the environment and decides which action to take given the current state. %
    \item \textbf{Critic}: The critic evaluates the quality of the actions taken by the actor. It learns a value function $\mathcal{V}^\pi(s)$ or an action-value function $\mathcal{Q}^\pi(s, a)$ that provides feedback to the actor on how good its actions are. The critic is also parameterized by a set of parameters, often denoted by $\phi$.
\end{itemize}

The actor-critic approach aims to improve the policy by reducing the discrepancy between the value function estimated by the critic and the true return observed in the environment. The objective of the critic is to minimize a loss function that represents the difference between the predicted value and the actual return:
\begin{equation}
\mathcal{L}(\phi) = \mathbb{E}_{s_t \sim \mathcal{P}} \left[ \left( \mathcal{V}_\phi(s_t) - \left( r_t + \gamma \mathcal{V}_\phi(s_{t+1}) \right) \right)^2 \right].
\end{equation}

The actor then uses the feedback from the critic to adjust the policy parameters $\theta$ by following the policy gradient:
\begin{equation}
\nabla_\theta J(\theta) = \mathbb{E}_{a_t \sim \pi_\theta, s_t \sim \mathcal{P}} \left[ \nabla_\theta \log \pi_\theta (a_t | s_t) \hat{A}_t \right],
\end{equation}
where $\hat{A}_t$ is an estimate of the advantage function, which indicates how much better the chosen action is compared to the average action taken in that state.
Actor-critic methods benefit from the policy gradient's ability to optimize complex policies, while also taking advantage of the critic's stable evaluation of the policy's performance. This balance often leads to improved sample efficiency and learning stability compared to purely value-based or policy-based methods.

\paragraph{Proximal Policy Optimization (PPO)}

One prominent on-policy actor-critic RL algorithm is Proximal Policy Optimization (PPO). PPO aims to stabilize the training process by preventing abrupt policy updates. The PPO algorithm uses a clipped surrogate objective to constrain the change in the policy:
\begin{equation}
\begin{aligned}
\mathcal{L}^{\text{CLIP}}(\theta) &= 
\mathbb{E}_{a_t \sim \pi_{\theta_{\text{old}}},\, s_{t+1} \sim \mathcal{P}}
\bigg[ \min \Big( 
\frac{\pi_\theta(a_t \mid s_t)}{\pi_{\theta_{\text{old}}}(a_t \mid s_t)} \hat{A}_t, \\
&\qquad\qquad
\text{clip}\Big(\frac{\pi_\theta(a_t \mid s_t)}{\pi_{\theta_{\text{old}}}(a_t \mid s_t)}, 
1 - \epsilon, 1 + \epsilon\Big) \hat{A}_t 
\Big) \bigg],
\end{aligned}
\end{equation}
where $\frac{\pi_\theta(a_t|s_t)}{\pi_{\theta_{\text{old}}}(a_t|s_t)}$ is the probability ratio between the new policy $\pi_\theta$ and the old policy $\pi_{\theta_{\text{old}}}$. $\hat{A}_t$ is an estimate of the advantage function at time step $t$ and $\epsilon$ is a hyperparameter that controls the clipping range.
The objective encourages small updates to the policy, preventing large policy changes that could destabilize training. By optimizing this objective, PPO achieves a balance between exploration and exploitation, allowing the agent to effectively learn a policy.

\section{Related Work}  \label{sec:related_rl}
Recently, the integration of symbolic methods into RL has gained significant attention. %
Symbolic RL does cover different approaches including
program synthesis \citep{trivedi2021learning, penkov2019learning, verma2018programmatically}, concept bottleneck models \citep{ye2024concept}, piecewise linear networks \citep{wabartha2024piecewise}, logic \citep{delfosse2024interpretable} and mathematical expressions \citep{landajuela2021discovering, guo2024efficient, luoend, kamienny2022symbolic}. 
Another line of work aims to synthesize symbolic policies using logical rules, leveraging differentiable inductive logic programming for gradient-based optimization~\citep{jiang2019neural,cao2022galois}. In contrast to first-order rules, DTs offer greater flexibility by not only combining atomic conditions but also comparing features against thresholds — a critical capability for handling continuous observation spaces.
In the following, we focus exclusively on tree-based methods and actor-critic, on-policy RL, as introduced in Section~\ref{ssec:on_off_policy}. %
Several approaches have been proposed to leverage the strengths of interpretable, tree-based representations within RL. 
However, each approach comes with its own critical limitations. 
We summarize existing methods into three streams of work:

\begin{enumerate}
    \item \textbf{Post-processing} \hspace{1.5mm} 
        One line learns full-complexity policies first and then performs some kind of post-processing for interpretability. One prominent example is the VIPER algorithm \citep{bastani2018viper}. In this case, a policy is learned using neural networks before DTs are distilled from the policy. However, distillation methods often suffer from significant performance mismatches between the training and evaluation policies (see Figure~\ref{fig:sadt}). 
        To mitigate this mismatch, existing methods often learn large DTs (VIPER learns DTs with 1,000 nodes) and therefore aim for systematic verification rather than interpretability. In contrast, SYMPOL is able to learn small, interpretable DTs (average of only 50 nodes) without information loss.
        Following VIPER, various authors proposed  distillation methods \citep{li2021neural, liu2019toward, liu2023effective, jhunjhunwala2020improved}, which exhibit similar limitations.
    \item \textbf{Soft Decision Trees (SDTs)} \hspace{1.5mm}
        Methods optimizing SDTs~\citep{silva2020optimization, silva2021encoding, coppens2019distilling, tambwekar2023natural,liu2022mixrts,farquhar2017treeqn} are difficult to interpret due to missing hard and axis-aligned properties \citep{silva2020optimization, silva2021encoding, coppens2019distilling, tambwekar2023natural} as already discussed in Section~\ref{ssec:sdt_background} and Section~\ref{sec:sdt_related}. Since SDTs model probability distributions in their internal nodes and, therefore, allow gradient optimization, they can be integrated into existing RL approaches. Nevertheless, the trees are usually not easily interpretable and techniques such as discretizing the learned trees into more interpretable representations are applied \citep{silva2020optimization}, occasionally resulting in high performance mismatches (Figure~\ref{fig:d-sdt}). In contrast, SYMPOL directly optimizes hard, axis-aligned DTs and therefore does not exhibit a performance loss (Figure~\ref{fig:sympol}). 
    \item \textbf{Custom optimization} \hspace{1.5mm} 
        The third line involves custom, tree-specific optimization techniques and/or objectives \citep{ernst2005tree, roth2019conservative, gupta2015policy, kanamori2022counterfactual} and, hence, is more time-consuming and less flexible. 
        As a result, their policy models cannot be easily integrated into existing learning RL approaches. Examples are evolutionary methods \citep{costa2024evolving, custode2023evolutionary} and linear integer programming \citep{vos2023optimal}.
        
\end{enumerate}

\paragraph{Distinction of SYMPOL from differentiable and soft decision trees} %
In existing work, such as \citet{silva2020optimization}, differentiable DTs typically correspond to SDTs, achieving differentiability by relaxing discrete decisions in terms of feature selection at each internal node and path selection. This approach is fundamentally different from SYMPOL, which does \emph{not} use differentiable DTs. Instead, SYMPOL leverages GradTree to optimize standard, non-differentiable DTs through gradient descent, as we will show in Section~\ref{sec:sympol} and more detailed in Appendix~\ref{A:SYMPOL}. \\

Furthermore, trees have also been used in agentic components beyond the policy, such as reward functions \citep{maviper2023, kalra2023can, kalra2022interpretable}, which are outside the scope of this work. Similarly, ensemble methods \citep{fuhrer2024gradient, min2022q} have been proposed. However, policies consisting of hundreds of trees and nodes lack interpretability and therefore are out of scope for this thesis. 

\begin{figure}[t]
    \centering
    \includegraphics[width=0.9\linewidth]{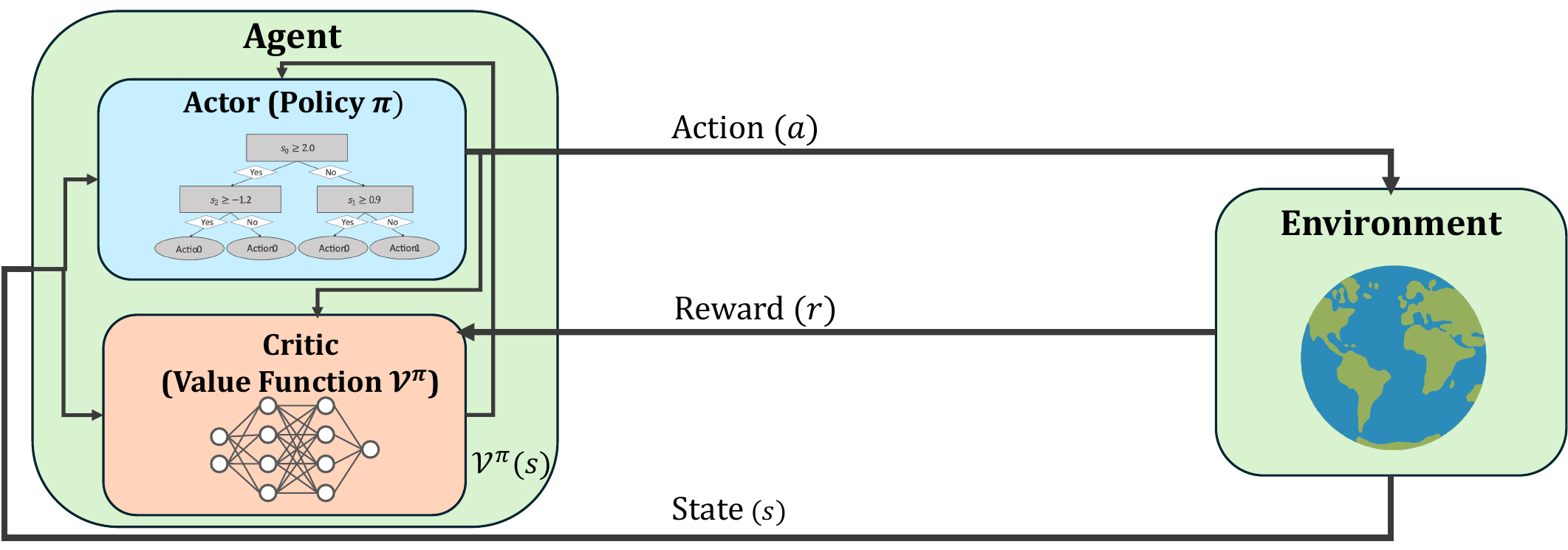}
    \caption[SYMPOL overview]{\textbf{SYMPOL overview.} This is an overview of SYMPOL, with its two main components: (1) An interpretable DT policy and (2) a neural network as critic.}%
    \label{fig:sympol_overview}
\end{figure}

\section{SYMPOL: Symbolic On-Policy RL}\label{sec:sympol}
One of the key advantages of gradient-based DTs over alternative methods lies in their enhanced flexibility, which stems from their gradient-based nature. This flexibility facilitates seamless integration into existing frameworks that rely on gradient descent and backpropagation, such as those used in RL. By embedding GradTree within actor-critic, on-policy RL algorithms like PPO, as shown in Figure~\ref{fig:sympol_overview}, we enable the direct optimization of a symbolic, tree-based policy.
A formal framework for the online training of GradTree with an on-policy objective is provided in Appendix~\ref{A:sympol_formalization}. To efficiently train DT policies using SYMPOL, we incorporated several critical modifications, as demonstrated in the ablation study (Table~\ref{fig:ablation}). In this section, we focus exclusively on the changes made to the overall framework described in Section~\ref{sec:method_gradtree}.
The main conceptual difference to existing work that learn symbolic policies end-to-end \citep{delfosse2024interpretable,fuhrer2024gradient,delfosse2024interpretableCBM,topin2021custard,luoend} is that SYMPOL does \emph{not} require any modification of the RL algorithms themselves, making the proposed method method-agnostic. As a result, interpretable policies with SYMPOL are learned in the same way as NN policies are commonly learned.
While we focus on PPO as the, we believe, most prominent on-policy RL method, our approach is not restricted to it and can be extended to any gradient-based RL approach. To substantiate our claim of seamless integration, we present additional results using Advantage Actor-Critic (A2C), as proposed by \citet{mnih2016asynchronous}, in Appendix~\ref{A:A2C}. These results align with the findings from the experiments conducted in Section~\ref{sec:results}.

\subsection{Improving Gradient-Based Decision Trees for Dynamic Reinforcement Learning Environments} \label{sec:dt_policy}

\paragraph{Weight decay}%
In contrast to GradTree, which employs an Adam~\citep{kingma2014adam} optimizer with stochastic weight averaging~\citep{izmailov2018averaging}, we opted for an Adam optimizer with decoupled weight decay~\citep{loshchilov2017decoupled}, as introduced in Section~\ref{sec:weight_dacay}. In the context of SYMPOL, weight decay does not serve as a direct regularizer for model complexity as it does in standard neural networks, since the interpretation of model parameters differs. We distinguish between three types of parameters: The predicted distributions in the leaves, the split index encoding, and the split values. We do not apply weight decay to the split values because their significance is independent of magnitude. However, for the split indices and leaves, weight decay facilitates exploration during training by penalizing large parameter values. 
As a result, the distribution for the split index selection and class prediction are narrow and have lower variance.
This aids in dynamically adjusting which feature is considered at a split and in altering the class predictions at the leaves. 
Additionally, we decrease the learning rate if no improvement in validation reward is observed for five consecutive iterations, allowing for finer model adjustments in later training stages.

\paragraph{Actor-critic network architecture}%
Commonly, the actor and critic use a similar network architecture or even share the same weights~\citep{schulman2017proximal}. While SYMPOL aims for a simple and interpretable policy, we do not have the same requirements for the critic. Therefore, we decided to only employ a tree-based actor and use a full-complexity neural network as a value function. As a result, we can still capture complexity through the value function, without losing interpretability, as we maintain a simple and interpretable policy. We visualized the concept of SYMPOL in Figure~\ref{fig:sympol_overview}.

\paragraph{Continuous action spaces}%
Furthermore, we extend the DT policy of SYMPOL to environments with continuous action spaces. Therefore, instead of predicting a categorical distribution over the classes, we predict the mean of a normal distribution at each leaf and utilize an additional variable $\sigma_{\text{log}} \in \mathbb{R}^{|\mathcal{A}|}$ to learn the log of the standard deviation.

\subsection{Addressing Training Stability} \label{sec:training_stability}
One main challenge when using DTs as a policy is the stability. While a stable training is also desired and often hard to achieve for a neural network policy, this is even more pronounced for SYMPOL. This is mainly caused by the inherent tree-based architecture. Changing a split at the top of the tree can have a severe impact on the whole model, as it can completely change the paths taken for certain observations. This is especially relevant in the context of RL, where the data distribution can vary highly between iterations. To mitigate the impact of highly non-stationary training samples, especially at early stages of training, we made two crucial modifications for improved stability of SYMPOL.

\begin{figure}[t]
   \centering
   \includegraphics[width=0.75\textwidth]{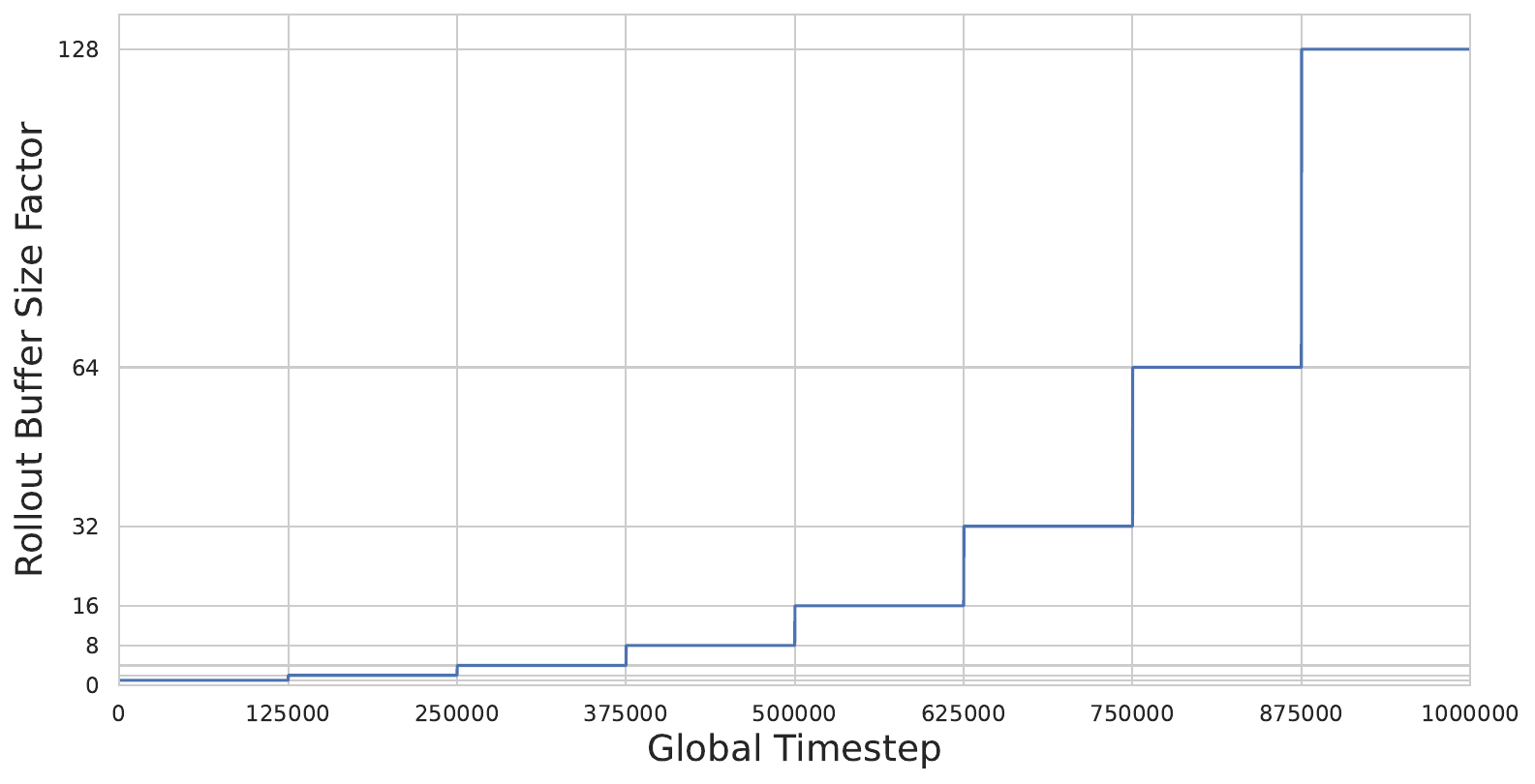}
   \caption[Rollout Size Factor by Global Timestep]{\textbf{Rollout Size Factor by Global Timestep.} This figure shows how the factor by which the rollout buffer size (= the number of environment steps per training iteration) is multiplied increases at a stepwise exponential rate over time.}
   \label{fig:rollout_increase}
\end{figure}

\paragraph{Exploration stability}%
 Motivated by the idea that rollouts of more accurate policies contain increasingly diverse, higher quality samples, we implemented a dynamic number of environment steps between training iterations. Let us consider a Pendulum as an example. 
While at early stages of training a relatively small sample size facilitates faster learning as the Pendulum constantly flips, more optimal policies lead to longer rollouts and therefore more expressive and diverse experiences in the rollout buffer. Similarly, the increasing step counts stabilize the optimization of policy and critic, as the number of experiences for gradient computation grow with agent expertise and capture the diversity within trajectories better. Therefore, our novel collection approach starts with $m_{init}$ environment steps and expands until $m_{final}$ actions are taken before each training iteration. For computational efficiency reasons, instead of increasing the size of the rollout buffer at every time step, we introduce a step-wise exponential function (Figure~\ref{fig:rollout_increase}). 
The exponential increase supports exploration in the initial iterations, while maintaining stability at later iterations. Hence, we define the number of steps in the environment $m_t$ at time step $t$ as 
\begin{equation}
\begin{split} 
    m_t = m_\text{init} \times 2^{\left\lfloor \frac{(t+1) \times i}{1 + t_\text{total}} \right\rfloor - 1}  \;
     \text{with} \;
     i = 1 + \log_2\left(\frac{m_\text{init}}{m_\text{final}}\right).
     \label{step-wise-exponential}
\end{split} 
\end{equation}
For our experiments, we define $m_\text{init}$ as a hyperparameter (similar to the static step size for other methods) and set $m_\text{final} = 128 \times m_\text{init}$ and therefore $i=8$ which we observed is a good default value.

\paragraph{Gradient stability}%
We also utilize large batch sizes for SYMPOL resulting in less noisy gradients, leading to a smoother convergence and better stability. In this context, we implement gradient accumulation to virtually increase the batch size further while maintaining memory-efficiency. As reduced noise in the gradients also leads to less exploration in the parameter space, we implement a dynamic batch size, increasing in the same rate as the environment steps between training iterations (Equation~\ref{step-wise-exponential}). Therefore, we can benefit from exploration and fast convergence early on and increase gradient stability during the training.

\section{Evaluation} \label{sec:sympol_eval}
We designed our experiments to evaluate whether SYMPOL can learn accurate DT policies without information loss and observe whether the trees learned by SYMPOL are small and interpretable.
Furthermore, we conducted an ablation study to show how the proposed design choices (Section~\ref{sec:sympol}) affect the performance.
As mentioned above, we focus on PPO as the most prominent actor-critic, on-policy RL algorithm in our evaluation. To support our claim that SYMPOL can be seamlessly integrated into existing on-policy RL methods, we additionally provide results using A2C in Appendix~\ref{A:A2C}.

\subsection{Experimental Settings}

\paragraph{Setup} %
We implemented SYMPOL in a highly efficient single-file JAX implementation\footnote{Our implementation is publically available under: \url{https://github.com/s-marton/SYMPOL}.} that allows a flawless integration with highly optimized training frameworks~\citep{gymnax2022github, weng2022envpool, bonnet2024jumanji}.
We evaluated our method on several environments commonly used for benchmarking RL methods. Specifically, we used control environments including CartPole (CP), Acrobot (AB), LunarLander (LL), MountainCarContinuous (MC-C) and Pendulum (PD-C), as well as the MiniGrid~\citep{minigrid} environments Empty-Random (E-R), DoorKey (DK), LavaGap (LG) and DistShift (DS).

\paragraph{Methods} %
The goal of this evaluation is to compare SYMPOL to alternative methods that allow an interpretation of RL policies as a symbolic, axis-aligned DTs. Therefore, we build on previous work \citep{silva2020optimization} and use two methods grounded in the interpretable RL literature, as follows:
\begin{itemize}
    \item \textbf{State-Action DTs (SA-DT)}: Behavioral cloning SA-DTs are the most common method to generate interpretable policies post-hoc. Hereby, we first train an MLP policy, which is then distilled into a DT as a post-processing step after the training. SA-DT can be considered as a version of DAGGER~\citep{ross2011reduction} and therefore a simplified version of VIPER~\citep{bastani2018viper}. In a comparative experiment (see Appendix~\ref{A:viper_comparison}), we showed that for the case of learning small, interpretable DTs the performance of SA-DT is similar to those of VIPER, which is in-line with results reported e.g., by ~\citet{kohler2024interpretable}. %
    \item \textbf{Discretized Soft DTs (D-SDT)}: SDTs allow gradient computation by assigning probabilities to each node. While SDTs exhibit a hierarchical structure, they are usually considered as less interpretable, since multiple features are considered in a single split and the whole tree is traversed simultaneously. Therefore, \citet{silva2020optimization} use SDTs as policies which are discretized post-hoc to allow an easy interpretation. %
\end{itemize}
We further included an MLP and SDT as full-complexity models, providing an orientation to state-of-the-art results. %

\paragraph{Evaluation procedure}%
We report the average undiscounted cumulative reward over 5 random trainings with 5 random evaluation episodes each (=25 evaluations for each method). We trained each method for 1mio timesteps.
For SYMPOL, SDT and MLP, we optimized the hyperparameters based on the validation reward with optuna~\citep{akiba2019optuna} for 60 trials using a predefined grid. %
For D-SDT we discretized the SDT and for SA-DT, we distilled the MLP with the highest performance. More details on the methods and the parameter grids, including the final parameters for each method, are listed in Appendix~\ref{A:hyperparameters_sympol}. 

\subsection{Results} \label{sec:results}

\begin{table}[t]
    \centering
    \small
    \caption[Control Performance]{\textbf{Control Performance.} We report the average undiscounted cumulative test reward over 25 random trials. The best interpretable method, and methods not statistically different, are marked bold.}
    
    \begin{tabular}{lrrrrr}
    \toprule
             & \multicolumn{1}{l}{CP}  & \multicolumn{1}{l}{AB}  & \multicolumn{1}{l}{LL}  & \multicolumn{1}{l}{MC-C}    & \multicolumn{1}{l}{PD-C}\\ \midrule   
         SYMPOL (ours)     & \bftab 500        &  \bftab -\phantom{0}80   &  \bftab -\phantom{0}57    &  \bftab \phantom{-}94       &  \bftab -\phantom{0}323     \\
         D-SDT             &  128    &    -205        &  -221                     &  -10                        &  -1343                      \\
         SA-DT (d=5)       &  446    &    -97        &  -197                     &  \bftab \phantom{-}97       &  -1251                      \\
         SA-DT (d=8)       &  476    &   \bftab -\phantom{0}75  &  -150                     &  \bftab \phantom{-}96       &  -\phantom{0}854            \\\midrule 
         
         MLP               & 500               &    -\phantom{0}72        &  \phantom{-}241                     &  \phantom{-}95            &  -\phantom{0}191            \\
         SDT               & 500               &     -\phantom{0}77                   & -124            &  -\phantom{0}4            &  -\phantom{0}310            \\             
         \bottomrule
    \end{tabular}
    \label{tab:results_control}
\end{table}
\paragraph{SYMPOL learns accurate decision tree policies}%
We evaluated our approach against existing methods on control environments in Table~\ref{tab:results_control}. SYMPOL is consistently among the best interpretable models and achieves significantly higher rewards compared to alternative methods for learning DT policies on several environments, especially on LL and PD-C. Further, SYMPOL consistently solves CP and AB and is competitive to full-complexity models on most environments. %

\begin{table}[t]
\centering
\small
\caption[MiniGrid Performance]{\textbf{MiniGrid Performance.} We report the average undiscounted cumulative test reward over 25 random trials. The best interpretable method, and methods not statistically different, are marked bold.}
\begin{tabular}{lrrrrr}
\toprule
 & \multicolumn{1}{l}{E-R} & \multicolumn{1}{l}{DK} & \multicolumn{1}{l}{LG-5} & \multicolumn{1}{l}{LG-7} & \multicolumn{1}{l}{DS} \\ \midrule
SYMPOL (ours)   & \bftab 0.964  & \bftab 0.959  & \bftab 0.951  & \bftab 0.953  & 0.939 \\
D-SDT           & 0.662         & 0.654         & 0.262         & 0.381         & 0.932 \\
SA-DT (d=5)     & 0.583         & \bftab 0.958  & \bftab 0.951  & 0.458         & \bftab 0.952 \\
SA-DT (d=8)     & 0.845         & \bftab 0.961  & \bftab 0.951  & 0.799         & \bftab 0.954 \\ \midrule
MLP             & 0.963         & 0.963         & 0.951         & 0.760         & 0.951 \\ %
SDT             & 0.966         & 0.959         & 0.839         & 0.953         & 0.954 \\
\bottomrule
\end{tabular}
\label{tab:results_MiniGrid}
\end{table}
\paragraph{DT policies offer a good inductive bias for categorical environments}
While SYMPOL achieves great results in control benchmarks, it may not be an ideal method for environments modeling physical relationships. As recently also noted by \citet{fuhrer2024gradient}, tree-based models are best suited for categorical environments due to their effective use of axis-aligned splits. In our experiments on MiniGrid (Table~\ref{tab:results_MiniGrid}), SYMPOL achieves comparable or superior results to full-complexity models (e.g., on LG-7). 
The performance gap between SA-DT and SYMPOL is smaller in certain MiniGrid environments due to less complex environment transition functions and missing randomness, making the distillation easy. Considering more complex environments with randomness or lava like E-R or LG-7, SYMPOL outperforms alternative methods by a substantial margin.

\begin{figure}[t]
    \centering
    \begin{subfigure}[H]{0.32\textwidth}
        \centering
        \includegraphics[width=\linewidth]{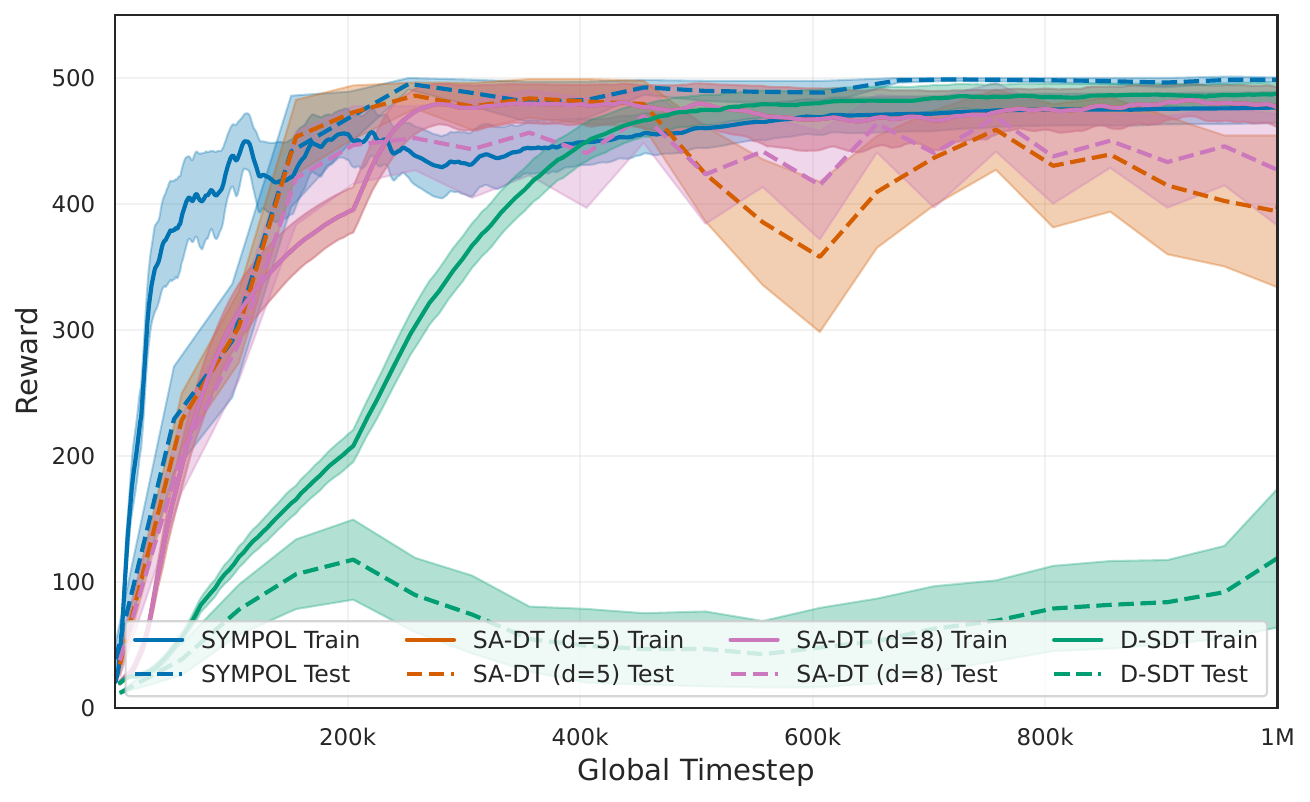}
        \caption{CartPole}
        \label{fig:learning_curves_COMBINED_NEWCartPole}
    \end{subfigure}
    \begin{subfigure}[H]{0.32\textwidth}
        \centering
        \includegraphics[width=\linewidth]{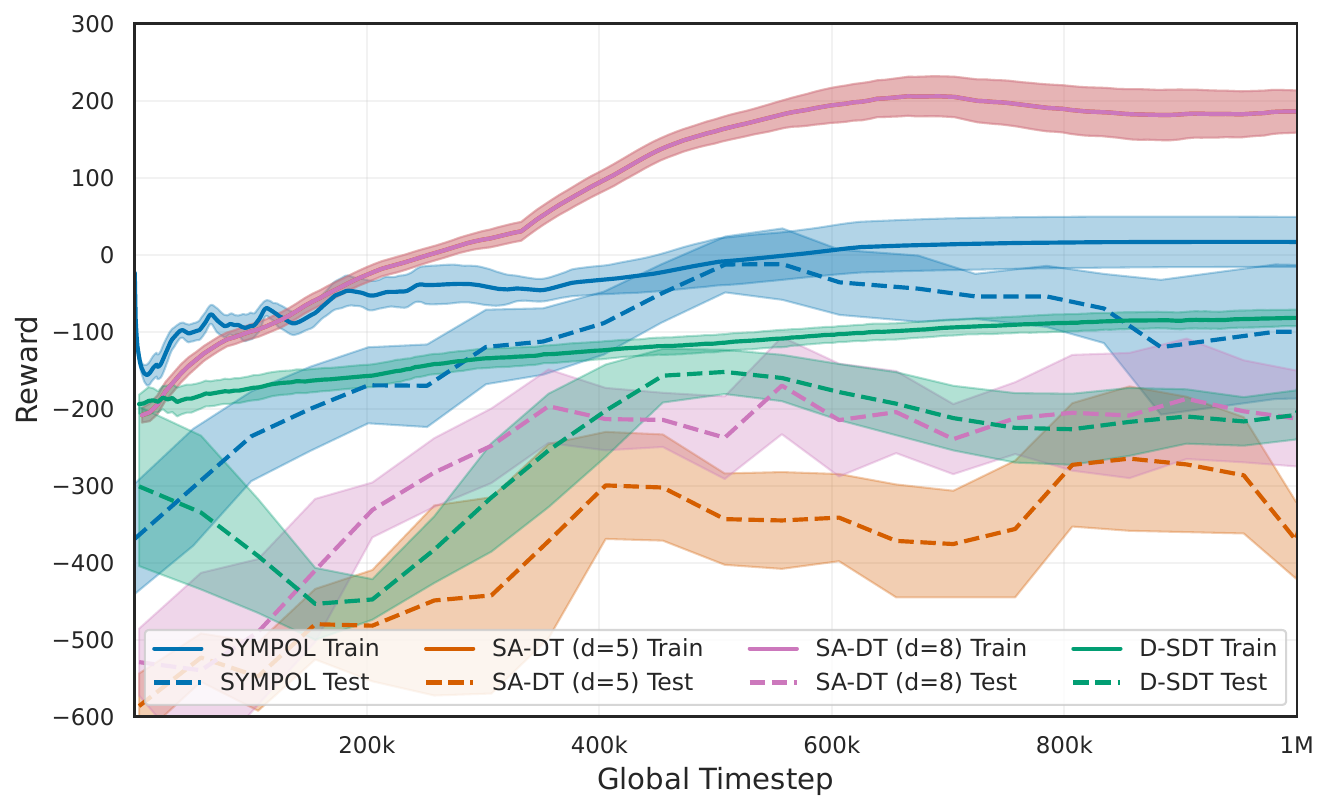}
        \caption{LunarLander}
        \label{fig:learning_curves_COMBINED_NEWLunarLander}
    \end{subfigure}   
    \begin{subfigure}[H]{0.32\textwidth}
        \centering
        \includegraphics[width=\linewidth]{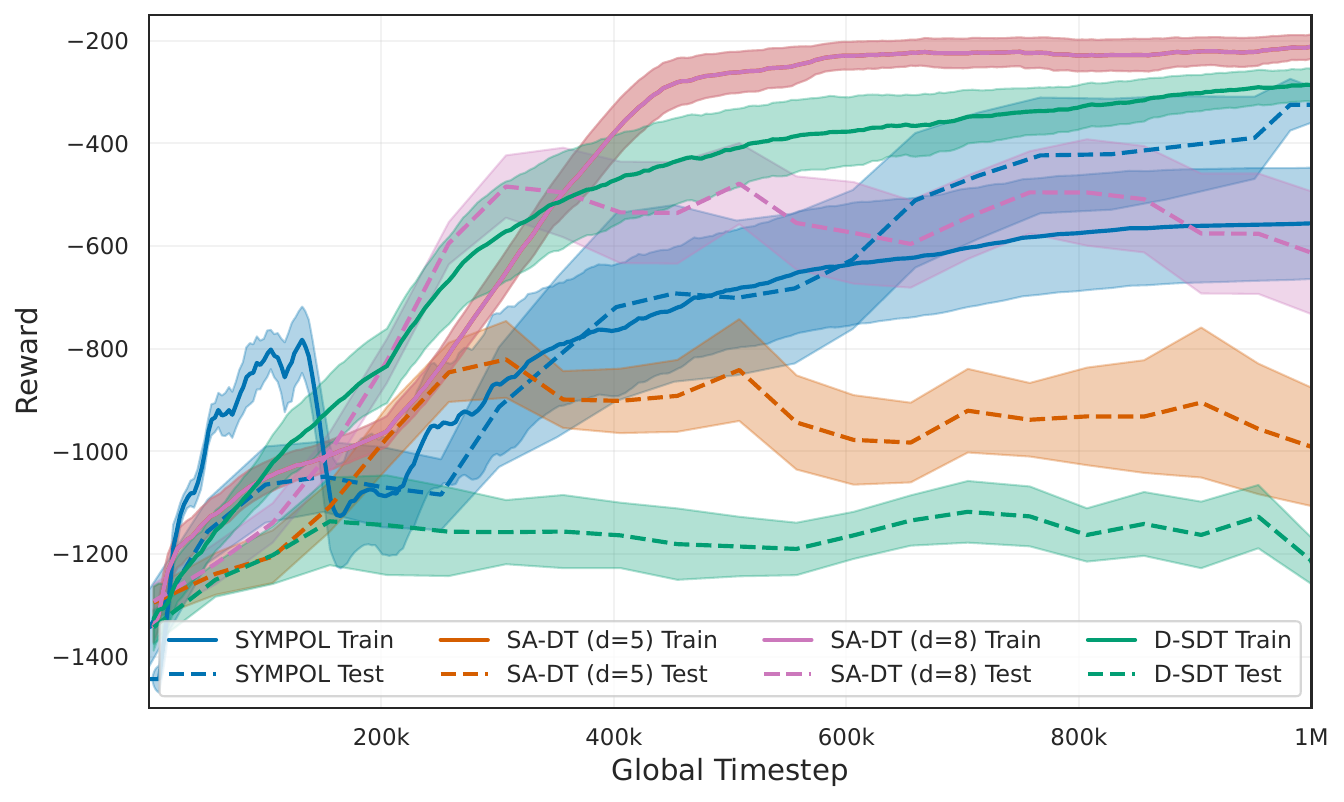}
        \caption{Pendulum (cont.)}
        \label{fig:learning_curves_COMBINED_NEWPendulum}
    \end{subfigure}    
    \caption[Selected Training Curves]{\textbf{Selected Training Curves.} Shows the training reward of the full-complexity policy (e.g., MLP in the case of SA-DT) as solid line and the test reward of the interpretable policy as dashed line for three control environments. Additional, more detailed results are in Appendix~\ref{A:runtime_curves}.}
    
    \label{fig:training_curves_selected}
\end{figure}

\begin{table}[t]
\centering
\small
        \caption[Information Loss]{\textbf{Information Loss.} We calculated Cohen's D to measure effect size between the validation reward of the trained model and the test reward of the applied model. Typically, values $>0.8$ are considered as a large effect. Detailed results are listed in Appendix~\ref{A:information_loss}}          
        \resizebox{0.95\linewidth}{!}{
        \begin{tabular}{l|rrrr|rr}
            \toprule
            & SYMPOL (ours) & SA-DT (d=5) & SA-DT (d=8) & D-SDT & MLP & SDT \\ \midrule
            Cohen's D $\downarrow$ & \bftab -0.019 & \phantom{-}3.449 & \phantom{-}2.527 & \phantom{-}3.126 & \phantom{-}0.306 & \phantom{-}0.040 \\ 
            \bottomrule
        \end{tabular}        
        }
        \label{tab:train_test_diff}
\end{table}
\paragraph{SYMPOL does not exhibit information loss}%
Existing methods for learning DT policies usually involve post-processing to obtain the interpretable model. Therefore, they introduce a mismatch between the optimized and interpreted policy, which can result in information loss. An example is given in Figure~\ref{fig:methods_comparison} and more examples are in Figure~\ref{fig:training_curves_pairwise}.
The main advantage of SYMPOL is the direct optimization of a DT policy, which guarantees that there is no information loss between the optimized and interpreted policy. 
To show this, we calculated Cohen's D to measure the effect size comparing the validation reward of the trained model with the test reward of the applied, optionally post-processed model (Table~\ref{tab:train_test_diff}). We can observe very large effects for SA-DT and D-SDT and only a very small effect for SYMPOL, similar to full-complexity models MLP and SDT. This discrepancy can also be observed in the training curves in Figure~\ref{fig:training_curves_selected}.

\begingroup
\begin{wraptable}{r}{0.525\textwidth}
\vspace{-\intextsep} %
\begin{minipage}{0.525\textwidth}
\small
    \centering
        \caption[Runtime SYMPOL]{\textbf{Runtime SYMPOL.} We report the average runtime over 25 trials with 1mio timesteps each. We excluded LL, as this is the only non-vectorized environment.}    
    \begin{tabular}{lrrr}
    \toprule
                                    & SYMPOL (ours)     & SDT               & MLP          \\ \midrule
         CP                   &   28.8            &  23.9             &  25.2         \\     
         AB                    &   35.5            &  37.7             &  33.8        \\     
         MC-C   &  23.4             &  19.4            &  18.4          \\     
         PD-C      &  28.7             &  28.2             &  18.5         \\     \midrule
         \bftab Mean $\downarrow$                      & \bftab 29.1             & \bftab 27.3             & \bftab 24.0         \\   
         \bottomrule
    \end{tabular}
   
         \label{tab:runtime_main}
\end{minipage}
\end{wraptable}
\paragraph{SYMPOL is efficient}%
In RL, the actor-environment interaction frequently constitutes a significant portion of the total runtime. For smaller policies, in particular, the runtime is mainly determined by the time required to execute actions within the environment to obtain the next observation, while the time required to execute the policy itself having a comparatively minimal impact on runtime. Therefore, recent research put much effort into optimizing this interaction through environment vectorization. The design of SYMPOL, in contrast to existing methods for tree-based RL, allows a seamless integration with these highly efficient training frameworks. As a result, the runtime of SYMPOL is almost identical to using an MLP or SDT as policy, averaging less than 30 seconds for 1mio timesteps (see Table~\ref{tab:runtime_main}). 
Detailed results including training curves are in Appendix~\ref{A:runtime_curves}. %

\begin{figure}[t]
    \centering
    \includegraphics[width=0.9\textwidth]{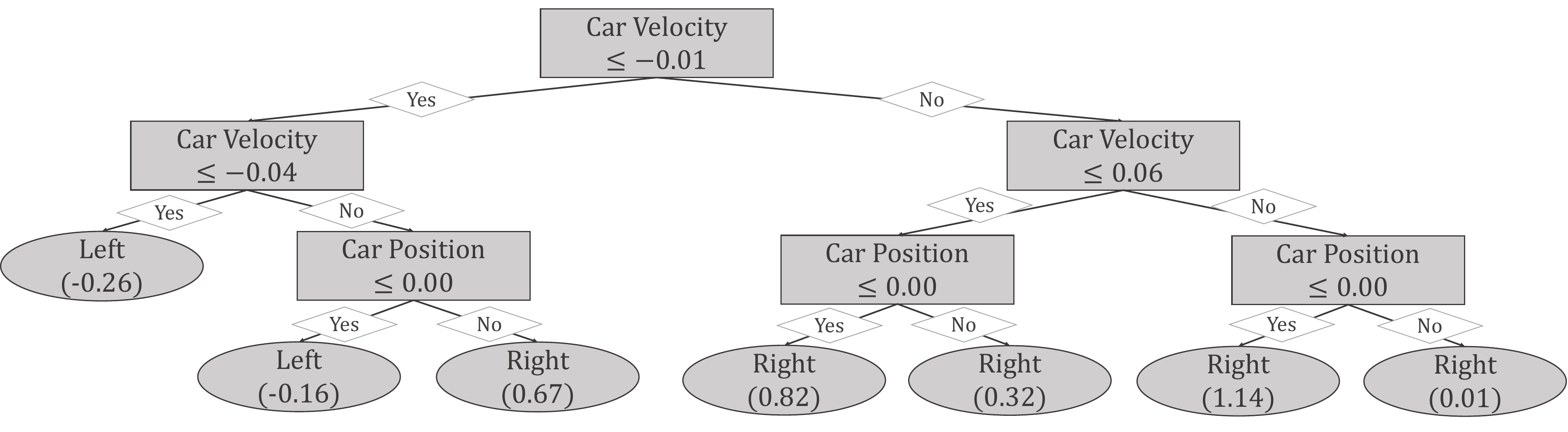}
    \caption[SYMPOL Policy for MC-C]{\textbf{SYMPOL Policy for MC-C.} The main rule encoded by this tree is that the car should accelerate to the left if its velocity is negative and to the right if it is positive, which essentially increases the speed of the car over time, making it possible to reach the goal at the top of the hill. The magnitude of the acceleration is mainly determined by the current position, reducing the cost of the actions.}
    \label{fig:sympol_policies}
\end{figure}
\paragraph{DT policies learned with SYMPOL are small and interpretable}%
While we trained SYMPOL with a depth of 7 and therefore 255 possible nodes, the effective tree size after pruning is significantly smaller with only $50.5$ nodes (internal and leaf combined) on average (see Table~\ref{tab:tree_size_sympol}). This can be attributed to a self-pruning mechanism that is inherently applied by SYMPOL in learning redundant paths during the training and therefore only optimizing relevant parts. Furthermore, DTs learned with SYMPOL are smaller than SA-DTs (d=5) with an average of $60.3$ nodes and significantly smaller than SA-DTs (d=8) averaging $291.6$ nodes. The pruned D-SDTs are significantly smaller with only $16.5$, but also have a very poor performance, as shown in the previous experiment.
An exemplary DT learned by SYMPOL is visualized in Figure~\ref{fig:sympol_policies}. Extended results, including a comparison with SDTs, are in Appendix~\ref{A:tree_size} and \ref{A:interpretability_comparison}.

\begin{table}[t]
    \centering
    \small
        \caption[Tree Size]{\textbf{Tree Size.} We report the average size of the learned DT for each environment.}
    \begin{tabular}{lrrrr}
    \toprule
                                    & SYMPOL (ours)     & D-SDT             & SA-DT (d=5)   & SA-DT (d=8)   \\ \midrule
         CP                   &   39.4            & 14.2              &  61.8         &  315.0        \\     
         AB                    &   78.6            & 17.0              &  56.5         &  173.0        \\     
         LL                &   55.0            & 19.8              &  59.8         &  270.2        \\
         MC-C   &   23.4            & \phantom{0}3.0    &  61.0         &  311.8        \\              
         PD-C      &   56.2            & 28.6              &  62.2         &  388.2        \\ \midrule  
         \bftab Mean $\downarrow$                      &   \bftab 50.5     & \bftab 16.5       &  \bftab 60.3  &  \bftab 291.6 \\      
         \bottomrule
    \end{tabular}
    \label{tab:tree_size_sympol}
\end{table}

\paragraph{Ablation study}%
In Section~\ref{sec:sympol}, we introduced several crucial components to facilitate an efficient and stable training of SYMPOL. To support the intuitive justifications for our modifications, we performed an ablation study to evaluate the relevance of the individual components. 
Therefore, we disabled individual components of our method and evaluated the performance without the specific component. This includes the following modifications:
\begin{enumerate}
    \item \textbf{w/o separate architecture:} Instead of using separate architectures for actor and critic, we use the same architecture and hyperparameters for the actor and critic.
    \item \textbf{w/o dynamic rollout:} We proposed a dynamic rollout buffer that increases with a stepwise exponential rate during training to increase stability while maintaining exploration early on. Here we used a standard, static rollout buffer.
    \item \textbf{w/o batch size adjustment:} Similar to the dynamic rollout buffer, we proposed using a dynamic batch size to increase gradient stability in later stages of the training. Here, we used  standard, static batch size.
    \item \textbf{w/o AdamW:} We introduced an Adam optimizer with weight decay to SYMPOL to support the adjustment of the features to split on and the class predicted. Here, we use a standard Adam optimizer without weight decay.
\end{enumerate}

\begin{figure}[t]
   \centering
   \includegraphics[width=0.75\textwidth]{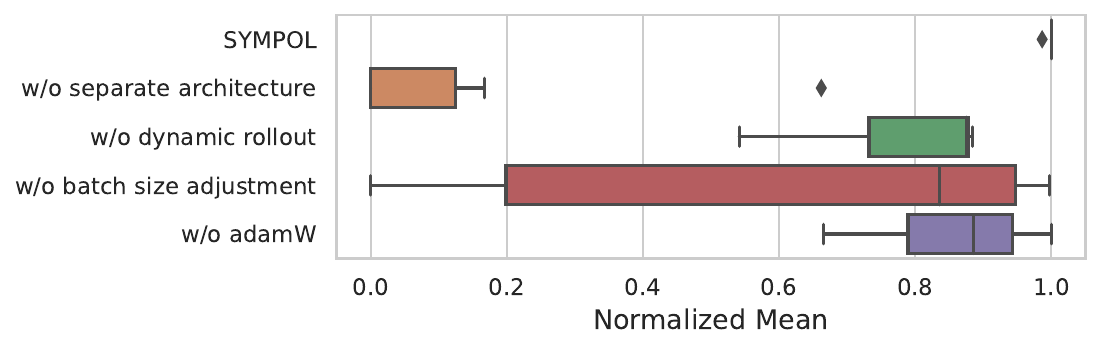}
   \caption[Ablation Study]{\textbf{Ablation Study.} We report the mean normalized performance over all control environments. Detailed results are reported in Table~\ref{tab:ablation}.}
   \label{fig:ablation}
\end{figure}

The results are visualized in Figure~\ref{fig:ablation} and detailed results for each of the control environments are reported in Table~\ref{tab:ablation}. 
The ablation study underscores the critical role of specific design choices in the performance of SYMPOL. The separate architecture emerged as the most important contributor, showing how important it is in learning accurate policies. Dynamic rollout and batch size further enhanced learning efficiency, showcasing their value in improving policy adaptation. While adjustments to the use of AdamW as the optimizer also had a positive impact on performance, the contribution was comparatively less pronounced. These findings emphasize that the synergy of architectural design and optimization dynamics is essential for the robust performance of SYMPOL.

\section{Case Study: Detecting Goal Misgeneralization} \label{sec:case_study}

To demonstrate the benefits of SYMPOLs enhanced transparency, we present a case study on goal misgeneralization \citep{di2022goal}. Good policy generalization is vital in RL, yet agents often exhibit poor out-of-distribution performance, even with minor environmental changes. Goal misgeneralization is a well-researched out-of-distribution robustness failure that occurs when an agent learns robust skills during training but follows unintended goals. This happens when the agent's behavioral objective diverges from the intended objective, leading to high rewards during training but poor generalization during testing. For instance, NNs were shown to systematically misgeneralize on Atari environments~\citep{farebrother2018generalization,delfosse2024hackatari}. \\

\begin{figure}[t]
    \centering
    \begin{minipage}[t]{0.445\linewidth}
        \centering
    \includegraphics[width=0.95\linewidth]{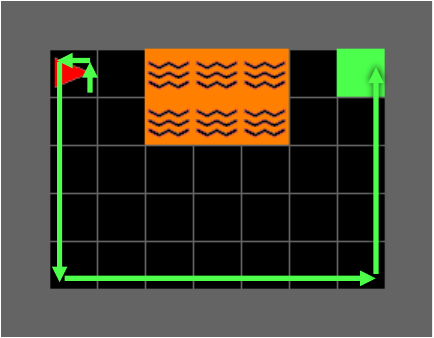}
    \caption[DistShift]{\textbf{DistShift.} We show the training environment for the agent along with the starting position and goal. The path taken by SYMPOL (visualized in Figure~\ref{fig:sympol_distshift}) is marked by green arrows and solves the environment.} %
    \label{fig:distshift_base}
    \end{minipage}%
    \hfill
    \begin{minipage}[t]{0.55\linewidth}

        \centering
        \includegraphics[width=0.75\linewidth]{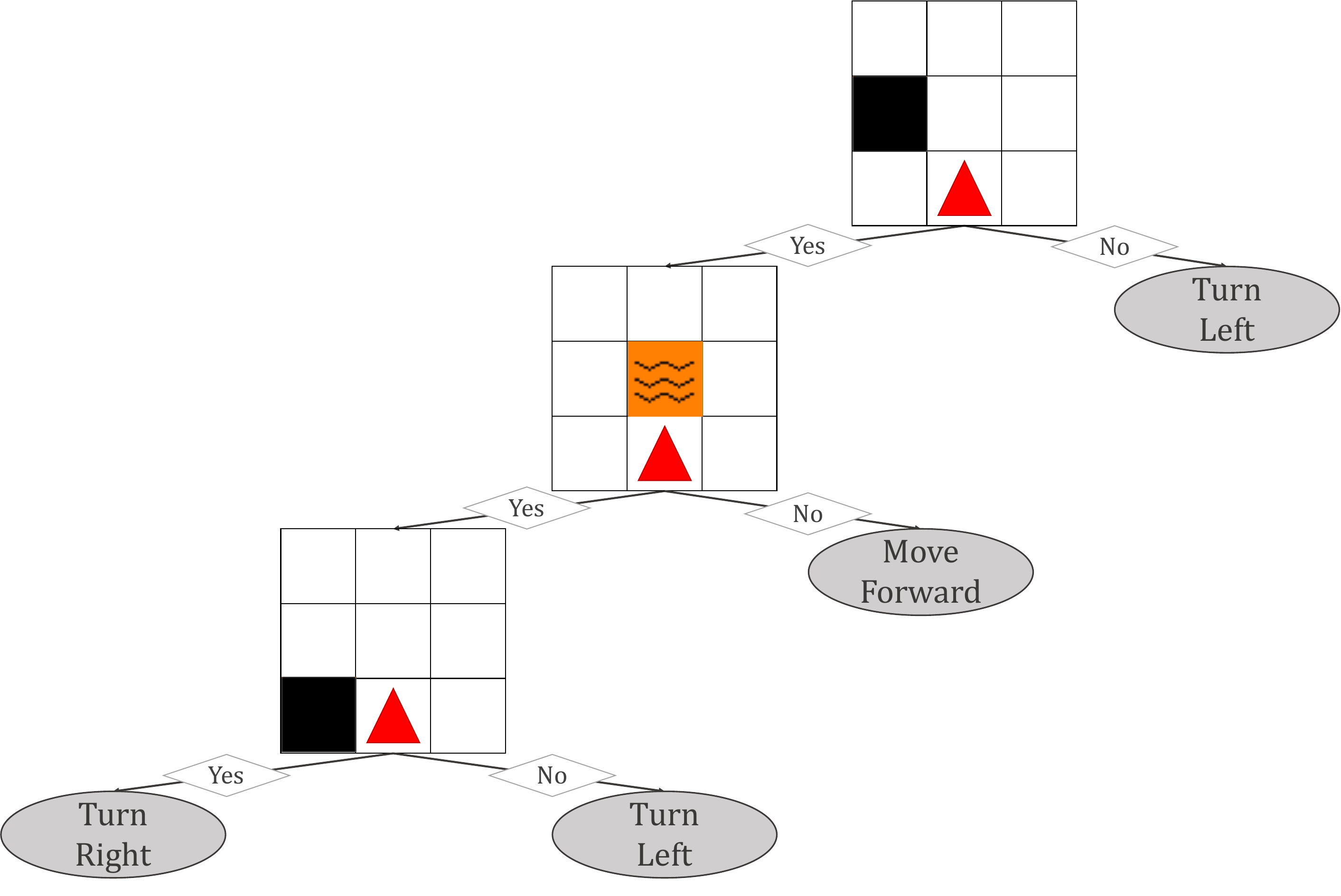}
        \caption[SYMPOL Policy]{\textbf{SYMPOL Policy.} This image shows the DT policy of SYMPOL. Split nodes are visualized as the 3x3 view grid of the agent with one square marking the considered object and position. If the visualized object is present at this position, the true path (left) is taken.}
        \label{fig:sympol_distshift}
    \end{minipage}
\end{figure}

To demonstrate that SYMPOL can help in detecting misaligned behavior, let us consider the DistShift environment from MiniGrid, shown in Figure~\ref{fig:distshift_base}. The environment is designed to test for misgeneralization~\citep{minigrid}, as the goal is repeatedly placed in the top right corner and the lava remains at the same position.
We can formulate the intended behavior according to the task description as avoiding the lava and reaching a specific goal location. SYMPOL, similar to other methods, solved the task consistently. %
The advantage of SYMPOL is the tree-based structure, which is easily interpretable.
When inspecting the SYMPOL policy (Figure~\ref{fig:sympol_distshift}), we can immediately observe that the agent has not captured the actual task correctly. Essentially, it has only learned to keep an empty space on the left of the agent (which translates into following the wall) and not to step into lava (but not to get around it). While this is sufficient to solve this exact environment, it is evident, that the agent has not generalized to the overall goal. 
\begin{figure}[t]
    \centering
    \begin{minipage}[t]{0.495\linewidth}
        \vspace{0pt} %
        \centering
        \includegraphics[width=0.8\linewidth]{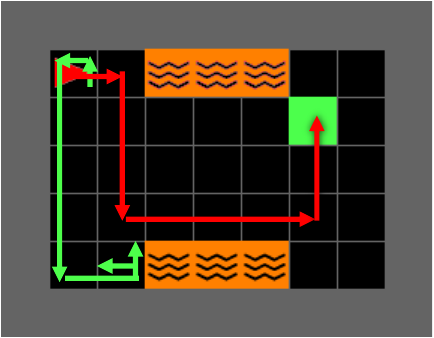}
        \caption[DistShift with Domain Randomization]{\textbf{DistShift with Domain Randomization.} This is a modified version of DistShift, with goal and lava at random positions. SYMPOL (Figure~\ref{fig:sympol_distshift}), visualized as the green line, is not able to solve this environment. Training SYMPOL with domain randomization (Figure~\ref{fig:sympol_distshift_random}), visualized as the red line, is able to solve the environment.}
        \label{fig:distshift_domain_randomization}
    \end{minipage}%
    \hfill
    \begin{minipage}[t]{0.5\linewidth}
        \vspace{0pt} %

        \centering
        \includegraphics[width=\linewidth]{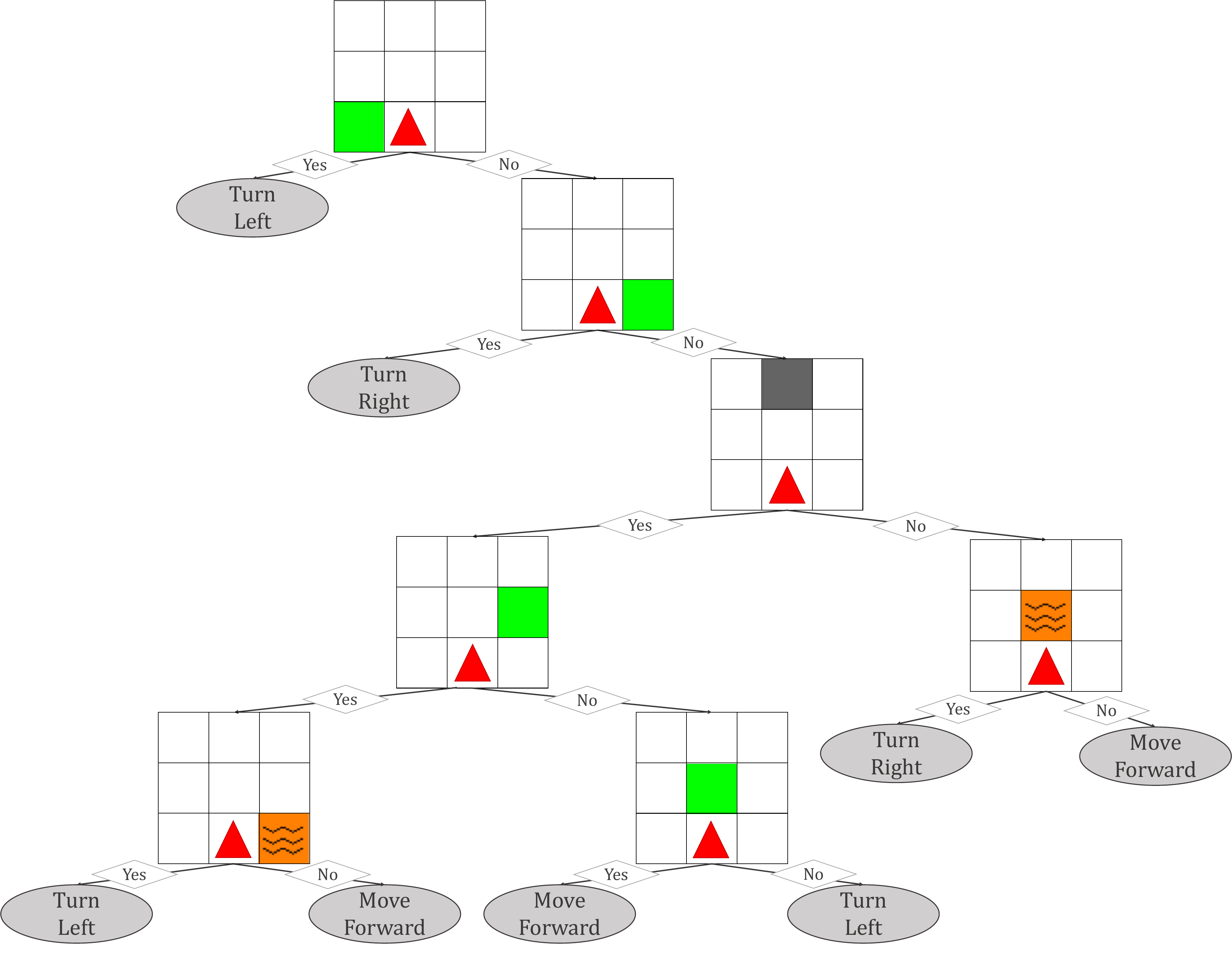}
        \caption[SYMPOL Policy with Domain Randomization]{\textbf{SYMPOL Policy with Domain Randomization.} The SYMPOL policy (Figure~\ref{fig:sympol_distshift_random}) retrained with domain randomization. The agent now has learned to avoid lava and walls, as well as identifying and walk into the goal.}
        \label{fig:sympol_distshift_random}
    \end{minipage}
\end{figure}
In order to test for misgeneralization, we created test environments in which the agent has to reach a random goal placed with lava placed at a varying locations. 
As already identified based on the interpretable policy, we can observe in Figure~\ref{fig:distshift_domain_randomization} that the agent gets stuck when the goal or lava positions change. %
Alternative non-interpretable policies exhibit the same behavior, which might remain unnoticed due to the black-box nature. Instead of simply looking at the learned policy with SYMPOL, alternative methods would require using external methods or designing complex test cases to detect such misbehavior. Alternative methods to generate DT policies like SA-DT also provide an interpretable policy, but as already shown during our experiments, frequently come with severe information loss. Due to this information loss, we cannot ensure that we are actually interpreting the policy, which is guaranteed using SYMPOL.
Based on these insights, we retrained SYMPOL with domain randomization. The resulting policy (Figure~\ref{fig:sympol_distshift_random}) now solves the randomized environments (Figure~\ref{fig:distshift_domain_randomization}), still maintaining interpretability.
In line with our results, \citet{delfosse2024interpretableCBM} showed that interpretable DT policies can mitigate goal misgeneralization, underscoring the potential benefit of interpretable RL policies for robust generalization.

\section{Summary}

In this chapter, we introduced SYMPOL, a novel method for tree-based RL. SYMPOL can be seamlessly integrated into existing on-policy RL algorithms, where the DT policy is directly optimized on-policy while maintaining interpretability. This direct optimization guarantees that the explanation exactly matches the policy learned during training, avoiding the information loss often encountered with existing methods that rely on post-processing to obtain an interpretable policy.
Furthermore, the performance of interpretable DT policies learned by SYMPOL is significantly higher compared to existing methods, particularly in environments involving more complex environment transition functions or randomness.
We believe that SYMPOL represents a significant step towards bridging the gap between the performance of on-policy RL and the interpretability and transparency of symbolic approaches, paving the way for the widespread adoption of trustworthy and explainable AI systems in safety-critical and high-stakes domains. \\

While we focused on an actor-critic, on-policy RL method, the flexibility of SYMPOL allows an integration into arbitrary policy optimization algorithms including off-policy methods like Soft Actor-Critic~\citep{sac} or Advantage Weighted Regression~\citep{peng2019awr}, which can be explored in future work. Also, it would be interesting to evaluate a more complex, forest-like tree structure as a performance-interpretability trade-off, similar to GRANDE (see Section~\ref{sec:method_grande}), especially based on the promising results of \citet{fuhrer2024gradient} for tree-based RL.

\chapter{Discussion, Conclusion and Future Work} \label{cha:conclusion}

In this thesis, we present gradient‐based decision trees as a novel framework that bridges traditional decision tree learning with modern gradient‐based optimization. By reformulating decision trees to support end‐to‐end gradient optimization while preserving their hard, axis‐aligned structure, our approach directly tackles multiple shortcomings of existing methods. It mitigates the constrained search space and local optimality inherent in greedy, sequential splitting, maintains the beneficial inductive biases and interpretability of axis‐aligned splits, and enhances flexibility by incorporating custom objective functions with seamless integration into advanced frameworks such as multimodal and reinforcement learning. Extensive theoretical analysis and empirical validation across diverse domains demonstrate that our method achieves a holistic optimization of tree parameters, offering a robust, scalable, and versatile alternative to conventional decision tree learning.
The following discussion (Section~\ref{sec:discussion}) examines the advantages (Section~\ref{ssec:advantages}) and challenges as well as limitations (Section~\ref{ssec:challenges}) of gradient-based DTs. In this context, we also discuss real-world implications (Section~\ref{ssec:implications}) and propose future research directions (Section~\ref{ssec:future_directions}) based on the work of this thesis. We conclude this thesis (Section~\ref{sec:conclusion}) by highlighting key contributions addressing the research questions formulated in the introduction (Section~\ref{sec:research_questions}). %

\section{Discussion} \label{sec:discussion}

In the previous chapters we have introduced and investigated how the novel concept of axis-aligned, hard, gradient-based DTs can be incorporated into different areas, yielding to advancements in performance and interpretability. 
In this chapter, we will discuss the advantages, limitations and broader impact of our work and findings.

\subsection{Advantages for Gradient-Based Decision Trees} \label{ssec:advantages}

\paragraph{Inductive bias of decision trees}
DTs provide a powerful inductive bias not only for tabular data but also in alternative scenarios such as reinforcement learning on categorical environments. The strength of this inductive bias can be attributed to two key components: The axis-aligned nature of the splits and the hierarchical structure of the trees:
\begin{itemize}
    \item \textbf{Inductive bias of axis-aligned splits} \hspace{1.5mm} 
The use of axis-aligned splits introduces several inherent advantages that enhance the effectiveness of gradient-based DTs.
First, they provide natural feature selection by considering one feature at a time, making the decision-making process transparent and interpretable~\citep{molnar2020}. This property is particularly valuable in domains where understanding the model's decisions is crucial. %
Second, axis-aligned splits are well-suited for capturing discontinuities and sharp transitions in the data, as they partition the input space into rectangular regions~\citep{hastie2009elements}. Many real-world problems naturally exhibit such piecewise behavior (e.g., a cutoff in blood pressure for a medical risk score), and by stacking multiple splits in a hierarchy, trees can flexibly approximate abrupt changes in the underlying function. This is supported by recent research that showed that axis-aligned splits provide a good inductive bias for irregular, high-frequency target functions, which are commonly present in tabular data~\citep{beyazit2024inductive,grinsztajn2022tree}.
Finally, the simplicity of axis-aligned splits helps prevent overfitting by limiting the model's capacity (when combined with standard regularization measures like limiting the tree depth) while still allowing it to capture complex patterns through hierarchical combination of splits. 

    \item \textbf{Inductive bias of hierarchical structure} \hspace{1.5mm} 
The hierarchical structure of DTs introduces another powerful inductive bias. 
This structure naturally aligns with a divide-and-conquer strategy~\citep{soft1}. It breaks down complex decision problems into simpler subproblems, enabling efficient learning and representation of nested decision rules.
At each node, the model focuses on a subset of the feature space, learning localized decision boundaries that are refined further down the tree. This systematic, step-by-step approach not only facilitates the efficient learning of simple patterns early on but also enables the capture of complex feature interactions at deeper levels of the tree.
Moreover, the structure provides transparency, as each path from the root to a leaf node represents a coherent sequence of decisions. This makes DTs particularly interpretable, as they align closely with human decision-making processes~\citep{hastie2009elements,kotsiantis2013decision}.

\end{itemize}

\paragraph{Joint model parameter optimization}
A significant advantage of our gradient-based approach is the ability to jointly optimize all tree parameters simultaneously, rather than relying on sequential, greedy decisions. As demonstrated in our experiments, this joint optimization can lead to better global solutions that might be missed by greedy approaches. Furthermore, our method shows increased robustness to overfitting compared to both greedy and alternative non-greedy methods (like optimal DTs). This robustness can be attributed to the mini-batch optimization process, which inherently introduces a form of regularization, and the ability to adjust all parameters simultaneously in response to new data. Moreover, the integration of tree-based and deep learning approaches offers an extensive repertoire of regularization techniques, including dropout, feature subset, and data subset selection which are used as hyperparameters in GRANDE.

\paragraph{Flexibility through gradient-based optimization}
A key advantage of our gradient-based approach lies in its flexibility. Unlike traditional DT methods that often require custom optimization procedures, our approach seamlessly integrates into existing gradient-based frameworks. This integration capability has been demonstrated across multiple domains, from incorporating a tree-based component into a multimodal learning architecture (Section~\ref{sec:multimodal}) to implementing interpretable policies in reinforcement learning algorithms (Section~\ref{cha:sympol}). The flexibility extends to the optimization process itself, where we can easily customize objectives and loss functions to suit specific requirements. This adaptability is particularly valuable in practical applications where standard loss functions may not adequately capture task-specific requirements. %

\paragraph{Leveraging advances in deep learning and gradient descent optimization} %
The proposed method can benefit significantly from ongoing advancements in deep learning and gradient-based optimization. This advantage is already evident in several ways %
, where the impact of adaptive learning rates has been particularly noteworthy. The dense representation of our trees, where only a subset of parameters is effectively utilized at any time, makes pure SGD nearly ineffective. Therefore, modern optimizers with momentum and adaptive learning rates have proven crucial for efficient training. Similarly, advances in GPU technology and optimization frameworks like TensorFlow and JAX have significantly reduced runtime requirements, with further improvements expected as hardware continues to evolve.
Looking forward, we can anticipate that our method will continue to benefit from new developments in optimization techniques, loss functions, and hardware acceleration. 

\subsection{Challenges and Limitations of Gradient-Based Decision Trees} \label{ssec:challenges}

\paragraph{Gradients can be uninformative}
Gradients, particularly those associated with split indices, can be uninformative. At best, they indicate whether the currently selected feature is suitable, but they cannot provide insight into which feature might be the next best candidate for a split. This limitation means that identifying the optimal or even a good feature frequently relies on trial and error, guided by observations and the information encoded in model parameters from previous gradient updates. Despite this challenge, our results demonstrate that our approach performs effectively in practice, especially in ensembles of trees. Notably, we achieved strong performance on datasets with over $1,000$ features. However, we see significant potential for improvement in this area. Refining the method for calculating split values or redefining the dense architecture could enhance both runtime (by accelerating the discovery of good splits) and overall model performance (by identifying better splits). 

\paragraph{Scalability}
While our method shows promising results, scalability remains a challenge, particularly along three dimensions: Number of features/classes, dataset size the tree depth. The dense representation we employ, while enabling gradient-based optimization, requires memory proportional to the number of features at each internal node and memory proportional to the number of classes at each leaf node. This can become problematic for high-dimensional datasets. For large datasets, although our method benefits from mini-batch processing, the computational overhead of maintaining and updating the dense representation can become significant. We have partially addressed these challenges through techniques like feature and data subsampling as well as an efficient, tensor-based implementation, but further optimization strategies may be needed for extremely large-scale applications. 
Finally, learning trees with depths exceeding 10 can present challenges, as the number of nodes (and consequently the model size) grows exponentially with tree depth. In ensemble methods like GRANDE, this is typically not an issue, since shallower trees are generally employed as weak learners. However, for individual trees, there may be specific cases where highly imbalanced structures are needed, such as when capturing a single complex rule. In such scenarios, deeper trees may be necessary to effectively represent these intricate patterns, which cannot be adequately addressed with the current approach.

\paragraph{Stability} As already observed in reinforcement learning, a global, gradient-based DT optimization can favor instability during the training process, especially in dynamic environments. This instability arises from the hierarchical nature of DTs, where changes at higher levels of the hierarchy can significantly impact subsequent decisions. For example, altering the split at the root node redistributes samples to different branches, rendering previously learned splits further down the tree potentially invalid. While this issue is less critical in practice, especially when DTs are used within ensemble methods, there remains room for improvement. Potential strategies can include modifying the dense representation of the tree, incorporating additional learnable parameters, or adjusting the training procedure, which could be explored in future work.

\paragraph{High cardinality}
Handling high-cardinality categorical features presents a particular challenge for our gradient-based approach. While traditional DTs can efficiently handle categorical features through techniques like purity calculation, our method currently requires categorical features to be encoded numerically. Standard encoding techniques like one-hot encoding can dramatically increase the feature dimensionality, intensifying the scalability issues discussed above. 
This limitation is particularly relevant for real-world applications where high-cardinality categorical features are common, such as user IDs in recommendation systems or ZIP codes in demographic analysis.

\paragraph{Alternative modalities}
Our current framework, while effective for structured data, faces challenges when dealing with alternative, unstructured data modalities such as images and text. The primary difficulty lies in combining our method with embedding-based approaches commonly used for these modalities. The parallel optimization of embeddings and tree parameters creates a moving target problem: As the embedding layer updates change the input representation, the tree's threshold-based splits may become less meaningful. One potential solution could be an alternating optimization approach, where we freeze one component while updating the other. However, this would require careful coordination of the learning process and might slow down convergence. An alternative would be to extract and freeze a pretrained embedding, which can then be used as tree input. However, the high-dimensional nature of embedded representations from images and text poses challenges for our current architecture. Future work could explore specialized architectures or optimization strategies for handling these alternative modalities while maintaining the interpretability benefits of our approach.

\subsection{Practical Implications: Real-World Application of Gradient-Based Trees} \label{ssec:implications}

The methods developed in this thesis have significant potential for real-world applications, particularly in domains where interpretability, efficiency, and integration with existing systems are crucial. Unlike deep learning architectures, which often operate as opaque black boxes, the proposed method offers distinct advantages in terms of interpretability due to its symbolic, tree-based nature. This inherent transparency makes the model easy to verify, a critical factor in applications where understanding and validating decision-making processes is essential. While our experimental evaluations on real-world benchmark datasets have demonstrated promising results, especially in the tabular domain, the transition from controlled evaluations to actual deployment scenarios represents an important next step. Real-world applications often introduce additional challenges not fully captured in benchmark datasets, such as missing values, data quality issues, concept drift, and complex operational constraints. Therefore, implementing our methods in production environments would provide valuable insights into their practical utility and likely reveal new areas for improvement and optimization. 
Despite these challenges, several application domains stand to benefit significantly from our approach. In the following, we will examine concrete application scenarios across different domains where the advantages of gradient-based DTs, particularly their interpretability and verifiability, could meaningfully impact real-world systems.

\paragraph{Tabular data applications in high-stake domains}
For instance in healthcare, GRANDE could be particularly valuable for clinical decision support systems~\citep{med1,med2,med3}, where both performance and interpretability are essential. In patient risk assessment, the method could process complex patient data while providing clear decision pathways that clinicians can verify and understand. The instance-wise weighting mechanism could help identify similar cases and their outcomes, providing additional context for medical decisions.
Financial applications, such as credit scoring~\citep{credit1} or fraud detection~\citep{fraud1}, represent another promising domain. Here, the combination of high performance and interpretability is crucial for regulatory compliance and fairness assessment. GRANDE's ability to handle complex patterns while maintaining interpretable or verifiable decisions could help balance performance requirements with the need for transparent decision-making processes.

\paragraph{Multimodal applications}
The ability to integrate gradient-based DTs into multimodal architectures opens up applications in areas like medical diagnosis, where both imaging data (e.g., X-rays, MRIs) and patient metadata need to be considered~\citep{hager2023best,cui2023deep}. Our approach could provide interpretable or verifiable decision processes for the tabular component while maintaining high overall system performance.
Similarly, in manufacturing quality control, where both visual inspection data and sensor readings need to be processed, our method could help create more transparent decision systems. %

\paragraph{Interpretable reinforcement learning applications}
Traffic light control systems~\citep{wiering2004intelligent} represent an ideal application for SYMPOL, where reinforcement learning needs to make interpretable decisions based on potentially categorical state information. The decisions about traffic light timing can be naturally represented through axis-aligned splits based on queue lengths, waiting times, and traffic density. The interpretability of the policy is crucial for validation by traffic engineers and for explaining system behavior to stakeholders. Similar applications exist in other urban planning scenarios, such as parking space management or public transportation routing.
These practical applications demonstrate how our methods can bridge the gap between high-performance machine learning and the interpretability or verifiability requirements of real-world systems. 

\subsection{Future Research Directions} \label{ssec:future_directions}
The previous sections have outlined several research directions inspired by the contributions of this thesis. However, there are still other paths worth exploring. In the following paragraphs, we will take a closer look at these potential areas for future research:

\paragraph{Online learning and adaptive systems}
The gradient-based nature of our approach makes it particularly suitable for online learning scenarios where models need to adapt to changing conditions~\citep{domingos2000mining,zhang2019incremental}. For example, in predictive maintenance systems, the model could continuously update its decision criteria based on new sensor data while maintaining interpretable decision rules~\citep{kaparthi2020designing}. This could help maintenance teams understand and validate the system's recommendations while benefiting from adaptive learning capabilities. Online learning with gradient-based DTs represents an unexplored area of research and offers a compelling direction for future investigation.

\paragraph{Model distillation and knowledge transfer}
Our gradient-based approach offers unique advantages for model distillation, particularly when transferring knowledge from complex neural networks to interpretable DTs~\citep{soft3}. Unlike traditional distillation methods that typically rely on hard labels or post-hoc tree fitting~\citep{molnar2020}, our framework can directly leverage soft probability distributions (through logits) from teacher models during the gradient-based optimization process. This capability enables more nuanced knowledge transfer, as the DT can learn from the teacher model's uncertainty and relative confidence across different predictions. For instance, in high-stakes domains like autonomous systems, complex neural networks that achieve high performance could be distilled into interpretable DTs that maintain much of the original model's performance while providing verifiable decision paths. Alternatively, if a single interpretable DT lacks the capacity to capture the model's complexity, our method enables distilling the neural network into a tree ensemble. While this ensemble may not be fully interpretable, it still facilitates systematic verification~\citep{bastani2018viper}.

\paragraph{Fine-tuning decision trees or decision tree ensembles}
Another interesting direction for future research is the fine-tuning of pre-trained DTs or DT ensembles using gradient-based optimization. The idea is to initialize the model with a DT learned through traditional methods like CART or with a DT ensemble like XGBoost. The dense representation introduced in this thesis can then be used to fine-tune the model parameters globally, allowing for a joint optimization of all tree parameters. In the case of DT ensembles, the fine-tuning process can also involve learning accurate weights for the individual trees in the ensemble.
The main challenge in this approach lies in finding an intelligent way to initialize the additional parameters in the dense representation. Therefore, methods for mapping the pre-trained tree structure to the dense representation while preserving the learned split and still allowing a stable optimization are required. Additionally, regularization techniques may be necessary to prevent overfitting during the fine-tuning process, especially when dealing with complex ensembles or limited fine-tuning data.
This procedure is particularly promising for adapting already trained models to new data collected over time, without the need for complete re-training. 
This makes the method especially suitable for scenarios where it is hard to store all collected data or when  computational resources are limited.
In addition to fine-tuning pre-trained trees, our approach can also be used to incorporate expert knowledge by initializing the model with predefined rules, allowing for a seamless integration of domain expertise into the optimization process.

\paragraph{Representation learning and unsupervised learning}
Another direction for future research is the application of gradient-based DTs to unsupervised and representation learning tasks~\citep{bengio2012deep}. While this thesis primarily focuses on supervised learning settings, the ability to learn meaningful representations from unlabeled data is crucial for many real-world applications where labeled data is scarce or expensive to obtain~\citep{bengio2012deep}.
Gradient-based DTs can be adapted to unsupervised learning by modifying the objective function to capture the inherent structure and patterns in the data. 
For example, we could train DTs to minimize the spatial density in each leaf using gradient descent, effectively creating a hierarchical, DT-based clustering algorithm.
Alternatively, the trees could be trained to minimize reconstruction error in an autoencoder framework~\citep{hinton2006reducing}, where the goal is to learn a compact representation of the input data that allows for accurate reconstruction. 
Representation learning~\citep{bengio2013representation} with gradient-based DTs could also be applied to transfer learning scenarios, where the learned representations are used as features for downstream tasks. By pre-training the trees on large unlabeled datasets and fine-tuning them on specific tasks, the models could benefit from the rich representations learned from the unsupervised data, leading to improved performance and sample efficiency.

\paragraph{Time series data}  %
Extending gradient-based DTs to handle time series data is another promising research direction. Time series data is prevalent in various domains, such as finance~\citep{zhang2024analyzing}, and presents unique challenges due to its temporal dependencies and potential non-stationarity.
To adapt gradient-based DTs for time series data, several modifications can be explored. %
One exciting possibility is the integration of recurrent components, such as recurrent neural networks (RNNs) or long short-term memory (LSTM) units, into the gradient-based DT framework. By combining the temporal modeling capabilities of recurrent components with the interpretability and efficiency of DTs, hybrid models can be developed that leverage the strengths of both approaches. This could lead to a new type of recurrent model, specifically, a recurrent DT, which, to the best of our knowledge, has not been explored so far. %

\section{Conclusion} \label{sec:conclusion}

In this concluding chapter, we revisit the main research questions that guided this dissertation. We summarize our findings, providing a comprehensive synthesis that addresses the potential of learning DTs with gradient descent and evaluates the broader implications for machine learning.

\paragraph{We can efficiently learn hard, axis-aligned decision trees with gradient descent}
The results presented in this dissertation indicate that it is indeed possible to efficiently learn hard, axis-aligned DTs using gradient-based optimization (\textbf{\ref{rq1}}). 
Our GradTree framework overcomes the non-differentiability of traditional DTs by introducing soft differentiable approximations and leveraging the ST operator. This addresses the non-differentiability challenge inherent to traditional DTs, which often hinders holistic optimization.
Our work demonstrates that joint optimization of all tree parameters (splitting thresholds, feature selection, and predictions in the leaf nodes) is feasible using gradient descent (\textbf{\ref{rq1.1}}). 
Specifically, the ST operator enables gradient propagation while preserving hard, axis-aligned splits, making the entire optimization process feasible with gradient descent.
This optimization approach enables a holistic search for decision boundaries without being constrained by the greedy, stepwise selection of traditional tree learning.
Consequently, our findings demonstrate that gradient-based optimization effectively mitigates the issue of local optimality arising from the constrained search space inherent in traditional DT algorithms (\textbf{\ref{rq1.2}}).
Conventional tree learning methods, such as CART and C4.5, rely on greedy, stepwise splitting procedures that make locally optimal decisions at each node, constraining the search space and often leading to suboptimal tree structures. 
By contrast, our approach enables a more holistic search of the decision space by jointly optimizing all model parameters, thereby preventing premature convergence to suboptimal configurations. While gradient descent does not inherently eliminate the risk of local optima given the non-convex nature of the optimization problem, our empirical evaluations show that alternative local optima encountered in GradTree often yield robust and interpretable models, similar to the behavior observed in neural network training. 
Gradient-based DT learning, while inherently iterative and potentially computationally expensive, was shown to be runtime efficient on moderately large datasets (\textbf{\ref{rq1.3}}). We implemented optimization strategies such as batching, parallelization, and the use of efficient tensor-based operations to improve runtime performance. The evaluations showed that although our method involves multiple epochs of parameter updates, these computational costs are mitigated by the ability to efficiently parallelize computations across nodes. Moreover, we demonstrated that our approach scales effectively with data size and feature count, making it suitable for more complex problems. The experimental results confirmed that the runtime of our gradient-based approach is comparable to and sometimes better than existing non-greedy algorithms like optimal DTs, particularly for deeper trees where holistic optimization brings greater benefits.
The interpretability of gradient-based DTs was a crucial aspect of our investigation (\textbf{\ref{rq1.4}}). Our empirical results show that the learned trees maintain the intuitive, axis-aligned decision boundaries that make traditional DTs interpretable. The axis-aligned and hard splits enforced during optimization ensured that the decision boundaries remain clear and comprehensible, with each decision node only depending on one feature. This results in models that are easy to understand and analyze. 

\paragraph{Gradient-based decision trees can be extended to a performant ensemble}
The second major research question focused on the feasibility of extending gradient-based DTs to form GRANDE, a gradient-based DT ensemble (\textbf{\ref{rq2}}). 
By leveraging weighted tree ensembles with differentiable operations discretized using the ST-operator, we achieved robust performance across various complex datasets. 
Building on the results of individual gradient-based DTs, we investigated whether transitioning to an ensemble structure would yield a substantial performance gain, similar to the well-established improvements seen when extending decision trees to random forests (\textbf{\ref{rq2.1}}). Our results confirmed that ensembling enhances the predictive power of gradient-based DTs while retaining key advantages of axis-alignment and flexibility. In particular, GRANDE's instance-specific weighting mechanism dynamically optimized the contribution of each tree, leading to improved generalization and robustness across different datasets.
The development of GRANDE showed that it is possible to formulate an entire ensemble in a manner that supports efficient tensor-based operations (\textbf{\ref{rq2.2}}). Through careful design of the ensemble structure and parallel computation of trees, we ensured that the model remained computationally efficient and scalable, especially with number of features and trees. 
The use of gradient-based optimization allowed us to incorporate a sophisticated, instance-specific weighting technique within the ensemble, improving model performance and interpretability (\textbf{\ref{rq2.3}}). Thereby, we enhanced the ensemble's ability to focus on different regions of the feature space more effectively. This dynamic weighting mechanism, which was jointly learned during training, allowed the ensemble to achieve better performance by emphasizing trees that were more specialized for specific subsets of data, ultimately leading to improved generalization and interpretability.

\paragraph{We can seamlessly integrate gradient-based decision trees into existing approaches across multiple domains}
Our research extended gradient-based DTs to new domains, including multimodal learning and reinforcement learning (\textbf{\ref{rq3}}). Our approach bridges the gap between tree-based methods and modern gradient-based optimization techniques, enabling seamless integration into broader machine learning architectures and frameworks. By incorporating our method into established architectures, we demonstrated their versatility and effectiveness in improving model performance, interpretability, and computational efficiency across diverse applications.
The integration of gradient-based DTs as a tabular component into a multimodal architecture showed promising results (\textbf{\ref{rq3.1}}). In experiments combining tabular data with image data, gradient-based DTs enabled an end-to-end training process that enhanced the overall model performance. Specifically, the tree-based structure was used as the tabular component within a multimodal architecture, where it interacted seamlessly with other neural layers processing image data. This integration led to improved performance, as the gradient-based DTs were able to effectively capture structured information from tabular data while benefiting from joint training with other modalities.
The feasibility of integrating gradient-based DTs into reinforcement learning algorithms was also investigated (\textbf{\ref{rq3.2}}). We specifically investigated SYMPOL, which demonstrated that DTs can be successfully used as interpretable policies in an on-policy reinforcement learning setting, providing significant advantages over black-box neural policies in terms of transparency and trustworthiness. We successfully integrated DTs as policy representations in existing reinforcement learning algorithms, such as PPO. This integration enabled interpretable policies to be directly optimized using policy gradients, maintaining a high level of performance while providing a transparent decision-making process. 
Notably, the direct optimization of DTs as policy using our method comes without information loss, which is a distinct advantage over existing methods for tree-based RL. \\

In summary, this dissertation has successfully demonstrated the feasibility of learning axis-aligned DTs and their ensembles using gradient-based optimization. We have shown that this approach addresses key challenges in traditional DT learning, such as non-differentiability and limited holistic optimization. The extensions to ensembles and integration into broader frameworks indicate a wide applicability for these methods, with considerable advantages in performance, efficiency, and interpretability. The contributions of this work lay a foundation for further exploration into gradient-based learning of discrete structures, bridging the gap between classic machine learning and modern gradient-based optimization.

\appendix %

\chapter{Notation} %

\label{A:notation} %

This appendix provides a detailed overview of the notation used throughout this dissertation. It outlines the symbols and conventions employed for sets, scalars, vectors, matrices, tensors, and operations, ensuring clarity and consistency. The notational choices align with standard practices in machine learning and optimization literature to maintain readability and precision.

\section{Notation Principles}

This dissertation follows common notation conventions while maintaining consistency with the machine learning literature. We use different typefaces and letter cases to distinguish between different mathematical objects:

\begin{itemize}
    \item \textbf{Scalars} are denoted by lowercase letters (e.g., $n$, $d$, $k$), lowercase Greek letters (e.g., $\tau$, $\iota$) or lowercase Fraktur letters (e.g., $\mathfrak{i}$, $\mathfrak{p}$) for special cases, where scalars are the output of a function.
    \item \textbf{Vectors} are denoted by bold lowercase letters (e.g., $\mathbf{x}$, $\mathbf{y}$) or bold lowercase Greek letters (e.g., $\boldsymbol{\tau}$, $\boldsymbol{\iota}$).
    \item \textbf{Matrices} are denoted by bold uppercase letters (e.g., $\mathbf{I}$, $\mathbf{T}$, $\mathbf{W}$)
    \item \textbf{Tensors} of more than three dimensions are denoted by bold, slanted, capital, sans-serif letters (e.g., $\tens{L}$, $\tens{T}$, $\tens{I}$).
    \item \textbf{Sets} are denoted by uppercase letters (e.g., $S$, $A$, $X$) or standard number sets (e.g., $\mathbb{R}$, $\mathbb{N}$).
    \item \textbf{Functions} are denoted by lowercase letters (e.g., $t$, $g$), often using a calligraphic font for special functions (e.g., $\mathbb{L}$, $\mathcal{V}$, $\mathcal{Q}$). A more detailed summary of relevant special functions is provided in Table~\ref{tab:math_func}.
    \item \textbf{Operations} are denoted by standard operators with exceptions or conflicts explicitly noted (e.g., $\sum$, $\prod$, $\Delta$, $\bigodot$). A more detailed list of relevant special functions is provided in Table~\ref{tab:math_op}.
\end{itemize}

We made these choices to align with standard notation while preserving readability and clarity. In cases where notation conventions might conflict, we prioritize consistency with the machine learning literature to ensure our work is accessible to researchers in the field.

\newpage

\section{Detailed Notation of Operations and Functions }

\begin{table}[h]
\centering
\caption[Relevant Mathematical Operations]{\textbf{Relevant Mathematical Operations}}
\label{tab:math_op}
\begin{tabular}{|c|l|}
\hline
\textbf{Operation} & \textbf{Explanation} \\ \hline
$\sum$ & Summation \\ \hline
$\prod$ & Product \\ \hline
$\cdot$ & Dot product \\ \hline
$\otimes$ & Outer product (Kronecker product) \\ \hline
$\odot$ & Element-wise multiplication (Hadamard product) \\ \hline
$\bmod$ & Modulo operation (remainder after division) \\ \hline
$\times$ & Cartesian product \\ \hline
$\rightarrow$ & Maps to \\ \hline
$\lfloor \cdot \rceil$ & Rounding \\ \hline
$\cup$ & Set union \\ \hline
$\cap$ & Set intersection \\ \hline
$\in$ & Set membership \\ \hline
$\subseteq$ & Subset \\ \hline
\end{tabular}
\end{table}

\vspace{2cm}

\begin{table}[h]
\centering
\caption[Relevant Functions]{\textbf{Relevant Functions}}
\label{tab:math_func}
\begin{tabular}{|c|l|}
\hline
\textbf{Function} & \textbf{Explanation} \\ \hline
$\mathbb{E}[\cdot]$ & Expected value \\ \hline
$\mathbb{L}(\cdot)$ & Indicator function \\ \hline
$\nabla$ & Gradient \\ \hline
$\mathrm{clip}(\cdot)$ & Clipping function \\ \hline
$\mathrm{min}(\cdot)$ & Minimum value \\ \hline
$\mathrm{exp}(\cdot)$ & Exponential function \\ \hline
$\log(\cdot)$ & Logarithm function \\ \hline
$\sigma(\cdot)$ & Softmax function \\ \hline

$S(\cdot)$ & Sigmoid activation function \\ \hline
$\mathrm{entmax}(\cdot)$ & Entmax activation function \\ \hline
$\mathrm{Heaviside}(\cdot)$ & Heaviside step function \\ \hline
$\mathrm{softsign}(\cdot)$ & Softsign activation function \\ \hline
\end{tabular}
\end{table}

\chapter[Methods for Greedy, Purity-Based Decision Tree Induction]{Comparing Methods for Greedy, Purity-Based Decision Tree Induction} \label{A:greedy_dt}

Starting with the publication of the first DT learning algorithm in 1963~\citep{morgan1963problems} the task of learning a DT from data remained well-researched until today. 
The most prominent and still frequently used DT learning algorithms, namely C4.5~\citep{quinlan1993c45} (Section~\ref{sec:c45}) as an extension of ID3.0~\citep{quinlan1986_id30} (Section~\ref{sec:id3}) and CART~\citep{cart_breiman1984} (Section~\ref{sec:cart}), date back to the 1980s. %
They all follow a greedy procedure to learn a DT and build onto \emph{Hunt's algorithm} as a basis for DT induction. 
In general, Hunt's algorithm, ID3 and C4.5 all allow for non-binary splits, i.e., more than two branches from a split node. In the following, we only focus on binary DTs for simplicity and to stay consistent with the remaining algorithms that will be introduced. 
Although some algorithms are specifically designed for regression tasks, all approaches can theoretically be extended to handle regression, even though they require varying degrees of modification. For simplicity, however, the following discussion will focus on classification.
In general, the induction of purity-based DTs relies on an impurity measure $I$ to determine the next split. The specific impurity measure employed can significantly influence the selection of the next split, and varies across existing methods. In this chapter, we focus solely on greedy methods and compare existing greedy approaches for DT induction.

\subsection*{Hunt's Algorithm}
\paragraph{Algorithm overview} Hunt's algorithm~\citep{hunt1966experiments} is a recursive method used to build DTs by progressively partitioning a set of training records into subsets that are increasingly homogeneous with respect to a target class. This algorithm is fundamental in the construction of DTs, which are widely used for classification tasks. Hunt's algorithm is defined recursively through the following steps \citep{Tan2018}:
\begin{itemize}[leftmargin=2cm]
    \item[Step 1:] If all the records in the current subset belong to the same class \(y_i\), then the current node \(v\) is a leaf node labeled with the class \(y_i\). %
    \item[Step 2:] If the subset contains records belonging to more than one class, an attribute test condition is selected to partition the records into smaller subsets. Each outcome of the test condition corresponds to a child node, and the records are distributed among these children based on the test condition's outcomes.
\end{itemize}

Hunt's algorithm is then recursively applied to each child node. %

\paragraph{Stopping criteria} This procedure is repeated until a stopping criterion is met.  Stopping criteria are:
\begin{itemize}
    \item All leaf nodes are homogeneous.
    \item There are no more instances remaining.
    \item There are no more features to split on.
    \item The maximum recursion depth is reached (optional).
    \item The remaining features would yield an increase in impurity\footnote{This special case is not explicitly discussed, but selecting a split resulting in an increase in impurity would contradict the motivation of using an impurity measure.}.
\end{itemize}

\subsection*{Example 1a: Constructing a Decision Tree using Hunt's Algorithm}
Here, we apply Hunt's algorithm to the dataset from Table~\ref{tab:titanic_2d_cat}. The splitting criterion in this example is based on the fraction of passengers belonging to each class (\textit{Yes} or \textit{No}), i.e., we use the misclassification error ($I_{\text{Mis}}$) as the most simple impurity measure defined as %

\begin{equation} \label{eq:emis}
\small{
I_{\text{Mis}}(\text{$TP$}, \text{$FN$}, \text{$TN$}, \text{$FP$}) = 1 - \max \left( \frac{\text{$TP$}}{\text{$TP$} + \text{$FN$} +  \text{$TN$} + \text{$FP$}}, \frac{\text{$TN$}}{\text{$TP$} + \text{$FN$} + \text{$TN$} + \text{$FP$}} \right).
}
\end{equation}

Subsequently, the decrease in misclassification error can be defined as:

\begin{equation} \label{eq:emis_weightedsum}
\small{
\Delta I_{\text{Mis}}(v|\mathbb{S}_v) = I_{\text{Mis}}(\text{TP}_{v}, \text{TN}_{v}, \text{FP}_{v}, \text{FN}_{v}) - \frac{1}{k} \sum_{i=0}^{k}  I_{\text{Mis}}(\text{TP}_{v,i}, \text{TN}_{v,i}, \text{FP}_{v,i}, \text{FN}_{v,i}),
}
\end{equation}

where $\mathbb{S}_v$ is the split function at node $v$, $k$ is the number of splits ($k=2$ for binary trees), $\text{TP}$ are true positives, $\text{TN}$ are true negatives, $\text{FP}$ are false positives and $\text{FN}$ are false negatives.
To construct a DT with Hunt's algorithm, we then proceed as follows:
\begin{enumerate}
    \item \textbf{Calculate the misclassification error for the whole dataset} \hspace{0.25cm}
    As the majority of passengers (13/20) did not survive, the root node is initially labeled as \textit{No}.
    Calculate the fraction of passengers belonging to each class (Yes or No) in the entire dataset:

    \[
        I_{\text{Mis}}(0, 13, 0, 7) = 1 - \max \left( \frac{0}{20}, \frac{13}{20} \right) = 0.350 
        \quad \triangleright \text{Eq.~\ref{eq:emis}} 
    \]    

    \item \textbf{Calculate decrease in misclassification error for each potential split}
    
        \textbf{Information Gain for \emph{Fare Category}} \hspace{0.25cm} 
        The \emph{Fare Category} attribute has two categories: Low and High. We calculate the entropy for each subset.
        
        \begin{itemize}
            \item \textbf{Low Fare}: 11 passengers, 4 Survived (Yes), 7 Did Not Survive (No)
        \end{itemize}
        
        \begin{align*}\small
        I_{\text{Mis}}(0, 7, 0, 4) = 1 - \max \left( \frac{0}{11}, \frac{7}{11} \right) \approx 0.363 
        \quad \triangleright \text{Eq.~\ref{eq:emis}} 
        \end{align*}
        
        \begin{itemize}
            \item \textbf{High Fare}: 9 passengers, 3 Survived (Yes), 6 Did Not Survive (No)
        \end{itemize}
        
        \begin{align*}\small
        I_{\text{Mis}}(0, 6, 0, 3) = 1 - \max \left( \frac{0}{9}, \frac{6}{9} \right) \approx 0.333        
        \quad \triangleright \text{Eq.~\ref{eq:emis}} 
        \end{align*}

        \begin{align*}\small
        \Delta I_{\text{Mis}}(0|\text{Fare Category}) = 0.350 - \frac{0.363+0.333}{2} = 0.002 
        \quad \triangleright \text{Eq.~\ref{eq:emis_weightedsum}} 
        \end{align*}

        \textbf{Information Gain for \emph{Age}} \hspace{0.25cm}
        The \emph{Age} attribute has two categories: Young and Old. We calculate the entropy for each subset:
        
        \begin{itemize}
            \item \textbf{Young}: 4 passengers, 3 Survived (Yes), 1 Did Not Survive (No)
        \end{itemize}
        
        \begin{align*}\small
        I_{\text{Mis}}(3, 0, 1, 0) = 1 - \max \left( \frac{3}{4}, \frac{0}{4} \right) = 0.250        
        \quad \triangleright \text{Eq.~\ref{eq:emis}} 
        \end{align*}

        \begin{itemize}
            \item \textbf{Old}: 16 passengers, 4 Survived (Yes), 12 Did Not Survive (No)
        \end{itemize}
        
        \begin{align*}\small
        I_{\text{Mis}}(0, 12, 0, 4) = 1 - \max \left( \frac{0}{16}, \frac{12}{16} \right) = 0.250        
        \quad \triangleright \text{Eq.~\ref{eq:emis}} 
        \end{align*}

        \begin{align*}\small
        \Delta I_{\text{Mis}}(0|\text{Age Category}) = 0.350 - \frac{0.250+0.250}{2} = 0.100        
        \quad \triangleright \text{Eq.~\ref{eq:emis_weightedsum}} 
        \end{align*}

    \item \textbf{Select the best split for the current node} \hspace{0.25cm}
    To decide the best attribute to split, we choose the attribute that results in the subsets with the greatest decrease of the misclassification error (i.e., and increase in purity). In this case, splitting on \textit{Age Category} results in the greatest purity increase ($0.100$ > $0.002$)

    \item \textbf{Recursive partitioning} \hspace{0.25cm}
    The DT  branches based on the \emph{Age Category} attribute and applies Hunt's algorithm recursively as detailed above until one of the previously defined stopping criteria is reached.

\end{enumerate}

\begin{figure}[t]
    \centering
    \includegraphics[width=0.5\textwidth]{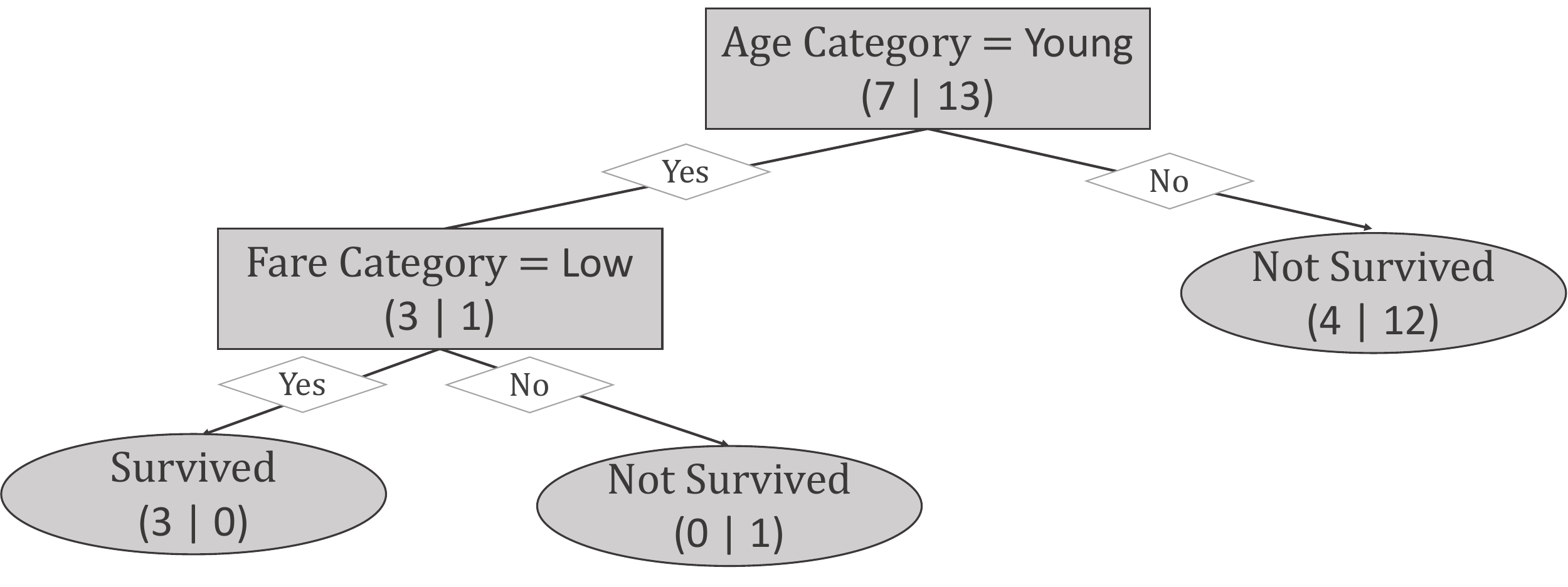}
    \caption[DT learned with Hunt's Algorithm]{\textbf{DT learned with Hunt's Algorithm and Misclassification Error as Impurity Measure.} The tree achieves an accuracy of $80\%$ on the training data and involves two decisions. The main rule encoded by the tree is that a passenger is predicted to survive if he is young and has paid a low fare, and is unlikely to survive otherwise. The right branch of the root does not split further, as the misclassification error does not decrease when splitting on fare (0.250 before and after).}
    \label{fig:decision_tree_hunt}
\end{figure}

This procedure results in the DT visualized in Figure~\ref{fig:decision_tree_hunt}. The main rule encoded by the tree is that a passenger is predicted to survive if he is young and has paid a low fare, and is unlikely to survive otherwise. Also, we can see that the split on the \emph{Fare Category} in the second level only separates a single sample into the right branch. While this significantly increases the purity, the decision is potentially not very robust and might originate from overfitting to the training data.

\paragraph{Limitations}  However, Hunt's algorithm can face challenges in certain special cases \citep{Tan2018}:
\begin{itemize}
    \item \textbf{Empty Child Nodes:} Sometimes, a test condition may create a child node with no associated records. This can occur if no instance in the training set match a particular combination of attributes. %
    \item \textbf{Identical Attribute Values:} If all records in a subset have identical values for the attributes but different survival outcomes, further splitting is impossible. %
    \item \textbf{Balanced Class Distribution:} If a node is assigned a leaf node, but the purity of all classes is equal, it is not possible to clearly assign a class to the leaf node.
\end{itemize}

\paragraph{Extensions} Hunt's algorithm is a fundamental technique for DT construction, serving as the basis for more advanced methods. These methods typically improve the algorithm by appropriately handling special cases. Furthermore, in designing an advanced algorithm for DT induction, two essential aspects must be addressed \citep{Tan2018}:
\begin{itemize}
    \item \textbf{Attribute Selection:} At each step of tree growth, the algorithm must determine which attribute to use for splitting the records. The algorithm must include a method to specify test conditions suitable for various attribute types, and an objective measure to assess the quality of each test condition.
    \item \textbf{Stopping Criteria:} The algorithm must also define when to stop growing the tree. While expanding a node until all records belong to the same class or have identical attributes is a sufficient stopping criterion, additional criteria can be used to terminate the tree-building process earlier. Early stopping or post-hoc pruning can be beneficial in preventing the model from overfitting and reduce the runtime.
\end{itemize}
Essentially, these aspects are tackled by advanced algorithms for DT induction in several ways resulting in distinct characteristics, as we will show in the following subsections.

\section{Iterative Dichotomiser 3 and C4.5 Algorithms}

One of the most commonly used methods for DT induction is C4.5 developed by \citet{quinlan1993c45}. C4.5 is an extension of Iterative Dichotomiser 3 (ID3) introduced by \citet{quinlan1986_id30}. Both are grounded in the basic idea of Hunt's algorithm and will be introduced more detailed in the following.

\subsection*{Iterative Dichotomiser 3} \label{sec:id3}
The ID3 method for DT induction is a fundamental algorithm in machine learning, particularly in the construction of DTs for classification tasks. ID3 utilizes a top-down, greedy approach to induce a DT from a given set of examples. The core idea behind ID3 is to use entropy and information gain as the criteria for selecting the attribute that best classifies the examples at each step of the tree-building process. In the following, we will summarize the key aspects of ID3 based on \citet{quinlan1986_id30}.

\paragraph{Entropy and information gain}
Entropy is a measure derived from information theory, representing the amount of uncertainty or impurity in a set of examples. For a binary classification problem, where the dataset 
\(
    \mathcal{D} = \left\{ (\boldsymbol{x}^{j}, y^{j}) \mid \boldsymbol{x}^{j} \in \mathcal{X}, y^{j} \in \mathcal{Y}, i = 0, 1, \dots, m-1 \right\}
\)
contains \( \text{P} \) positive labels (i.e., \( y_i = 1 \)) and \( \text{N} \) negative labels (i.e., \( y_i = 0 \)), the entropy is defined as:

\begin{equation} \label{eq:entropy}
I_{\text{Entropy}}(\text{P}, \text{N}) = - \frac{\text{P}}{\text{P}+\text{N}} \log_2 \left(\frac{\text{P}}{\text{P}+\text{N}}\right) - \frac{\text{N}}{\text{P}+\text{N}} \log_2 \left(\frac{\text{N}}{\text{P}+\text{N}}\right).
\end{equation}

The information gain is the reduction in entropy achieved by partitioning the data at node \( v \) with a split \( \mathbb{S}_v \). Let \(\text{P}_{v,i}\) and \(\text{N}_{v,i}\) denote the number of positive and negative instances by splitting at node $v$ into $k$ subsets $R_1,\dots,R_k$ where \(i \in \{1,...,k\}\). The expected entropy after partitioning based on \( \mathbb{S}_v \), denoted as \(E(v)\), is given by:

\begin{equation} \label{eq:expected_entropy}
E(v|\mathbb{S}_v) = \sum_{i=0}^{k-1} \frac{\text{P}_{v,i} + \text{N}_{v,i}}{\text{P}_{v} + \text{N}_{v}} I_{\text{Entropy}}(\text{P}_{v,i}, \text{N}_{v,i}).
\end{equation}

The information gain for attribute \(A\) is then:

\begin{equation} \label{eq:information_gain}
\text{Gain}(v|\mathbb{S}_v) = I_{\text{Entropy}}(\text{P}_v, \text{N}_v) - E(v|\mathbb{S}_v).
\end{equation}

ID3 selects the attribute with the highest information gain to split the current set of examples.

\paragraph{Iterative nature and efficiency}
In addition to the previously explained procedure, ID3 employs an iterative approach by initially selecting a random subset of the training data, called a window, to construct the DT. The tree is then tested on the remaining examples in the training set. If the tree correctly classifies all examples, the process terminates; otherwise, misclassified examples are added to the window, and the tree is rebuilt. This iterative method is computationally more efficient than constructing a tree from the entire training set at once, particularly for large datasets with many attributes and examples. However, it does not guarantee convergence to a final tree unless the window can grow to include the entire training set.

\subsection*{Example 1b: Constructing a Decision Tree using ID3}

Again, consider the simplified version of the Titanic dataset with discretized features (Table~\ref{tab:titanic_2d_cat}), which is a necessary preprocessing step for ID3.
In the following, we will give an example for a step-by-step construction of the root node for the DT. This process is iteratively repeated for each node, following the exact same procedure. For simplicity, we will use the entire dataset instead of a window from the beginning here.

\begin{enumerate}

    \item \textbf{Calculate the entropy for the whole dataset} \hspace{0.25cm}
    The initial entropy \(I_{\text{Entropy}}(p, n)\) is calculated based on the proportion of passengers who survived (Yes) and who did not survive (No):
    
    20 passengers, 7 Survived (Yes), 13 Did Not Survive (No)

    \begin{align*}\small
    I_{\text{Entropy}}(7, 13) = -\frac{7}{20} \log_2 \left(\frac{7}{20}\right) - \frac{13}{20} \log_2 \left(\frac{13}{20}\right) \approx 0.934
    \quad \triangleright \text{Eq.~\ref{eq:entropy}} 
    \end{align*}
    
    \item \textbf{Calculate information gain for each potential split}
    
        \textbf{Information Gain for \emph{Fare Category}} \hspace{0.25cm}  
        The \emph{Fare Category} attribute has two categories: Low and High. We calculate the entropy for each subset.
        
        \begin{itemize}
            \item \textbf{Low Fare}: 11 passengers, 4 Survived (Yes), 7 Did Not Survive (No)
        \end{itemize}
        
        \begin{align*}\small
        I_{\text{Entropy}}(4, 7) = -\frac{4}{11} \log_2 \left(\frac{4}{11}\right) - \frac{7}{11} \log_2 \left(\frac{7}{11}\right) \approx 0.946
        \quad \triangleright \text{Eq.~\ref{eq:entropy}} 
        \end{align*}
        
        \begin{itemize}
            \item \textbf{High Fare}: 9 passengers, 3 Survived (Yes), 6 Did Not Survive (No)
        \end{itemize}
        
        \begin{align*}\small
        I_{\text{Entropy}}(3, 6) = -\frac{3}{9} \log_2 \left(\frac{3}{9}\right) - \frac{6}{9} \log_2 \left(\frac{6}{9}\right) \approx 0.918 
        \quad \triangleright \text{Eq.~\ref{eq:entropy}} 
        \end{align*}
        
        The expected entropy after splitting by \emph{Fare Category}is:
        
        \begin{align*}\small
        E(0 | \text{Fare Category}) = \frac{11}{20} \cdot 0.946 + \frac{9}{20} \cdot 0.918 \approx 0.933
        \quad \triangleright \text{Eq.~\ref{eq:expected_entropy}} 
        \end{align*}
        
        The information gain for \emph{Fare Category}is:
        
        \begin{align*}\small
        \text{Gain}(0 | \text{Fare Category}) = 0.934 - 0.933 = 0.001
        \quad \triangleright \text{Eq.~\ref{eq:information_gain}} 
        \end{align*}
        
        \textbf{Information Gain for \emph{Age}} \hspace{0.25cm}
        The \emph{Age} attribute has two categories: Young and Old. We calculate the entropy for each subset:
        
        \begin{itemize}
            \item \textbf{Young}: 4 passengers, 3 Survived (Yes), 1 Did Not Survive (No)
        \end{itemize}
        
        \begin{align*}\small
        I_{\text{Entropy}}(3, 1) = -\frac{3}{4} \log_2 \left(\frac{3}{4}\right) - \frac{1}{4} \log_2 \left(\frac{1}{4}\right) = 0.811
        \quad \triangleright \text{Eq.~\ref{eq:entropy}} 
        \end{align*}
        
        \begin{itemize}
            \item \textbf{Old}: 16 passengers, 4 Survived (Yes), 12 Did Not Survive (No)
        \end{itemize}
        
        \begin{align*}\small
        I_{\text{Entropy}}(4, 12) = -\frac{4}{16} \log_2 \left(\frac{4}{16}\right) - \frac{12}{16} \log_2 \left(\frac{12}{16}\right) \approx 0.811
        \quad \triangleright \text{Eq.~\ref{eq:entropy}} 
        \end{align*}
        
        The expected entropy after splitting by \emph{Age Category} is:
        
        \begin{align*}\small
        E(0 | \text{Age Category}) = \frac{4}{20} \cdot 0.811 + \frac{16}{20} \cdot 0.811 \approx 0.811
        \quad \triangleright \text{Eq.~\ref{eq:expected_entropy}} 
        \end{align*}
        
        The information gain for \emph{Age} is:
        
        \begin{align*}\small
        \text{Gain}(0 | \text{Age Category}) = 0.934 - 0.811 = 0.123
        \quad \triangleright \text{Eq.~\ref{eq:information_gain}} 
        \end{align*}
    
    \item \textbf{Select the best split for the current node} \hspace{0.25cm}
        Since \emph{Age Category} has the highest information gain ($0.123 > 0.001$) and is selected as the root node of the DT.
    
    \item \textbf{Recursive partitioning} \hspace{0.25cm}
        The DT then branches based on the \emph{Age Category} attribute and recursively applies the same procedure detailed above until a stopping criterion is reached, similar to Hunt's algorithm.
    
\end{enumerate}

\begin{figure}[t]
    \centering
    \includegraphics[width=0.6\textwidth]{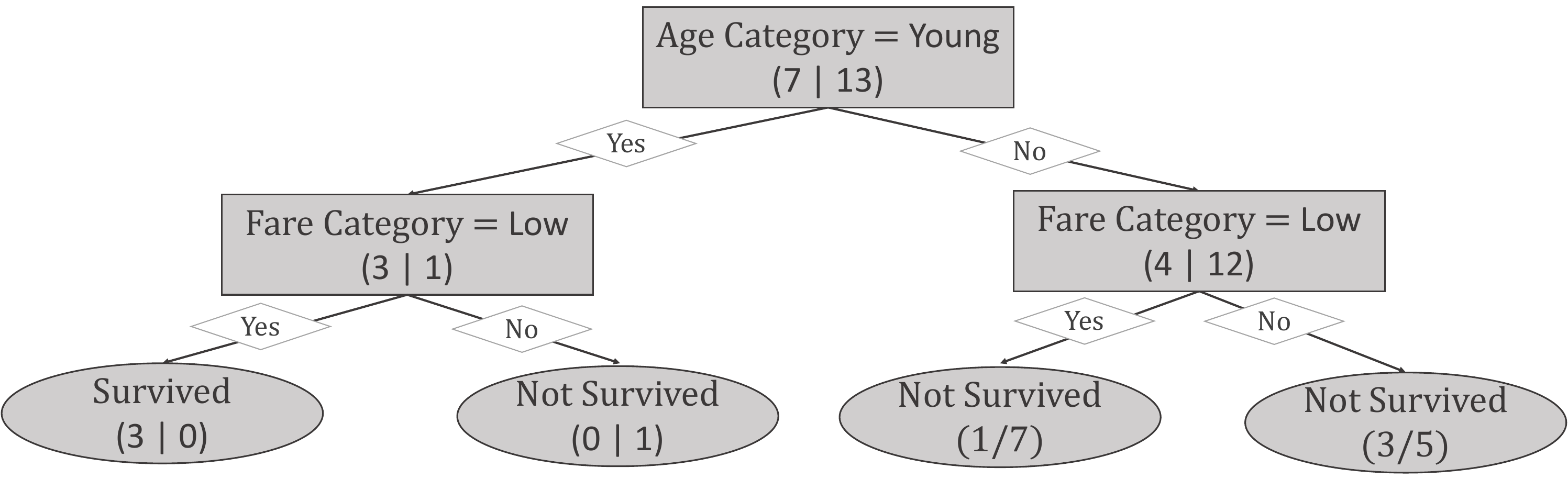}
    \caption[DT learned with ID3]{\textbf{DT learned with ID3.} The tree achieves an accuracy of $80\%$ on the training data and involves three decisions. The main rule encoded by the tree is that a passenger is predicted to survive if he is young and has paid a low fare, and is unlikely to survive otherwise. While the misclassification error does nor decrease with the additional split on the right branch of the root node (still both leafs are 'Not Survived'), the impurity still decreases significantly justifying this additional split, resulting in additional information on the likelihood of survival which is significantly worse fore passengers that are old and paid a low fare.}
    \label{fig:decision_tree_id3}
\end{figure}

Iteratively following this procedure results in the DT visualized in Figure~\ref{fig:decision_tree_id3}. The tree achieves an accuracy of $80\%$ on the training data. While this performance is similar to the tree learned with Hunt's algorithm (Figure~\ref{fig:decision_tree_hunt}), the structure of the tree and the purity in the leaf nodes differs, as the DT learned with ID3 has an additional split on the \emph{Fare Category} in the right branch of the root node providing additional information on the certainty of the prediction.

\paragraph{Limitations and extensions}

While ID3 is effective in constructing simple DTs, it has limitations. ID3 can handle only categorical attributes and continuous attributes must be discretized beforehand. ID3~\citep{quinlan1986_id30} has been extended in later systems like C4.5~\citep{quinlan1993c45} to handle noise, missing values, and continuous attributes directly.

\subsection*{C4.5: Programs for Machine Learning} \label{sec:c45}

The C4.5 algorithm, developed by \citet{quinlan1993c45}, is an extension of the ID3 algorithm and is still widely used for DT induction in machine learning today. It addresses some limitations of ID3, such as handling continuous attributes, missing values, and pruning trees to prevent overfitting. C4.5 builds upon the entropy and information gain concepts of ID3, incorporating several enhancements that improve its robustness and applicability to real-world datasets.

\paragraph{Handling continuous attributes}
Unlike ID3, which only handles categorical attributes, C4.5 can effectively deal with continuous attributes by dynamically determining the optimal threshold for splitting. Given a continuous attribute \(\boldsymbol{x}_{j}\) with \(k\) distinct values, C4.5 evaluates each potential split point between the ordered values \(x_1, x_2, \ldots, x_k\). For each threshold \(\tau\), the data is partitioned into two subsets: Those with values greater than or equal to \(\tau\) and those with values smaller than \(\tau\). The threshold that results in the maximum Gain Ratio is selected as the splitting point. This ratio is calculated by dividing the Gain by a split ratio which acts as a normalizing factor penalizing features with many possible values which would typically lead to more complex trees and therefore overfitting
The Gain Ratio for a test \(T\) with \(k\) outcomes is calculated as:

\begin{equation} \label{eq:gain_ratio}
\text{Ratio}(v|\mathbb{S}_v) = \frac{\text{Gain}(v|\mathbb{S}_v)}{\text{Split}(v|\mathbb{S}_v)},
\end{equation}

where

\begin{equation} \label{eq:split_ratio}
\text{Split}(v|\mathbb{S}_v) = -\sum_{i=0}^{k-1} \frac{\text{P}_{v,i}+\text{N}_{v,i}}{\text{P}_{v}+\text{N}_{v}} \log_2\left(\frac{\text{P}_{v,i}+\text{N}_{v,i}}{\text{P}_{v}+\text{N}_{v}}\right).
\end{equation}

This mechanism allows C4.5 to handle both continuous and discrete attributes more flexibly and effectively.

\subsection*{Example 1c: Constructing a Decision Tree using C4.5}

Again, consider the simplified version of the Titanic dataset from Table~\ref{tab:titanic_2d_num_cat} which now covers the age as continuous and the fare as categorical attribute.
In the following, we will give a step-by-step construction of the root node for the DT using the Gain Ratio criterion.

\begin{enumerate}

    \item \textbf{Calculate the entropy for the whole dataset} \hspace{0.25cm}
    The initial entropy \(I_{\text{Entropy}}(p, n)\) is calculated based on the proportion of passengers who survived (Yes) and who did not survive (No):
    
    20 passengers, 7 Survived (Yes), 13 Did Not Survive (No)

    \begin{align*}\small
    I_{\text{Entropy}}(7, 13) = -\frac{7}{20} \log_2 \left(\frac{7}{20}\right) - \frac{13}{20} \log_2 \left(\frac{13}{20}\right) \approx 0.934
    \quad \triangleright \text{Eq.~\ref{eq:entropy}} 
    \end{align*}
    
     \item \textbf{Calculate gain ratio for each potential split}

        \begin{table}[tb]
            \centering
            \small
            \begin{tabular}{cccc}
                \toprule
                \textbf{Split} & \textbf{Info Gain} & \textbf{Split Info} & \textbf{Gain Ratio} \\ 
                \midrule
                Fare Category    & 0.001 & 0.993 & 0.001 \\
                Age $\geq$ \phantom{0}8.0  & 0.032 & 0.286 & 0.112 \\
                Age $\geq$ 10.5 & 0.008 & 0.469 & 0.016 \\
                Age $\geq$ 12.0 & 0.053 & 0.610 & 0.088 \\
                Age $\geq$ 18.5 & 0.123 & 0.722 & 0.170 \\
                Age $\geq$ 24.5 & 0.064 & 0.811 & 0.079 \\
                Age $\geq$ 25.5 & 0.030 & 0.881 & 0.034 \\
                Age $\geq$ 27.5 & 0.010 & 0.934 & 0.011 \\
                Age $\geq$ 30.5 & 0.001 & 0.971 & 0.001 \\
                Age $\geq$ 38.5 & 0.001 & 0.993 & 0.001 \\
                Age $\geq$ 45.0 & 0.001 & 0.993 & 0.001 \\
                Age $\geq$ 45.5 & 0.001 & 0.993 & 0.001 \\
                Age $\geq$ 50.5 & 0.022 & 0.971 & 0.022 \\
                Age $\geq$ 56.0 & 0.080 & 0.934 & 0.085 \\
                Age $\geq$ 59.0 & 0.049 & 0.881 & 0.056 \\
                Age $\geq$ 64.5 & 0.025 & 0.811 & 0.031 \\
                Age $\geq$ 68.5 & 0.008 & 0.722 & 0.011 \\
                Age $\geq$ 71.5 & 0.000 & 0.610 & 0.000 \\
                Age $\geq$ 77.0 & 0.008 & 0.469 & 0.016 \\
                Age $\geq$ 81.5 & 0.079 & 0.286 & \bftab 0.277 \\
                \bottomrule
            \end{tabular}
            \caption[Split Summary C4.5]{\textbf{Split Summary C4.5.} Information Gain, Split Information and Gain Ratio for each possible split of the root node.}
            \label{tab:info_gain_table}
        \end{table}

        \textbf{Gain and Gain Ratio for \emph{Age}} \hspace{0.25cm}
        We evaluate possible thresholds for the \emph{Age} attribute. Sorted values for \emph{Age} are 6, 10, 11, 13, 24, 25, 26, 29, 32, 45, 45, 46, 55, 57, 61, 68, 69, 74, 80, 83. We consider thresholds at the midpoints:
        
        \textbf{Threshold = 81.5}:
        \begin{itemize}
            \item \textbf{Age $\geq$ 81.5}: 1 passenger, 1 Survived (Yes), 0 Did not survive (No)
            \item \textbf{Age < 81.5}: 19 passengers, 6 Survived (Yes), 13 Did not survive (No)
        \end{itemize}
        
        Entropy after split       \footnote{
        This covers the special case of $log_2(0)$. However, the term \( p_i \log_2(p_i) \) is well-defined because the convention is that
            \[
            0 \cdot \log_2(0) = 0.
            \]      
            This is derived from the limit:
            \[
            \lim_{p \to 0} p \log_2(p) = 0.
            \]
        and therefore when any $p_i=0$, that term contributes nothing to the sum in the entropy calculation, and there is no $log_2(0)$ issue.
        }:
        \begin{align*}\small   
        I_{\text{Entropy}}(1, 0) = -\frac{1}{1} \log_2 \left(\frac{1}{1}\right) - \frac{0}{1} \log_2 \left(\frac{0}{1}\right) = 0.000
        \quad \triangleright \text{Eq.~\ref{eq:entropy}} 
        \end{align*}

        \begin{align*}\small
        I_{\text{Entropy}}(6, 13) = -\frac{6}{19} \log_2\left(\frac{6}{19}\right) - \frac{13}{19} \log_2\left(\frac{13}{19}\right) \approx 0.900
        \quad \triangleright \text{Eq.~\ref{eq:entropy}} 
        \end{align*}     
                        
        Expected entropy:
        \begin{align*}\small   
        E(0|\text{Age} \geq 81.5) = \frac{1}{20} \cdot 0 + \frac{19}{20} \cdot 0.900 = 0.855
        \quad \triangleright \text{Eq.~\ref{eq:expected_entropy}} 
        \end{align*}     
        
        Information gain:
        \begin{align*}\small   
        \text{Gain}(0|\text{Age} \geq 81.5) = 0.934 - 0.855 = 0.079
        \quad \triangleright \text{Eq.~\ref{eq:information_gain}} 
        \end{align*}     
        
        Split information:
        \begin{align*}\small   
        \text{Split}(0|\text{Age} \geq 81.5) = - \left( - \frac{1}{20} \log_2\left(\frac{1}{20}\right) - \frac{19}{20} \log_2\left(\frac{19}{20}\right) \right) = 0.286
        \quad \triangleright \text{Eq.~\ref{eq:split_ratio}} 
        \end{align*}   
        
        Gain Ratio:
        \begin{align*}\small   
        \text{Ratio}(0|\text{Age} \geq 81.5) = \frac{0.079}{0.286} = 0.277
        \quad \triangleright \text{Eq.~\ref{eq:gain_ratio}} 
        \end{align*}   
                
        This calculation is repeated for all potential thresholds (8.0, 10.5, 12.0, 18.5, 24.5, 25.5, 27.5, 30.5, 38.5, 45.0, 45.5, 50.5, 56.0, 59.0, 64.5, 68.5, 71.5, 77.0, 81.5).  We summarized the results for each threshold in Table~\ref{tab:info_gain_table}.

        \textbf{Gain and Gain Ratio for \emph{Fare Category}} \hspace{0.25cm}
        The \emph{Fare Category} attribute has two categories: Low and High. We calculate the entropy for each subset.
        
        \begin{itemize}
            \item \textbf{Low Fare}: 11 passengers, 4 Survived (Yes), 7 Did Not Survive (No)
        \end{itemize}
        
        \begin{align*}\small
        I_{\text{Entropy}}(4, 7) = -\frac{4}{11} \log_2 \left(\frac{4}{11}\right) - \frac{7}{11} \log_2 \left(\frac{7}{11}\right) \approx 0.946
        \quad \triangleright \text{Eq.~\ref{eq:entropy}} 
        \end{align*}
        
        \begin{itemize}
            \item \textbf{High Fare}: 9 passengers, 3 Survived (Yes), 6 Did Not Survive (No)
        \end{itemize}
        
        \begin{align*}\small
        I_{\text{Entropy}}(3, 6) = -\frac{3}{9} \log_2 \left(\frac{3}{9}\right) - \frac{6}{9} \log_2 \left(\frac{6}{9}\right) \approx 0.918
        \quad \triangleright \text{Eq.~\ref{eq:entropy}} 
        \end{align*}
        
        The expected entropy after splitting by \emph{Fare Category} is:
        
        \begin{align*}\small
        E(0|\text{Fare Category}) = \frac{11}{20} \cdot 0.946 + \frac{9}{20} \cdot 0.918 \approx 0.933
        \quad \triangleright \text{Eq.~\ref{eq:expected_entropy}} 
        \end{align*}
        
        The information gain for \emph{Fare Category} is:
        
        \begin{align*}\small
        \text{Gain}(0|\text{Fare Category}) = 0.934 - 0.933 = 0.001
        \quad \triangleright \text{Eq.~\ref{eq:information_gain}} 
        \end{align*}
        
        Split information:
        \begin{align*}\small   
        \text{Split}(0|\text{Fare Category}) = -\left(\frac{11}{20} \log_2 \frac{11}{20} + \frac{9}{20} \log_2 \frac{9}{20}\right) \approx 0.993
        \quad \triangleright \text{Eq.~\ref{eq:split_ratio}} 
        \end{align*}     
        
        Gain Ratio:
        \begin{align*}\small   
        \text{Ratio}(0|\text{Fare Category}) = \frac{0.001}{0.993} \approx 0.001
        \quad \triangleright \text{Eq.~\ref{eq:gain_ratio}} 
        \end{align*}     
        
    \item \textbf{Select the best split for the current node} \hspace{0.25cm} 
        After computing the Gain Ratio for all possible splits of the attributes \emph{Fare Category} and \emph{Age}, the split with the highest Gain Ratio is selected as the root node of the DT. In this case, the maximum gain ratio is achieved by the split $\text{Age} \leq 81.5$ with a value of $0.277$ (Table~\ref{tab:gini_values_table_appendix_appendix}).             

    \item \textbf{Recursive partitioning} \hspace{0.25cm}
        The DT then branches based on the chosen attribute and threshold. Each subset of the data is processed recursively, following the same procedure, until a stopping criterion is met.
    
\end{enumerate}

\begin{figure}[t]
    \centering
    \includegraphics[width=0.6\textwidth]{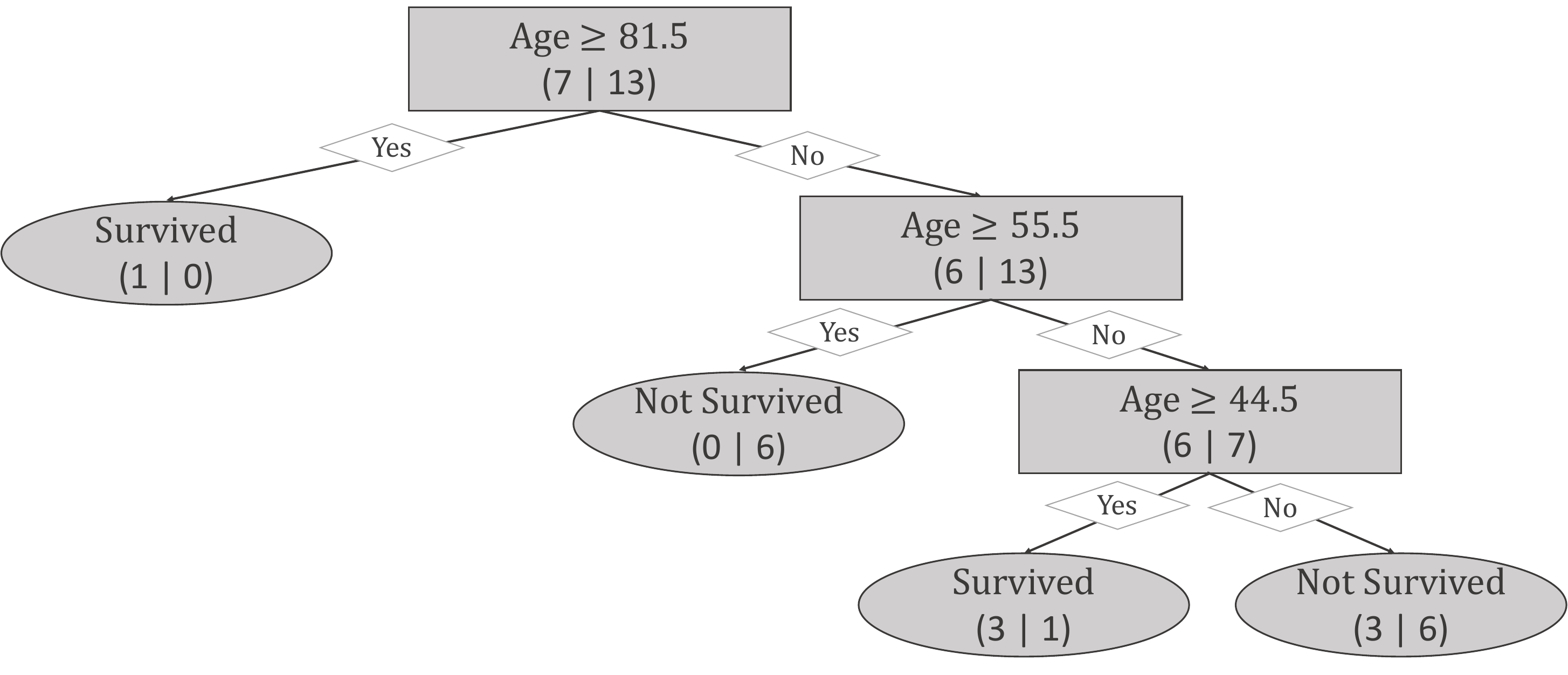}
    \caption[DT learned with C4.5]{\textbf{DT learned with C4.5.} The tree achieves an accuracy of 80\% on the training data and includes 3 decisions. It encodes the rule that if a person is older than 81.5, the person survived, otherwise, if the passenger is older than 55.5, they did not survive. If Age is less than 55.5, those older than 44.5 survived, while those younger did not survive. }
    \label{fig:decision_tree_c45}

\end{figure}

This procedure results in the DT visualized in Figure~\ref{fig:decision_tree_c45}. In contrast to Hunt's algorithm (Figure~\ref{fig:decision_tree_hunt}) and ID3 (Figure~\ref{fig:decision_tree_id3}), C4.5 allows for splits on continuous variables. The tree learned by C4.5 has a total of 3 splits, which are all made based on the age. The split at the root node only separates a single instance in the left branch. While this results in a perfect purity for the corresponding leaf node, this might also indicate overfitting\footnote{Post-pruning for C4.5 has not been applied here for simplicity. While this can control overfitting on the training data, it is not incorporated in the training procedure and might still lead to suboptimal trees.}. Overfitting is a common issue with DTs for numerical features, since it allows for very fine granular splits, as becomes evident in this example. Surprisingly, allowing for splits on continuous variables does not improve the performance over Hunt's algorithm and ID3, when limiting the depth to 3. This can be attributed to the suboptimal, greedy procedure of C4.5 where only the locally optimal splits (e.g., splitting a single sample at the root node) are selected without taking into account the overall model performance.

\paragraph{Pruning and overfitting control}

Another significant enhancement in C4.5 over ID3 is its ability to prune the DT after its initial construction. Pruning helps to reduce overfitting, which occurs when the DT becomes too specific to the training data and loses its generalization capability. C4.5 performs post-pruning by evaluating the expected error rates of subtrees and replacing subtrees with leaf nodes when it leads to a simpler model with comparable or better predictive accuracy. 
The Minimum Description Length (MDL) principle is often used to decide whether a subtree should be pruned. It is a formalization of Occam's Razor, which states that the best model for a dataset is the one that minimizes the total description length, combining the cost of encoding the model itself and the cost of encoding the data given the model. This approach helps balance the complexity of the model with its ability to generalize well to unseen data.

\paragraph{Handling missing values}

C4.5 introduces a method for handling missing attribute values, a common issue in real-world datasets. When encountering a missing value for an attribute in a training instance, C4.5 uses a probabilistic approach to distribute the instance among all possible outcomes of a test, weighted by the relative frequency of the outcomes in the training set. This soft assignment allows the algorithm to make better use of incomplete data without discarding instances, which can be crucial in datasets with many missing values.

\section{Classification and Regression Trees (CART)} \label{sec:cart_appendix}

The CART algorithm is a robust DT learning technique introduced by \citet{breiman1984classification}, used for both, classification and regression tasks. Unlike other tree-based methods like ID3 and C4.5, CART produces only binary trees by recursively splitting the data at each node into two child nodes. The splits are chosen to maximize the purity of the resulting nodes, and the tree can be pruned using cross-validation to avoid overfitting.

\paragraph{Impurity measures}

To determine the best split at each node, CART uses different impurity measures based on the type of task:
\begin{itemize}
    
    \item \textbf{Classification:} For classification tasks, CART uses the Gini impurity to quantify the homogeneity of the target variable within the subsets created by the split.
  
       The Gini impurity at node \( v \) is defined as:

      \begin{equation} \label{eq:gini_appendix}
      I_{\text{Gini}}(v) = 1 - \sum_{c=0}^{|\mathcal{C}|-1} p(c|v)^2,
      \end{equation}
    
      where \( p(c|v) \) is the proportion of class \( c \) instances at node \( v \), and \( |\mathcal{C}| \) is the total number of classes.
    
    \item \textbf{Regression:} For regression tasks, CART uses the variance (mean squared error) of the target variable to evaluate potential splits.

      The variance impurity at node \( v \) is defined as:
    
      \begin{equation} \label{eq:variance_appendix}
      I_{\text{Var}}(v) = \frac{1}{m_{v}} \sum_{i=0}^{m_{v}-1} (y_i - \bar{y}_v)^2,
      \end{equation}
    
      where \( y_i \) is the target value of the \( i \)-th instance at node \( v \), \( \bar{y}_v \) is the mean of the target values at node \( v \), and \( m_v \) is the number of instances at node \( v \).
    
\end{itemize}

\paragraph{Split selection}
The algorithm evaluates all possible splits at each node and selects the split that maximizes the reduction in impurity. For a node \( v \), if a split \( \mathbb{S}_v \) divides it into two child nodes \( v_{\text{left}} \) and \( v_{\text{right}} \), the decrease in Gini impurity due to the split is given by:

\begin{equation} \label{eq:gini_decrease_appendix}
\Delta I_{\text{Gini}}(v | \mathbb{S}_v) = I_{\text{Gini}}(v) - \left( \frac{m_{v,\text{left}}}{m_v} \, I_{\text{Gini}}(v_{\text{left}}) + \frac{m_{v,\text{right}}}{m_v} \, I_{\text{Gini}}(v_{\text{right}}) \right),
\end{equation}

where \( I_{\text{Gini}}(v) \) is the impurity of node \( v \), \( I_{\text{Gini}}(v_{\text{left}}) \) and \( I_{\text{Gini}}(v_{\text{right}}) \) are the impurities of the left and right child nodes after the split \( \mathbb{S}_v \) and \( m_{v,\text{left}} \) and \( m_{v,\text{right}} \) are the numbers of instances in the left and right child nodes, respectively.
The goal is to find the split \( \mathbb{S}_v \) that maximizes \( \Delta I_{\text{Gini}}(\mathbb{S}_v, v)  \), thereby achieving the greatest reduction in impurity and resulting in more homogeneous child nodes.

\subsection*{Example 1d: Constructing a Decision Tree using CART}

Again, we use the simplified version of the Titanic dataset from Table~\ref{tab:titanic_2d_num_cat} including the age as continuous and the fare as categorical attribute.
In the following, we provide a step-by-step example for constructing the root node of the DT using the Gini impurity criterion.

\begin{enumerate}
    \item \textbf{Calculate the Gini impurity for the whole dataset}  \hspace{0.25cm}
    The initial Gini impurity \(I_{\text{Gini}}\) is calculated based on the proportion of passengers who survived (Yes) and who did not survive (No):
    
    20 passengers, 7 Survived (Yes), 13 Did Not Survive (No)
    
    \begin{align*}\small   
    I_{\text{Gini}}(v_0) = 1 - \left(\frac{7}{20}\right)^2 - \left(\frac{13}{20}\right)^2 = 0.455
    \quad \triangleright \text{Eq.~\ref{eq:gini_appendix}} 
    \end{align*}   
    
    \textbf{Calculate decrease in Gini impurity for each potential split}

        \begin{table}[tb]
            \centering
            \small
            \begin{tabular}{c|cccc}
                \toprule
                \textbf{Threshold} & \textbf{Left Gini} & \textbf{Right Gini} & \textbf{Weighted Gini} & \textbf{Gini Gain} \\ 
                \midrule
                Fare Category    & 0.463 & 0.444 & 0.455 & 0.000 \\
                Age $\geq$ \phantom{0}8.0   & 0.000 & 0.465 & 0.442 & 0.013 \\
                Age $\geq$ 10.5  & 0.500 & 0.444 & 0.450 & 0.005 \\
                Age $\geq$ 12.0  & 0.444 & 0.415 & 0.420 & 0.035 \\
                Age $\geq$ 18.5  & 0.375 & 0.375 & 0.375 & \bftab 0.080 \\
                Age $\geq$ 24.5  & 0.480 & 0.391 & 0.413 & 0.042 \\
                Age $\geq$ 25.5  & 0.500 & 0.408 & 0.436 & 0.019 \\
                Age $\geq$ 27.5  & 0.490 & 0.426 & 0.448 & 0.007 \\
                Age $\geq$ 30.5  & 0.469 & 0.444 & 0.454 & 0.001 \\
                Age $\geq$ 38.5  & 0.444 & 0.463 & 0.455 & 0.000 \\
                Age $\geq$ 45.0  & 0.463 & 0.444 & 0.455 & 0.000 \\
                Age $\geq$ 45.5  & 0.463 & 0.444 & 0.455 & 0.000 \\
                Age $\geq$ 50.5  & 0.486 & 0.375 & 0.442 & 0.013 \\
                Age $\geq$ 56.0  & 0.497 & 0.245 & 0.409 & 0.046 \\
                Age $\geq$ 59.0  & 0.490 & 0.278 & 0.426 & 0.029 \\
                Age $\geq$ 64.5  & 0.480 & 0.320 & 0.440 & 0.015 \\
                Age $\geq$ 68.5  & 0.469 & 0.375 & 0.450 & 0.005 \\
                Age $\geq$ 71.5  & 0.457 & 0.444 & 0.455 & 0.000 \\
                Age $\geq$ 77.0  & 0.444 & 0.500 & 0.450 & 0.005 \\
                Age $\geq$ 81.5  & 0.432 & 0.000 & 0.411 & 0.044 \\
                \bottomrule
            \end{tabular}
            \caption[Split Summary CART]{\textbf{Split Summary CART.} Gini Values and Gini Gain for each possible split of the root node.}
            \label{tab:gini_values_table_appendix_appendix}
        \end{table}
        
        \textbf{Gini Impurity for \emph{Age} Attribute}  \hspace{0.25cm}
        We evaluate possible thresholds for the \emph{Age} attribute. Sorted values for \emph{Age} are 6, 10, 11, 13, 24, 25, 26, 29, 32, 45, 45, 46, 55, 57, 61, 68, 69, 74, 80, 83. We consider thresholds at the midpoints:
        
        \textbf{Threshold = 18.5}:
        \begin{itemize}
            \item \textbf{Age $\geq$ 18.5}: 4 passengers, 3 Survived (Yes), 1 Did not survive (No)
        \end{itemize}        
        \begin{align*}\small   
        I_{\text{Gini}}(v_{0,\text{left}}) = 1 - \left(\frac{3}{4}\right)^2 - \left(\frac{1}{4}\right)^2 = 0.375
        \quad \triangleright \text{Eq.~\ref{eq:gini_appendix}} 
        \end{align*}   
        
        \begin{itemize}
            \item \textbf{Age < 18.5}: 16 passenger, 4 Survived (Yes), 12 Did not survive (No)
        \end{itemize}
        \begin{align*}\small   
        I_{\text{Gini}}(v_{0,\text{right}}) = 1 - \left(\frac{4}{16}\right)^2 - \left(\frac{12}{16}\right)^2 = 0.375
        \quad \triangleright \text{Eq.~\ref{eq:gini_appendix}} 
        \end{align*}   
        
        Decrease in Gini impurity:
        \begin{align*}\small   
        \Delta I_{\text{Gini}}(v_0 | \text{Age} \geq 18.5) = 0.455 -\frac{4}{20} \cdot 0.375 + \frac{16}{20} \cdot 0.375  = 0.080
        \quad \triangleright \text{Eq.~\ref{eq:gini_decrease_appendix}} 
        \end{align*}    
        
        This calculation is repeated for all potential thresholds (8.0, 10.5, 12.0, 18.5, 24.5, 25.5, 27.5, 30.5, 38.5, 45.0, 45.5, 50.5, 56.0, 59.0, 64.5, 68.5, 71.5, 77.0, 81.5). We summarized the results in Table~\ref{tab:gini_values_table_appendix_appendix}.

        \textbf{Gini Impurity for \emph{Fare Category} attribute}  \hspace{0.25cm}
        We calculate the Gini impurity for each subset:
        \begin{itemize}
            \item \textbf{Low Fare}: 11 passenger, 4 Survived (Yes), 7 Did not survive (No)
        \end{itemize}        
        \begin{align*}\small   
        I_{\text{Gini}}(v_{0,\text{left}}) = 1 - \left(\left(\frac{4}{11}\right)^2 + \left(\frac{7}{11}\right)^2\right) = 0.463
        \quad \triangleright \text{Eq.~\ref{eq:gini_appendix}} 
        \end{align*}   
        
        \begin{itemize}
            \item \textbf{High Fare}: 9 passengers, 3 Survived (Yes), 6 Did not survive (No)
        \end{itemize}
        \begin{align*}\small   
        I_{\text{Gini}}(v_{0,\text{right}}) = 1 - \left(\left(\frac{3}{9}\right)^2 + \left(\frac{6}{9}\right)^2\right) = 0.444
        \quad \triangleright \text{Eq.~\ref{eq:gini_appendix}} 
        \end{align*}   
        
        Decrease in Gini impurity:
        \begin{align*}\small  
        \Delta I_{\text{Gini}}(v_0 | \text{Fare Category}) = 0.455 - \frac{11}{20} \cdot 0.463 + \frac{9}{20} \cdot 0.444 = 0.000
        \quad \triangleright \text{Eq.~\ref{eq:gini_decrease_appendix}} 
        \end{align*}           

    \item \textbf{Select the best split for the current node} \hspace{0.25cm}
        After computing the decrease in Gini impurity for all possible splits of the attributes \emph{Fare Category} and \emph{Age} the attribute-threshold pair with the highest reduction in impurity (Age $\geq$ 18.5 with a decrease of 0.080) is selected as the root node of the DT. The impurity decrease for each potential split is summarized in Table~\ref{tab:gini_values_table_appendix_appendix}.                

    \item \textbf{Recursive Partitioning} \hspace{0.25cm}
        The DT then branches based on the chosen attribute and threshold. Each subset of the data is processed recursively, following the same procedure, until all nodes are pure, or another stopping criterion is met.
    
\end{enumerate}

\begin{figure}[t]
    \centering
    \includegraphics[width=0.7\textwidth]{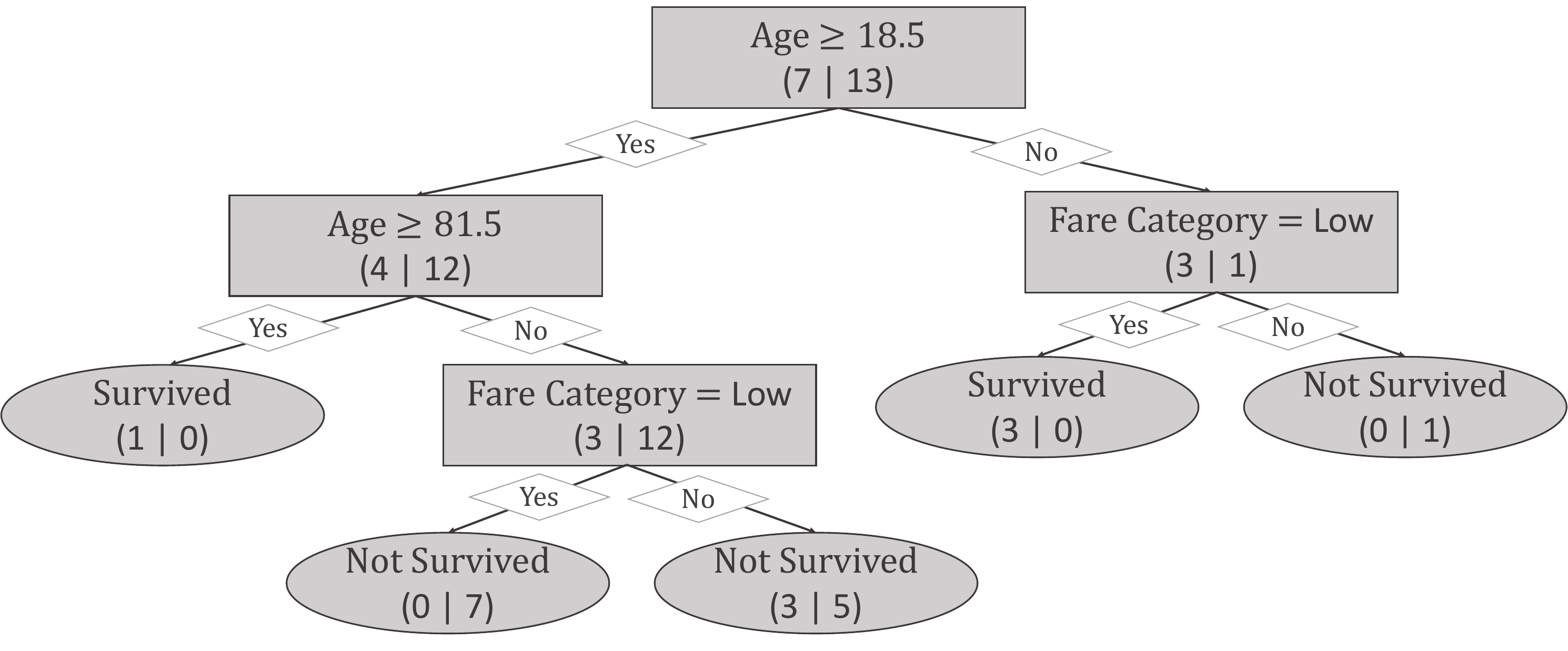}
    \caption[DT learned with CART]{\textbf{DT learned with CART.} The DT achieves an accuracy of 85\% on the training data and covers a total of 4 splits. It encodes the rule that if a passenger is older than 18.5, it is checked if the passenger is also older than 81.5. Those older survived and those younger did not survive, but with different certainties based on the fare. If passengers are younger than 18.5, the tree splits based on Fare Category, where those with a low fare did survive, while those with a higher fare survived.}
    \label{fig:decision_tree_cart_appendix}
\end{figure}

This procedure results in the DT visualized in Figure~\ref{fig:decision_tree_cart_appendix}. We can immediately see the impact of different purity measures, as the tree learned by CART (using Gini) is significantly different from the tree learned by C4.5 (using Information Gain). CART does not separate a single passenger at the root node, even if this would result in a perfect purity at the corresponding leaf immediately. This is caused by the fact that Information Gain rewards higher purity more compared to Gini (logarithmic vs. quadratic weighting) where splits producing larger homogeneous groups are preferred. As a result, the performance on the training data for CART ($85\%$) is higher compared to C4.5 (Figure~\ref{fig:decision_tree_lookahead})  with only $80\%$ accuracy. This also shows the advantage of splitting on continuous features, as the performance over Hunt's algorithm and ID3 increased as more information can be exploited.

\paragraph{Pruning and cross-validation} CART uses a pruning strategy to prevent overfitting by growing a large tree and then removing branches that provide little predictive power. The pruning process is typically guided by cross-validation, where the tree is pruned back to the level that minimizes the cross-validated error estimate. The optimal tree size is selected to balance model complexity and predictive accuracy.

\paragraph{Handling missing values} CART handles missing values by using a technique called surrogate splits. When a split decision is based on a feature with missing values, CART tries to use other features that correlate with the original one to make the decision. These alternative features, known as surrogates, are chosen based on how well they mimic the splitting feature. If no good surrogate is found, CART can assign cases with missing values to both branches of the split, weighing their contributions based on how frequently each branch is used by non-missing cases.

\section{Comparison of ID3, C4.5, and CART} \label{sec:greedy_dt_comparison}
The DT algorithms ID3, C4.5, and CART each possess distinct methodologies suited to different data types and applications, as summarized in Table~\ref{tab:greedy_summary}. Today, CART can be considered as the most commonly used DT learning algorithm, which is also implemented in libraries like \texttt{sklearn}\footnote{\url{https://scikit-learn.org/stable/modules/tree.html\#tree-algorithms} (Last accessed: March 3, 2025).}. While greedy methods excel in their efficiency, they usually suffer from local optimality. This is tackled by alternative approaches, as we will show in the following sections. Furthermore, greedy methods often lack flexibility. They are tied to impurity measures and are not easily adaptable to custom objectives and can only hardly be integrated into existing frameworks, often requiring a gradient-based optimization (like reinforcement learning).

\begin{table}
    \centering
    \small
    \caption[Comparison of ID3, C4.5 and CART]{\textbf{Comparison of ID3, C4.5 and CART.} In this table, we compare common greedy, purity-based DT induction methods based on the most relevant aspects.}
    
    \begin{tabular}{|c|C{2.5cm}|C{2.5cm}|C{2.5cm}|}
        \hline
         &  \textbf{ID3 (Section~\ref{sec:id3})} &  \textbf{C4.5 (Section~\ref{sec:c45})} & \textbf{CART (Section~\ref{sec:cart_appendix})} \\ \hline
         \textbf{Impurity Measures} &  Information Gain&  Gain Ratio& Gini Impurity\\ \hline
         \textbf{Continuous Attributes} &  -&  yes & yes\\ \hline
         \textbf{Pruning Strategy} &  -&  post-pruning& cost-complexity pruning\\ \hline
         \textbf{Splitting Strategy} &  multi-way&  multi-way& binary\\ \hline
         \textbf{Missing Values} &  -&  probabilistic& surrogate splits\\ \hline
    \end{tabular}

    \label{tab:greedy_summary}
\end{table}

\section{Further Optimizations}
Since CART and C4.5 have been introduced, many variations to those algorithms have been proposed. The most prominent ones are C5.0~\citep{kuhn2013applied_c50} and GUIDE~\citep{loh2002regression_guide1,loh2009improving_guide2}.

\subsection*{See5 and C5.0}
C5.0 (See5 for the Windows implementation\footnote{\url{https://www.rulequest.com/see5-info.html} (Last accessed: March 3, 2025)}) is an advancement of the C4.5 algorithm developed by \citet{quinlan1993c45}, offering several improvements and new features over its predecessor~\citep{kuhn2013applied_c50} released under a proprietary license. While both C5.0 and C4.5 are DT-based models designed for classification tasks, C5.0 introduces enhancements that improve efficiency, accuracy, and flexibility in various contexts.

\paragraph{Model complexity and tree pruning}
C5.0 generally produces smaller and more efficient trees compared to C4.5. This is achieved through several modifications to the tree-building process. One notable improvement is the introduction of a global pruning mechanism that uses a cost-complexity approach. This method prunes subtrees until the error rate exceeds a standard error of the baseline rate. The pruning strategy in C5.0 contrasts with the pessimistic pruning of C4.5, which estimates error rates with upper confidence bounds and removes subtrees if the error rate is reduced.

\paragraph{Handling of categorical variables}
Another significant improvement in C5.0 is its handling of categorical variables. C5.0 can group categories into fewer splits based on a heuristic algorithm that determines the best category groupings for each split. This results in smaller and more interpretable trees. The earlier version, C4.5, treats each category independently, which can lead to unnecessarily large trees if the data has many categorical levels. C5.0's ability to optimize the grouping of categories reduces tree complexity and improves performance.

\paragraph{Improved handling of missing values}
While both C4.5 and C5.0 deal with missing data during the model-building process, C5.0 employs a more refined approach. C4.5 uses fractional data accounting methods, where missing values contribute fractionally to all possible outcomes based on their distribution among non-missing values. C5.0, however, improves upon this by making more sophisticated adjustments to the training process, resulting in more accurate handling of missing values during tree construction.

\subsection*{Generalized, Unbiased Interaction Detection and Estimation (GUIDE)}

GUIDE~\citep{loh2002regression_guide1, loh2009improving_guide2} is an advanced algorithm for constructing regression trees, designed to address several inherent limitations in traditional tree-based methods such as CART. The primary objectives of GUIDE are to eliminate selection bias in variable selection, improve the detection of local interactions between variables, and enhance the flexibility and interpretability of the regression models produced.

\paragraph{Addressing selection bias}
One of the most significant improvements offered by GUIDE is its ability to mitigate selection bias, a common problem in traditional regression tree algorithms. In standard methods, such as CART, there is a propensity to favor variables that provide more potential splits, which can distort the interpretation of the model. For instance, continuous variables or categorical variables with many levels are more likely to be selected due to their higher number of possible splits.
GUIDE counters this issue by employing a chi-square test for residuals at each node to determine which variable should be used for the next split. This method prevents the algorithm from favoring variables with a higher number of splits and maintains an unbiased selection probability for each variable. Moreover, GUIDE uses a bootstrap calibration technique to adjust p-values, which further ensures that the variable selection remains unbiased and is not unduly influenced by the data distribution.

\paragraph{Interaction detection}
Another key feature of GUIDE is its ability to detect and incorporate interactions between variables more effectively. Traditional algorithms often fail to detect local interactions, particularly in datasets where such interactions are subtle or involve more than two variables. GUIDE addresses this through a series of statistical tests that specifically aim to identify significant interactions at each node. By incorporating these interaction tests, GUIDE is capable of constructing trees that capture complex relationships between variables more accurately, which leads to models that are both more parsimonious and more predictive.

\subsection*{Conclusion}
While these C5.0 and GUIDE were designed to tackle weaknesses of existing methods, until today, none of these algorithms was able to consistently outperform CART and C4.5 in independent experiments as shown for instance by \citet{zharmagambetov2021non}. Furthermore, they are still considered as greedy approaches and therefore still have similar limitations regarding a constrained search space, local optimality and flexibility.

\chapter{Appendix GradTree} %
\label{A:GradTree} %

\section{Datasets}\label{A:dataset}
The datasets along with their specifications and source are summarized in Table~\ref{tab:data-spec}. 
For all datasets, we performed a standard preprocessing: We applied Leave-one-out encoding to all categorical features. %
Similar to \citet{popov2019neural}, we further perform a quantile transform, making each feature follows a normal distribution. %
We use a random $80\%/20\%$ train-test split for all datasets. To account for class imbalance, we rebalanced the training data using SMOTE~\citep{chawla2002smote} when the minority class accounts for less than $\frac{25}{c-1} \%$ of the data points, where $c$ is the number of classes.
Since GradTree and DNDT require a validation set for early stopping, we performed another $80\%/20\%$ split on the training data for those approaches. The remainder of the approaches utilize the complete training data. 
For DL8.5 additional preprocessing was necessary since they can only handle binary features. Therefore, we one-hot encoded all categorical features and discretized numeric features by one-hot encoding them using quantile binning with $5$ bins.

\begin{table}[H]
\centering
\caption[Dataset Specifications.]{\textbf{Dataset Specifications.}}
\resizebox{1\columnwidth}{!}{
\begin{tabular}{L{3.2cm}R{1.7cm}R{1.7cm}R{2.5cm}R{1.7cm}L{9cm}}
\toprule
\multicolumn{1}{L{3.2cm}}{\textbf{Dataset Name}}& \multicolumn{1}{R{1.7cm}}{\textbf{Number of Features}}& \multicolumn{1}{R{1.7cm}}{\textbf{Number of Samples}}& \multicolumn{1}{R{3cm}}{\textbf{Samples by Class}}& \multicolumn{1}{R{1.7cm}}{\textbf{Number of Classes}}& \textbf{Link}\\
\toprule

Blood Transfusion& 4& 748& 178  / 570& 2& \url{https://archive.ics.uci.edu/ml/datasets/Blood+Transfusion+Service+Center}\\
Banknote Authentication& 4& 1372& 610  / 762& 2& \url{https://archive.ics.uci.edu/ml/datasets/banknote+authentication}\\
Titanic& 7& 891& 342 / 549& 2& \url{https://www.kaggle.com/c/titanic}\\
Raisin& 7& 900& 450 /450& 2& \url{https://archive.ics.uci.edu/ml/datasets/Raisin+Dataset}\\
Rice& 7& 3810& 1630 / 2180& 2& \url{https://archive.ics.uci.edu/ml/datasets/Rice+\%28Cammeo+and+Osmancik\%29}\\
Echocardiogram& 8& 132& 107 / 25& 2& \url{https://archive.ics.uci.edu/ml/datasets/echocardiogram}\\
Wisconsin Diagnostic Breast Cancer& 10& 569& 212  / 357& 2& \url{https://archive.ics.uci.edu/ml/datasets/breast+cancer+wisconsin+(diagnostic)}\\
\bottomrule
\end{tabular}%
}

\label{tab:data-spec}
\end{table}

\section{Additional Results}

\subsection{Ablation Study}

\begin{table}[H]

\centering
\caption[Ablation Study Optimized Parameters]{\textbf{Ablation Study Optimized Parameters.} We report macro F1-scores (mean $\pm$ stdev over $10$ trials) with optimized parameters and the rank of each approach in brackets. The datasets are sorted by the number of features. The top part comprises binary classification tasks and the bottom part multi-class datasets.}
\resizebox{1.0\columnwidth}{!}{

\begin{tabular}{lccc}
    \toprule
    & ST Entmax (ours) & ST Gumbel Softmax & Temperature Annealing \\
    \midrule
        Blood Transfusion & \bftab 0.628 $\pm$ .036 (1) & 0.482 $\pm$ .083 (3) & 0.606 $\pm$ .073 (2) \\
        Banknote Authentication & \bftab 0.987 $\pm$ .007 (1) & 0.770 $\pm$ .144 (3) & 0.946 $\pm$ .041 (2) \\
        Titanic & \bftab 0.776 $\pm$ .025 (1) & 0.543 $\pm$ .093 (3) & 0.762 $\pm$ .036 (2) \\
        Raisins & 0.840 $\pm$ .022 (2) & 0.792 $\pm$ .069 (3) & \bftab 0.846 $\pm$ .028 (1) \\
        Rice & 0.926 $\pm$ .007 (2) & 0.859 $\pm$ .058 (3) & \bftab 0.927 $\pm$ .006 (1) \\
        Echocardiogram & \bftab 0.658 $\pm$ .113 (1) & 0.574 $\pm$ .108 (3) & 0.619 $\pm$ .151 (2) \\
        Wisconsin Diagnostic Breast Cancer & \bftab 0.904 $\pm$ .022 (1) & 0.844 $\pm$ .059 (3) & 0.893 $\pm$ .031 (2) \\
\bottomrule
\end{tabular}
}
\label{tab:ablation_HPO}
\end{table}

\begin{table}[H]
\centering
\caption[Ablation Study Default Parameters]{\textbf{Ablation Study Default Parameters.} We report macro F1-scores (mean $\pm$ stdev over $10$ trials) with default parameters and the rank of each approach in brackets. The datasets are sorted by the number of features. The top part comprises binary classification tasks and the bottom part multi-class datasets.}
\resizebox{1.0\columnwidth}{!}{
\begin{tabular}{lccc}
    \toprule
    & ST Entmax (ours) & ST Gumbel Softmax & Temperature Annealing \\
    \midrule
        Blood Transfusion & \bftab 0.627 $\pm$ .056 (1) & 0.497 $\pm$ .086 (3) & 0.618 $\pm$ .045 (2) \\
        Banknote Authentication & \bftab 0.980 $\pm$ .008 (1) & 0.722 $\pm$ .126 (3) & 0.970 $\pm$ .007 (2) \\
        Titanic & \bftab 0.782 $\pm$ .035 (1) & 0.623 $\pm$ .096 (3) & 0.769 $\pm$ .033 (2) \\
        Raisins & 0.850 $\pm$ .025 (2) & 0.802 $\pm$ .055 (3) & \bftab 0.853 $\pm$ .016 (1) \\
        Rice & \bftab 0.926 $\pm$ .006 (1) & 0.802 $\pm$ .101 (3) & \bftab 0.926 $\pm$ .006 (1) \\
        Echocardiogram &\bftab  0.648 $\pm$ .130 (1) & 0.571 $\pm$ .093 (3) & 0.595 $\pm$ .121 (2) \\
        Wisconsin Diagnostic Breast Cancer & 0.902 $\pm$ .029 (2) & 0.841 $\pm$ .129 (3) & \bftab 0.903 $\pm$ .028 (1) \\
\bottomrule
\end{tabular}
}
\label{tab:ablation_noHPO}
\end{table}

\subsection{Train Data Performance}

\begin{table}[H]

\centering
\caption[Train Performance Comparison]{\textbf{Train Performance Comparison.} We report macro F1-scores (mean $\pm$ stdev over $10$ trials) on the training data and the rank of each approach in brackets. The datasets are sorted by the number of features. The top part comprises binary classification tasks and the bottom part multi-class datasets.}
\resizebox{1.0\columnwidth}{!}{

\begin{tabular}{lccccc}
\toprule
{} &                           \multicolumn{2}{l}{Gradient-Based} &            \multicolumn{2}{l}{Non-Greedy} &                       \multicolumn{1}{l}{Greedy} \\ \cmidrule(lr){2-3}  \cmidrule(lr){4-5} \cmidrule(lr){6-6}
  
{} &                           \multicolumn{1}{l}{GradTree (ours)} &                   \multicolumn{1}{l}{DNDT} &            \multicolumn{1}{l}{GeneticTree} &                          \multicolumn{1}{l}{DL8.5 (Optimal)} &                       \multicolumn{1}{l}{CART} \\
\midrule
Blood Transfusion                  &         0.686 $\pm$ .032 (3) &  0.552 $\pm$ .085 (5) &  0.615 $\pm$ .122 (4) &         0.711 $\pm$ .045 (2) &  \bftab 0.832 $\pm$ .040 (1) \\
Banknote Authentication            &         0.995 $\pm$ .003 (2) &  0.907 $\pm$ .012 (5) &  0.933 $\pm$ .016 (4) &         0.967 $\pm$ .003 (3) &  \bftab 0.999 $\pm$ .001 (1) \\
Titanic                            &         0.761 $\pm$ .128 (5) &  0.791 $\pm$ .025 (4) &  0.908 $\pm$ .023 (3) &  \bftab 0.970 $\pm$ .006 (1) &         0.949 $\pm$ .012 (2) \\
Raisins                            &  \bftab 0.892 $\pm$ .024 (1) &  0.829 $\pm$ .027 (5) &  0.863 $\pm$ .004 (4) &         0.889 $\pm$ .004 (2) &         0.866 $\pm$ .004 (3) \\
Rice                               &         0.925 $\pm$ .003 (3) &  0.913 $\pm$ .012 (5) &  0.920 $\pm$ .002 (4) &  \bftab 0.930 $\pm$ .002 (1) &         0.927 $\pm$ .002 (2) \\
Echocardiogram                     &         0.827 $\pm$ .088 (4) &  0.817 $\pm$ .034 (5) &  0.844 $\pm$ .032 (3) &         0.935 $\pm$ .013 (2) &  \bftab 0.981 $\pm$ .009 (1) \\
\bottomrule
\end{tabular}%
}
\label{tab:eval-results-train}
\end{table}

\subsection{Tree Size Comparison}

\begin{table}[H]

\centering
\caption[Tree Size Comparison]{\textbf{Tree Size Comparison.} We report the average tree size based on the optimized hyperparameters (mean $\pm$ stdev over $10$ trials) and the rank of each approach in brackets. The datasets are sorted by the number of features. The top part comprises binary classification tasks and the bottom part multi-class datasets.}
\resizebox{1.0\columnwidth}{!}{

\begin{tabular}{lccccc}
\toprule
{} &                           \multicolumn{2}{l}{Gradient-Based} &            \multicolumn{2}{l}{Non-Greedy} &                       \multicolumn{1}{l}{Greedy} \\ \cmidrule(lr){2-3}  \cmidrule(lr){4-5} \cmidrule(lr){6-6}
  
{} &                           \multicolumn{1}{l}{GradTree (ours)} &                   \multicolumn{1}{l}{DNDT} &            \multicolumn{1}{l}{GeneticTree} &                          \multicolumn{1}{l}{DL8.5 (Optimal)} &                       \multicolumn{1}{l}{CART} \\
\midrule
Blood Transfusion                  &           \phantom{0}29.20 $\pm$ \phantom{00}8.22 (3) &           \phantom{0.0}24.00 $\pm$ \phantom{0.00}0.00 (2) &         \phantom{0}5.20 $\pm$ \phantom{0}3.03 (1) &          24.00 $\pm$ \phantom{0}1.61 (3) &   165.60 $\pm$ \phantom{0}21.84 (5) \\
Banknote Authentication            &          \phantom{0}60.00 $\pm$ \phantom{0}11.22 (5) &           \phantom{0.0}24.00 $\pm$ \phantom{0.00}0.00 (2) &       \bftab 12.80 $\pm$ \phantom{0}5.33 (1) &          26.80 $\pm$ \phantom{0}0.60 (3) &           \phantom{0}44.80 $\pm$ \phantom{00}4.69 (4) \\
Titanic                            &          \phantom{0}39.60 $\pm$ \phantom{0}10.17 (3) &          \phantom{0.}142.00 $\pm$ \phantom{0.00}0.00 (5) & \bftab 10.40 $\pm$ \phantom{0}3.11 (1) &          28.20 $\pm$ \phantom{0}0.98 (3) &           \phantom{0}21.60 $\pm$ \phantom{00}0.92 (2) \\
Raisins                            &         114.80 $\pm$ \phantom{0}33.69 (4) &          \phantom{0.}142.00 $\pm$ \phantom{0.00}0.00 (5) &   \phantom{0}3.60 $\pm$ \phantom{0}1.80 (2) &          29.40 $\pm$ \phantom{0}1.50 (3) &           \bftab \phantom{00}3.00 $\pm$ \phantom{00}0.00 (1) \\
Rice                               &          \phantom{0}41.00 $\pm$ \phantom{0}10.05 (4) &          \phantom{0.}142.00 $\pm$ \phantom{0.00}0.00 (5) &   \phantom{0}3.00 $\pm$ \phantom{0}0.00 (2) &          30.20 $\pm$ \phantom{0}0.98 (3) &           \bftab \phantom{00}3.00 $\pm$ \phantom{00}0.00 (1) \\
\bottomrule
\end{tabular}%
}
\label{tab:eval-results-tree-size}
\end{table}

\subsection{Default Parameter Comparison}

\begin{table}[H]

\centering
\caption[Performance Comparison Default Hyperparameters]{\textbf{Performance Comparison Default Hyperparameters.} We report macro F1-scores (mean $\pm$ stdev over $10$ trials) with default parameters and the rank of each approach in brackets. The datasets are sorted by the number of features. The top part comprises binary classification tasks and the bottom part multi-class datasets.}
\resizebox{1.0\columnwidth}{!}{

\begin{tabular}{lccccc}
\toprule
{} &                           \multicolumn{2}{l}{Gradient-Based} &            \multicolumn{2}{l}{Non-Greedy} &                       \multicolumn{1}{l}{Greedy} \\ \cmidrule(lr){2-3}  \cmidrule(lr){4-5} \cmidrule(lr){6-6}
  
{} &                           \multicolumn{1}{l}{GradTree (ours)} &                   \multicolumn{1}{l}{DNDT} &            \multicolumn{1}{l}{GeneticTree} &                          \multicolumn{1}{l}{DL8.5 (Optimal)} &                       \multicolumn{1}{l}{CART} \\
\midrule
Blood Transfusion                  &  \bftab 0.627 $\pm$ .056 (1) &         0.558 $\pm$ .059 (5) &         0.574 $\pm$ .093 (3) &         0.590 $\pm$ .034 (2) &         0.573 $\pm$ .036 (4) \\
Banknote Authentication            &         0.980 $\pm$ .008 (2) &         0.886 $\pm$ .024 (5) &         0.928 $\pm$ .022 (4) &         0.962 $\pm$ .011 (3) &  \bftab 0.981 $\pm$ .006 (1) \\
Titanic                            &  \bftab 0.782 $\pm$ .035 (1) &         0.740 $\pm$ .026 (4) &         0.763 $\pm$ .041 (2) &         0.754 $\pm$ .031 (3) &         0.735 $\pm$ .041 (5) \\
Raisins                            &         0.850 $\pm$ .025 (2) &         0.832 $\pm$ .026 (4) &  \bftab 0.859 $\pm$ .022 (1) &         0.849 $\pm$ .027 (3) &         0.811 $\pm$ .028 (5) \\
Rice                               &         0.926 $\pm$ .006 (2) &         0.902 $\pm$ .018 (5) &  \bftab 0.927 $\pm$ .005 (1) &         0.925 $\pm$ .008 (3) &         0.905 $\pm$ .010 (4) \\
Echocardiogram                     &  \bftab 0.648 $\pm$ .130 (1) &         0.543 $\pm$ .117 (5) &         0.563 $\pm$ .099 (3) &         0.609 $\pm$ .112 (2) &         0.559 $\pm$ .075 (4) \\
\bottomrule
\end{tabular}%
}
\label{tab:eval-results_noHPO_complete}
\end{table}

\subsection{Runtime Comparison}

\begin{table}[H]

\centering
\caption[Runtime Comparison]{\textbf{Runtime Comparison.} We report runtime without restarts based on the optimized hyperparameters (mean $\pm$ stdev over $10$ trials) and the rank of each approach in brackets. The datasets are sorted by the number of features. The top part comprises binary classification tasks and the bottom part multi-class datasets.}
\resizebox{1.0\columnwidth}{!}{

\begin{tabular}{lccccc}
\toprule
{} &                           \multicolumn{2}{l}{Gradient-Based} &            \multicolumn{2}{l}{Non-Greedy} &                       \multicolumn{1}{l}{Greedy} \\ \cmidrule(lr){2-3}  \cmidrule(lr){4-5} \cmidrule(lr){6-6}
  
{} &                           \multicolumn{1}{l}{GradTree (ours)} &                   \multicolumn{1}{l}{DNDT} &            \multicolumn{1}{l}{GeneticTree} &                          \multicolumn{1}{l}{DL8.5 (Optimal)} &                       \multicolumn{1}{l}{CART} \\
\midrule
Blood Transfusion                  &        \phantom{0}13.01 $\pm$ \phantom{0}1.00 (5) &    \phantom{0}3.70 $\pm$ \phantom{0}1.00 (3) &   \phantom{0}10.51 $\pm$ \phantom{0}4.00 (4) &    \phantom{00}0.03 $\pm$ 0.00 (2) &  \bftab 0.00 $\pm$ 0.00 (1) \\
Banknote Authentication            &        \phantom{0}13.21 $\pm$ \phantom{0}2.00 (4) &   34.43 $\pm$ \phantom{0}1.00 (5) &   \phantom{0}10.32 $\pm$ \phantom{0}3.00 (3) &    \phantom{00}0.03 $\pm$ 0.00 (2) &  \bftab 0.00 $\pm$ 0.00 (1) \\
Titanic                            &        \phantom{0}29.58 $\pm$ \phantom{0}1.00 (5) &    \phantom{0}6.82 $\pm$ \phantom{0}0.00 (4) &    \phantom{00}3.68 $\pm$ \phantom{0}1.00 (3) &    \phantom{00}0.26 $\pm$ 0.00 (2) &  \bftab 0.00 $\pm$ 0.00 (1) \\
Raisins                            &        \phantom{0}30.61 $\pm$ \phantom{0}2.00 (5) &    \phantom{0}8.50 $\pm$ \phantom{0}1.00 (4) &    \phantom{00}0.62 $\pm$ \phantom{0}0.00 (3) &    \phantom{00}0.20 $\pm$ 0.00 (2) &  \bftab 0.00 $\pm$ 0.00 (1) \\
Rice                               &        \phantom{0}21.49 $\pm$ \phantom{0}4.00 (5) &   11.96 $\pm$ \phantom{0}1.00 (4) &    \phantom{00}1.33 $\pm$ \phantom{0}0.00 (3) &    \phantom{00}0.48 $\pm$ 0.00 (2) &  \bftab 0.01 $\pm$ 0.00 (1) \\
\bottomrule
\end{tabular}%
}
\label{tab:eval-results-runtime}
\end{table}
\newpage

\section{Hyperparameters}\label{A:hyperparams}
In the following, we report the hyperparameters used for each approach. The hyperparameters were selected based on a random search over a predefined parameter range for GradTree, CART and GeneticTree and are summarized in Table~\ref{tab:hpos_GradTree} to Table~\ref{tab:hpos_sklearn}. All parameters that were considered are noted in the tables. The number of trials ($300$) was equal for each approach. For GradTree and DNDT, we did not optimize the batch size as well as the number of epochs, but used early stopping with a predefined patience of $200$. Additionally, we used $3$ random restarts to prevent bad initial parametrizations and selected the best model based on the validation loss.
A gradient-based optimization allows using an arbitrary loss function for the optimization. During preliminary experiments, we observed that it is beneficial to adjust the loss function for specific datasets. More specifically, we allowed adjusting the cross-entropy loss by adding a focal factor~\citep{lin2017focal} of $3$. Additionally, we considered using PolyLoss~\citep{leng2022polyloss} within the HPO to tailor the loss function for the specific task. %
For GeneticTree, we used a complexity penalty as it is commonly used in genetic algorithms. This has a strong impact on the tree size of the learned DT and can explain the small complexity of GeneticTree. To ensure that this complexity penalty does have a significant impact on the performance of the method, we conducted an additional experiment comparing GeneticTree with and without complexity penalty (see Table~\ref{tab:genetictree_comparison}). Removing the complexity penalty in the genetic algorithm resulted in a small increase of the average performance to $0.713$ (GradTree $0.758$) for binary and $0.573$ (GradTree $0.619$) for multi-class. However, removing the complexity penalty also led to a significant increase in the tree size (e.g. from 7 to 1,600 for binary classification). Thus, interpretability of the models is substantially reduced and runtimes substantially increased. Most importantly, the change in the performance does not impact the overall results: When excluding the complexity penalty of the genetic benchmark completely, the relative performance of the methods (number of wins and rank) remained unchanged.
For DL8.5 the relevant tunable hyperparameters according to the authors are the maximum depth and the minimum support. However, the maximum depth strongly impacts the runtime, which is why we fixed the maximum depth to $4$, similar to the maximum depth used during the experiments of \citet{murtree} and \citet{aglin2020learning}. Running the experiments with a higher depth becomes infeasible for many datasets. In preliminary experiments, we also observed that changing the depth does not have a positive impact on the performance (e.g. increasing the depth to $5$ results in a decrease in the test performance). Similarly, reducing the depth resulted in a reduced performance. Furthermore, to assure a fair comparison, we fixed the minimum support to $1$ which is equal to the pruning of GradTree. Additionally, we observed in our preliminary experiments, that increasing the minimum support reduces overfitting, but has no positive impact on the test performance (i.e. the train performance decreases, but the test performance is not improved).
For DNDT, the number of cut points is the tunable hyperparameter of the model, according to \citet{dndt}. However, it has to be restricted to $1$ in order to generate binary trees for comparability reasons. Therefore, we only optimized the temperature and the learning rate.
Furthermore, we extended their implementation to use early stopping based on the validation loss, similar to GradTree, to reduce the runtime and prevent overfitting.
Further details can be found in the code, which we provided in the supplementary material.

\begin{table}[t]
    \centering
    \small
    \caption[GeneticTree Comparison]{\textbf{GeneticTree Comparison.} We compare GeneticTree with and without complexity penalty.}
    
    \begin{tabular}{lrrrr}
         \toprule
         & \multicolumn{2}{l}{Binary} & \multicolumn{2}{l}{Multi} \\ \cmidrule(lr){2-3} \cmidrule(lr){4-5}
         & \multicolumn{1}{l}{Penalty} & \multicolumn{1}{l}{No Penalty} & \multicolumn{1}{l}{Penalty} & \multicolumn{1}{l}{No Penalty} \\\midrule
         F1-Score   &  0.671 & 0.713 &  0.546 & 0.573 \\
         Tree Size  &  \phantom{0,00}7 & 1,600 &  \phantom{0,0}20 & 2,201  \\
         \bottomrule
    \end{tabular}
\label{tab:genetictree_comparison}
    
\end{table}

\begin{table}[t]
\centering
\caption[GradTree Hyperparameters]{\textbf{GradTree Hyperparameters}}

\resizebox{\columnwidth}{!}{
\begin{tabular}{lrrrrlllr}
\toprule
 Dataset Name &  depth &  lr\_index &  lr\_values &  lr\_leaf & activation &                      loss &  polyLoss &  epsilon \\
\midrule
                 Blood Transfusion &      8 &      0.010 &       0.100 &     0.010 &                   entmax &              crossentropy &     False &                2 \\
           Banknote Authentication &      7 &      0.050 &       0.050 &     0.100 &                   entmax & focal\_crossentropy &      True &                2 \\
                           Titanic &     10 &      0.005 &       0.010 &     0.010 &                   entmax &              crossentropy &     False &                2 \\
                           Raisins &     10 &      0.005 &       0.005 &     0.100 &                   entmax &              crossentropy &      True &                5 \\
                              Rice &      7 &      0.050 &       0.010 &     0.010 &                   entmax &              crossentropy &     False &                2 \\
                    Echocardiogram &      8 &      0.010 &       0.050 &     0.100 &                   entmax &              crossentropy &      True &                5 \\
Wisconsin Breast Cancer &     10 &      0.050 &       0.010 &     0.100 &                   entmax &              crossentropy &      True &                2 \\
                        Loan House &      8 &      0.005 &       0.100 &     0.010 &                   entmax & focal\_crossentropy &     False &                2 \\
                     Heart Failure &     10 &      0.005 &       0.250 &     0.100 &                   entmax &              crossentropy &     False &                2 \\
                     Heart Disease &      9 &      0.010 &       0.050 &     0.005 &                   entmax & focal\_crossentropy &     False &                2 \\
                             Adult &      8 &      0.050 &       0.005 &     0.050 &                   entmax &              crossentropy &      True &                5 \\
                    Bank Marketing &      8 &      0.250 &       0.250 &     0.050 &                   entmax &              crossentropy &     False &                2 \\
                   Cervical Cancer &      8 &      0.005 &       0.010 &     0.100 &                   entmax &              crossentropy &      True &                2 \\
              Congressional Voting &     10 &      0.005 &       0.050 &     0.010 &                   entmax & focal\_crossentropy &      True &                5 \\
                       Absenteeism &     10 &      0.050 &       0.010 &     0.050 &                   entmax & focal\_crossentropy &      True &                5 \\
                         Hepatitis &     10 &      0.005 &       0.050 &     0.010 &                   entmax & focal\_crossentropy &      True &                5 \\
                            German &      7 &      0.005 &       0.050 &     0.010 &                   entmax &              crossentropy &      True &                2 \\
                          Mushroom &      9 &      0.010 &       0.010 &     0.050 &                   entmax & focal\_crossentropy &     False &                2 \\
                       Credit Card &      8 &      0.050 &       0.100 &     0.010 &                   entmax & focal\_crossentropy &     False &                2 \\
                       Horse Colic &      8 &      0.250 &       0.250 &     0.010 &                   entmax & focal\_crossentropy &     False &                2 \\
                           Thyroid &      8 &      0.010 &       0.010 &     0.050 &                   entmax &              crossentropy &     False &                2 \\
                          Spambase &     10 &      0.005 &       0.010 &     0.010 &                   entmax &              crossentropy &     False &                2 \\
\midrule
         Iris &      7 &      0.005 &       0.005 &      0.05 &                   entmax &              crossentropy &     False &                2 \\
Balance Scale &      8 &      0.050 &       0.010 &      0.10 &                   entmax &              crossentropy &      True &                5 \\
          Car &      9 &      0.010 &       0.010 &      0.01 &                   entmax & focal\_crossentropy &     False &                2 \\
        Glass &     10 &      0.050 &       0.050 &      0.05 &                   entmax & focal\_crossentropy &      True &                5 \\
Contraceptive &      7 &      0.010 &       0.050 &      0.01 &                   entmax &              crossentropy &     False &                2 \\
  Solar Flare &      8 &      0.005 &       0.010 &      0.20 &                   entmax & focal\_crossentropy &      True &                2 \\
         Wine &     10 &      0.010 &       0.050 &      0.01 &                   entmax & focal\_crossentropy &     False &                2 \\
          Zoo &      9 &      0.050 &       0.010 &      0.10 &                   entmax & focal\_crossentropy &      True &                2 \\
 Lymphography &      8 &      0.050 &       0.010 &      0.05 &                   entmax &              crossentropy &      True &                2 \\
      Segment &      7 &      0.005 &       0.005 &      0.05 &                   entmax &              crossentropy &     False &                2 \\
  Dermatology &      7 &      0.010 &       0.010 &      0.10 &                   entmax &              crossentropy &      True &                2 \\
      Landsat &      8 &      0.005 &       0.010 &      0.05 &                   entmax &              crossentropy &      True &                5 \\
    Annealing &     10 &      0.250 &       0.050 &      0.01 &                   entmax &              crossentropy &     False &                2 \\
       Splice &      9 &      0.010 &       0.005 &      0.05 &                   entmax &              crossentropy &     False &                2 \\
\bottomrule
\end{tabular}%
}

\label{tab:hpos_GradTree}
\end{table}

\begin{table}[t]
\centering
\small
\caption[DNDT Hyperparameters]{\textbf{DNDT Hyperparameters}}

\resizebox{0.8\columnwidth}{!}{
\begin{tabular}{lrrr}
\toprule
                      Dataset Name &  learning\_rate &  temperature & num\_cut  \\
\midrule
                 Blood Transfusion &           0.050 &        0.100 &     1  \\
           Banknote Authentication &           0.001 &        0.100 &     1  \\
                           Titanic &           0.050 &        0.001 &     1  \\
                           Raisins &           0.050 &        0.001 &     1  \\
                              Rice &           0.100 &        0.010 &     1  \\
                    Echocardiogram &           0.005 &        1.000 &     1  \\
Wisconsin Diagnostic Breast Cancer &           0.100 &        0.010 &     1  \\
                        Loan House &           0.001 &        1.000 &     1  \\
                     Heart Failure &           0.005 &        1.000 &     1  \\

\midrule
         Iris &           0.050 &         0.100 &     1  \\
Balance Scale &           0.005 &         0.100 &     1  \\
          Car &           0.010 &         0.010 &     1  \\
        Glass &           0.100 &         0.010 &     1  \\
Contraceptive &           0.001 &         0.010 &     1  \\
  Solar Flare &           0.050 &         0.100 &     1  \\
         Wine &           0.001 &         0.100 &     1  \\
\bottomrule
\end{tabular}%
}

\label{tab:hpos_dndt}
\end{table}

\begin{table}[t]
\centering
\small
\caption[GeneticTree Hyperparameters]{\textbf{GeneticTree Hyperparameters}}

\resizebox{0.9\columnwidth}{!}{
\begin{tabular}{lrrrrr}

\toprule
                      Dataset Name &  n\_thresholds &  n\_trees &  max\_iter &  cross\_prob &  mutation\_prob \\
\midrule
                 Blood Transfusion &             10 &       500 &        500 &          1.0 &             0.2  \\
           Banknote Authentication &             10 &       500 &        500 &          0.8 &             0.6  \\
                           Titanic &             10 &       500 &        250 &          0.2 &             0.3  \\
                           Raisins &             10 &       100 &         50 &          1.0 &             0.6  \\
                              Rice &             10 &       100 &        250 &          0.4 &             0.3  \\
                    Echocardiogram &             10 &        50 &        100 &          0.8 &             0.4  \\
Wisconsin Breast Cancer &             10 &       250 &        100 &          0.2 &             0.7  \\
                        Loan House &             10 &       500 &        500 &          1.0 &             0.2  \\
                     Heart Failure &             10 &       500 &         50 &          0.4 &             0.6  \\
                     Heart Disease &             10 &       400 &        500 &          0.6 &             0.4  \\
                             Adult &             10 &       100 &        500 &          0.6 &             0.6  \\
                    Bank Marketing &             10 &       500 &        250 &          0.8 &             0.6  \\
                   Cervical Cancer &             10 &       100 &        500 &          0.4 &             0.9  \\
              Congressional Voting &             10 &       500 &        500 &          1.0 &             0.7  \\
                       Absenteeism &             10 &       500 &        500 &          1.0 &             0.7  \\
                         Hepatitis &             10 &       500 &        250 &          0.2 &             0.3  \\
                            German &             10 &       250 &        500 &          0.4 &             0.8  \\
                          Mushroom &             10 &       400 &        500 &          0.6 &             0.4  \\
                       Credit Card &             10 &        50 &         50 &          0.4 &             0.1  \\
                       Horse Colic &             10 &        50 &        100 &          0.8 &             0.4  \\
                           Thyroid &             10 &       500 &        500 &          1.0 &             0.7  \\
                          Spambase &             10 &       250 &        500 &          0.8 &             0.6  \\
\midrule
         Iris &             10 &       100 &        500 &          1.0 &             0.5  \\
Balance Scale &             10 &       500 &        250 &          0.6 &             0.4  \\
          Car &             10 &       500 &        500 &          1.0 &             0.7  \\
        Glass &             10 &       500 &        500 &          0.8 &             0.6  \\
Contraceptive &             10 &       250 &        500 &          0.4 &             0.5  \\
  Solar Flare &             10 &       500 &        250 &          0.6 &             0.4  \\
         Wine &             10 &       100 &         50 &          0.2 &             0.8  \\
          Zoo &             10 &       500 &        500 &          1.0 &             0.7  \\
 Lymphography &             10 &       500 &        250 &          0.6 &             0.4  \\
      Segment &             10 &       100 &        500 &          0.4 &             0.9  \\
  Dermatology &             10 &       100 &        500 &          1.0 &             0.5  \\
      Landsat &             10 &       500 &        500 &          1.0 &             0.7  \\
    Annealing &             10 &       100 &        500 &          0.4 &             0.9  \\
       Splice &             10 &       100 &        250 &          0.2 &             0.8  \\
\bottomrule
\end{tabular}%
}%

\label{tab:hpos_gentree}
\end{table}

\begin{table}[t]
\centering
\small
\caption[CART Hyperparameters]{\textbf{CART Hyperparameters}}

\resizebox{0.85\columnwidth}{!}{
\begin{tabular}{lrllrrr}
\toprule
                      Dataset Name &  max\_depth & criterion & max\_features &  samples\_leaf &  samples\_split &  ccp\_alpha \\
\midrule
                 Blood Transfusion &           9 &   entropy &          None &                   1 &                    5 &         0.0 \\
           Banknote Authentication &           9 &      gini &          None &                   1 &                    5 &         0.0 \\
                           Titanic &           7 &   entropy &          None &                   1 &                   50 &         0.0 \\
                           Raisins &           8 &      gini &          None &                   5 &                    2 &         0.2 \\
                              Rice &           8 &      gini &          None &                   5 &                    2 &         0.2 \\
                    Echocardiogram &           9 &   entropy &          None &                   1 &                    5 &         0.0 \\
Wisconsin Breast Cancer &           7 &   entropy &          None &                   5 &                    2 &         0.4 \\
                        Loan House &          10 &   entropy &          None &                  10 &                    2 &         0.0 \\
                     Heart Failure &           9 &      gini &          None &                   5 &                   50 &         0.0 \\
                     Heart Disease &           8 &   entropy &          None &                   5 &                   10 &         0.0 \\
                             Adult &          10 &   entropy &          None &                  10 &                    2 &         0.0 \\
                    Bank Marketing &           8 &   entropy &          None &                   5 &                   10 &         0.0 \\
                   Cervical Cancer &           9 &   entropy &          None &                   1 &                    5 &         0.0 \\
              Congressional Voting &          10 &      gini &          None &                   1 &                    2 &         0.0 \\
                       Absenteeism &           7 &   entropy &          None &                   1 &                   10 &         0.0 \\
                         Hepatitis &           9 &   entropy &          None &                   1 &                    5 &         0.0 \\
                            German &           7 &   entropy &          None &                   1 &                   10 &         0.0 \\
                          Mushroom &           9 &   entropy &          None &                   1 &                    5 &         0.0 \\
                       Credit Card &           9 &      gini &          None &                   1 &                    5 &         0.0 \\
                       Horse Colic &          10 &   entropy &          None &                  10 &                    2 &         0.0 \\
                           Thyroid &          10 &   entropy &          None &                  10 &                    2 &         0.0 \\
                          Spambase &          10 &      gini &          None &                   1 &                    2 &         0.0 \\
\midrule
         Iris &           8 &   entropy &          None &                   5 &                   10 &         0.0 \\
Balance Scale &           9 &   entropy &          None &                   1 &                    5 &         0.0 \\
          Car &           9 &      gini &          None &                   1 &                    5 &         0.0 \\
        Glass &           9 &      gini &          None &                   1 &                    5 &         0.0 \\
Contraceptive &           9 &   entropy &          None &                   1 &                    5 &         0.0 \\
  Solar Flare &           7 &   entropy &          None &                   1 &                   50 &         0.0 \\
         Wine &           9 &      gini &          None &                   1 &                    5 &         0.0 \\
          Zoo &          10 &      gini &          None &                   1 &                    2 &         0.0 \\
 Lymphography &           7 &   entropy &          None &                   1 &                   10 &         0.0 \\
      Segment &           9 &   entropy &          None &                   1 &                    5 &         0.0 \\
  Dermatology &           9 &      gini &          None &                   1 &                    5 &         0.0 \\
      Landsat &          10 &      gini &          None &                   1 &                    2 &         0.0 \\
    Annealing &           9 &      gini &          None &                   1 &                    5 &         0.0 \\
       Splice &           9 &      gini &          None &                   1 &                    5 &         0.0 \\
\bottomrule
\end{tabular}%
}

\label{tab:hpos_sklearn}
\end{table}

\chapter{Appendix GRANDE} %
\label{A:GRANDE} %

\section{Instance-Wise Weighting Statistics} \label{A:weighting}

\begin{table}[t]%
\centering
\caption[Weighting Statistics Estimator]{\textbf{Weighting Statistics Estimator.} This table shows summarizing statistics on the instance-wise weighting for each dataset.}
\resizebox{\columnwidth}{!}{
\begin{tabular}{lrrrrrr}
\toprule
 & \begin{tabular}[c]{@{}l@{}}Number\\Internal\\Nodes\end{tabular} & \begin{tabular}[c]{@{}l@{}}Number\\Leaf\\Nodes\end{tabular} & \begin{tabular}[c]{@{}l@{}}Percentage\\Highest\\Estimator\end{tabular}  & \begin{tabular}[c]{@{}l@{}}Percentage\\Highest\\Estimator\\Top 5\%\end{tabular} & \begin{tabular}[c]{@{}l@{}}Count\\Highest\\ Estimator\end{tabular} & \begin{tabular}[c]{@{}l@{}}Count\\Highest\\ Estimator\\Top 5\%\end{tabular} \\
\midrule
dresses-sales & 1.12 & 2.12 & 0.470 & 0.667 & 10.20 & 2.40 \\
climate-model-simulation & 3.12 & 4.12 & 0.241 & 0.467 & 15.80 & 4.00 \\
cylinder-bands & 5.44 & 6.44 & 0.124 & 0.400 & 31.20 & 4.00 \\
wdbc & 1.16 & 2.16 & 0.587 & 0.733 & 4.20 & 1.60 \\
ilpd & 3.96 & 4.96 & 0.319 & 0.600 & 15.00 & 3.40 \\
tokyo1 & 3.20 & 4.20 & 0.273 & 0.780 & 19.80 & 2.00 \\
qsar-biodeg & 2.80 & 3.80 & 0.191 & 0.745 & 29.80 & 2.60 \\
ozone-level-8hr & 2.00 & 3.00 & 0.198 & 0.662 & 27.80 & 5.40 \\
madelon & 3.92 & 4.92 & 0.322 & 0.741 & 19.60 & 3.20 \\
Bioresponse & 3.96 & 4.96 & 0.099 & 0.311 & 82.00 & 11.80 \\
wilt & 1.76 & 2.76 & 0.274 & 0.604 & 19.20 & 5.00 \\
churn & 2.28 & 3.28 & 0.111 & 0.408 & 72.00 & 11.40 \\
phoneme & 3.28 & 4.28 & 0.163 & 0.636 & 40.80 & 4.40 \\
SpeedDating & 6.96 & 7.96 & 0.168 & 0.640 & 90.20 & 5.00 \\
PhishingWebsites & 5.60 & 6.60 & 0.178 & 0.594 & 116.80 & 12.20 \\
Amazon\_employee\_access & 0.24 & 1.24 & 0.882 & 0.999 & 2.80 & 1.20 \\
nomao & 3.20 & 4.20 & 0.113 & 0.260 & 218.20 & 15.80 \\
adult & 0.36 & 1.36 & 0.905 & 0.912 & 5.20 & 1.60 \\
numerai28.6 & 1.08 & 2.08 & 0.559 & 0.824 & 19.60 & 4.40 \\ \midrule
Mean & 2.92 & 3.92 & 0.325 & 0.631 & 44.22 & 5.35 \\
\bottomrule
\end{tabular}
}
\label{tab:statistics_estimators}
\end{table}

\begin{table}[t]%
\centering
\small
\caption[Weighting Statistics Distribution]{\textbf{Weighting Statistics Distribution.} We provide the skewness, as measure of the asymmetry of a distribution, and kurtosis, as measure of the tailedness of a distribution, for all datasets. The measures are split in an average over all samples and an average over the top 5\% of the samples with the highest maximum weight.}
\resizebox{0.9\textwidth}{!}{
\begin{tabular}{lrrrr}
\toprule
 & Skewness & Skewness Top 5\% & Kurtosis & Kurtosis Top 5\% \\
\midrule
dresses-sales & 0.946 & 1.187 & 0.617 & 1.597 \\
climate-model-simulation & 1.259 & 1.909 & 0.742 & 3.095 \\
cylinder-bands & 0.853 & 1.199 & -0.562 & 0.230 \\
wdbc & 0.082 & 0.251 & -1.084 & -0.980 \\
ilpd & 0.180 & 0.385 & -0.649 & -0.528 \\
tokyo1 & 0.975 & 1.839 & 1.322 & 4.919 \\
qsar-biodeg & 0.875 & 1.194 & 0.200 & 1.302 \\
ozone-level-8hr & 0.752 & 1.462 & -0.048 & 1.972 \\
madelon & 1.901 & 2.208 & 2.797 & 4.465 \\
Bioresponse & 0.777 & 1.139 & -0.496 & 0.317 \\
wilt & 0.508 & 1.150 & -0.499 & 1.100 \\
churn & 2.369 & 3.488 & 5.912 & 13.419 \\
phoneme & 0.563 & 0.789 & -0.447 & 0.119 \\
SpeedDating & 1.069 & 1.180 & 1.148 & 1.627 \\
PhishingWebsites & 1.263 & 1.978 & 1.136 & 4.343 \\
Amazon\_employee\_access & -0.029 & -0.119 & 3.348 & 3.081 \\
nomao & 2.411 & 4.190 & 6.920 & 24.791 \\
adult & 0.621 & 0.679 & 3.345 & 3.280 \\
numerai28.6 & -0.297 & -0.269 & 0.833 & 0.829 \\ \midrule
Mean & 0.899 & 1.360 & 1.291 & 3.630 \\
\bottomrule
\end{tabular}
}
\label{tab:statistics_distribution}
\end{table}

In the following, we provide a more detailed evaluation of the instance-wise weighting, including statistics on the highest weighted estimators and the weight distributions.
The following statistics were obtained by calculating the post-softmax weights for each sample and averaging the values over all samples. In addition, we calculated the same statistics for the top 5\% of samples with the highest weights. This provides additional insights, since we argue that unique local interactions might exist only for a subset of the samples.
We can observe that on average, the highest weighted estimator comprises a moderate number of $\approx$3 internal and $\approx 4$ leaf nodes, allowing an easy interpretation (see Table~\ref{tab:statistics_estimators}). Furthermore, on average, the highest weighted estimator is the same for $\approx 33\%$ of the data instances, with a total of $\approx 44$ different estimators having the highest weight for at least one instance. If we further inspect the top $5\%$ of instances with the highest weight for a single estimator, we can additionally observe that the highest weighted estimator is the same for $\approx 63\%$ of these instances and only $\approx 5$ different estimators have the highest weight for at least one instance. This suggests the presence of a limited number of local experts in most datasets, which have high weights for specific instances where local rules are applicable.
In addition, Table~\ref{tab:statistics_distribution} summarizes the skewness as measure of the asymmetry of a distribution, and kurtosis as measure of the tailedness of a distribution, for all datasets. In general, it stands out that both values increase substantially when considering only the top 5\% of samples with the highest weight for a single estimator. This again indicates the presence of local expert estimators for a subset of the data, where unique local interactions were identified. In addition, there are major differences in the values depending on the datasets, which we will discuss more detailed in the following:

\begin{table}[t]%
\centering
\small
\caption[Skewness Summary]{\textbf{Skewness Summary.} This table summarizes the skewness of estimator weights in different categories based on the magnitude of skewness on each dataset.}
\begin{tabular}{|l|r|r|r|r|}
\hline
 & \begin{tabular}[c]{@{}l@{}}Very skewed\\ $(-\infty, -1.0)$ or $(1.0, \infty)$\end{tabular}  & \begin{tabular}[c]{@{}l@{}}Skewed\\ $[-1.0, -0.5)$ or $(0.5, 1.0]$\end{tabular}  & \begin{tabular}[c]{@{}l@{}}Slight skew\\ $[-0.5, 0.5]$\end{tabular}  & \textbf{Sum} \\
\hline
Left skew & 0 & 1 & 1 & \textbf{2} \\ \hline
Right skew & 6 & 9 & 2 & \textbf{17} \\
\hline
\textbf{Sum} & \textbf{6} & \textbf{10} & \textbf{3} &  \\
\hline
\end{tabular}
\label{tab:skewness}
\end{table}

\begin{table}[t]%
\centering
\small
\caption[Skewness Top 5\%]{\textbf{Skewness Top 5\%.} This table summarizes the skewness of estimator weights in different categories based on the magnitude of skewness on each dataset, considering only the top 5\% of samples with the highest weight for a single estimator.}
\begin{tabular}{|l|r|r|r|r|}
\hline
 & \begin{tabular}[c]{@{}l@{}}Very skewed\\ $(-\infty, -1.0)$ or $(1.0, \infty)$\end{tabular}  & \begin{tabular}[c]{@{}l@{}}Skewed\\ $[-1.0, -0.5)$ or $(0.5, 1.0]$\end{tabular}  & \begin{tabular}[c]{@{}l@{}}Slight skew\\ $[-0.5, 0.5]$\end{tabular}  & Sum \\
\hline
Left skew & 0 & 1 & 1 & \textbf{2} \\ \hline
Right skew & 13 & 1 & 3 & \textbf{17} \\
\hline
\textbf{Sum} & \textbf{13} & \textbf{2} & \textbf{4} &  \\
\hline
\end{tabular}
\label{tab:skewness_top5}
\end{table}

\paragraph{Skewness}
In total, for 16/19 datasets the weights have a skewed (10) or very skewed (6) distribution, i.e., are long-tailed (see Table~\ref{tab:skewness}).
Generally, left-skewed distributions suggest a few estimators with very low weights, thereby contributing substantially less and resulting in a more compact ensemble.
However, right-skewed distributions are more desired as they indicate a small number of trees having high weights (which we consider as local experts), while the majority of estimators have small weights. 
Considering the top 5\% of samples with the highest weights of a single estimator, we can see that the number of very skewed distributions increases from 6 to 13 (see Table~\ref{tab:skewness_top5}), indicating samples that can be assigned to a class more easily based on a small number of trees (similar to the example in the case study).
In general, we are most interested in right-skewed distributions as this indicates a small number of estimators with high weights, which we can consider as local experts.

\begin{table}[t]%
\centering
\small
\caption[Kurtosis]{\textbf{Kurtosis.} This table categorizes the kurtosis of the weighting distribution for each dataset.}
\begin{tabular}{lrr}
\toprule
Excess Kurtosis & Count & Count Top 5\% \\
\midrule
Very leptokurtic $(3.0, \infty)$ & 4 & 8 \\
Moderately leptokurtic $(0.5, 3.0]$ & 7 & 6 \\
Mesokurtic $[-0.5, 0.5]$ & 5 & 3 \\
Platykurtic $(-\infty, -0.5)$ & 3 & 2 \\
\bottomrule
\end{tabular}
\label{tab:kurtosis}
\end{table}

\paragraph{(Excess) Kurtosis}
We can observe that 11/19 distributions are leptokurtic, i.e., have heavy tails, with 4 considered as very leptokurtic (see Table~\ref{tab:kurtosis}). In general, distributions with heavy tails are interesting, as this indicates outliers (estimators with substantially higher / lower weights). Trees comprising substantially higher weights can be considered as local experts that have learned unique local interactions. Again, when considering the top 5\% of samples with the highest weights of a single estimator, the number of leptokurtic distributions increases from 11 to 15 with the number of very leptokurtic distributions increasing from 4 to 8. Again, this indicates that there exists samples that can be assigned to a class more easily with simple rules (from local experts) in many datasets.
Overall, the additional statistics are in line with the intuitive justifications for an implementation of instance-wise weighting. For several datasets, the distribution of weight is long- and/or heavy-tailed, indicating the presence of local expert estimators. However, the benefit of learning local expert estimators is dataset-dependent and not feasible in all scenarios.
This is also confirmed by our results, as there are datasets where the distribution is symmetric and mesokurtic. This is valid for instance for numerai28.6, which is a very complex dataset without many simple rules (as indicated by the poor performance of all methods). In addition, it should be noted that a more comprehensive analysis is necessary to confirm these additional findings.

\clearpage

\section{Additional Results}\label{A:additional_results}

\begin{figure}[H]
    \centering
    \includegraphics[width=0.5\linewidth]{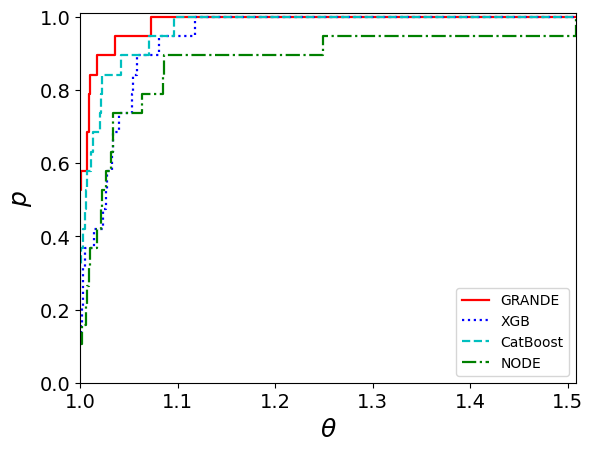}
    \caption[Performance Profile (HPO 250 Trials)]{\textbf{Performance Profile (HPO 250 Trials)}. The performance profile is based on the macro F1-Score with optimized hyperparameters (complete grids, 250 trials). The x-axis represents a tolerance factor, and the y-axis is a proportion of the evaluated datasets.}
    \label{fig:performane_profile}
\end{figure}

\vspace{0.5cm}

\subsection{Additional Metrics}

\begin{table}[H]
\centering
\small
\caption[ROC-AUC Performance Comparison (HPO 250 Trials)]{\textbf{ROC-AUC Performance Comparison (HPO 250 Trials).} We report the test ROC-AUC (mean $\pm$ stdev for a 5-fold CV) with optimized parameters (complete grids, 250 trials) and the ranking of each approach in parentheses. The datasets are sorted based on the data size.}
\resizebox{\columnwidth}{!}{
\begin{tabular}{lcccc}
\toprule
{} &                        \multicolumn{1}{l}{GRANDE} &                           \multicolumn{1}{l}{XGB} &                      \multicolumn{1}{l}{CatBoost} &                          \multicolumn{1}{l}{NODE} \\
\midrule
dresses-sales                    &         0.623 $\pm$ 0.049 (2) &         0.589 $\pm$ 0.060 (4) &         0.607 $\pm$ 0.040 (3) &  \bftab 0.632 $\pm$ 0.041 (1) \\
climate-simulation-crashes &  \bftab 0.958 $\pm$ 0.031 (1) &         0.923 $\pm$ 0.031 (4) &         0.938 $\pm$ 0.036 (2) &         0.933 $\pm$ 0.030 (3) \\
cylinder-bands                   &  \bftab 0.896 $\pm$ 0.032 (1) &         0.872 $\pm$ 0.019 (3) &         0.879 $\pm$ 0.046 (2) &         0.837 $\pm$ 0.037 (4) \\
wdbc                             &  \bftab 0.993 $\pm$ 0.006 (1) &         0.990 $\pm$ 0.010 (4) &         0.990 $\pm$ 0.010 (3) &         0.993 $\pm$ 0.007 (2) \\
ilpd                             &  \bftab 0.748 $\pm$ 0.046 (1) &         0.721 $\pm$ 0.047 (4) &         0.728 $\pm$ 0.055 (3) &         0.745 $\pm$ 0.048 (2) \\
tokyo1                           &         0.983 $\pm$ 0.005 (2) &         0.982 $\pm$ 0.006 (3) &  \bftab 0.984 $\pm$ 0.005 (1) &         0.980 $\pm$ 0.005 (4) \\
qsar-biodeg                      &  \bftab 0.934 $\pm$ 0.008 (1) &         0.925 $\pm$ 0.008 (3) &         0.933 $\pm$ 0.011 (2) &         0.920 $\pm$ 0.009 (4) \\
\bottomrule
\end{tabular}
}
\label{tab:eval-results_binary_test_optimized_roc_auc}
\end{table}

\begin{table}[H]
\centering
\small
\caption[Accuracy Performance Comparison (HPO 250 Trials)]{\textbf{Accuracy Performance Comparison (HPO 250 Trials).} We report the test balanced accuracy (mean $\pm$ stdev for a 5-fold CV) with optimized parameters (complete grids, 250 trials) and the ranking of each approach in parentheses. The datasets are sorted based on the data size.}
\resizebox{\columnwidth}{!}{
\begin{tabular}{lcccc}
\toprule
{} &                        \multicolumn{1}{l}{GRANDE} &                           \multicolumn{1}{l}{XGB} &                      \multicolumn{1}{l}{CatBoost} &                          \multicolumn{1}{l}{NODE} \\
\midrule
dresses-sales                    &  \bftab 0.613 $\pm$ 0.048 (1) &         0.585 $\pm$ 0.060 (3) &         0.589 $\pm$ 0.037 (2) &  0.581 $\pm$ 0.037 (4) \\
climate-simulation-crashes &  \bftab 0.875 $\pm$ 0.079 (1) &         0.762 $\pm$ 0.060 (4) &         0.832 $\pm$ 0.035 (2) &  0.773 $\pm$ 0.052 (3) \\
cylinder-bands                   &  \bftab 0.817 $\pm$ 0.032 (1) &         0.777 $\pm$ 0.045 (3) &         0.798 $\pm$ 0.042 (2) &  0.751 $\pm$ 0.040 (4) \\
wdbc                             &  \bftab 0.973 $\pm$ 0.010 (1) &         0.952 $\pm$ 0.029 (4) &         0.963 $\pm$ 0.023 (3) &  0.963 $\pm$ 0.017 (2) \\
ilpd                             &  \bftab 0.709 $\pm$ 0.046 (1) &         0.673 $\pm$ 0.042 (3) &         0.689 $\pm$ 0.062 (2) &  0.548 $\pm$ 0.054 (4) \\
tokyo1                           &         0.925 $\pm$ 0.002 (2) &         0.918 $\pm$ 0.012 (4) &  \bftab 0.932 $\pm$ 0.010 (1) &  0.920 $\pm$ 0.007 (3) \\
qsar-biodeg                      &         0.853 $\pm$ 0.024 (2) &  \bftab 0.856 $\pm$ 0.022 (1) &         0.847 $\pm$ 0.028 (3) &  0.831 $\pm$ 0.032 (4) \\
\bottomrule
\end{tabular}
}
\label{tab:eval-results_binary_test_optimized_acc}
\end{table}

\vspace{0.5cm}

\subsection{Runtime}

\begin{table}[H]
\centering
\small
\caption[Runtime Performance Comparison]{\textbf{Runtime Performance Comparison.} We report the runtime (mean $\pm$ stdev for a 5-fold CV) with optimized parameters (complete grids, 250 trials) and the ranking of each approach in parentheses. The datasets are sorted based on the data size. For all methods, we used a single NVIDIA RTX A6000.}
\resizebox{\columnwidth}{!}{
\begin{tabular}{lcccc}
\toprule
{} &                        \multicolumn{1}{l}{GRANDE} &                           \multicolumn{1}{l}{XGB} &                      \multicolumn{1}{l}{CatBoost} &                          \multicolumn{1}{l}{NODE} \\
\midrule
dresses-sales                    &    \phantom{0}11.121 $\pm$ \phantom{0}1.0 (3) &  \bftab 0.052 $\pm$ 0.0 (1) &   \phantom{0}4.340 $\pm$ 2.0 (2) &       \phantom{0}31.371 $\pm$ \phantom{0}17.0 (4) \\
climate-simulation-crashes &    \phantom{0}16.838 $\pm$ \phantom{0}3.0 (3) &  \bftab 0.077 $\pm$ 0.0 (1) &   \phantom{0}8.740 $\pm$ 9.0 (2) &       \phantom{0}97.693 $\pm$ \phantom{0}10.0 (4) \\
cylinder-bands                   &    \phantom{0}20.554 $\pm$ \phantom{0}2.0 (3) &  \bftab 0.093 $\pm$ 0.0 (1) &   \phantom{0}4.887 $\pm$ 2.0 (2) &       \phantom{0}38.983 $\pm$ \phantom{0}12.0 (4) \\
wdbc                             &    \phantom{0}29.704 $\pm$ \phantom{0}4.0 (3) &  \bftab 0.151 $\pm$ 0.0 (1) &   \phantom{0}1.046 $\pm$ 0.0 (2) &       \phantom{0}83.548 $\pm$ \phantom{0}16.0 (4) \\
ilpd                             &    \phantom{0}11.424 $\pm$ \phantom{0}1.0 (3) &  \bftab 0.049 $\pm$ 0.0 (1) &   \phantom{0}3.486 $\pm$ 3.0 (2) &       \phantom{0}59.085 $\pm$ \phantom{0}24.0 (4) \\
tokyo1                           &    \phantom{0}21.483 $\pm$ \phantom{0}3.0 (3) &  \bftab 0.078 $\pm$ 0.0 (1) &   \phantom{0}1.485 $\pm$ 1.0 (2) &        \phantom{0}84.895 $\pm$ \phantom{00}5.0 (4) \\
qsar-biodeg                      &    \phantom{0}21.565 $\pm$ \phantom{0}2.0 (3) &  \bftab 0.087 $\pm$ 0.0 (1) &   \phantom{0}1.195 $\pm$ 0.0 (2) &       \phantom{0}96.204 $\pm$ \phantom{0}20.0 (4) \\
\bottomrule
\end{tabular}

}
\label{tab:eval-results_RUNTIME}
\end{table}

\subsection{Ablation Study}

\begin{table}[H]
\centering
\small
\caption[Dataset Differences]{\textbf{Dataset Differences.} This table summarizes the performance difference between the base learners GradTree and CART along with their basic ensembling counterparts GRANDE and RandomForest respectively. We report the difference in test macro F1-score (mean $\pm$ stdev over 10 trials) with default parameters.}
\label{tab:gradtree_grande_comparison_difference}
\resizebox{0.8\columnwidth}{!}{

\begin{tabular}{lrr}
\toprule
 & GRANDE - GradTree & RandomForest - CART \\
\midrule
             dresses-sales &                 0.0683 &                   0.0266 \\
climate-simulation-crashes &                 0.2185 &                  -0.1475 \\
            cylinder-bands &                 0.1156 &                   0.0805 \\
                      wdbc &                 0.0369 &                   0.0367 \\
                      ilpd &                 0.0333 &                   0.0168 \\
                    tokyo1 &                 0.0046 &                   0.0129 \\
               qsar-biodeg &                 0.0666 &                   0.0682 \\
           ozone-level-8hr &                 0.1398 &                  -0.0696 \\
                   madelon &                 0.1141 &                  -0.0315 \\
               Bioresponse &                 0.0787 &                   0.0782 \\
                      wilt &                 0.0570 &                  -0.0009 \\
                     churn &                 0.0510 &                   0.0371 \\
                   phoneme &                 0.0680 &                   0.0525 \\
               SpeedDating &                 0.1127 &                  -0.0114 \\
          PhishingWebsites &                 0.0457 &                   0.0081 \\
  Amazon\_employee\_access &                 0.0108 &                   0.0189 \\
                     nomao &                 0.0496 &                   0.0233 \\
                     adult &                 0.0251 &                   0.0397 \\
               numerai28.6 &                -0.0019 &                   0.0072 \\ \midrule
                Mean                        & 0.0681 & 0.0129 \\
\bottomrule
\end{tabular}
}
\label{tab:dataset_differences}
\end{table}

\vspace{0.5cm}

\begin{table}[H]
\centering
\small
\caption[Ablation Study Split Activation]{\textbf{Ablation Study Split Activation.} We report the test macro F1-Score (mean $\pm$ stdev for a 5-fold CV) with optimized parameters (complete grids, 250 trials) and the ranking of each approach in parentheses. The datasets are sorted based on the data size.}
\begin{tabular}{lccc}
\toprule
 & \multicolumn{1}{l}{Softsign} & \multicolumn{1}{l}{Entmoid} & \multicolumn{1}{l}{Sigmoid} \\ 
\midrule
dresses-sales & \bftab 0.612 $\pm$ 0.049 (1) & 0.580 $\pm$ 0.047 (2) & 0.568 $\pm$ 0.052 (3) \\
climate-simulation-crashes & \bftab 0.853 $\pm$ 0.070 (1) & 0.840 $\pm$ 0.038 (2) & 0.838 $\pm$ 0.041 (3) \\
cylinder-bands & \bftab 0.819 $\pm$ 0.032 (1) & 0.802 $\pm$ 0.028 (3) & 0.807 $\pm$ 0.020 (2) \\
wdbc & \bftab 0.975 $\pm$ 0.010 (1) & 0.970 $\pm$ 0.007 (3) & 0.972 $\pm$ 0.008 (2) \\
ilpd & 0.657 $\pm$ 0.042 (2) & 0.652 $\pm$ 0.047 (3) & \bftab 0.663 $\pm$ 0.047 (1) \\
tokyo1 & \bftab 0.921 $\pm$ 0.004 (1) & 0.920 $\pm$ 0.010 (3) & 0.921 $\pm$ 0.008 (2) \\
qsar-biodeg & 0.854 $\pm$ 0.022 (3) & 0.855 $\pm$ 0.018 (2) & \bftab 0.845 $\pm$ 0.018 (1) \\
\bottomrule
\end{tabular}
\label{tab:ablation_study_activation}
\end{table}

\begin{table}[H]
\centering
\small
\caption[Ablation Study Weighting]{\textbf{Ablation Study Weighting.} We report the test macro F1-Score (mean $\pm$ stdev for a 5-fold CV) with optimized parameters (complete grids, 250 trials) and the ranking of each approach in parentheses. The datasets are sorted based on the data size.}
\begin{tabular}{lcc}
\toprule
 & \multicolumn{1}{l}{Leaf Weights} & \multicolumn{1}{l}{Estimator Weights}  \\ 
\midrule
        dresses-sales & \bftab 0.612 $\pm$ 0.049 (1) & 0.605 $\pm$ 0.050 (2) \\
        climate-simulation-crashes & \bftab 0.853 $\pm$ 0.070 (1) & 0.801 $\pm$ 0.040 (2) \\
        cylinder-bands & \bftab 0.819 $\pm$ 0.032 (1) & 0.787 $\pm$ 0.051 (2) \\
        wdbc & \bftab 0.975 $\pm$ 0.010 (1) & 0.970 $\pm$ 0.009 (2) \\
        ilpd & \bftab 0.657 $\pm$ 0.042 (1) & 0.612 $\pm$ 0.064 (2) \\
        tokyo1 & 0.921 $\pm$ 0.004 (2) &\bftab  0.922 $\pm$ 0.016 (1) \\
        qsar-biodeg & \bftab 0.854 $\pm$ 0.022 (1) & 0.850 $\pm$ 0.019 (2) \\
        ozone-level-8hr & \bftab 0.726 $\pm$ 0.020 (1) & 0.711 $\pm$ 0.020 (2) \\
        madelon & \bftab 0.803 $\pm$ 0.010 (1) & 0.606 $\pm$ 0.028 (2) \\
        Bioresponse & \bftab 0.794 $\pm$ 0.008 (1) & 0.784 $\pm$ 0.010 (2) \\
        wilt & \bftab 0.936 $\pm$ 0.015 (1) & 0.930 $\pm$ 0.019 (2) \\
        churn & \bftab 0.914 $\pm$ 0.017 (1) & 0.873 $\pm$ 0.018 (2) \\
        phoneme & \bftab 0.846 $\pm$ 0.008 (1) & 0.845 $\pm$ 0.011 (2) \\
        SpeedDating & 0.723 $\pm$ 0.013 (2) & \bftab 0.728 $\pm$ 0.009 (1) \\
        PhishingWebsites & \bftab 0.969 $\pm$ 0.006 (1) & 0.965 $\pm$ 0.006 (2) \\
        Amazon\_employee\_access & 0.665 $\pm$ 0.009 (2) & \bftab 0.675 $\pm$ 0.008 (1) \\
        nomao & \bftab 0.958 $\pm$ 0.002 (1) & 0.954 $\pm$ 0.002 (2) \\
        adult & \bftab 0.790 $\pm$ 0.006 (1) & 0.790 $\pm$ 0.005 (2) \\
        numerai28.6 & \bftab 0.519 $\pm$ 0.003 (1) & 0.519 $\pm$ 0.003 (2) \\
        \midrule
        Normalized Mean $\uparrow$ & \bftab 0.824 (1) & 0.177 (2) \\ 
        MRR $\uparrow$ & \bftab 0.921  (1) & 0.579  (2) \\        
\bottomrule
\end{tabular}

\vspace{0.5cm}

\label{tab:ablation_study_weighting}
\end{table}

\subsection{Confusion Matrix Case Study}

\begin{table}[H]
\small
\caption[Pairwise Confusion Matrix PhishingWebsites]{\textbf{Pairwise Confusion Matrix PhishingWebsites.} We compare the predictions of each approach with the predictions of a CART DT. It becomes evident, that a simple model is not sufficient to solve the task well, as CART makes more than twice as many mistakes as state-of-the-art models.}
\centering
\begin{tabular}{|L{3cm}|R{2cm}|R{2cm}|R{1.5cm}|}
\hline
 &  \multicolumn{1}{l|}{Correct DT} & \multicolumn{1}{l|}{Incorrect DT} & \multicolumn{1}{l|}{Total}  \\
\hline
Correct GRANDE & 2012 & 128 & \bftab 2140 \\
\hline
Incorrect GRANDE & 15 & 56 & \bftab 71 \\
\hline
Total & \bftab 2027 & \bftab 184 & \\ \hline
\end{tabular}

\vspace{0.1cm}

\begin{tabular}{|L{3cm}|R{2cm}|R{2cm}|R{1.5cm}|}
\hline
 &  \multicolumn{1}{l|}{Correct DT} & \multicolumn{1}{l|}{Incorrect DT} & \multicolumn{1}{l|}{Total}  \\
 \hline
Correct XGBoost & 2018 & 115 & \bftab 2133 \\
\hline
Incorrect XGBoost & 9 & 69 & \bftab 78 \\
\hline
Total & \bftab 2027 & \bftab 184 & \\ \hline
\end{tabular}

\vspace{0.1cm}

\begin{tabular}{|L{3cm}|R{2cm}|R{2cm}|R{1.5cm}|}
\hline
 &  \multicolumn{1}{l|}{Correct DT} & \multicolumn{1}{l|}{Incorrect DT} & \multicolumn{1}{l|}{Total}  \\
 \hline
Correct CatBoost & 2019 & 105 & \bftab 2124 \\
\hline
Incorrect CatBoost & 8 & 79 & \bftab 87 \\
\hline
Total & \bftab 2027 & \bftab 184 & \\ \hline
\end{tabular}

\vspace{0.1cm}

\begin{tabular}{|L{3cm}|R{2cm}|R{2cm}|R{1.5cm}|}
\hline
 &  \multicolumn{1}{l|}{Correct DT} & \multicolumn{1}{l|}{Incorrect DT} & \multicolumn{1}{l|}{Total}  \\

 \hline
Correct NODE & 2012 & 112 & \bftab 2124 \\
\hline
Incorrect NODE & 15 & 72 & \bftab 87 \\
\hline
Total & \bftab 2027 & \bftab 184 & \\ \hline
\end{tabular}

\label{tab:case_study_performance}
\end{table}

\newpage

\subsection{Additional Benchmarks}\label{A:benchmarks}

In addition to the representative evaluation in the main part, we provide an extended comparison with additional benchmarks, including tree ensemble methods Random Forest~\citep{breiman2001randomforest} and ExtraTrees~\citep{geurts2006extratree}, as well as SAINT~\cite{somepalli2021saint} as an additional DL benchmark, which is the superior gradient-based method according to \citet{borisov2022deep}. In the following, we present results using default hyperparameters for all methods (Table~\ref{tab:additional_benchmarks_default}) and optimized hyperparameters (Table~\ref{tab:additional_benchmarks_HPO} and Table~\ref{tab:additional_benchmarks_HPO_DL}). The results for HPO are presented twofold. In Table~\ref{tab:additional_benchmarks_HPO}, we report the results of an extensive search comprising 250 trials. Due to the very long runtime of SAINT, we were not able to optimize 250 trials for SAINT. Therefore, we present results from an additional HPO consisting of 30 trials for each method (similar to \citet{mcelfresh2023neural_whitepaper}) including SAINT (Table~\ref{tab:additional_benchmarks_HPO}). Details on the HPO can be found in Appendix~\ref{A:hyperparameters}. The implementation for SAINT, similar to NODE, was adopted from \citet{borisov2022deep}.

In all experiments, GRANDE demonstrated superior results over existing methods, achieving the highest normalized accuracy, MRR and number of wins in almost all scenarios. In line with the experiments from the main part, CatBoost achieved the second-best and XGBoost the third-best results.  The only exception is using default parameters, where XGBoost achieved a slightly higher normalized mean, but GRANDE still an significant higher MRR. While SAINT provided competitive results for some datasets (1 wins with HPO), it also achieved a very performance in some cases (e.g. for the \emph{madelon} dataset, the prediction of SAINT was only predicting the majority class). Similarly, ExtraTree and RandomForest, especially with optimized hyperparameters achieved competitive results, but were almost consistently outperformed by gradient-boosting methods and GRANDE.

\vspace{0.5cm}

\begin{table}[H]
\centering
\caption[Extended Performance Comparison (HPO 250 trials)]{\textbf{Extended Performance Comparison (HPO 250 trials).} We report the test macro F1-score (mean for a 5-fold CV) with optimized parameters based on a HPO with 250 trials. The datasets are sorted based on the data size.}
\small
\resizebox{\columnwidth}{!}{
\begin{tabular}{lllllll}
\toprule
 & GRANDE & XGB & CatBoost & NODE & ExtraTree & RandomForest \\
\midrule
dresses-sales & \bftab 0.612 (1) & 0.581 (4) & 0.588 (3) & 0.564 (5) & 0.589 (2) & 0.559 (6) \\
climate-model-simulation & \bftab 0.853 (1) & 0.763 (4) & 0.778 (3) & 0.802 (2) & 0.754 (5) & 0.751 (6) \\
cylinder-bands & \bftab 0.819 (1) & 0.773 (4) & 0.801 (2) & 0.754 (6) & 0.762 (5) & 0.782 (3) \\
wdbc & \bftab 0.975 (1) & 0.953 (4) & 0.963 (3) & 0.966 (2) & 0.951 (5) & 0.949 (6) \\
ilpd & \bftab 0.657 (1) & 0.632 (5) & 0.643 (3) & 0.526 (6) & 0.645 (2) & 0.640 (4) \\
tokyo1 & 0.921 (3) & 0.915 (5) & \bftab 0.927 (1) & 0.921 (2) & 0.916 (4) & 0.909 (6) \\
qsar-biodeg & 0.854 (2) & 0.853 (3) & 0.844 (5) & 0.836 (6) & 0.851 (4) & \bftab 0.855 (1) \\
ozone-level-8hr & \bftab 0.726 (1) & 0.688 (6) & 0.721 (2) & 0.703 (3) & 0.695 (5) & 0.697 (4) \\
madelon & 0.803 (3) & 0.833 (2) &\bftab  0.861 (1) & 0.571 (6) & 0.766 (4) & 0.764 (5) \\
Bioresponse & 0.794 (3) & 0.799 (2) & \bftab 0.801 (1) & 0.780 (6) & 0.794 (5) & 0.794 (4) \\
wilt & 0.936 (2) & 0.911 (6) & 0.919 (3) & \bftab 0.937 (1) & 0.917 (5) & 0.918 (4) \\
churn & 0.914 (2) & 0.900 (3) & 0.869 (6) & \bftab 0.930 (1) & 0.899 (4) & 0.899 (5) \\
phoneme & 0.846 (6) & 0.872 (2) & \bftab 0.876 (1) & 0.862 (5) & 0.868 (4) & 0.868 (3) \\
SpeedDating & \bftab 0.723 (1) & 0.704 (6) & 0.718 (2) & 0.707 (5) & 0.708 (3) & 0.707 (4) \\
PhishingWebsites & \bftab 0.969 (1) & 0.968 (2) & 0.965 (4) & 0.968 (3) & 0.963 (5) & 0.963 (5) \\
Amazon\_employee\_access & 0.665 (2) & 0.621 (6) & \bftab 0.671 (1) & 0.649 (3) & 0.642 (5) & 0.648 (4) \\
nomao & 0.958 (3) & \bftab 0.965 (1) & 0.964 (2) & 0.956 (4) & 0.951 (6) & 0.955 (5) \\
adult & 0.790 (5) & \bftab 0.798 (1) & 0.796 (2) & 0.794 (3) & 0.785 (6) & 0.790 (4) \\
numerai28.6 & \bftab 0.519 (1) & 0.518 (5) & 0.519 (2) & 0.503 (6) & 0.519 (4) & 0.519 (3) \\ \midrule
Normalized Mean $\uparrow$ & \bftab 0.817 (1) & 0.518 (3) & 0.705 (2) & 0.382 (6) & 0.385 (5) & 0.404 (4) \\
MRR $\uparrow$ & \bftab 0.668 (1) & 0.365 (3) & 0.541 (2) & 0.352 (4) & 0.251 (6) & 0.275 (5) \\
\bottomrule
\end{tabular}
}
\label{tab:additional_benchmarks_HPO}
\end{table}

\begin{table}[H]
\centering
\caption[Extended Performance Comparison (HPO 30 trials)]{\textbf{Extended Performance Comparison (HPO 30 trials).} We report the test macro F1-score (mean for a 5-fold CV) with optimized parameters (30 trials). The datasets are sorted based on the data size.}
\resizebox{\columnwidth}{!}{
\begin{tabular}{llllllll}
\toprule
 & GRANDE & XGB & CatBoost & NODE & ExtraTree & RandomForest &  SAINT \\
\midrule
dresses-sales & \bftab 0.596 (1) & 0.567 (5) & 0.592 (2) & 0.564 (6) & 0.583 (3) & 0.570 (4) & 0.516 (7) \\
climate-model-simulation-crashes &  \bftab 0.841 (1) & 0.764 (6) & 0.782 (5) & 0.802 (3) & 0.802 (2) & 0.793 (4) & 0.744 (7) \\
cylinder-bands &  \bftab 0.799 (1) & 0.769 (4) & 0.787 (3) & 0.754 (6) & 0.766 (5) & 0.753 (7) & 0.791 (2) \\
wdbc & 0.966 (2) & 0.959 (4) & 0.964 (3) &  \bftab 0.966 (1) & 0.955 (5) & 0.951 (7) & 0.952 (6) \\
ilpd &  \bftab 0.656 (1) & 0.625 (5) & 0.650 (2) & 0.526 (6) & 0.650 (3) & 0.642 (4) & 0.425 (7) \\
tokyo1 &  \bftab 0.933 (1) & 0.919 (6) & 0.925 (3) & 0.921 (4) & 0.929 (2) & 0.920 (5) & 0.915 (7) \\
qsar-biodeg &  \bftab 0.857 (1) & 0.841 (6) & 0.850 (5) & 0.836 (7) & 0.850 (4) & 0.855 (2) & 0.850 (3) \\
ozone-level-8hr & 0.716 (2) & 0.676 (6) &  \bftab 0.736 (1) & 0.703 (3) & 0.695 (5) & 0.697 (4) & 0.659 (7) \\
madelon & 0.787 (3) & 0.833 (2) &  \bftab 0.861 (1) & 0.571 (6) & 0.656 (5) & 0.774 (4) & 0.333 (7) \\
Bioresponse & 0.794 (5) &  \bftab 0.799 (1) & 0.797 (2) & 0.780 (6) & 0.796 (3) & 0.795 (4) & 0.780 (6) \\
wilt & 0.934 (3) & 0.913 (6) & 0.919 (4) & 0.937 (2) &  \bftab 0.938 (1) & 0.916 (5) & 0.574 (7) \\
churn & 0.914 (2) & 0.899 (4) & 0.869 (7) &  \bftab 0.930 (1) & 0.903 (3) & 0.899 (5) & 0.873 (6) \\
phoneme & 0.855 (6) & 0.870 (2) &  \bftab 0.881 (1) & 0.862 (4) & 0.852 (7) & 0.868 (3) & 0.858 (5) \\
SpeedDating &  \bftab 0.728 (1) & 0.704 (6) & 0.715 (2) & 0.707 (5) & 0.709 (4) & 0.709 (3) & 0.678 (7) \\
PhishingWebsites & 0.966 (4) &  \bftab 0.968 (1) & 0.964 (5) & 0.968 (2) & 0.953 (7) & 0.962 (6) & 0.966 (3) \\
Amazon\_employee\_access & 0.622 (6) & 0.620 (7) & 0.671 (2) & 0.649 (3) & 0.633 (5) & 0.647 (4) &  \bftab 0.696 (1) \\
nomao & 0.955 (5) & 0.963 (2) &  \bftab 0.964 (1) & 0.956 (3) & 0.944 (7) & 0.954 (6) & 0.956 (4) \\
adult & 0.790 (5) &  \bftab 0.798 (1) & 0.796 (2) & 0.794 (3) & 0.784 (6) & 0.790 (4) & 0.774 (7) \\
numerai28.6 &  \bftab 0.521 (1) & 0.518 (5) & 0.519 (3) & 0.503 (7) & 0.518 (4) & 0.520 (2) & 0.509 (6) \\ \midrule
Normalized Mean $\uparrow$ &  \bftab 0.804 (1) & 0.609 (3) & 0.787 (2) & 0.510 (6) & 0.523 (5) & 0.585 (4) & 0.244 (7) \\
MRR $\uparrow$ &  \bftab 0.597 (1) & 0.368 (3) & 0.491 (2) & 0.341 (4) & 0.299 (5) & 0.257 (6) & 0.240 (7) \\
\bottomrule
\end{tabular}
}
\label{tab:additional_benchmarks_HPO_DL}
\end{table}

\vspace{0.5cm}

\begin{table}[H]
\centering
\caption[Extended Performance Comparison (Default Parameters)]{\textbf{Extended Performance Comparison (Default Parameters).} We report the test macro F1-score (mean for a 5-fold CV) with default parameters. The datasets are sorted based on the data size.}
\resizebox{\columnwidth}{!}{
\begin{tabular}{llllllll}
\toprule
 & GRANDE & XGB & CatBoost & NODE & ExtraTree & RandomForest &  SAINT \\
\midrule
dresses-sales &  \bftab 0.596 (1) & 0.570 (3) & 0.573 (2) & 0.559 (4) & 0.545 (6) & 0.554 (5) & 0.422 (7) \\
climate-model-simulation-crashes & 0.758 (5) & 0.781 (2) & 0.781 (3) & 0.766 (4) & 0.728 (6) & 0.714 (7) &  \bftab 0.850 (1) \\
cylinder-bands &  \bftab 0.813 (1) & 0.770 (5) & 0.795 (2) & 0.696 (7) & 0.780 (4) & 0.787 (3) & 0.733 (6) \\
wdbc & 0.962 (3) & 0.966 (1) & 0.955 (4) & 0.964 (2) & 0.940 (7) & 0.942 (6) & 0.945 (5) \\
ilpd &  \bftab 0.646 (1) & 0.629 (3) & 0.643 (2) & 0.501 (6) & 0.582 (5) & 0.592 (4) & 0.416 (7) \\
tokyo1 &  \bftab 0.922 (1) & 0.917 (6) & 0.917 (5) & 0.921 (2) & 0.917 (3) & 0.917 (4) & 0.907 (7) \\
qsar-biodeg &  \bftab 0.851 (1) & 0.844 (2) & 0.843 (3) & 0.838 (4) & 0.836 (7) & 0.838 (5) & 0.836 (6) \\
ozone-level-8hr &  \bftab 0.735 (1) & 0.686 (3) & 0.702 (2) & 0.662 (4) & 0.601 (7) & 0.605 (6) & 0.651 (5) \\
madelon & 0.768 (5) & 0.811 (2) &  \bftab 0.851 (1) & 0.650 (6) & 0.773 (4) & 0.773 (3) & 0.333 (7) \\
Bioresponse & 0.789 (4) & 0.789 (5) & 0.792 (3) & 0.786 (6) & 0.799 (2) &  \bftab 0.801 (1) & 0.786 (6) \\
wilt &  \bftab 0.933 (1) & 0.903 (5) & 0.898 (6) & 0.904 (4) & 0.916 (2) & 0.911 (3) & 0.486 (7) \\
churn & 0.896 (3) & 0.897 (2) & 0.862 (7) &  \bftab 0.925 (1) & 0.892 (5) & 0.893 (4) & 0.867 (6) \\
phoneme & 0.860 (6) & 0.864 (3) & 0.861 (5) & 0.842 (7) & 0.875 (2) &  \bftab 0.880 (1) & 0.862 (4) \\
SpeedDating &  \bftab 0.725 (1) & 0.686 (4) & 0.693 (3) & 0.703 (2) & 0.642 (6) & 0.640 (7) & 0.643 (5) \\
PhishingWebsites &  \bftab 0.969 (1) & 0.969 (2) & 0.963 (6) & 0.961 (7) & 0.968 (3) & 0.968 (3) & 0.964 (5) \\
Amazon\_employee\_access & 0.602 (7) & 0.608 (6) & 0.652 (4) & 0.621 (5) & 0.662 (3) &  \bftab 0.682 (1) & 0.682 (2) \\
nomao & 0.955 (6) &  \bftab 0.965 (1) & 0.962 (2) & 0.955 (7) & 0.956 (4) & 0.961 (3) & 0.956 (5) \\
adult & 0.785 (5) & 0.796 (2) & 0.796 (3) &  \bftab 0.799 (1) & 0.784 (6) & 0.793 (4) & 0.768 (7) \\
numerai28.6 & 0.503 (7) & 0.516 (2) &  \bftab 0.519 (1) & 0.506 (6) & 0.514 (3) & 0.512 (4) & 0.509 (5) \\ \midrule
Normalized Mean $\uparrow$ & 0.673 (2) &  \bftab 0.691 (1) & 0.666 (3) & 0.460 (6) & 0.518 (5) & 0.596 (4) & 0.228 (7) \\
MRR $\uparrow$ &  \bftab 0.586 (1) & 0.422 (2) & 0.397 (3) & 0.326 (5) & 0.267 (6) & 0.365 (4) & 0.235 (7) \\
\bottomrule
\end{tabular}
}
\label{tab:additional_benchmarks_default}
\end{table}

\newpage

\section{Hyperparameters}\label{A:hyperparameters}
We optimized the hyperparameters using Optuna~\citep{akiba2019optuna} with 250 trials and selected the search space and default parameters for related work in accordance with \citet{borisov2022deep}. The best parameters were selected based on a 5x2 cross-validation as suggested by \citet{raschka2018modelevaluation} where the test data of each fold was held out of the HPO to get unbiased results. To deal with class imbalance, we further included class weights. In line with \citet{borisov2022deep}, we did not tune the number of estimators for XGBoost and CatBoost, but used early stopping. To validate our approach of not tuning the estimators, we conducted an additional HPO with tuning the estimators for XGBoost and CatBoost (see Table~\ref{tab:tune_estimators_xgb} and Table~\ref{tab:tune_estimators_catboost}) and show that this results in similar (slightly worse) results.
While using 250 trials can be considered as a very exhaustive search, we include additional results for a HPO with only 30 trials, similar to ~\cite{mcelfresh2023neural_whitepaper}. Based on the results in Table~\ref{tab:hpo_30trials}, we can verify that even with a small number of trials, GRANDE achieves strong results. Yet, we can also observe that searching for more trials is especially beneficial for GRANDE.
Furthermore, GRANDE has a total of 11 hyperparameters that were optimized during the HPO. In contrast, for XGBoost and CatBoost we only optimized 4 and 3 hyperparameters, respectively. The choice to optimize only a small number of hyperparameters was in accordance with \citet{borisov2022deep} based on the assumption that an exhaustive search on the most relevant parameters is beneficial for the performance. To account for this, we performed an additional HPO for GRANDE, where we similarly only optimized 4 parameters (one overall learning\_rate, focal\_factor, cosine\_decay\_steps and dropout). The results are displayed in Table~\ref{tab:hpo_reduced_grid} and show that GRANDE still achieves SOTA results. However, it becomes evident, that GRANDE benefits from using a larger grid, especially since using different learning rates for the different components (split values, split indices, leaf probabilities and leaf weights) has a positive impact on the performance.
For GRANDE, we used a batch size of 64 and early stopping after 25 epochs.
Similar to NODE~\cite{popov2019neural}, GRANDE uses an Adam optimizer with stochastic weight averaging over 5 checkpoints~\citep{izmailov2018averaging} and a learning rate schedule that uses a cosine decay with optional warmup~\citep{loshchilov2016sgdr}. Furthermore, GRANDE allows using a focal factor~\citep{lin2017focal}, similar to GradTree.
In the supplementary material, we provide the notebook used for the optimization along with the search space for each approach.

\subsection{Estimator Tuning Comparison}

\begin{table}[H]
\centering
\small
\caption[Comparison XGB HPO (250 Trials)]{\textbf{Comparison XGB HPO (250 Trials).} We compare a HPO for XGBoost with and without tuning the number of estimators based on 250 trials. Tuning the estimators does not improve the performance, and using early stopping in addition to setting the number of estimators to a high number is sufficient.}
\resizebox{0.9\linewidth}{!}{
\begin{tabular}{lrr|r}
\toprule
 & \begin{tabular}[c]{@{}l@{}}Without\\tuning estimators\end{tabular}  & \begin{tabular}[c]{@{}l@{}}With\\tuning estimators\end{tabular}  & Difference \\
\midrule
dresses-sales & 0.5755 & \bftab 0.5813 & -0.0058 \\
climate-model-simulation-crashes & \bftab 0.7932 & 0.7626 & 0.0306 \\
cylinder-bands & 0.7670 & \bftab 0.7734 & -0.0064 \\
wdbc & \bftab 0.9568 & 0.9531 & 0.0037 \\
ilpd & 0.6316 & \bftab 0.6321 & -0.0005 \\
tokyo1 & \bftab 0.9147 & \bftab 0.9147 & 0.0000 \\
qsar-biodeg & 0.8486 & \bftab 0.8526 & -0.0039 \\
ozone-level-8hr & \bftab 0.7044 & 0.6885 & 0.0159 \\
madelon & 0.8226 & \bftab 0.8334 & -0.0108 \\
Bioresponse & 0.7924 & \bftab 0.7988 & -0.0064 \\
wilt & 0.9102 & \bftab 0.9114 & -0.0012 \\
churn & \bftab 0.9003 & 0.8998 & 0.0005 \\
phoneme & \bftab 0.8725 & 0.8718 & 0.0007 \\
SpeedDating & \bftab 0.7071 & 0.7043 & 0.0028 \\
PhishingWebsites & 0.9674 & \bftab 0.9683 & -0.0009 \\
Amazon\_employee\_access & \bftab 0.6225 & 0.6209 & 0.0016 \\
nomao & 0.9642 & \bftab 0.9650 & -0.0008 \\
adult & 0.7977 & \bftab 0.7980 & -0.0003 \\
numerai28.6 & \bftab 0.5187 & 0.5180 & 0.0007 \\ \midrule
Mean $\uparrow$ & \bftab 0.7930 & 0.7920 & 0.0010 \\
\bottomrule
\end{tabular}
}
\label{tab:tune_estimators_xgb}
\end{table}

\vspace{0.5cm}

\begin{table}[H]
\centering
\small
\caption[Comparison CatBoost HPO (250 Trials)]{\textbf{Comparison CatBoost HPO (250 Trials).} We compare a HPO for CatBoost with and without tuning the number of estimators based on 250 trials. Tuning the estimators does not improve the performance, and using early stopping in addition to setting the number of estimators to a high number is sufficient.}
\resizebox{0.9\linewidth}{!}{
\begin{tabular}{lrr|r}
\toprule
 & \begin{tabular}[c]{@{}l@{}}Without\\tuning estimators\end{tabular}  & \begin{tabular}[c]{@{}l@{}}With\\tuning estimators\end{tabular}  & Difference \\ \midrule
dresses-sales & 0.5611 & \bftab 0.5880 & -0.0270 \\
climate-model-simulation-crashes & \bftab 0.7902 & 0.7775 & 0.0127 \\
cylinder-bands & \bftab 0.8080 & 0.8014 & 0.0066 \\
wdbc & 0.9551 & \bftab 0.9625 & -0.0074 \\
ilpd & 0.6379 & \bftab 0.6428 & -0.0049 \\
tokyo1 & 0.9249 & \bftab 0.9274 & -0.0025 \\
qsar-biodeg & \bftab 0.8467 & 0.8444 & 0.0023 \\
ozone-level-8hr & \bftab 0.7330 & 0.7206 & 0.0124 \\
madelon & \bftab 0.8650 & 0.8607 & 0.0042 \\
Bioresponse & 0.7992 & \bftab 0.8013 & -0.0021 \\
wilt & \bftab 0.9227 & 0.9187 & 0.0040 \\
churn & \bftab 0.8734 & 0.8693 & 0.0042 \\
phoneme & \bftab 0.8828 & 0.8763 & 0.0065 \\
SpeedDating & 0.7182 & \bftab 0.7184 & -0.0001 \\
PhishingWebsites & \bftab 0.9657 & 0.9650 & 0.0007 \\
Amazon\_employee\_access & 0.6704 & \bftab 0.6709 & -0.0004 \\
nomao & 0.9633 & \bftab 0.9644 & -0.0011 \\
adult & 0.7957 & \bftab 0.7960 & -0.0003 \\
numerai28.6 & 0.5172 & \bftab 0.5193 & -0.0021 \\ \midrule
Mean $\uparrow$ & \bftab 0.8016 & 0.8013 & 0.0003 \\
\bottomrule
\end{tabular}
}
\label{tab:tune_estimators_catboost}
\end{table}

\vspace{0.5cm}

\subsection{Minimal HPO Comparison}

\begin{table}[H]
\centering
\small
\caption[HPO 30 Trials Performance Comparison.]{\textbf{HPO 30 Trials Performance Comparison.} We report the test macro F1-score (mean $\pm$ stdev over 10 trials) based on a reduced HPO with only 30 trials. The datasets are sorted based on the data size.}
\resizebox{\columnwidth}{!}{
\begin{tabular}{lcccc}
\toprule
{} &                        \multicolumn{1}{l}{GRANDE} &                           \multicolumn{1}{l}{XGB} &                      \multicolumn{1}{l}{CatBoost} &                          \multicolumn{1}{l}{NODE} \\
\midrule
dresses-sales & \bftab 0.596 $\pm$ 0.047 (1) & 0.567 $\pm$ 0.055 (3) & 0.592 $\pm$ 0.046 (2) & 0.564 $\pm$ 0.051 (4) \\
climate-model-simulation & \bftab 0.841 $\pm$ 0.058 (1) & 0.764 $\pm$ 0.073 (4) & 0.782 $\pm$ 0.057 (3) & 0.802 $\pm$ 0.035 (2) \\
cylinder-bands & \bftab 0.799 $\pm$ 0.037 (1) & 0.769 $\pm$ 0.040 (3) & 0.787 $\pm$ 0.044 (2) & 0.754 $\pm$ 0.040 (4) \\
wdbc & 0.966 $\pm$ 0.019 (2) & 0.959 $\pm$ 0.026 (4) & 0.964 $\pm$ 0.022 (3) & \bftab 0.966 $\pm$ 0.016 (1) \\
ilpd & \bftab 0.656 $\pm$ 0.044 (1) & 0.625 $\pm$ 0.052 (3) & 0.650 $\pm$ 0.059 (2) & 0.526 $\pm$ 0.069 (4) \\
tokyo1 & \bftab 0.933 $\pm$ 0.011 (1) & 0.919 $\pm$ 0.012 (4) & 0.925 $\pm$ 0.011 (2) & 0.921 $\pm$ 0.010 (3) \\
qsar-biodeg & \bftab 0.857 $\pm$ 0.025 (1) & 0.841 $\pm$ 0.015 (3) & 0.850 $\pm$ 0.030 (2) & 0.836 $\pm$ 0.028 (4) \\
ozone-level-8hr & 0.716 $\pm$ 0.007 (2) & 0.676 $\pm$ 0.028 (4) & \bftab 0.736 $\pm$ 0.027 (1) & 0.703 $\pm$ 0.029 (3) \\
madelon & 0.787 $\pm$ 0.014 (3) & 0.833 $\pm$ 0.015 (2) & \bftab 0.861 $\pm$ 0.012 (1) & 0.571 $\pm$ 0.022 (4) \\
Bioresponse & 0.794 $\pm$ 0.010 (3) & \bftab 0.799 $\pm$ 0.011 (1) & 0.797 $\pm$ 0.008 (2) & 0.780 $\pm$ 0.011 (4) \\
wilt & 0.934 $\pm$ 0.01 (2) & 0.913 $\pm$ 0.011 (4) & 0.919 $\pm$ 0.009 (3) & \bftab 0.937 $\pm$ 0.017 (1) \\
churn & 0.914 $\pm$ 0.013 (2) & 0.899 $\pm$ 0.020 (3) & 0.869 $\pm$ 0.014 (4) & \bftab 0.930 $\pm$ 0.011 (1) \\
phoneme & 0.855 $\pm$ 0.008 (4) & 0.870 $\pm$ 0.008 (2) & \bftab 0.881 $\pm$ 0.007 (1) & 0.862 $\pm$ 0.013 (3) \\
SpeedDating & \bftab 0.728 $\pm$ 0.007 (1) & 0.704 $\pm$ 0.012 (4) & 0.715 $\pm$ 0.015 (2) & 0.707 $\pm$ 0.015 (3) \\
PhishingWebsites & 0.966 $\pm$ 0.004 (3) & \bftab 0.968 $\pm$ 0.008 (1) & 0.964 $\pm$ 0.004 (4) & 0.968 $\pm$ 0.006 (2) \\
Amazon\_employee\_access & 0.622 $\pm$ 0.028 (3) & 0.620 $\pm$ 0.008 (4) & \bftab 0.671 $\pm$ 0.013 (1) & 0.649 $\pm$ 0.009 (2) \\
nomao & 0.955 $\pm$ 0.001 (4) & 0.963 $\pm$ 0.004 (2) & \bftab 0.964 $\pm$ 0.004 (1) & 0.956 $\pm$ 0.001 (3) \\
adult & 0.790 $\pm$ 0.006 (4) &\bftab  0.798 $\pm$ 0.004 (1) & 0.796 $\pm$ 0.005 (2) & 0.794 $\pm$ 0.004 (3) \\
numerai28.6 & \bftab 0.521 $\pm$ 0.004 (1) & 0.518 $\pm$ 0.002 (3) & 0.519 $\pm$ 0.002 (2) & 0.503 $\pm$ 0.010 (4) \\ \midrule
Normalized Mean $\uparrow$ & \bftab 0.694 (1) & 0.430  (3) & 0.676 (2) & 0.336 (4) \\ 
MRR $\uparrow$ & \bftab 0.636  (1) & 0.434 (3) & 0.579 (2) & 0.434  (3) \\
\bottomrule
\end{tabular}
}
\label{tab:hpo_30trials}
\end{table}

\vspace{0.5cm}

\subsection{Reduced Grid HPO Comparison}

\begin{table}[H]
\centering
\small
\caption[HPO Reduced Grid Performance Comparison (250 Trials)]{\textbf{HPO Reduced Grid Performance Comparison (250 Trials).} We report the test macro F1-score (mean $\pm$ stdev over 10 trials) based on a HPO with a reduced parameter grid for GRANDE (only one overall learning\_rate, focal\_factor, cosine\_decay\_steps and dropout) and 250 trials for all methods. The datasets are sorted based on the data size.}
\resizebox{\columnwidth}{!}{
\begin{tabular}{lcccc}
\toprule
{} &                        \multicolumn{1}{l}{GRANDE} &                           \multicolumn{1}{l}{XGB} &                      \multicolumn{1}{l}{CatBoost} &                          \multicolumn{1}{l}{NODE} \\
\midrule
dresses-sales & \bftab 0.590 $\pm$ 0.042 (1) & 0.581 $\pm$ 0.059 (3) & 0.588 $\pm$ 0.036 (2) & 0.564 $\pm$ 0.051 (4) \\
climate-model-simulation & 0.800 $\pm$ 0.055 (2) & 0.763 $\pm$ 0.064 (4) & 0.778 $\pm$ 0.050 (3) & \bftab 0.802 $\pm$ 0.035 (1) \\
cylinder-bands & 0.783 $\pm$ 0.040 (2) & 0.773 $\pm$ 0.042 (3) & \bftab 0.801 $\pm$ 0.043 (1) & 0.754 $\pm$ 0.040 (4) \\
wdbc & 0.964 $\pm$ 0.007 (2) & 0.953 $\pm$ 0.030 (4) & 0.963 $\pm$ 0.023 (3) & \bftab 0.966 $\pm$ 0.016 (1) \\
ilpd & \bftab 0.645 $\pm$ 0.030 (1) & 0.632 $\pm$ 0.043 (3) & 0.643 $\pm$ 0.053 (2) & 0.526 $\pm$ 0.069 (4) \\
tokyo1 & \bftab 0.928 $\pm$ 0.017 (1) & 0.915 $\pm$ 0.011 (4) & 0.927 $\pm$ 0.013 (2) & 0.921 $\pm$ 0.010 (3) \\
qsar-biodeg & \bftab 0.859 $\pm$ 0.024 (1) & 0.853 $\pm$ 0.020 (2) & 0.844 $\pm$ 0.023 (3) & 0.836 $\pm$ 0.028 (4) \\
ozone-level-8hr & \bftab 0.733 $\pm$ 0.016 (1) & 0.688 $\pm$ 0.021 (4) & 0.721 $\pm$ 0.027 (2) & 0.703 $\pm$ 0.029 (3) \\
madelon & 0.809 $\pm$ 0.010 (3) & 0.833 $\pm$ 0.018 (2) & \bftab 0.861 $\pm$ 0.012 (1) & 0.571 $\pm$ 0.022 (4) \\
Bioresponse & 0.783 $\pm$ 0.009 (3) & 0.799 $\pm$ 0.011 (2) & \bftab 0.801 $\pm$ 0.014 (1) & 0.780 $\pm$ 0.011 (4) \\
wilt & 0.921 $\pm$ 0.014 (2) & 0.911 $\pm$ 0.010 (4) & 0.919 $\pm$ 0.007 (3) & \bftab 0.937 $\pm$ 0.017 (1) \\
churn & 0.888 $\pm$ 0.017 (3) & 0.900 $\pm$ 0.017 (2) & 0.869 $\pm$ 0.021 (4) & \bftab 0.930 $\pm$ 0.011 (1) \\
phoneme & 0.859 $\pm$ 0.010 (4) & 0.872 $\pm$ 0.007 (2) & \bftab 0.876 $\pm$ 0.005 (1) & 0.862 $\pm$ 0.013 (3) \\
SpeedDating & \bftab 0.725 $\pm$ 0.016 (1) & 0.704 $\pm$ 0.015 (4) & 0.718 $\pm$ 0.014 (2) & 0.707 $\pm$ 0.015 (3) \\
PhishingWebsites & \bftab 0.970 $\pm$ 0.006 (1) & 0.968 $\pm$ 0.006 (2) & 0.965 $\pm$ 0.003 (4) & 0.968 $\pm$ 0.006 (3) \\
Amazon\_employee\_access & 0.643 $\pm$ 0.024 (3) & 0.621 $\pm$ 0.008 (4) & \bftab 0.671 $\pm$ 0.011 (1) & 0.649 $\pm$ 0.009 (2) \\
nomao & 0.956 $\pm$ 0.004 (4) &  \bftab 0.965 $\pm$ 0.003 (1) &         0.964 $\pm$ 0.002 (2) &         0.956 $\pm$ 0.001 (3) \\
adult & 0.791 $\pm$ 0.005 (4) &  \bftab 0.798 $\pm$ 0.004 (1) &         0.796 $\pm$ 0.004 (2) &         0.794 $\pm$ 0.004 (3) \\
numerai28.6 & 0.519 $\pm$ 0.005 (2) & 0.518 $\pm$ 0.001 (3) & \bftab 0.519 $\pm$ 0.002 (1) & 0.503 $\pm$ 0.010 (4) \\ \midrule

Normalized Mean $\uparrow$ & 0.659 $\pm$ 0.509 (2) & 0.488 $\pm$ 0.497 (3) & \bftab 0.721 $\pm$ 0.373 (1) & 0.346 $\pm$ 0.561 (4) \\
MRR $\uparrow$ & \bftab 0.610 $\pm$ 0.531 (1) & 0.425 $\pm$ 0.469 (4) & 0.596 $\pm$ 0.456 (2) & 0.452 $\pm$ 0.627 (3) \\
\bottomrule
\end{tabular}
}
\label{tab:hpo_reduced_grid}
\end{table}

\vspace{0.5cm}

\subsection{Best Hyperparameters}

\begin{table}[H]
\centering
\small
\caption[Hyperparameters GRANDE (Part 1)]{\textbf{Hyperparameters GRANDE (Part 1).} We report the hyperparameters for GRANDE based on the extensive HPO with 250 trials.}
\resizebox{1.0\columnwidth}{!}{
\begin{tabular}{lrrrrrrrr}
\toprule
 & depth & n\_estimators & lr\_weights & lr\_index & lr\_values & lr\_leaf  \\
\midrule
dresses-sales & 4 & 512 & 0.0015 & 0.0278 & 0.1966 & 0.0111 \\
climate-simulation-crashes & 4 & 2048 & 0.0007 & 0.0243 & 0.0156 & 0.0134 \\
cylinder-bands & 6 & 2048 & 0.0009 & 0.0084 & 0.0086 & 0.0474 \\
wdbc & 4 & 1024 & 0.0151 & 0.0140 & 0.1127 & 0.1758 \\
ilpd & 4 & 512 & 0.0007 & 0.0059 & 0.0532 & 0.0094 \\
tokyo1 & 6 & 1024 & 0.0029 & 0.1254 & 0.0056 & 0.0734 \\
qsar-biodeg & 6 & 2048 & 0.0595 & 0.0074 & 0.0263 & 0.0414 \\
ozone-level-8hr & 4 & 2048 & 0.0022 & 0.0465 & 0.0342 & 0.0503 \\
madelon & 4 & 2048 & 0.0003 & 0.0575 & 0.0177 & 0.0065 \\
Bioresponse & 6 & 2048 & 0.0304 & 0.0253 & 0.0073 & 0.0784 \\
wilt & 6 & 2048 & 0.0377 & 0.1471 & 0.0396 & 0.1718 \\
churn & 6 & 2048 & 0.0293 & 0.0716 & 0.0179 & 0.0225 \\
phoneme & 6 & 2048 & 0.0472 & 0.0166 & 0.0445 & 0.1107 \\
SpeedDating & 6 & 2048 & 0.0148 & 0.0130 & 0.0095 & 0.0647 \\
PhishingWebsites & 6 & 2048 & 0.0040 & 0.0118 & 0.0104 & 0.1850 \\
Amazon\_employee\_access & 6 & 2048 & 0.0036 & 0.0056 & 0.1959 & 0.1992 \\
nomao & 6 & 2048 & 0.0059 & 0.0224 & 0.0072 & 0.0402 \\
adult & 6 & 1024 & 0.0015 & 0.0087 & 0.0553 & 0.1482 \\
numerai28.6 & 4 & 512 & 0.0001 & 0.0737 & 0.0513 & 0.0371 \\
\bottomrule
\end{tabular}
}
\label{tab:hp_grand_1}
\end{table}

\vspace{0.5cm}

\begin{table}[H]
\centering
\small
\caption[Hyperparameters GRANDE (Part 2)]{\textbf{Hyperparameters GRANDE (Part 2).} We report the hyperparameters for GRANDE based on the extensive HPO with 250 trials.}
\resizebox{\columnwidth}{!}{
\begin{tabular}{lrrrrrlrrr}
\toprule
  & dropout & selected\_variables & data\_fraction & focal\_factor & cosine\_decay\_steps \\
\midrule
dresses-sales & 0.25 & 0.7996 & 0.9779 & 0 & 0.0 \\
climate-simulation-crashes & 0.00 & 0.6103 & 0.8956 & 0 & 1000.0 \\
cylinder-bands & 0.25 & 0.5309 & 0.8825 & 0 & 1000.0 \\
wdbc & 0.50 & 0.8941 & 0.8480 & 0 & 0.0 \\
ilpd & 0.50 & 0.6839 & 0.9315 & 3 & 1000.0 \\
tokyo1 & 0.50 & 0.5849 & 0.9009 & 0 & 1000.0 \\
qsar-biodeg & 0.00 & 0.5892 & 0.8098 & 0 & 0.0 \\
ozone-level-8hr & 0.25 & 0.7373 & 0.8531 & 0 & 1000.0 \\
madelon & 0.25 & 0.9865 & 0.9885 & 0 & 100.0 \\
Bioresponse & 0.50 & 0.5646 & 0.8398 & 0 & 0.0 \\
wilt & 0.25 & 0.9234 & 0.8299 & 0 & 0.0 \\
churn & 0.00 & 0.6920 & 0.8174 & 0 & 1000.0 \\
phoneme & 0.00 & 0.7665 & 0.8694 & 3 & 1000.0 \\
SpeedDating & 0.00 & 0.8746 & 0.8229 & 3 & 0.1 \\
PhishingWebsites & 0.00 & 0.9792 & 0.9588 & 0 & 0.1 \\
Amazon\_employee\_access & 0.50 & 0.9614 & 0.9196 & 3 & 0.0 \\
nomao & 0.00 & 0.8659 & 0.8136 & 0 & 100.0 \\
adult & 0.50 & 0.5149 & 0.8448 & 3 & 100.0 \\
numerai28.6 & 0.50 & 0.7355 & 0.8998 & 0 & 0.1 \\
\bottomrule
\end{tabular}
}
\label{tab:hp_grand_2}
\end{table}

\begin{table}[tb]
\centering
\small
\caption[Hyperparameters XGBoost]{\textbf{Hyperparameters XGBoost.} We report the hyperparameters for GRANDE based on the extensive HPO with 250 trials.}
\begin{tabular}{lrrrrrr}
\toprule
                                &  learning\_rate &  max\_depth &  reg\_alpha &  reg\_lambda  \\
\midrule
                   dresses-sales &                0.1032 &                 3 &            0.0000 &             0.0000 \\
climate-simulation-crashes &                0.0356 &                11 &            0.5605 &             0.0000 \\
                  cylinder-bands &                0.2172 &                11 &            0.0002 &             0.0057 \\
                            wdbc &                0.2640 &                 2 &            0.0007 &             0.0000 \\
                            ilpd &                0.0251 &                 4 &            0.3198 &             0.0000 \\
                          tokyo1 &                0.0293 &                 3 &            0.2910 &             0.3194 \\
                     qsar-biodeg &                0.0965 &                 5 &            0.0000 &             0.0000 \\
                 ozone-level-8hr &                0.0262 &                 9 &            0.0000 &             0.6151 \\
                         madelon &                0.0259 &                 6 &            0.0000 &             0.9635 \\
                     Bioresponse &                0.0468 &                 5 &            0.9185 &             0.0000 \\
                            wilt &                0.1305 &                 8 &            0.0000 &             0.0003 \\
                           churn &                0.0473 &                 6 &            0.0000 &             0.3132 \\
                         phoneme &                0.0737 &                11 &            0.9459 &             0.2236 \\
                     SpeedDating &                0.0277 &                 9 &            0.0000 &             0.9637 \\
                PhishingWebsites &                0.1243 &                11 &            0.0017 &             0.3710 \\
          Amazon\_employee\_access &                0.0758 &                11 &            0.9785 &             0.0042 \\
                           nomao &                0.1230 &                 5 &            0.0000 &             0.0008 \\
                           adult &                0.0502 &                11 &            0.0000 &             0.7464 \\
                     numerai28.6 &                0.1179 &                 2 &            0.0001 &             0.0262 \\

\bottomrule
\end{tabular}

\label{tab:hp_xgb}
\end{table}

\begin{table}[tb]
\centering
\small
\caption[Hyperparameters CatBoost]{\textbf{Hyperparameters CatBoost.} We report the hyperparameters for GRANDE based on the extensive HPO with 250 trials.}
\begin{tabular}{lrrrrr}
\toprule
                                &  learning\_rate &  max\_depth &  l2\_leaf\_reg  \\
\midrule
                   dresses-sales &                0.0675 &               3 &             19.8219 \\
climate-simulation-crashes &                0.0141 &               2 &             19.6955 \\
                  cylinder-bands &                0.0716 &              11 &             19.6932 \\
                            wdbc &                0.1339 &               3 &              0.7173 \\
                            ilpd &                0.0351 &               4 &              5.0922 \\
                          tokyo1 &                0.0228 &               5 &              0.5016 \\
                     qsar-biodeg &                0.0152 &              11 &              0.7771 \\
                 ozone-level-8hr &                0.0118 &              11 &              3.0447 \\
                         madelon &                0.0102 &              10 &              9.0338 \\
                     Bioresponse &                0.0195 &              11 &              8.1005 \\
                            wilt &                0.0192 &              11 &              1.1095 \\
                           churn &                0.0248 &               9 &              7.0362 \\
                         phoneme &                0.0564 &              11 &              0.6744 \\
                     SpeedDating &                0.0169 &              11 &              1.5494 \\
                PhishingWebsites &                0.0239 &               8 &              1.6860 \\
          Amazon\_employee\_access &                0.0123 &              11 &              1.6544 \\
                           nomao &                0.0392 &               8 &              2.6583 \\
                           adult &                0.1518 &              11 &             29.3098 \\
                     numerai28.6 &                0.0272 &               4 &             18.6675 \\

\bottomrule
\end{tabular}
\label{tab:hp_catboost}
\end{table}

\begin{table}[tb]
\centering
\small
\caption[Hyperparameters NODE]{\textbf{Hyperparameters NODE.} We report the hyperparameters for GRANDE based on the grid suggested by the authors~\cite{popov2019neural}.}
\begin{tabular}{lrrr}
\toprule
                                &  num\_layers &  total\_tree\_count &  tree\_depth \\
\midrule
                   dresses-sales &                2 &                   2048 &                6 \\
climate-simulation-crashes &                2 &                   1024 &                8 \\
                  cylinder-bands &                2 &                   1024 &                8 \\
                            wdbc &                2 &                   2048 &                6 \\
                            ilpd &                4 &                   2048 &                8 \\
                          tokyo1 &                4 &                   1024 &                8 \\
                     qsar-biodeg &                2 &                   2048 &                6 \\
                 ozone-level-8hr &                4 &                   1024 &                6 \\
                         madelon &                2 &                   2048 &                6 \\
                     Bioresponse &                2 &                   1024 &                8 \\
                            wilt &                4 &                   1024 &                6 \\
                           churn &                2 &                   1024 &                8 \\
                         phoneme &                4 &                   2048 &                8 \\
                     SpeedDating &                4 &                   1024 &                6 \\
                PhishingWebsites &                2 &                   1024 &                8 \\
          Amazon\_employee\_access &                2 &                   2048 &                6 \\
                           nomao &                2 &                   2048 &                6 \\
                           adult &                4 &                   1024 &                8 \\
                     numerai28.6 &                2 &                   1024 &                8 \\
\bottomrule
\end{tabular}
\label{tab:hp_node}
\end{table}

\chapter{Appendix SYMPOL} %
\label{A:SYMPOL} %

\section{Learning Decision Trees with Policy Gradients} \label{A:sympol_formalization}
In the following, we formalize the online training of hard, axis-aligned DTs\footnote{In the following, we will refer to hard, axis-aligned DTs simply as \emph{DTs}, and specifically note exceptions.} with the PPO objective. In contrast to existing work on RL with DTs, this allows an optimization of the DT on-policy without information loss. Our method can be combined with arbitrary gradient-based RL frameworks. 
We primarily focus on PPO as the, we believe, most prominent on-policy RL method. To support our claim of seamless integration, we provide additional results using Advantage Actor-Critic (A2C), as proposed by \citet{mnih2016asynchronous}, in Appendix~\ref{A:A2C}.
To efficiently learn DT policies with SYMPOL, we employed several crucial (see ablation study in Table~\ref{tab:ablation}) modifications, which we will elaborate below.

\subsection{Learning DTs with Policy Gradients} \label{sec:dt_policy_appendix}
SYMPOL utilizes GradTree (Section~\ref{sec:method_gradtree}) as a core component to learn a DT policy. In the following section, we incorporate GradTree into the PPO framework.

\paragraph{Arithmetic DT policy formulation}%
Traditionally, DTs involve nested concatenations of rules. In GradTree, DTs are formulated as arithmetic functions based on addition and multiplication to facilitate gradient-based learning. Therefore, our resulting DT policy is fully-grown (i.e., complete, full) and can be pruned post-hoc. Our basic pruning involves removing redundant paths, which significantly reduces the complexity. We define a path as redundant if the decision is already determined either by previous splits or based on the range of the selected feature. More details are given in Appendix~\ref{A:tree_size}.
Overall, we formulate a DT policy $\pi$ of depth \(d\) with respect to its parameters as:
\begin{equation}\label{eq:tree}
 \pi(\bm{s} |  \bm{a}, \bm{\tau}, \bm{\iota}) = \sum_{l=0}^{2^d-1} a_l \, \mathbb{L}(\bm{s} | l, \bm{\tau}, \bm{\iota}),
\end{equation}
where $\mathbb{L}$ is a function that indicates whether a state \(\bm{s} \in \mathbb{R}^{|\mathcal{S}|}\) belongs to a leaf $l$, \(\bm{a} \in \mathcal{A}^{2^d}\) denotes the selected action for each leaf node, \(\bm{\tau} \in \mathbb{R}^{2^d-1}\) represents split thresholds and \(\bm{\iota} \in \mathbb{N}^{2^d-1}\) the feature index for each internal node. %

\paragraph{Dense architecture}%
To support a gradient-based optimization and ensure an efficient computation via matrix operations, we make use of a dense DT representation. Traditionally, the feature index vector \(\bm{\iota}\) is one-dimensional. However, as in GradTree, we expand it into a matrix form. Specifically, this representation one-hot encodes the feature index, converting \(\bm{\iota} \in \mathbb{R}^{2^d-1}\) into a matrix \(\bm{I} \in \mathbb{R}^{(2^d-1) \times |\mathcal{S}|}\). Similarly, for split thresholds, instead of a single value for all features, individual values for each feature are stored, leading to \(\bm{T} \in \mathbb{R}^{(2^d-1) \times |\mathcal{S}|}\).
An example for the dense representation is visualized in Figure~\ref{fig:new_rep_dense}. Please note, that in contrast to SDTs, the dense representation of SYMPOL corresponds to an equivalent standard DT representation at each point in time, ensuring that the underlying model is a hard, axis-aligned DT.
By enumerating the internal nodes in breadth-first order, we can redefine the indicator function $\mathbb{L}$ for a leaf $l$, resulting in
\begin{equation}\label{eq:gradtree}
\pi(\bm{s} | \bm{a}, T, I) = \sum_{l=0}^{2^d-1} a_l \, \mathbb{L}(\bm{s} | l, \bm{T}, \bm{I}),
\end{equation}
\begin{equation}\label{eq:indicator_func_sympol}
\begin{split}
    \text{where} \quad
    \mathbb{L}(\boldsymbol{s} | l, \bm{T}, \bm{I}) &= \prod^{d-1}_{j=0} \left(1 - \mathfrak{p}(l,j) \right) \, \mathbb{S}(\boldsymbol{s} | \bm{I}_{\mathfrak{i}(l,j)},\bm{T}_{\mathfrak{i}(l,j)}) \\
    &\quad + \mathfrak{p}(l,j) \, \left( 1 - \mathbb{S}(\boldsymbol{s} | \bm{I}_{\mathfrak{i}(l,j)},\bm{T}_{\mathfrak{i}(l,j)}) \right).
\end{split} 
\end{equation}
Here, $\mathfrak{i}$ is the index of the internal node preceding a leaf node $l$ at a certain depth $j$ and  $\mathfrak{p}$ indicates whether the left ($\mathfrak{p} = 0$) or the right branch ($\mathfrak{p} = 1$) was taken. %

\paragraph{Axis-aligned splitting}%
Typically, DTs use the Heaviside function for splitting, which is non-differentiable. We use the split function introduced in GradTree to account for reasonable gradients:

\begin{equation}
     \mathbb{S}(\boldsymbol{s}| \boldsymbol{\iota}, \boldsymbol{\tau}) = \left\lfloor S \left( \boldsymbol{\iota} \cdot  \boldsymbol{s} - \boldsymbol{\iota} \cdot \boldsymbol{\tau} \right)  \right\rceil ,
     \label{eq:split_sigmoid_round_sympol}
\end{equation}

where \(S(z) = \frac{1}{1 + e^{-z}}\) represents the logistic function, \(\left\lfloor z \right\rceil\) stands for rounding a real number $z$ to the nearest integer and $\bm{a} \cdot \bm{b}$ denotes the dot product. We further need to ensure that $\boldsymbol{\iota}$ is a one-hot encoded vector to account for axis-aligned splits. This is achieved by applying a hardmax transformation before calculating $\mathbb{S}$. 
Both rounding and hardmax operations are non-differentiable and therefore, SYMPOL is \emph{not} considered as a soft or differentiable DT method. To overcome this, we employ a straight-through operator \citep{bengio2013propagating} during backpropagation. This allows the model to use non-differentiable operations in the forward pass while ensuring gradient propagation in the backward pass. As a result, we can directly learn an interpretable DT from policy gradient. This makes SYMPOL framework-agnostic and facilitates a seamless integration into existing RL frameworks.

\section{Additional Results}\label{A:results}
In this section, we present additional results to support the claims made in the main part, along with extended results for the summarizing tables. We focus on the control environments because they offer a diverse suite of benchmarks that cover different tasks and include both continuous and discrete action spaces. We chose not to include the MiniGrid environments here because their inclusion could distort the results, particularly the averages calculated in the main part, as all MiniGrid environments involve similar tasks and feature a discrete action and observation space. The primary reason for including MiniGrid in the main part is to provide additional experimental results that confirm the robustness and applicability of our method across different domains, as well as to highlight that tree-based methods offer a beneficial inductive bias for these categorical environments.

\begin{table}[tb]
    \centering
    \small
    \caption[A2C Control Performance.]{\textbf{A2C Control Performance.} We report the average undiscounted cumulative test reward over 25 random trials. The best interpretable method, and methods not statistically different, are marked bold.}
    
    \begin{tabular}{lrrrrr}
    \toprule
             & \multicolumn{1}{L{0.5cm}}{CP}  & \multicolumn{1}{L{0.6cm}}{AB}  & \multicolumn{1}{L{0.6cm}}{LL}  & \multicolumn{1}{L{0.85cm}}{MC-C}    & \multicolumn{1}{L{0.75cm}}{PD-C}\\ \midrule   
         SYMPOL (ours)     & \bftab 500        &  \bftab -\phantom{0}84   &  \bftab -\phantom{0}85    &  \bftab \phantom{-}58       &  \bftab -\phantom{0}502     \\
         D-SDT             &  11    &    -427        &  -396                     &  -0                        &  -1137                      \\
         SA-DT (d=5)       &  295    &    -102        &  -348                     &   \phantom{-0}0       &  -1467                      \\
         SA-DT (d=8)       &  223    &    -\phantom{0}99  &  -367                     &   \phantom{-0}2       &  -1526            \\\midrule 
         
         MLP               & 500               &    -\phantom{0}78        &  \phantom{-}208                     &  \phantom{-0}0            &  -\phantom{0}202            \\
         SDT               & 500               &     -\phantom{0}85                   & -159            &  \phantom{-0}0            &  -\phantom{0}201            \\             
         \bottomrule
    \end{tabular}
    \label{tab:results_control_a2c}
\end{table}

\begin{table}[tb]
    \centering
    \small
    \caption[A2C Control Performance Comparison.]{\textbf{A2C Control Performance Comparison.} We report the average undiscounted cumulative test reward over 25 random trials, comparing A2C with PPO using optimized hyperparameters. A number is marked bold if the performance achieved with the underlying RL algorithm (PPO or A2C) is significantly better or not statistically different from the best result.}

    \resizebox{1.0\linewidth}{!}{
    \begin{tabular}{lrr|rr||rr|rr|rr|rr}
    \toprule
                                    & \multicolumn{2}{c}{MLP}       & \multicolumn{2}{c}{SDT}       & \multicolumn{2}{c}{SYMPOL (ours)}     & \multicolumn{2}{c}{SA-DT (d=5)}     & \multicolumn{2}{c}{SA-DT (d=8)}           & \multicolumn{2}{c}{D-SDT}\\ 
                                    &    A2C      &    PPO       &    A2C      &    PPO       &    A2C      &    PPO       &    A2C      &    PPO       &    A2C      &    PPO       &    A2C      &    PPO       \\ \midrule
                                    
         CP                   &   \bftab 500 & \bftab 500 & \bftab 500 & \bftab 500 & \bftab 500 & \bftab 500 & 295 & \bftab 446 & 223 & \bftab 476 & 11 & \bftab 128 \\
         AB                   &   -78 & \bftab -72 & -85 & \bftab -77 & -84 & \bftab -80 & -102 & \bftab -97 & -99 & \bftab -75 & -427 & \bftab -205 \\
         LL                   &   208 & \bftab 241 & -159 & \bftab -124 & -85 & \bftab -57 & -348 & \bftab -197 & -367 & \bftab -150 & -396 & \bftab -221 \\
         MC-C                 &   0 & \bftab 95  & \bftab 0 & -4 & 58 & \bftab 94 & 0 & \bftab 97 & -2 & \bftab 96 & \bftab 0 & -10 \\
         PD-C                 &   -202 & \bftab -191 & \bftab -201 &  -310 & -502 & \bftab -323 & -1467 & \bftab -1251 & -1526 & \bftab -854 & \bftab -1137 & -1343 \\

         \bottomrule
    \end{tabular}
     }
    
    \label{tab:results_control_comparison_a2c}
\end{table}

\subsection{Evaluation with Alternative RL Algorithms: Advantage Actor Critic (A2C)} \label{A:A2C}
To support our claim that SYMPOL can be integrated into arbitrary on-policy frameworks, we provide results on A2C in the following. The reported results use optimized hyperparameters for each method. In general, A2C is considered as less stable compared to PPO, which is an additional challenge for SYMPOL, as training stability is especially crucial for DT policies. As A2C does not update the policy in minibatches over multiple epochs, we did not include a dynamic batch size here, but update SYMPOL with a single update over the rollout to stay consistent with the A2C algorithm.
We compared the performance of all methods using A2C on control environments in Table~\ref{tab:results_control_a2c}, and additionally provided a direct comparison between PPO and A2C in Table~\ref{tab:results_control_comparison_a2c}.
When using A2C, SYMPOL consistently outperforms other interpretable models. The performance gap becomes even more pronounced when using A2C instead of PPO, as SYMPOL achieves substantially higher performance than all other interpretable models in each environment. On MC-C, SYMPOL is the only method that achieves a positive reward, whereas even full-complexity models were unable to solve the task. This can be attributed to the lower training stability of A2C compared to PPO. 
This could also explain the poor results of distillation methods, as the policy learned by full-complexity models, even when achieving a high test reward, is potentially less consolidated, making it harder to distill.
However, to confirm this assumption, further experiments would be required.
Based on these results, we can confirm that SYMPOL can seamless be integrated into other RL algorithms, demonstrating the high flexibility of our proposed method.
Additionally, our method can benefit from advances in RL, as it can be seamlessly integrated into novel frameworks.

\subsection{Information Loss}\label{A:information_loss}
We provide detailed results on the information loss which can result as a consequence of discretization (for D-SDT) or distillation (for SA-DT). In Table~\ref{tab:train_test_complete}, we report the validation reward of the trained model along with the test reward of the discretized model. We can clearly observe that there are major differences for SA-DT and D-SDT on several datasets, indicating information loss.
In Table~\ref{tab:cohensD}, we report Cohen's D to measure the effect size comparing the validation reward of the trained model with the test reward of the applied, optionally post-processed model. Again, we can clearly see large effects for SA-DT and D-SDT on several datasets, especially for PD-C and LL, but also CP. Furthermore, the training curves in Figure~\ref{fig:training_curves_pairwise} visually show the information loss during the training.

\begin{table}[tb]
    \centering
    \small
    \caption[Information Loss (Comparison)]{\textbf{Information Loss (Comparison)}. We report the validation reward of the trained model and the test reward of the applied model.}
    
    \resizebox{1.0\linewidth}{!}{
    \begin{tabular}{lrr|rr||rr|rr|rr|rr}
    \toprule
                                    & \multicolumn{2}{c}{MLP}       & \multicolumn{2}{c}{SDT}       & \multicolumn{2}{c}{SYMPOL (ours)}     & \multicolumn{2}{c}{SA-DT (d=5)}     & \multicolumn{2}{c}{SA-DT (d=8)}           & \multicolumn{2}{c}{D-SDT}\\ 
                                    &    valid      &    test       &    valid      &    test       &    valid      &    test       &    valid      &    test       &    valid      &    test       &    valid      &    test       \\ \midrule
                                    
         CP        &   500  & 500  & 500  & 500  & 500  & 500  & 500  & 446   & 500  & 476  & 500  & 128 \\
         AB        &   -71  & -72  & -89  & -77  & -79  & -80  & -71  & -97   & -71  & -75  & -89  & -205 \\
         LL        &   256  & 241  & -91  & -124 & -9   & -57  & 256  & -197  & 256  & -150 & -91  & -221 \\
         MC-C      &   95   & 95   & -4   & -4   & 87   & 94   & 95   & 97    & 95   & 96   & -4   & -10 \\
         PD-C      &   -169 & -191 & -295 & -310 & -305 & -323 & -169 & -1251 & -169 & -854 & -295 & -1343 \\

         \bottomrule
    \end{tabular}
    }
    \label{tab:train_test_complete}
\end{table}

\begin{table}[tb]
    \centering
    \small
        \caption[Information Loss (Cohen's D)]{\textbf{Information Loss (Cohen's D)}. We calculated Cohen's D to measure effect size between the validation reward of the trained model and the test reward of the applied model. Typically, values $>0.5$ are considered a medium and values $>0.8$ a large effect. positive effects that are at least medium are marked as bold.}
    
    \begin{tabular}{lrrrrrr}
    \toprule
                                    &   MLP             &   SDT         & SYMPOL (ours)     & SA-DT (d=5)   & SA-DT (d=8)   & D-SDT         \\ \midrule
         CP                   &   0.000 & 0.000 & 0.000 & \bftab 0.632 & \bftab 1.214 & \bftab 4.075 \\
         AB                    &   0.035 & -0.630 & 0.104 & \bftab 0.728 & 0.338 & \bftab 0.982 \\
         LL                &   0.341 & \bftab 0.750 & 0.370 & \bftab 4.776 & \bftab 8.155 & \bftab 2.254 \\
         MC-C   &   -0.042 & -0.002 & -1.035 & -2.011 & -1.172 & \bftab 0.745 \\
         PD-C      &  \bftab 1.195 & 0.081 & 0.468 & \bftab 13.120 & \bftab 4.101 & \bftab 7.573 \\  \midrule  
         \bftab Mean $\downarrow$                      &   \bftab 0.306          &  \bftab 0.040      &  \bftab -0.019          &   \bftab 3.449      &   \bftab 2.527     &   \bftab 3.126      \\    
         \bottomrule
    \end{tabular}
    
    \label{tab:cohensD}
\end{table}

\subsection{Comparison of SYMPOL, SA-DT (DAGGER) and VIPER (Q-DAGGER)} \label{A:viper_comparison}

In this section, we provide a direct comparison of SYMPOL with SA-DT and VIPER for control environments (see Table~\ref{tab:viper_control}) and MiniGrid environments (see Table~\ref{tab:viper_minigrid}).
SA-DT~\citep{silva2020optimization} can be considered as a version of DAGGER~\citep{ross2011reduction} and is conceptually similar to VIPER~\citep{bastani2018viper}, also referred to as Q-DAGGER, which improves data collection including additional weighting. 
Our results remain consistent with our original claims, demonstrating that SYMPOL outperforms alternative approaches. This is also in-line with the results reported by \citet{kohler2024interpretable}, where the authors show, that the sampling in VIPER does not yield a better performance compared do DAGGER/SA-DT for interpretable DTs.
The results reported in the original VIPER paper~\citep{bastani2018viper} stating to achieve a perfect reward for CP are on a different version of the environment (CartPole-v0) with only 200 opposed to 500 time steps and less randomness (CartPole-v1), making the underlying task easier.
Also, we want to note, that the reported results are in-line with related work, reporting comparable or worse results than ours. For instance, \citet{vos2024optimizing} report a mean reward of only 367 for VIPER on CartPole-v1. Also, \citet{kenny2023interpretable} showcase a poor performance of VIPER in general and specifically for LL the performance is worse than what we reported.
Our findings align with these, suggesting that differences in performance may reflect randomness and missing generalizability in the evaluation.

\begin{table}
\small
    \centering
    \small
    \caption[Control Performance VIPER]{\textbf{Control Performance.} We report the average undiscounted cumulative test reward over 25 random trials. The best interpretable method, and methods not statistically different, are marked bold. Please note that VIPER cannot be applied to continuous environments.}
    
    \begin{tabular}{lrrrrr}
    \toprule
             & CP  & AB  & LL  & MC-C    & PD-C \\ \midrule   
         SYMPOL (ours)     & \bftab 500        &  \bftab -\phantom{0}80   &  \bftab -\phantom{0}57    &  \bftab \phantom{-}94       &  \bftab -\phantom{0}323     \\
         D-SDT             &  128    &    -205        &  -221                     &  -10                        &  -1343                      \\
         SA-DT (d=5)       &  446    &    -97        &  -197                     &  \bftab \phantom{-}97       &  -1251                      \\
         SA-DT (d=8)       &  476    &   \bftab -\phantom{0}75  &  -150                     &  \bftab \phantom{-}96       &  -\phantom{0}854            \\ 
          VIPER (d=5)       &  457    &    \bftab -\phantom{0}77        &  -200                     &  -       &  -                      \\
         VIPER (d=8)       &  480    &   \bftab -\phantom{0}75  &  -169                     &  -       &  -            \\\midrule 
         
         MLP               & 500               &    -\phantom{0}72        &  \phantom{-}241                     &  \phantom{-}95            &  -\phantom{0}191            \\
         SDT               & 500               &     -\phantom{0}77                   & -124            &  -\phantom{0}4            &  -\phantom{0}310            \\             
         \bottomrule
    \end{tabular}
    \label{tab:viper_control}
        
\end{table}

\begin{table}
\centering
\small
\caption[MiniGrid Performance VIPER]{\textbf{MiniGrid Performance.} We report the average undiscounted cumulative test reward over 25 random trials. The best interpretable method, and methods not statistically different, are marked bold.}
\begin{tabular}{lR{0.6cm}R{0.75cm}R{0.75cm}R{0.6cm}R{0.8cm}}
\toprule
 & \multicolumn{1}{L{0.55cm}}{E-R} & \multicolumn{1}{L{0.5cm}}{DK} & \multicolumn{1}{L{0.7cm}}{LG-5} & \multicolumn{1}{L{0.7cm}}{LG-7} & \multicolumn{1}{L{0.5cm}}{DS} \\ \midrule
SYMPOL (ours)   & \bftab 0.964  & \bftab 0.959  & \bftab 0.951  & \bftab 0.953  & 0.939 \\
D-SDT           & 0.662         & 0.654         & 0.262         & 0.381         & 0.932 \\
SA-DT (d=5)     & 0.583         & \bftab 0.958  & \bftab 0.951  & 0.458         & \bftab 0.952 \\
SA-DT (d=8)     & 0.845         & \bftab 0.961  & \bftab 0.951  & 0.799         & \bftab 0.954 \\
VIPER (d=5)     & 0.651         & \bftab 0.958  & \bftab 0.948  & 0.456         & \bftab 0.954 \\
VIPER (d=8)     & 0.845         & \bftab 0.963  & \bftab 0.948  & 0.801         & \bftab 0.954 \\
\midrule
MLP             & 0.963         & 0.963         & 0.951         & 0.760         & 0.951 \\ %
SDT             & 0.966         & 0.959         & 0.839         & 0.953         & 0.954 \\
\bottomrule
\end{tabular}
\label{tab:viper_minigrid}
\end{table}

\newpage

\subsection{Tree Size} \label{A:tree_size}
We report the average tree sizes over 25 trials for each environment. The DTs for SYMPOL and D-SDT are automatically pruned by removing redundant paths. There are mainly two identifiers, making a path redundant:
\begin{itemize}
    \item The split threshold of a split is outside the range specified by the environment. For instance, if $x_1 \in [0.0, 1.0]$ the decision $x_1 \le -0.1$ will always be false as $-0.1 \le 0.0$.
    \item A decision at a higher level of the tree already predefines the current decision. For instance, if the split at the root node is $x_1 \le 0.5$ and the subsequent node following the true path is $x_1 \le 0.6$ we know that this node will always be evaluated to true as $0.5 \le 0.6$.
\end{itemize}
We excluded the MiniGrid environments here, as they require a more sophisticated, automated pruning as there exist more requirements making a path redundant. For instance, if for the decision whether there is a wall in front of the agent is true, the decision for all other objects at the same position has to be always false.
\begin{table}[tb]
    \centering
    \small
        \caption[Tree Size]{\textbf{Tree Size}. We report the average size of the learned DT for each environment.}
    \begin{tabular}{lrrrr}
    \toprule
                                    & SYMPOL (ours)     & D-SDT             & SA-DT (d=5)   & SA-DT (d=8)   \\ \midrule
         CP                   &   39.4            & 14.2              &  61.8         &  315.0        \\     
         AB                    &   78.6            & 17.0              &  56.5         &  173.0        \\     
         LL                &   55.0            & 19.8              &  59.8         &  270.2        \\
         MC-C   &   23.4            & \phantom{0}3.0    &  61.0         &  311.8        \\              
         PD-C      &   56.2            & 28.6              &  62.2         &  388.2        \\ \midrule  
         \bftab Mean $\downarrow$                      &   \bftab 50.5     & \bftab 16.5       &  \bftab 60.3  &  \bftab 291.6 \\      
         \bottomrule
    \end{tabular}
    \label{tab:tree_size_sympol_repeated}
\end{table}

\subsection{Interpretability Comparison between Hard, Axis-Aligned and Soft DTs} \label{A:interpretability_comparison}
In the main part, we showed a visualization of a hard, axis-aligned DT learned by SYMPOL on the MC-C environment. While this was a comparatively small tree, we provide another example of a comparatively large tree with $59$ nodes in Figure~\ref{fig:cartpole_large}. While the tree is comparatively large, we can observe that the main logic is contained in the nodes at higher levels, focusing on the pole angle and the pole angular velocity. The less important features are in the lower levels where splits are often mage on the cart position, which is not required to solve the task perfectly. This also highlights the potential for advanced post-hoc pruning methods to increase interpretability and potentially even generalization.
When comparing the hard, axis-aligned DTs (Figure~\ref{fig:cartpole_large} and Figure~\ref{fig:sympol_policies}) learned by SYMPOL with corresponding SDTs learned by existing direct optimization methods (Figure~\ref{fig:cartpole_sdt} and Figure~\ref{fig:mountaincar_sdt}), the superiority of axis-aligned splits over oblique, multidimensional boundaries becomes evident.
The DTs learned by SYMPOL are substantially more interpretable, both at the level of individual splits and the overall tree structure. For example, when examining the root node of the CartPole task, the standard DT (Figure~\ref{fig:cartpole_large}) makes a straightforward comparison, "$s_3 \text{(Pole Velocity)} \leq -0.800$", whereas the corresponding SDT (Figure~\ref{fig:cartpole_sdt}) expresses the decision as "$\sigma(-1.14s_0 - 0.30s_1 + 0.94s_2 + 0.11s_3 + 0.99)$", which is significantly more complex and harder to interpret. This disparity becomes even more pronounced when considering complete paths or the entire tree. In SYMPOL's DTs, decisions at nodes are binary (yes/no), whereas SDTs employ probabilistic routing.
Probabilistic routing introduces two key disadvantages in interpretability compared to axis-aligned DTs: (1) The need to consider multiple paths simultaneously, and (2) the inability to directly interpret the leaf outputs, as they are weighted by path probabilities.
We also want to note that in order to allow a visualization of the SDT, we reduced the tree size to only 4, while the tree size used during the evaluation was 7.

\begin{figure}[H]
\centering
    \includegraphics[width=0.8\textheight, angle=90]{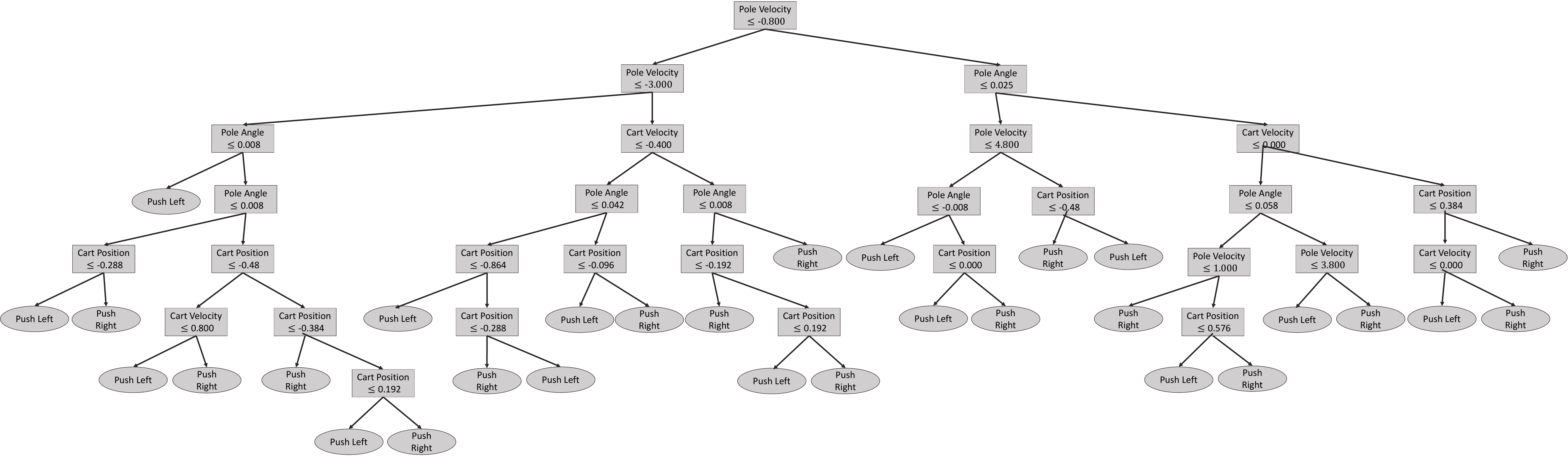} %
    \caption[Exemplary large CartPole DT]{\textbf{Exemplary large CartPole DT.} This figure visualizes a larger tree with $59$ nodes learned by SYMPOL on the CartPole environment.}
    \label{fig:cartpole_large} %
\end{figure}

\begin{figure}[H]
\centering
    \includegraphics[width=0.8\textheight, angle=90]{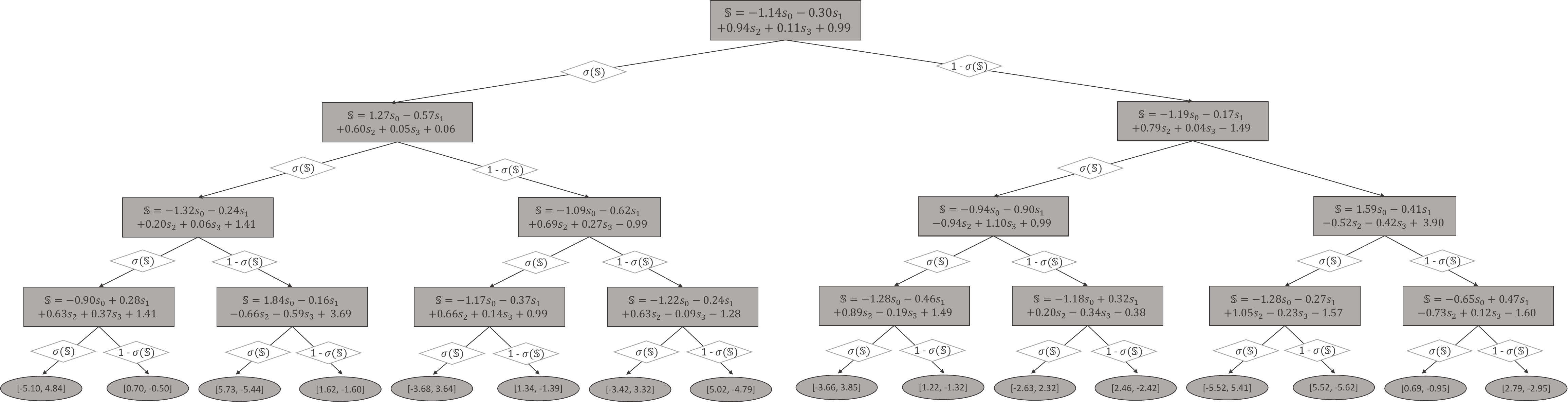} %
    \caption[Exemplary SDT for CartPole]{\textbf{Exemplary SDT for CartPole.} This figure visualizes a soft / differentiable DT learned on CartPole. The tree involves oblique decisions involving multiple features at each split. Additionally, there is no hard decision on which path is selected, but multiple paths are taken with an associated probability. The final prediction is obtained by weighting the leaf outputs with the corresponding path probability.}
    \label{fig:cartpole_sdt} %
\end{figure}

\begin{figure}[H]
\centering
    \includegraphics[width=0.8\textheight, angle=90]{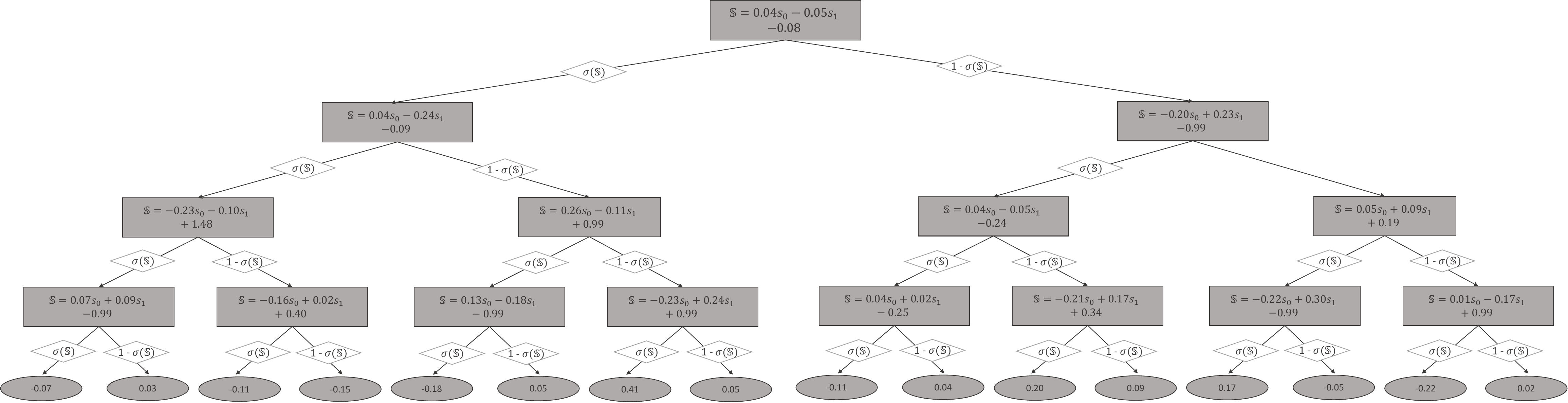} %
    \caption[Exemplary SDT for MountainCar]{\textbf{Exemplary SDT for MountainCar.} This figure visualizes a soft / differentiable DT learned on MountainCar. The tree involves oblique decisions involving multiple features at each split. Additionally, there is no hard decision on which path is selected, but multiple paths are taken with an associated probability. The final prediction is obtained by weighting the leaf outputs with the corresponding path probability.}
    \label{fig:mountaincar_sdt} %
\end{figure}

\subsection{Ablation Study}\label{A:ablation}
Our ablation study was designed to support our intuitive justifications for the modifications made to the RL framework and our method. Therefore, we disabled individual component of our method and evaluated the performance without the specific component. This includes the following modifications introduced in Section~\ref{sec:sympol}:
\begin{enumerate}
    \item \textbf{w/o separate architecture:} Instead of using separate architectures for actor and critic, we use the same architecture and hyperparameters for the actor and critic.
    \item \textbf{w/o dynamic rollout:} We proposed a dynamic rollout buffer that increases with a stepwise exponential rate during training to increase stability while maintaining exploration early on. Here we used a standard, static rollout buffer.
    \item \textbf{w/o batch size adjustment:} Similar to the dynamic rollout buffer, we proposed using a dynamic batch size to increase gradient stability in later stages of the training. Here, we used  standard, static batch size.
    \item \textbf{w/o adamW:} We introduced an Adam optimizer with weight decay to SYMPOL to support the adjustment of the features to split on and the class predicted. Here, we use a standard Adam optimizer without weight decay.
\end{enumerate}
Detailed results for each of the control datasets are reported in Table~\ref{tab:ablation}. The results clearly confirm our intuitive justifications, as each adjustment has a crucial impact on the performance of SYMPOL.

\vspace{0.5cm}

\begin{table}[h]
    \centering
    \small
        \caption[Ablation Study]{\textbf{Ablation Study}. We report the average test performance over a total of 25 random trials. This normalized performance consists in normalizing each reward between 0 and 1 via an affine renormalization between the top- and worse-performing models. Instead of the worse-performing model, we use the 20\% test reward quantile to account for outliers.}  
    
        \resizebox{1.0\linewidth}{!}{
        \begin{tabular}{lrrrrr|r}
        \toprule
             Agent Type    & \multicolumn{1}{l}{CP}  & \multicolumn{1}{l}{AB}   & \multicolumn{1}{l}{LL} & \multicolumn{1}{l}{MC-C}  & \multicolumn{1}{l|}{PD-C} & \multicolumn{1}{l}{Normalized Mean ($\uparrow$)} \\ \midrule             
            SYMPOL                                  & \bftab 500.0        &  \bftab -\phantom{0}79.9   &  \bftab -\phantom{0}57.4                     & \bftab 94.3      &  \bftab -\phantom{0}323.3             &  \bftab 0.988          \\
             w/o separate architecture              & 135.6\phantom{}     &  -196.4                    &  -276.8                                      &  -552.4          &   -1219.4             &  0.080          \\
             w/o dynamic rollout                    & 456.1\phantom{}     &  -\phantom{0}92.0          &  -178.2                                      &  -144.1          &   -\phantom{0}434.4   &  0.598          \\
             w/o batch size adjustment              & 498.8\phantom{}     &  -\phantom{0}81.7          &  -320.7                                      &  -1818.8         &   -\phantom{0}434.4   &  0.372          \\
             w/o adamW                              & 416.1\phantom{}     &  -\phantom{0}78.3          &  -\phantom{0}97.3                            &  0.0             &   -\phantom{0}393.5   &  0.865          \\ 
             
             \bottomrule
        \end{tabular}
        }
        \label{tab:ablation}
\end{table}

\newpage

\subsection{Runtime and Training Curves}\label{A:runtime_curves}
The experiments were conducted on a single NVIDIA RTX A6000. The environments for CP, AB, MC-C and PD-C were vectorized~\citep{gymnax2022github} and therefore the training is highly efficient, taking only 30 seconds for 1mio timesteps on average (excluding the sequential evaluation which cannot be vectorized). The remaining environments are not vectorized, and we used the standard Gymnasium~\citep{towers2024gymnasium} implementation. In Table~\ref{tab:runtime} can clearly see the impact of environment vectorization, as the runtime for LL, which is not vectorized, is more than 10 times higher with over 400 seconds.
In addition to the training times, we report detailed training curves for each method. Figure~\ref{fig:training_curves} compares the training reward and the test reward of SYMPOL with the full-complexity models MLP and SDT. SYMPOL shows a similar convergence compared to full-complexity models on most environments. For AB, SYMPOL converges even faster than an MLP which can be attributed to the dynamic rollout buffer and batch size. For MC-C we can see that the training of SYMPOL is very unstable at the beginning. We believe that this can be attributed to the sparse reward function of this certain environment and the fact that as a result, minor changes in the policy can result in a severe drop in the reward. Combined with the small rollout buffer and batch size early in the training of SYMPOL, this can result in an unstable training. However, we can see that the training stabilizes later on, which again confirms the design of our dynamic buffer size increasing over time.
Furthermore, we provide a pairwise comparison of SYMPOL with SA-DT and D-SDT in Figure~\ref{fig:training_curves_pairwise}. Here, we can again observe the severe information loss for D-SDT and SA-DT by comparing the training curve with the test reward.

\vspace{0.5cm}

\begin{table}[h]
    \centering
    \small
        \caption[Runtime.]{\textbf{Runtime.} We report the average runtime over 25 trials. One trial spans 1mio timesteps for each environment. We excluded LL from the mean runtime calculation, as this is the only non-vectorized environment. To provide a fair comparison of different methods, we aligned the step and batch size.}    
    \begin{tabular}{lrrr}
    \toprule
                                    & SYMPOL (ours)     & SDT               & MLP          \\ \midrule
         CP                   &   28.8            &  23.9             &  25.2         \\     
         AB                    &   35.5            &  37.7             &  33.8        \\     
         MC-C   &  23.4             &  19.4            &  18.4          \\     
         PD-C      &  28.7             &  28.2             &  18.5         \\     \midrule
         \bftab Mean $\downarrow$                      & \bftab 29.1             & \bftab 27.3             & \bftab 24.0         \\    \midrule 
         LL                &  402.3            &   394.0            &  405.6       \\  
         \bottomrule
    \end{tabular}
   
         \label{tab:runtime}
\end{table}

\begin{figure}[H]
    \centering
    \begin{subfigure}[tb]{0.32\textwidth}
        \centering
        \includegraphics[width=\linewidth]{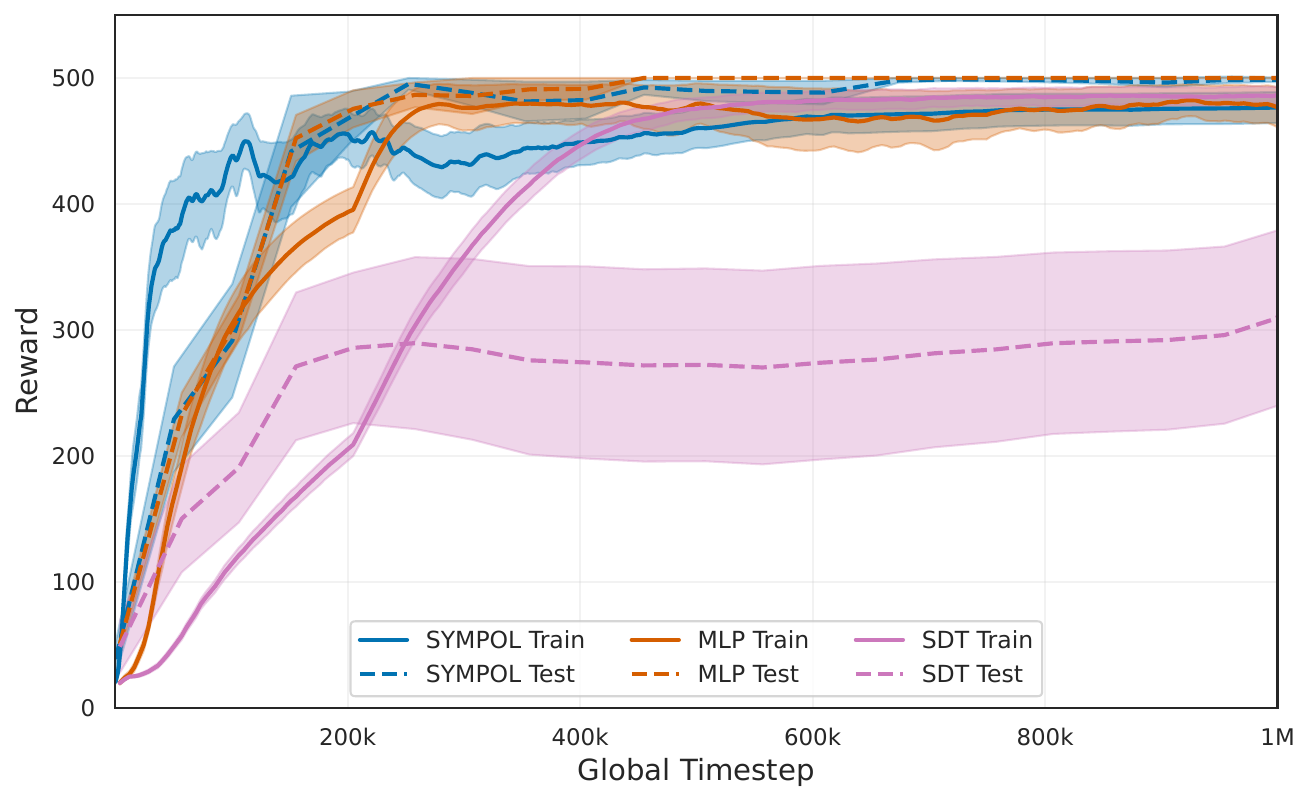}
        \caption{CP}
        \label{fig:learning_curves_CartPole}
    \end{subfigure}
    \:
    \begin{subfigure}[tb]{0.32\textwidth}
        \centering
        \includegraphics[width=\linewidth]{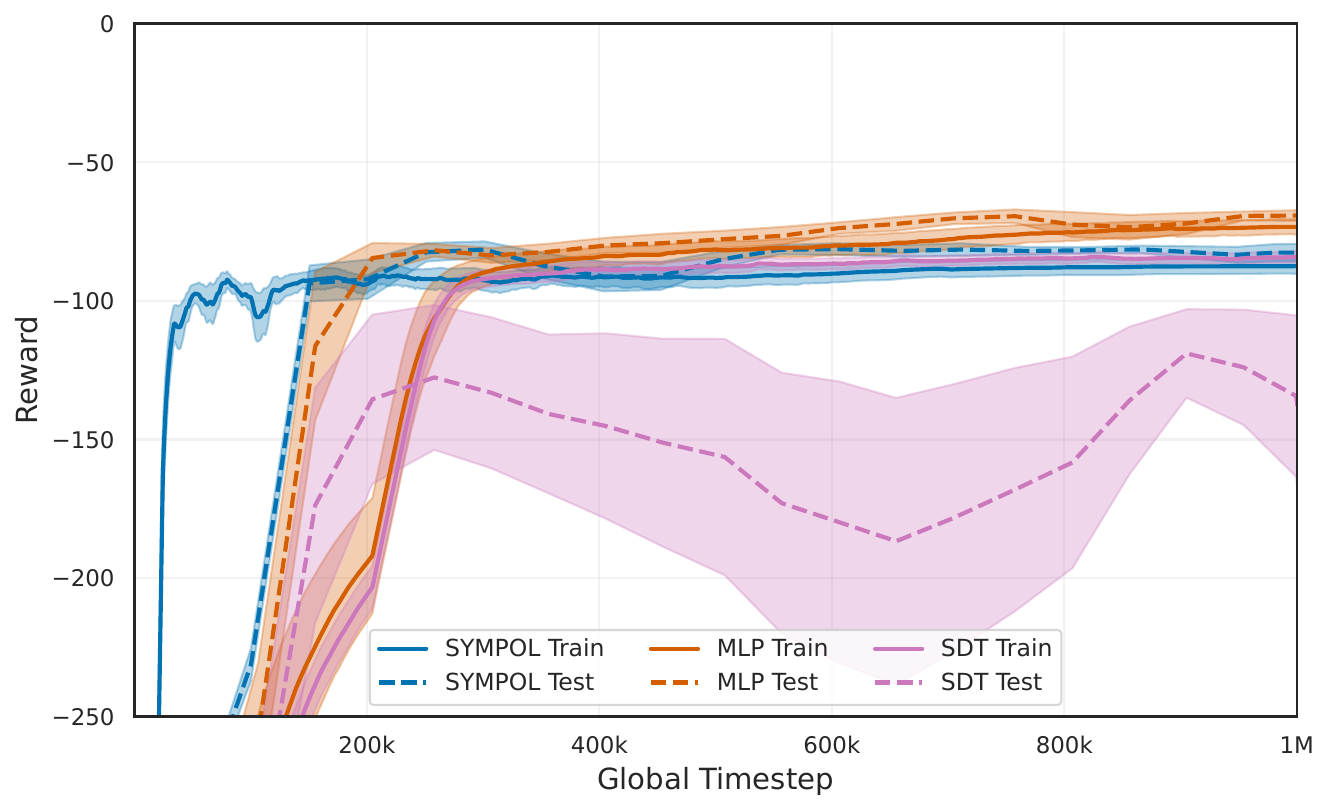}
        \caption{AB}
        \label{fig:learning_curves_Acrobot}
    \end{subfigure}
    \:
    \begin{subfigure}[tb]{0.32\textwidth}
        \centering
        \includegraphics[width=\linewidth]{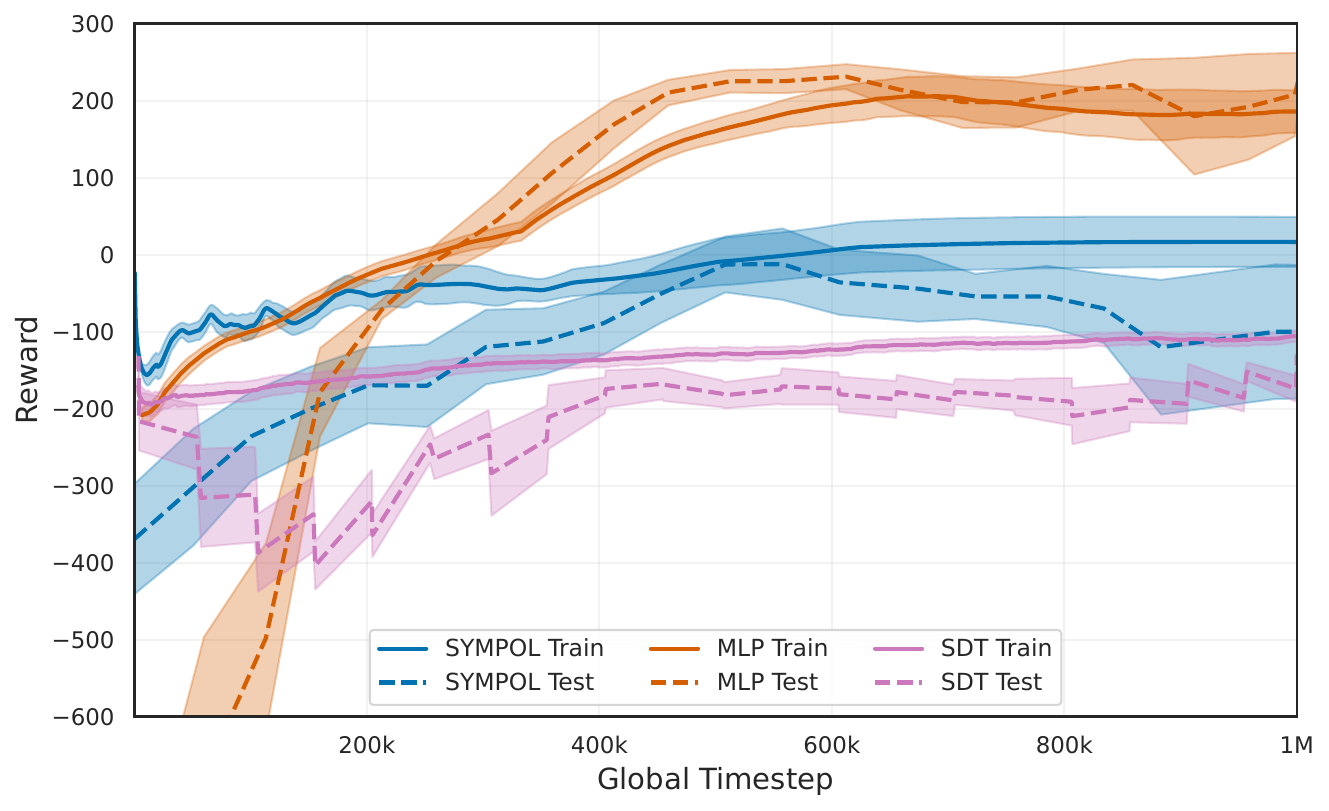}
        \caption{LL}
        \label{fig:learning_curves_LunarLander}
    \end{subfigure}   
    \qquad
    \begin{subfigure}[tb]{0.32\textwidth}
        \centering
        \includegraphics[width=\linewidth]{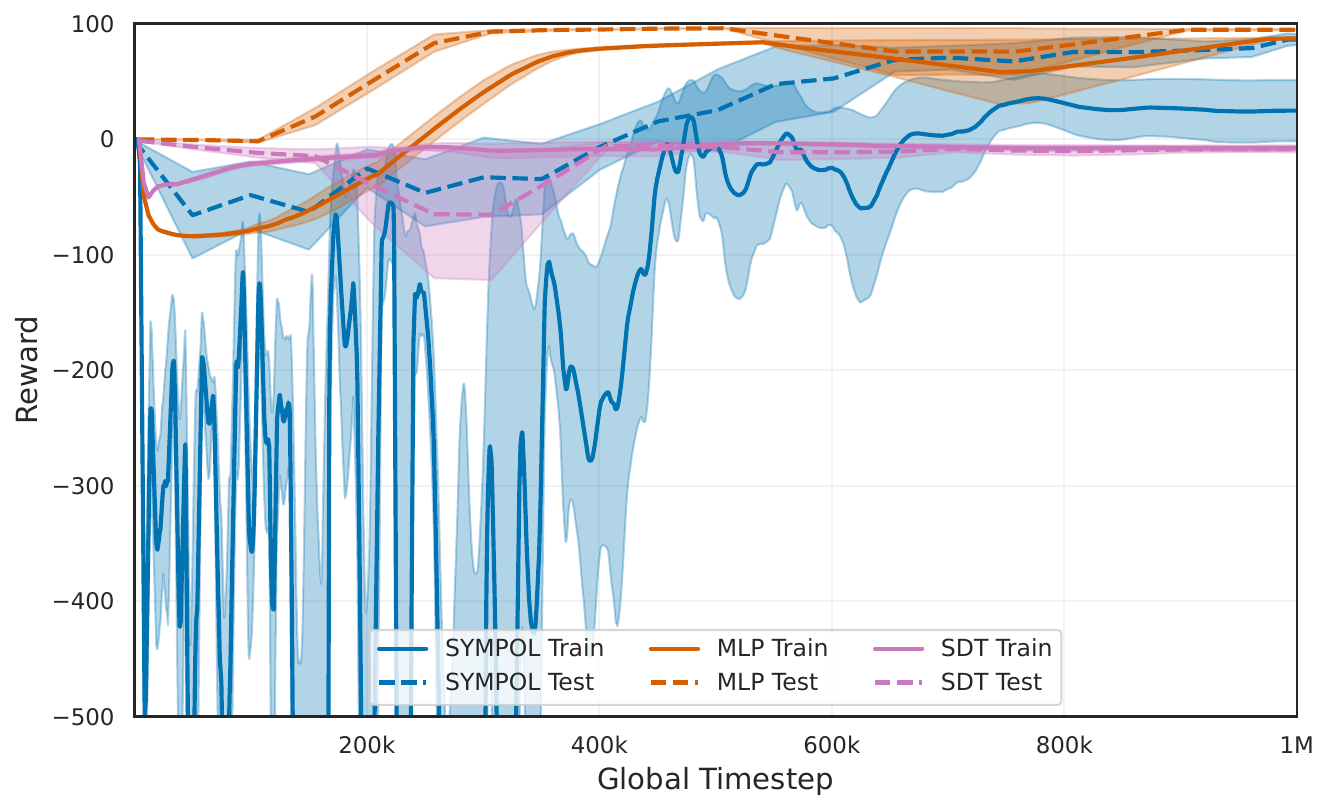}
        \caption{MC-C}
        \label{fig:learning_curves_MountainCarCont}
    \end{subfigure} 
    \:
    \begin{subfigure}[tb]{0.32\textwidth}
        \centering
        \includegraphics[width=\linewidth]{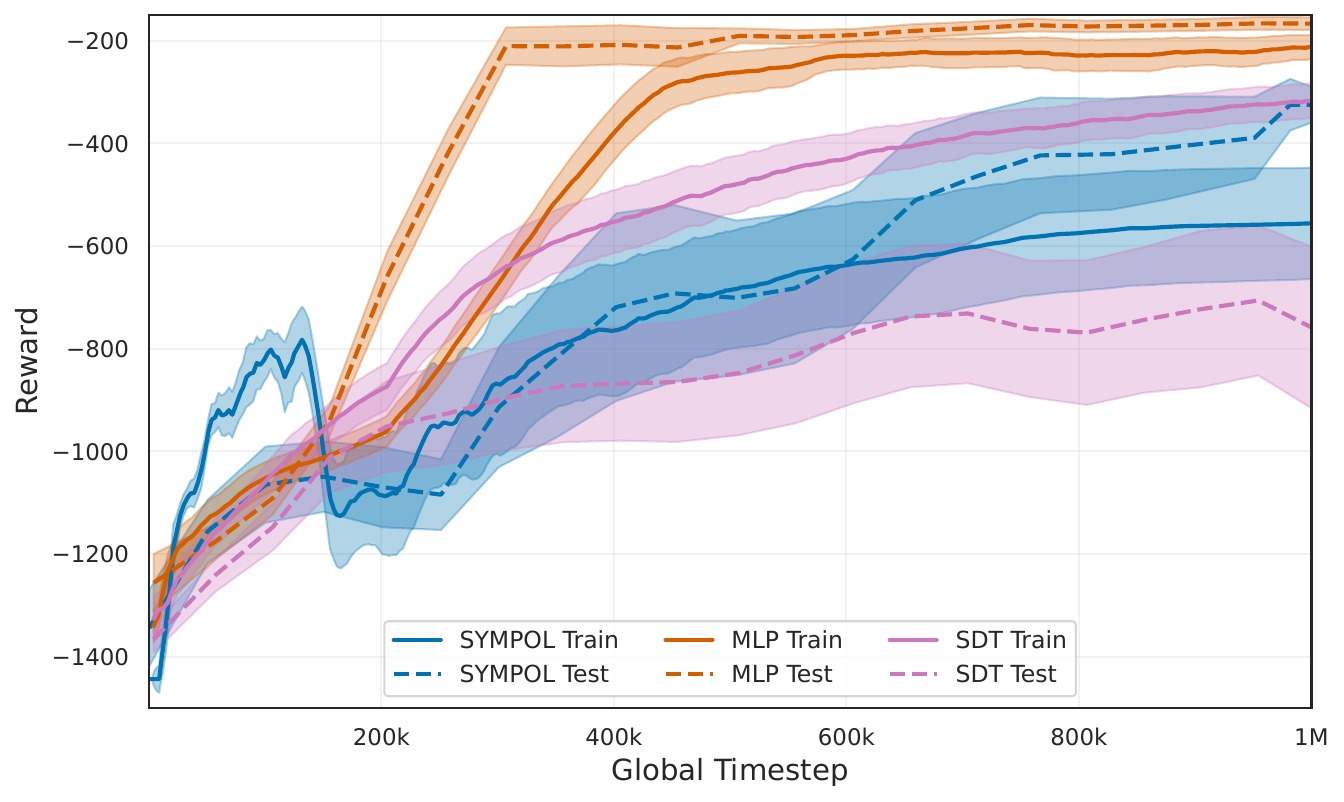}
        \caption{PD-C}
        \label{fig:learning_curves_Pendulum}
    \end{subfigure}    
    \caption[Training Curves (Full-Complexity).]{\textbf{Training Curves (Full-Complexity).} Shows the training reward as solid line and the test reward as dashed line for SYMPOL (blue), MLP (orange) and SDT (green).}
    \label{fig:training_curves}
\end{figure}

\begin{figure}[H]
    \centering
    \begin{subfigure}[tb]{0.32\textwidth}
        \centering
        \includegraphics[width=\linewidth]{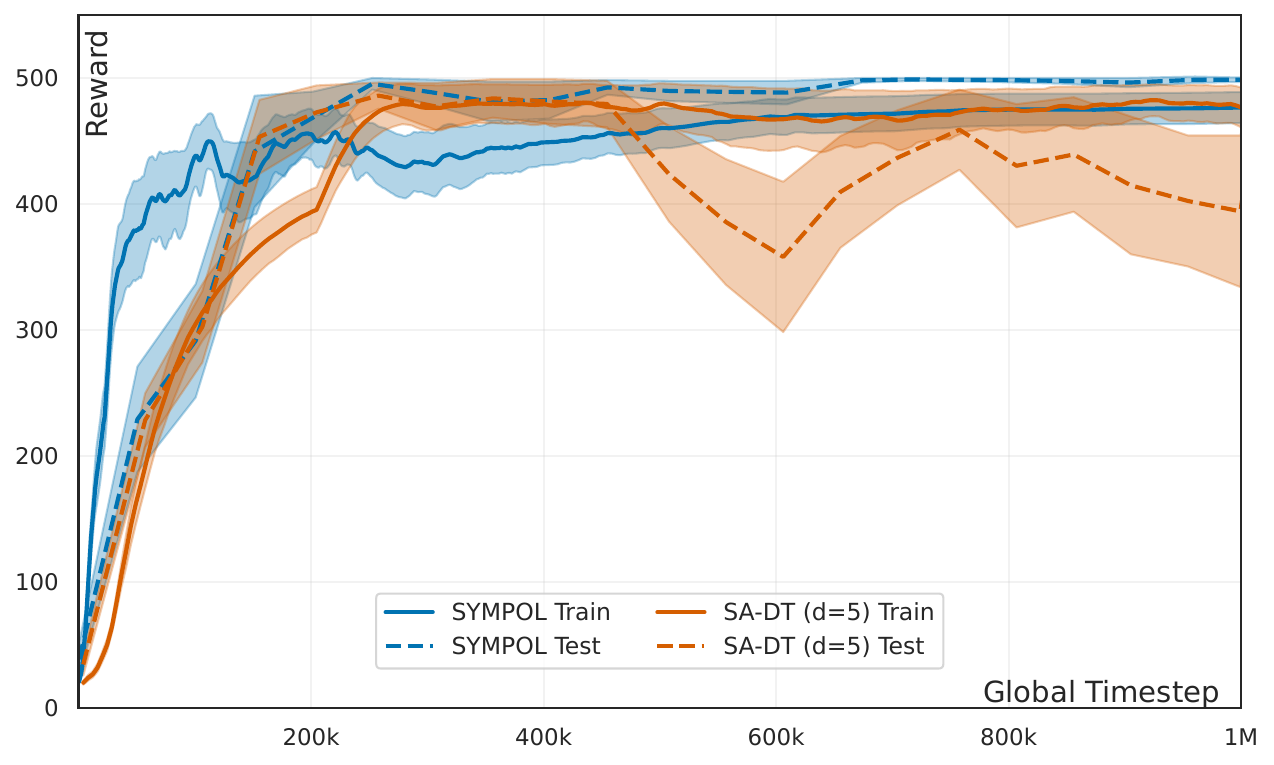}
        \caption{CP SYMPOL vs. SA-DT (d=5)}
        \label{fig:learning_curves_sadt5CartPole}
    \end{subfigure}
    \:
    \begin{subfigure}[tb]{0.32\textwidth}
        \centering
        \includegraphics[width=\linewidth]{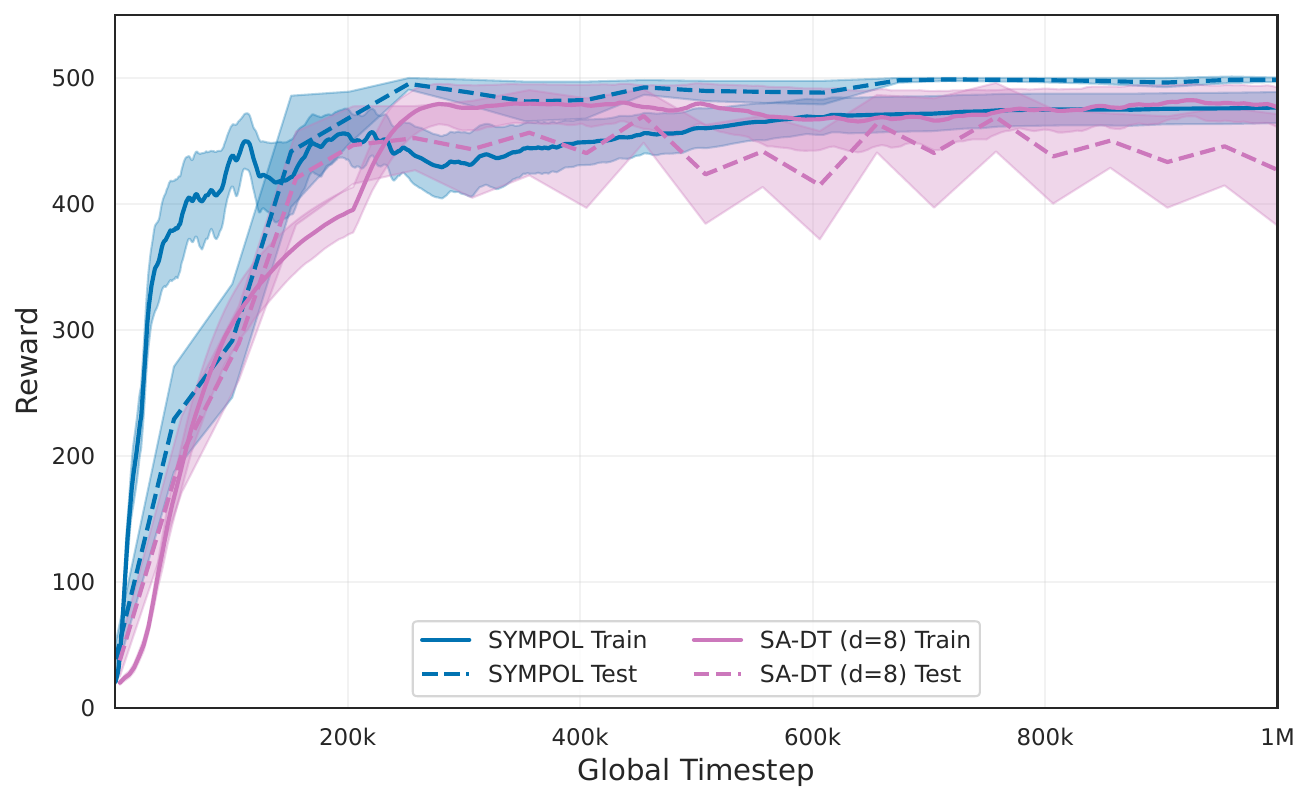}
        \caption{CP SYMPOL vs. SA-DT (d=8)}
        \label{fig:learning_curves_sadt8CartPole}
    \end{subfigure}
    \:
    \begin{subfigure}[tb]{0.32\textwidth}
        \centering
        \includegraphics[width=\linewidth]{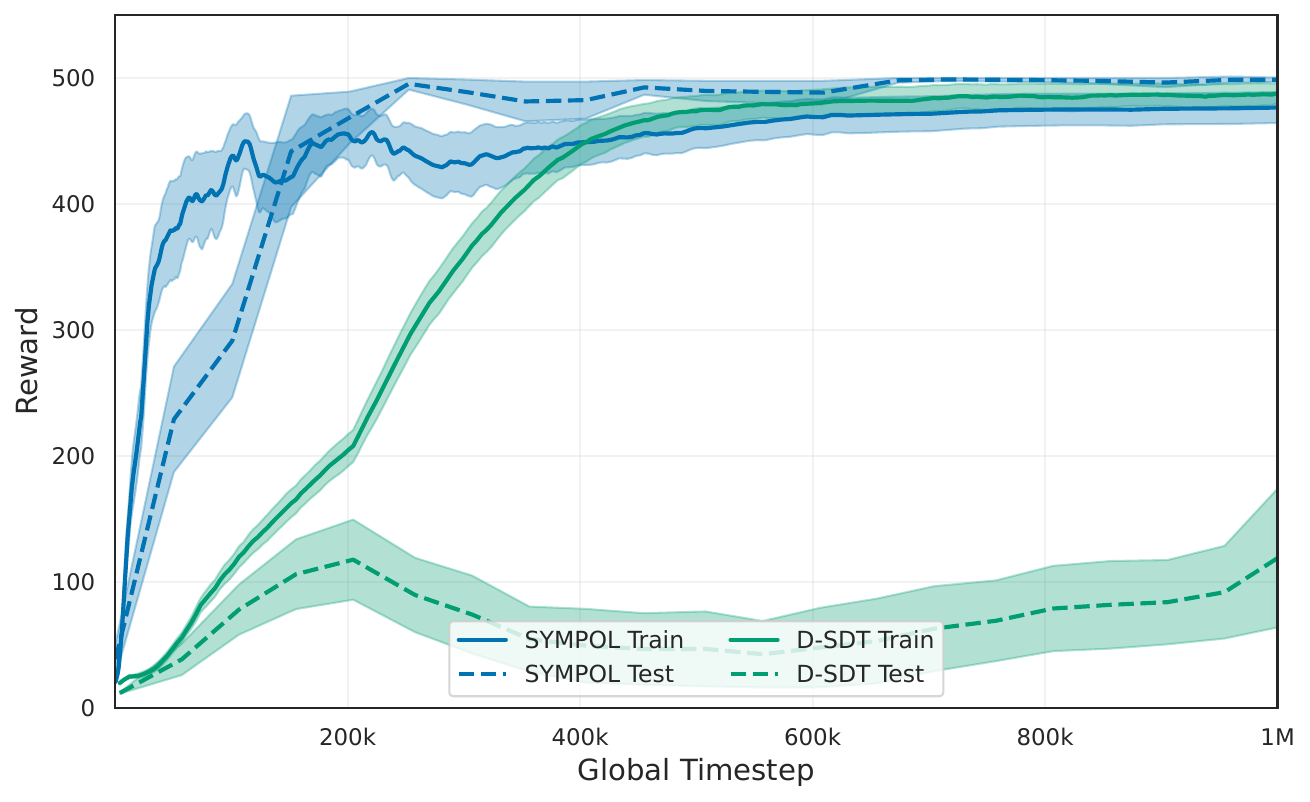}
        \caption{CP SYMPOL vs. D-SDT}
        \label{fig:learning_curves_dsdtCartPole}
    \end{subfigure}   
    \quad
    \begin{subfigure}[tb]{0.32\textwidth}
        \centering
        \includegraphics[width=\linewidth]{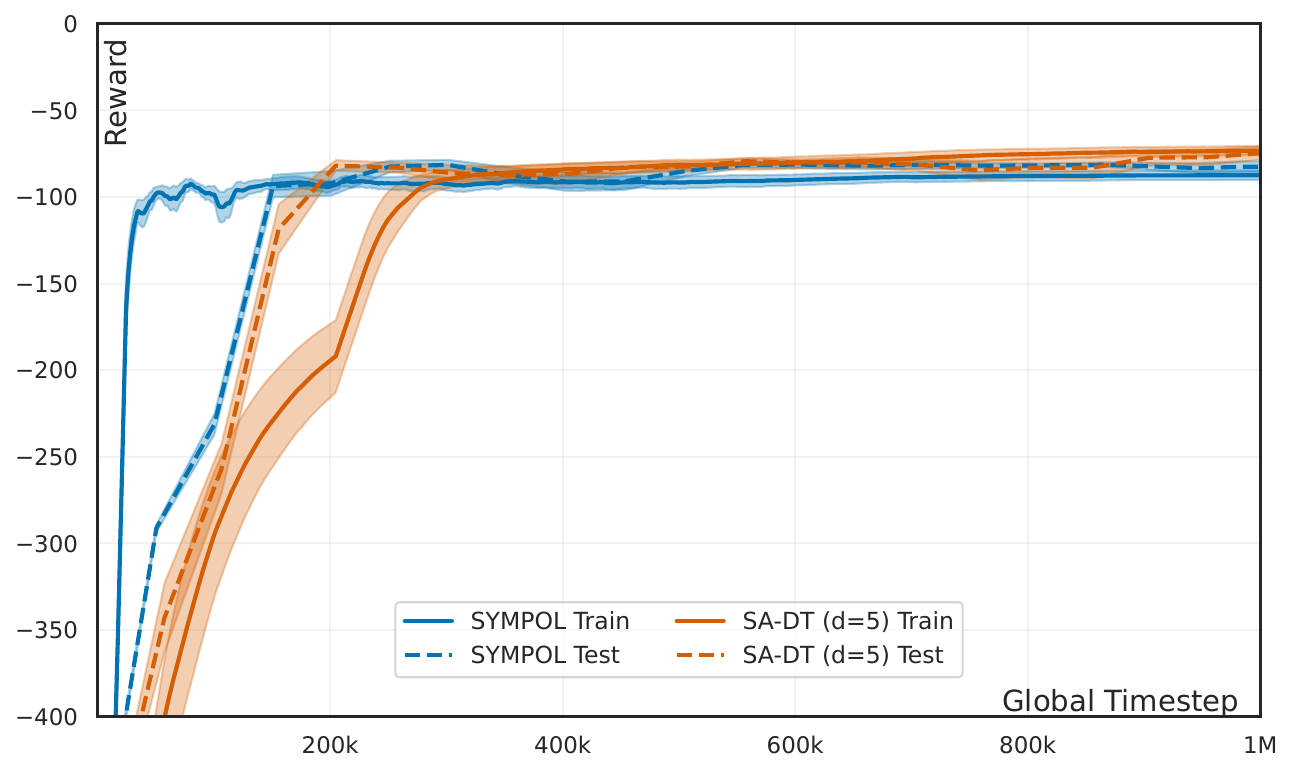}
        \caption{AB SYMPOL vs. SA-DT (d=5)}
        \label{fig:learning_curves_sadt5Acrobot}
    \end{subfigure}
    \:
    \begin{subfigure}[tb]{0.32\textwidth}
        \centering
        \includegraphics[width=\linewidth]{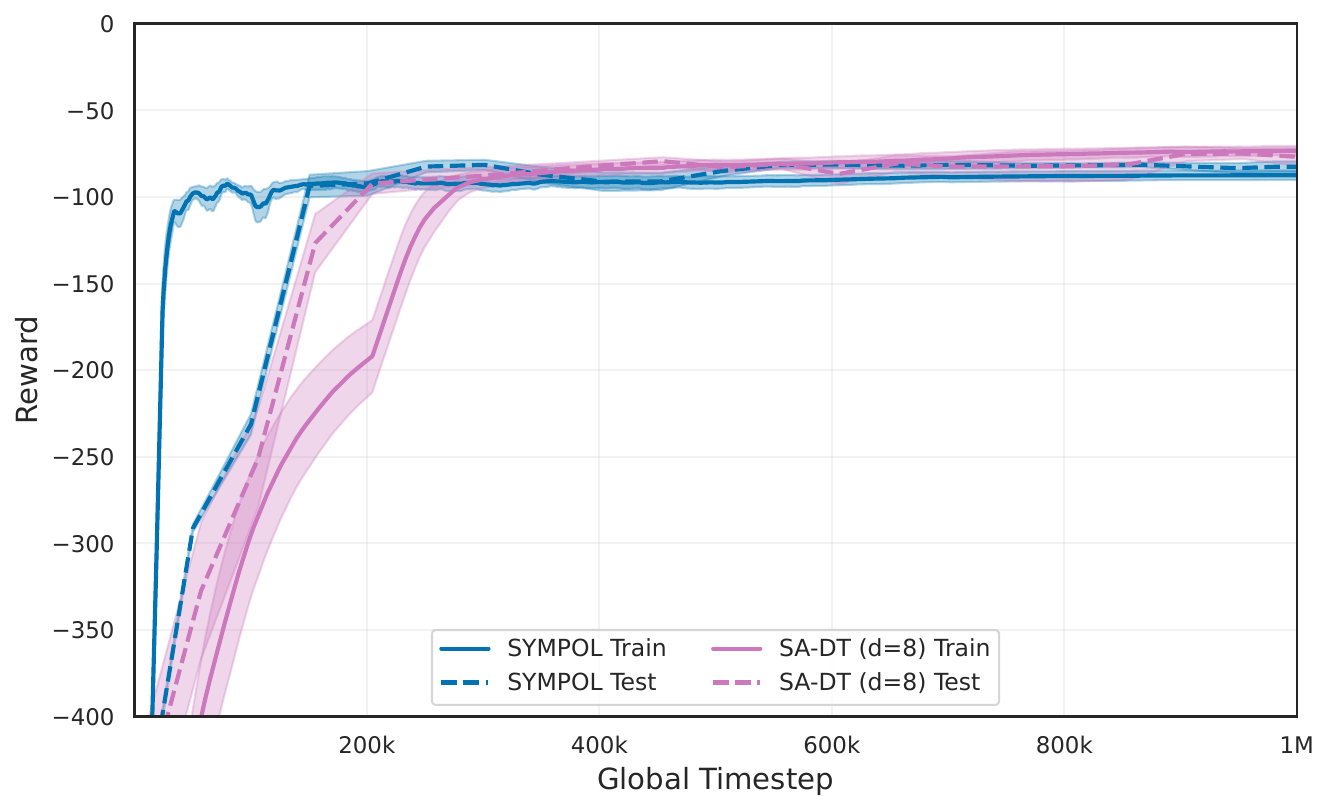}
        \caption{AB SYMPOL vs. SA-DT (d=8)}
        \label{fig:learning_curves_sadt8Acrobot}
    \end{subfigure}
    \:
    \begin{subfigure}[tb]{0.32\textwidth}
        \centering
        \includegraphics[width=\linewidth]{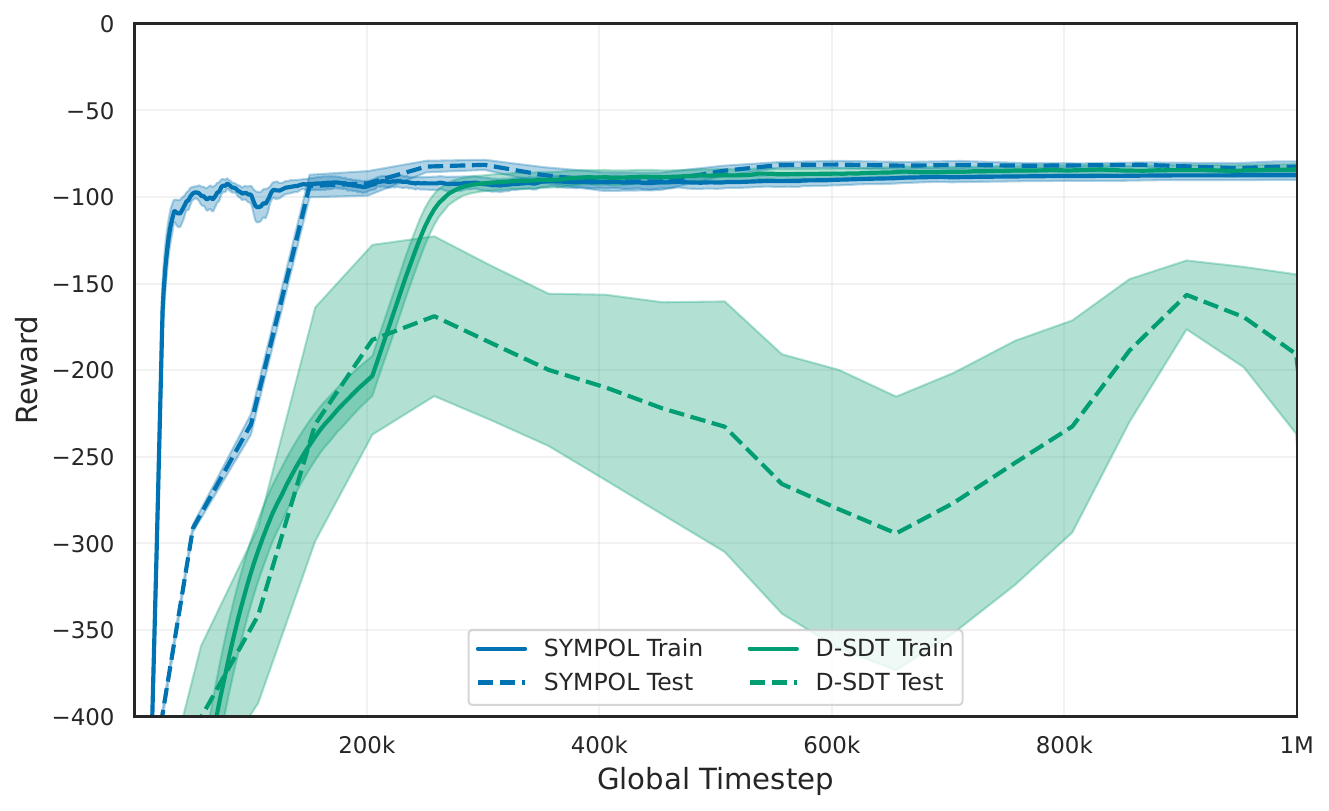}
        \caption{AB SYMPOL vs. D-SDT}
        \label{fig:learning_curves_dsdtAcrobot}
    \end{subfigure}   
    \quad
    \begin{subfigure}[tb]{0.32\textwidth}
        \centering
        \includegraphics[width=\linewidth]{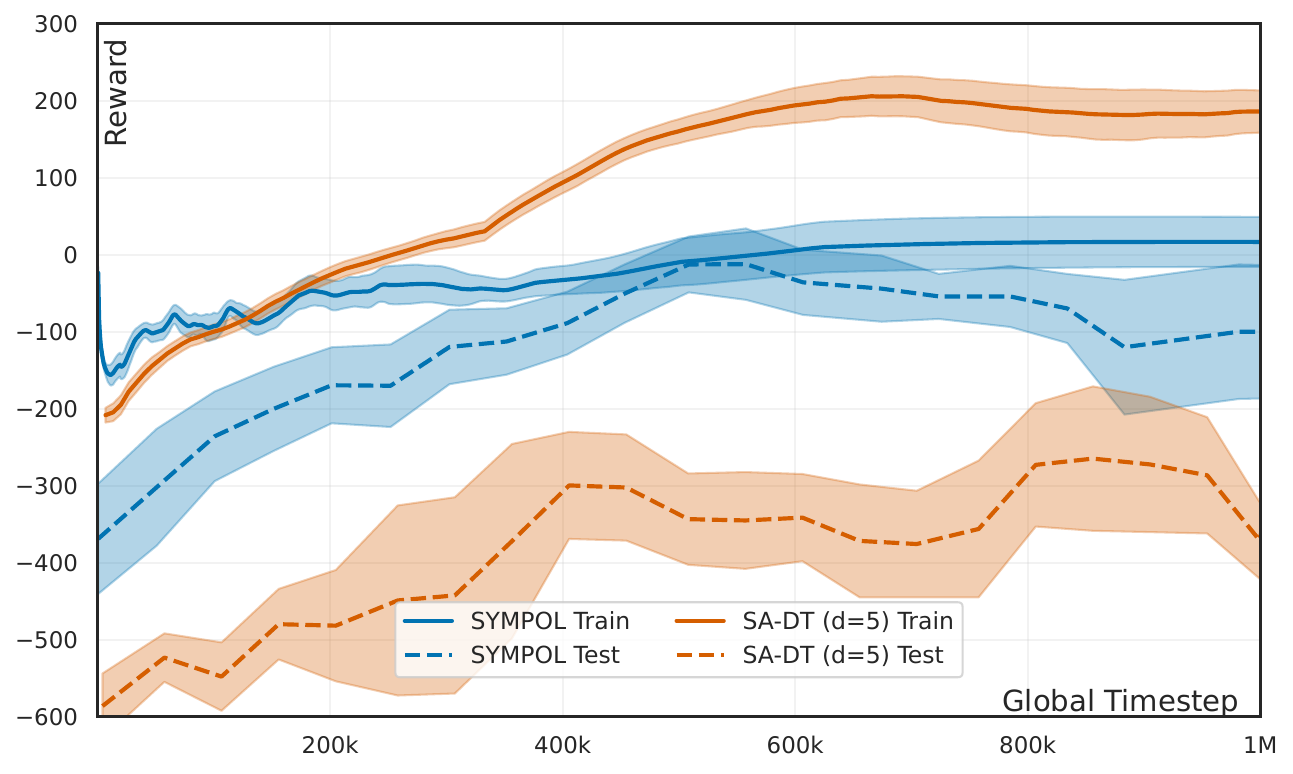}
        \caption{LL SYMPOL vs. SA-DT (d=5)}
        \label{fig:learning_curves_sadt5LunarLander}
    \end{subfigure}
    \:
    \begin{subfigure}[tb]{0.32\textwidth}
        \centering
        \includegraphics[width=\linewidth]{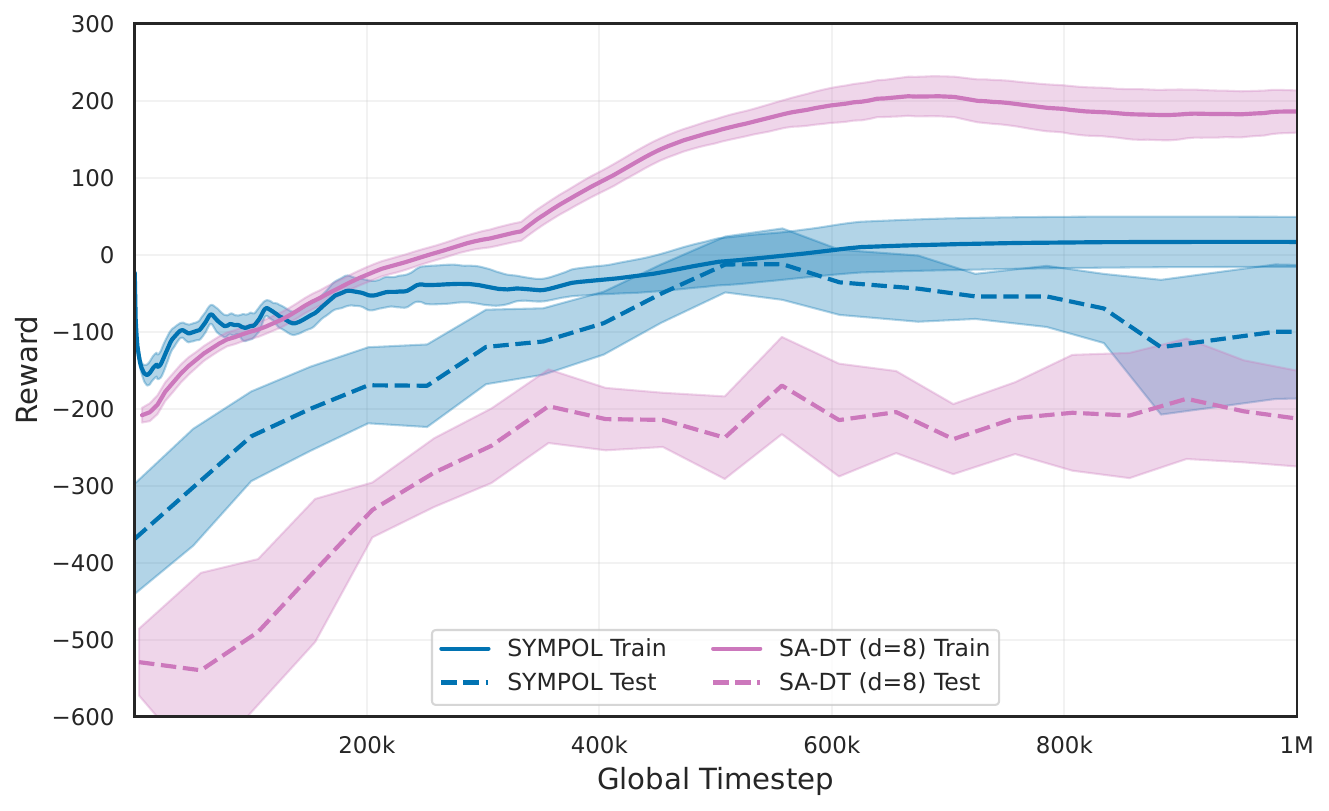}
        \caption{LL SYMPOL vs. SA-DT (d=8)}
        \label{fig:learning_curves_sadt8LunarLander}
    \end{subfigure}
    \:
    \begin{subfigure}[tb]{0.32\textwidth}
        \centering
        \includegraphics[width=\linewidth]{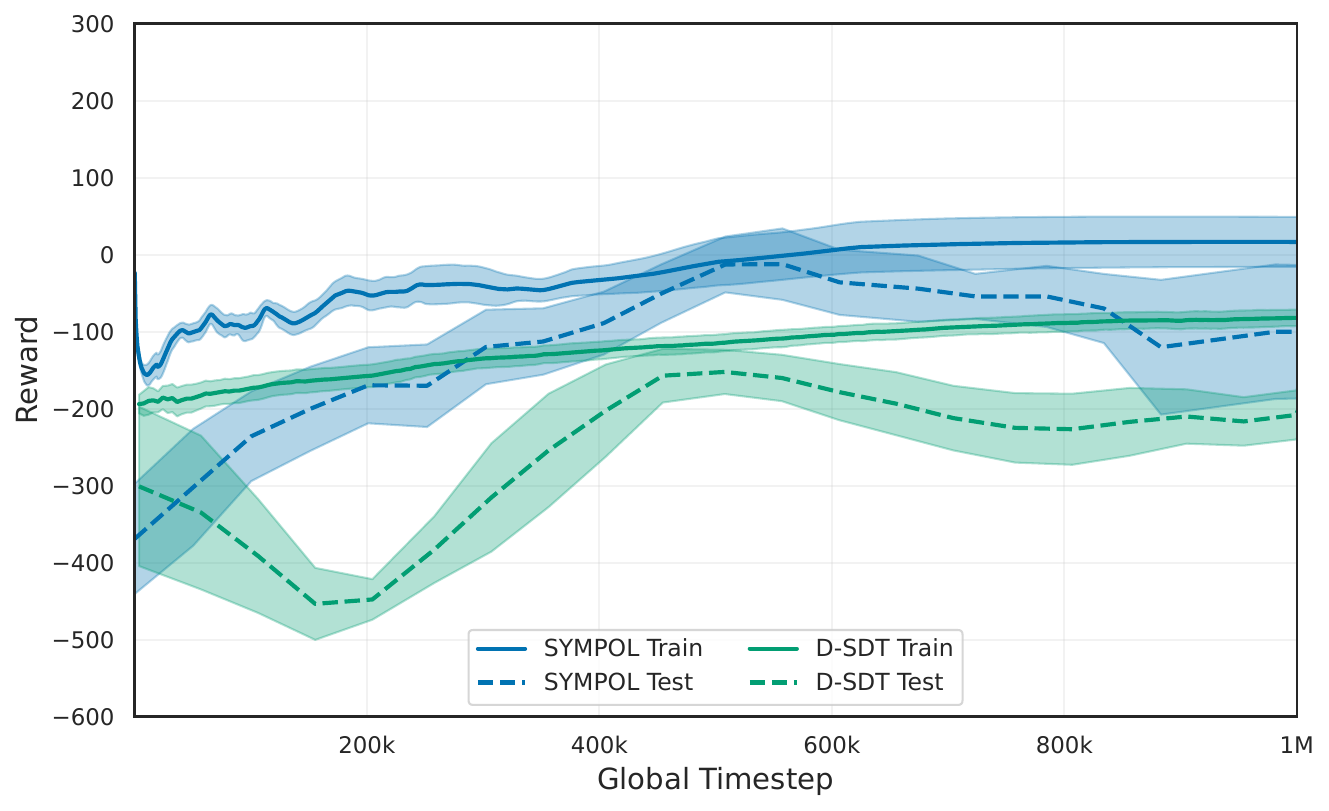}
        \caption{LL SYMPOL vs. D-SDT}
        \label{fig:learning_curves_dsdtLunarLander}
    \end{subfigure}   
    \quad
    \begin{subfigure}[tb]{0.32\textwidth}
        \centering
        \includegraphics[width=\linewidth]{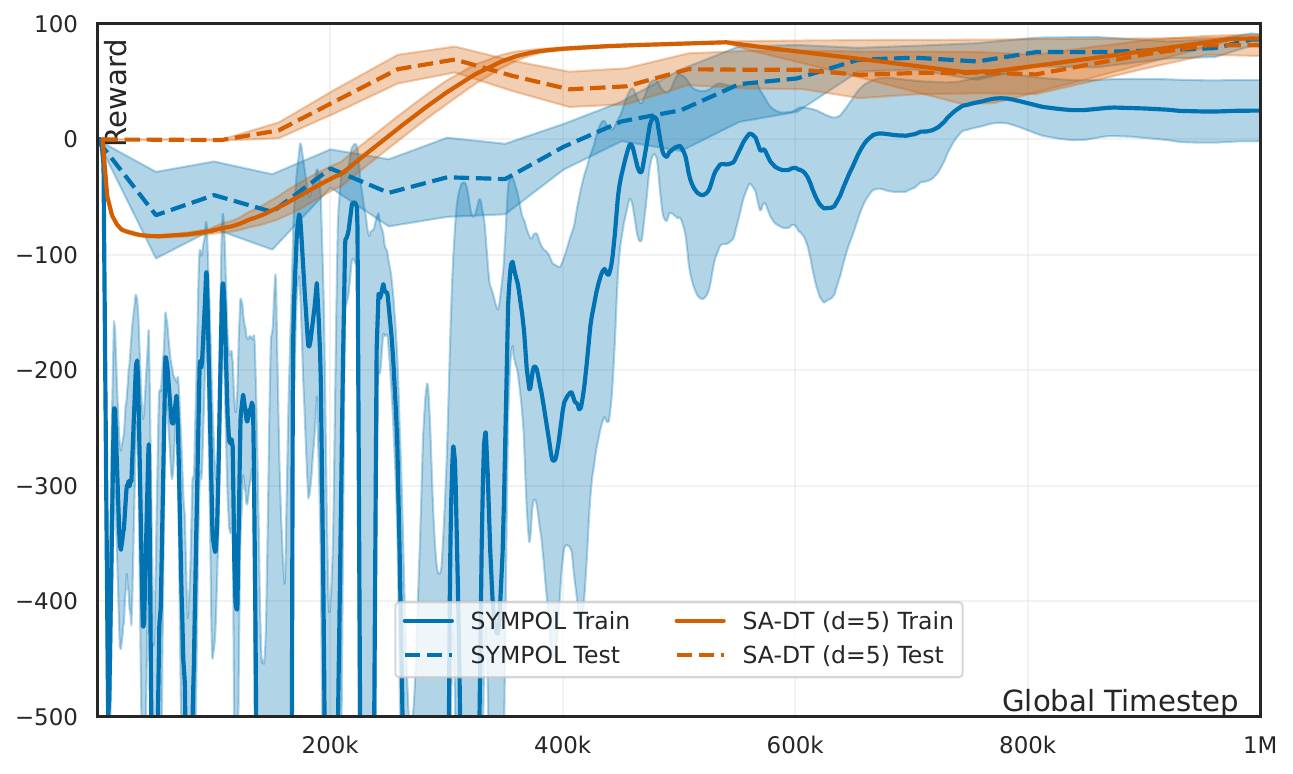}
        \caption{MC-C SYMPOL vs. SA-DT (d=5)}
        \label{fig:learning_curves_sadt5MountainCarCont}
    \end{subfigure}
    \:
    \begin{subfigure}[tb]{0.32\textwidth}
        \centering
        \includegraphics[width=\linewidth]{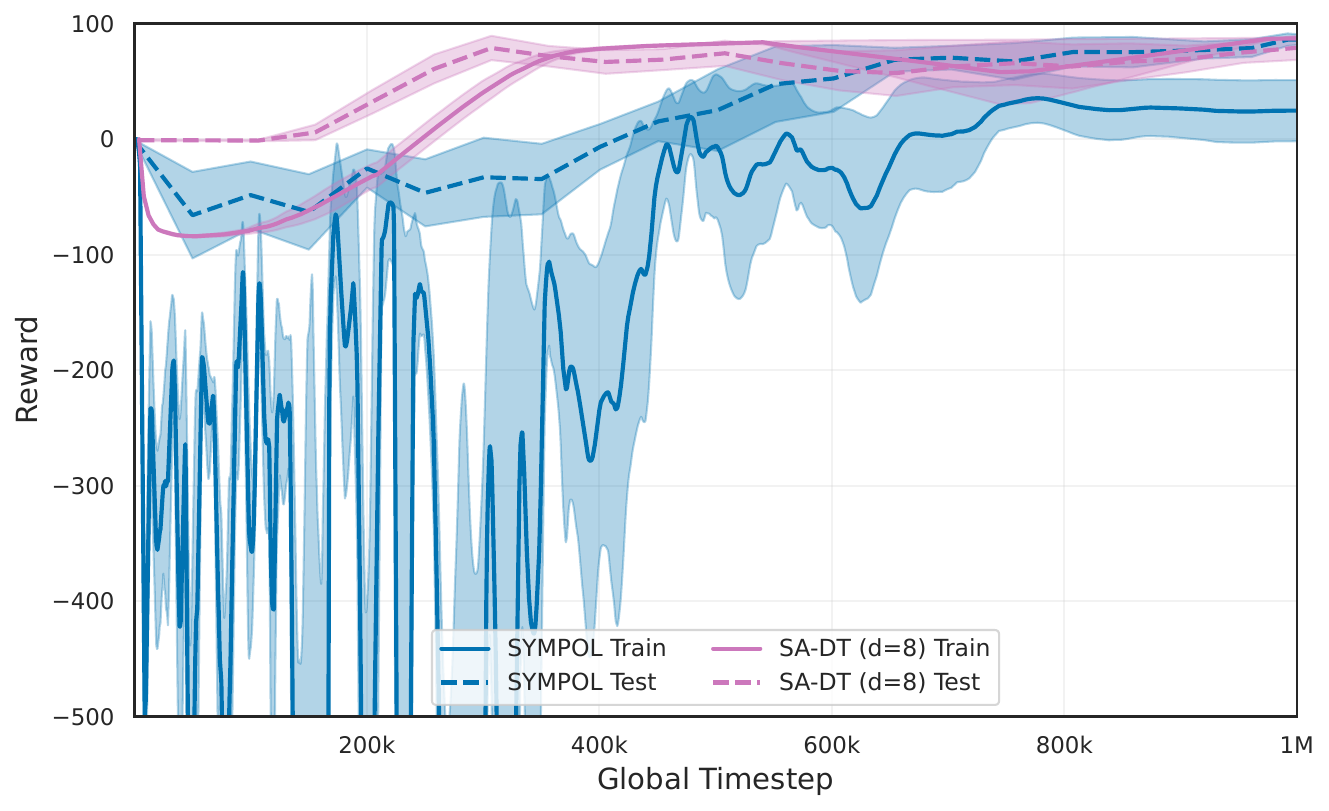}
        \caption{MC-C SYMPOL vs. SA-DT (d=8)}
        \label{fig:learning_curves_sadt8MountainCarCont}
    \end{subfigure}
    \:
    \begin{subfigure}[tb]{0.32\textwidth}
        \centering
        \includegraphics[width=\linewidth]{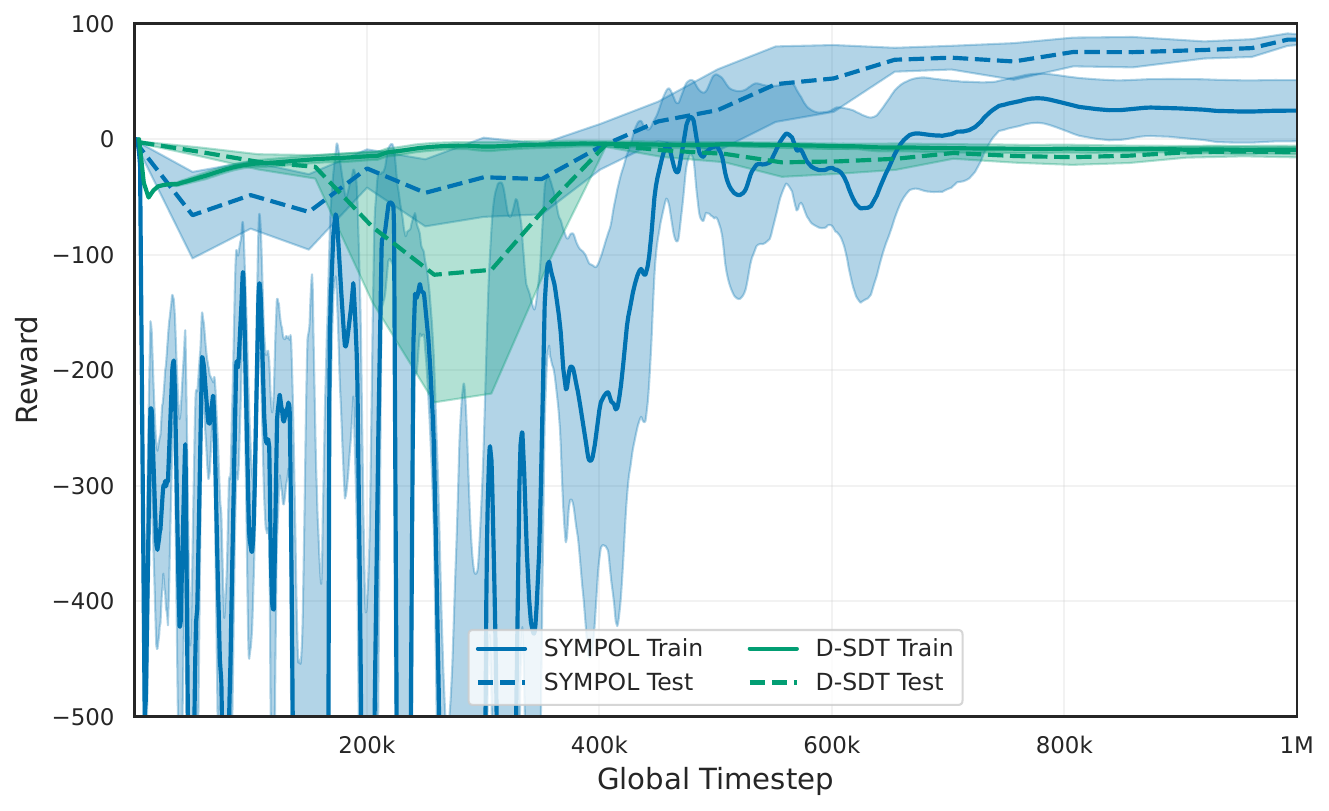}
        \caption{MC-C SYMPOL vs. D-SDT}
        \label{fig:learning_curves_dsdtMountainCarCont}
    \end{subfigure}       
    \quad
    \begin{subfigure}[tb]{0.32\textwidth}
        \centering
        \includegraphics[width=\linewidth]{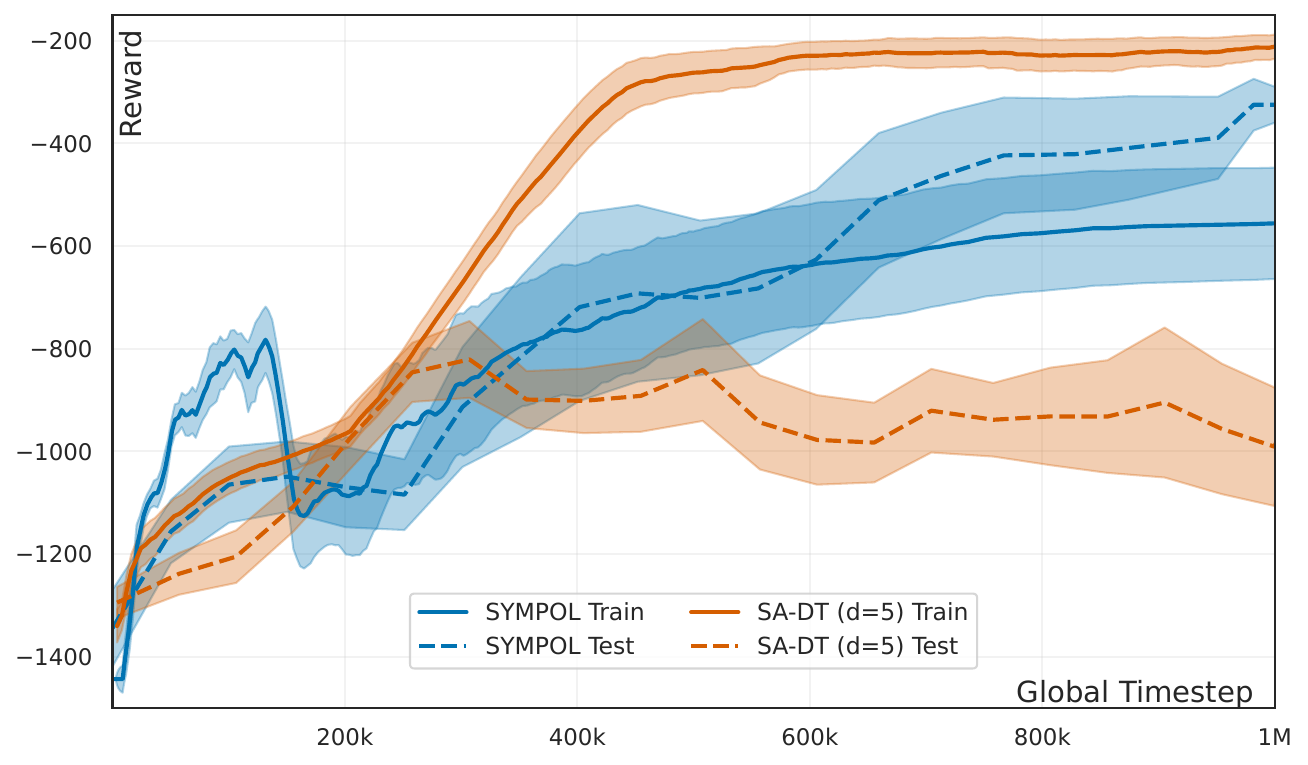}
        \caption{PC-C SYMPOL vs. SA-DT (d=5)}
        \label{fig:learning_curves_sadt5Pendulum}
    \end{subfigure}
    \:
    \begin{subfigure}[tb]{0.32\textwidth}
        \centering
        \includegraphics[width=\linewidth]{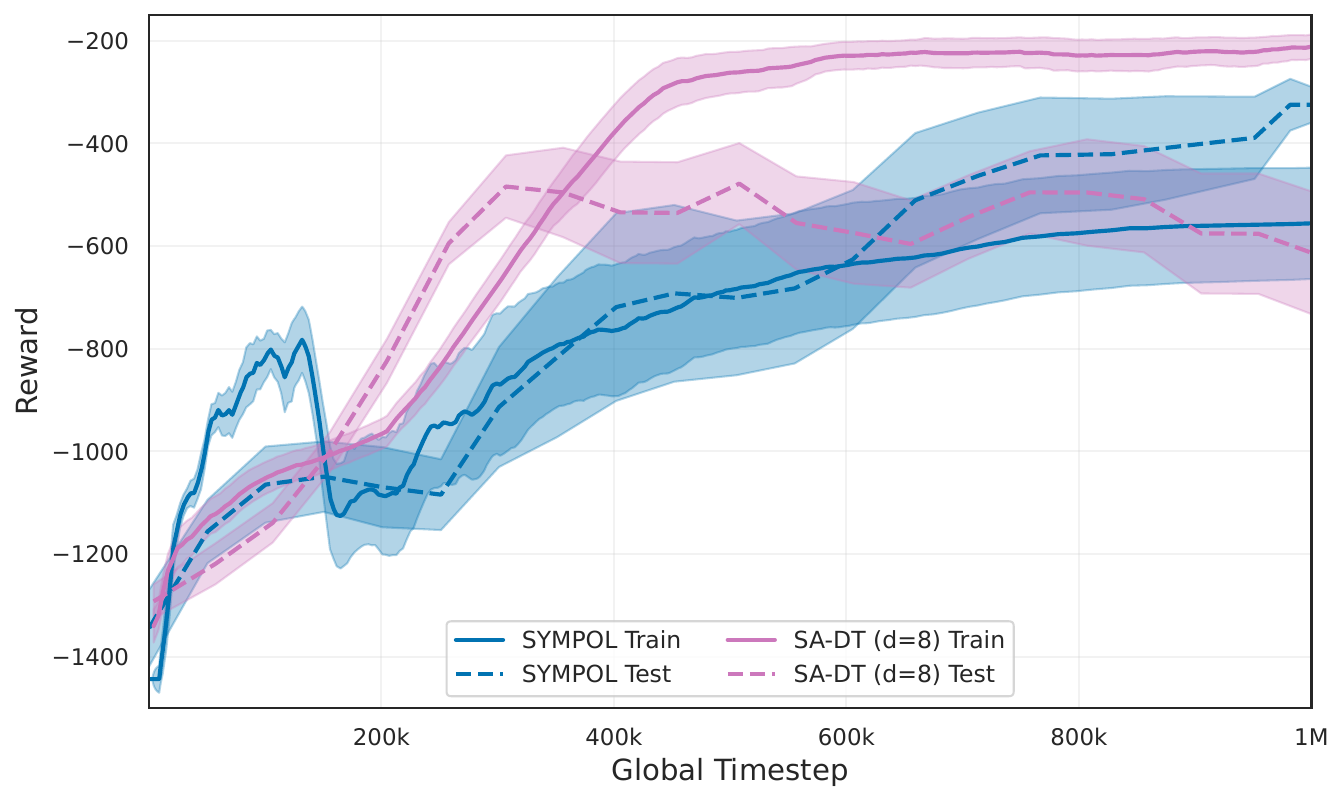}
        \caption{PC-C SYMPOL vs. SA-DT (d=8)}
        \label{fig:learning_curves_sadt8Pendulum}
    \end{subfigure}
    \:
    \begin{subfigure}[tb]{0.32\textwidth}
        \centering
        \includegraphics[width=\linewidth]{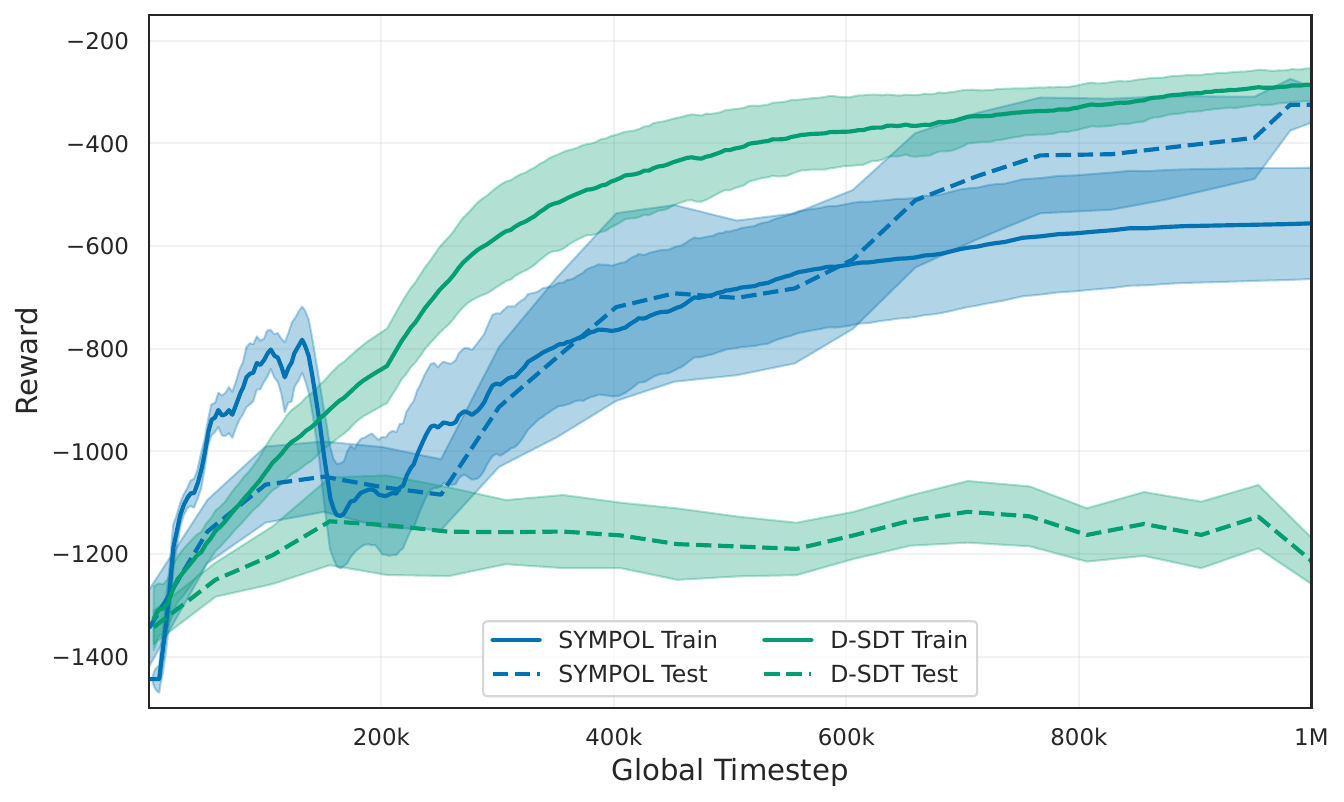}
        \caption{PC-C SYMPOL vs. D-SDT}
        \label{fig:learning_curves_dsdtPendulum}
    \end{subfigure}   
    
    \caption[Training Curves.]{\textbf{Training Curves.} Shows the training reward as solid line and the test reward as dashed line for SYMPOL (blue), SA-DT-5 (orange) SA-DT-8 (green) and D-SDT (red). Thereby, the test reward is calculated with the discretized/distilled policy for SA-DT and D-SDT. For several datasets, we can again observe the severe information loss introduced with the post-processing (e.g. for PD-C and LL).}
    \label{fig:training_curves_pairwise}
\end{figure}

\newpage

\section{MiniGrid} \label{A:envs}
We used the MiniGrid implementation from \citep{minigrid}. For each environment, we limited to observations and the actions to the available ones according to the documentation. Furthermore, we decided to use a view size of 3 to allow a good visualization of the results.
In the following, we provide more examples for our MiniGrid Use-Case, along with more detailed visualizations.
In the following, we visualized the SYMPOL agent sequentially acting in the environment as one image for one step from left to right and top to bottom. Figure~\ref{fig:sympol_basic} shows how SYMPOL (see Figure~\ref{fig:sympol_distshift} in the main part or \texttt{tree\_function(obs)} defined below) solves the environment. Figure~\ref{fig:sympol_random_1_basic} and Figure~\ref{fig:sympol_random_2_basic} show the same agent failing on the environment with domain randomization, proving that the agent did not generalize, as we could already observe by inspecting the symbolic, tree-based policy. Retraining the agent with domain randomization (see Figure~\ref{fig:sympol_distshift_random} in the main part or \texttt{tree\_function\_retrained(obs)} defined below), SYMPOL is able to solve the environment (see Figure~\ref{fig:sympol_random_1_random} and Figure~\ref{fig:sympol_random_2_random}), maintaining interpretability.

\subsection{Visualizations Environment}

\vspace{0.5cm}

\begin{figure}[H]
    \centering
    \begin{tabular}{ccccc}
        \includegraphics[width=0.18\textwidth]{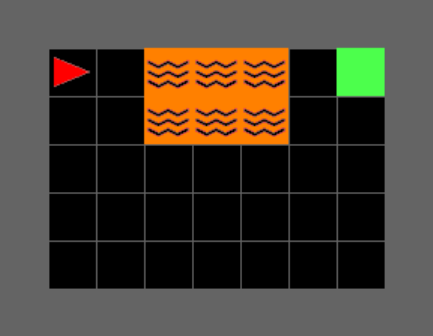} &
        \includegraphics[width=0.18\textwidth]{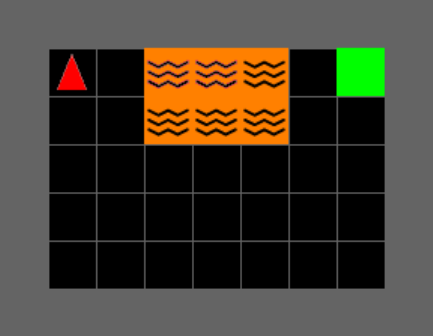} &
        \includegraphics[width=0.18\textwidth]{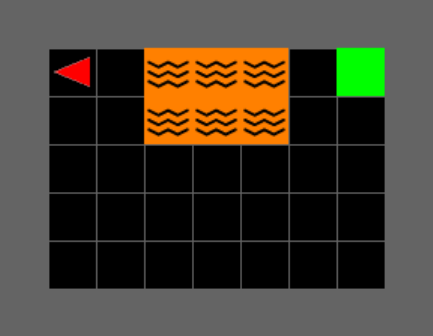} &
        \includegraphics[width=0.18\textwidth]{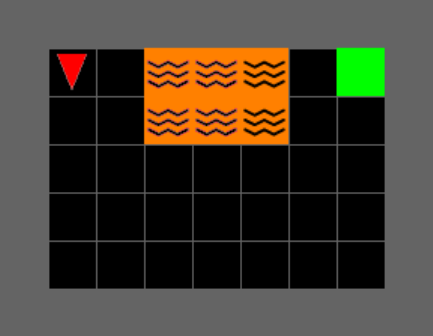} &
        \includegraphics[width=0.18\textwidth]{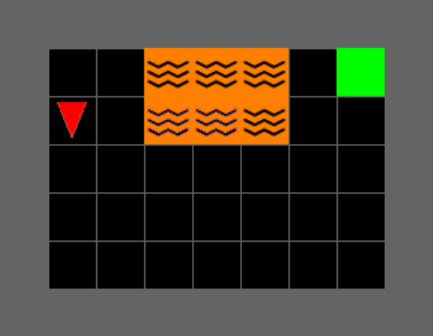} \\
        \includegraphics[width=0.18\textwidth]{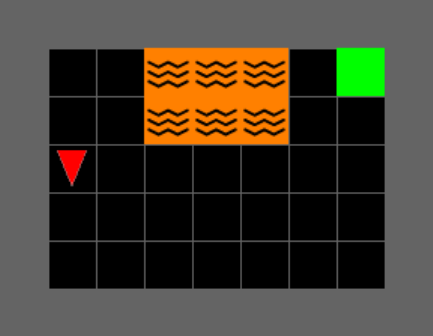} &
        \includegraphics[width=0.18\textwidth]{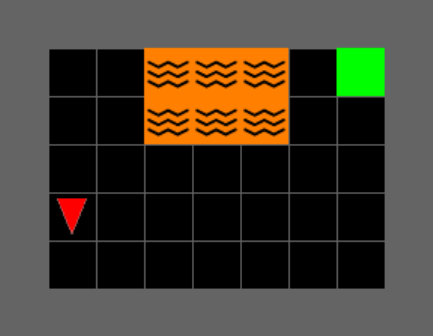} &
        \includegraphics[width=0.18\textwidth]{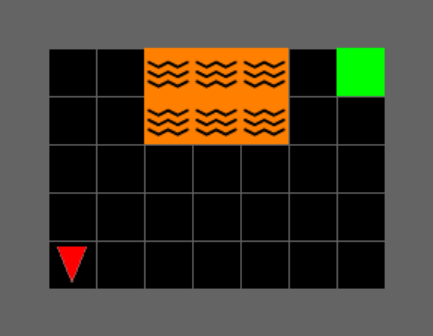} &
        \includegraphics[width=0.18\textwidth]{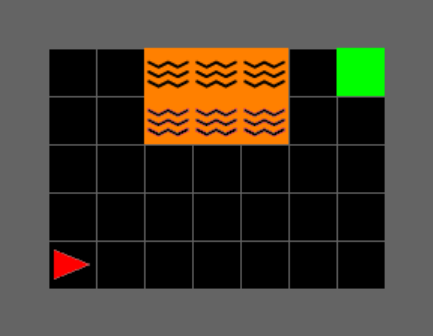} &
        \includegraphics[width=0.18\textwidth]{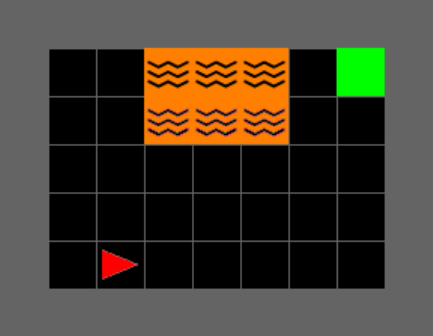} \\
        \includegraphics[width=0.18\textwidth]{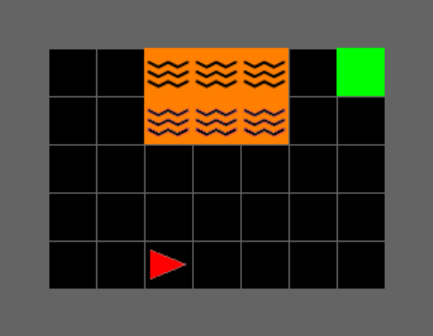} &
        \includegraphics[width=0.18\textwidth]{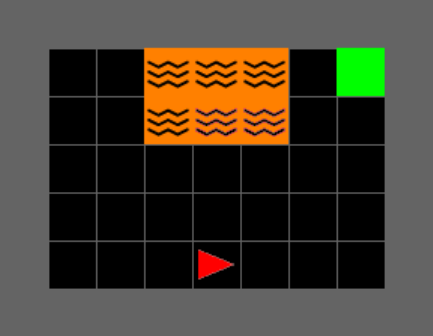} &
        \includegraphics[width=0.18\textwidth]{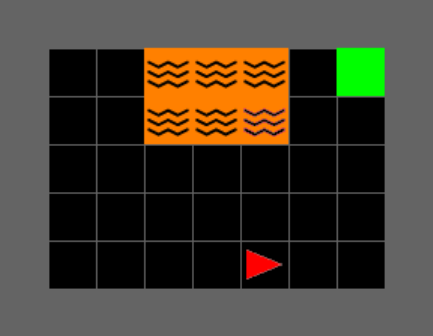} &
        \includegraphics[width=0.18\textwidth]{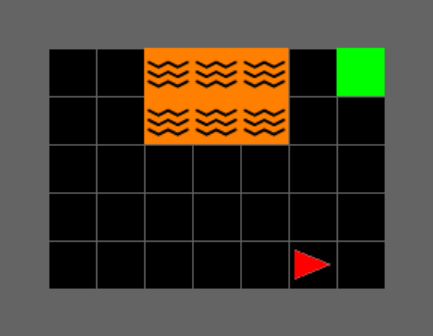} &
        \includegraphics[width=0.18\textwidth]{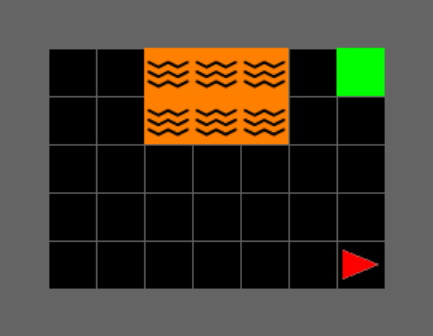} \\
        \includegraphics[width=0.18\textwidth]{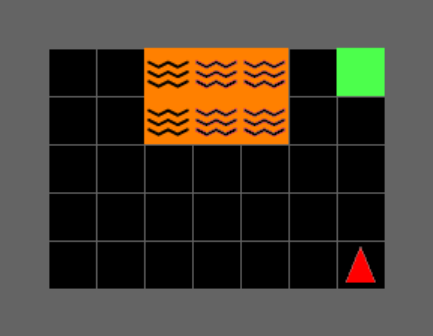} &
        \includegraphics[width=0.18\textwidth]{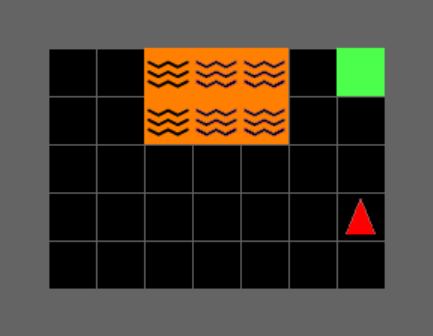} &
        \includegraphics[width=0.18\textwidth]{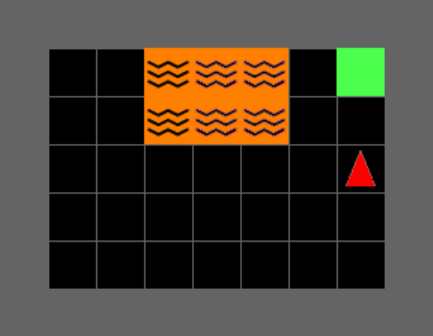} &
        \includegraphics[width=0.18\textwidth]{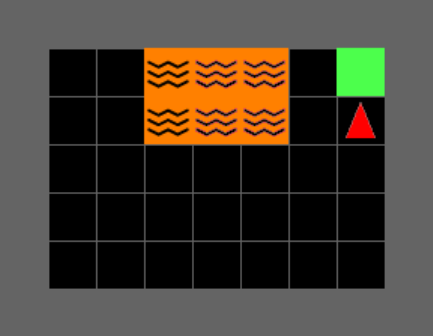} &
        \includegraphics[width=0.18\textwidth]{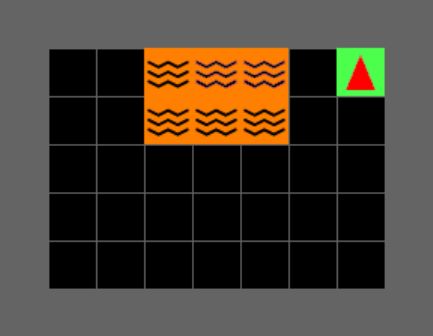} \\
    \end{tabular}
    \caption[DistShift SYMPOL.]{\textbf{DistShift SYMPOL.} This figure visualizes the path taken by the SYMPOL agent trained on the basic DistShift environment (see Figure~\ref{fig:sympol_distshift} in the main part or \texttt{tree\_function(obs)} defined below) from left to right and top to bottom. The agent follows the wall and reaches the goal at the top right corner.}
    \label{fig:sympol_basic}
\end{figure}

\begin{figure}[H]
    \centering
    \begin{tabular}{ccccc}
        \includegraphics[width=0.18\textwidth]{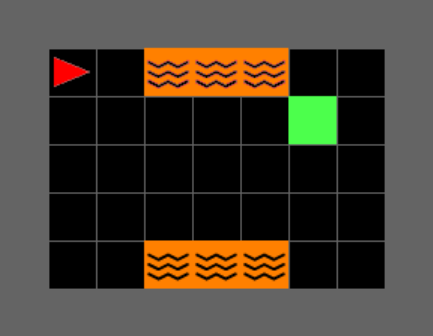} &
        \includegraphics[width=0.18\textwidth]{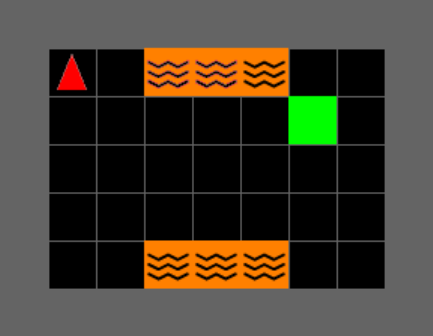} &
        \includegraphics[width=0.18\textwidth]{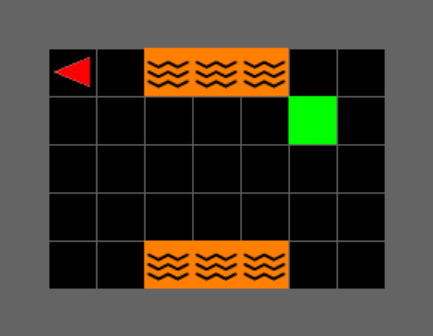} &
        \includegraphics[width=0.18\textwidth]{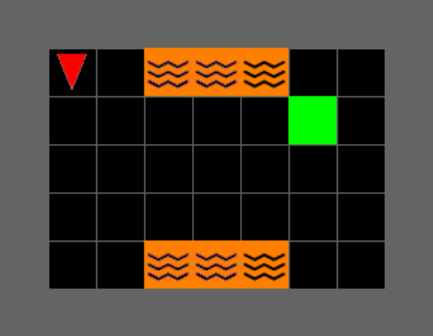} &
        \includegraphics[width=0.18\textwidth]{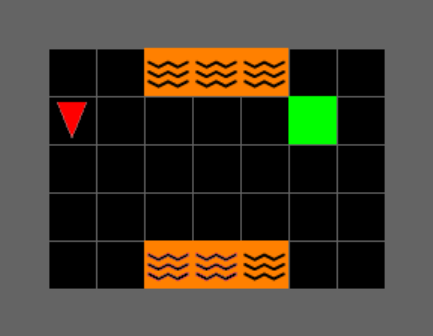} \\
        \includegraphics[width=0.18\textwidth]{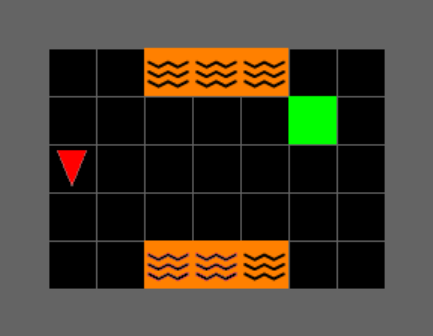} &
        \includegraphics[width=0.18\textwidth]{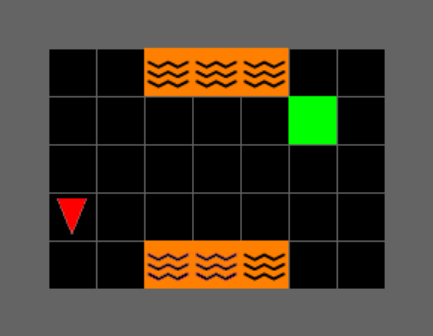} &
        \includegraphics[width=0.18\textwidth]{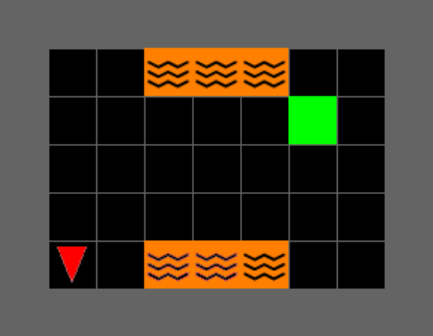} &
        \includegraphics[width=0.18\textwidth]{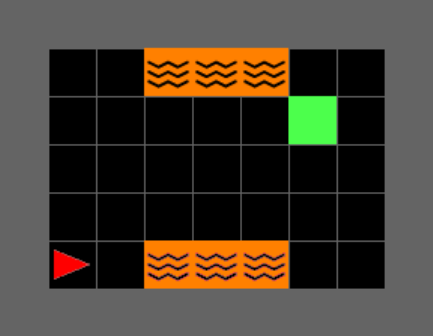} &
        \includegraphics[width=0.18\textwidth]{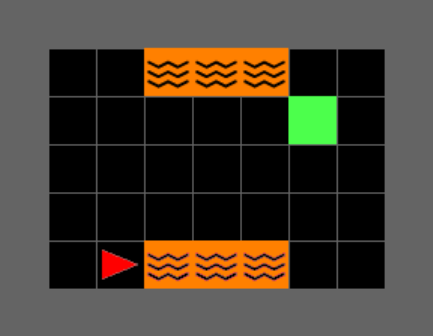} \\
        \includegraphics[width=0.18\textwidth]{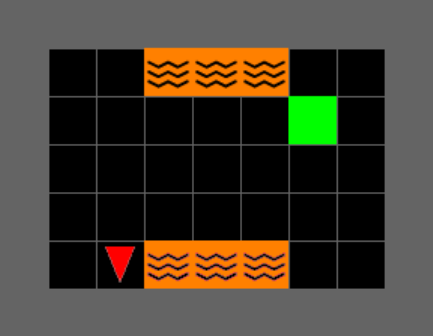} &
        \includegraphics[width=0.18\textwidth]{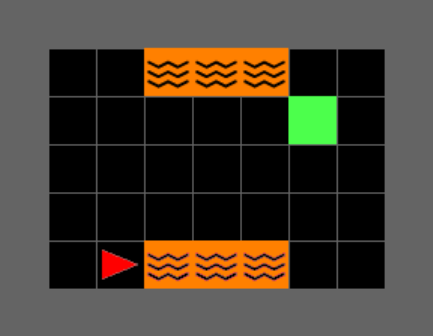} &
        \includegraphics[width=0.18\textwidth]{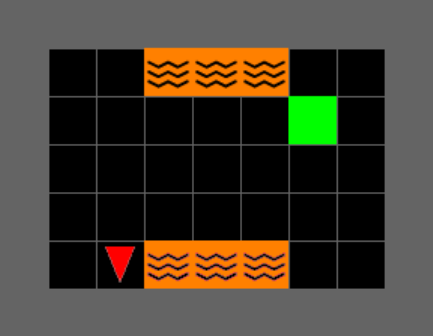} &
        \includegraphics[width=0.18\textwidth]{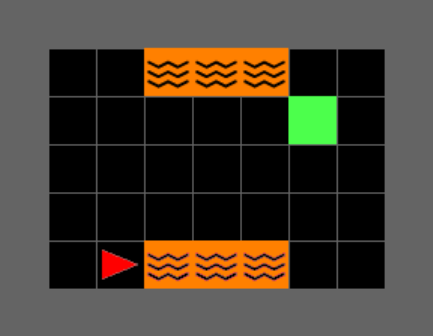} &
        \includegraphics[width=0.18\textwidth]{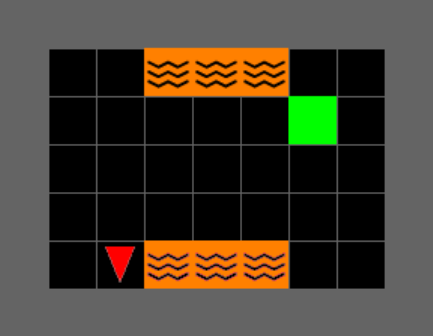} \\
    \end{tabular}
    \caption[DistShift (Domain Randomization) SYMPOL Example 1.]{\textbf{DistShift (Domain Randomization) SYMPOL Example 1.} This figure visualizes the path taken by the SYMPOL agent trained on the basic DistShift environment (see Figure~\ref{fig:sympol_distshift} in the main part or \texttt{tree\_function(obs)} defined below) from left to right and top to bottom. The agent follows the wall gets stuck by the lava.}
    \label{fig:sympol_random_1_basic}
\end{figure}

\vspace{0.5cm}

\begin{figure}[H]
    \centering
    \begin{tabular}{ccccc}
        \includegraphics[width=0.18\textwidth]{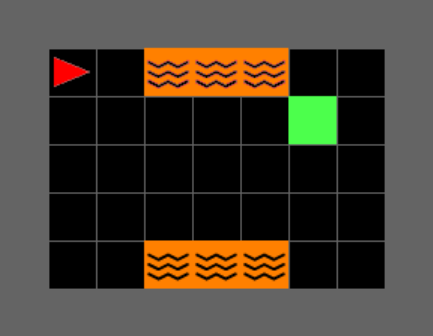} &
        \includegraphics[width=0.18\textwidth]{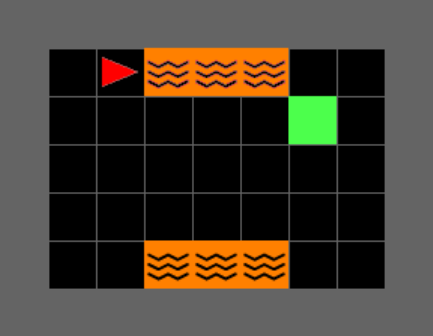} &
        \includegraphics[width=0.18\textwidth]{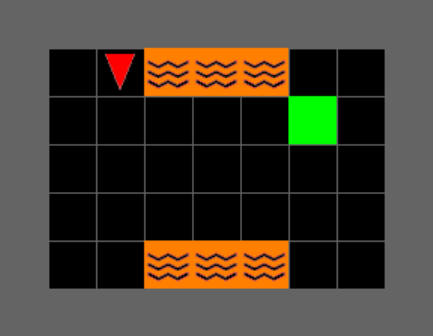} &
        \includegraphics[width=0.18\textwidth]{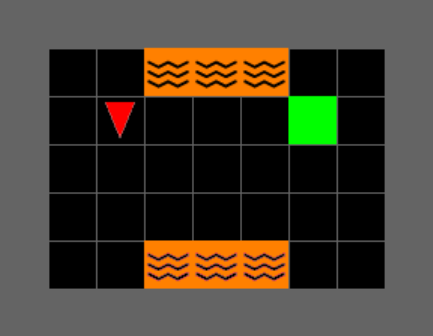} &
        \includegraphics[width=0.18\textwidth]{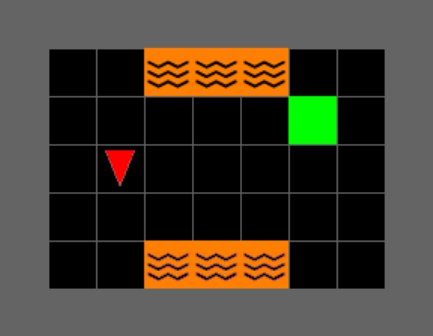} \\
        \includegraphics[width=0.18\textwidth]{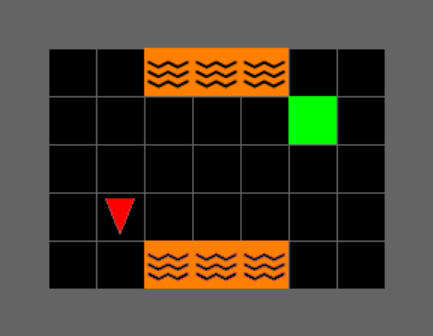} &
        \includegraphics[width=0.18\textwidth]{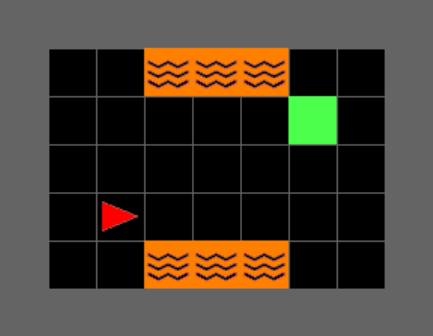} &
        \includegraphics[width=0.18\textwidth]{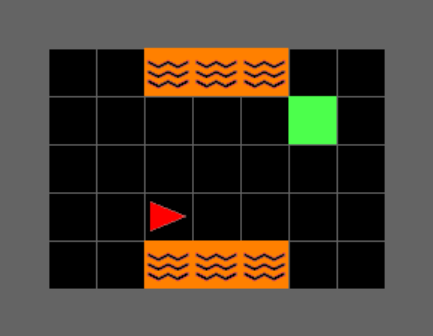} &
        \includegraphics[width=0.18\textwidth]{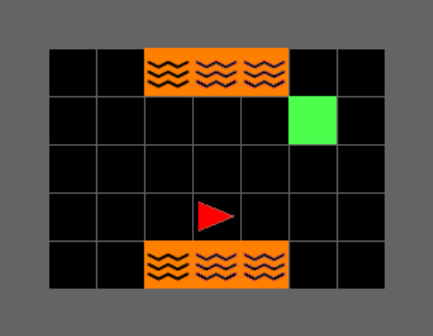} &
        \includegraphics[width=0.18\textwidth]{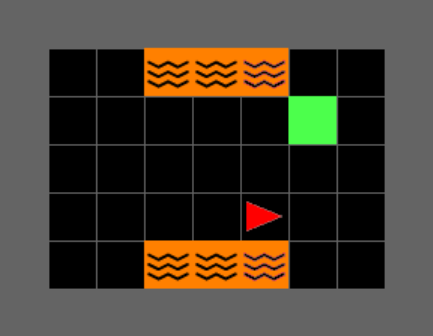} \\
        \includegraphics[width=0.18\textwidth]{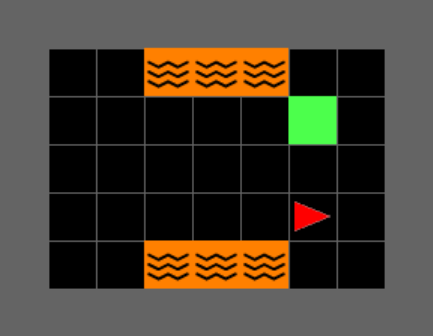} &
        \includegraphics[width=0.18\textwidth]{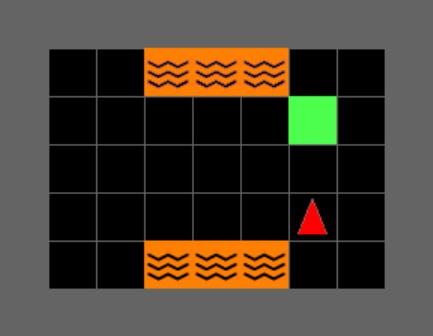} &
        \includegraphics[width=0.18\textwidth]{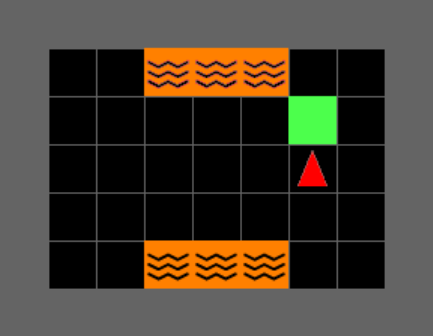} &
        \includegraphics[width=0.18\textwidth]{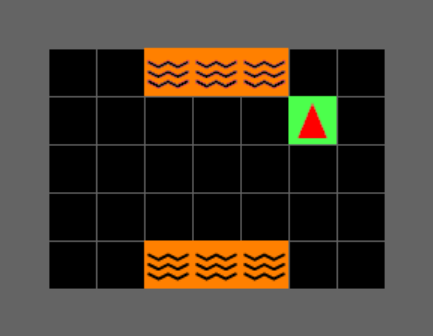} &
    \end{tabular}
    \caption[DistShift (Domain Randomization) SYMPOL (Retrained) Example 1.]{\textbf{DistShift (Domain Randomization) SYMPOL (Retrained) Example 1.} This figure visualizes the path taken by the SYMPOL agent trained on the randomized DistShift environment (see Figure~\ref{fig:sympol_distshift_random} in the main part or \texttt{tree\_function\_retrained(obs)} defined below) from left to right and top to bottom. The agent avoids the lava, identifies the goal and walks into the goal.}
    \label{fig:sympol_random_1_random}
\end{figure}

\begin{figure}[H]
    \centering
    \begin{tabular}{ccccc}
        \includegraphics[width=0.18\textwidth]{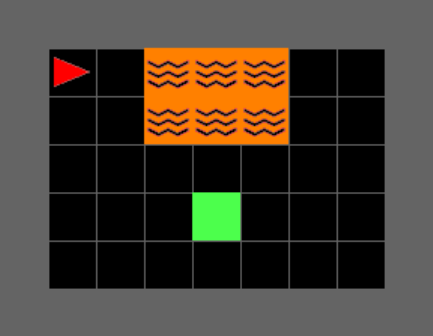} &
        \includegraphics[width=0.18\textwidth]{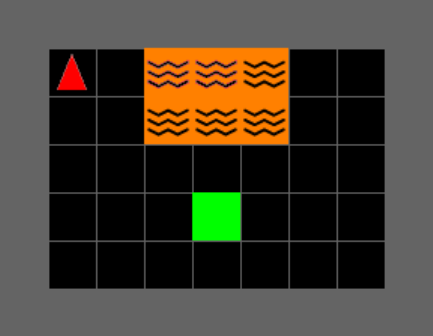} &
        \includegraphics[width=0.18\textwidth]{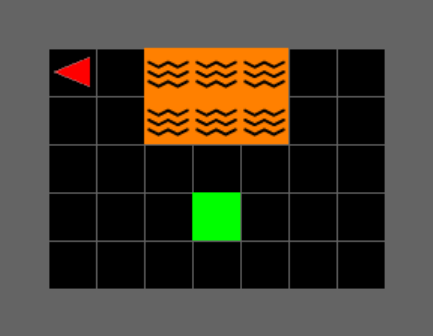} &
        \includegraphics[width=0.18\textwidth]{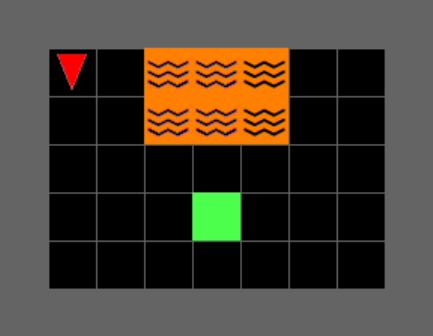} &
        \includegraphics[width=0.18\textwidth]{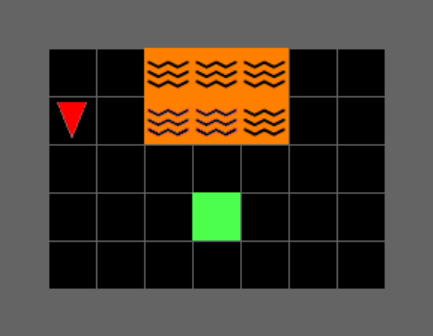} \\
        \includegraphics[width=0.18\textwidth]{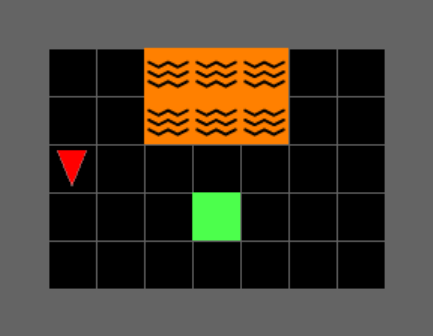} &
        \includegraphics[width=0.18\textwidth]{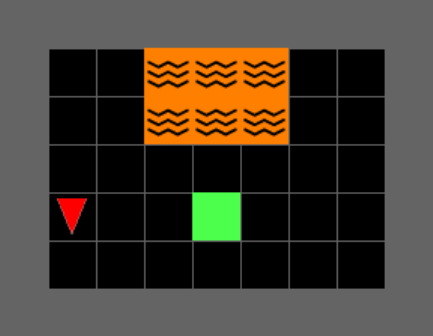} &
        \includegraphics[width=0.18\textwidth]{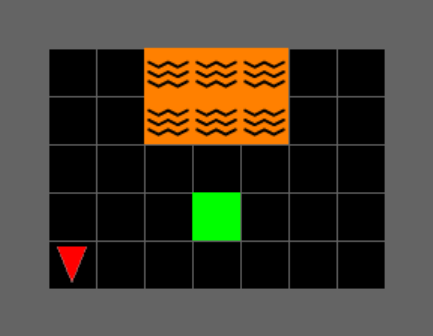} &
        \includegraphics[width=0.18\textwidth]{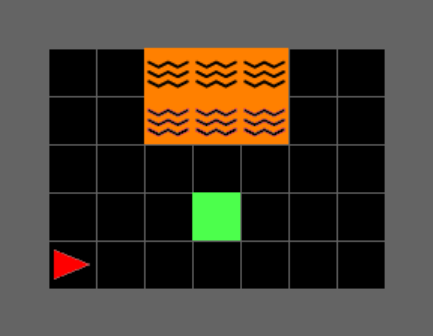} &
        \includegraphics[width=0.18\textwidth]{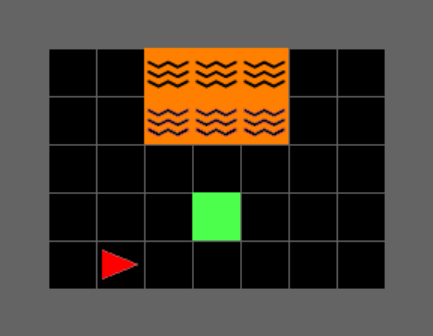} \\
        \includegraphics[width=0.18\textwidth]{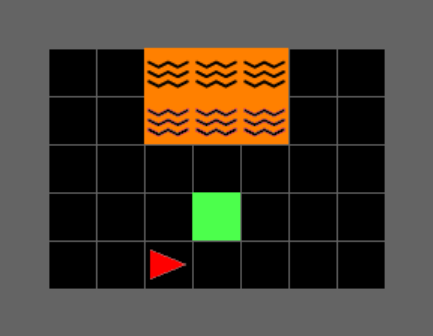} &
        \includegraphics[width=0.18\textwidth]{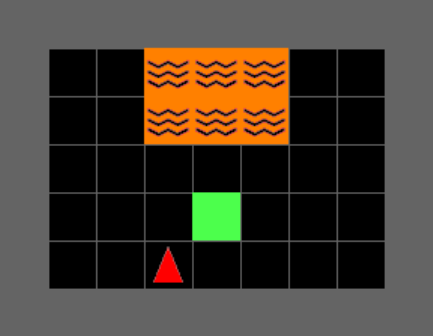} &
        \includegraphics[width=0.18\textwidth]{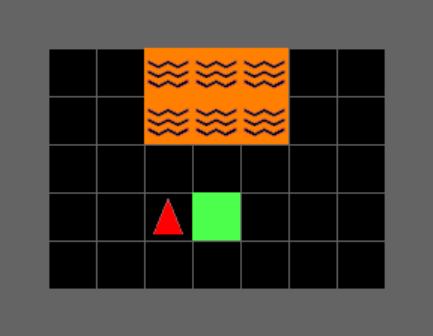} &
        \includegraphics[width=0.18\textwidth]{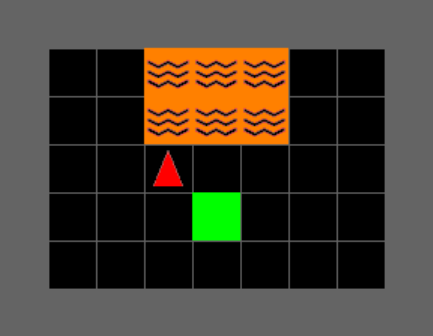} &
        \includegraphics[width=0.18\textwidth]{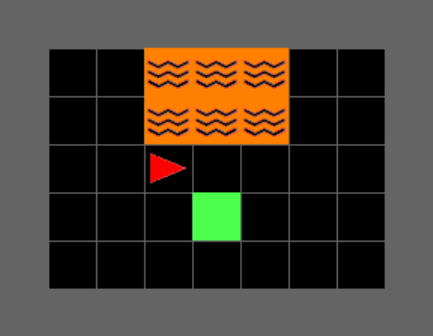} \\
        \includegraphics[width=0.18\textwidth]{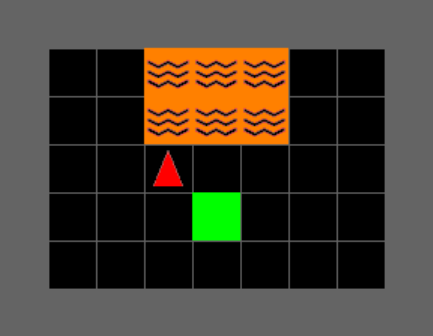} &
        \includegraphics[width=0.18\textwidth]{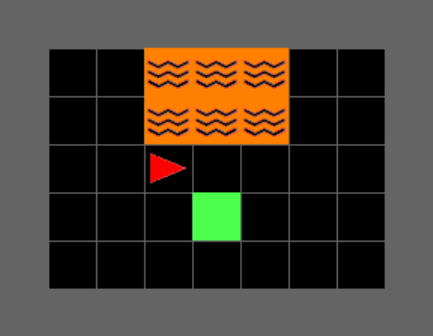} &
        \includegraphics[width=0.18\textwidth]{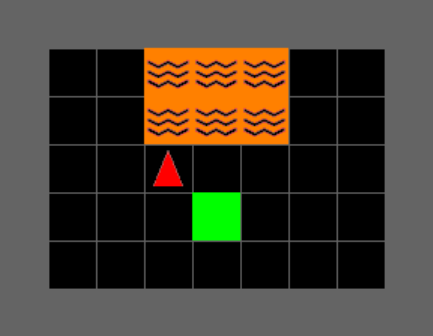} &
        \includegraphics[width=0.18\textwidth]{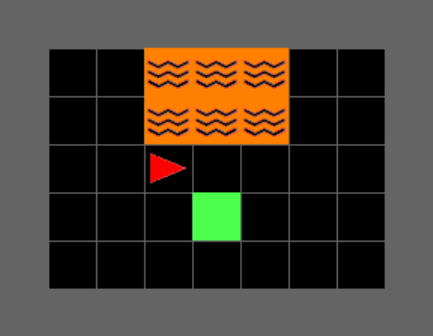} &
        \includegraphics[width=0.18\textwidth]{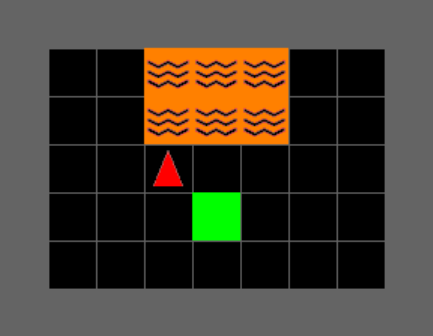} \\
    \end{tabular}
    \caption[DistShift (Domain Randomization) SYMPOL Example 2.]{\textbf{DistShift (Domain Randomization) SYMPOL Example 2.} This figure visualizes the path taken by the SYMPOL agent trained on the basic DistShift environment (see Figure~\ref{fig:sympol_distshift_random} in the main part or \texttt{tree\_function(obs)} defined below) from left to right and top to bottom. The agent follows the wall until there is no empty space on the left. Instead of an empty space there is the goal, but instead of walking into the goal, the agent surpasses it and again gets stuck at the lava.}
    \label{fig:sympol_random_2_basic}
\end{figure}

\vspace{0.5cm}

\begin{figure}[H]
    \centering
    \begin{tabular}{ccccc}
        \includegraphics[width=0.18\textwidth]{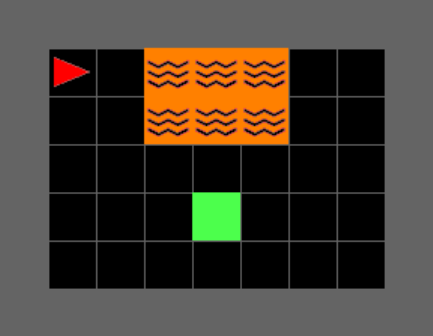} &
        \includegraphics[width=0.18\textwidth]{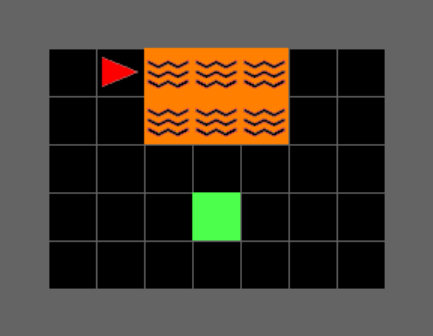} &
        \includegraphics[width=0.18\textwidth]{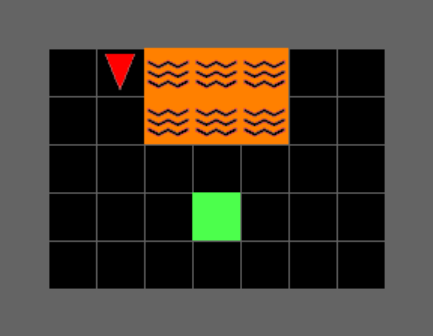} &
        \includegraphics[width=0.18\textwidth]{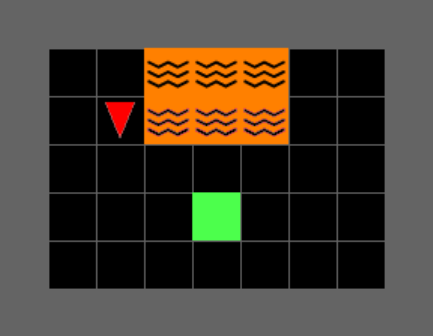} &
        \includegraphics[width=0.18\textwidth]{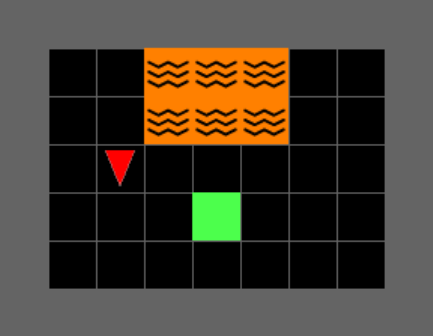} \\
        \includegraphics[width=0.18\textwidth]{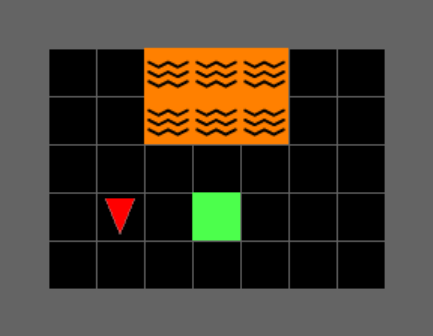} &
        \includegraphics[width=0.18\textwidth]{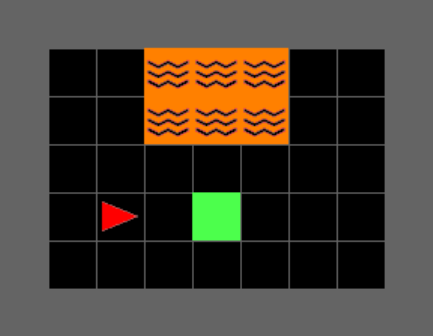} &
        \includegraphics[width=0.18\textwidth]{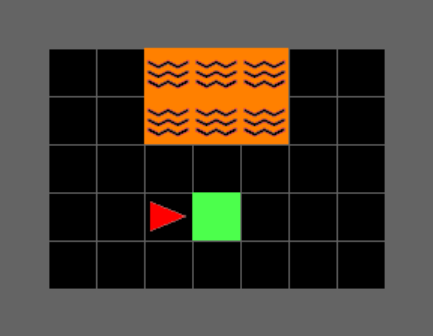} &
        \includegraphics[width=0.18\textwidth]{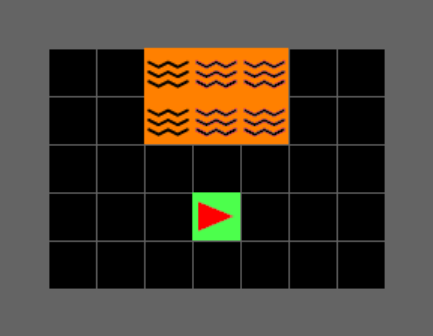} &

    \end{tabular}
    \caption[DistShift (Domain Randomization) SYMPOL (Retrained) Example 2.]{\textbf{DistShift (Domain Randomization) SYMPOL (Retrained) Example 2.} This figure visualizes the path taken by the SYMPOL agent trained on the randomized DistShift environment (see Figure~\ref{fig:sympol_distshift_random} in the main part or \texttt{tree\_function\_retrained(obs)} defined below) from left to right and top to bottom. The agent avoids the lava, identifies the goal, and walks into the goal.}
    \label{fig:sympol_random_2_random}
\end{figure}

\newpage

\subsection{SYMPOL Algorithmic Presentation}

\begin{lstlisting}
def tree_function(obs):
    if obs[field one to front and one to left] is 'empty':
        if obs[field one to front] is 'lava':
            if obs[field one to left] is 'empty':
                action = 'turn right'
            else:
                action = 'turn left'
        else:
            action = 'move forward'
    else:
        action = 'turn left'
    return action
\end{lstlisting}

\vspace{0.5cm}

\begin{lstlisting}
def tree_function_retrained(obs):
    if obs[field one to left] is 'goal':
        action = 'turn left'
    else:
        if obs[field one to right] is 'goal':
            action = 'turn right'
        else:
            if obs[field two to front] is 'wall':
                if obs[field one to front and one to right] is 'goal':
                    if obs[field one to right] is 'lava':
                        action = 'turn left'
                    else:
                        action = 'move forward'
                else:
                    if obs[field one to front] is 'goal':
                        action = 'move forward'
                    else:
                        action = 'turn left'                
            else:
                if obs[field one to front] is 'lava':
                    action = 'turn right'    
                else:
                    action = 'move forward'    
    return action
\end{lstlisting}

\newpage

\section{Methods and Hyperparameters}\label{A:hyperparameters_sympol}
The main methods we compared SYMPOL against are behavioral cloning state-action DTs (SA-DT) and discretized soft decision trees (D-SDT). In addition to the information given in Section~\ref{sec:sympol_eval}, we want to provide some more detailed results of the implementation and refer to our source code for the exact definition.
\begin{itemize}
    \item \textbf{State-Action DTs (SA-DT)} Behavioral cloning SA-DTs are the most common method to generate interpretable policies post-hoc. Hereby, we first train an MLP policy, which is then distilled into a DT as a post-processing step after the training. Specifically, we train the DT on a dataset of expert trajectories generated with the MLP policy. The number of expert trajectories was set to $25$ which we experienced as a good trade-off between dataset size for the distillation and model complexity during preliminary experiments. The $25$ expert trajectories result in a total of approximately $12500$ state-action pairs, varying based on the environment specification.
    \item \textbf{Discretized Soft Decision Trees (D-SDT)} SDTs allow gradient computation by assigning probabilities to each node. While SDTs exhibit a hierarchical structure, they are usually considered as less interpretable, since multiple features are considered in a single split and the whole tree is traversed simultaneously. Therefore, \citet{silva2020optimization} use SDTs as policies which are discretized post-hoc to allow an easy interpretation considering only a single feature at each split. Discretization is achieved by employing an argmax to obtain the feature index and normalizing the split threshold based on the feature vector.
    We improved their method by replacing the scaled sigmoid and softmax, with an entmoid and entmax transformation~\citep{peters2019sparse}, resulting in sparse feature selectors with more responsive gradients, as it is common practice~\citep{popov2019neural,node_gam}.
\end{itemize}

In the following, we list the parameter grids used during the hyperparameter optimization (HPO) as well as the optimal parameters selected for each environment. For SYMPOL, SDT and MLP, we optimized the hyperparameters based on the validation reward with optuna~\citep{akiba2019optuna} for 60 trials. Thereby, we ensured that the environments evaluated during the HPO were distinct to the environments used for reporting the test performance in the rest of the evaluation.

\subsection{HPO Grids}

\begin{table}[H]
    \centering
        \caption[HPO Grid SYMPOL]{\textbf{HPO Grid SYMPOL}}
        \begin{tabular}{lr}
        \toprule
         hyperparameter & \multicolumn{1}{l}{values}  \\
        \midrule
        learning\_rate\_actor\_weights &  [0.0001, 0.1]   \\
        learning\_rate\_actor\_split\_values &   [0.0001, 0.05]  \\
        learning\_rate\_actor\_split\_idx\_array &  [0.0001, 0.1]   \\
        learning\_rate\_actor\_leaf\_array &  [0.0001, 0.05]   \\
        learning\_rate\_actor\_log\_std & [0.0001, 0.1]    \\
        
        learning\_rate\_critic &  [0.0001, 0.01]   \\
        
        n\_update\_epochs &  [0, 10]   \\
        reduce\_lr &  \{True, False\}   \\
        
        n\_steps &   \{128, 512\}  \\
        n\_envs &  [4, 16]   \\
        
        norm\_adv &  \{True, False\}   \\
        ent\_coef &  \{0.0, 0.1, 0.2, 0.5\}   \\
        gae\_lambda &  \{0.8, 0.9, 0.95, 0.99\}   \\
        gamma &  \{0.9, 0.95, 0.99, 0.999\}   \\
        vf\_coef &  \{0.25, 0.50, 0.75\}   \\
        max\_grad\_norm &  [None]   \\
        
        SWA &   \{True\}  \\
        adamW &  \{True\}   \\
        depth &   \{7\}  \\
        minibatch\_size &  \{64\}   \\
        \bottomrule
        \end{tabular}
    \label{tab:grid_sympol}
\end{table}

\vspace{0.5cm}

\begin{table}[H]
    \centering
        \caption[HPO Grid MLP]{\textbf{HPO Grid MLP}}
\begin{tabular}{lr}
\toprule
 hyperparameter & \multicolumn{1}{l}{values}  \\
\midrule
neurons\_per\_layer & [16, 256]    \\
num\_layers &  [1, 3]   \\

learning\_rate\_actor &  [0.0001, 0.01]   \\
learning\_rate\_critic & [0.0001, 0.01]    \\
minibatch\_size &  \{64, 128, 256, 512\}   \\
n\_update\_epochs &  [1, 10]   \\

n\_steps &   \{128, 512\}  \\
n\_envs &  [4, 16]   \\

norm\_adv &  \{True, False\}   \\
ent\_coef &  \{0.0, 0.1, 0.2, 0.5\}   \\
gae\_lambda &  \{0.8, 0.9, 0.95, 0.99\}   \\
gamma &  \{0.9, 0.95, 0.99, 0.999\}   \\
vf\_coef &  \{0.25, 0.50, 0.75\}   \\
max\_grad\_norm &  \{0.1, 0.5, 1.0, None\}   \\

\bottomrule
\end{tabular}
    \label{tab:grid_mlp}
\end{table}

\begin{table}[H]
    \centering
    \caption[HPO Grid SDT]{\textbf{HPO Grid SDT}}

\begin{tabular}{lr}
\toprule
 hyperparameter & \multicolumn{1}{l}{values}  \\
\midrule
critic &   \{'MLP', 'SDT'\}  \\
depth &  [4, 8]   \\
temperature &  \{0.01, 0.05, 0.1, 0.5, 1.0\}   \\

learning\_rate\_actor &  [0.0001, 0.01]   \\
learning\_rate\_critic & [0.0001, 0.01]    \\
minibatch\_size &  \{64, 128, 256, 512\}   \\
n\_update\_epochs &  [1, 10]   \\

n\_steps &   \{128, 512\}  \\
n\_envs &  [4, 16]   \\

norm\_adv &  \{True, False\}   \\
ent\_coef &  \{0.0, 0.1, 0.2, 0.5\}   \\
gae\_lambda &  \{0.8, 0.9, 0.95, 0.99\}   \\
gamma &  \{0.9, 0.95, 0.99, 0.999\}   \\
vf\_coef &  \{0.25, 0.50, 0.75\}   \\
max\_grad\_norm &  \{0.1, 0.5, 1.0, None\}   \\

\bottomrule
\end{tabular}
    \label{tab:grid_sdt}
\end{table}

\vspace{0.5cm}

\subsection{Best Hyperparameters}

\begin{table}[H]     
\centering 
    \caption[Best Hyperparameters SYMPOL (Control)]{\textbf{Best Hyperparameters SYMPOL (Control)}}

\begin{tabular}{llllll}
\toprule
 & CP & AB & LL & MC-C & PD-C \\
\midrule
ent\_coef & 0.200 & 0.000 & 0.000 & 0.500 & 0.100 \\
gae\_lambda & 0.950 & 0.950 & 0.900 & 0.990 & 0.800 \\
gamma & 0.990 & 0.990 & 0.999 & 0.999 & 0.999 \\
learning\_rate\_actor\_weights & 0.048 & 0.003 & 0.072 & 0.000 & 0.022 \\
learning\_rate\_actor\_split\_values & 0.000 & 0.000 & 0.001 & 0.000 & 0.000 \\
learning\_rate\_actor\_split\_idx\_array & 0.026 & 0.052 & 0.010 & 0.000 & 0.010 \\
learning\_rate\_actor\_leaf\_array & 0.020 & 0.005 & 0.009 & 0.028 & 0.006 \\
learning\_rate\_actor\_log\_std & 0.001 & 0.002 & 0.021 & 0.094 & 0.000 \\
learning\_rate\_critic & 0.001 & 0.000 & 0.002 & 0.002 & 0.000 \\
max\_grad\_norm & None & None & None & None & None \\
n\_envs & 7 & 8 & 6 & 5 & 15 \\
n\_steps & 512 & 128 & 512 & 128 & 128 \\
n\_update\_epochs & 7 & 7 & 7 & 2 & 7 \\
norm\_adv & False & False & True & False & True \\
reduce\_lr & True & True & True & True & False \\
vf\_coef & 0.500 & 0.250 & 0.500 & 0.500 & 0.750 \\
SWA & True & True & True & True & True \\
adamW & True & True & True & True & True \\
dropout & 0.000 & 0.000 & 0.000 & 0.000 & 0.000 \\
depth & 7 & 7 & 7 & 7 & 7 \\
minibatch\_size & 64 & 64 & 64 & 64 & 64 \\
n\_estimators & 1 & 1 & 1 & 1 & 1 \\
\bottomrule
\end{tabular}
    \label{tab:hyperparameters_sympol}
\end{table}

\begin{table}[H]     
\centering 
    \caption[Best Hyperparameters SYMPOL (MiniGrid)]{\textbf{Best Hyperparameters SYMPOL (MiniGrid)}}

\begin{tabular}{llllll}
\toprule
 & E-R & DK & LG-5 & LG-7 & DS \\
\midrule
ent\_coef & 0.100 & 0.200 & 0.100 & 0.100 & 0.500 \\
gae\_lambda & 0.990 & 0.950 & 0.900 & 0.900 & 0.950 \\
gamma & 0.900 & 0.990 & 0.950 & 0.990 & 0.999 \\
learning\_rate\_actor\_weights & 0.063 & 0.042 & 0.055 & 0.001 & 0.036 \\
learning\_rate\_actor\_split\_values & 0.001 & 0.001 & 0.006 & 0.001 & 0.000 \\
learning\_rate\_actor\_split\_idx\_array & 0.001 & 0.001 & 0.012 & 0.001 & 0.009 \\
learning\_rate\_actor\_leaf\_array & 0.003 & 0.004 & 0.009 & 0.008 & 0.001 \\
learning\_rate\_actor\_log\_std & 0.043 & 0.021 & 0.005 & 0.002 & 0.038 \\
learning\_rate\_critic & 0.001 & 0.001 & 0.001 & 0.001 & 0.001 \\
max\_grad\_norm & None & None & None & None & None \\
n\_envs & 14 & 14 & 16 & 7 & 10 \\
n\_steps & 128 & 512 & 512 & 128 & 512 \\
n\_update\_epochs & 8 & 9 & 5 & 4 & 5 \\
norm\_adv & True & True & True & True & False \\
reduce\_lr & False & True & True & True & True \\
vf\_coef & 0.500 & 0.500 & 0.250 & 0.500 & 0.250 \\
SWA & True & True & True & True & True \\
adamW & True & True & True & True & True \\
dropout & 0.000 & 0.000 & 0.000 & 0.000 & 0.000 \\
depth & 7 & 7 & 7 & 7 & 7 \\
minibatch\_size & 64 & 64 & 64 & 64 & 64 \\
n\_estimators & 1 & 1 & 1 & 1 & 1 \\
\bottomrule
\end{tabular}
    \label{tab:hyperparameters_sympol_minigrid}
\end{table}

\vspace{0.5cm}

\begin{table}[H]     
\centering 
    \caption[Best Hyperparameters MLP (Control)]{\textbf{Best Hyperparameters MLP (Control)}}

\begin{tabular}{llllll}
\toprule
 & CP & AB & LL & MC-C & PD-C \\
\midrule
adamW & False & False & False & False & False \\
ent\_coef & 0.200 & 0.000 & 0.100 & 0.100 & 0.100 \\
gae\_lambda & 0.900 & 0.900 & 0.900 & 0.950 & 0.950 \\
gamma & 0.999 & 0.990 & 0.999 & 0.999 & 0.990 \\
learning\_rate\_actor & 0.001 & 0.000 & 0.001 & 0.005 & 0.000 \\
learning\_rate\_critic & 0.003 & 0.005 & 0.003 & 0.001 & 0.002 \\
max\_grad\_norm & 1.000 & 1.000 & 0.500 & 0.100 & None \\
minibatch\_size & 256 & 256 & 128 & 512 & 128 \\
n\_envs & 13 & 12 & 13 & 15 & 8 \\
n\_steps & 128 & 512 & 512 & 512 & 512 \\
n\_update\_epochs & 7 & 9 & 8 & 2 & 2 \\
neurons\_per\_layer & 139 & 185 & 46 & 240 & 75 \\
norm\_adv & False & True & False & True & True \\
num\_layers & 2 & 2 & 3 & 2 & 2 \\
reduce\_lr & False & False & False & False & False \\
vf\_coef & 0.250 & 0.500 & 0.500 & 0.250 & 0.250 \\
\bottomrule
\end{tabular}
    \label{tab:hyperparameters_mlp}
\end{table}

\begin{table}[H]     
\centering 
    \caption[Best Hyperparameters MLP (MiniGrid)]{\textbf{Best Hyperparameters MLP (MiniGrid)}}

\begin{tabular}{llllll}
\toprule
 & E-R & DK & LG-5 & LG-7 & DS \\
\midrule
adamW & False & False & False & False & False \\
ent\_coef & 0.100 & 0.100 & 0.100 & 0.100 & 0.100 \\
gae\_lambda & 0.950 & 0.900 & 0.950 & 0.950 & 0.990 \\
gamma & 0.990 & 0.900 & 0.990 & 0.900 & 0.990 \\
learning\_rate\_actor & 0.000 & 0.000 & 0.002 & 0.000 & 0.000 \\
learning\_rate\_critic & 0.001 & 0.000 & 0.003 & 0.001 & 0.001 \\
max\_grad\_norm & 0.100 & 0.100 & 1 & 0.500 & 0.100 \\
minibatch\_size & 64 & 256 & 128 & 512 & 256 \\
n\_envs & 13 & 8 & 8 & 12 & 10 \\
n\_steps & 512 & 256 & 512 & 128 & 128 \\
n\_update\_epochs & 5 & 7 & 9 & 8 & 7 \\
neurons\_per\_layer & 112 & 169 & 76 & 28 & 158 \\
norm\_adv & False & True & False & True & True \\
num\_layers & 3 & 1 & 1 & 1 & 2 \\
reduce\_lr & False & False & False & False & False \\
vf\_coef & 0.500 & 0.500 & 0.250 & 0.750 & 0.500 \\
\bottomrule
\end{tabular}
    \label{tab:hyperparameters_mlp_minigrid}
\end{table}

\vspace{0.5cm}

\begin{table}[H]     
\centering 
    \caption[Best Hyperparameters SDT (Control]{\textbf{Best Hyperparameters SDT (Control)}}

\begin{tabular}{llllll}
\toprule
 & CP & AB & LL & MC-C & PD-C \\
\midrule
adamW & False & False & False & False & False \\
critic & mlp & mlp & mlp & mlp & mlp \\
depth & 7 & 6 & 8 & 7 & 7 \\
ent\_coef & 0.000 & 0.100 & 0.200 & 0.000 & 0.200 \\
gae\_lambda & 0.950 & 0.950 & 0.990 & 0.900 & 0.900 \\
gamma & 0.990 & 0.990 & 0.999 & 0.990 & 0.900 \\
learning\_rate\_actor & 0.001 & 0.002 & 0.001 & 0.001 & 0.000 \\
learning\_rate\_critic & 0.000 & 0.000 & 0.001 & 0.007 & 0.000 \\
max\_grad\_norm & 0.100 & 0.100 & 1.000 & 0.500 & 0.100 \\
minibatch\_size & 128 & 128 & 128 & 64 & 128 \\
n\_envs & 15 & 6 & 7 & 14 & 7 \\
n\_steps & 512 & 128 & 512 & 512 & 256 \\
n\_update\_epochs & 4 & 10 & 2 & 1 & 7 \\
norm\_adv & True & False & True & False & False \\
reduce\_lr & False & False & False & False & False \\
temperature & 1 & 0.500 & 1 & 1 & 0.100 \\
vf\_coef & 0.500 & 0.500 & 0.750 & 0.250 & 0.500 \\
\bottomrule
\end{tabular}
    \label{tab:hyperparameters_sdt}
\end{table}

\begin{table}[H]     
\centering 
    \caption[Best Hyperparameters SDT (MiniGrid)]{\textbf{Best Hyperparameters SDT (MiniGrid)}}

\begin{tabular}{llllll}
\toprule
 & E-R & DK & LG-5 & LG-7 & DS \\
\midrule
adamW & False & False & False & False & False \\
critic & sdt & mlp & sdt & sdt & sdt \\
depth & 7 & 6 & 7 & 8 & 7 \\
ent\_coef & 0.100 & 0.100 & 0.200 & 0.100 & 0.100 \\
gae\_lambda & 0.900 & 0.950 & 0.990 & 0.950 & 0.900 \\
gamma & 0.990 & 0.900 & 0.999 & 0.950 & 0.950 \\
learning\_rate\_actor & 0.004 & 0.001 & 0.000 & 0.002 & 0.001 \\
learning\_rate\_critic & 0.000 & 0.002 & 0.000 & 0.005 & 0.002 \\
max\_grad\_norm & 0.100 & 0.100 & 0.500 & 0.100 & None \\
minibatch\_size & 512 & 256 & 512 & 256 & 512 \\
n\_envs & 10 & 10 & 10 & 13 & 5 \\
n\_steps & 512 & 256 & 256 & 128 & 512 \\
n\_update\_epochs & 5 & 10 & 8 & 4 & 7 \\
norm\_adv & True & True & True & True & True \\
reduce\_lr & False & False & False & False & False \\
temperature & 1 & 1 & 1 & 1 & 1 \\
vf\_coef & 0.750 & 0.750 & 0.750 & 0.250 & 0.750 \\
\bottomrule
\end{tabular}
    \label{tab:hyperparameters_sdt_minigrid}
\end{table}

\printbibliography[heading=bibintoc]

\end{document}